\pdfoutput=1
\documentclass[a4paper]{icthesis}
\usepackage{bookmark}

\usepackage{CJKutf8}

\begin{document}

\begin{onehalfspace}
\includepdf[pages=-]{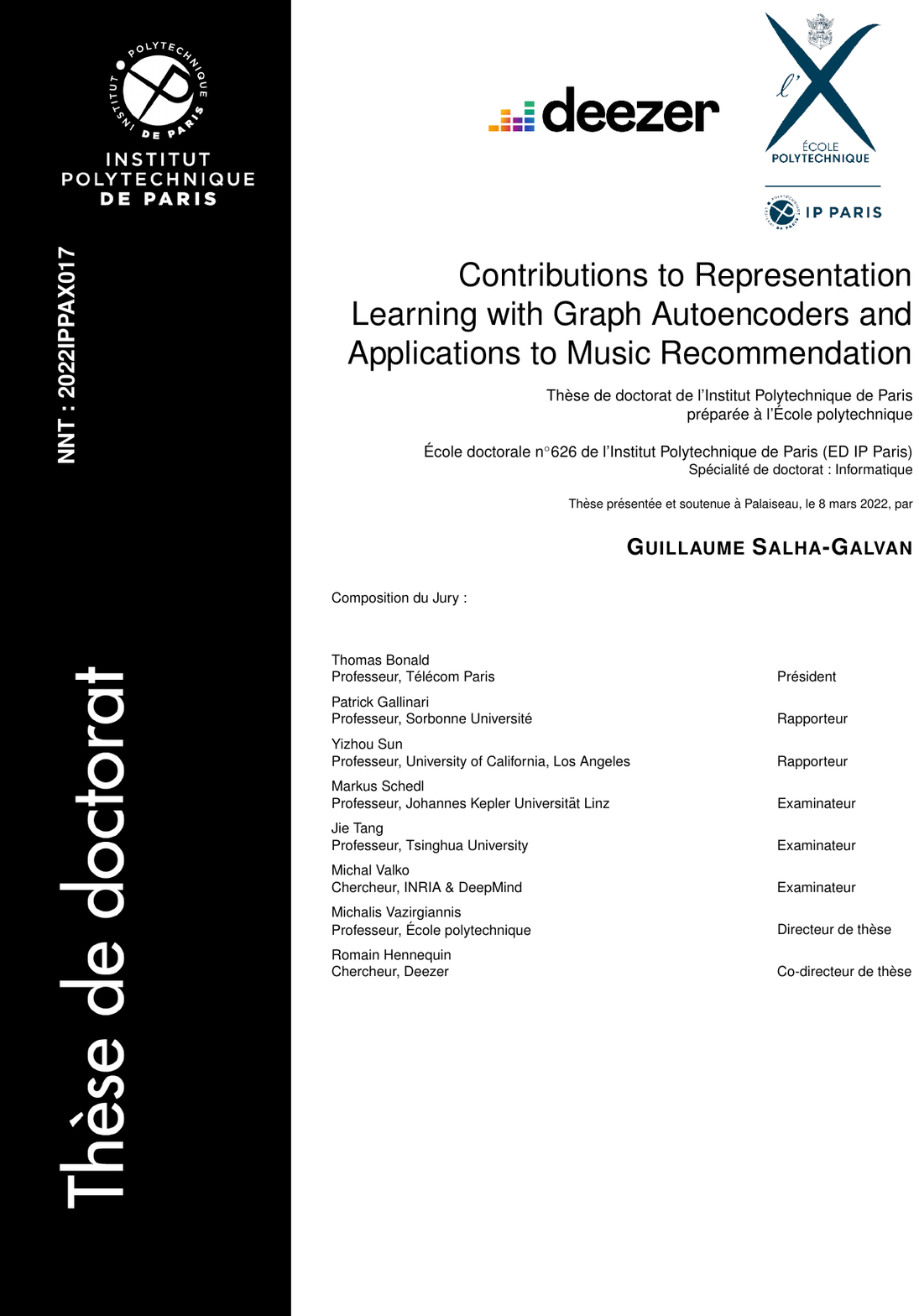}

\frontmatter

\mainmatter


\pagestyle{fancy}
\renewcommand{\headrulewidth}{0.2pt}

\renewcommand{\chaptermark}[1]{ \markboth{#1}{} }
\renewcommand{\sectionmark}[1]{ \markright{#1} }
\newcommand{\up}[1]{\textsuperscript{#1}}

\fancyhf{}
\cfoot{\thepage}

%
%
%

\fancyhead[LE]{\textbf{\nouppercase{\leftmark}} } 
\fancyhead[LO]{\textbf{\nouppercase{\leftmark}  }} 

\fancyhead[LE]{\textbf{\chaptername\ \thechapter.\ \nouppercase{\leftmark}} } 
\fancyhead[LO]{\textbf{\thesection.\ \nouppercase{\rightmark}} } 

\part{General Introduction}
\label{partI}

\chapter[Context and Scope of this Thesis]{Context and Scope of this Thesis}\label{chapter_1}
\chaptermark{Context and Scope of this Thesis}

\textit{This introductory chapter provides a general overview of this PhD thesis. Firstly, we present the context and objectives of this work, which is at the intersection of graph representation learning and music recommendation. Then, we detail the scientific contributions as well as the organization of the remainder of this thesis. We also list the publications that resulted from the research conducted during these three years.}

\section{Context and Objectives}
\label{sec11}

Graph structures became ubiquitous in various fields ranging from web mining to biology, due to the proliferation of data representing entities, also known as (a.k.a.) \textit{nodes} or \textit{vertices}, connected by links a.k.a. \textit{edges} summarizing their relations or their interactions. For instance, web graphs depict pages of the World~Wide~Web as nodes, and a node~$i$ will point to a node~$j$ via an edge if any hyperlink on page~$i$ refers to page~$j$. Social networks such as Facebook and Twitter are graphs of users, connected through ``friendship'' or ``following'' relations. Citation graphs represent scientific articles connected through citation links. Protein-protein interaction graphs efficiently summarize biological proteins and their kinetic interactions \cite{hamilton2020graph}. 

\begin{figure}[t]
    \centering
    \includegraphics[width=\textwidth]{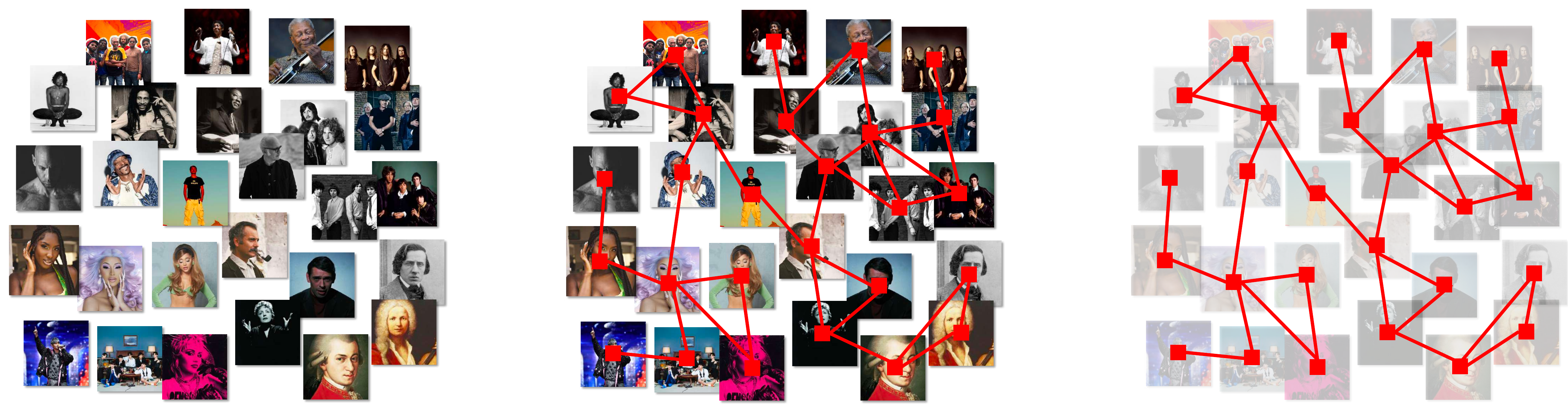}
    \caption[An example of a similar artists graph]{An example of a similar artists graph. While connections between artists are not naturally given (\textit{left}), a graph of similar artists can be constructed, e.g., by connecting artists that are simultaneously listened to or liked by numerous users (\textit{middle}). This graph can be used as an abstraction (\textit{right}), e.g., for music recommendation.}
    \label{fig:graphdeezerintro}
\end{figure}

At Deezer\footnote{\href{https://www.deezer.com/}{https://www.deezer.com/}}, where the research presented in this thesis was conducted, graphs also naturally emerge on numerous occasions. Deezer is a French music streaming service with, at the time of writing, more than 15~million active users from 180 countries. 
These users can ``follow'' each other on the service, hence creating a large social graph. They have access to a catalog of 73~million music tracks, and the musical description of this catalog also involves graph representations. Indeed, music tracks can be connected to artists, albums, music genres, or record labels, that can themselves be connected together through various semantic links (e.g., the song ``\textit{Kashmir}'' is part of ``\textit{Physical Graffiti}'', an album from the English band ``\textit{Led Zeppelin}''), generating a large knowledge graph \cite{wang2017knowledge} of musical entities. Previous studies emphasized the benefits of leveraging such representations for music information retrieval~\cite{raimond2007music}. Besides, as we will develop in Chapter~\ref{chapter_10}, Deezer possesses various graph ontologies \cite{epure2020modeling,schreiber2016genre} that represent music genres. These are graphs of conceptually related music genres, connected through various relation-specific edges (e.g., ``\textit{rap west coast}'' is a subgenre of ``\textit{hip hop}''; ``\textit{punk}'' and ``\textit{electronic music}'' are the origin~of~``\textit{synthpunk}''). 
Lastly, Deezer also constructs similarity graphs from usage data, e.g., graphs of similar artists. Contrary to the aforementioned ones, these graphs are not naturally given. As illustrated in Figure~\ref{fig:graphdeezerintro}, they are artificially built by connecting artists that are simultaneously listened to by numerous users on the service. They are subsequently processed in recommender systems and, as we will further develop in Chapters~\ref{chapter_8}~and~\ref{chapter_9}, they play a central role to help users discover new musical content on the service. The study of these similarity graphs, and of their application to industrial-level music recommendation problems, initially motivated the establishment of this PhD project.

Overall, extracting relevant information from the nodes and edges of a graph is crucial to tackle a wide range of machine learning problems~\cite{hamilton2020graph,sun2012mining,wu2019comprehensive,zhang2018network}. This includes the \textit{link prediction} task~\cite{kumar2020link,liben2007link}, which consists in inferring the presence of new or unobserved edges between some pairs of nodes, based on observed edges in the graph. This also includes \textit{community detection}~\cite{blondel2008louvain,malliaros2013clustering}, which consists in clustering nodes into similar subgroups according to a chosen similarity metric, as well as several other tasks mentioned throughout this thesis.
As an illustration, Deezer often wants to predict new connections in the similar artists graph illustrated in Figure~\ref{fig:graphdeezerintro}, corresponding to new artists pairs that users would enjoy listening to together, and which could be achieved by performing link prediction in the graph. Deezer would also like to learn clusters of similar artists, with the aim of providing usage-based recommendations (e.g., if users listen to several artists from a cluster, other unlistened artists from this same cluster could be recommended to them), which could be achieved through community detection.

\begin{figure}[H]
    \centering
    \includegraphics[width=0.5\textwidth]{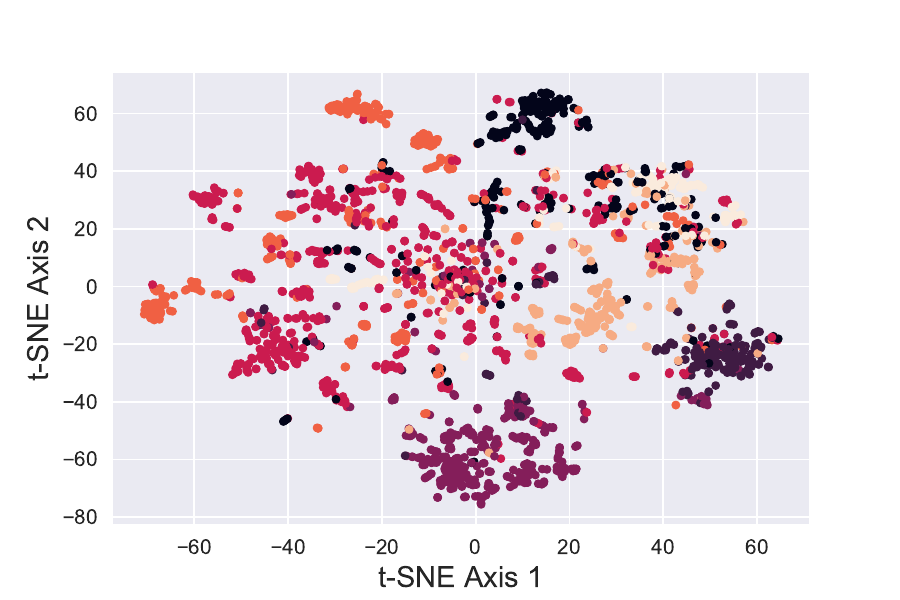}
    \caption[Visualization of node embedding representations obtained from a VGAE]{Visualization of node embedding representations for the Cora citation graph~\cite{sen2008collective} of 2 708  scientific articles connected through citations, and obtained from the VGAE detailed in Chapter~\ref{chapter_7}. Each point corresponds to an article, and colors denote their corresponding fields (not provided during training). This visualization was obtained by using the t-SNE method~\cite{van2008visualizing}, permitting visualizing 32-dimensional VGAE embedding vectors in~two~dimensions.}
    \label{fig:cora_vgae_c1}
\end{figure}

\begin{multicols}{2}
Addressing such graph-based problems has been the objective of significant research efforts over the past decades \cite{kumar2020link,leskovec2020mining,li2018influence,liben2007link,malliaros2013clustering}. Traditional approaches often focused on  hand-engineered features. For instance, locating missing edges in graphs has been historically addressed via the construction of node similarity measures, such as the popular Adamic-Adar, Jaccard or Katz indices \cite{liben2007link}. Nonetheless, as further detailed in Chapter~\ref{chapter_2}, promising improvements were recently achieved by methods aiming to directly \textit{learn} node representations \cite{hamilton2020graph,hamilton2017representation,kipf2020phd,kipf2016-1,wu2019comprehensive} summarizing the graph under consideration. As illustrated in Figure~\ref{fig:cora_vgae_c1}, these \textit{representation learning} methods compute vectorial representations of nodes  
\columnbreak

\end{multicols}
\vspace{-1.25cm}
in a \textit{node embedding space} where node positions should reflect and summarize the initial graph structure. Then, usually, they assess the probability of a missing edge between two nodes, or their likelihood of belonging to the same community, by evaluating the proximity of these nodes in the space \cite{choong2018learning,kipf2016-2,wang2017mgae}. As we will explain throughout this thesis, these methods often learn such node embedding spaces by leveraging either random walk strategies \cite{grover2016node2vec,perozzi2014deepwalk}, matrix factorization techniques \cite{cao2015grarep,ou2016asymmetric} or graph neural networks (GNNs)~\cite{hamilton2017inductive,kipf2016-1}.

In particular, \textit{graph autoencoders} (GAEs) and \textit{variational graph autoencoders} (VGAEs) \cite{kipf2016-2,tian2014learning,wang2017mgae,wang2016structural} recently emerged as two powerful families of GNN-based node embedding methods. They both rely on an encoding-decoding strategy that, in a broad sense, consists of \textit{encoding} nodes into an embedding space from which \textit{decoding}, i.e., reconstructing the original graph should ideally be possible, by
leveraging either a deterministic (for a GAE) or a probabilistic (for a VGAE) approach. The intuition behind this strategy is the following: if, starting from the embedding space, one can reconstruct a graph close to the true one, then one might conclude that embedding vectors preserve some important characteristics of the initial graph structure. Originally mainly designed for link prediction, at least in their modern formulation leveraging GNN encoders \cite{kipf2016-2}, the overall effectiveness of GAEs and VGAEs on this specific task as well as on several others has been widely experimentally confirmed over the past~few~years~(see~Chapter~\ref{chapter_2}). 

Besides their empirical performances, GAEs and VGAEs are of particular interest in the context of this PhD for two key reasons. First and foremost, they are suited for representation learning \textit{in the absence of nodes labels}, contrary to many methods that are optimized in a supervised or semi-supervised fashion and require ground truth node labels during training \cite{hamilton2017inductive,kipf2016-1,wu2019comprehensive}. This is desirable as, in most applications developed in this thesis, such labels will be unavailable. For instance, we do not have access to ground truth groups of artists/nodes that should actually be recommended together. Secondly, contrary to some popular alternatives~\cite{grover2016node2vec,perozzi2014deepwalk,von2007tutorial}, GAEs and VGAEs can process \textit{attributed graphs}, i.e., graphs in which each node is also described by its own feature vector. Once again, this is desirable as the graphs under consideration in our experiments will often be enriched by such descriptions. For instance, on a similar artists graph, artists could also be described by their own feature vectors providing some additional musical information, such as the music genres of each artist. GAEs and VGAEs could permit learning richer artist embedding representations by jointly capturing usage (through edges in the graph) and musical (through feature vectors) information on these artists. As we will emphasize in Chapter~\ref{chapter_8}, leveraging feature vectors could also permit generalizing these methods to \textit{inductive} settings~\cite{hamilton2020graph}, that require learning representations for new artists/nodes,~unseen~during~training.

Nonetheless, at the beginning of this PhD, leveraging GAEs and VGAEs for industrial-level applications, e.g., for music recommendation at Deezer, was still a challenging task.
Indeed, as we will explain in the next chapters, these models suffered from scalability issues \cite{salha2021fastgae,salha2019-1}. In 2018, in the scientific literature, experiments on GAEs and VGAEs were limited to graphs with at most a few thousand nodes and edges. The question of how to effectively scale them to larger graphs, such as those available at Deezer whose catalog includes millions of artists, albums, or music tracks, remained widely open. Besides, these models were originally designed for undirected and static graphs \cite{kipf2016-2}, while real-world graphs can also be directed (i.e., relations between nodes can be asymmetric) and/or dynamic (i.e., the graph structure can evolve over time). Moreover, some recent studies emphasized the relative limitations of standard GAEs and VGAEs for community detection \cite{choong2018learning,choong2020optimizing}. Lastly, while existing variants of these models often rely on complex neural encoders \cite{salha2019keep,salha2020simple}, their actual benefit with respect to (w.r.t.) simpler strategies, that might be preferred in production environments, had never been~thoroughly~analyzed.

\textit{The objective of this PhD thesis is twofold. Firstly, we wish to address these limitations, with the general aim of improving GAEs and VGAEs, and of facilitating their application to real-world problems involving node embedding representations. Secondly, we aim to test and evaluate our proposed approaches on various industrial graphs extracted from the Deezer service. As detailed in Section~\ref{sec12}, we will put the emphasis on graph-based music recommendation 
problems, often formulated as link prediction or community detection tasks.}

\section{Overview of Scientific Contributions}
\label{sec12}

We now provide an outline of the remainder of this thesis, while simultaneously presenting our scientific contributions. First of all, in Chapter~\ref{chapter_2} from this Part~\ref{partI}, we will review some key concepts related to machine learning on graphs. We will particularly focus on link prediction and community detection, which will be the two main evaluation tasks considered in experiments throughout the next chapters. We will also provide an introduction to graph representation learning, focusing on \textit{node} embedding methods, as well as an introduction to (variational) graph autoencoders. Then, in Part~\ref{partII}, corresponding to Chapters~\ref{chapter_3}~to~\ref{chapter_7}, we will detail our contributions to improve representation learning with GAEs and VGAEs. Finally, in Part~\ref{partIII}, corresponding to Chapters~\ref{chapter_8}~to~\ref{chapter_12}, we will present several applications related to music recommendation~at~Deezer.

\subsection{Representation Learning with Graph Autoencoders}

The Part~\ref{partII} of this thesis, focusing on contributions to node representation learning with graph autoencoders, will begin with two chapters fully dedicated to \textit{scalability} concerns. Firstly, in Chapter~\ref{chapter_3}, we will present the general framework proposed at the beginning of this PhD to scale GAEs and VGAEs to large graphs with millions of nodes and edges. This framework leverages \textit{graph degeneracy} concepts \cite{malliaros2019} to train models only from a dense subset of nodes instead of using the entire graph. Together with a simple yet effective propagation mechanism, this approach improves scalability and training speed while, to some extent, preserving performance on downstream tasks such as link prediction and community detection. We will report an evaluation of the framework on several variants of existing GAEs and VGAEs. At the time of publication of the paper associated with this work, these experiments provided, to the best of our knowledge, the first application of these models to graphs with millions of nodes and edges.

In Chapter~\ref{chapter_4}, we will subsequently introduce FastGAE, an alternative stochastic strategy to scale GAEs and VGAEs. FastGAE leverages graph mining-based sampling schemes and an effective subgraph decoding strategy to significantly lower the computational complexity of graph autoencoders, while preserving or even slightly improving their performances. We will propose an in-depth theoretical and experimental analysis of the proposed solution, showing that it behaves favorably when compared to the degeneracy framework from Chapter~\ref{chapter_3}. FastGAE constitutes a faster and simpler improvement of our previous efforts.

Then, in Chapter~\ref{chapter_5}, we will extend GAEs and VGAEs to \textit{directed} graphs. As previously mentioned, their original versions were designed for undirected graphs \cite{kipf2016-2}. They ignore the potential direction of the link when decoding edges from node embedding spaces. As we will argue, this is limiting for numerous real-life applications. In this chapter, we will present a method to effectively learn node embedding spaces from directed graphs using the GAE and VGAE paradigms. We will draw inspiration from physics to introduce a new \textit{gravity-inspired decoder} scheme, able to reconstruct asymmetric relations from node embedding spaces. We will achieve competitive empirical results on three different directed link prediction tasks, for which standard GAEs and VGAEs perform poorly.

In Chapter~\ref{chapter_6}, we will propose to \textit{simplify} GAEs and VGAEs. 
While most existing variants of these models rely on multi-layer graph convolutional network (GCN) encoders to learn node embedding vectors, or on more sophisticated neural architectures \cite{salha2019keep,salha2020simple}, we will show that these encoders are actually unnecessarily complex for many applications. We will propose to replace them by simpler linear models w.r.t. the direct neighborhood (one-hop) adjacency matrix of the graph, involving fewer operations, fewer parameters, and no activation function. In various experiments, we will show that this simpler approach consistently reaches competitive performances w.r.t. GCN-based GAEs and VGAEs for numerous real-world graphs. This includes the Cora, Citeseer and Pubmed benchmark datasets \cite{kipf2016-2,sen2008collective} commonly used to evaluate GAEs and VGAEs in the literature. Based on these results, we will question the relevance of repeatedly using these same datasets to compare complex graph autoencoders. 

Lastly, in Chapter~\ref{chapter_7}, we will focus on \textit{community detection}. While GAEs and VGAEs emerged as powerful methods for link prediction, their performances were sometimes less impressive on community detection problems where, according to recent and concurring studies~\cite{choong2018learning,choong2020optimizing,salha2021fastgae,salha2019-1}, they can be outperformed by simpler alternatives such as the Louvain method \cite{blondel2008louvain}. At the beginning of this PhD project, it was still unclear to which extent one could improve community detection with a GAE or a VGAE, especially in the absence of node features. It was moreover uncertain whether one could do so while simultaneously preserving good performances on link prediction. In this chapter, we will show that jointly addressing these two tasks with high accuracy is possible. For this purpose, we will introduce and theoretically study a community-preserving message passing scheme, doping our GAE and VGAE encoders by considering both the initial graph structure and modularity-based prior communities when computing embedding spaces. We will also propose novel training and optimization strategies, including the introduction of a modularity-inspired regularizer complementing the existing reconstruction losses for joint link prediction and community detection. We will demonstrate the effectiveness of our approach, referred to as \textit{Modularity-Aware} GAE and VGAE, through in-depth experiments.

\subsection{Applications to Music Recommendation at Deezer}

\begin{figure}[t]
    \centering
    \includegraphics[width=0.74\textwidth]{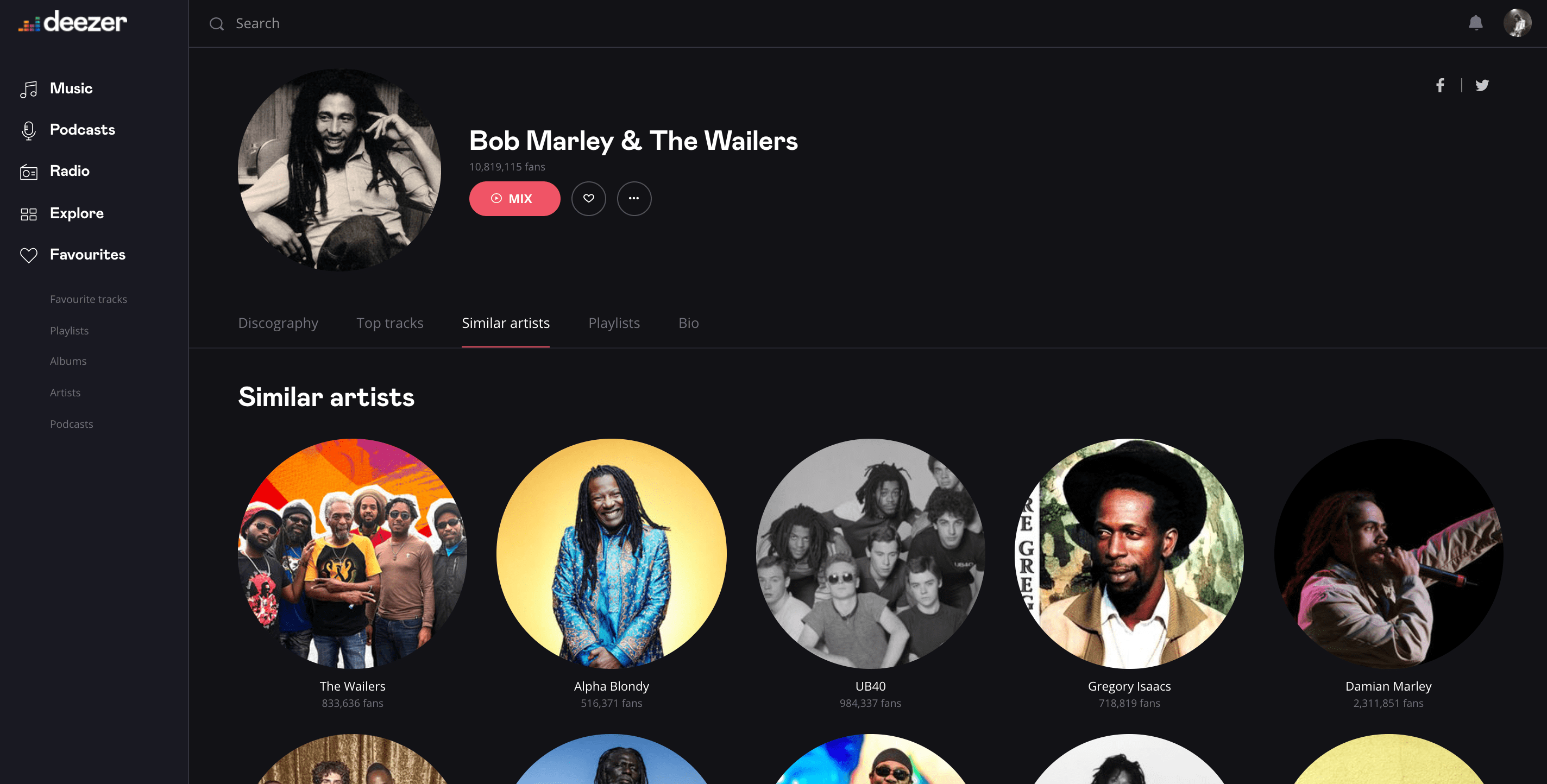}
    \caption[Recommending similar artists on Deezer]{Some examples of ``similar artists'' recommended on the website version of Deezer.}
    \label{fig:deezer_similar_c1}
\end{figure}

The Part~\ref{partIII} of this thesis will provide five additional chapters, focusing on applications to music recommendation. Firstly, in Chapter~\ref{chapter_8}, we will adopt a graph-based approach to rank similar artists on Deezer. As illustrated in Figure~\ref{fig:deezer_similar_c1}, on an artist's profile page, music streaming services such as Deezer frequently recommend a ranked list of similar artists that fans also liked.
However, implementing such a feature is challenging for new artists, for which usage data on the service (e.g., streams or likes) is not yet available.
In this chapter, we will model this ``\textit{cold start}'' \textit{similar artists ranking} problem as a link prediction task in a directed and attributed graph, connecting artists to their most similar neighbors and incorporating side musical information as attribute vectors.
Then, we will leverage the directed GAE/VGAE from Chapter~\ref{chapter_5} to learn node embedding representations from this graph, and to automatically rank the most similar neighbors of new artists using the gravity-inspired asymmetric decoder. We will empirically show the flexibility and effectiveness of this method on data extracted from Deezer's~production~system.

In Chapter~\ref{chapter_9}, we will give a broader overview of how Deezer has historically leveraged more general \textit{similarity graphs} for music recommendation. While some technical details will be omitted for confidentiality reasons, we will explain how such graphs played a central role in production-facing algorithms such as Deezer's ``\textit{Flow}'' feature, notably to detect communities of similar artists to recommend to users. In this direction, we will show the benefits of adopting GAEs and VGAEs, integrating advances from previous chapters, to improve community detection in large graphs of artists and albums with recommendation purposes. Moreover, we will mention future plans to extend the approach to other graphs such as graphs of music tracks, of users, or more general knowledge graphs, as well as to culture-specific graphs. Lastly, we will discuss experimental studies, done with a Master's intern at Deezer, to extend the proposed methods to \textit{dynamic} graphs, incorporating new nodes and edges over time.

In Chapter~\ref{chapter_10}, we will then focus on music genre \textit{ontologies}. As explained in Section~\ref{sec11}, these ontologies are graphs of conceptually related music genres, connected through various relation-specific edges  \cite{schreiber2016genre}. In this chapter, we will leverage these graphs to model the music genre \textit{perception} across (language-bound) cultures. The problem is the following: the music genre perception expressed through human annotations of artists or albums varies significantly across cultures \cite{Ferwerda2016InvestigatingTR}.
These variations cannot be modeled as mere translations since we also need to account for cultural differences in the perception.
This is an important issue for music streaming services such as Deezer, as these variations impact a wide range of tasks, ranging from localized playlist captioning to genre-driven music recommendation. In this work, we will study the feasibility of obtaining relevant cross-lingual, culture-specific music genre annotations based only on language-specific semantic representations, namely on 1) word embeddings of music genres and 2) graph ontologies~\cite{epure2020modeling}. 
Our study, focused on six languages, will show that unsupervised cross-lingual music genre annotation is feasible with high accuracy when combining both types of representations. 
This combination will concurrently be performed through a GAE/VGAE approach, and through an alternative simple yet effective method based~on~retrofitting~\cite{faruqui2015}.

Finally, during the three years of this PhD, several other projects less related to GAEs and VGAEs were also carried out jointly with colleagues from Polytechnique or Deezer, sometimes leading to scientific publications \cite{bendada2020carousel,briand2021semi,epure2020multilingual,tran2021hierarchical}. For consistency reasons, we omit most of these additional projects in this thesis. We nonetheless chose to present two of them in the last Chapters~\ref{chapter_11}~and~\ref{chapter_12}, for two reasons. Foremost, these two works directly relate to music recommendation at Deezer, and consider some issues previously introduced in the thesis such as the cold start problem. Besides, these two projects led, not only to experiments on Deezer's data, but also to online A/B tests and subsequently to model deployment on the Deezer service. They will therefore complement the previous chapters by providing a larger overview of some of Deezer's strategies and production-facing algorithms to recommend music.

\begin{figure}[H]
    \centering
    \includegraphics[width=0.48\textwidth]{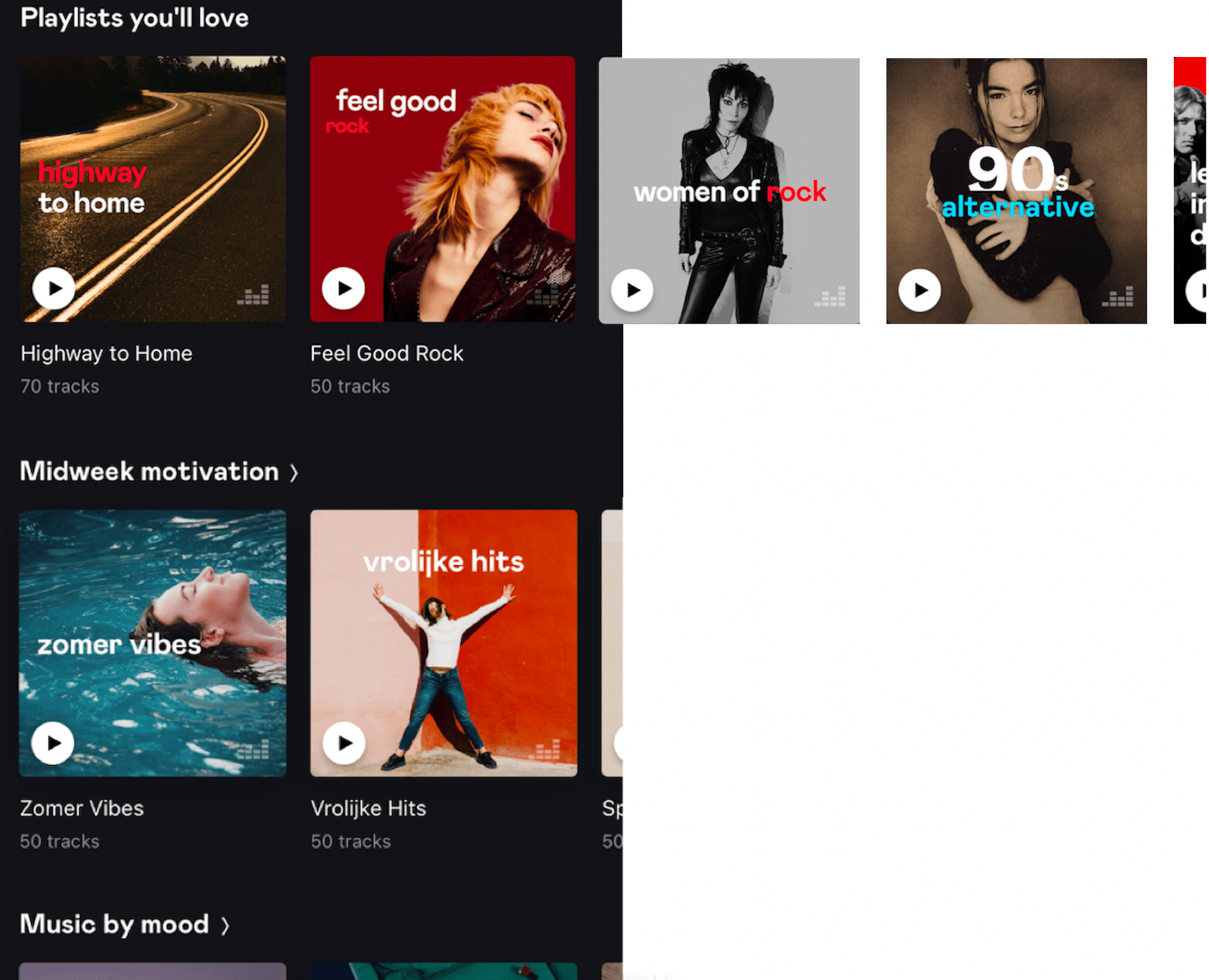}
    \caption[Personalized carousels on Deezer]{An example of a personalized swipeable carousel on the Deezer mobile app. These carousels, also referred to as sliders or
shelves \cite{mcinerney2018explore}, consist of ranked lists of items or cards to recommend to users, e.g., playlists’ cover images as in this figure. These playlists were created by professional curators from Deezer with the purpose of complying with a specific music genre, cultural area, or mood. A few playlists are initially displayed to each user, who can click on them or swipe on the screen to see some of the additional recommended playlists from the carousel.}
    \label{fig:carousel_c1}
\end{figure}

\begin{multicols}{2}
Specifically, in Chapter~\ref{chapter_11}, we will present some research on \textit{carousel} personalization. As illustrated in Figure~\ref{fig:carousel_c1}, music streaming services frequently leverage swipeable carousels, i.e., ranked lists of items or cards (albums, artists, playlists...), to recommend personalized content to users on the homepage. However, selecting the most relevant items to display in these carousels is a challenging task, as items are numerous and users have different preferences. In this chapter, we will model carousel personalization as a contextual multi-armed bandit problem with multiple plays, cascade-based updates, and delayed batch feedback. We will empirically show the effectiveness of our framework at capturing characteristics of real-world carousels by addressing a large-scale playlist recommendation task on Deezer, through offline and online experiments. 

\columnbreak
\end{multicols}
\vspace{-0.25cm}
Then, in Chapter~\ref{chapter_12}, we will present the system recently deployed on Deezer to address the \textit{user} cold start problem, and thus to recommend music to new users with few to no interactions with the catalog. The solution leverages a semi-personalized recommendation strategy, based on a deep neural network architecture and on a clustering of users from heterogeneous sources of information. We will show the practical impact of this system and its effectiveness at predicting the future musical preferences of cold start users on Deezer, through both offline and online experiments. These experiments will include tests on carousels from Chapter~\ref{chapter_11}. We will also emphasize how this system enables us to provide more interpretable recommendations.

\clearpage

\section{List of Publications}
\label{c1s13}

The research conducted during this PhD led to the publication of several scientific articles, that are listed thereafter. For the sake of simplicity, this list only mentions my current name G.~Salha-Galvan. Nonetheless, I actually published under my old name G.~Salha until 2020, and older publications are reported in their original form in the bibliography. In the remainder of this thesis, when a chapter presents some published results, the corresponding publication(s) will be specified at the beginning of the chapter.

\paragraph{Journals} $ $

\begin{itemize}
\item \underline{G. Salha-Galvan}, R. Hennequin, J. B. Remy, M. Moussallam, M. Vazirgiannis, \textit{FastGAE: Scalable Graph Autoencoders with Stochastic Subgraph Decoding}. \textbf{Neural~Networks~142 (2021)},~1~--~19, Elsevier (impact factor: 8.05).

\item  \underline{G. Salha-Galvan}, J. F. Lutzeyer, G. Dasoulas, R. Hennequin, M. Vazirgiannis,
\textit{Modularity-Aware Graph Autoencoders for Joint Community Detection and Link Prediction}.
This article is currently under review for publication in \textbf{Neural Networks},~Elsevier in 2022.
\end{itemize}

\paragraph{International Conferences} $ $

\begin{itemize}
\item \underline{G. Salha-Galvan}, R. Hennequin, V. A. Tran, M. Vazirgiannis,
\textit{A Degeneracy Framework for Scalable Graph Autoencoders}.
Proceedings of the 28\up{th} International Joint Conference on Artificial Intelligence~(\textbf{IJCAI~2019}), 3353 -- 3359.

\item \underline{G. Salha-Galvan}, S. Limnios, R. Hennequin, V. A. Tran, M. Vazirgiannis,
\textit{Gravity-Inspired Graph Autoencoders for Directed Link Prediction}.
Proceedings of the 28\up{th} ACM International Conference on Information and Knowledge Management (\textbf{CIKM 2019}),~589~--~598.

\item \underline{G. Salha-Galvan}, R. Hennequin, M. Vazirgiannis,
\textit{Simple and Effective Graph Autoencoders with One-Hop Linear Models}. Proceedings of the 2020 European Conference on Machine Learning and Principles and Practice of Knowledge Discovery in Databases (\textbf{ECML~-~PKDD~2020}), 319~--~334.

\item E. V. Epure, \underline{G. Salha-Galvan}, M. Moussallam, R. Hennequin,
\textit{Modeling the Music Genre Perception across Language-Bound Cultures}.
Proceedings of the 2020 Conference on Empirical Methods in Natural Language Processing (\textbf{EMNLP~2020}), 4765 -- 4779.

\item E. V. Epure, \underline{G. Salha-Galvan}, R. Hennequin
\textit{Multilingual Music Genre Embeddings for Effective Cross-Lingual Music Item Annotation}.
Proceedings of the 21\up{st} International Society for Music Information Retrieval Conference (\textbf{ISMIR 2020}), 803 -- 810.

\item W. Bendada, \underline{G. Salha-Galvan}, T. Bontempelli,
\textit{Carousel Personalization in Music Streaming Apps with Contextual Bandits}.
Proceedings of the 14\up{th} ACM Conference on Recommender Systems (\textbf{RecSys 2020}), 420 -- 425. \textcolor{ImperialColor}{Best short paper honorable mention}.

\item \underline{G. Salha-Galvan}, R. Hennequin, B. Chapus, V. A. Tran, M. Vazirgiannis, \textit{Cold Start Similar Artists Ranking with Gravity-Inspired Graph Autoencoders}.
Proceedings of the 15\up{th} ACM Conference on Recommender Systems (\textbf{RecSys 2021}), 443 -- 452. \textcolor{ImperialColor}{Best student paper honorable mention}. 

\item V. A. Tran, \underline{G. Salha-Galvan}, R. Hennequin, M. Moussallam,
\textit{Hierarchical Latent Relation Modeling for Collaborative Metric Learning}.
Proceedings of the 15\up{th} ACM Conference on Recommender Systems (\textbf{RecSys 2021}), 302 -- 309.

\item L. Briand, \underline{G. Salha-Galvan}, W. Bendada, M. Morlon, V. A. Tran,
\textit{A Semi-Personalized System for User Cold Start Recommendation on Music Streaming Apps}.
Proceedings of the 27\up{th} ACM SIGKDD Conference on Knowledge Discovery and Data Mining (\textbf{KDD~2021}), 2601~--~2609.

\end{itemize}

\paragraph{Workshops and Demonstrations} $ $

\begin{itemize}
\item \underline{G. Salha-Galvan}, R. Hennequin, M. Vazirgiannis,
\textit{Keep It Simple: Graph Autoencoders Without Graph Convolutional Networks}.
Workshop on Graph Representation Learning,
33\up{rd} Conference on Neural Information Processing Systems (\textbf{NeurIPS 2019}).

\item E. V. Epure, \underline{G. Salha-Galvan}, F. Voituret, M. Baranes, R. Hennequin, \textit{Muzeeglot : Annotation Multilingue et Multi-Sources d’Entités Musicales à partir de Représentations de Genres Musicaux}. Actes de la 27\up{ème} Conférence sur le Traitement Automatique des Langues Naturelles - Démonstrations (\textbf{TALN 2020}), 18 -- 21.

\end{itemize}

\paragraph{Code and Data Releases} $ $

\begin{itemize}
    \item Along with each of these articles, we publicly released our source code on a GitHub repository, for reproducibility and future usage of the proposed methods. Along with five of these articles, we also publicly released new datasets, that were either directly extracted from Deezer's private resources, or scraped and processed from the internet. More details are provided throughout this thesis, and on: \href{https://github.com/deezer}{https://github.com/deezer}.
    \end{itemize}


\chapter[Background on Graph Representation Learning]{Background on Graph Representation Learning}\label{chapter_2}
\chaptermark{Background on Graph Representation Learning}

\textit{This chapter formally introduces some key concepts related to graph structures and graph representation learning, mainly focusing on methods learning node embedding representations including graph neural networks and graph autoencoders. It does not aim to provide an exhaustive review of the fast-growing graph representation learning field (which could be the objective of an entire book~\cite{hamilton2020graph}), but rather to present the important notions that will subsequently be used in the remainder of this PhD thesis.}

\section{Introduction}
\label{c2s21}

We begin this chapter with some general concepts and definitions related to graph structures.

\subsection{Graph Structures: Generalities and Key Definitions}

In its most general formulation, a \textit{graph} is defined as follows.

\begin{definition}
A \textit{graph} is an ordered pair $\mathcal{G} = (\mathcal{V}, \mathcal{E})$ comprising:
\begin{itemize}
    \item a set of \textit{nodes} or \textit{vertices} $\mathcal{V}$. For the sake of simplicity, we assume $\mathcal{V} = \{1,2,\dots,|\mathcal{V}|\}$;
    \item a set of \textit{edges} or \textit{links} $\mathcal{E} \subseteq \mathcal{V} \times \mathcal{V}$ connecting these
nodes. In essence, they summarize the relations or interactions between nodes. We denote an edge going from a node $i \in \mathcal{V}$ to a node $j \in \mathcal{V}$ as $(i, j) \in \mathcal{E}$. Each edge is equipped with a positive \textit{weight} $w_{ij}$, normalized to lie in the $[0,1]$ set without loss of generality (w.l.o.g.), and indicating the intensity of the link from $i$ to $j$. For \textit{unweighted} graphs, we implicitly set $w_{ij} = 1$~for~all~$(i,j)~\in~\mathcal{E}$.
\end{itemize}
\label{defgraph}
\end{definition}

In this thesis, we denote by $n$ the number of the nodes in the graph, i.e., $n = |\mathcal{V}|$, and by $m$ the number of edges in the graph, i.e., $m = |\mathcal{E}|$. 

\paragraph{Remark: graph or network?}
In this remark, we discuss our choice to refer to the mathematical object defined in Definition~\ref{defgraph} as a ``\textit{graph}'' and not as a ``\textit{network}'', contrary to several references \cite{blondel2008louvain,borgatti2009network,Lutzeyer2020,Newman8577,sen2008collective}. Some research studies explicitly present the two terms as synonyms \cite{adams2020gathering,chen2012discovering}. Hamilton~\cite{hamilton2020graph} makes a distinction between their usage. In his book, ``\textit{graph}'' refers to the abstract data structure on which we focus on in this thesis, while ``\textit{network}'' describes real-world instances of this structure, e.g., social networks or computer networks. He also explains that ``\textit{network}'' has been historically favored in the data mining and network science communities (studying such real-world data), while  ``\textit{graph}'' is prevalent in the graph theory (studying theoretical properties of the mathematical abstraction) and in the machine learning communities. We choose to employ ``\textit{graph}'' as well in this thesis for consistency with recent advances in 
machine learning, but also to avoid terminological clashes with the term ``\textit{neural network}'', which will be widely used in our work.

In this thesis, we will often represent a graph $\mathcal{G}$ by its \textit{adjacency matrix}, defined as follows. 
 \begin{definition}
The \textit{adjacency matrix} $A \in [0,1]^{n \times n}$ of a graph $\mathcal{G} = (\mathcal{V}, \mathcal{E})$ is defined as:
  \begin{equation}
    A_{ij} =
    \begin{cases}
      w_{ij}, & \text{if}\ (i, j) \in \mathcal{E}, \\
      0, & \text{otherwise,}
    \end{cases}
  \end{equation}
for all $(i, j) \in \{1,\dots,n\}^2$.
\label{def:adj}
 \end{definition}

\begin{figure*}[t]
\centering
\resizebox{\textwidth}{!}{
  \subfigure[Undirected]{
  \begin{tikzpicture}[bn/.style={circle,fill=ImperialColor,text=white,minimum size=9mm},every node/.append style={bn}]
 \path node (4) {4} --  (2,2) node (5) {5} -- (0,2) node (3) {3}
 -- (-2,2) node (1) {1} -- (0,4) node (2) {2}
 -- (2,4) node (6) {6} ;
 \draw[thick] 
 foreach \X [count=\Y] in {2,...,6} {(\Y) -> (\X)}
 (2) -- (5) (5) -- (2) (3) -- (2) (4) -- (1) (4) -- (3) (5) -- (6) (1) -- (3) (3) -- (5) (2) -- (6);
\end{tikzpicture}}
  \subfigure[Directed]{
  \begin{tikzpicture}[bn/.style={circle,fill=ImperialColor,text=white,minimum
size=9mm},every node/.append style={bn}]
 \path node (4) {4} --  (2,2) node (5) {5} -- (0,2) node (3) {3}
 -- (-2,2) node (1) {1} -- (0,4) node (2) {2}
 -- (2,4) node (6) {6} ;
 \draw[->,thick]
  (1) -- (3);
  \draw[->,thick]
  (1) -- (4) ;
  \draw[->,thick]
(5) -- (3);
  \draw[->,thick]
  (2) -- (5) ;
  \draw[->,thick] (6) -- (2) ;
    \draw[->,thick] (4) -- (3);
    \draw[->,thick] (3) -- (2);
    \draw[->,thick] (2) -- (1);
    \draw[->,thick] (4) -- (5);
    \draw[->,thick] (5) -- (6);
\end{tikzpicture}}
    \subfigure[Weighted]{
  \begin{tikzpicture}[bn/.style={circle,fill=ImperialColor,text=white,minimum
size=9mm},every node/.append style={bn}]
 \path node (4) {4} --  (2,2) node (5) {5} -- (0,2) node (3) {3}
 -- (-2,2) node (1) {1} -- (0,4) node (2) {2}
 -- (2,4) node (6) {6} ;
  \begin{scope}[every edge/.style={draw=black, thick}, every node/.style={draw=black, thick,fill=white}]
    \draw  (1) edge node{\tiny{0.5}} (2);
    \draw  (1) edge node{\tiny{0.7}} (3);
    \draw  (1) edge node{\tiny{0.8}} (4);
    \draw  (2) edge node{\tiny{0.2}} (3);
    \draw  (2) edge node{\tiny{0.1}} (5);
    \draw  (3) edge node{\tiny{0.9}} (4);
    \draw  (2) edge node{\tiny{0.2}} (6);
    \draw  (3) edge node{\tiny{0.1}} (5);
    \draw  (4) edge node{\tiny{0.5}} (5);
     \draw  (5) edge node{\tiny{0.9}} (6);
   \end{scope}
\end{tikzpicture}}
      \subfigure[Bipartite]{
  \begin{tikzpicture}[bn/.style={circle,fill=ImperialColor,text=white,minimum
size=9mm},every node/.append style={bn}]
\node (1) at (0,0) {1};
\node (2) at (0,-1) {2};
\node (3) at (0,-2) {3};
\node (4) at (3,1) {4};
\node (5) at (3,0) {5};
\node (6) at (3,-1) {6};
\node (7) at (3,-2) {7};
\node (8) at (3,-3) {8};
 \draw[thick]
  (1) -- (4);
  \draw[thick]
  (1) -- (5) ;
  \draw[thick]
(2) -- (4);
  \draw[thick]
  (3) -- (7) ;
   \draw[thick] (2) -- (6) ;
  \draw[thick] (3) -- (5) ;
  \draw[thick] (3) -- (8) ;  
\end{tikzpicture}}}
  \caption[Examples of different types of graphs]{Examples of some different types of graphs.}
  \label{fig:examplegraph_c2}
\end{figure*}
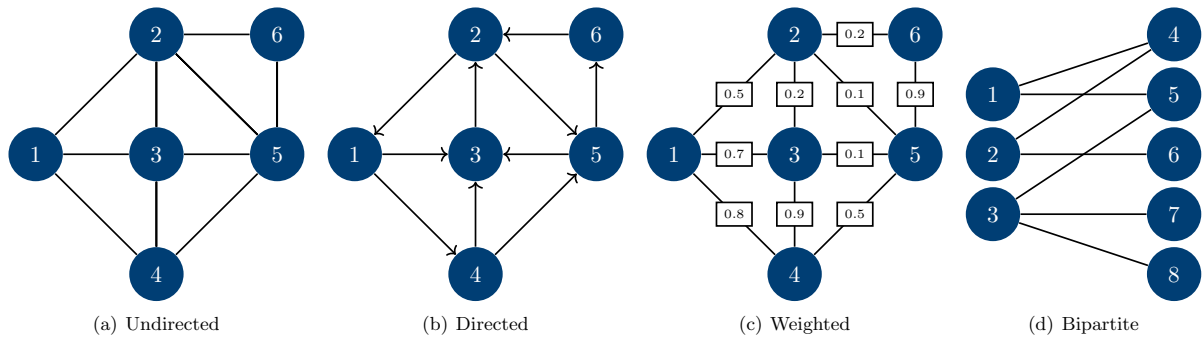

This matrix can represent graphs with various properties, e.g., those illustrated~in~Figure~\ref{fig:examplegraph_c2}:
\begin{itemize}
    \item if $A$ is a \textit{binary} matrix, i.e., $A \in \{0,1\}^{n \times n} $, then $\mathcal{G}$ is an \textit{unweighted} graph;
    \item if $A_{ii} >0$ for any $i \in \{1,\dots,n\}$, then $\mathcal{G}$ includes a \textit{self-loop} connecting the node $i$ to itself;
    \item if $A$ is a \textit{symmetric} matrix, i.e.,  $A_{ij} = A_{ji}$ for all $(i, j) \in \{1,\dots,n\}^2$, then $\mathcal{G}$ is an \textit{undirected} graph. It corresponds to cases where all edges are bidirectional. For instance, the Facebook social network is an undirected graph (if the user $i$ is a ``friend'' of the user $j$, then $j$ is also a ``friend'' of $i$) while the Twitter social network is \textit{directed} (connections can be asymmetric, i.e., $i$ can ``follow'' $j$ while $j$ does not ``follow'' $i$ back);
    \item if $\mathcal{V}$ can be partitioned into two disjoint sets $\mathcal{V}_1$ and $\mathcal{V}_2$ with $A_{ij} > 0 \implies i \in \mathcal{V}_1$ and $j\in\mathcal{V}_2$, then $\mathcal{G}$ is a \textit{bipartite} graph where any two nodes from the same partition set are unconnected. For instance, if $\mathcal{V}_1$ denotes a set of Deezer users, and $\mathcal{V}_2$ denotes a set of music tracks, then an edge $(i,j)$ in such a graph could represent the fact that the user $i$ likes the track $j$.
\end{itemize}
Definitions~\ref{defgraph} and \ref{def:adj} can also be straightforwardly extended to represent more complex structures.
For instance, in Chapter~\ref{chapter_9}, we will consider \textit{dynamic graphs}, where nodes and edges can appear or disappear over time. They are often summarized by $T$ discrete snapshots of standard graphs $\mathcal{G}_1 = (\mathcal{V}_1, \mathcal{E}_1)$, $\mathcal{G}_2 = (\mathcal{V}_2, \mathcal{E}_2)$, $\dots$, $\mathcal{G}_T = (\mathcal{V}_T, \mathcal{E}_T)$ depicting the structural evolution over time \cite{kazemi2020representation}, and therefore by $T$ adjacency matrices $A_1$, $A_2$, $\dots$, $A_T$.
Moreover, in Chapter~\ref{chapter_10}, we will consider structures of the form $\mathcal{G} = (\mathcal{V}, \mathcal{E}_1, \mathcal{E}_2, \dots, \mathcal{E}_p)$ where nodes can be connected through edges of $p$ different \textit{natures}~\cite{epure2020modeling}. They can be summarized as well by $p$ adjacency matrices $A_1$, $A_2$, $\dots$, $A_p$, each of them indicating edges of a particular nature. 

We also use $A$ to introduce the concept of neighboring nodes, which will be useful in our work.
\begin{definition}
The \textit{incoming neighbors} $\mathcal{N}_{\text{in}}(i)$ and \textit{outcoming neighbors} $\mathcal{N}_{\text{out}}(i)$ of a node $i \in \mathcal{V}$ from a directed graph $\mathcal{G} = (\mathcal{V}, \mathcal{E})$, with adjacency matrix $A$, correspond to nodes ``pointing towards $i$'' and ``towards which $i$ points'' through an edge, respectively. More formally:
\begin{equation}
    \mathcal{N}_{\text{in}}(i) = \{ j \in \mathcal{V}; A_{ji} > 0 \}
    \text{ and }
    \mathcal{N}_{\text{out}}(i) = \{ j \in \mathcal{V}; A_{ij} > 0 \}.
\end{equation}
For undirected graphs (where $A_{ij} = A_{ji}$), we simply denote by $\mathcal{N}(i)$ the \textit{neighborhood} of the node $i$, defined as $\mathcal{N}(i) =\mathcal{N}_{\text{in}}(i) = \mathcal{N}_{\text{out}}(i)$. This same set will occasionally be referred to as the \textit{direct neighborhood} or the \textit{one-hop neighborhood} of the node $i$, in chapters where more general definitions of neighborhoods will concurrently be introduced.
\label{defneighbors}
\end{definition}

In addition to the adjacency matrix $A$, we introduce several other matrices also frequently used for graph analysis.
 \begin{definition}
 \label{def:degree_mat}
The \textit{in-degree matrix} $D_{\text{in}}$ and \textit{out-degree matrix} $D_{\text{out}}$ of a graph $\mathcal{G} = (\mathcal{V}, \mathcal{E})$, with adjacency matrix $A$, are the $n \times n$ diagonal matrices defined as: 
  \begin{equation}
     D_{\text{in}} =\mathrm{diag}(A^T \mathbf{1}_n) \text{ and }  D_{\text{out}} =\mathrm{diag}(A \mathbf{1}_n),
  \end{equation}
where $\mathbf{1}_n$ denotes the $n$-dimensional vector containing $n$ entries all equal to $1$. For undirected graphs (where $A^T = A$) we simply denote by $D$ the \textit{degree matrix} defined as $D = D_{\text{in}} = D_{\text{out}}$.
 \end{definition}
 
In a nutshell, the $i$-th diagonal element of $D_{\text{in}}$ (respectively, of $D_{\text{out}}$) corresponds to the sum of weights of all edges pointing towards node $i$ (resp., going
out of node $i$) in the graph under consideration. For unweighted graphs, this coincides with the number of incoming (resp., outcoming) neighbors of $i$. In the case of an undirected degree matrix $D$, diagonal elements equal the number of neighbors of $i$. 

From Definition~\ref{def:degree_mat}, we subsequently introduce the \textit{symmetric normalization} of $A$. 

\begin{definition}
The \textit{symmetric normalization} of the adjacency matrix $A$ of an undirected graph $\mathcal{G} = (\mathcal{V}, \mathcal{E})$ with degree matrix $D$ is:
\begin{equation}
\tilde{A} = (D+I_n)^{-\frac{1}{2}}(A+I_n)(D+I_n)^{-\frac{1}{2}
},
\end{equation}
where $I_n$ denotes the $n \times n$ identity matrix.
\label{def:norm_c2}
\end{definition}

In $\tilde{A}$, each element $(i,j)$ of the original adjacency matrix is re-weighted according to the degrees of nodes $i$ and $j$. Specifically, we have $\tilde{A}_{ij} = \frac{A_{ij}}{\sqrt{(D_{ii} +1)(D_{jj}+1)}}$ for all $(i,j) \in \mathcal{V} \times \mathcal{V}$ with $i\neq j$, and $\tilde{A}_{ii} = \frac{A_{ii}+1}{D_{ii} +1}$ ($= \frac{1}{D_{ii} +1}$ in the absence of self-loop) for all $i \in \mathcal{V}$. $\tilde{A}$ will often be preferred to $A$ in the \textit{graph neural networks} we will introduce thereafter, for reasons related to information propagation that will be detailed along with the presentation of these methods (see Section~\ref{c2s23}~and~\ref{c2s24}). Other normalizations of $A$, such as variants involving $D_{\text{in}}$ or $D_{\text{out}}$ for directed graphs, will also occasionally be mentioned throughout this thesis (see, e.g., Chapter~\ref{chapter_5}). Besides, other transformations of the adjacency matrix with useful properties have also been proposed to represent graphs. This includes \textit{Laplacian matrices} \cite{merris1994laplacian,von2007tutorial}, defined as follows.

\begin{definition}
The \textit{unnormalized Laplacian matrix} of an undirected graph $\mathcal{G} = (\mathcal{V}, \mathcal{E})$, with adjacency matrix $A$ and degree matrix $D$, is defined as:
\begin{equation}
L_{\text{un}} = D - A.
\label{eq:c2laplacian}    
\end{equation}
In addition, several matrices are sometimes referred to as normalized Laplacian matrices \cite{von2007tutorial}. This includes the \textit{random-walk normalized Laplacian matrix}, defined as:
\begin{equation}
L_{\text{rw}} = D^{-1}L_{\text{un}},
\label{eq:c2laplacianrw}    
\end{equation}
as well as the \textit{symmetrically normalized Laplacian matrix}, defined as:
\begin{equation}
L_{\text{sym}} = D^{-1/2}L_{\text{un}}D^{-1/2}.
\label{eq:c2laplaciansym}    
\end{equation}
\label{def:lap}
\end{definition}

Laplacian matrices are central to \textit{spectral graph theory}~\cite{Chung1997} and to numerous related methods such as \textit{Laplacian eigenmaps} and \textit{spectral clustering} \cite{ng2002spectral,spielman2007spectral,von2007tutorial} that we will introduce in Section~\ref{c2s22} and consider in several experiments throughout this thesis. We refer to Von Luxburg~\cite{von2007tutorial} and Lutzeyer~\cite{Lutzeyer2020} for an in-depth presentation of the differences, the applications, and the mathematical properties of these Laplacian matrices, ranging from their symmetry to their positive semi-definiteness and their link to the number of connected components in~$\mathcal{G}$. 

Lastly, some graphs mentioned in our work will be \textit{attributed graphs}. Each node will also be described by its own \textit{feature} vector, a.k.a. \textit{attribute} vector or \textit{descriptive} vector \cite{hamilton2020graph}. We will encapsulate these descriptions in a \textit{node feature matrix} $X$. 

\begin{definition}
The \textit{node feature matrix} $X$ of an attributed graph $\mathcal{G} = (\mathcal{V}, \mathcal{E})$, in which each node $i$ is also described by its own vector $x_i \in \mathbb{R}^f$, is the $n \times f$ matrix stacking up all $x_i$ vectors:
\begin{equation}
    X= [x_1, \ldots, x_n]^T,
\end{equation}
i.e., the $i$-th row of $X$ corresponds to the feature vector $x_i$ of node $i$.
\label{def:featurematrix}
\end{definition}

\subsection{Machine Learning on Graphs}
\label{c2s212}

A wide range of machine learning problems involve graphs, and require extracting relevant information from the nodes and edges of such structures~\cite{hamilton2020graph,hamilton2017representation,sun2012mining,wu2019comprehensive,zhang2018network}. In the following, we present some of these problems. We primarily focus on \textit{link prediction} and on \textit{community detection}, which will be the two evaluation tasks mainly considered in our experiments, as they are closely related to our applications at Deezer. We refer to several surveys \cite{fortunato2010community,hamilton2020graph,kumar2020link,malliaros2013clustering,shi2016survey,wu2019comprehensive} for a broader overview of graph-based~machine~learning problems.

\paragraph{Link Prediction} The \textit{link prediction} task~\cite{kumar2020link,liben2007link} consists in inferring the presence of new or unobserved edges between some pairs of nodes, based on the observed edges in the graph. Over the past decades, this problem initiated significant research efforts from the scientific community, with various real-world applications ranging from recommending new ``friends'' in social networks to predicting interactions between proteins \cite{chen2012discovering,kipf2016-2,kumar2020link,liben2007link,lichtenwalter2010new,shi2016survey,yang2015evaluating,zhang2018link}. More formally, we define a link prediction model as follows.

\begin{definition}
Let us consider a graph $\mathcal{G} = (\mathcal{V},\mathcal{E})$ in which some edges would be ``missing'', e.g., masked or yet to appear. 
A \textit{link prediction} model for such a graph is a function:
\begin{equation}
p: \mathcal{V} \times\mathcal{V}  \setminus \mathcal{E} \rightarrow [0, 1].
\end{equation}
For a given pair $(i,j)$ of unconnected nodes, it returns an estimated probability $p(i,j)$ of a missing edge connecting $i$ to $j$ in the graph $\mathcal{G} = (\mathcal{V},\mathcal{E})$ under consideration. 
\label{deflp}
\end{definition}

Definition~\ref{deflp} is very general. For instance, $p$~could correspond to a node similarity measure directly computed from the graph, e.g., to the proportion of common edges between two nodes\footnote{Such node similarity measures usually rely on homophily assumptions. This term describes the tendency of nodes to connect to ``similar'' nodes in the graph, which is observed in numerous real-world applications \cite{kumar2020link}.}. Alternatively, $p$~could also correspond to a more sophisticated model with trainable parameters, e.g., to a neural network~\cite{kumar2020link}. To assess the performance of such a model, researchers often adopt an evaluation methodology~\cite{kipf2016-2,pan2018arga,zhang2018link} consisting in:
\begin{itemize}
    \item firstly, training this model on an (artifically) incomplete version of a graph, for which only a certain percentage of randomly sampled edges (for instance, 85\% in \cite{kipf2016-2}) are visible;
    \item then, constructing validation and test sets gathering:
    \begin{itemize}
        \item node pairs corresponding to missing edges ($5\%$ and $10\%$, respectively, in \cite{kipf2016-2});
        \item the same number of randomly picked unconnected node pairs in the graph;
    \end{itemize} 
    \item finally, evaluating the model's ability to distinguish edges from non-edges in~these~sets, using the model trained on the incomplete graph.
\end{itemize}  
While all node pairs in such validation and test sets are observed to be unconnected, half of them actually correspond to missing edges from the original graph. Under this formulation, link prediction acts as a \textit{binary classification} task, assessing to which extent the model correctly locates these missing edges despite their absence during training.

Researchers evaluate model performance on this link prediction task through binary classification metrics \cite{hossin2015review}, such as the popular \textit{Area Under the ROC Curve} (AUC) \cite{fawcett2006introduction,mcclish1989analyzing} and \textit{Average Precision} (AP) \cite{robertson2008new} scores, that we will adopt and discuss as well throughout~our~work\footnote{We will use the implementations provided in the \textit{scikit-learn} Python library for machine learning \cite{pedregosa2011scikit} and respectively described here: \href{https://scikit-learn.org/stable/modules/generated/sklearn.metrics.roc_auc_score.html}{https://scikit-learn.org/stable/modules/generated/sklearn.metrics.roc\_auc\_score.html} and here: \href{https://scikit-learn.org/stable/modules/generated/sklearn.metrics.average_precision_score.html}{https://scikit-learn.org/stable/modules/generated/sklearn.metrics.average\_precision\_score.html}.}.

\paragraph{Community Detection} Another fundamental problem in graph-based machine learning, community detection consists in identifying $K < n$ clusters a.k.a. \textit{communities} of nodes that, in some sense, are more similar to each other than to the other nodes \cite{fortunato2010community,salhagalvan2022modularity}. As link prediction, this problem initiated significant research efforts \cite{du2007community,fortunato2010community,liu2020deep,malliaros2013clustering,papadopoulos2012community,plantie2013survey,salhagalvan2022modularity,wang2017mgae}.
Community detection has numerous real-world applications. This includes the segmentation of websites in a web graph according to thematic categories \cite{huang2006web,malliaros2013clustering}, as well as the detection of densely connected subgroups of users (actual ``\textit{communities}'') in online social networks \cite{du2007community,papadopoulos2012community,plantie2013survey}. More formally, we define a community detection model as follows.

\begin{definition}
Let us consider a graph $\mathcal{G} = (\mathcal{V},\mathcal{E})$. Denoting by $\mathcal{P}_K(\mathcal{V})$ the set of partitions of $\mathcal{V}$ of cardinality $K \leq n$, a \textit{community detection} model is a function:
\begin{equation}
c: \mathcal{G} \rightarrow \mathcal{P}_K(\mathcal{V}),
\end{equation}
Given the observed graph, it returns a partition of the node set $\mathcal{V}$ into $K$ sets, denoted as follows:
\begin{equation}
C_1 \subseteq \mathcal{V}, \ldots, C_K \subseteq \mathcal{V}.
\end{equation}
We denote $n_k = |C_k|$, for $k \in \{1, \ldots,K\}$.
\label{defcd}
\end{definition}

\begin{figure}[H]
    \centering
    \includegraphics[width=0.42\textwidth]{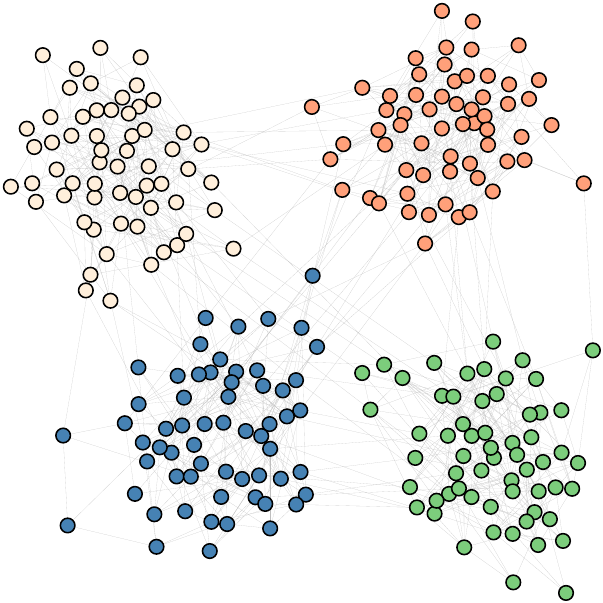}
    \caption[A graph with four ground truth communities]{A graph with four ground truth communities, denoted by node colors. This synthetic graph was obtained through a stochastic block model \cite{abbe2017community}, which generates community-based random graphs.}
    \label{fig:sbm_c2}
\end{figure}

\begin{multicols}{2}
In the scientific literature, the quality of such a partition is usually assessed through some predefined similarity metrics, e.g., unsupervised density-based metrics computed from the intra- and inter-cluster edge density
\citep{malliaros2013clustering}, or scores such as the \textit{Adjusted Mutual Information} (AMI) \cite{vinh2010information} and the \textit{Adjusted Rand Index} (ARI) \cite{hubert1985comparing} that compare the partition to some ground truth node labels (when such labels are available) hidden during training. We will adopt and discuss these metrics as well throughout our work\footnote{Again, we will use the implementations provided in \textit{scikit-learn} \cite{pedregosa2011scikit} and respectively described here: \href{https://scikit-learn.org/stable/modules/generated/sklearn.metrics.adjusted_mutual_info_score.html}{https://scikit-learn.org/stable/modules/generated/sklearn.metrics.adjusted\_mutual\_info\_score.html} and here: \href{https://scikit-learn.org/stable/modules/generated/sklearn.metrics.adjusted_rand_score.html}{https://scikit-learn.org/stable/modules/generated/sklearn.metrics.adjusted\_rand\_score.html}.}. Figure~\ref{fig:sbm_c2} provides a illustrative example of a graph generated from a stochastic block model~\cite{abbe2017community}, where nodes are partitioned in $K = 4$ communities.
\columnbreak
\end{multicols}
\vspace{-0.5cm}

As is the case for link prediction, Definition~\ref{defcd} is voluntarily very general. Specific examples of community detection methods will be presented in the next sections and chapters. Also, while this definition implicitly assumes that $K$ is known and fixed, we emphasize that some community detection methods instead involve an automatic selection of the appropriate number of communities to detect~\cite{blondel2008louvain}. Besides, while the above partitioning implies that communities are \textit{disjoint}, several overlapping community detection methods were also proposed in the scientific literature~\cite{hollocou2019modularity,kaufmann2016spectral,zhang2020detecting}. They are out of the scope of this thesis.

\paragraph{Other Tasks}

Several other graph-based machine learning problems have also been introduced and widely studied over the last decades. While they will be less or not discussed in this thesis, we nevertheless mention several of them in this paragraph along with some indicative references. Examples of such other problems include: \textit{influence maximization}, which consists in identifying the most ``influent'' nodes in a graph, e.g., the ones that will maximize the spread of a rumor in a social network according to some information diffusion process~\cite{cautis2019adaptive,golovin2011adaptive,kempe2003maximizing,li2018influence,salha2018adaptive}; \textit{graph-level clustering}, where the goal is to partition a set of several entire graphs, into groups of structurally similar ones \cite{dai2016discriminative,duvenaud2015convolutional,hamilton2017representation,nikolentzos2019graph}; \textit{graph generation}, consisting in generating new graphs verifying some desirable properties, e.g., new biologically plausible molecules~\cite{molecule3,liao2019efficient,molecule1,molecule2,simonovsky2018graphvae}; and several other problems ranging from supervised/semi-supervised \textit{node classification} \cite{Dasoulas2021,hamilton2017inductive,kipf2016-1,velivckovic2019graph,wu2019simplifying} to \textit{edge classification} \cite{aggarwal2016edge,cesa2012correlation,lee2019hyperlink}.

\section{Learning Node Embedding Representations}
\label{c2s22}

To address these machine learning problems, significant efforts were recently devoted to the development of \textit{representation learning} methods. In the remainder of this Chapter~\ref{chapter_2}, we propose an introduction to this learning paradigm. For consistency with the content of this thesis, we focus on methods learning vectorial representations of \textit{nodes} in an embedding space, and omit other existing methods, e.g., those designed for edge-level or graph-level representation learning \cite{maddalena2020whole,narayanan2017graph2vec,rennard2020graph,verma2019heterogeneous}. Moreover, at times, we partly draw inspiration from Hamilton's recent book on graph representation learning~\cite{hamilton2020graph} to organize and present some key concepts. We refer to this book for more details and references on this topic.

Specifically, in this Section~\ref{c2s22}, we present the general objectives of representation learning on graphs. We also mention several popular methods, referred to as ``\textit{shallow}'' by Hamilton \cite{hamilton2020graph}, to learn node embedding spaces. In Section~\ref{c2s23}, we will consider representation learning with \textit{graph neural networks}~(GNNs). Finally, in Section~\ref{c2s24}, we will focus on the two GNN-based methods at the center of this thesis: \textit{graph autoencoders}~(GAEs) and \textit{variational graph autoencoders}~(VGAEs).

\subsection{From Graph Structures to Vectorial Representations}

Traditional approaches to tackle machine learning problems such as link prediction and community detection often directly operate on the graph structure under consideration. For instance, predicting the most likely locations of missing links has been historically addressed by the construction of hand-engineered \textit{node similarity measures} \cite{hamilton2020graph,kumar2020link,liben2007link} derived from the observed connections. Examples of such measures include the (previously cited) proportion of common neighbors between nodes in the graph, as well as several refined alternatives such as the popular Adamic-Adar, Jaccard, or Katz indices~\cite{liben2007link}. Besides, as of today, one of the most popular approaches for community detection remains the \textit{Louvain} algorithm \cite{blondel2008louvain}, which also directly operates on the graph. This greedy method clusters nodes by iteratively maximizing a graph-based measure referred to as the \textit{modularity}~\cite{Newman8577,salhagalvan2022modularity}. It will compare the observed density of connections in communities to the expected density in a configuration model graph, with equal degree distribution but allocating edges randomly without any specified community structure (we provide a more complete presentation of the Louvain method in Chapter~\ref{chapter_7}).

\begin{figure}[t]
    \centering
    \includegraphics[width=0.85\textwidth]{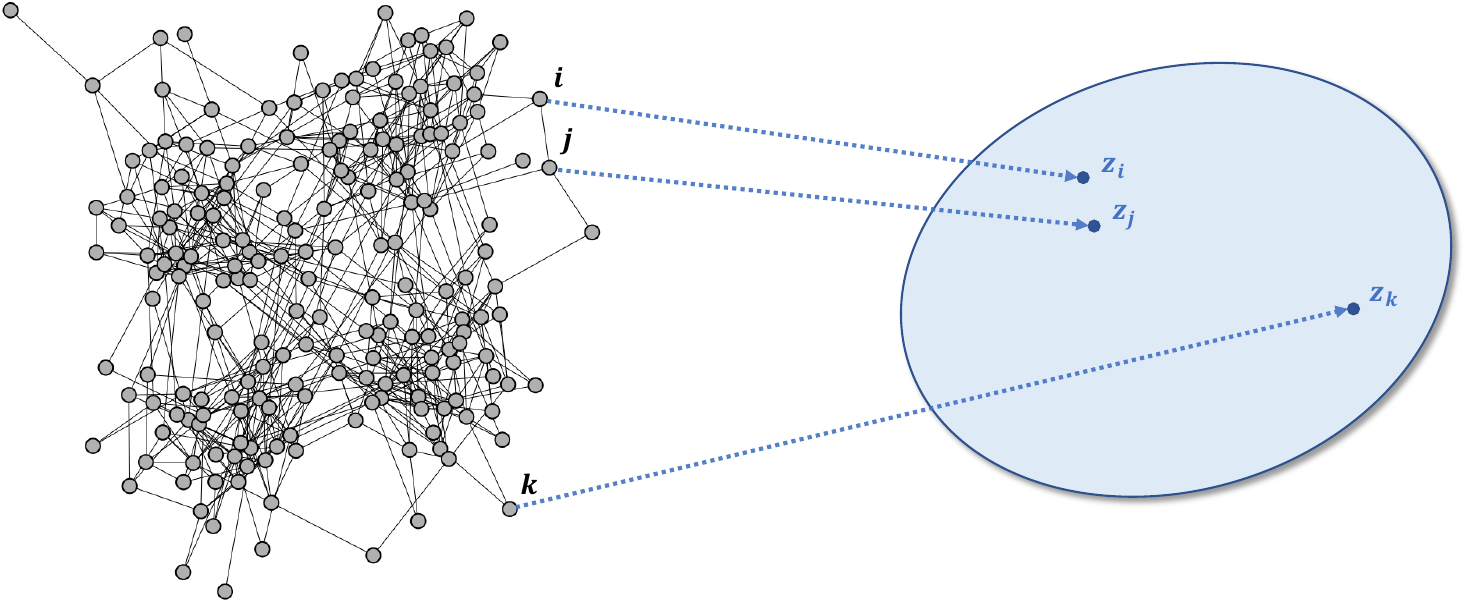}
    \caption[Illustration of the node embedding paradigm]{Illustration of the node embedding paradigm. The objective is to encode nodes from a graph (\textit{left}) as vectors in a low-dimensional vector space (\textit{right}), where nodes with ``structural proximity'' in the graph should be close. In essence, one should be able to subsequently leverage these vectors for downstream applications, such as link prediction (e.g., by assessing the likelihood of a link from $i$ to $j$ by computing $z^T_i z_j$ or $\|z_i - z_j\|_2$) or community detection (e.g., by running a $k$-means algorithm to cluster all $z_i$ vectors).}
    \label{fig:embedding_c2}
\end{figure}

Nonetheless, promising improvements on link prediction, community detection, and various other tasks were recently achieved by methods adopting another strategy. Instead of directly operating on the graph structure itself, these methods aim to \textit{learn} node representations summarizing the graph under consideration \cite{grover2016node2vec,hamilton2020graph,hamilton2017representation,kipf2020phd,kipf2016-1,perozzi2014deepwalk,wu2019comprehensive}. As illustrated in Figure~\ref{fig:embedding_c2}, the objective of these \textit{representation learning} methods is to encode nodes as vectors in a low-dimensional vector space, called a \textit{node embedding} space. In such a space, nodes
with ``structural proximity'' in the graph (which could have various meanings~\cite{donnat2018learning}) should be close. More formally, each node $i \in \mathcal{V}$ from a graph $\mathcal{G}$ will be associated with an embedding vector $z_i \in \mathbb{R}^d$, where $d \ll n$ denotes the dimension of the node embedding space. In the following, we will often encapsulate these embedding vectors in the \textit{node embedding matrix} $Z$ defined~as~follows.

\begin{definition}
The \textit{node embedding matrix} $Z$ of a graph $\mathcal{G} = (\mathcal{V}, \mathcal{E})$, in which each node $i \in \mathcal{V}$ is associated with an embedding vector $z_i \in \mathbb{R}^d$, is the $n \times d$ matrix stacking up all $z_i$ vectors:
\begin{equation}
    Z = [z_1, \ldots, z_n]^T,
\end{equation}
i.e., the $i$-th row of $Z$ corresponds to the embedding vector $z_i$ of node $i$.
\label{def:embedding}
\end{definition}

Over the last years, various methods, including those described thereafter in this chapter, have been proposed to learn such $z_i$ vectors. Authors of these methods successfully optimized and leveraged these low-dimensional vectors, which are arguably easier to handle than a complex graph structure, for various downstream applications~\cite{hamilton2020graph,wu2019comprehensive}. For instance, some methods manage to directly assess the probability of a missing edge between two nodes by evaluating the proximity of these nodes in the embedding space, e.g., through pairwise Euclidean distances or inner products \cite{choong2018learning,kipf2016-2,wang2017mgae}. Similarly, in the presence of node embedding vectors, several studies reframed community detection as the standard problem consisting in clustering $n$ vectors in a $d$-dimensional vector space into $K$ groups~\cite{liu2020deep,macqueen1967some}. Such a problem can then be solved by one of the numerous clustering methods such as the popular $k$-means algorithm~\cite{appert2020information,macqueen1967some}.

Based on the promising results of such a paradigm \cite{hamilton2020graph,kumar2020link,liu2020deep,wu2019comprehensive,zhang2018network}, the graph representation learning field has grown at a fast pace over the past decade \cite{hamilton2020graph}, notably the sub-field involving research on \textit{graph neural networks} (see Sections~\ref{c2s23} and \ref{c2s24}). In the last few years, international conferences on machine learning and data mining often received hundreds of submissions from graph representation learning articles \cite{ivanov2020iclr,ivanov2021express}, and regularly organized dedicated workshops on related topics\footnote{This includes the NeurIPS 2019 workshop on \textit{Graph Representation Learning}: \href{https://nips.cc/Conferences/2019/ScheduleMultitrack?event=13172}{https://nips.cc/Conferences/2019/ScheduleMultitrack?event=13172}; the ICML 2020 workshop on \textit{Graph Representation Learning and Beyond}: \href{https://icml.cc/Conferences/2020/ScheduleMultitrack?event=5715}{https://icml.cc/Conferences/2020/ScheduleMultitrack?event=5715}; and the KDD~2021 workshop on \textit{Deep Learning on Graphs: Methods and Applications}: \href{https://kdd.org/kdd2021/workshops}{https://kdd.org/kdd2021/workshops}.}. Some recent advances were also already successfully transposed to industrial-level applications ranging from product recommendation to malicious behavior detection, including on online services such as Alibaba \cite{wang2018billion}, Pinterest~\cite{ying2018graph}, and Twitter~\cite{bronstein2020twitter}.

\subsection{Node Embeddings from Matrix Factorization}
\label{c2s222}
In the remainder of this chapter, we review several popular approaches aiming to learn node embedding vectors. Firstly, an important part of the existing literature involves \textit{factorization} techniques. 

\paragraph{Laplacian Eigenmaps and Spectral Clustering} In particular, some seminal approaches make use of the \textit{eigendecomposition} of Laplacian matrices introduced in Defintion~\ref{def:lap} \cite{belkin2003laplacian,Lutzeyer2020,ng2002spectral,von2007tutorial}. One can show that these matrices have $n$ non-negative eigenvalues $0 = \lambda_n \leq \lambda_{n-1} \leq \dots \leq \lambda_1$, and that the geometric multiplicity of the 0 eigenvalue corresponds to the number of \textit{connected components} in the graph. Furthermore, the corresponding \textit{eigenvectors}\footnote{The eigenvector $u_{i} \in \mathbb{R}^n$ associated with an eigenvalue $\lambda_i$ of $L_{\text{un}}$ verifies: $L_{\text{un}} u_{i} = \lambda_i u_{i}$.} are indicators of nodes belonging to each of these
components  \cite{von2007tutorial}. As a consequence, these eigenvectors can be used to cluster nodes according to their connected component membership. More interestingly, Laplacian eigenvectors are also useful to cluster nodes into $K$ communities in a connected graph. In this direction, a fundamental approach referred to as \textit{spectral clustering} \cite{ng2002spectral,shi2000normalized} consists in:
\begin{itemize}
    \item computing the $K$ eigenvectors corresponding to the $K$ smallest eigenvalues of $L_{\text{un}}$ (or $L_{rw}$ or $L_{sym}$), and subsequently form the matrix $Z \in \mathbb{R}^{n \times K}$, in which the $K$ columns correspond to these $K$ $n$-dimensional eigenvectors;
    \item represent each node $i \in \mathcal{V}$ by the $i$-th row of $Z$, denoted $z_i$;
    \item run a $k$-means algorithm \cite{macqueen1967some} to cluster the $z_i$ vectors into communities.
\end{itemize}

\begin{figure}[t]
    \centering
    \includegraphics[width=0.85\textwidth]{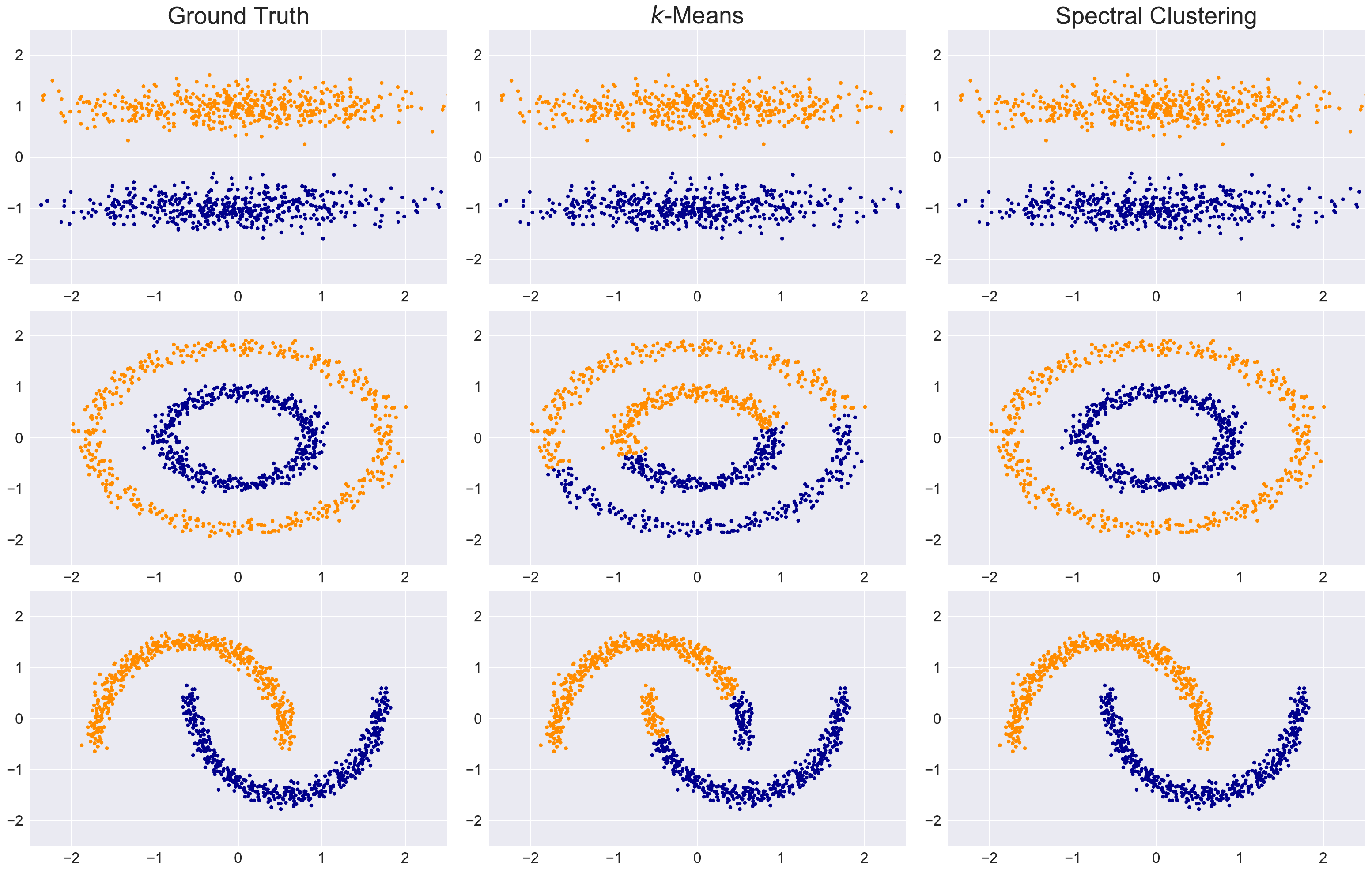}
    \caption[Comparing $k$-means to spectral clustering]{A comparison of $k$-means and spectral clustering. \textit{Left}: three synthetic datasets of 1000 two-dimensional data points, clustered in two ground truth communities displayed in \textcolor{orange}{orange} and~in~\textcolor{blue}{blue}, respectively. \textit{Middle}: the communities retrieved by running $k$-means algorithms in these two-dimensional spaces. \textit{Right}: the communities retrieved by instead 1)~creating a graph connecting data points (acting as nodes) to their ten nearest neighbors, and 2)~applying spectral clustering on graphs. In 2/3 cases, a standard $k$-means fails to properly identify communities. This approach directly relies on Euclidean distances in the input space and extracts ``compact'' communities, which is relevant for dataset n°1 but not for datasets n°2 and n°3. On the contrary, the graph-based spectral clustering approach extracts and leverages the ``connectivity'' between data points which, while being more computationally expensive, permits identifying ground truth communities on all three datasets.}
    \label{fig:twomoons_c2}
\end{figure}

One can prove that communities identified via spectral clustering correspond to those obtained from the \textit{relaxation} of NP-hard graph partitioning optimization problems, consisting in minimizing metrics that explicitly evaluate the quality of a node partition in a graph (specifically, a \textit{RatioCut} or \textit{NCut} metric depending on Laplacian matrices \cite{von2007tutorial}). Spectral clustering can also interestingly be studied under the lens of perturbation theory~\cite{von2007tutorial}, bringing additional theoretical justifications for such an approach. Overall, as illustrated in Figure~\ref{fig:twomoons_c2}, leveraging point/node-level connections permits identifying cluster structures where, in essence, ``\textit{connectivity}'' is more relevant than ``\textit{compactness}'' and where a standard $k$-means would therefore fail. Also, while a naive eigendecomposition of a Laplacian matrix incurs a cubic computational complexity $O(n^3)$ \cite{hamilton2020graph},  several approximate methods have been studied to overcome such a cost \cite{boutsidis2015spectral,de2019sparse,fowlkes2004spectral}. Moreover, besides clustering, the above eigendecomposition procedure has also been interpreted as an effective way to project nodes in a low-dimensional node embedding space preserving connection information from the initial graph structure.  The $z_i$ vectors from the above matrix $Z$ correspond to \textit{embedding vectors} (hence the same notations as Definition~\ref{def:embedding}) and such an approach is often referred to as \textit{Laplacian eigenmaps} in the literature \cite{belkin2003laplacian,hamilton2020graph}. 

\paragraph{Inner product methods} In parallel, several other studies introduced factorization-based methods that rather leverage the \textit{inner product} $z^T_i z_j$, between two node embedding vectors $z_i$ and $z_j$, as an indicator of the structural proximity of nodes $i$ and $j$ in the graph. From a matrix perspective, these methods aim to learn an $n \times d$ node embedding matrix $Z$ such as:
\begin{equation}
f(A) \approx Z Z^T,
\end{equation}
for some pre-defined function $f: \mathbb{R}^{n \times n} \rightarrow \mathbb{R}^{n \times n}$. Examples of such methods include the \textit{distributed graph factorization} model \cite{ahmed2013distributed} as well as the \textit{GraRep}~\cite{cao2015grarep} and \textit{HOPE}~\cite{ou2016asymmetric} methods. As explained by Hamilton~\cite{hamilton2020graph}, these methods primarily differ in how $f$ is defined. While Ahmed et al.~\cite{ahmed2013distributed} directly leverage the adjacency matrix, i.e., $f(A) = A$, authors of GraRep and HOPE exploit more general matrices, based on powers of $A$~\cite{cao2015grarep} or on neighborhood overlap measures~\cite{ou2016asymmetric}. They aim to minimize loss functions corresponding or proportional to:
\begin{equation}
\mathcal{L} \approx \| f(A) - Z Z^T \|^2_2,
\end{equation}
which can be achieved through matrix factorization algorithms such as \textit{Singular Value Decomposition} (SVD) \cite{golub1971singular,stewart1993early}. We refer to Klema and Laub~\cite{klema1980singular} for a detailed introduction to SVD and its applications.

\subsection{Node Embeddings from Random Walks}
\label{c2s223}
\begin{multicols}{2}

\begin{figure}[H]
    \centering
    \begin{tikzpicture}[bn/.style={circle,fill=ImperialColor,text=white,minimum
size=9mm},every node/.append style={bn}]
 \path node (4)[fill=orange] {4} --  (2,2) node (5) {5} -- (0,2) node (3)[fill=orange] {3}
 -- (-2,2) node (1)[fill=orange] {1} -- (0,4) node (2) {2}
 -- (2,4) node (6) {6} ;
 \draw[->,thick]
  (1) -- (3);
  \draw[->,thick,color=orange]
  (1) -- (4) ;
  \draw[->,thick]
(5) -- (3);
  \draw[->,thick]
  (2) -- (5) ;
  \draw[->,thick] (6) -- (2) ;
    \draw[->,thick,color=orange] (4) -- (3);
    \draw[->,thick] (3) -- (2);
    \draw[->,thick] (2) -- (1);
    \draw[->,thick] (4) -- (5);
    \draw[->,thick] (5) -- (6);
\end{tikzpicture}
    \caption[An example of a random walk of length 3]{An example of a random walk of length $T=3$, on the small directed graph from Figure~\ref{fig:examplegraph_c2}(b). This random walk, colored in \textcolor{orange}{orange}, starts at node n°1, then ``visits'' node n°4 and ends at node n°3.}
    \label{fig:rw_c2}
\end{figure}
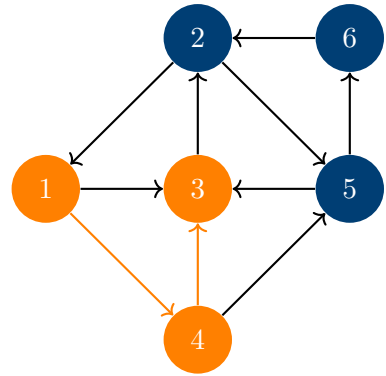

We now turn to another family of embedding methods, relying on \textit{stochastic} strategies. In such methods, nodes end up with similar embedding vectors when they frequently co-occur on short \textit{random walks} computed over the graph \cite{perozzi2014deepwalk}. As illustrated in Figure~\ref{fig:rw_c2}, a random walk is a list of nodes of length $T > 1$. The first node is selected from $\mathcal{V}$, and then the $i$-th node for $i \in \{2,...,T\}$ is subsequently sampled from the neighbors of the $(i-1)$-th node, either uniformly at random or following some pre-defined sampling~technique~\cite{huang2021broader}. 
\columnbreak
\end{multicols}
\vspace{-0.5cm}
Random walk-based methods often explicitly draw inspiration from \textit{word2vec}-like models \cite{w2v2,w2v1} from natural language processing (NLP). These shallow neural network models learn word embedding representations, ensuring that words appearing in the same contexts, e.g., the same sentences, end up with close word embedding vectors. Here, nodes and random walks act as the counterparts of words and sentences, respectively.

More formally, and albeit under different formulations \cite{grover2016node2vec,hamilton2020graph,huang2021broader,perozzi2014deepwalk}, these node embedding methods usually encode nodes as $z_i$ embedding vectors that should approximately verify:
\begin{equation}
p_T(j|i) \approx \frac{e^{z^T_i z_j}}{\sum\limits_{k \in \mathcal{V}} e^{z^T_i z_{k}}},   
\end{equation}
where $p_T(j|i)$ denotes the probability that the node $j \in \mathcal{V}$ appears on a random walk of length $T$ starting from node $i \in \mathcal{V}$. Examples of such methods include the popular \textit{DeepWalk}~\cite{perozzi2014deepwalk} and \textit{node2vec}~\cite{grover2016node2vec} models. Precisely, DeepWalk and node2vec adopt different strategies to learn $z_i$ vectors from random walks, either based on hierarchical softmax computations involving a fast binary-tree structure \cite{perozzi2014deepwalk} or a noise contrastive learning approach with negative sampling \cite{grover2016node2vec} (we refer to Hamilton~\cite{hamilton2020graph} as well as to original research articles on these methods \cite{grover2016node2vec,perozzi2014deepwalk} for technical details, that we omit in this brief overview). Moreover, node2vec includes two additional hyperparameters, the ``\textit{return}'' parameter $p$ and the ``\textit{in-out}'' parameter $q$, providing more flexibility when sampling random walks. They permit interpolating between random walks more akin to a breadth-first search or a depth-first search \cite{cormen2009introduction} over the graph. DeepWalk and node2vec have been explicitly presented as \textit{scalable} methods in their original articles, backed by experimental evaluations on graphs with up to a few million nodes or edges \cite{grover2016node2vec,perozzi2014deepwalk}, observing linear increases in running times w.r.t. $n$ in the case of node2vec \cite{grover2016node2vec}. 

Over the last few years, several other random walk-based methods have been proposed, e.g., with different architectures or sampling schemes. This includes \textit{dynnode2vec}~\cite{mahdavi2018dynnode2vec},  \textit{struct2vec}~\cite{ribeiro2017struc2vec} and \textit{Walklets}~\cite{perozzi2017don}. As explained by Hamilton~\cite{hamilton2020graph}, the popular \textit{LINE} method~\cite{tang2015line} is also often presented among random walk approaches. While it does not explicitly draw random walks, LINE nonetheless shares strong conceptual motivations with DeepWalk and node2vec (see~\cite{hamilton2020graph}). Moreover, related techniques have also been extended to multi-relational data and knowledge graphs~\cite{wang2017knowledge}. More recently, Qiu et al.~\cite{qiu2018network} proved that methods such as DeepWalk, LINE, and node2vec implicitly approximate and factorize some matrices, for which they derived closed forms. Therefore, to some extent, they unified random walk-based learning with the matrix factorization framework from Section~\ref{c2s222}.
 
\subsection{On the Limitations of these Methods}
\label{c2s224}

The random walks and matrix factorization methods we presented in this Section~\ref{c2s22} are referred to as ``\textit{shallow}'' by Hamilton~\cite{hamilton2020graph}. In his book, this term encompasses node embedding methods consisting in a simple embedding lookup based on a node's ID, and where an encoder directly optimizes a unique embedding vector for each node. While these methods led to promising applications over the past years \cite{hamilton2020graph,hamilton2017representation,huang2021broader}, they nonetheless suffer from several limitations:
\begin{itemize}
    \item firstly, the encoding function, which maps nodes into the embedding space, does not share parameters between nodes. Hamilton~\cite{hamilton2020graph} argues that this is inefficient from a statistical point of view, and that parameter sharing could permit learning representations capturing properties of the original graph structure more efficiently;
    \item besides, while attributed graphs are ubiquitous, these methods do not leverage node features/attributes. Such additional information, summarized in the matrix~$X$ from Definion~\ref{def:featurematrix}, could however be relevant and useful in the learning process;
    \item lastly, these embedding methods are \textit{transductive}~\cite{hamilton2017inductive}. By design, they only learn representations of nodes available during training. Without additional optimization steps, they usually can not provide representations for new nodes. This prevents applying such methods to \textit{inductive} settings, that specifically require such generalization to new nodes after training. Such a generalization would be desirable for several applications mentioned throughout this thesis.
\end{itemize}

In the following sections, we review alternative node embedding methods that address these limitations. They rely on graph neural networks, which emerged as a popular formalism in the last years \cite{wu2019comprehensive}. While random walk and matrix factorization methods will still be mentioned in the following chapters, as relevant baselines for various experimental analyses, graph neural networks will be at the center of our work, especially through the lens of graph autoencoders.

\section{Representation Learning with Graph Neural Networks}
\label{c2s23}

We now turn to \textit{graph neural networks} (GNNs), a general framework to extend deep neural networks \cite{goodfellow2016deep,lecun2015deep} to graph structures. As an introduction, it is worth noting that a straightforward use of standard \textit{deep learning} models on graphs is usually a bad idea. For instance, by directly training a \textit{convolutional neural network} (CNN) \cite{fukushima1982neocognitron,lecun1999object} on an adjacency matrix~$A$ treated as a two-dimensional regular grid, one would aggregate information from a node $i$ with information from nodes corresponding to rows/columns $i - 1$ and $i +1$ in convolutional layers. While such an aggregation is relevant for neighboring pixels in an image, it is inappropriate for graphs as the node ordering in $A$ is usually \textit{arbitrary}. The node associated with the $(i -1)$-th row/column of $A$ will not necessarily be connected or even ``close'' to the one associated with the $i$-th row/column. Alternatively, one could also want to learn node embedding vectors by directly processing a flattened version of $A$ using a \textit{multi-layer perceptron} (MLP) \cite{rosenblatt1958perceptron}. Again, in such an approach, the resulting node embedding representations would depend on the arbitrary ordering of nodes in $A$~\cite{hamilton2020graph}. On the contrary, researchers will often favor neural network models that are \textit{permutation equivariant}, meaning that a permutation of the node ordering would permute the output representations in a consistent way \cite{keriven2019universal}.

\subsection{Neural Message Passing and Graph Neural Network (GNN) Models}

Although numerous GNN models have been proposed and theoretically motivated over the last years, sometimes under different formalisms, most of them actually leverage a \textit{neural message passing} process in
which vector ``\textit{messages}'' are iteratively exchanged between nodes and updated \cite{gilmer2017neural,hamilton2020graph,nikolentzos2020message,xu2019powerful}. Comprehensively, a GNN model usually takes as input:
\begin{itemize}
    \item an adjacency matrix $A$, or any variant such as one of those presented in Section~\ref{c2s21};
    \item some \textit{input vectors} providing an initial representation for each node $i \in \mathcal{V}$. In numerous applications, graphs will be attributed and these vectors will correspond to \textit{node features} $x_i \in \mathbb{R}^f$, summarized in the $n \times f$ matrix $X$ from Definition~\ref{def:featurematrix}. In the absence of such features, one can use node-level statistics~\cite{hamilton2020graph}. Alternatively, another popular approach consists in simply setting $X = I_n$, i.e., the $n \times n$ identity matrix\footnote{Nonetheless, as we will emphasize in our work, such a choice will also make the GNN model \textit{transductive}.}, thus mapping each node to a one-hot indicator \cite{kipf2016-2,kipf2016-1}.
\end{itemize}
Then, a GNN usually consists in the succession of $L\geq 1$ \textit{message passing iterations} a.k.a. $L$~\textit{layers} by analogy with standard neural networks. At each iteration $l \in \{1,\dots,L\}$, we will obtain a \textit{hidden embedding vector} $h^{(l)}_i \in \mathbb{R}^{d_l}$ for each node $i \in \mathcal{V}$ in the graph\footnote{We note that dimensions $d_1, \dots, d_L$ of these vectors can differ across layers.}. In particular, we set $d_L = d$ for the last layer.
To compute such representations, we adopt a two-step process. Firstly, we \textit{aggregate} vectors $h^{(l-1)}_j$ from nodes in the \textit{neighborhood} $\mathcal{N}(i)$ of node $i$ to derive the message $m^{(l-1)}_{\mathcal{N}(i)}$ to pass in layer $l$. This aggregation can correspond to a sum, an average, a concatenation or to some more complex operation \cite{hamilton2020graph}. Also, although $\mathcal{N}(i)$ will usually correspond to nodes directly connected to $i$, consistently with our notation from Definition~\ref{defneighbors}, more general definitions of neighboring nodes have also been adopted in the scientific literature~\cite{hamilton2017inductive,klicpera2019diffusion}. To sum up, in this first step, we derive:
\begin{equation}
m^{(l-1)}_{\mathcal{N}(i)} = \text{AGGREGATE}^{(l)} \Big( \{ h^{(l-1)}_j; j \in \mathcal{N}(i) \}) \Big).
\label{eq:MP1}
\end{equation}
Secondly, we merge $m^{(l-1)}_{\mathcal{N}(i)}$ with node $i$'s own previous representation $h^{(l-1)}_i$, to obtain $h^{(l)}_i$:
\begin{equation}
h^{(l)}_i = \text{UPDATE}^{(l)} \Big( h^{(l-1)}_i, m^{(l-1)}_{\mathcal{N}(i)} \Big).
\label{eq:MP2}
\end{equation}
Again, various strategies can be adopted for such an update step~\cite{hamilton2020graph,hamilton2017inductive}. In the above equations, $h^{(0)}_i$ corresponds to the input vector representation of node $i$, i.e., $h^{(0)}_i = x_i \in \mathbb{R}^f$ in the presence of node features. Also, the final $h^{(L)}_i \in \mathbb{R}^{d}$ will correspond to the actual node embedding vector eventually returned by this GNN model for node $i$, i.e.:
\begin{equation}
\forall i \in \mathcal{V}, z_i = h^{(L)}_i.
\end{equation}
By design, such a process is \textit{permutation equivariant} \cite{hamilton2020graph}. Intuitively, at each layer, a node $i$'s representation will aggregate structural and feature information from its direct neighbors (and itself), assuming $\mathcal{N}(i)$ corresponds to the one-hop neighborhood of $i$. As, at layer n°2, these neighbors will have already aggregated information from their own neighbors, then $h^{(2)}_i$ will indirectly incorporate structural and feature information from the 2-hop neighborhood of $i$, i.e., nodes reachable from a path of length 2 starting from $i$ in the graph. More generally, each $h^{(l)}_i$ will capture information up to the $l$-hop neighborhood of $i$ and, in particular, the ultimate $z_i$ vector will capture information up to the $L$-hop neighborhood. 

\subsection{A Popular GNN: Graph Convolutional Network (GCN)}
\label{c2s232}

Since the seminal models of Gori et al.~\cite{gori2005new}, Merkwirth and Langauer~\cite{merkwirth2005automatic} and Scarselli et al.~\cite{scarselli2008graph}, often praised as the first message-passing GNNs~\cite{hamilton2020graph}, several $\text{AGGREGATE}(\cdot)$ and $\text{UPDATE}(\cdot)$ functions have been proposed and compared in the scientific literature. We refer to relevant surveys \cite{hamilton2020graph,wu2019comprehensive,zhang2020deep} for an in-depth review. In this section, we focus on one of the most popular of these GNNs, that we will ourselves often implement in the experiments reported throughout this thesis: the \textit{graph convolutional network}~(GCN) from Kipf and Welling~\cite{kipf2016-1}.

To introduce the GCN model consistently with its formulation from Kipf and Welling~\cite{kipf2016-1}, we temporarily assume that $\mathcal{G}$ is \textit{undirected}. Consistently with the notation from Definition~\ref{defneighbors}, $\mathcal{N}(i)$ denotes the neighborhood of node $i \in \mathcal{V}$. For simplicity, we assume that $\mathcal{G}$ has no self-loop. From there, at each layer $l \in \{1,\dots, L\}$, Kipf and Welling~\cite{kipf2016-1} consider the following message passing operation:
\begin{equation}
\begin{split}
h^{(l)}_i &= f^{(l)} \Big( \sum_{j \in \mathcal{N}(i) \cup \{i\}} \tilde{A}_{ij} h^{(l-1)}_j \Big)\\
&= f^{(l)} \Big( \frac{1}{D_{ii} +1}h^{(l-1)}_i + \sum_{j \in \mathcal{N}(i)} \frac{A_{ij}}{\sqrt{(D_{ii} +1)(D_{jj} +1)}} h^{(l-1)}_j \Big).
\end{split}
\end{equation}
In the above equation, $f^{(l)}: \mathbb{R}^{d_{l-1}} \rightarrow \mathbb{R}^{d_{l}}$ is a parameterized function combining:
\begin{itemize}
    \item the multiplication\footnote{This operation also often involves the addition of a \textit{bias} term~\cite{goodfellow2016deep}, that we omit for the sake of clarity.} of hidden vectorial embedding representations by a trainable \textit{weight matrix} $W^{(l-1)} \in \mathbb{R}^{d_{l-1} \times d_l}$. Several strategies, detailed thereafter, can be adopted to \textit{tune} such a matrix, depending on the final application and on data availability; 
    \item the use of a \textit{non-linear activation function}  on top of the above multiplication, analogously to standard neural networks \cite{lecun2015deep}. Without loss of generality, Kipf and Welling~\cite{kipf2016-1} leverage a \textit{Rectified Linear Unit} (ReLU) function: $\text{ReLU}(x) = \max(x,0)$ \cite{goodfellow2016deep}. This non-linear activation is often omitted for the last layer \cite{kipf2016-2,kipf2016-1}.
\end{itemize}

Summing up these operations using a matrix-level notation, we end up with the following definition of (multi-layer) GCN models. 

\begin{definition}
A \textit{multi-layer graph convolutional network} (GCN), with $L \geq 2$ layers, is a function taking as input a symmetrically normalized adjacency matrix $\tilde{A} \in [0,1]^{n\times n}$ (as defined in Definition~\ref{def:norm_c2}), potentially equipped with a node feature matrix $X \in \mathbb{R}^{n \times f}$ (as defined in Definition~\ref{def:featurematrix}), and returning a node embedding matrix $Z \in \mathbb{R}^{n \times d}$ such as:
\begin{equation} 
\begin{cases}
H^{(0)} = X \text{ (or $I_n$ in the featureless case)} \\
H^{(l)} = \text{ReLU} (\tilde{A} H^{(l-1)} W^{(l-1)}), \hspace{5pt} \text{for } l \in \{1,...,L-1\} \\
Z = H^{(L)} = \tilde{A} H^{(L-1)} W^{(L-1)},
\end{cases}
\label{eq:gcn}
\end{equation}
where each $H^{(l)}$ denotes the $n \times d_l$ matrix stacking all $h^{(l)}_i$ vectors, i.e., the $i$-th row of $H^{(l)}$ corresponds to the hidden embedding vector of node $i$ at layer $l$. In particular, $d_0 = f$ (in the presence of node features) or $n$ (in the absence of node features), and $d_L = d$. Also, $W^{(0)} \in \mathbb{R}^{d_0 \times d_1},\ldots,W^{(L-1)} \in \mathbb{R}^{d_{L-1} \times d_L}$ are trainable weight matrices.
\label{def:gcn}
\end{definition}

Unless explicitly stated otherwise\footnote{A notable counterexample is the one from Chapters~\ref{chapter_5}, where we will adapt this message passing to \textit{directed} graphs. The symmetric normalization of $A$ will be replaced by \textit{out-degree normalization} $\tilde{A}_{\text{out}} = (D_{\text{out}} + I_n)^{-1} (A + I_n)$, so that each node averages representations from neighbors \textit{to which it points} through a directed edge, i.e., the outcoming neighbors. Another example is the one from Chapter~\ref{chapter_6} where we will consider the case $L =1$, leading to simpler linear models that are no longer \textit{multi-layer} GCNs.}, we will adopt this definition of the GCN in the remainder of the thesis. In a nutshell, at each layer $l$, the GCN computes a vectorial representation for each node $i\in \mathcal{V}$, by averaging the representations from layer $l-1$ of $i$'s direct neighbors and of $i$ itself. This averaging operation is composed with a linear transformation via the weight matrices and a ReLU activation. In their original work~\cite{kipf2016-1}, Kipf and Welling tune $W^{(0)},\dots, W^{(L-1)}$ in a (semi)-supervised fashion. Specifically, they iteratively minimize, by \textit{gradient descent}~\cite{goodfellow2016deep}, a cross-entropy classification loss. This loss compares the $z_i$ vectors transformed through a softmax function~\cite{goodfellow2016deep} to $d$ ground truth labels (article/node topics in citation networks in~\cite{kipf2016-1}). Nonetheless, this strategy requires the availability of such labels. In Section~\ref{c2s24}, we will fully detail how GCNs can alternatively be optimized in an \textit{unsupervised} fashion, when acting as encoders in the broader GAE and VGAE frameworks~\cite{kipf2020phd,kipf2016-2}.

Leveraging $\tilde{A}$ instead of $A$ in a GCN has two main advantages related to \textit{information propagation} in the graph. Firstly, using an unnormalized $A$ would lead to a \textit{summation} of vectorial representations which, over layers, would change the scale of these vectors, especially for very connected nodes. Moreover, in the absence of self-loops, multiplication by $A$ would imply a summation of vectors from all neighboring nodes \textit{but not from the node itself}, which is nonetheless important in practice~\cite{kipf2016-1,wu2019simplifying}. Using $\tilde{A}$ addresses this problem, thanks to the addition of the identity matrix $I_n$ in the normalization (see Definition~\ref{def:norm_c2}).
In the last few years, several studies such as the ones from Klicpera et al.~\cite{klicpera2019predict} and Wu et al.~\cite{wu2019simplifying} empirically confirmed that using $\tilde{A}$ for message passing leads to better performance than some alternatives. Wu et al.~\cite{wu2019simplifying} also proved that adding self-loops reduces the magnitude of the dominant eigenvalue of the corresponding Laplacian matrix: in essence, adding self-loops decreases the relative influence of distant nodes, which might help obtaining more stable representations and might be desirable for various applications. More recently, Dasoulas et al.~\cite{Dasoulas2021} proposed a \textit{parameterized} graph shift operator for GNN models such as GCNs. Specific parameter values of this operator result in the most commonly used
graph matrices. Authors manage to incorporate these parameters in the training of GNNs, which permits learning the optimal message passing matrix for some given data- and task-specific application.

Since its formulation by Kipf and Welling~\cite{kipf2016-1} five years ago, the multi-layer GCN architecture emerged as one of the most popular ones in the graph representation learning community, with various successful applications to graph data from various domains ranging from biology to web mining and social networks~\cite{hamilton2020graph,schlichtkrull2018modeling,xu2018representation,zhang2020deep,zhou2020graph}. As we will develop in Section~\ref{c2s24}, GCNs also play a central role in the GAE and VGAE models proposed by Kipf and Welling~\cite{kipf2020phd,kipf2016-2} (see Section~\ref{c2s24}) and in numerous of their extensions \cite{do2019matrix,grover2019graphite,semiimplicit2019,pan2018arga,aaai20,huang2019rwr}. Besides these experimental successes, the popularity of GCN models can also be explained by their relative \textit{simplicity} w.r.t. other GNNs, notably the ones directly based on \textit{spectral graph convolutions} \cite{bruna2013spectral,defferrard2016} discussed in Section~\ref{c2s233}. Also, the evaluation of each GCN layer has a linear time complexity w.r.t. the number of edges $m$ in the graph when leveraging sparse matrix-level operations \cite{kipf2016-1}, which is relatively low compared to several other models.
Lastly, as we will further explain and empirically show in Chapters~\ref{chapter_8}~and~\ref{chapter_9}, GCN models can easily be leveraged in \textit{inductive} settings when node features are available~\cite{salha2021cold}.

\subsection{Theoretical Considerations}
\label{c2s233}

An active area of research in graph representation learning revolves around the study of potential theoretical guarantees for existing GNN models such as GCNs, as well as the development of new theoretically grounded models. As explained by Hamilton~\cite{hamilton2020graph}, GNNs independently emerged from three distinct theoretical motivations.

First and foremost, several GNN models were (and are still) explicitly conceived from the \textit{graph signal processing} theory, generalizing the notion of convolution in a Euclidean space to graphs~\cite{bruna2013spectral}. In parallel, other research studies motivated the GNN approach from its relation to Weisfeiler-Lehman \textit{graph isomorphism} tests~\cite{hamilton2017inductive,shervashidze2011weisfeiler,weisfeiler1968reduction}. Lastly, analogies were also made between neural message passing and inference in \textit{probabilistic graphical
models}~\cite{dai2016discriminative,hamilton2020graph}.

In this section, we provide a brief overview of the first of these aspects, i.e., graph signal processing. We refer to Hamilton~\cite{hamilton2020graph} for a review of the two other ones, that we omit in this thesis. While they are relevant and linked to active areas of research~\cite{maron2019provably,morris2019weisfeiler,murphy2019relational,xu2019powerful}, they are less connected to the contributions presented~in~the~next~chapters.

\paragraph{GNN Models and Spectral Graph Convolutions} 
Let us consider the orthogonal \textit{spectral decomposition} of the symmetric message passing operator $\tilde{A}$ used in Equation~\eqref{eq:gcn},
\begin{equation}
 \tilde{A} = U\Lambda U^T,
\end{equation}
where $\Lambda$ denotes the diagonal matrix containing the eigenvalues $\lambda_i$ of $\tilde{A}$ (assumed to be ordered), and where $U=[u_1, \dots, u_n]^T$ denotes the corresponding orthogonal matrix containing the eigenvectors $u_i$ of $\tilde{A}$. Then, the computation performed in Equation~\eqref{eq:gcn} can be~reformulated~as:
\begin{equation} \label{eq:GCN_spectral}
    \text{ReLU} (U\Lambda U^T H^{(l-1)} W^{(l-1)}) = \text{ReLU} \left(\sum_{i=1}^n \lambda_i u_iu_i^T H^{(l-1)} W^{(l-1)}\right).
\end{equation}
Therefore, performing one message passing of hidden representations on the graph $\mathcal{G}$ defined by $\tilde{A},$ i.e., $\tilde{A} H^{(l-1)},$ can be interpreted as a \textit{Fourier transform} of these hidden representations called \textit{graph Fourier transform} \cite{Shuman2013}. The eigenvectors of $\tilde{A}$ act as a Fourier basis and the eigenvalues of $\tilde{A}$ define the Fourier coefficients. From such a setting, \textit{spectral graph convolutions} are defined as element-wise product operations in this Fourier space.

When trying to perform a theoretical analysis of neural message passing steps in GNN models such as GCNs, it often turns out to be more insightful to instead consider the above decomposition~\cite{salhagalvan2022modularity}, and therefore analyze eigenvalues and eigenvectors of the corresponding message passing operator (see, e.g., Chapter~\ref{chapter_7}). Historically, the study of spectral graph theory \cite{Chung1997,spielman2007spectral}, and in particular the area of graph signal processing \cite{Ortega2018,Sandryhaila2014}, has yielded insights in the study of graphs. Therefore it is unsurprising that, in the study of the GNNs, the spectral analysis of these architectures appears as a promising avenue of research as well \cite{Balcilar2021,Dasoulas2021,gama2020}. 

Such a spectral perspective has 
also given rise to a variety of architectures directly relying on graph signal processing, e.g., by proposing convolutional operations a.k.a. \textit{filters} based on learnable functions applied to the diagonal terms of $\Lambda$. This includes the seminal model of Bruna~et~al.~\cite{bruna2013spectral} and several more recent ones~\cite{defferrard2016,henaff2015deep,Levie2019}. In particular, as this spectral approach is computationally costly\footnote{Filters are directly defined in the Fourier space, and computing the graph Fourier transform involves $O(n^2)$ operations. As already explained, computing the full eigendecomposition of an adjacency or a Laplacian matrix also itself suffers from an $O(n^3)$ complexity, when using a baseline implementation.}, Defferrard et al.~\cite{defferrard2016} proposed to approximate smooth filters in the spectral domain using Chebyshev polynomials~\cite{hammond2011wavelets}. 
Kipf and Welling~\cite{kipf2016-1} themselves mathematically derived their GCN architecture as a faster and localized first-order approximation of spectral graph convolutions, bringing further simplifications over Defferrard et al.~\cite{defferrard2016}. Their analysis brings some theoretical foundation and motivation to the relatively intuitive message passing strategy described in Section~\ref{c2s232}.

\subsection{Recent Advances in Representation Learning with GNN Models}
\label{c2s234}

As mentioned in our introduction to node embedding methods in Section~\ref{c2s22}, the graph representation learning field has grown at a very fast pace over the past few years. This is especially true for the sub-field involving graph neural networks, to such an extent that it becomes increasingly difficult to keep track of all advances from the research community. Before introducing GAE and VGAE models in the next Section~\ref{c2s24} and focusing on them in most of this thesis, this section will nonetheless try to provide a brief summary of some of the most notable other advances related to representation learning with GNNs. 

Besides advances on provably powerful GNN models, e.g., based on the aforementioned spectral analyses or connections to Weisfeiler-Lehman tests~\cite{hamilton2020graph,maron2019provably,morris2019weisfeiler,murphy2019relational,xu2019powerful}, numerous research studies proposed novel message passing GNN architectures for various applications. Among the most impactful ones, GraphSAGE models from Hamilton~et~al.~\cite{hamilton2017inductive} incorporate  generalized neighborhood aggregation schemes, suitable for inductive representation learning. 
Graph attention networks (GAT) from Veli{\v{c}}kovi{\'c}~et~al.~\cite{velivckovic2019graph} leverage \textit{attention} mechanisms during message passing, permitting weighting the influence of each neighbor during information aggregation. Graph diffusion convolutions (GDC) from Klicpera~et~al.~\cite{klicpera2019diffusion} resort to more general \textit{diffusion} functions to identify which nodes one should aggregate information from, which is useful in the presence of graphs with noisy or arbitrarily defined edges. Jumping knowledge (JK) networks from Xu~et~al.~\cite{xu2018representation} use representations from all hidden layers, and not only the last one, to compute embedding vectors. Graph Transformers from Dwivedi and Bresson~\cite{dwivedi2021generalization} generalize transformer neural networks~\cite{wolf2020transformers} to graphs. Several other models were additionally designed for specific structures such as knowledge graphs~\cite{schlichtkrull2018modeling} or dynamic graphs~\cite{kazemi2020representation}, and recent works also introduced GNN models \textit{without} message passing procedures~\cite{hu2021graph,wu2019simplifying}.

As large graphs with millions (or even billions) or nodes and edges are ubiquitous~\cite{salha2021fastgae}, several studies also aimed to scale GNN models, typically by using mini-batch sampling or other various approximate learning strategies. This includes the recent FastGCN~\cite{chen2018fastgcn}, Cluster-GCN~\cite{chiang2019cluster} and GraphSAINT~\cite{zeng2020graphsaint} models, among others \cite{chen2018stochastic,hamilton2017inductive,ying2018graph}. Moreover, while research on GNN models is often referred to as ``\textit{deep learning on graphs}''~\cite{zhang2020deep}, we note that the most effective architectures actually leverage a few layers at most, contrary to other fields such as computer vision~\cite{lecun2015deep}. This is explained by the standard challenges related to training deep architectures, such as \textit{vanishing gradients} in backpropagation~\cite{goodfellow2016deep}, but also by the nature of real-world graph data (numerous graphs have a \textit{small-world} structure~\cite{adamic1999small}, i.e., a few hops/layers already permit reaching any node from another one) and other graph-specific problems such as \textit{over-smoothing} and \textit{over-squashing} (see~\cite{alon2021bottleneck,li2018deeper}). While some works proposed techniques to train deeper GNNs, e.g., based on regularization~\cite{rong2020dropedge,zhao2020pairnorm} or residual connections~\cite{gong2020geometrically,xu2018representation}, they often fail to outperform models with a few layers~\cite{shchur2018pitfalls}. On the contrary, several other studies tend to show that \textit{simpler} models can achieve competitive empirical performances. This includes the one of Wu~et~al.~\cite{wu2019simplifying} who introduced simple graph convolution (SGC), a variant of GCN removing nonlinearities and collapsing weight matrices. To finish, we point out that recent articles~\cite{hu2021ogb,hu2020open,shchur2018pitfalls}, including the one behind the Open Graph Benchmark initiative~\cite{hu2020open}, also criticized the current evaluations of GNNs, pointing at discrepancies in experimental procedures and proposing new relevant tasks or datasets. These last three important aspects (scalability, simplification, better evaluation) will also be at the center of several of our studies on~GAE~and~VGAE~models. 

\section{Representation Learning with Graph Autoencoders}
\label{c2s24}

So far, we presented some general concepts and definitions related to graph representation learning. In this Section~\ref{c2s24}, we now leverage this background to formally introduce the GNN-based models that will be at the center of our research: \textit{graph autoencoders}. In its most general formulation, this term can actually refer to \textit{two} families of models, learning node embedding representation from graph data~\cite{kipf2020phd,kipf2016-2,salha2020simple,tian2014learning,wang2016structural}:
\begin{itemize}
    \item the \textit{deterministic graph autoencoders}, often simply referred to as graph autoencoders (GAEs) in the following when there is no ambiguity, and presented in Section~\ref{c2s241};
    \item the \textit{variational graph autoencoders} (VGAEs), presented in Section~\ref{c2s242}.
\end{itemize}
They both rely on an \textit{encoding-decoding strategy} that, in a broad sense, consists of \textit{encoding} nodes into an embedding space from which \textit{decoding}, i.e., reconstructing the original graph should ideally be possible, by
leveraging either a deterministic (for a GAE) or a probabilistic (for a VGAE) approach. Intuitively, the ability to accurately reconstruct a graph from a node embedding space indicates that this space preserves some important information from the graph structure. 
As explained in the introduction, GAE and VGAE models are specifically suited for representation learning on graphs \textit{in the absence of node labels}, i.e., in an unsupervised fashion. Simultaneously, GAE and VGAE models can process attributed graphs. As VGAEs emerged as effective alternatives to GAEs in several previous works described throughout this thesis, we see value in considering both variants of graph autoencoders in our work. 

\subsection{Deterministic Graph Autoencoder (GAE)}
\label{c2s241}

Albeit under various formulations, the encoding-decoding strategy has been widely adopted over the last years to learn embedding spaces in the absence of node labels \cite{kipf2020phd,kipf2016-2,tian2014learning,wang2017mgae,wang2016structural}. In this thesis, we follow the formulation of Kipf and Welling~\cite{kipf2016-2}. While it is not the oldest article, their work is explicitly mentioned as the seminal reference in the majority of recent advances involving graph autoencoders, including \cite{choong2018learning,choong2020optimizing,do2021improving,grover2019graphite,semiimplicit2019,pei2021generalization,li2021mask,li2020dirichlet,pan2018arga,aaai20,huang2019rwr}. 

\begin{figure}[t]
    \centering
\resizebox{1.0\textwidth}{!}{
    \tikzstyle{block} = [draw, fill=ImperialColor, rectangle, 
    minimum height=4em, minimum width=8em]
\tikzstyle{sum} = [draw, fill=ImperialColor, circle, node distance=1cm]
\tikzstyle{input} = [coordinate]
\tikzstyle{output} = [coordinate]
\tikzstyle{pinstyle} = [pin edge={to-,thin,black}]
\begin{tikzpicture}[auto, node distance=8cm,>=latex']
    \node  (A) {$A, X$};
    \node [block, right of=A, node distance=3cm, align=center] (encoder){\textcolor{white}{GCN} \\ \textcolor{white}{encoder}};
    \node [node distance=3cm, right of = encoder] (Z) {$Z$};
    \node [block, right of=Z, node distance=3cm, align=center] (decoder) {\textcolor{white}{Inner product} \\ \textcolor{white}{decoder}};
    \node [node distance=3cm, right of = decoder] (Arec) {$\hat{A}$};
    \draw [->] (A) -- (encoder);
    \draw [->] (encoder) -- (Z);
    \draw [->] (Z) -- (decoder);
    \draw [->] (decoder) -- (Arec);
\end{tikzpicture}
}
\vspace{0.5cm}
\resizebox{1.0\textwidth}{!}{
    \begin{minipage}{.15\textwidth}
    		\includegraphics[width=1.0\textwidth]{figures/Chapter2/sbmok.pdf}
    \end{minipage}
\hspace{2.5cm}
\begin{minipage}{.15\textwidth}
				\includegraphics[width=1.0\textwidth]{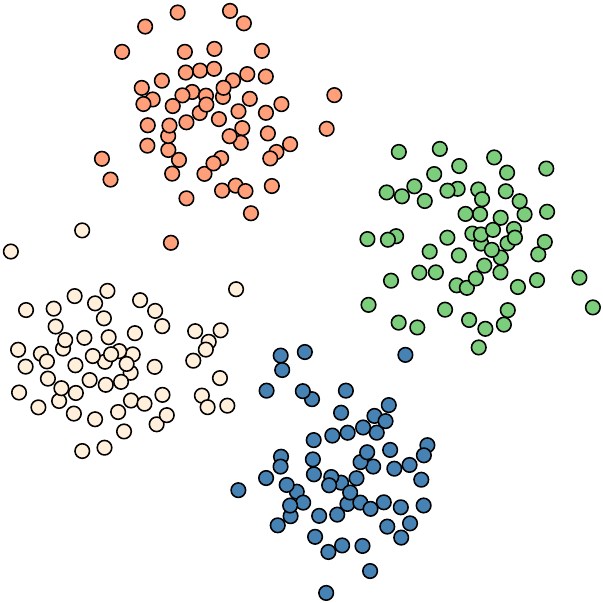}
	 \end{minipage}
	\hspace{2cm}
	\begin{minipage}{.15\textwidth}
				\includegraphics[width=1.0\textwidth]{figures/Chapter2/sbmok.pdf}
 \end{minipage}}
    \caption[Schematic representation of a GAE]{Schematic representation of a GAE model, as formulated by Kipf and Welling~\cite{kipf2016-2}.}
    \label{fig:c2_gae}
\end{figure}
 Graph autoencoders involve the combination of two building blocks: an \textit{encoder} and a \textit{decoder}. The above Figure~\ref{fig:c2_gae} provides an illustration of a GAE model.

\paragraph{Encoder} In its most general formulation, the first of these two components, i.e., the \textit{encoder}, is defined as follows.

\begin{definition}
An \textit{encoder} is a function processing the adjacency matrix $A$ and the node feature\footnote{As in Section~\ref{c2s23}, we simply set $X = I_n$ when dealing with featureless graphs.} matrix $X$ of a graph $\mathcal{G} = (\mathcal{V},\mathcal{E})$, and mapping each node $i \in \mathcal{V}$ from $\mathcal{G}$ to a low-dimensional embedding vector $z_i \in \mathbb{R}^d$ with $d \ll n$. Adopting the notation from Definition~\ref{def:embedding}, we have:
\begin{equation}
Z = \text{Encoder}(A,X).    
\end{equation}
\label{def:encoder}
\end{definition}

 In practice, a GNN (with weights optimized using the procedure described thereafter) often acts as the encoder in GAE models. In particular, Kipf and Welling \cite{kipf2016-2} leverage a multi-layer GCN~\cite{kipf2016-1}, as defined in Definition~\ref{def:gcn}, to encode nodes, i.e.:
\begin{equation} 
Z = \text{GCN}(A,X).    
\label{eq:gcn_AE}
\end{equation} 

To this day, multi-layer GCNs remain the most  popular encoders in GAE extensions building upon Kipf and Welling \cite{kipf2016-2}, including  \cite{choong2018learning,grover2019graphite,semiimplicit2019,pei2021generalization,pan2018arga,salha2021fastgae,salha2019-2,aaai20,huang2019rwr}, mainly thanks to the relative simplicity and reduced complexity of these models w.r.t. several GNN alternatives (see Section~\ref{c2s232}). Nonetheless, they can be replaced by various alternatives, including by faster \cite{chen2018fastgcn,chiang2019cluster,hamilton2017inductive}, by more sophisticated \cite{bruna2013spectral,defferrard2016,velivckovic2019graph} or, on the contrary, by simpler \cite{choong2020optimizing,salha2020simple} models.

\paragraph{Decoder} 
The second component of a GAE, i.e., the \textit{decoder}, aims to reconstruct an $n \times n$ adjacency matrix $\hat{A}$, estimated from the learned embedding vectors. It is defined as follows.

\begin{definition}
Assuming an $n \times d$ node embedding matrix $Z$ stacking up node embedding vectors $z_i \in \mathbb{R}^d$ for nodes of a graph $\mathcal{G} = (\mathcal{V},\mathcal{E})$, a \textit{decoder} is a function reconstructing an $n \times n$ estimated adjacency matrix $\hat{A}$ from $Z$:
\begin{equation}
\hat{A} = \text{Decoder}(Z).    
\end{equation}
\end{definition}

While another neural network could act as a decoder \cite{li2020graph,park2019symmetric,wang2016structural}, Kipf and Welling \cite{kipf2016-2} and most of the aforementioned extensions rely on simpler \textit{inner~product} decoders:
\begin{equation}
\hat{A} = \sigma(ZZ^T),
\end{equation}
where $\sigma$ denotes the \textit{sigmoid} function: $\sigma(x) = 1/(1 + e^{-x})$. Therefore, for all node pairs $(i,j) \in \mathcal{V}\times\mathcal{V}$, we have:
\begin{equation}
\hat{A}_{ij} = \sigma (z^T_i z_j) \in~]0, 1[.
\end{equation}
In such a setting, a large and positive inner~product $z^T_i z_j$ in the node embedding space indicates the likely presence of an edge between nodes $i$ and $j$ in $\mathcal{G}$, according to the model. Again, the choice of inner~product decoders is not restrictive, and recent efforts considered replacing them with alternatives verifying some desirable properties such as the ability to capture triads structures \cite{aaai20}, to simultaneously reconstruct node features \cite{sun2021dual}, or to reconstruct biologically plausible graphs in the case of autoencoders for molecular structures \cite{molecule1,simonovsky2018graphvae}.

\paragraph{Optimization}

We recall that GAE models aim to learn node embedding spaces from which one can accurately reconstruct graphs. The intuition behind this strategy is the following: if, starting from $Z$, one can reconstruct a graph close to the true one, i.e., $\hat{A} \approx A$, then these embedding vectors should manage to preserve some important characteristics of the initial graph structure, and should therefore be useful to perform downstream tasks such as link prediction.

Consequently, GAEs are trained to minimize \textit{reconstruction losses}, which specifically evaluate the similarity between the decoded adjacency matrix $\hat{A}$ and the original one $A$. For instance, Kipf and Welling \cite{kipf2016-2} tune weight matrices of their GCN encoders by iteratively minimizing, by gradient descent\footnote{Specifically, Kipf and Welling~\cite{kipf2016-2} use the \textit{Adam} algorithm, a prevalent method for first-order gradient-based optimization in neural networks architectures, fully described in the original work of Kingma~and~Ba~\cite{kingma2014adam}.} \cite{goodfellow2016deep}, the following \textit{cross-entropy} loss:
\begin{align}
\mathcal{L}_{\text{GAE}} = \frac{-1}{n^2}\sum_{(i,j) \in \mathcal{V}\times \mathcal{V}} \Big[A_{ij}\log(\hat{A}_{ij}) + (1-A_{ij})\log(1 - \hat{A}_{ij})\Big].
\label{lossGAE}
\end{align}
In the case of sparse graphs where unconnected node pairs significantly outnumber the connected ones, i.e., the graph's edges, it is common to reweight the ``positive terms'' in Equation~\eqref{lossGAE} by a factor $w_{\text{pos}} > 1$ \cite{kipf2016-2,salha2021fastgae,wang2016structural}. We note that an exact evaluation of $\mathcal{L}_{\text{GAE}}$ requires the reconstruction of the entire matrix $\hat{A}$, which suffers from a quadratic $O(dn^2)$ time complexity. Such a decoding approach is therefore unsuitable for graphs with more than a few thousand nodes. Scalability concerns will be thoroughly discussed and addressed in Chapters~\ref{chapter_3}~and~\ref{chapter_4}. Lastly, we note that this minimization scheme based on reconstruction losses permits optimizing a GCN, or any encoder, in an unsupervised fashion, i.e., without node-level ground truth labels (as in Section~\ref{c2s232}).  

\subsection{Variational Graph Autoencoder (VGAE)}
\label{c2s242}

Kipf and Welling \cite{kipf2016-2} also considered a probabilistic variant of GAE, extending the \textit{variational autoencoder} (VAE) from Kingma and Welling \cite{kingma2013vae}. Besides constituting generative models with promising applications to graph generation \cite{molecule3,molecule1,simonovsky2018graphvae} (see the next paragraphs), variants of \textit{variational graph autoencoder} (VGAE) models also turned out to be effective alternatives to GAE on several downstream applications such as link prediction or community detection tasks \cite{choong2020optimizing,semiimplicit2019,kipf2016-2,salha2019-2}. In the following, we briefly review key concepts related to VAE, and subsequently present the VGAE model.

\paragraph{Latent Models and VAE} Variational autoencoders are \textit{latent variable} models. In essence, these are probabilistic models explaining some observed variable $x \in \mathcal{X}$ through some unobserved latent variable $z \in \mathcal{Z}$. Denoting by $p_{\theta}(x)$ the distribution of $x$, parameterized by some $\theta \in \Theta$, we have:
\begin{equation}
p_{\theta}(x) = \int_{\mathcal{Z}} p_{\theta}(x|z)p(z)dz.
\label{eq:cond}
\end{equation}
The right-hand side of Equation~\eqref{eq:cond} introduces the conditional distribution $p_{\theta}(x|z)$, to explicitize the unobserved underlying role that $z$ plays in the generation of $x$. In such latent variable models, the objective is usually to estimate the parameters of  $p_{\theta}(x|z)$, referred to as the \textit{generative} model, which maximize $p_{\theta}(x)$ for some observed data and some known \textit{prior distribution} $p(z)$ on $z$. 
As explained by Kipf~\cite{kipf2020phd}, in many applications, one would like to use neural networks as generative models, e.g., in combination with a Gaussian distribution:
\begin{equation}
 p_{\theta}(x|z) = \mathcal{N}(x; \text{NN}(z),\Sigma),
\end{equation}
where this notation refers to the density of a Gaussian distribution, with a mean vector $\text{NN}(z)$ determined by a neural network ($\theta$ corresponds to its weights), and with a fixed variance matrix $\Sigma$. 
Unfortunately, in a large number of applications, this formulation of $p_{\theta}(x)$ will be \textit{intractable}\footnote{We note that this is not specific to neural networks. For instance, we refer to Blei et al. \cite{vistat} for a presentation of the \textit{Gaussian Mixture Model} (GMM), a famous example of latent variable model. In a GMM, $p_{\theta}(x)$ is intractable due to a sum over the latent variables in the integrand of $p_{\theta}(x)$.}. It will be impossible to optimize $\theta$ analytically, e.g., via maximum~likelihood~estimation~\cite{doersch2016tutorial}.

To address this issue, variational autoencoders (VAE) from Kingma~and~Welling~\cite{kingma2013vae} leverage concepts from \textit{variational inference}~\cite{doersch2016tutorial}, and consider the optimization of a \textit{tractable} lower bound of $p_{\theta}(x)$, referred to as the \textit{evidence lower bound} (ELBO). It involves an approximate \textit{posterior distribution} $q_{\phi}(z|x)$ with parameters $\phi \in \Phi$. Formally, using Jensen's inequality~\cite{jensen1906fonctions}, we have:
\begin{equation}
\log p_{\theta}(x) = \log \int_{\mathcal{Z}} \frac{q_{\phi}(z|x)}{q_{\phi}(z|x)} p_{\theta}(x|z)p(z)dz \geq  \int_{\mathcal{Z}} q_{\phi}(z|x) \log \frac{p_{\theta}(x|z)p(z)}{q_{\phi}(z|x)} dz.
\label{eq:elbo1}
\end{equation}
The right-hand side of Equation~\eqref{eq:elbo1} corresponds to the ELBO, that we reformulate~as~follows:
\begin{align}
\text{ELBO} &= \int_{\mathcal{Z}} q_{\phi}(z|x) \log \frac{p_{\theta}(x|z)p(z)}{q_{\phi}(z|x)} dz \nonumber \\
&= \int_{\mathcal{Z}} q_{\phi}(z|x) \log p_{\theta}(x|z) dz + \int_{\mathcal{Z}} q_{\phi}(z|x) \log \frac{p(z)}{q_{\phi}(z|x)} dz \nonumber \\
&= \int_{\mathcal{Z}} q_{\phi}(z|x) \log p_{\theta}(x|z) dz - \int_{\mathcal{Z}} q_{\phi}(z|x) \log \frac{q_{\phi}(z|x)}{p(z)} dz \nonumber \\
&= \mathbb{E}_{q_{\phi}(z|x)}\Big[ \log p_{\theta}(x|z) \Big] - \mathcal{D}_{\text{KL}}\Big(q_{\phi}(z|x) || p(z)\Big),
\end{align}
where $\mathbb{E}_{q}$ denotes the expectation under the distribution $q$ and $\mathcal{D}_{\text{KL}}$ denotes the Kullback-Leibler divergence~\cite{kullback1951information}. In the context of VAE models, neural networks often characterize $q_{\phi}(z|x)$, which is then referred to as the \textit{inference} model \cite{kingma2013vae}. For instance, assuming two neural networks $\text{NN}^{(\mu)}$ and $\text{NN}^{(\Sigma)}$ providing a mean vector and a variance matrix, one can set:
\begin{equation}
q_{\phi}(z|x) = \mathcal{N}(z; \text{NN}^{(\mu)}(x),\text{NN}^{(\Sigma)}(x)).
\label{eq:inference_example}
\end{equation}

Here, $\phi$ would correspond to neural weights. Parameters $\theta$ and $\phi$ are jointly optimized by maximizing the ELBO by gradient ascent, using some available data samples $x$ from a dataset. We underline that the choices of $q_{\phi}(z|x)$ and $p(z)$ are often driven by computational constraints. In this thesis, we will often assume that they are Gaussian distributions with different parameters. The Kullback-Leibler divergence between two Gaussian distributions has a closed form  (see Doersch~\cite{doersch2016tutorial}) which facilitates gradient computation. Also, gradients of the above expectation are usually estimated through Monte Carlo approximations~\cite{doersch2016tutorial,kingma2013vae}, and using a \textit{reparameterization trick}~\cite{kingma2013vae}.
This trick consists in positioning all sampling steps from the VAE model in the input layer, to avoid backpropagating errors through a layer that samples $z$ vectors from $q_{\phi}(z|x)$, which is a non-continuous operation and has no gradient. For instance, let us assume the inference model from Equation~\eqref{eq:inference_example}. Instead of sampling $z \sim \mathcal{N}(\text{NN}^{(\mu)}(x),\text{NN}^{(\Sigma)}(x))$, one can sample $\varepsilon \sim \mathcal{N}(0,I)$ in the input layer, and subsequently compute:
\begin{equation}
z = \text{NN}^{(\mu)}(x) + \text{NN}^{(\Sigma)}(x)^{1/2} \times \varepsilon.
\end{equation}

Once parameters of the VAE are optimized, the inference model can infer latent variables $z$ of new data samples $x$. Analogously to previous sections, these latent variables will often act as \textit{embedding vectors} summarizing input data. Simultaneously, the generative model can be used to generate data from a latent variable $z$, e.g., new unseen data samples drawn from the prior distribution $p(z)$. For instance, VAE models have been successfully transposed to image generation problems~\cite{doersch2016tutorial,oussidi2018deep}. We refer to the comprehensive tutorial of~Doersch~\cite{doersch2016tutorial} for a broader introduction to VAE models and a discussion of their foundations~and~their~applications.

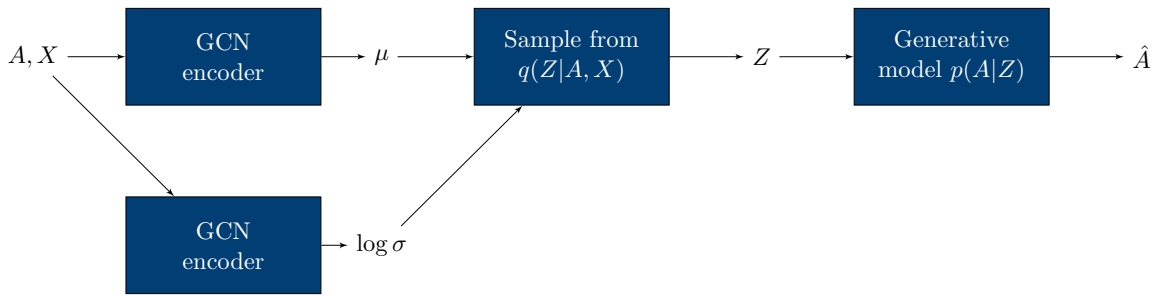
\begin{figure}[t]
    \centering
\resizebox{1.0\textwidth}{!}{
    \tikzstyle{block} = [draw, fill=ImperialColor, rectangle, 
    minimum height=4em, minimum width=8em]
\tikzstyle{sum} = [draw, fill=ImperialColor, circle, node distance=1cm]
\tikzstyle{input} = [coordinate]
\tikzstyle{output} = [coordinate]
\tikzstyle{pinstyle} = [pin edge={to-,thin,black}]
\begin{tikzpicture}[auto, node distance=2cm,>=latex']
    \node  (A) {$A, X$};
    \node [block, right of=A, node distance=3cm, align=center] (encoder) {\textcolor{white}{GCN} \\ \textcolor{white}{encoder}};
    \node [block, right of=A, below of=A, node distance=3cm, align=center] (encoder2) {\textcolor{white}{GCN} \\ \textcolor{white}{encoder}};
    \node [node distance=2.5cm, right of = encoder] (mu) {$\mu$};
    \node [node distance=3cm, below of = mu] (sigma) {$\log \sigma$};
    \node [block, right of=mu, node distance=3cm, align=center] (sampler) {\textcolor{white}{Sample from}\\ \textcolor{white}{$q(Z|A, X)$}};
    \node [node distance=3cm, right of = sampler] (Z) {$Z$};
    \node [block, node distance=3cm, right of = Z, align=center] (decoder) {\textcolor{white}{Generative} \\ \textcolor{white}{model} \textcolor{white}{$p(A|Z)$}};
    \node [node distance=3cm, right of = decoder] (Ahat) {$\hat{A}$};

    \draw [->] (A) -- (encoder);
    \draw [->] (A) -- (encoder2);
    \draw [->] (encoder) -- (mu);
    \draw [->] (encoder2) -- (sigma);
    \draw [->] (mu) -- (sampler);
    \draw [->] (sigma) -- (sampler);
    \draw [->] (sampler) -- (Z);
    \draw [->] (Z) -- (decoder);
    \draw [->] (decoder) -- (Ahat);
\end{tikzpicture}}
    \caption[Schematic representation of a VGAE]{Schematic representation of a VGAE model, as formulated by Kipf and Welling~\cite{kipf2016-2}.}
    \label{fig:c2_vgae}
\end{figure}

\paragraph{From VAE to VGAE} As explained at the beginning of Section~\ref{c2s242}, Kipf and Welling~\cite{kipf2016-2} extended the VAE  framework to graph structures. Their \textit{variational graph autoencoder} (VGAE) provides an alternative strategy to learn node embedding vectors in an unsupervised fashion, assuming that these vectors are drawn from specific distributions. As standard VAE models, a VGAE incorporates an inference model and a generative model. In the following, by analogy with the GAE approach, we will also often refer to these components as the VGAE's encoder and decoder, respectively. The above Figure~\ref{fig:c2_vgae} provides an illustration of a VGAE model. 

\paragraph{Encoder}

In their work, Kipf and Welling \cite{kipf2016-2} assume that each embedding vector $z_i \in \mathbb{R}^d$ corresponds to the \textit{latent} vector of a node $i \in \mathcal{V}$, following the VAE paradigm. This vector is a sample drawn from a $d$-dimensional Gaussian distribution, with mean vector $\mu_i \in \mathbb{R}^d$ and variance matrix $\Sigma_i = \text{diag}(\sigma_i^2)  \in \mathbb{R}^{d \times d}$ (with $\sigma_i \in \mathbb{R}^d$). They rely on \textit{two encoders} to learn these parameters. Denoting the $n \times d$ matrices stacking up the $d$-dimensional mean
and (log)-variance vectors for each node by $\mu$ and by $\log \sigma$\footnote{We allow a slight clash of notation here (as $\sigma$ also denotes the sigmoid activation function), for consistency with the commonly used notation from the literature (see, e.g., \cite{kipf2016-2}), and due to very low risks of confusion.}, respectively, they set:
\begin{equation} \mu = \text{Encoder}_{\mu}(A,X) \text{ and } \log \sigma = \text{Encoder}_{\sigma}(A,X).
\label{vaeencoder}
\end{equation}
As is the case for GAEs, multi-layer GCNs often act as encoders:
\begin{equation}
\mu = \text{GCN}_{\mu}(A,X) \text{ and } \log \sigma = \text{GCN}_{\sigma}(A,X).
\end{equation}Then, they adopt a \textit{mean-field inference model} for $Z$  \cite{kipf2016-2}. With  $\mathcal{N}(\cdot|\mu_i, \text{diag}(\sigma_i^2))$ denoting the density of a Gaussian distribution with mean vector $\mu_i$ and variance matrix $\text{diag}(\sigma_i^2)$, we have:
\begin{align} \label{eqn:distnZ}
q(Z|A,X) = \prod_{i=1}^n q(z_i|A,X), \text{ where } q(z_i|A,X) = \mathcal{N}(z_i|\mu_i, \text{diag}(\sigma_i^2)),
\end{align}

\paragraph{Decoder}

In the VGAE setting, the actual embedding vectors $z_i$ are sampled  from the above normal distributions. From such embedding representations, VGAE models then require a \textit{generative} model $p(A | Z,X)$, to act as a graph \textit{decoder}. As for GAE, Kipf and Welling~\cite{kipf2016-2} rely on simple inner~products together with sigmoid activation functions to reconstruct edges: 
\begin{align} \hat{A}_{ij} = p(A_{ij} = 1 | z_i, z_j) = \sigma(z_i^Tz_j),
\end{align}
where the embedding vectors $z_i, z_j$ are sampled from the distribution in Equation \eqnref{eqn:distnZ}.
Then, the authors assume the following generative model which factorizes over the edges, and is conditionally independent of $X$ for simplicity~\cite{kipf2020phd}:
\begin{align}p(A | Z,X) = \prod\limits_{i=1}^n  \prod\limits_{j=1}^n  p(A_{ij} | z_i, z_j).\end{align}

\paragraph{Optimization}

During training, and similarly to VAE models \cite{kingma2013vae}, Kipf and Welling~\cite{kipf2016-2} iteratively maximize the tractable \textit{evidence lower bound} (ELBO) \cite{kingma2013vae} of the model's likelihood, written as follows in the context of a VGAE:
\begin{align}\mathcal{L}_{\text{VGAE}} = \mathbb{E}_{q(Z | A,X)} \Big[\log
p(A | Z,X)\Big] - \mathcal{D}_{KL}\Big(q(Z | A,X)||p(Z)\Big).
\label{elbo}
\end{align}
This ELBO is iteratively maximized w.r.t. weights of the two GCN encoders, by gradient ascent. In the above Equation (\ref{elbo}), $p(Z)$ corresponds to a unit Gaussian  
prior (i.e., $\mathcal{N}(0,I_d)$) on the distribution of the latent vectors, that can also be interpreted as a regularization term on the magnitude of the embedding vectors~\cite{grover2019graphite,salhagalvan2022modularity}. As is the case for $\mathcal{L}_{\text{GAE}}$ from Equation~\eqref{lossGAE}, an exact evaluation of the ELBO suffers from an $O(dn^2)$ time complexity.

\subsection{Applications and Limitations}
\label{c2s243}

The question of how to properly determine the quality of node embedding representations learned from GAE and VGAE models is crucial. While one could directly report reconstruction losses~\cite{wang2016structural}, recent research articles instead strove to apply the GAE and VGAE models to downstream evaluation tasks, which permits reporting more insightful metrics~\cite{kipf2016-2,tian2014learning,wang2016structural}.
In particular, Kipf and Welling \cite{kipf2016-2} originally evaluated their GAE and VGAE models on \textit{link~prediction} problems in three popular citation networks (Cora, Citeseer, and Pubmed)~\cite{sen2008collective}, adopting the train/validation/test methodology described in Section~\ref{c2s212}. In their experiments, GAEs and VGAEs reached competitive link prediction scores w.r.t. DeepWalk \cite{perozzi2014deepwalk} and Laplacian eigenmaps \cite{von2007tutorial}, that we presented in Section~\ref{c2s22}. They also recalled an additional benefit of the GAE and VGAE models over these baselines, which is the ability to leverage both the graph structure and node features when learning embedding spaces. 

Over the last five few years, the overall effectiveness of the GAE and VGAE paradigms at addressing link prediction has been widely confirmed experimentally \cite{berg2018matrixcomp,do2021improving,grover2019graphite,ijcai2020DEAL,semiimplicit2019,pei2021generalization,pan2018arga,rennard2020graph,salha2021fastgae,salha2020simple,salha2019-2,aaai20,tran2018multi,huang2019rwr}. Numerous research efforts proposed and evaluated variants of GAEs and VGAEs designed for this specific task, improving their performances by considering more refined encoders \cite{ijcai2020DEAL,semiimplicit2019,li2020dirichlet,salha2019-1,wu2021deepened}, decoders \cite{grover2019graphite,semiimplicit2019,salha2019-2,aaai20,sun2021dual} or regularization techniques \cite{pei2021generalization,pan2018arga,huang2019rwr}. Most of these extensions were actually published during the time of this PhD thesis. The exact technical contribution of several of them will be further detailed in this thesis, when used as baselines for our own proposed methods.

Other research studies on GAE and VGAE models successfully addressed different downstream tasks that are closely related to link prediction, such as edge classification \cite{rennard2020graph} or graph-based recommendation \cite{berg2018matrixcomp,ijcai2020DEAL,zhang2021find}. GAE and VGAE models have also be applied to (semi-supervised) node classification \cite{semiimplicit2019,tran2018multi}, canonical correlation analysis \cite{kaloga2021multiview} and to community detection~\cite{choong2018learning,choong2020optimizing,guo2021end,li2021mask,mrabah2021rethinking,wang2017mgae}. Last, but not least, researchers were also interested in leveraging VGAEs as generative models, especially to generate molecular graph data. We refer for instance to \cite{molecule3,molecule1,molecule2,simonovsky2018graphvae} for recent advances in graph generation with VGAE models. This aspect will \textit{not} be at the center of our work in this thesis; we will instead mainly focus on the \textit{inference} component of VGAEs. As already explained, VGAE models emerged as effective  alternatives to deterministic GAEs on several downstream applications in some of the above references, which we will further analyze in the next chapters. Consequently, in this thesis, we saw value in considering both GAE and VGAE models to learn node embedding spaces.

Despite these promising advances, at the beginning of this PhD, i.e., in 2018, transposing these recent advances to industrial-level applications, e.g., at Deezer, was still a challenging task. As mentioned in this section, the standard GAE and VGAE methods suffer from scalability issues. We argue in the next chapters that standard techniques from deep learning, such as mini-batch sampling, often fail to provide satisfying solutions to this problem \cite{salha2021fastgae,salha2019-1}. As a consequence, in 2018, experiments on GAEs and VGAEs were limited to graphs with at most a few thousand nodes and edges. Among other challenges, standard GAE and VGAE models were designed for undirected and static graphs \cite{kipf2016-2}, and often neglected simple but effective encoding schemes~\cite{salha2020simple}. Recent studies \cite{choong2018learning,choong2020optimizing} also emphasized their relative limitations on community detection applications. 
The next part of this PhD thesis will present our contributions to improve GAEs and VGAEs and facilitate their application to real-world problems.

\part{Contributions to Representation Learning with Graph Autoencoders}
\label{partII}

\chapter[A Degeneracy Framework for Scalable Graph Autoencoders]{A Degeneracy Framework for Scalable~Graph~Autoencoders}\label{chapter_3}
\chaptermark{A Degeneracy Framework for Scalable Graph Autoencoders}

\textit{This chapter presents research conducted with Romain Hennequin, Viet-Anh Tran, and Michalis Vazirgiannis, and published in the proceedings of the 28\up{th} International Joint Conference on Artificial Intelligence (IJCAI 2019)~\cite{salha2019-1}.}

\section{Introduction}
\label{c3s31}

We begin this Part~\ref{partII} of the thesis, which presents our contributions to representation learning with GAE and VGAE models, with two chapters fully dedicated to \textit{scalability} concerns. Indeed, while these models emerged as powerful node embedding methods, they also suffer from scalability issues, preventing them to be applied to large graphs with millions of nodes and edges such as those available at Deezer.
More precisely, we identify two sources of complexity:
\begin{itemize}
    \item firstly, as explained in Section~\ref{c2s241}, GNN models often act as GAE/VGAE \textit{encoders}. The encoding step can therefore be computationally costly if the GNN models under consideration themselves involve complex operations. Nonetheless, at the time of this work, several scalable GNNs had already been proposed in the scientific literature. To this day, GCNs remain the most popular encoders for GAE and VGAE models. As explained in Section~\ref{c2s232}, the cost of evaluating each layer of a GCN evolves linearly w.r.t. the number of edges~$m$~\cite{kipf2016-1}. This can be improved by instead encoding nodes with a FastGCN \cite{chen2018fastgcn}, with a Cluster-GCN \cite{chiang2019cluster}, or by using simpler graph convolutions \cite{wu2019simplifying} or other recently proposed stochastic strategies \cite{chen2018stochastic,hamilton2017inductive,ying2018graph,zeng2020graphsaint} for improved scalability;
    \item moreover, and more importantly, the \textit{decoding} step of standard GAE and VGAE models usually suffers from a high computational complexity. In particular, Kipf and Welling~\cite{kipf2016-2} and numerous extensions of their work leverage inner product decoders $\hat{A} = \sigma(Z Z^T)$. They require the multiplication of the dense matrices $Z$ and $Z^T$, to compute reconstruction losses at each training iteration. As explained in Section~\ref{c2s241}, this decoding scheme has an $O(dn^2)$ complexity, as most of the alternative decoders from the literature, which also require computing inner products or Euclidean distances~\cite{grover2019graphite,salha2019-2,aaai20}. Storing $n \times n$ dense matrices $\hat{A}$ can also potentially lead to memory issues for a large $n$. 
    Some existing decoders are even more complex: for instance, the VGAE model of Simonovsky~and~Komodakis~\cite{simonovsky2018graphvae} includes a graph matching step with an $O(n^4)$ complexity\footnote{This is nonetheless acceptable in their work, as they focus on small molecular graphs with fewer than 40 nodes~\cite{simonovsky2018graphvae}.}.
    As a consequence, the aforementioned efforts to scale GNNs (that were achieved in a supervised setting, and out of the wider GAE and VGAE unsupervised frameworks where GNN encoders are only a building block) are \textit{not} sufficient to scale GAEs and VGAEs. These models will usually still suffer from (at least) a quadratic complexity due to their costly decoding operations.
\end{itemize}

As a result, at the time of this work, GAE and VGAE models had only been applied to relatively small graphs with, at most, a few thousand nodes and edges. While the majority of works were still aiming to reconstruct \textit{entire graphs} at each GAE/VGAE training iteration, we acknowledge that Kipf~and~Welling~\cite{kipf2016-2} and Grover~et~al.~\cite{grover2019graphite} briefly mentioned \textit{node sampling} and \textit{edge sampling} ideas as possible extensions. We will later observe that a direct uniform sampling of nodes or edges in reconstruction losses is often suboptimal (and, in Chapter~\ref{chapter_4}, we will ourselves propose some refined stochastic sampling strategies for loss approximation). We note that Shi~et~al.~\cite{aaai20} also incorporated mini-batch sampling ideas in their GAE/VGAE models, but rather to reconstruct \textit{triads} of nodes and improve performances than to provide scalable models. They did not report running times nor experiments on large graphs. To sum up, the question of how to effectively scale GAE and VGAE models remained unsatisfactorily addressed.

In this Chapter~\ref{chapter_3}, we present the solution proposed at the beginning of this PhD to address this issue. More specifically, our contribution is threefold:
\begin{itemize}
    \item we introduce a general framework to scale GAE and VGAE models to large graphs with millions of nodes and edges, by optimizing the reconstruction loss (for GAE) or the ELBO objective (for VGAE) only from a dense and representative subset of nodes, and then by propagating embedding representations in the entire graph. These nodes are selected using \textit{graph degeneracy} \cite{malliaros2019} concepts. Such an approach considerably improves scalability while, to some extent, preserving performances on downstream evaluation tasks;
    \item we apply this framework to five real-world graph datasets and discuss empirical results on several variants of GAE and VGAE models for two learning tasks: link prediction and community detection. Our implementation of this framework is publicly available~\cite{salha2019-1}. At the time of publication of this work (early 2019), these experiments provided, to the best of our knowledge, the first application of GAE and VGAE models to graphs with millions of nodes and edges;
    \item we show that our models achieve competitive results w.r.t. several popular scalable unsupervised node embedding methods, such as node2vec and DeepWalk from Section~\ref{c2s223}. It emphasizes the relevance of pursuing further research towards scalable graph autoencoders.
\end{itemize}

This chapter is organized as follows. In Section~\ref{c3s32}, we present our proposed degeneracy framework for scalable graph autoencoders, and we explain how to learn node embedding spaces using a GAE or a VGAE model trained on a subset of nodes. We report and discuss our experimental evaluation in Section~\ref{c3s33}, and we conclude in Section~\ref{c3s34}. In Section~\ref{c3s35}, we provide additional tables and proofs, placed out of the ``main'' chapter for the sake of brevity and readability.

\section{Scaling GAEs and VGAEs with Graph Degeneracy}
\label{c3s32}

Throughout this paper,  we adopt the notation from Chapter~\ref{chapter_2} and further assume that $\mathcal{G}~=~(\mathcal{V},\mathcal{E})$ is an \textit{undirected} graph, without self-loops. Additionally, and for the sake of simplicity, we assume in this section that nodes are \textit{featureless}, i.e., $X = I_n$. Therefore, GAE/VGAE models only learn node embedding vectors from $A$. Node features will be re-introduced in Section~\ref{c3s33}.

\subsection{Overview of the Degeneracy Framework}

To deal with large graphs, we propose to optimize the reconstruction loss (for GAE models) or the ELBO variational lower bound (for VAE models) only from a wisely selected subset of nodes, instead of using the entire graph $\mathcal{G}$ which would be intractable. More precisely, we proceed~as~follows:
\begin{itemize}
    \item \textbf{Step 1:} firstly, we identify the nodes on which the GAE/VGAE model should be trained, by computing a $k$\textit{-core decomposition}~\cite{malliaros2019} of the graph. The selected subgraph is the so-called $k$\textit{-degenerate} version of the original one. We justify this choice in Section~\ref{c3s322}, and explain how we choose the value of $k$;
    \item \textbf{Step 2:} secondly, we train a graph autoencoder (a GAE or a VGAE, following the architecture of Kipf~and~Welling~\cite{kipf2016-2} or any variant/extension) on this $k$-degenerate subgraph. Hence, we derive node embedding vectors for the nodes included in this subgraph;
    \item \textbf{Step 3:} regarding the nodes of $\mathcal{G}$ that are not in this subgraph, we infer their embedding representations using a simple and fast propagation heuristic, presented in Section~\ref{c3s323}.
\end{itemize}

\begin{figure}[t]
    \centering
    \includegraphics[width=1.0\textwidth]{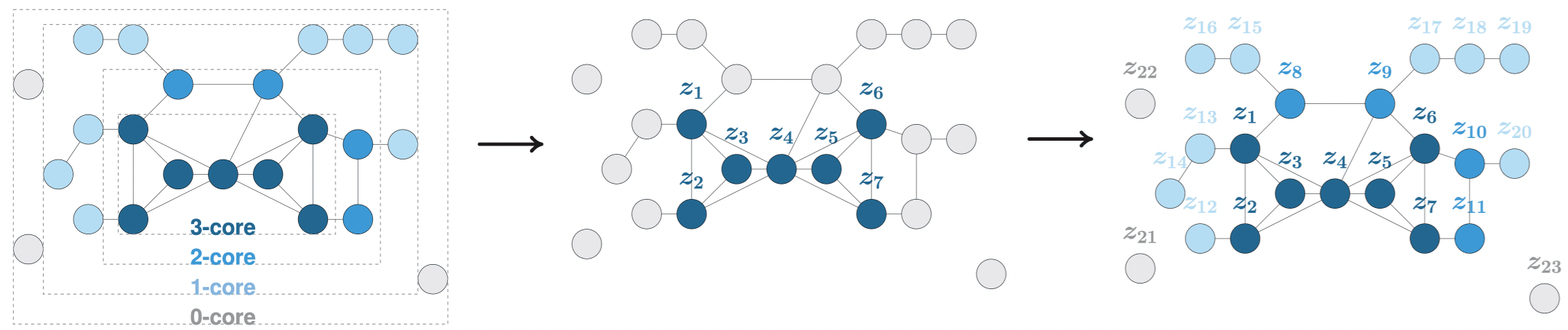}
    \caption[Overview of our degeneracy framework for scalable GAE and VGAE]{Schematic overview of our degeneracy framework for scalable graph autoencoders. After extracting the smaller $k$-core a.k.a. $k$-degenerate subgraph of $\mathcal{G}$, using the $O(\max(m,n))$ algorithm described in Section~\ref{c3s322} (\textit{left}), we train a GAE or a VGAE model on this subgraph only (\textit{middle}). In this figure, the model is trained on the 3-core subgraph. While this training procedure still suffers from (at least) a quadratic complexity, the input subgraph is significantly smaller, which makes the training tractable. However, we obtain embedding vectors~$z_i$ for nodes belonging to this subgraph, but not for others. We subsequently infer embedding vectors of the remaining nodes (\textit{right}), by leveraging the $O(m)$ algorithm described in Section~\ref{c3s323}.}
    \label{fig:ijcai_c3}
\end{figure}

In the above Figure~\ref{fig:ijcai_c3}, we illustrate the three steps of this ``\textit{degeneracy framework}'' for scalable graph autoencoders. In a nutshell, the training step (Step 2) still has a potentially high complexity (e.g., a quadratic complexity, when using the GAE or VGAE~from Kipf~and~Welling~\cite{kipf2016-2}). However, the subgraph processed by the autoencoder will be significantly smaller than $\mathcal{G}$, making the training tractable. Moreover, we will show in the next sections that Steps~1~and~3 have linear running times w.r.t. the number of edges $m$ in the graph. Therefore, this framework significantly improves speed and scalability and, as we will experimentally confirm in Section~\ref{c3s33}, can effectively process large graphs with up to millions~of~nodes~and~edges.

\subsection{Graph Degeneracy and $k$-Core Decomposition}
\label{c3s322}


In this section, we detail our Step~1, i.e., the identification of a representative subgraph on which the GAE or VGAE model should be trained. We resort to the \textit{$k$-core decomposition}~\cite{batagelj2003,malliaros2019}, a powerful tool to analyze the structure of a graph. Formally, the \textit{$k$-core}, or \textit{$k$-degenerate} version of graph $\mathcal{G}$, is defined as follows.

\begin{definition}
The \textit{$k$-core} or \textit{$k$-degenerate} version of a graph $\mathcal{G} = (\mathcal{V},\mathcal{E})$ is the largest subgraph in which every node has a degree at least equal to $k$ \textit{within the subgraph}. Therefore, in a $k$-core, each node is connected to at least $k$ nodes, that are themselves connected to at least $k$ nodes. We denote by $\mathcal{C}_k \subseteq \mathcal{V}$ the set of nodes belonging to the $k$-core of $\mathcal{G}$.
\label{def:kcore}
\end{definition}

\begin{definition}
The \textit{degeneracy} number $\delta^*(\mathcal{G}) \in \{0, \dots, n\}$ of a graph $\mathcal{G} = (\mathcal{V},\mathcal{E})$ is the maximum $k$ for which the $k$-core is not empty. 
\end{definition}

\begin{figure}[H]
  \centering
  \scalebox{0.48}{\begin{tikzpicture}[state/.style={circle, draw, minimum size=2cm}]
    \node[fill=ImperialColor, text=white, shape=circle,draw=black, minimum size=0.7cm] (A) at (0,0) {\textbf{A}} ;
    \node[fill=ImperialColor, text=white, shape=circle,draw=black, minimum size=0.7cm] (B) at (0,-3) {};
    \node[fill=ImperialColor, text=white, shape=circle,draw=black, minimum size=0.7cm] (C) at (1.5,-1.5) {};
    \node[fill=ImperialColor, text=white, shape=circle,draw=black, minimum size=0.7cm] (D) at (3,-1.5) {\textbf{B}};
    \node[fill=ImperialColor, text=white, shape=circle,draw=black, minimum size=0.7cm] (E) at (4.5,-1.5) {};
    \node[fill=ImperialColor, text=white, shape=circle,draw=black, minimum size=0.7cm] (F) at (6,0) {\textbf{C}} ;
\node[fill=ImperialColor, text=white, shape=circle,draw=black, minimum size=0.7cm] (G) at (6,-3) {} ;
\node[fill=lightblue3, text=white, shape=circle,draw=black, minimum size=0.7cm] (H) at (1.5,1.5) {\textbf{D}} ;
\node[fill=lightblue3, text=white, shape=circle,draw=black, minimum size=0.7cm] (I) at (4.5,1.5) {\textbf{E}} ;
\node[fill=lightblue2, shape=circle,draw=black, minimum size=0.7cm] (J) at (0,3) {} ;
\node[fill=lightblue2, shape=circle,draw=black, minimum size=0.7cm] (K) at (6,3) {\textbf{F}} ;
\node[fill=lightblue2, shape=circle,draw=black, minimum size=0.7cm] (L) at (7.5,3) {\textbf{G}};
\node[fill=lightblue3, shape=circle,draw=black, minimum size=0.7cm] (M) at (7.5,-0.5) {};
\node[fill=lightblue3, shape=circle,draw=black, minimum size=0.7cm] (N) at (7.5,-3) {};
\node[fill=lightblue2, shape=circle,draw=black, minimum size=0.7cm] (O) at (9,-0.5) {};
\node[fill=lightblue2, shape=circle,draw=black, minimum size=0.7cm] (P) at (-1.5,0) {};
\node[fill=lightblue2, shape=circle,draw=black, minimum size=0.7cm] (Q) at (-1.5,-3) {};
\node[fill=lightblue2, shape=circle,draw=black, minimum size=0.7cm] (R) at (-2.5,-1.5) {};
\node[fill=lightblue2, shape=circle,draw=black, minimum size=0.7cm] (S) at (9,3) {\textbf{H}};
\node[fill=lightblue2, shape=circle,draw=black, minimum size=0.7cm] (T) at (-1.5,3) {};
\node[fill=light-gray, shape=circle,draw=black, minimum size=0.7cm] (U) at (-3.5,1.5) {};
\node[fill=light-gray, shape=circle,draw=black, minimum size=0.7cm] (V) at (-3.5,-4) {};
\node[fill=light-gray, shape=circle,draw=black, minimum size=0.7cm] (W) at (10,-5) {};
    \path [] (A) edge node {} (B);
    \path [](B) edge node[left] {} (C);
    \path [](A) edge node[left] {} (C);
    \path [](B) edge node[left] {} (D);
    \path [](A) edge node[left] {} (D);
    \path [](C) edge node[left] {} (D);
    \path [](D) edge node[left] {} (E);
    \path [](D) edge node[left] {} (F);
    \path [](D) edge node[left] {} (G);
    \path [](E) edge node[left] {} (F);
     \path [](E) edge node[left] {} (G);
 \path [](F) edge node[left] {} (G);
 \path [](A) edge node[left] {} (H); 
\path [](D) edge node[left] {} (I);
 \path [](F) edge node[left] {} (I);
 \path [](J) edge node[left] {} (H);
 \path [](H) edge node[left] {} (I);
 \path [](I) edge node[left] {} (K);
\path [](K) edge node[left] {} (L);
\path [](F) edge node[left] {} (M);
\path [](G) edge node[left] {} (N);
\path [](M) edge node[left] {} (N);
\path [](M) edge node[left] {} (O);
\path [](P) edge node[left] {} (A);
\path [](Q) edge node[left] {} (B);
\path [](R) edge node[left] {} (P);
\path [](L) edge node[left] {} (S);
\path [](T) edge node[left] {} (J);
\draw [gray,thick,dashed] (-0.5,0.5) -- (6.75,0.5) -- (6.75,-3.5) -- (-0.5,-3.5) -- (-0.5,0.5);
\draw [gray,thick,dashed] (-1,2) -- (8.25,2) -- (8.25,-4.5) -- (-1,-4.5) -- (-1,2);
\draw [gray,thick,dashed] (-3,3.5) -- (9.5,3.5) -- (9.5,-5.5) -- (-3,-5.5) -- (-3,3.5);
\draw [gray,thick,dashed] (-4,4) -- (10.5,4) -- (10.5,-6.5) -- (-4,-6.5) -- (-4,4);
    \node at (3,-3.25) {\Large{\color{ImperialColor}{\textbf{3-core}}}};
    \node at (3,-4.25) {\Large{\color{lightblue3}{\textbf{2-core}}}};
    \node at (3,-5.25) {\Large{\color{lightblue2-text}{\textbf{1-core}}}};
    \node at (3,-6.25) {\Large{\color{gray}{\textbf{0-core}}}};
\end{tikzpicture}}
  \caption[A graph of degeneracy 3 and its cores]{A graph of degeneracy 3 and its cores. Some nodes are labeled for the purpose of Section~\ref{c3s323}.}
  \label{fig:kcoredecomposition}
\end{figure}
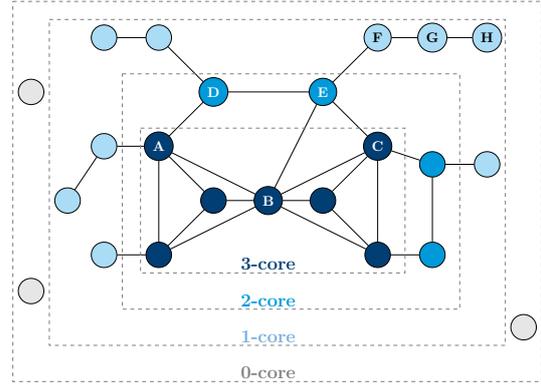

As illustrated in Figure~\ref{fig:kcoredecomposition}, nodes from each core $\mathcal{C}_0, \dots, \mathcal{C}_{\delta^*(\mathcal{G})}$ form a nested chain, i.e.:
\begin{equation}
    \mathcal{C}_{\delta^*(\mathcal{G})} \subseteq \mathcal{C}_{\delta^*(\mathcal{G})-1} \subseteq ... \subseteq \mathcal{C}_{0} = \mathcal{V}.
\end{equation}

\begin{multicols}{2}

In Step 2, we train a GAE or a VGAE model, either only on the $\delta^*(\mathcal{G})$-core version of $\mathcal{G}$, or on a larger $k$-core subgraph, i.e., for a $k < \delta^*(\mathcal{G})$. Why do we choose to resort to a core decomposition in the context of GAE and VGAE models? Our justification for this strategy is twofold. The first reason is \textit{computational}: the $k$-core decomposition can be computed in a linear running time for an undirected graph~\cite{batagelj2003}. Algorithm~\ref{c3_algo1} describes\footnote{In Algorithm~\ref{c3_algo1}, $D_{vv}$ denotes the \textit{degree} of node $v$, consistently with the notation of Definition~\ref{def:degree_mat}.} the procedure we adopt in this work. Specifically, to construct a $k$-core, the strategy is to recursively remove all nodes with degree lower than $k$ and their edges from $\mathcal{G}$ until no node can be removed. It involves sorting nodes by degrees, which can be achieved in $O(n)$ time using a variant of bin-sort, and going through all nodes and edges once (see \cite{batagelj2003} for details).  The time complexity of Algorithm~\ref{c3_algo1} is $O(\max(m,n))$, with $\max(m,n) = m$ in many real-world graphs, and with the same space complexity with sparse matrices, as explained by Batagelj and Zaversnik~\cite{batagelj2003}.
\columnbreak

\vspace{-0.5cm}

\begin{algorithm}[H]
\caption{$k$-core Decomposition}
\textbf{Input}: Graph $\mathcal{G} = (\mathcal{V},\mathcal{E})$\\
\textbf{Output}: Set of $k$-cores $\mathcal{C}=\{\mathcal{C}_0,\mathcal{C}_1,...,\mathcal{C}_{\delta^*(\mathcal{G})}\}$

\begin{algorithmic}[1] 
\STATE Initialize $\mathcal{C}=\{\mathcal{V}\}$ and $k = \min_{v \in \mathcal{V}} D_{vv}$
\FOR{$i = 1$ \textbf{to} $n$} 
\STATE $v =$ node with smallest degree in $\mathcal{G}$
\IF{$D_{vv} > k$}
\STATE Append $\mathcal{V}$ to $\mathcal{C}$
\STATE $k = D_{vv}$
\ENDIF
\STATE $\mathcal{V} = \mathcal{V} \setminus \{v\}$ and remove edges linked~to~$v$
\ENDFOR
\end{algorithmic}
\label{c3_algo1}
\end{algorithm}

\end{multicols}
\vspace{-0.5cm}

Our second reason to rely on the $k$-core decomposition is that, despite being relatively simple, it has been proven to be a very \textit{useful tool to extract representative subgraphs} over the past years, with numerous applications ranging from community detection \cite{giatsidis2014} to keyword extraction in graph-of-words \cite{tixier2016} and core-based kernels for graph similarity \cite{nikolentzos2018}. Our work, therefore, builds upon these successes. We refer to the recent survey of Malliaros~et~al.~\cite{malliaros2019} for a broader overview of the history, theory, extensions, and applications of core decomposition.

 \paragraph{On the Selection of $k$} To select $k$, one must often face an inherent \textit{performance/speed trade-off}, that we will illustrate in our experiments. Intuitively, reconstructing very small subgraphs during the GAE/VGAE training will speed up computations but, as we will later verify, this might also deteriorate performances. Besides, on large graphs, training a GAE or a VGAE will usually be impossible on the lowest cores (i.e., on the largest subgraphs) due to overly large memory requirements. In our experiments from Section~\ref{c3s33}, we will adopt a simple strategy when dealing with large graphs, consisting in training models on the lowest computationally tractable cores, i.e., on the largest possible subgraphs. In practice, these subgraphs are significantly smaller than the original ones (at least 95\% of nodes are removed, in the datasets under consideration). Moreover, when running experiments on medium-size graphs where all cores would be tractable, we will plainly avoid choosing $k < 2$ (since $\mathcal{V} = \mathcal{C}_0 = \mathcal{C}_1$, or $\mathcal{C}_0 \approx \mathcal{C}_1$, in all our graphs). As we will show, setting $k=2$, i.e., removing \textit{leaves} from the graph, will often empirically appear as a good option, preserving performances w.r.t. models trained on $\mathcal{G}$ while significantly reducing running times by pruning up to 50\% of nodes in our graphs.

\subsection{Propagation of Embedding Vectors}
\label{c3s323}

From Steps 1 and 2, we obtain embedding vectors $z_i \in \mathbb{R}^d$ for each node $i \in \mathcal{C}_k$ of a pre-selected $k$-core. Step 3 consists in the inference of such representations for the remaining nodes of $\mathcal{G}$, in a scalable way. We recall that we consider featureless nodes in this section.

Our strategy starts by assigning embedding representations to nodes directly connected to the $k$-core. We average the values of their embedded neighbors \textit{and} of the nodes being embedded at the same step of the process. For instance, in the graph of Figure~\ref{fig:kcoredecomposition}, to compute $z_D$ and $z_E$ we would solve the system:
\begin{equation}
\begin{cases}
z_D = \frac{1}{2}(z_A + z_E) \\
z_E = \frac{1}{3}(z_B + z_C + z_D)
\end{cases}    
\end{equation} 
or a weighted mean, if edges are weighted. Then, we would repeat this process on the neighbors of these newly embedded nodes, and so on until no new node would be reachable. Taking into account the fact that nodes $D$ and $E$ are themselves connected is important. Indeed, node $A$ from the maximal core is also a second-order neighbor of $E$. Exploiting such a proximity when computing $z_E$ empirically improves performances, as it also strongly impacts all the following nodes whose latent vectors will then be derived from $z_E$, i.e., nodes $F$, $G$ and $H$ in Figure~\ref{fig:kcoredecomposition}. 

More generally, let $\mathcal{V}_1$ denote the set of nodes whose embedding vectors are already computed, $\mathcal{V}_2$ the set of nodes connected to $\mathcal{V}_1$ and without embedding vectors, $A_1$ the $|\mathcal{V}_1| \times |\mathcal{V}_2|$ adjacency matrix linking nodes from $\mathcal{V}_1$ to nodes from $\mathcal{V}_2$, and $A_2$ the $|\mathcal{V}_2| \times |\mathcal{V}_2|$ adjacency matrix of nodes in $\mathcal{V}_2$. We normalize $A_1$ and $A_2$ by the total degree in $\mathcal{V}_1 \cup \mathcal{V}_2$, i.e., we divide rows by row sums of the $(A^T_1|A_2)$ matrix row-concatenating $A^T_1$ and $A_2$. We denote by $\tilde{A}_1$ and $\tilde{A}_2$ these normalized versions. We already learned the $|\mathcal{V}_1|\times d$ embedding  matrix $Z_1$ for nodes in $\mathcal{V}_1$. To implement our strategy, we want to derive a $|\mathcal{V}_2|\times d$ embedding matrix $Z_2$ for nodes in $\mathcal{V}_2$, verifying:
\begin{equation}
Z_2 = \tilde{A}_1 Z_1 + \tilde{A}_2 Z_2.  
\end{equation}
The solution of this system is:
\begin{equation}
Z^* = (I - \tilde{A}_2)^{-1} \tilde{A}_1 Z_1,
\label{eq:optizstar}
\end{equation}
which exists since $(I - \tilde{A}_2)$ is \textit{strictly diagonally dominant} and thus \textit{invertible} from the Levy-Desplanques theorem~\cite{taussky1949recurring}. Unfortunately, the exact computation of $Z^*$ has a cubic complexity. We approximate it by initializing $Z_2$ with random values in $[-1,1]$ and iterating:
\begin{equation}
Z_2 \leftarrow \tilde{A}_1 Z_1 + \tilde{A}_2 Z_2
\end{equation} until convergence to a \textit{fixed point}, guaranteed to happen exponentially fast,~as~stated~below.

\begin{algorithm}[tb]
\caption{Propagation of Embedding Vectors}
\textbf{Input}: Graph $\mathcal{G}$, list of embedded nodes $\mathcal{V}_1$, node embedding matrix $Z_1 \in \mathbb{R}^{|\mathcal{V}_1|\times d}$ (already learned), number of iterations $t$\\
\textbf{Output}: A $d$-dimensional node embedding vector for each node in  $\mathcal{G}$

\begin{algorithmic}[1] 
\STATE $\mathcal{V}_2=$ set of non-embedded nodes reachable from $\mathcal{V}_1$
\WHILE{$|\mathcal{V}_2| > 0$} 
\STATE $A_1$ = $|\mathcal{V}_1| \times |\mathcal{V}_2|$ adjacency matrix connecting nodes from $\mathcal{V}_1$ to nodes from $\mathcal{V}_2$
\STATE $A_2$ = $|\mathcal{V}_2| \times |\mathcal{V}_2|$ adjacency matrix connecting nodes from $\mathcal{V}_2$ together \STATE $\tilde{A}_1, \tilde{A}_2 =$ normalized versions of $A_1$, $A_2$, with rows divided by row sums of $(A^T_1 | A_2)$\STATE Randomly initialize $Z_2 \in \mathbb{R}^{|\mathcal{V}_2|\times d}$ \textit{(rows of $Z_2$ will be embedding vectors of nodes in $\mathcal{V}_2$)}
\FOR{$i = 1$ \textbf{to} $t$}
\STATE $Z_2 = \tilde{A}_1 Z_1 + \tilde{A}_2 Z_2$
\ENDFOR
\STATE $\mathcal{V}_1 = \mathcal{V}_2$
\STATE $\mathcal{V}_2=$ set of non-embedded nodes reachable from $\mathcal{V}_1$
\ENDWHILE
\STATE Assign random vectors to remaining unreachable nodes
\end{algorithmic}
\label{c3algo2}
\end{algorithm}

\begin{theorem}
  Let us denote by $Z^{(t)}$ the $|\mathcal{V}_2|\times f$ matrix obtained by iterating $Z^{(t)} = \tilde{A}_1 Z_1 + \tilde{A}_2 Z^{(t-1)}$ $t$ times starting from some random initial matrix $Z^{(0)}$. Let $\|\cdot\|_F$ be the Frobenius matrix norm. Then, exponentially fast, 
  \begin{equation}
 \|Z^{(t)} -  Z^* \|_F \xrightarrow[{t \rightarrow +\infty}]{} 0,    
  \end{equation}
where $Z^*$ is the optimal solution from Equation~\eqref{eq:optizstar}.
\label{c3prop1}
\end{theorem}

We prove Proposition~\ref{c3prop1} in Section~\ref{c3s35}. Our propagation process is summarized in Algorithm~\ref{c3algo2}. If some nodes are unreachable by such a process because $\mathcal{G}$ is not connected, then we assign them random vectors. Using sparse representations for $\tilde{A}_1$ and $\tilde{A}_2$, the memory requirement for Algorithm~\ref{c3algo2} is $O(m +nf)$, and the computational complexity of each evaluation of line 7 also increases linearly w.r.t. the number of edges $m$ in the graph. In practice, $t$ is small: we set $t=10$ in our experiments (we illustrate the impact of $t$ in Section~\ref{c3s35}). The number of iterations in the while loop of line 2 corresponds to the size of the longest shortest-path connecting a node to the $k$-core, a number bounded above by the diameter of the graph which increases at a $O(\log(n))$ speed in numerous real-world graphs \cite{chakrabarti2006}. In the next section, we will empirically confirm that both Steps~1~and~3 run in linear time, and therefore scale to large graphs.

\section{Experimental Analysis}
\label{c3s33}

In this section, we empirically evaluate our degeneracy framework for scalable graph autoencoders. Although all main results and conclusions are presented and discussed here, we report additional and more complete tables in the ``supplementary'' Section~\ref{c3s35} for the sake of brevity. An implementation of our framework is publicly available on GitHub\footnote{\href{https://github.com/deezer/linear_graph_autoencoders}{https://github.com/deezer/linear\_graph\_autoencoders}}.

\subsection{Experimental Setting}
\label{c3s331}

\paragraph{Datasets} In this chapter, we provide experimental results on five graphs of increasing sizes. First and foremost, for comparison purposes, we study the three medium-size citation networks\footnote{\label{linqs}\href{https://linqs.soe.ucsc.edu/data}{https://linqs.soe.ucsc.edu/data}} used by Kipf~and~Welling~\cite{kipf2016-2}: Cora ($n =$ 2 708 and $m =$ 5 429), Citeseer ($n =$ 3~327 and $m =$ 4~732), and Pubmed ($n =$ 19 717 and $m  =$ 44 338)~\cite{sen2008collective}. In these graphs, nodes are documents and edges are citation links. As Kipf~and~Welling~\cite{kipf2016-2}, we ignored edges' directions in experiments, i.e., we considered undirected versions of these graphs (we refer to our Chapter~\ref{chapter_5} for an extension of GAE and VGAE models to directed graphs). Documents/nodes have sparse bag-of-words feature vectors, of sizes 3 703, 1 433, and 500 respectively. Each document also has a class label corresponding to its topic: in Cora (respectively in Citeseer, in Pubmed), nodes are clustered in six classes (resp. in seven classes, in three classes) that we use as ground~truth communities in the following. Classes are roughly balanced. These datasets are common benchmarks for evaluating GAEs and VGAEs \cite{grover2019graphite,semiimplicit2019,kipf2016-2,pan2018arga,aaai20,tran2018multi,huang2019rwr,wang2017mgae} (we discuss the relevance and the limitations of these benchmarks in Chapter~\ref{chapter_6}). For these three medium-size graphs, we can directly compare the performance of our framework to standard GAEs and VGAEs, as training these standard models is still computationally affordable.

Then, we consider two larger graphs from Stanford's SNAP project~\cite{snap}.  The first one is the Google web graph\footnote{\href{https://snap.stanford.edu/data/web-Google.html}{https://snap.stanford.edu/data/web-Google.html}} ($n =$ 875 713 and $m =$ 4 322 051). Nodes are web pages and edges represent hyperlinks between these pages. Data do not include ground~truth communities. The second one is the US Patent citation network\footnote{\href{https://snap.stanford.edu/data/cit-Patents.html}{https://snap.stanford.edu/data/cit-Patents.html}} ($n =$ 2 745 762 and $m =$ 13 965 410), originally released by the National Bureau of Economic Research (NBER) and representing citations between patents. Nodes belong to one out of six patent categories, that act as ground truth communities; we removed nodes without categories from the original Patent graph. For both graphs, we once again ignored edges' directions.

\begin{figure*}[t]
\centering
  \subfigure[Cora]{
  \scalebox{0.37}{\includegraphics{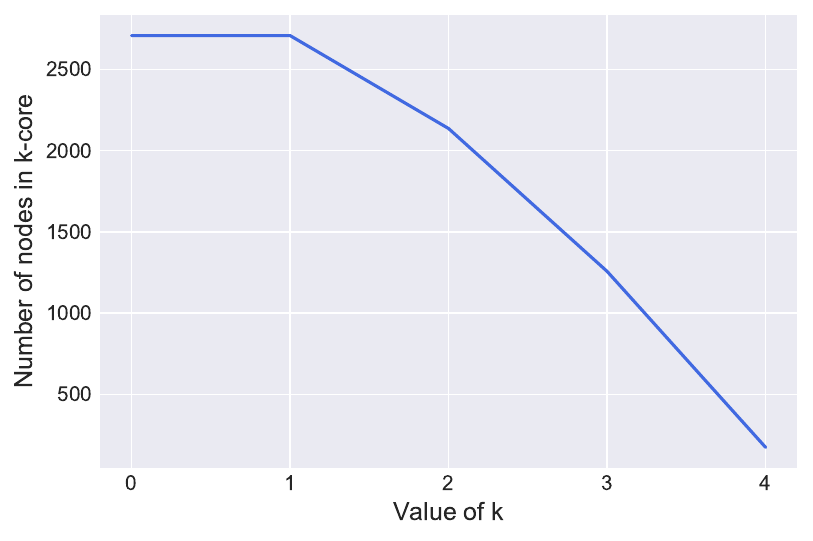}}}\subfigure[Citeseer]{
  \scalebox{0.37}{\includegraphics{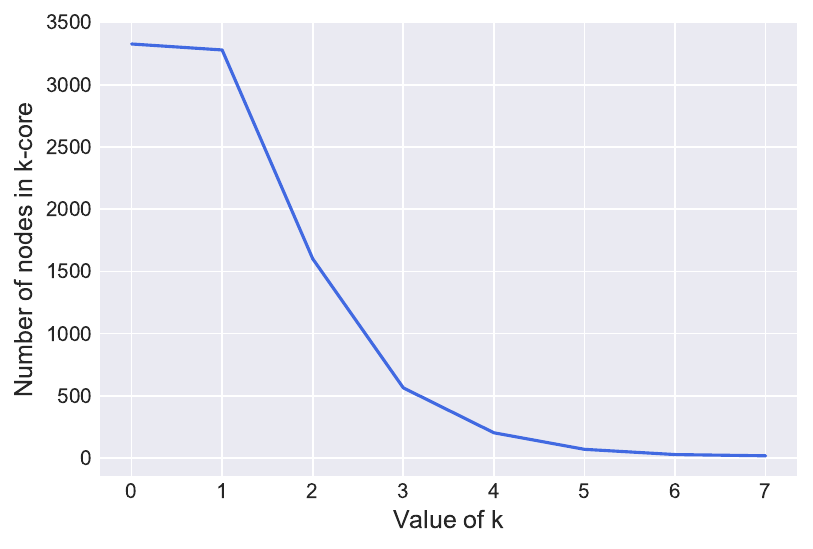}}}\subfigure[Pubmed]{
  \scalebox{0.37}{\includegraphics{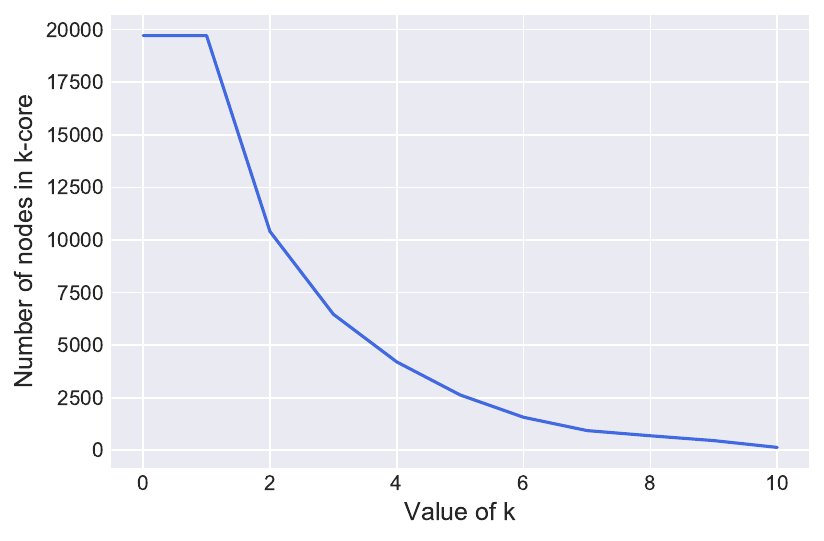}}}
    \subfigure[Google]{
  \scalebox{0.37}{\includegraphics{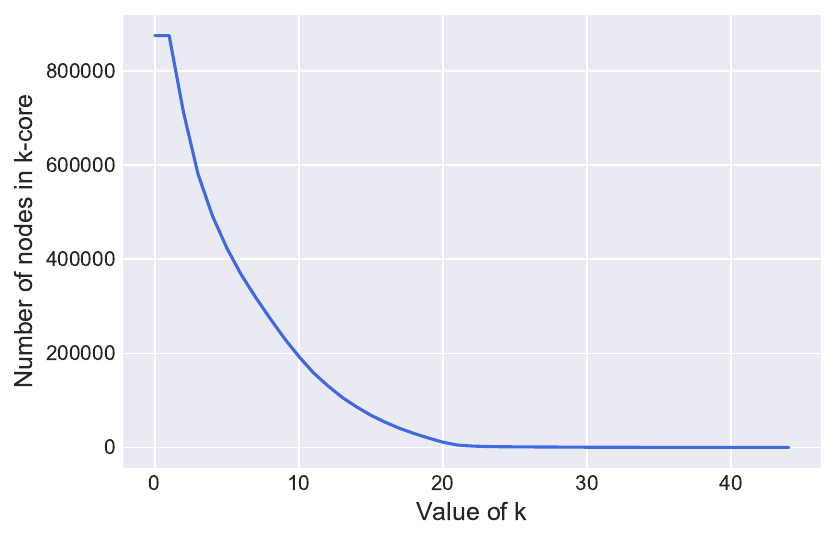}}}\subfigure[Patent]{
  \scalebox{0.37}{\includegraphics{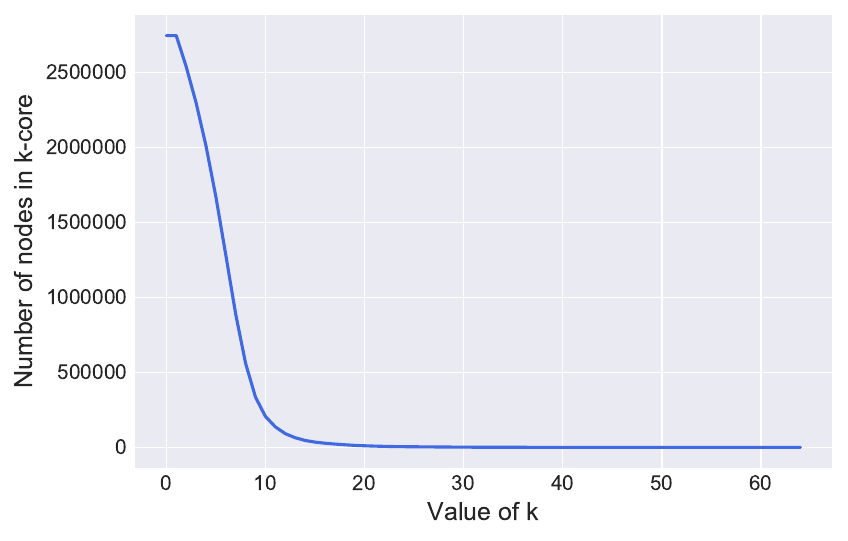}}}
  \caption[Core decomposition of the Cora, Citeseer, Pubmed, Google, and Patent graphs]{Core decomposition of the Cora, Citeseer, Pubmed, Google, and Patent graphs.}
  \label{c3fig33}
\end{figure*}

For these five graphs, we illustrate in Figure~\ref{c3fig33} the number of nodes in each $k$-core subgraph. We note that, for Google (resp. for Patent), we were unable to train autoencoders on the 0 to 15-cores (resp. on the 0 to 13-cores) in our machines due to memory errors. Overall, using our NVIDIA GTX 1080 GPU, we encountered memory errors during the GAE/VGAE training when we were trying to decode graphs with more than 35 000 nodes.

\paragraph{Tasks}

We consider two downstream tasks for evaluation. Firstly, we consider \textit{link prediction}, following the methodology described in Section~\ref{c2s212}. We train all models described thereafter on masked graphs for which $15\%$ of edges were randomly removed for medium-size graphs (resp. $5\%$ for large graphs). Then, we create validation and test sets from the removed edges (resp. from $5\%$ and $10\%$ of edges for medium-size graphs, and resp. $2\%$ and $3\%$ for large graphs) and from the same number of randomly sampled unconnected node pairs. Using the decoder's predictions $\hat{A}_{ij}$, we evaluate the model's ability to classify edges from non-edges, using the mean \textit{Area Under the ROC Curve} (AUC) and \textit{Average Precision} (AP) scores (see Section~\ref{c2s212}) on test sets. As explained in Section~\ref{c2s243}, link prediction is the most common task to evaluate GAE and VGAE models in the literature. We, therefore, found it essential to consider it as~well~in~our~work.

We also consider \textit{community detection} experiments, on datasets with ground truth communities. For this task, after training models on complete versions of the graphs, we run $k$-means algorithms~\cite{appert2021new} in node embedding spaces to cluster the $z_i$ vectors. We compare these clusters to the ground truth using the mean \textit{Adjusted Mutual Information} (AMI) score (see Section~\ref{c2s212}).

\paragraph{Models}
We apply our framework to ten graph autoencoders: the seminal GAE and VGAE models from Kipf~and~Welling~\cite{kipf2016-2} with 2-layer GCN encoders and inner product decoder; two deeper variants of GAE and VGAE with 3-layer GCN encoders, denoted DeepGAE and DeepVGAE; the Graphite and Variational Graphite extended models from Grover et al.~\cite{grover2019graphite} incorporating a
reverse message passing scheme; Pan et~al.'s adversarially regularized variants of GAE and VGAE (denoted ARGA and ARVGA)~\cite{pan2018arga}; and two other models, denoted ChebAE and ChebVAE, and consisting in variants of GAE and VGAE with ChebNets~\cite{defferrard2016} of order 3 instead of GCN encoders.  All models were trained on $200$ epochs to return 16-dimensional embedding vectors, except for Patent (32-dimensional embedding vectors). We included a 32-dimensional hidden layer in GCN encoders (two for the DeepGAE and DeepVGAE models), used the Adam optimizer~\cite{kingma2014adam}, training models without dropout~\cite{srivastava2014dropout} and with a learning rate of $0.01$. We performed full-batch gradient descent\footnote{In these experiments, we omitted two simple heuristics to approximate losses, consisting in 1) reconstructing ``mini-batches'' of nodes/edges randomly  (uniformly) sampled, and 2) incorporating negative sampling techniques~\cite{kipf2016-2}. These two methods will be presented in detail in Chapter~\ref{chapter_4}, where they will be considered as baselines (including on the same graphs as experiments from this chapter). We will show that they usually fail to effectively scale GAE/VGAE models while preserving performances, for reasons that will be discussed~in~Chapter~\ref{chapter_4}.} and used the reparameterization trick from Kingma and Welling~\cite{kingma2013vae}, described in Section~\ref{c2s242}. We used the public implementations of these models (see their respective references).

We also compare our results to the DeepWalk \cite{perozzi2014deepwalk}, LINE \cite{tang2015line} and node2vec  \cite{grover2016node2vec} methods mentioned in Section~\ref{c2s223}. We focus on these methods because they directly claim scalability. For each model, hyperparameters were tuned on AUC scores using validation sets. For DeepWalk~\cite{perozzi2014deepwalk}, we trained models from 10 random walks of length 80 per node with a window size of 5, on a single epoch for each graph. We used similar hyperparameters for node2vec \cite{grover2016node2vec}, setting $p = q =1$, and LINE \cite{tang2015line} enforcing second-order proximity. We used the public implementations provided by the authors. Due to underperforming results on some graphs with 16-dimensional embeddings, we had to increase dimensions, up to 64, to compete with autoencoders.
 We also implemented a Laplacian eigenmaps/spectral clustering baseline when this approach was tractable (embedding axes are the first 64 eigenvectors of $\mathcal{G}$'s Laplacian matrix) as well as the Louvain method \cite{blondel2008louvain} for community detection. 

For all models, we used Python\footnote{Our code notably builds upon Thomas Kipf's implementation of GAE models: \href{https://github.com/tkipf/gae}{https://github.com/tkipf/gae}} and especially the TensorFlow library~\cite{abadi2016tensorflow}, training models on an NVIDIA GTX 1080 GPU and running other operations on a double Intel~Xeon~Gold~6134~CPU. 

\subsection{Results and Discussion}

\begin{table}[t]
\centering
\caption[Link prediction on Pubmed using the degeneracy framework]{Link prediction on the Pubmed graph ($n=$ 19~717, $m =$ 44~338), using the VGAE model from Kipf and Welling~\cite{kipf2016-2} with 2-layer GCN encoders and inner product decoder, its $k$-core variants using our degeneracy framework, and other baselines. All VGAE models learn embedding vectors of dimension $d =16$. Scores are averaged over 100 runs. \textbf{Bold} numbers correspond to the best scores and best running time. Scores \textit{in italic} are within one standard deviation range from the best score.}
 \resizebox{1.0\textwidth}{!}{
\begin{tabular}{c|c|cc|cccc|c}
\toprule
\textbf{Model}  & \textbf{Size of input} & \multicolumn{2}{c}{\textbf{Mean Perf. on Test Set}} & \multicolumn{5}{c}{\textbf{Mean Running Times (in sec.)}}\\
& \textbf{$k$-core} & \footnotesize \textbf{AUC  (in \%)} & \footnotesize \textbf{AP  (in \%)} & \footnotesize $k$-core dec. & \footnotesize Model train & \footnotesize Propagation & \footnotesize \textbf{Total} & \footnotesize \textbf{Speed gain} \\ 
\midrule
\midrule
VGAE on $\mathcal{G}$ & - & $83.02 \pm 0.13$ & $\textbf{87.55} \pm \textbf{0.18}$ & - & $710.54$ & - & $710.54$ & - \\
on 2-core & $9,277 \pm 25$  & $\textbf{83.97} \pm \textbf{0.39}$ & $85.80 \pm 0.49$ & $1.35$ & $159.15$ & $0.31$ & $160.81$ & $\times 4.42$ \\
on 3-core & $5,551 \pm 19$ & $\textit{83.92} \pm \textit{0.44}$ & $85.49 \pm 0.71$ & $1.35$ & $60.12$ & $0.34$ & $61.81$ & $\times 11.50$\\
on 4-core & $3,269 \pm 30$ & $82.40 \pm 0.66$ & $83.39 \pm 0.75$ & $1.35$ & $22.14$ & $0.36$ & $23.85$ & $\times 29.79$\\
on 5-core & $1,843 \pm 25$ & $78.31 \pm 1.48$ & $79.21 \pm 1.64$ & $1.35$ & $7.71$ & $0.36$ & $9.42$  & $\times 75.43$\\
... & ... & ... & ... & ... & ... & ... & ... & ... \\
on 8-core & $414 \pm 89$ & $67.27 \pm 1.65$ & $67.65 \pm 2.00$ & $1.35$ & $1.55$ & $0.38$ & $3.28$ & $\times 216.63$ \\
on 9-core & $149 \pm 93$ & $61.92 \pm 2.88$ & $63.97 \pm 2.86$ & $1.35$ & $1.14$ & $0.38$ & $\textbf{2.87}$ & $\times \textbf{247.57}$ \\
\midrule
DeepWalk & - &$81.04 \pm 0.45$ & $84.04 \pm 0.51$ & - & $342.25$ & - & $342.25$ & - \\
LINE & - & $81.21 \pm 0.31$ & $84.60 \pm 0.37$ & - & $63.52$ & - & $63.52$ &- \\
node2vec & - & $81.25 \pm 0.26$ & $85.55 \pm 0.26$ & - & $48.91$ & - & $48.91$ &- \\
Laplacian & - & $83.14 \pm 0.42$ & $86.55 \pm 0.41$ & - & $31.71$ & - & $31.71$ & -\\
\bottomrule
\end{tabular}
}
\label{c3tablepubmed}
\end{table}

\paragraph{Medium-Size Graphs} For Cora, Citeseer, and Pubmed, we apply our framework to all possible subgraphs from the $2$-core to the $\delta^*(\mathcal{G})$-core and on entire graphs, which is still computationally tractable. Table~\ref{c3tablepubmed} reports mean AUC and AP scores and their standard errors on 100 runs (masked edges are different for each run\footnote{Excluding the validation set, which is only extracted once. Validation edges never appear in test sets.}) along with mean running times, for the link prediction task on Pubmed with the VGAE model from Kipf and Welling~\cite{kipf2016-2}. Sizes of $k$-cores vary over runs due to the edge masking process in link prediction; this phenomenon does not occur for community detection.  We report similar tables for other datasets/tasks in Section~\ref{c3s35}. 

We observe that our framework significantly improves running times w.r.t. training a VGAE on the entire graph $\mathcal{G}$. Running times decrease when $k$ increases (up to $\times 247.57$ speed gain in Table~\ref{c3tablepubmed}), which was expected since the $k$-core becomes smaller. Overall, we observe this improvement on all other datasets, on both tasks, and for other GAE/VGAE variants (see Section~\ref{c3s35}). In particular, we confirm that Steps 1 and 3 of our framework, i.e., the core decomposition and the propagation of embedding representations, are~computationally~efficient. Simultaneously, for low cores, and especially for 2-cores, performances are consistently competitive w.r.t. models trained on entire graphs, and sometimes even slightly better (e.g., $+0.95$ point in AUC for link prediction on Pubmed in Table~\ref{c3tablepubmed}). This highlights the relevance of our propagation process, and the fact that training models on smaller graphs is easier. Training a model on a 2-core actually consists in removing ``leaves'' in the graph, which might appear as a simple but effective way to reduce noise during the training phase. Training models on higher cores leads to even faster results, but at the price of a loss in performance. This corresponds to what we referred to as the \textit{performance/speed trade-off} in Section~\ref{c3s322}.

\paragraph{Large graphs}  We now consider the two larger graphs. Table~\ref{c3comdetpub} reports mean AMI scores and standard errors over 10 runs, for community detection on Patent with the VGAE model from Kipf and Welling~\cite{kipf2016-2}.
Moreover, the two Tables~\ref{c3summary1}~and~\ref{c3summary2} provide more summarized results to compare results obtained with the GAE/VGAE variants. Specifically, in Table~\ref{c3summary1}, we report link prediction results on Google, reporting performances from all variants trained on the $17$-core. In Table~\ref{c3summary2} we report community detection results on Patent, reporting performances from all variants trained on the $15$-core. We report more complete tables in Section~\ref{c3s35}.  

Overall, we reach similar conclusions w.r.t. medium-size graphs, both in terms of good performance and of scalability. However, comparison with standard models trained on $\mathcal{G}$, i.e., without using our framework, is impossible on these graphs due to overly large memory requirements. We therefore only compare performances obtained from several computationally tractable cores, illustrating once again the inherent performance/speed trade-off when choosing $k$ and validating previous insights: increasing $k$ accelerates training times, but tends to decrease performances.

\begin{table}[t]
\centering
\caption[Community detection on Patent using the degeneracy framework]{Community detection on the Patent graph ($n =$ 2 745 762, $m =$ 13 965 410), using the VGAE model from Kipf and Welling~\cite{kipf2016-2} with 2-layer GCN encoders and inner product decoder, trained on the $14$-core to $18$-core subgraphs using our degeneracy framework, and other baselines. All VGAE models learn embedding vectors of dimension $d = 32$. Scores are averaged over 10 runs. We omit two baselines: DeepWalk due to too long running times on our machine, and spectral clustering due to memory errors. \textbf{Bold} numbers correspond to the best scores and best running time. Scores \textit{in italic} are within one standard deviation range from the best score.}
\resizebox{1.0\textwidth}{!}{
\begin{tabular}{c|c|c|cccc}
\toprule
\textbf{Model}  & \textbf{Size of input} & \textbf{Mean Performance} & \multicolumn{4}{c}{\textbf{Mean Running Times (in sec.)}}\\
& \textbf{$k$-core} & \footnotesize \textbf{AMI (in \%)} & \footnotesize $k$-core dec. & \footnotesize Model train & \footnotesize Propagation & \footnotesize \textbf{Total} \\ 
\midrule
\midrule
VGAE on 14-core & $46~685$  & $\textbf{25.22} \pm \textbf{1.51}$ & $507.08$ & $6~390.37$ & $120.80$ & $7~018.25~(1\text{h}57)$ \\
on 15-core & $35~432$ & $\textit{24.53} \pm \textit{1.62}$ & $507.08$ & $2~589.95$ & $123.95$ & $3~220.98~(54\text{min})$  \\
on 16-core & $28~153$ & $\textit{24.16} \pm \textit{1.96}$ & $507.08$ & $1~569.78$ & $123.14$ & $2~200.00~(37\text{min})$ \\
on 17-core & $22~455$ & $\textit{24.14} \pm \textit{2.01}$ & $507.08$ & $898.27$ & $124.02$ & $1~529.37~(25\text{min})$ \\
on 18-core & $17~799$ & $22.54 \pm 1.98$ & $507.08$ & $\textbf{551.83}$ & $126.67$ & $\textbf{1~185.58~(20\text{min})}$ \\
\midrule
LINE & - & $23.19 \pm 1.82$ & - & $33~063.80$ & - & $33~063.80~(9\text{h}11)$ \\
Louvain & - & $11.99 \pm 1.79$ & - & $13~634.16$ & - & $13~634.16~(3\text{h}47)$ \\
node2vec & - & $\textit{24.10} \pm \textit{1.64}$ & - & $26~126.01$ & - & $26~126.01~(7\text{h}15)$ \\
\bottomrule
\end{tabular}
}
\label{c3comdetpub}
\end{table}

\paragraph{Comparison to Baselines} While it is impossible to compare our results to GAE and VGAE models trained on $\mathcal{G}$, we observe that our framework provides competitive results w.r.t. other popular scalable baselines. Our framework is faster on large graphs while achieving comparable or outperforming performances in most experiments (e.g., +1.12 AMI point for our VGAE on 14-core in Table~\ref{c3comdetpub} w.r.t. node2vec, which is also slower). These competitive performances on large graphs emphasize the relevance of pursuing research towards more scalable graph autoencoders. On the other hand, we note that several of these baselines, notably Louvain and node2vec, are better to cluster nodes in Cora and Pubmed ($+10$ points in AMI for Louvain on Cora in Section~\ref{c3s35}) which questions the global ability of existing GAE or VGAE models to identify communities in a robust way. In Chapter~\ref{chapter_6}, we will provide an in-depth investigation of the relative limitations of GAE and VGAE models on community detection.
 
\paragraph{GAE/VGAE variants} For both tasks, we observe that the adversarial training strategy adopted by ARGA/ARGVA, as well as Graphite's variant decoding schemes, and ChebNet-based encoders tend to slightly improve predictions w.r.t. standard GAE and VGAE models. However, results are often very close to those obtained from the standard 2-layer GCN-based GAE and VGAE models. This questions the relevance of overcomplexifying these graph autoencoders. In Chapter~\ref{chapter_5}, we will instead consider \textit{simpler} yet effective variants of GAEs and VGAEs.

\begin{multicols}{2}

\captionof{table}[Link prediction on Google using all GAE and VGAE variants]{Link prediction on the Google graph ($n =$~875~713, $m =$~4~322~051) using all GAE/VGAE variants trained on the 17-core ($|\mathcal{C}_{17}| = 23~787 \pm 208$) using our degeneracy framework, and the best baseline. Scores are averaged over 10 runs. \textbf{Bold} numbers correspond to the best scores and best running time. Scores \textit{in italic} are within one standard deviation range from the best score.}
\label{c3summary1}
\vspace{0.2cm}
\resizebox{0.5\textwidth}{!}{
\begin{tabular}{c|cc|c}  
\toprule
\textbf{Model} & \multicolumn{2}{c}{\textbf{Perf. on Test Set}} & \textbf{Total}\\
\footnotesize (using framework, k=17) & \footnotesize \textbf{AUC (in \%)} & \footnotesize \textbf{AP (in \%)} &  \textbf{run. time} \\ 
\midrule
\midrule
GAE   & $94.02 \pm 0.20$ & $ 94.31\pm 0.21$ &  $23 \text{min}$ \\
VGAE   & $93.22 \pm 0.40$ & $93.20 \pm 0.45$ & $\textbf{22 \text{min}}$  \\
DeepGAE   & $93.74 \pm 0.17$ & $92.94 \pm 0.33$ & $24 \text{min}$ \\
DeepVGAE   & $93.12 \pm 0.29$ & $92.71 \pm 0.29$ & $24 \text{min}$ \\
Graphite   & $93.29 \pm 0.33$ & $93.11\pm 0.42$ & $23 \text{min}$ \\
Var-Graphite   & $93.13 \pm 0.35$ & $92.90 \pm 0.39$ &$\textbf{22 \text{min}}$ \\
ARGA  & $93.82 \pm 0.17$ & $94.17 \pm 0.18$ & $23\text{min}$ \\
ARVGA  & $93.00 \pm 0.17$ & $93.38 \pm 0.19$ & $23\text{min}$ \\
ChebGAE   & $\textbf{95.24} \pm \textbf{0.26}$ & $\textbf{96.94} \pm \textbf{0.27}$ & $41\text{min}$ \\
ChebVGAE   & $\textit{95.03} \pm \textit{0.25}$ & $\textit{96.58} \pm \textit{0.21}$ & $40\text{min}$ \\
\midrule
node2vec on $\mathcal{G}$& $\textit{94.89} \pm \textit{0.63}$ & $\textit{96.82} \pm \textit{0.72}$ & $4\text{h}06$ \\
\textit{(best baseline)} & & & \\
\bottomrule
\end{tabular}
}

\columnbreak
\captionof{table}[Community detection on Patent using all GAE and VGAE variants]{Community detection on the Patent graph ($n =$~2~745~762, $m =$~13~965~410) using all GAE/VGAE variants trained on the 15-core ($|\mathcal{C}_{15}| = 35~432 \pm 208$) using our degeneracy framework, and the best baseline. Scores are averaged over 10 runs. \textbf{Bold} numbers correspond to the best score and best running time.  Scores \textit{in italic} are within one standard deviation range from the best score.}
\label{c3summary2}
\vspace{-0.35cm}
\begin{center}
\resizebox{0.39\textwidth}{!}{
\begin{tabular}{c|c|c}
\toprule
\textbf{Model}  & \textbf{Performance} & \textbf{Total} \\
\footnotesize (using framework, k=15)&  \footnotesize \textbf{AMI (in \%)} & \textbf{run. time} \\ 
\midrule
\midrule
GAE & $\textit{23.76} \pm \textit{2.25}$ & $56\text{min}$ \\
VGAE & $\textit{24.53} \pm \textit{1.51}$ & $\textbf{54\text{min}}$ \\
DeepGAE & $\textit{24.27} \pm \textit{1.10}$ & $1\text{h}01$ \\
DeepVGAE & $\textit{24.54} \pm \textit{1.23}$ & $58\text{min}$ \\
Graphite & $\textit{24.22} \pm \textit{1.45}$  & $59\text{min}$ \\
Var-Graphite & $\textit{24.25} \pm \textit{1.51}$ & $58\text{min}$ \\
ARGA & $\textit{24.26} \pm \textit{1.18}$  & $1\text{h}01$ \\
ARVGA & $\textit{24.76} \pm \textit{1.32}$ & $58\text{min}$ \\
ChebGAE & $\textit{25.23} \pm \textit{1.21}$ & $1\text{h}41$ \\
ChebVGAE &  $\textbf{25.30} \pm \textbf{1.22}$ & $1\text{h}38$ \\
\midrule
node2vec on $\mathcal{G}$ & $\textit{24.10} \pm \textit{1.64}$ & $7\text{h}15$ \\
\textit{(best baseline)}  & & \\
\bottomrule
\end{tabular}
}
\end{center}

\end{multicols}

Furthermore, in this study, we did not observe any clear empirical distinction between deterministic and variational autoencoders. For instance, while various deterministic GAEs outperform their variational counterparts in Table~\ref{c3summary1} (e.g., 94.02\% AUC for a standard GAE vs 93.22\% for a VGAE), variational models reach better results in Table~\ref{c3summary2} (e.g., 24.53\% AMI for a standard VGAE vs 23.76\% for a GAE).

Besides, while we mainly focused on featureless graphs, our framework easily extends to \textit{attributed graphs}, by adding features from the $k$-core as input of GAE/VGAE models. In this direction, we report experiments on GAE and VGAE models \textit{with node features} (when available, i.e., for Cora, Citeseer, and Pubmed) for both tasks in Section~\ref{c3s35}. The addition of node features improves scores (e.g., from 85.24\% to 88.10\% AUC for a GAE trained of the 2-core~of~Cora).

\paragraph{Limitations} Despite these promising results, our framework still suffers from several limitations. For instance, in the experiments mentioned in the previous paragraph, features were not included in Step~3's propagation. Future work might study more efficient strategies to integrate node features. Moreover, we noticed that our performances tend to deteriorate when our framework is applied to the highest cores, i.e., the smallest subgraphs. In addition, our work implicitly assumes the existence of a tractable core subgraph, which might not always exist in practice. For instance, if a graph $\mathcal{G}$ has a $k$-core subgraph too large for GAE/VGAE training, and a $(k+1)$-core too small (or even empty), our degeneracy framework will fail to provide relevant node embedding representations. Experiments from Chapter~\ref{chapter_4} will consider~this case.

\section{Conclusion}
\label{c3s34}

In this chapter, we presented a general framework to scale graph autoencoders and variational graph autoencoders. This framework leverages graph degeneracy concepts to train models only from a dense subset of nodes instead of using the entire graph. Together with a simple yet effective propagation mechanism, our approach improves scalability and training speed while, to some extent, preserving performances on link prediction and community detection. 

We evaluated our framework on ten variants of existing GAE and VGAE, providing the first application of these models to large graphs with up to millions of nodes and edges. Last, but not least, we achieved empirically competitive results w.r.t. several popular scalable node embedding methods such as node2vec and DeepWalk, in a majority of our experiments. This emphasizes the relevance of pursuing further research towards more scalable GAE and VGAE~models.

Simultaneously, in our experiments, we also identified several limitations of our GAE and VGAE models. Some of them are directly related to our degeneracy framework, such as the potentially suboptimal use of node features and the assumption, sometimes unverified, that the graph under consideration includes tractable core subgraphs. In the next Chapter~\ref{chapter_4}, we will introduce FastGAE, an alternative strategy to scale GAEs and VGAEs, that addresses these limitations.

Some other limitations are not directly related to our degeneracy framework, but rather to the GAE and VGAE models themselves. This includes the inability of these models to reconstruct directed graphs, as well as their relatively lower performance in some community detection experiments. These important aspects will be further discussed and addressed in this thesis, respectively in Chapters~\ref{chapter_5}~and~\ref{chapter_7}.

\section{Appendices}
\label{c3s35}

This supplementary section provides additional tables related to our experiments. We also prove our Proposition~\ref{c3prop1}, stated in Section~\ref{c3s32}. They were placed out of the main content of Chapter~\ref{chapter_3} for the sake of brevity and readability.

\subsection*{Proof of Proposition \ref{c3prop1}}

We have:
\begin{align}
Z^{(t)} - Z^*  &= [\tilde{A}_1 Z_1 + \tilde{A}_2 Z^{(t-1)}] - [ \tilde{A}_2 Z^* + (I - \tilde{A}_2) Z^*] \nonumber \\ 
&= \tilde{A}_1 Z_1 + \tilde{A}_2 Z^{(t-1)} - \tilde{A}_2 Z^* - (I-\tilde{A}_2)(I-\tilde{A}_2)^{-1} \tilde{A}_1 Z_1  \nonumber \\
&= \tilde{A}_2 (Z^{(t - 1)} - Z^*).    
\end{align}
 So, $Z^{(t)} -  Z^* =  \tilde{A}^{t}_2 (Z^{(0)} -  Z^*)$. Then, as a consequence of the Cauchy-Schwarz inequality~\cite{steele2004cauchy}:
\begin{equation}
\|Z^{(t)} -  Z^*\|_F = \|\tilde{A}^{t}_2 (Z^{(0)} -  Z^*)\|_F \leq \|\tilde{A}^{t}_2 \|_F  \|Z^{(0)} -  Z^*\|_F.
\end{equation}
Furthermore, $\tilde{A}^{t}_2 = P D^{t} P^{-1}$, with $\tilde{A}_2 = P D P^{-1}$ denoting the eigendecomposition of the symmetric matrix $\tilde{A}_2$. For the diagonal matrix $D^{t}$ we have:
\begin{equation}
    \|D^{t}\|_F = \sqrt{\sum_{i=1}^{|\mathcal{V}_2|} |\lambda^t_i|^{2}} \leq \sqrt{|\mathcal{V}_2|} (\max_i |\lambda_i|)^t
\end{equation} with $\lambda_i$ denoting the $i$-th eigenvalue of $\tilde{A}_2$. Since $\tilde{A}_2$ has non-negative entries, we derive from the Perron–Frobenius theorem \cite{lovasz2007} that:
\begin{itemize}
    \item the maximum absolute value among eigenvalues of $\tilde{A}_2$ is reached by a nonnegative real eigenvalue;
    \item $\max_i \lambda_i$ is bounded above by the maximum degree in $\tilde{A}_2$'s graph.
\end{itemize}
By definition, each node in $\mathcal{V}_2$ has at least one connection to $\mathcal{V}_1$. Moreover, rows of $\tilde{A}_2$ are normalized by row sums of $(A^T_1|A_2)$, so the maximum degree in $\tilde{A}_2$'s graph is strictly lower than 1. We conclude, with the above two bullet points, that $0 \leq |\lambda_i| < 1$ for all $i \in \{1,...,|\mathcal{V}_2|\}$, so $0 \leq \max_i |\lambda_i| < 1$. This result implies that:
\begin{equation}
\|D^{t}\|_F \rightarrow_{t} 0
\end{equation} exponentially fast, and so does $\|\tilde{A}^{t}_2 \|_F \leq \|P\|_F \|D^{t}\|_F \|P^{-1}\|_F$, and then $\|Z^{(t)} - Z^*\|_F$.

\subsection*{Additional Tables}

\paragraph{Link Prediction} Tables \ref{lpB} to \ref{lpE} provide more complete results for the link prediction task. For medium-size graphs, we apply our framework to all possible subgraphs from the $2$-core to the $\delta^*(\mathcal{G})$-core and on entire graphs for comparison, for the VGAE model with 2-layer GCN encoders and inner product decoder~\cite{kipf2016-2}. Sizes of $k$-cores vary over runs due to the edge masking process. We obtained comparable performance/speed trade-offs for its GAE counterpart and for other GAE/VGAE variants: for the sake of brevity, we therefore only report results on 2-core for these models. Also, we only report the best baseline for the sake of brevity, and refer to the IJCAI paper~\cite{salha2019-1} for exhaustive results. For the Google and Patent graphs, comparison with full models on $\mathcal{G}$ is impossible due to overly large memory requirements. As a consequence, we apply our framework to the five largest $k$-cores (in terms of number of nodes) that were tractable using our machines. We report mean AUC and AP and their standard errors on 10 runs (train incomplete graphs and masked edges are different for each run) along with mean running times, for the standard VGAE model~\cite{kipf2016-2}. For other models, we only report results on the second largest cores for the sake of brevity. We chose the second largest cores (17-core for Google, 14-core for Patent), instead of the largest cores to lower~running~times.

\paragraph{Community Detection}
Tables \ref{cdB} to \ref{cdE} provide more complete results for the community detection task. As explained in Section~\ref{c3s34}, we run $k$-means over embedding vectors and report AMI scores w.r.t. ground truth communities. We used scikit-learn's implementation~\cite{pedregosa2011scikit} with $k$-means++ initialization~\cite{arthur2007kmeansplus}.
We do not report results for the Google graph, due to the lack of ground truth communities. Also, we obtained very low scores on Citeseer, which suggests that node features are more useful than the graph structure to explain labels. As a consequence, we also omit this graph here (community detection on Citesser will nonetheless be considered later in this thesis, in Chapters~\ref{chapter_4}~and~\ref{chapter_7}) and focus on Cora, Pubmed, and Patent in Tables~\ref{cdB}~to~\ref{cdE}. Tables are constructed in a similar fashion w.r.t. the previous ones for link prediction. 
Graph AE/VAE models and baselines were trained with identical hyperparameters w.r.t. \textit{link prediction} task (which we will question in Chapter~\ref{chapter_7}). As link prediction, we only report the best baseline in tables for the sake of brevity, and refer to the IJCAI paper~\cite{salha2019-1} for exhaustive results.

\begin{figure}[t]
  \centering
  \subfigure[Pubmed: VGAE model trained on $8$-core]{
  \scalebox{0.57}{\includegraphics{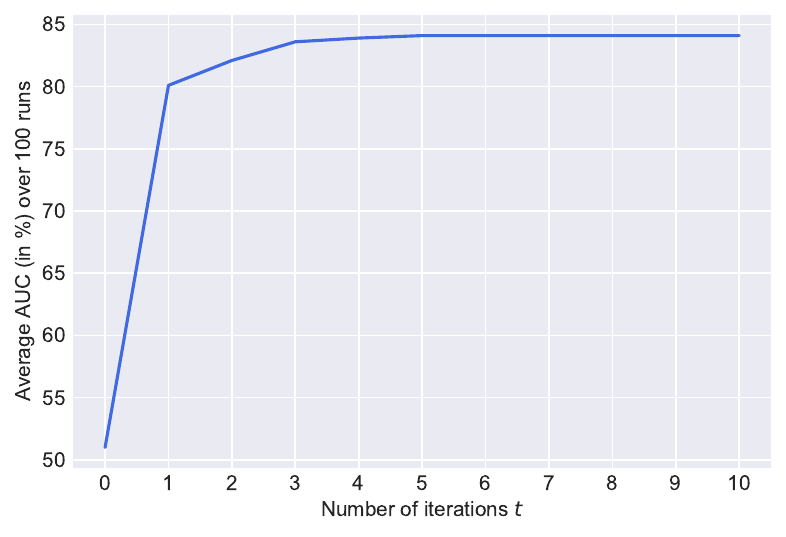}}}
  \subfigure[Google: VGAE model trained on $18$-core]{
  \scalebox{0.57}{\includegraphics{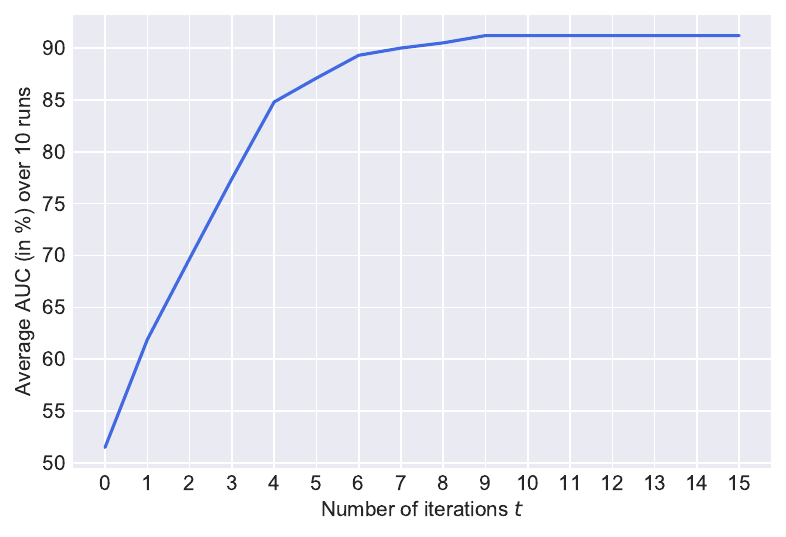}}}
  \caption[Impact of the number of iterations $t$ during progagation on mean AUC scores]{Impact of the number of iterations $t$ during progagation on mean AUC scores.}
  \label{fig:impactoft}
\end{figure}

\paragraph{Impact of the number of iterations $t$} To finish, we illustrate the impact of the number of iterations $t$ during propagation on performances in Figure~\ref{fig:impactoft}. We display the evolution of mean AUC on two  different graphs and cores w.r.t. the value of $t$. Overall, all scores stabilize by setting $t>5$ (resp. $t>10$) in all medium-size graphs (resp. all large graphs). We specify that the number of iterations has a negligible impact on running time. In our experiments, we set $t = 10$ for all models leveraging our degeneracy framework.

\begin{table}[t]
\centering
\caption[Link prediction on Cora using the degeneracy framework]{Link prediction on the Cora graph ($n=$ 2~708, $m =$ 5~429), using the standard VGAE~\cite{kipf2016-2} model on all cores, and GAE/VGAE variants on the 2-core. All GAE/VGAE models learn embedding vectors of dimension $d =16$. Scores are averaged over 100 runs. \textbf{Bold} numbers correspond to the best scores, in each table subsection\protect\footnotemark.} 
 \resizebox{1.0\textwidth}{!}{
\begin{tabular}{c|c|cc|cccc}
\toprule
\textbf{Model}  & \textbf{Size of input} & \multicolumn{2}{c}{\textbf{Mean Perf. on Test Set}} & \multicolumn{4}{c}{\textbf{Mean Running Times (in sec.)}}\\
& \textbf{$k$-core} & \footnotesize \textbf{AUC (in \%)} & \footnotesize \textbf{AP (in \%)} & \footnotesize $k$-core dec. & \footnotesize Model train & \footnotesize Propagation & \footnotesize \textbf{Total} \\ 
\midrule
\midrule
VGAE on $\mathcal{G}$ & - & $84.07 \pm 1.22$ & $\textbf{87.83} \pm \textbf{0.95}$ & - & $15.34$ & - & $15.34$ \\
on 2-core & $1~890 \pm 16$  & $\textbf{85.24} \pm \textbf{1.12}$ & $ 87.37 \pm 1.13$ & $0.16$ & $8.00$ & $0.10$ & $8.26$ \\
on 3-core & $862 \pm 26$ & $84.53 \pm 1.33$ & $85.04 \pm 1.87$ & $0.16$ & $2.82$ & $0.11$ & $3.09$  \\
on 4-core & $45 \pm 13$ & $72.33 \pm 4.67$ & $71.98 \pm 4.97$ & $0.16$ & $\textbf{0.98}$ & $0.12$ & $\textbf{1.26}$ \\
\midrule
GAE on 2-core & $1~890 \pm 16$ &$85.17 \pm 1.02$ & $87.26 \pm 1.12$ & $0.16$ & $8.05$ & $0.10$ & $8.31$ \\
DeepGAE on 2-core & $1~890 \pm 16$ & $86.25 \pm 0.81$ & $87.92 \pm 0.78$ & $0.16$ & $8.24$ & $0.10$ & $8.50$ \\
DeepVGAE on 2-core & $1~890 \pm 16$ & $86.16 \pm 0.95$ & $87.71 \pm 0.98$ & $0.16$ & $8.20$ & $0.10$ & $8.46$ \\
Graphite on 2-core & $1~890 \pm 16$& $86.35 \pm 0.82$ & $88.18 \pm 0.84$ & $0.16$ & $9.41$ & $0.10$ & $9.67$ \\
Var-Graphite on 2-core & $1~890 \pm 16$ & $\textbf{86.39} \pm \textbf{0.84}$ & $88.05 \pm 0.80$ & $0.16$ & $9.35$ & $0.10$ & $9.61$ \\
ARGA on 2-core & $1~890 \pm 16$ & $85.82 \pm 0.88$ & $ 88.22\pm 0.70$ & $0.16$ & $7.99$ & $0.10$ & $8.25$ \\
ARVGA on 2-core & $1~890 \pm 16$ &$85.74 \pm 0.74$ & $88.14 \pm 0.74$ & $0.16$ & $7.98$ & $0.10$ & $8.24$ \\
ChebGAE on 2-core & $1~890 \pm 16$ & $86.15 \pm 0.54$ & $88.01 \pm 0.39$ & $0.16$ & $15.78$ & $0.10$ & $16.04$ \\
ChebVGAE on 2-core & $1~890 \pm 16$ &$86.30 \pm 0.49$ & $\textbf{88.29} \pm \textbf{0.50}$ & $0.16$ & $15.65$ & $0.10$ & $15.91$ \\
\midrule
GAE with node features on 2-core & $1~890 \pm 16$ &$\textbf{88.10} \pm \textbf{0.87}$ & $89.36 \pm 0.88$ & $0.16$ & $8.66$ & $0.10$ & $8.92$ \\
VGAE with node features on 2-core & $1~890 \pm 16$ &$87.97 \pm 0.99$ & $\textbf{89.53} \pm \textbf{0.96}$ & $0.16$ & $8.60$& $0.10$ & $8.86$ \\
\midrule
Laplacian eigenmaps & - & $\textbf{86.53} \pm \textbf{1.02}$ & $\textbf{87.41} \pm \textbf{1.12}$ & - & $2.78$ & - & $2.78$ \\
(\textit{best baseline}) & & & & & & & \\
\bottomrule
\end{tabular}
}
\label{lpB}
\end{table}

\footnotetext{In this Table~\ref{lpB} and in the following ones, we chose not to report numbers in \textit{italic}, contrary to Section~\ref{c3s32}. This is due to the fact that several ``best'' scores are now presented (one for each table subsection), making the \textit{italic} notation from  Section~\ref{c3s32} potentially ambiguous in these tables.}

\begin{table}
\centering
\caption[Link prediction on Citeseer using the degeneracy framework]{Link prediction on the Citeseer graph ($n=$ 3~327, $m =$ 4~732), using the standard VGAE~\cite{kipf2016-2} model on all cores*, and GAE/VGAE variants on the 2-core. All GAE/VGAE models learn embedding vectors of dimension $d =16$. Scores are averaged over 100 runs. \textbf{Bold} numbers correspond to the best scores, in each table subsection. * 6-core and 7-core are not reported due to their frequent vanishing after edge masking.}
 \resizebox{1.0\textwidth}{!}{
\begin{tabular}{c|c|cc|cccc}
\toprule
\textbf{Model}  & \textbf{Size of input} & \multicolumn{2}{c}{\textbf{Mean Perf. on Test Set}} & \multicolumn{4}{c}{\textbf{Mean Running Times (in sec.)}}\\
& \textbf{$k$-core} & \footnotesize \textbf{AUC (in \%)} & \footnotesize \textbf{AP (in \%)} & \footnotesize $k$-core dec. & \footnotesize Model train & \footnotesize Propagation & \footnotesize \textbf{Total} \\ 
\midrule
\midrule
VGAE on $\mathcal{G}$ & - & $\textbf{78.10} \pm \textbf{1.52}$ & $\textbf{83.12} \pm \textbf{1.03}$ & - & $22.40$ & - & $22.40$ \\
on 2-core & $1~306 \pm 19$  & $77.50 \pm 1.59$ & $ 81.92 \pm 1.41$ & $0.15$ & $4.72$ & $0.11$ & $4.98$ \\
on 3-core& $340 \pm 13$ & $76.40 \pm 1.72$ & $80.22 \pm 1.42$ & $0.15$ & $1.75$ & $0.14$ & $2.04$  \\
on 4-core & $139 \pm 13$ & $73.34 \pm 2.43$ & $75.49 \pm 2.39$ & $0.15$ & $1.16$ & $0.16$ & $1.47$ \\
on 5-core & $46 \pm 10$ & $65.47 \pm 3.16$ & $68.50 \pm 2.77$ & $0.15$ & $\textbf{0.99}$ & $0.16$ & $\textbf{1.30}$ \\
\midrule
GAE on 2-core & $1~306 \pm 19$ &$78.35 \pm 1.51$ & $82.44 \pm 1.32$ & $0.15$ & $4.78$ & $0.11$ & $5.04$ \\
DeepGAE on 2-core & $1~306 \pm 19$ & $\textbf{79.32} \pm \textbf{1.39}$ & $82.80 \pm 1.33$ & $0.15$ & $4.99$ &  $0.11$ & $5.25$ \\
DeepVGAE on 2-core & $1~306 \pm 19$ & $78.52 \pm 1.02$ & $82.43 \pm 0.97$ & $0.15$ & $4.95$ &  $0.11$ & $5.21$ \\
Graphite on 2-core & $1~306 \pm 19$ & $78.61 \pm 1.58$ & $82.81 \pm 1.24$ & $0.15$ & $5.88$ & $0.11$ & $6.14$ \\
Var-Graphite on 2-core & $1~306 \pm 19$ & $78.51 \pm 1.62$ & $82.72 \pm 1.25$ & $0.15$ & $5.86$ &  $0.11$ & $6.12$ \\
ARGA on 2-core & $1~306 \pm 19$ & $78.89 \pm 1.33$ & $82.89 \pm 1.03$ & $0.15$ & $4.54$ &  $0.11$ & $4.80$ \\
ARVGA on 2-core & $1~306 \pm 19$ &$77.98 \pm 1.39$ & $82.39 \pm 1.09$ & $0.15$ & $4.40$ &  $0.11$ & $4.66$ \\
ChebGAE on 2-core & $1~306 \pm 19$ & $78.62 \pm 0.95$ & $83.22 \pm 0.89$ & $0.15$ & $8.87$ &  $0.11$ & $9.13$ \\
ChebVGAE on 2-core & $1~306 \pm 19$ &$78.75 \pm 1.03$ & $\textbf{83.23} \pm \textbf{0.76}$ & $0.15$ & $8.75$ &  $0.11$ & $9.01$ \\
\midrule
GAE with node features on 2-core & $1~306 \pm 19$ & $81.21 \pm 1.86$ & $\textbf{83.99} \pm \textbf{1.52}$ & $0.15$ & $5.51$ & $0.11$ & $5.77$ \\
VGAE with node features on 2-core & $1~306 \pm 19$ & $\textbf{81.88} \pm \textbf{2.23}$ & $83.83 \pm 1.85$ & $0.15$ & $5.70$& $0.11$ & $5.96$ \\
\midrule
Laplacian eigenmaps & - & $\textbf{80.56} \pm \textbf{1.41}$ & $\textbf{83.98} \pm \textbf{1.08}$ & - & $3.77$ & - & $3.77$ \\
(\textit{best baseline}) & & & & & & & \\
\bottomrule
\end{tabular}}
\end{table}


\begin{table}
\centering
\caption[Link prediction on Pubmed using the degeneracy framework]{Link prediction on the Pubmed graph ($n=$ 19~717, $m =$ 44~338), using the standard VGAE~\cite{kipf2016-2} model on all cores*, and GAE/VGAE variants on the 2-core. All GAE/VGAE models learn embedding vectors of dimension $d =16$. Scores are averaged over 100 runs. \textbf{Bold} numbers correspond to the best scores, in each table subsection. * 10-core is not reported due to its frequent vanishing after edge masking.}
 \resizebox{1.0\textwidth}{!}{
\begin{tabular}{c|c|cc|cccc}
\toprule
\textbf{Model}  & \textbf{Size of input} & \multicolumn{2}{c}{\textbf{Mean Perf. on Test Set}} & \multicolumn{4}{c}{\textbf{Mean Running Times (in sec.)}}\\
& \textbf{$k$-core} & \footnotesize \textbf{AUC (in \%)} & \footnotesize \textbf{AP (in \%)} & \footnotesize $k$-core dec. & \footnotesize Model train & \footnotesize Propagation & \footnotesize \textbf{Total} \\ 
\midrule
\midrule
VGAE on $\mathcal{G}$ & - & $83.02 \pm 0.13$ & $\textbf{87.55} \pm \textbf{0.18}$ & - & $710.54$ & - & $710.54$  \\
on 2-core & $9~277 \pm 25$  & $\textbf{83.97} \pm \textbf{0.39}$ & $85.80 \pm 0.49$ & $1.35$ & $159.15$ & $0.31$ & $160.81$ \\
on 3-core & $5~551 \pm 19$ & $83.92 \pm 0.44$ & $85.49 \pm 0.71$ & $1.35$ & $60.12$ & $0.34$ & $61.81$ \\
on 4-core & $3~269 \pm 30$ & $82.40 \pm 0.66$ & $83.39 \pm 0.75$ & $1.35$ & $22.14$ & $0.36$ & $23.85$ \\
on 5-core & $1~843 \pm 25$ & $78.31 \pm 1.48$ & $79.21 \pm 1.64$ & $1.35$ & $7.71$ & $0.36$ & $9.42$ \\
... & ... & ... & ... & ... & ... & ... & ... \\
on 8-core & $414 \pm 89$ & $67.27 \pm 1.65$ & $67.65 \pm 2.00$ & $1.35$ & $1.55$ & $0.38$ & $3.28$  \\
on 9-core & $149 \pm 93$ & $61.92 \pm 2.88$ & $63.97 \pm 2.86$ & $1.35$ & $\textbf{1.14}$ & $0.38$ & $\textbf{2.87}$ \\
\midrule
GAE on 2-core & $9~277 \pm 25$ &$84.30 \pm 0.27$ & $\textbf{86.11} \pm \textbf{0.43}$ & $1.35$ & $167.25$ &$0.31$ & $168.91$ \\
DeepGAE on 2-core & $9~277 \pm 25$ & $84.61 \pm 0.54$ & $85.18 \pm 0.57$ & $1.35$ & $166.38$ & $0.31$ & $168.04$ \\
DeepVGAE on 2-core & $9~277 \pm 25$ & $84.46 \pm 0.46$ & $85.31 \pm 0.45$ & $1.35$ & $157.43$ & $0.31$ & $159.09$ \\
Graphite on 2-core & $9~277 \pm 25$ & $84.51 \pm 0.58$ & $85.65 \pm 0.58$ & $1.35$ & $167.88$ & $0.31$ & $169.54$ \\
Var-Graphite on 2-core & $9~277 \pm 25$ & $84.30 \pm 0.57$ & $85.57 \pm 0.58$ & $1.35$ & $158.16$ & $0.31$ & $159.82$ \\
ARGA on 2-core & $9~277 \pm 25$ & $84.37 \pm 0.54$ & $86.07 \pm 0.45$ & $1.35$ & $164.06$ & $0.31$ & $165.72$ \\
ARVGA on 2-core & $9~277 \pm 25$ &$84.10 \pm 0.53$ & $85.88 \pm 0.41$ & $1.35$ & $155.83$ & $0.31$ & $157.49$ \\
ChebGAE on 2-core & $9~277 \pm 25$ & $\textbf{84.63} \pm \textbf{0.42}$ & $86.05 \pm 0.70$ & $1.35$ & $330.37$ & $0.31$ & $332.03$ \\
ChebVGAE on 2-core & $9~277 \pm 25$ &$84.54 \pm 0.48$ & $86.00 \pm 0.63$ & $1.35$ & $320.01$ & $0.31$ & $321.67$ \\
\midrule
GAE with node features on 2-core & $9~277 \pm 25$ &$84.94 \pm 0.54$ & $85.83 \pm 0.58$ & $1.35$ & $168.62$ & $0.31$ & $170.28$ \\
VGAE with node features on 2-core & $9~277 \pm 25$ &$\textbf{85.81} \pm \textbf{0.68}$ & $\textbf{88.01} \pm \textbf{0.53}$ & $1.35$ & $164.10$& $0.31$ & $165.76$ \\
\midrule
Laplacian eigenmaps & - & $\textbf{83.14} \pm \textbf{0.42}$ & $\textbf{86.55} \pm \textbf{0.41}$ & - & $31.71$ & - & $31.71$ \\
(\textit{best baseline}) & & & & & & & \\
\bottomrule
\end{tabular}}
\end{table}


\begin{table}
\centering
\caption[Link prediction on Google using the degeneracy framework]{Link prediction on the Google graph ($n=$ 875~713, $m =$ 4~322~051), using the standard VGAE~\cite{kipf2016-2} model trained on the 16 to 20-cores, and GAE/VGAE variants on the 17-core. All GAE/VGAE models learn embedding vectors of dimension $d = 16$. Scores are averaged over 10 runs. \textbf{Bold} numbers correspond to the best scores, in each table subsection.}
 \resizebox{1.0\textwidth}{!}{
\begin{tabular}{c|c|cc|cccc}
\toprule
\textbf{Model}  & \textbf{Size of input} & \multicolumn{2}{c}{\textbf{Mean Perf. on Test Set}} & \multicolumn{4}{c}{\textbf{Mean Running Times (in sec.)}}\\
& \textbf{$k$-core} & \footnotesize \textbf{AUC (in \%)} & \footnotesize \textbf{AP (in \%)} & \footnotesize $k$-core dec. & \footnotesize Model train & \footnotesize Propagation & \footnotesize \textbf{Total} \\ 
\midrule
\midrule
VGAE on 16-core & $36~854 \pm 132$  & $\textbf{93.56} \pm \textbf{0.38}$ & $\textbf{93.34} \pm \textbf{0.31}$ & $301.16$ & $2~695.42$ & $25.54$ & $3~022.12~(50\text{min})$ \\
on 17-core  & $23~787 \pm 208$ & $93.22 \pm 0.40$ & $93.20 \pm 0.45$ & $301.16$ & $1~001.64$ & $28.16$ & $1~330.86~(22 \text{min})$  \\
on 18-core  & $13~579 \pm 75$ & $91.24 \pm 0.40$ & $92.34 \pm 0.51$ & $301.16$ & $326.76$ & $28.20$ & $656.12~(11\text{min})$ \\
on 19-core & $6~613 \pm 127$ & $87.79 \pm 0.31$ & $89.13 \pm 0.29$ & $301.16$ & $82.19$ & $28.59$ & $411.94~(7 \text{min})$ \\
on 20-core & $3~589 \pm 106$ & $81.74 \pm 1.17$ & $83.51 \pm 1.22$ & $301.16$ & $\textbf{25.59}$ & $28.50$ & $\textbf{355.55~(6 \text{min})}$ \\
\midrule
GAE on $17$-core & $23~787 \pm 208$ &$94.02 \pm 0.20$ & $ 94.31\pm 0.21$ &  $301.16$ & $1~073.18$ & $28.16$ & $1~402.50~(23 \text{min})$ \\
DeepGAE on $17$-core & $23~787 \pm 208$& $93.74 \pm 0.17$ & $92.94 \pm 0.33$ &  $301.16$ & $1~137.24$ & $28.16$ & $1~466.56~(24 \text{min})$ \\
DeepVGAE on $17$-core & $23~787 \pm 208$ & $93.12 \pm 0.29$ & $92.71 \pm 0.29$ &  $301.16$ & $1~088.41$ & $28.16$ & $1~417.73~(24 \text{min})$ \\
Graphite on $17$-core & $23~787 \pm 208$ & $93.29 \pm 0.33$ & $93.11\pm 0.42$ &  $301.16$ & $1~033.21$ & $28.16$ & $1~362.53~(23 \text{min})$ \\
Var-Graphite on $17$-core & $23~787 \pm 208$ & $93.13 \pm 0.35$ & $92.90 \pm 0.39$ &  $301.16$ & $989.90$ & $28.16$ & $1~319.22~(22 \text{min})$ \\
ARGA on $17$-core & $23~787 \pm 208$ & $93.82 \pm 0.17$ & $94.17 \pm 0.18$ &  $301.16$ & $1~053.95$ & $28.16$ & $1~383.27~(23 \text{min})$ \\
ARVGA on $17$-core & $23~787 \pm 208$ &$93.00 \pm 0.17$ & $93.38 \pm 0.19$ &  $301.16$ & $1~027.52$ & $28.16$ & $1~356.84~(23 \text{min})$ \\
ChebGAE on $17$-core & $23~787 \pm 208$ & $\textbf{95.24} \pm \textbf{0.26}$ & $\textbf{96.94} \pm \textbf{0.27}$ &  $301.16$ & $2~120.66$ & $28.16$ & $2~449.98~(41 \text{min})$ \\
ChebVGAE on $17$-core & $23~787 \pm 208$ &$95.03 \pm 0.25$ & $96.82 \pm 0.72$ &  $301.16$ & $2~086.07$ & $28.16$ & $2~415.39~(40 \text{min})$ \\
\midrule
node2vec & - & $\textbf{94.89} \pm \textbf{0.63}$ & $\textbf{96.82} \pm \textbf{0.72}$ & - & $14~762.78$ & - & $14~762.78~(4\text{h}06)$ \\
(\textit{best baseline}) & & & & & & & \\
\bottomrule
\end{tabular}}
\end{table}

\begin{table}
\centering
\caption[Link prediction on Patent using the degeneracy framework]{Link prediction on the Patent graph ($n=$ 2~745~762, $m =$ 13~965~410), using the standard VGAE~\cite{kipf2016-2} model trained on the 14 to 18-cores, and GAE/VGAE variants on the 15-core. All GAE/VGAE models learn embedding vectors of dimension $d = 32$. Scores are averaged over 10 runs. \textbf{Bold} numbers correspond to the best scores, in each table subsection.}
 \resizebox{1.0\textwidth}{!}{
\begin{tabular}{c|c|cc|cccc}
\toprule
\textbf{Model}  & \textbf{Size of input} & \multicolumn{2}{c}{\textbf{Mean Perf. on Test Set}} & \multicolumn{4}{c}{\textbf{Mean Running Times (in sec.)}}\\
& \textbf{$k$-core} & \footnotesize \textbf{AUC (in \%)} & \footnotesize \textbf{AP (in \%)} & \footnotesize $k$-core dec. & \footnotesize Model train & \footnotesize Propagation & \footnotesize \textbf{Total} \\ 
\midrule
\midrule
VGAE on $14$-core & $38~408 \pm 147$  & $\textbf{88.48} \pm \textbf{0.35}$ & $\textbf{88.81} \pm \textbf{0.32}$ & $507.08$ & $3~024.31$ & $122.29$ & $3~653.68~(1\text{h}01)$ \\
on $15$-core & $29~191 \pm 243$ & $88.16 \pm 0.50$ & $88.37 \pm 0.57$ & $507.08$  & $1~656.46$ & $123.47$ &$2~287.01~(38\text{min})$ \\
on $16$-core  & $23~132 \pm 48$ & $87.85 \pm 0.47$ & $88.02 \pm 0.48$ & $507.08$ & $948.09$ & $124.26$  &$1~579.43~(26\text{min})$ \\
on $17$-core  & $18~066 \pm 143$ & $87.34 \pm 0.56$ & $87.64 \pm 0.47$ & $507.08$ & $574.25$ & $126.55$  & $1~207.88~(20\text{min})$ \\
on $18$-core  & $13~972 \pm 86$ & $87.27 \pm 0.55$ & $87.78 \pm 0.51$ & $507.08$ & $\textbf{351.73}$ & $127.01$ &$\textbf{985.82~(16\text{min})}$ \\
\midrule
GAE on $15$-core &  $29~191 \pm 243$ & $87.59 \pm 0.29$ & $87.30 \pm 0.28$ &  $507.08$  & $1~880.11$ &  $123.47$ & $2~510.66~(42 \text{min})$ \\
DeepGAE on $15$-core & $29~191 \pm 243$ & $87.71 \pm 0.31$ & $87.64 \pm 0.19$ &  $507.08$  & $2~032.15$ &  $123.47$ & $2~662.70~(44 \text{min})$ \\
DeepVGAE on $15$-core &  $29~191 \pm 243$ & $87.03 \pm 0.54$ & $87.20 \pm 0.44$ &  $507.08$  & $1~927.33$ &  $123.47$ & $2~557.88~(43 \text{min})$ \\
Graphite on $15$-core &  $29~191 \pm 243$ & $85.19 \pm 0.38$ & $86.01 \pm 0.31$ &  $507.08$  & $1~989.72$ &  $123.47$ & $2~620.27~(44 \text{min})$ \\
Var-Graphite on $15$-core &  $29~191 \pm 243$ & $85.37 \pm 0.30$ & $86.07 \pm 0.24$ &  $507.08$  & $1~916.79$ &  $123.47$ & $2~547.34~(42 \text{min})$ \\
ARGA on $15$-core &  $29~191 \pm 243$ & $\textbf{89.22} \pm \textbf{0.10}$ & $\textbf{89.40} \pm \textbf{0.11}$ &  $507.08$  & $2~028.46$ &  $123.47$ & $2~659.01~(44 \text{min})$ \\
ARVGA on $15$-core &  $29~191 \pm 243$ &$87.18 \pm 0.17$ & $87.39 \pm 0.33$ &  $507.08$  & $1~915.53$ &  $123.47$ & $2~546.08~(42 \text{min})$ \\
ChebGAE on $15$-core &  $29~191 \pm 243$  & $88.53 \pm 0.20$ & $88.91 \pm 0.20$ &  $507.08$  & $3~391.01$ & $123.47$ & $4~021.56~(1\text{h}07)$ \\
ChebVGAE on $15$-core & $29~191 \pm 243$ &$88.75 \pm 0.19$ & $89.07 \pm 0.24$ &  $507.08$  & $3~230.52$ & $123.47$ & $3~861.07~(1\text{h}04)$ \\
\midrule
node2vec & - & $\textbf{95.04} \pm \textbf{0.25}$ & $\textbf{96.01} \pm \textbf{0.19}$ & - & $26~126.01$ & - & $26~126.01~(7\text{h}15)$ \\
(\textit{best baseline}) & & & & & & & \\
\bottomrule
\end{tabular}}
\label{lpE}
\end{table}

\begin{table}
\centering
\caption[Community detection on Cora using the degeneracy framework]{Community detection on the Cora graph ($n=$ 2~708, $m =$ 5~429), using the standard VGAE~\cite{kipf2016-2} model trained on all cores, and GAE/VGAE variants on the 2-core. All GAE/VGAE models learn embedding vectors of dimension $d = 16$. Scores are averaged over 100 runs. \textbf{Bold} numbers correspond to the best scores, in each table subsection.}
 \resizebox{1.0\textwidth}{!}{
\begin{tabular}{c|c|c|cccc}
\toprule
\textbf{Model}  & \textbf{Size of input} & \textbf{Mean Performance} & \multicolumn{4}{c}{\textbf{Mean Running Times (in sec.)}}\\
& \textbf{$k$-core} & \footnotesize \textbf{AMI (in \%)} & \footnotesize $k$-core dec. & \footnotesize Model train & \footnotesize Propagation & \footnotesize \textbf{Total} \\ 
\midrule
\midrule
VGAE on $\mathcal{G}$ & - & $29.52 \pm 2.61$  & - & $15.34$ & - & $15.34$ \\
on 2-core & $2~136$  & $ 34.08 \pm 2.55$ & $0.16$ & $9.94$ & $0.10$ & $10.20$ \\
on 3-core& $1~257$ & $\textbf{36.29} \pm \textbf{2.52}$ & $0.16$ & $4.43$ & $0.11$ & $4.70$  \\
on 4-core & $174$ & $35.93 \pm 1.88$ & $0.16$ & $\textbf{1.16}$ & $0.12$ & $\textbf{1.44}$ \\

\midrule
VGAE with node features on $\mathcal{G}$ & - & $\textbf{47.25} \pm \textbf{1.80}$ &  - & $15.89$ & - & $15.89$ \\
on 2-core & $2~136$  & $45.09 \pm 1.91$ & $0.16$ & $10.42$ & $0.10$ & $10.68$ \\
on 3-core & $1~257$ & $40.96 \pm 2.06$ & $0.16$ & $4.75$ & $0.11$ & $5.02$  \\
on 4-core & $174$ & $38.11 \pm 1.23$ & $0.16$ & $1.22$ & $0.12$ & $1.50$ \\
\midrule
GAE on 2-core & $2~136$ &$34.91 \pm 2.51$ &  $0.16$ & $10.02$ & $0.10$ & $10.28$ \\
DeepGAE on 2-core & $2~136$ & $35.30 \pm 2.52$ & $0.16$ & $10.12$ & $0.10$ & $10.38$ \\
DeepVGAE on 2-core & $2~136$ & $34.49 \pm 2.85$ & $0.16$ & $10.09$ & $0.10$ & $10.35$ \\
Graphite on 2-core & $2~136$ & $33.91 \pm 2.17$ & $0.16$ & $10.97$ & $0.10$ & $11.23$ \\
Var-Graphite on 2-core & $2~136$ & $33.89 \pm 2.13$  & $0.16$ & $10.91$ & $0.10$ & $11.17$ \\
ARGA on 2-core & $2~136$ & $34.73 \pm 2.84$  & $0.16$ & $9.99$ & $0.10$ & $10.25$ \\
ARVGA on 2-core & $2~136$ &$33.36 \pm 2.53$  & $0.16$ & $9.97$ & $0.10$ & $10.23$ \\
ChebGAE on 2-core & $2~136$ & $\textbf{36.52} \pm \textbf{2.05}$ & $0.16$ & $19.22$ & $0.10$ & $19.48$ \\
ChebVGAE on 2-core & $2~136$ &$37.83 \pm 2.11$  & $0.16$ & $20.13$ & $0.10$ & $20.39$ \\
\midrule
Louvain & - & $\textbf{46.76} \pm \textbf{0.82}$ & - & $1.83$ & - & $1.83$ \\
(\textit{best baseline}) & & & & & & \\
\bottomrule
\end{tabular}
}
\label{cdB}
\end{table}

\begin{table}
\centering
\caption[Community detection on Pubmed using the degeneracy framework]{Community detection on the Pubmed graph ($n=$ 19~717, $m =$ 44~338), using the standard VGAE~\cite{kipf2016-2} model trained on all cores, and GAE/VGAE variants on the 2-core. All GAE/VGAE models learn embedding vectors of dimension $d = 16$. Scores are averaged over 100 runs. \textbf{Bold} numbers correspond to the best scores, in each table subsection.}
 \resizebox{1.0\textwidth}{!}{
\begin{tabular}{c|c|c|cccc}
\toprule
\textbf{Model}  & \textbf{Size of input} & \textbf{Mean Performance} & \multicolumn{4}{c}{\textbf{Mean Running Times (in sec.)}}\\
& \textbf{$k$-core} & \footnotesize \textbf{AMI (in \%)} & \footnotesize $k$-core dec. & \footnotesize Model train & \footnotesize Propagation & \footnotesize \textbf{Total} \\ 
\midrule
\midrule
VGAE on $\mathcal{G}$ & - & $22.36 \pm 0.25$  & - & $707.77$ & - & $707.77$ \\
on 2-core& $10~404$  & $ 23.71 \pm 1.83$ & $1.35$ & $199.07$ & $0.30$ & $200.72$ \\
on 3-core & $6~468$ & $\textbf{25.19} \pm \textbf{1.59}$ & $1.35$ & $79.26$ & $0.34$ & $80.95$  \\
on 4-core & $4~201$ & $24.67 \pm 3.87$ &  $1.35$ & $34.66$ & $0.35$ & $36.36$ \\
on 5-core & $2~630$ & $17.90 \pm 3.76$ &  $1.35$ & $14.55$ & $0.36$ & $16.26$ \\
... & ... & ... & ... & ... & ... & ... \\
on 10-core & $137$ & $10.79 \pm 1.16$ &  $1.35$ & $1.15$ & $0.38$ & $2.88$ \\
\midrule
VGAE with node features on $\mathcal{G}$ & - & $\textbf{26.05} \pm \textbf{1.40}$ &  - & $708.59$ & - & $708.59$ \\
on 2-core & $10~404$  & $24.25 \pm 1.92$ & $1.35$ & $202.37$ & $0.30$ & $204.02$ \\
on 3-core & $6~468$ & $23.26 \pm 3.42$ & $1.35$ & $82.89$ & $0.34$ & $84.58$  \\
on 4-core & $4~201$ & $20.17 \pm 1.73$ & $1.35$ & $36.89$ & $0.35$ & $38.59$ \\
on 5-core & $2~630$ & $18.15 \pm 2.04$ & $1.35$ & $16.08$ & $0.36$ & $17.79$ \\
... & ... & ... & ... & ... & ... & ... \\
on 10-core & $137$ & $11.67 \pm 0.71$ & $1.35$ & $\textbf{0.97}$ & $0.38$ & $\textbf{2.70}$ \\
\midrule
GAE on 2-core & $10~404$ &$22.76 \pm 2.25$ &  $1.35$ & $203.56$ & $0.30$ & $205.21$ \\
DeepGAE on 2-core & $10~404$ & $24.53 \pm 3.30$ & $1.35$ & $205.11$ & $0.30$ & $206.76$ \\
DeepVGAE on 2-core & $10~404$ & $25.63 \pm 3.51$ & $1.35$ & $200.73$ & $0.30$ & $202.38$ \\
Graphite on 2-core & $10~404$ & $26.55 \pm 2.17$  & $1.35$ & $209.12$ & $0.30$ & $210.77$ \\
Var-Graphite on 2-core & $10~404$ & $\textbf{26.69} \pm \textbf{2.21}$ & $1.35$ & $200.86$ & $0.30$ & $202.51$ \\
ARGA on 2-core & $10~404$ & $23.68 \pm 3.18$  & $1.35$ & $207.50$ & $0.30$ & $209.15$ \\
ARVGA on 2-core & $10~404$ &$25.98 \pm 1.93$  & $1.35$ & $199.94$ & $0.30$ & $201.59$ \\
ChebGAE on 2-core & $10~404$ & $25.88 \pm 1.66$ & $1.35$ & $410.81$ & $0.30$ & $412.46$ \\
ChebVGAE on 2-core & $10~404$ &$26.50 \pm 1.49$  & $1.35$ & $399.96$ & $0.30$ & $401.61$ \\
\midrule
node2vec & - & $\textbf{29.57} \pm \textbf{0.22}$ & - & $48.91$ & - & $48.91$ \\
(\textit{best baseline}) & & & & & & \\
\bottomrule
\end{tabular}
}
\end{table}


\begin{table}
\centering
\caption[Community detection on Patent using the degeneracy framework]{Community detection on the Patent graph ($n=$ 2~745~762, $m =$ 13~965~410), using the standard VGAE~\cite{kipf2016-2} model trained on the 14 to 18 cores,  and GAE/VGAE variants on 15-core. All GAE/VGAE models learn embedding vectors of dimension $d = 32$. Scores are averaged over 10 runs. \textbf{Bold} numbers correspond to the best scores, in each table subsection.}
 \resizebox{1.0\textwidth}{!}{
\begin{tabular}{c|c|c|cccc}
\toprule
\textbf{Model}  & \textbf{Size of input} & \textbf{Mean Performance} & \multicolumn{4}{c}{\textbf{Mean Running Times (in sec.)}}\\
& \textbf{$k$-core} & \footnotesize \textbf{AMI (in \%)} & \footnotesize $k$-core dec. & \footnotesize Model train & \footnotesize Propagation & \footnotesize \textbf{Total} \\ 
\midrule
\midrule
VGAE on 14-core & $46~685$  & $\textbf{25.22} \pm \textbf{1.51}$ & $507.08$ & $6~390.37$ & $120.80$ & $7~018.25~(1\text{h}57)$ \\
on 15-core & $35~432$ & $24.53 \pm 1.62$ & $507.08$ & $2~589.95$ & $123.95$ & $3~220.98~(54\text{min})$  \\
on 16-core & $28~153$ & $24.16 \pm 1.96$ & $507.08$ & $1~569.78$ & $123.14$ & $2~200.00~(37\text{min})$ \\
on 17-core & $22~455$ & $24.14 \pm 2.01$ & $507.08$ & $898.27$ & $124.02$ & $1~529.37~(25\text{min})$ \\
on 18-core & $17~799$ & $22.54 \pm 1.98$ & $507.08$ & $\textbf{551.83}$ & $126.67$ & $\textbf{1~185.58~(20\text{min})}$ \\
\midrule
GAE on $15$-core &$35~432$ &$23.76 \pm 2.25$ & $507.08$ & $2~750.09$ & $123.95$ & $3~381.13~(56\text{min})$ \\
DeepGAE on $15$-core & $35~432$ & $24.27 \pm 1.10$ & $507.08$ & $3~007.31$ & $123.95$ & $3~638.34~(1\text{h}01)$ \\
DeepVGAE on $15$-core & $35~432$ & $24.54 \pm 1.23$ & $507.08$ & $2~844.16$ & $123.95$ & $3~475.19~(58\text{min})$ \\
Graphite on $15$-core & $35~432$ & $24.22 \pm 1.45$  & $507.08$ & $2~899.87$ &$123.95$ & $3~530.90~(59\text{min})$ \\
Var-Graphite on $15$-core & $35~432$ & $24.25 \pm 1.51$ & $507.08$ & $2~869.92$ & $123.95$ & $3~500.95~(58\text{min})$ \\
ARGA on $15$-core & $35~432$ & $24.26 \pm 1.18$  & $507.08$ & $3~013.28$ & $123.95$ & $3~644.31~(1\text{h}01)$ \\
ARVGA on $15$-core & $35~432$ &$24.76 \pm 1.32$  & $507.08$ & $2~862.54$ & $123.95$ & $3~493.57~(58\text{min})$ \\
ChebGAE on $15$-core & $35~432$ & $25.23 \pm 1.21$ & $507.08$ & $5~412.12$ & $123.95$ & $6~043.15~(1\text{h}41)$ \\
ChebVGAE on $15$-core & $35~432$ & $\textbf{25.30} \pm \textbf{1.22}$  & $507.08$ & $5~289.91$ & $123.95$ & $5~920.94~(1\text{h}38)$ \\
\midrule
node2vec & - & $\textbf{24.10} \pm \textbf{1.64}$ & - & $26~126.01$ & - & $26~126.01~(7\text{h}15)$ \\
(\textit{best baseline}) & & & & & & \\
\bottomrule
\end{tabular}
}
\label{cdE}
\end{table}

\chapter[Scalable Graph Autoencoders with Stochastic Subgraph Decoding]{Scalable Graph Autoencoders with Stochastic~Subgraph Decoding}\label{chapter_4}

\chaptermark{Scalable Graph Autoencoders with Stochastic Subgraph Decoding}

\textit{This chapter presents research conducted with Romain Hennequin, Jean-Baptiste Remy, Manuel Moussallam, and Michalis Vazirgiannis, and published in Elsevier's Neural Networks journal (impact factor: 8.05) in 2021~\cite{salha2021fastgae}.}

\section{Introduction}

In this chapter, we introduce an alternative stochastic method to scale GAE and VGAE models. This method, referred to as FastGAE in the following, was developed in 2020, i.e., a year after the research presented in Chapter~\ref{chapter_3}. FastGAE constitutes an improvement of our previous efforts towards more scalable graph autoencoders.  

More specifically, we propose to leverage graph mining-based sampling schemes and an effective subgraph decoding strategy to significantly lower the computational complexity of graph autoencoders, while preserving or even slightly improving their performances. We previously argued that a random (uniform) node sampling is suboptimal in the context of GAE and VGAE models, which we will experimentally confirm in Section~\ref{c4s43}. In this chapter, we however explain that, by leveraging graph mining techniques, one can also derive more effective sampling schemes that, in essence, aim to reconstruct ``wisely selected'' random subparts of an original graph during training. We provide an in-depth theoretical and experimental analysis of the proposed solution, showing that it behaves favorably when compared to the degeneracy framework from Chapter~\ref{chapter_3}. We also show that FastGAE addresses some of the limitations of this degeneracy framework.

This chapter is organized as follows.  In Section~\ref{c4s42}, we present and analyze FastGAE, our method for scalable graph autoencoders with stochastic subgraph decoding. We report our experimental evaluation of this method and a discussion of our results in Section~\ref{c4s43}, and we conclude in Section~\ref{c4s44}. In Section~\ref{c4s45}, we provide some proofs as well as an additional figure, placed out of the ``main'' chapter for the sake of brevity and readability.

\section{FastGAE: Scaling GAE and VGAE with Stochastic Subgraph Decoding}
\label{c4s42}

In this section, we introduce our stochastic method to scale GAE and VGAE models. We refer to it as FastGAE when applied to GAEs, and as variational FastGAE when applied~to~VGAEs.

\subsection{A Stochastic Subgraph Decoding Strategy}
\label{c4s421}

\paragraph{Encoding the Entire Graph...}

As detailed in Section~\ref{c3s31} from the previous chapter, the encoding step of GAE and VGAE models can be computationally costly, if the GNN encoders under consideration themselves involve complex operations. Nonetheless, several scalable GNNs have been proposed in the scientific literature. 
We explain in this same section that GCN models~\cite{kipf2016-1} and their scalable extensions \cite{chen2018fastgcn,chiang2019cluster,wu2019simplifying,zeng2020graphsaint} can effectively process large graphs.

Throughout this chapter, we therefore rely on these models to \textit{encode all the nodes} from a graph~$\mathcal{G}$ into a node embedding space. More precisely, in the following experiments, we implement standard GCN encoders~\cite{kipf2016-1} for the sake of simplicity and an easier comparison to the literature. This choice is made without loss of generality. The method described in this section would remain valid for any other encoder producing a node embedding matrix $Z$, and notably for faster and/or less complex variants of GCNs \cite{chen2018fastgcn,chiang2019cluster} in the case of very large graphs (e.g., with hundreds of millions or with  billions of nodes) for which the $O(n)$ complexity of a forward GCN pass would become unaffordable.

\paragraph{...But Decoding Stochastic Subgraphs}
However, while computing node embedding vectors through a forward GCN pass is relatively fast, \textit{tuning the weights} of this encoder in the GAE and VGAE settings requires the reconstruction of the entire matrix $\hat{A}$ at each training iteration which, as explained in the previous chapters, suffers from a quadratic complexity and is intractable for large graphs with more than a few thousand nodes and edges.

 To overcome this issue, we propose to \textit{approximate} reconstruction losses and ELBO objectives, by computing their values only from wisely selected random subparts of the original graph. More precisely, at each training iteration, we aim to decode a different sampled subgraph of $\mathcal{G}$ with $n_{(S)}$ nodes, with $n_{(S)} < n$ being a fixed parameter. Let $\mathcal{G}_{(S)} =
(\mathcal{V}_{(S)},\mathcal{E}_{(S)})$ be such a sampled subgraph, with $\mathcal{V}_{(S)} \subset\mathcal{V}$, $|\mathcal{V}_{(S)}| = n_{(S)}$, and with $\mathcal{E}_{(S)}$ denoting the subset of edges connecting the nodes in $\mathcal{V}_{(S)}$. Instead of reconstructing the $n \times n$ matrix $\hat{A}$, we propose to reconstruct the smaller $n_{(S)} \times n_{(S)}$ matrix $\hat{A}_{(S)}$ with:
\begin{align}\hat{A}_{(S)ij} = \sigma(z^T_i z_j), \hspace{5pt} \forall (i,j) \in \mathcal{V}^2_{(S)},
\end{align}
and to only consider the quality of $\hat{A}_{(S)}$ w.r.t. its ground truth counterpart $A_{(S)}$, as measured by a cross entropy loss for GAEs, and by an ELBO objective for VGAEs. We propose to use the resulting approximate loss for gradients computations and for GCN weights updates by gradient descent. We draw \textit{a different subgraph $\mathcal{G}_{(S)}$ at each training iteration}, using the sampling methods~detailed~in~the~next~section.

\subsection{Node Sampling with Graph Mining}
\label{c4s422}

\paragraph{(Naive) Uniform Node Sampling}

A very simple way to obtain such subgraphs would consist in \textit{uniformly sampling} $n_{(S)}$ nodes from the set $\mathcal{V}$ at each training iteration. However, in such a strategy, there is no guarantee that the most important links (or absence of links) from the original graph structure will be preserved in the drawn subgraphs to reconstruct during the training phase. As we will confirm in Section~\ref{c4s43}, this usually significantly impacts the quality of the final node embedding, leading to underperforming performances on downstream evaluation tasks. As a consequence, in the following sections, we propose and study more refined strategies, aiming to \textit{leverage the graph structure} to obtain a more effective sampling.

\paragraph{Node Sampling with Graph Mining}

We propose to consider alternative sampling methods, that increase the probability of including some particular nodes in the drawn subgraph w.r.t. some others. Let $f: \mathcal{V} \rightarrow \mathbb{R}^+$ denote some measure of the relative \textit{importance} of nodes in the graph, obtained through graph mining methods. Assuming such a function is available, we draw inspiration from word sampling in natural language processing \cite{w2v2,w2v1} and propose to set the probability to pick each node $i \in \mathcal{V}$ as the first element of $\mathcal{V}_{(S)}$ as:
\begin{align}
p_i = \frac{f(i)^{\alpha}}{\sum\limits_{j \in \mathcal{V} } (f(j)^{\alpha})},
\label{eq:alphafastgae}
\end{align}
with $\alpha \in \mathbb{R}^+$. Then, assuming we sample $n_{(S)}$ \textit{distinct} nodes \textit{without replacement}, each remaining node $i \in \mathcal{V} \setminus \mathcal{V}_{(S)}$ has a probability $p_i / \sum_{j \notin \mathcal{V}_{(S)}} p_j$ to be picked as the second element of $\mathcal{V}_{(S)}$, and so on until $|\mathcal{V}_{(S)}| = n_{(S)}$. The previous division is a simple normalization to ensure that $\sum_{j \notin \mathcal{V}_{(S)}} p_j = 1$ at each sampling step. Alternatively, one could also sample $n_{(S)}$ nodes \textit{with replacement}: it simplifies computations, as sampling probabilities are then independent of previous draws and remain fixed to $p_i$, but a node could then be drawn several times. We stress out that, in our implementation, both variants return similar results. We adopt the former.

In a nutshell, important nodes according to $f$ are \textit{more likely to be selected for decoding}, and the hyperparameter $\alpha$ helps sharpening (for $\alpha > 1$) or smoothing (for $\alpha < 1$) the distribution. Setting $\alpha = 0$ leads to the aforementioned uniform node sampling. In our experiments, we will evaluate two importance measures $f$ from graph mining:
\begin{itemize}
    \item the \textit{degree} of each node, i.e., $f(i) = D_{ii} = \sum_{j \in \mathcal{V}} A_{ij}$ following the notation of Definition~\ref{def:degree_mat};
    \item the \textit{core number} of each node, i.e., $f(i) = c(i)$. As presented in Definition~\ref{def:kcore}, the $k$-core version of a graph is its largest subgraph for which every node has a degree higher or equal to $k$ within this subgraph. Here, the core number $c(i)$ of a node $i$ corresponds to the largest value of $k$ for which $i$ is in the $k$-core. This choice constitutes a more \textit{global} importance measure than the \textit{local} node degree.
\end{itemize}
Besides their popularity and their complementarity, we also choose to focus on these two metrics for \textit{computational efficiency}. Indeed, contrary to other potential importance metrics based on influence maximization \cite{kempe2003maximizing}, random walks \cite{leskovec2006sampling} or centrality measures \cite{newman}, both can be evaluated in a linear $O(m)$ running time \cite{batagelj2003}. As we will empirically check in Section~\ref{c4s43}, this permits fast and scalable computations of probability distributions, which is crucial for our FastGAE method whose primary objective is scalability. We refer the interested reader to the work of Leskovec~and~Faloutsos~\cite{leskovec2006sampling} and of Chiericetti~et~al.~\cite{chiericetti2016sampling} for a broader overview of other existing graph sampling methods.

\subsection{Theoretical Considerations}
\label{c4s423}

We now briefly present some theoretical considerations related to FastGAE that, for the sake of readability, will be further developed and proved in the ``supplementary'' Section~\ref{c4s45}.

\paragraph{On Approximate Losses} In the case of degree and core-based sampling strategies, some node pairs from the graph are more likely to appear in subgraphs than others. The probability to draw a node $i$, or an edge incident to $i$, increases with $p_i$ and with $f(i)$ for $\alpha > 0$. As a consequence, at each gradient descent iteration, the approximate loss (say $\mathcal{L}^{\text{\footnotesize FastGAE}}$) is \textit{biased} w.r.t. the standard GAE or VGAE loss that would have been computed on $\mathcal{G}$ (say $\mathcal{L}$), i.e., $\mathbb{E}(\mathcal{L}^{\text{\footnotesize FastGAE}}) \neq \mathcal{L}$ in general. For completeness, in Propositions~\ref{fastgaeprop1},~\ref{fastgaeprop2}~and~\ref{fastgaeprop3} of Section~\ref{c4s45}, \textit{we provide a theoretical analysis, in which we fully explicit the expected loss} $\mathbb{E}(\mathcal{L}^{\text{\footnotesize FastGAE}})$ that we actually stochastically optimize in FastGAE, as well as the formal probabilities to sample a given node or node pair at each training iteration.  Moreover, we will show in Section~\ref{c4s44} that, despite such a bias, optimizing this alternative loss does not deteriorate the quality of node embeddings. On the contrary, we will provide insights exhibiting the fact that re-weighting node pairs from high degree/core nodes can actually be beneficial.

\paragraph{On the Selection of $n_{(S)}$} When selecting $n_{(S)}$, one faces a performance/speed trade-off, as for our degeneracy framework. Reconstructing very small subgraphs speeds up the training but, as we later verify, this might also deteriorate performances. While we claimed in the previous paragraph that stochastically minimizing $\mathbb{E}(\mathcal{L}^{\text{\footnotesize FastGAE}})$ instead of $\mathcal{L}$ might be beneficial, we also acknowledge that, for small values of $n_{(S)}$, the actual loss $\mathcal{L}^{\text{\footnotesize FastGAE}}$ computed at a given training iteration can significantly deviate from its expectation. In this chapter, we propose to use these deviations as a criterion to select a relevant subgraph size. In Propositions~\ref{fastgaeprop4}~and~\ref{fastgaeprop5} of the ``supplementary'' Section~\ref{c4s45}, we leverage concentration inequalities to derive a theoretically-grounded \textit{threshold size}, denoted $n^*_{(S)}$ in the following, for which, at each training iteration, the deviation between the evaluation of $\mathcal{L}^{\text{\footnotesize FastGAE}}$ for each node and its expectation is proven to be bounded with a high probability. This proposed subgraph size is of the form:
\begin{align}
n^*_{(S)} = C \sqrt{n}
\label{eq:threshold}
\end{align}
where the constant $C >0$ depends on the deviation magnitude and probability, and is explicitly presented in Section~\ref{c4s45}. Our experiments will confirm the relevance of this choice.

\subsection{On Complexity and Links to Related Work}
\label{c4s424}

Before diving into experiments, we discuss the complexity of FastGAE and its links and differences w.r.t. some other scalable methods, including our degeneracy from~Chapter~\ref{chapter_3}.

\paragraph{Complexity of FastGAE} As previously detailed, both the GCN encoder and the sampling step of FastGAE have a linear time complexity w.r.t. the number of edges $m$ in the graph. Moreover, our decoder runs in $O(n_{(S)}^2)$ time, with $n_{(S)}$ being significantly smaller than $n$ in practice. In particular, setting  $n_{(S)} = n^*_{(S)}$ ensures a $O(n)$ time complexity for decoding (as $n^{*2}_{(S)} = (C \sqrt{n})^2 = C^2 n$) and an overall $O(m+n)$ linear time complexity for a complete FastGAE training iteration. Faster bounds can also be achieved by lowering $n_{(S)}$ or by replacing GCNs with another encoder. Therefore, as we will empirically verify in Section~\ref{c4s43}, our framework is significantly faster and more scalable than standard GAE and VGAE models. 

\paragraph{Differences with Related Work} At first glance, one might want to compare FastGAE with the methods aiming to scale the GNN/GCN models mentioned in Section~\ref{c2s234}. We would like to emphasize that FastGAE is \textit{not} directly comparable to these methods, e.g., to FastGCN \cite{chen2018fastgcn} that also samples nodes. FastGCN is a GCN-like model, optimized to classify node labels in a (semi) supervised fashion. It samples the \textit{neighborhood} of each node when averaging vector representations in forward passes. 
On the contrary, in this chapter, after \textit{full} GCN forward passes, we instead sample \textit{subgraphs to reconstruct}, in order to approximate the \textit{reconstruction} loss/objective of two unsupervised models, in which GCNs are only a building part (the encoder) of a larger framework (the GAE or the VGAE).
Both settings therefore address different problems. As explained in Section~\ref{c4s421}, methods such as FastGCN could actually be used \textit{in conjunction with} FastGAE, as alternative encoders replacing GCNs.

Furthermore, FastGAE is also more elaborated than data cleaning methods that simply consist in removing some nodes from a graph, e.g., the low-degree ones, to reduce its size. Indeed, in the case of FastGAE with degree sampling, low-degree nodes are still 1) \textit{fully} used in the GCN encoder, and 2) might also appear in \textit{some} subgraphs that we decode (but less often than high-degree nodes). As we leverage new different subgraphs at each iteration, we explore different parts of the \textit{entire} graph during training.

Lastly, we note that \textit{effective subset selection} for faster learning has already provided promising results in the machine learning community \cite{act3,act1}. Contrary to these works, we focus on an unsupervised graph-based problem, and our sampling methods remain fixed throughout learning as we rely on graph mining to select $\mathcal{G}_{(S)}$.

\paragraph{Differences with Chapter~\ref{chapter_3}} Overall, FastGAE is more flexible than the degeneracy framework from Chapter~\ref{chapter_3}, and addresses some of its limitations. For instance, as the degeneracy framework reconstructs one of the cores $\mathcal{C}_0, \dots, \mathcal{C}_{\delta^*(\mathcal{G})}$ during training, it assumes the existence of at least one tractable core subgraph in $\mathcal{G}$. If none of these cores has an appropriate size (because they are all either too large or too small), then it will fail to learn relevant embedding representations. On the contrary, FastGAE does not rely on such an assumption. This method permits reconstructing subgraphs \textit{of any size} $n_{(S)} < n$, where $n_{(S)}$ is a selectable parameter.

In addition, in Chapter~\ref{chapter_3}, we criticized the suboptimal  use of node features during the propagation step. On the contrary, in FastGAE, node features are processed exactly as in standard GCN-based GAE and VGAE models. Lastly, FastGAE is conceptually simpler (in essence, the method consists in a ``smarter'' mini-batch sampling strategy for graph decoding), which we consider being another advantage, as simple solutions often have the most impact.

\section{Experimental Analysis}
\label{c4s43}

In this section, we present an in-depth experimental evaluation of our proposed method to scale GAE and VGAE models. We publicly released the code of FastGAE on GitHub\footnote{\href{https://github.com/deezer/fastgae}{https://github.com/deezer/fastgae}}.

\subsection{Experimental Setting}
\label{c4s431}

\paragraph{Datasets} We provide experiments on seven graphs of increasing sizes. As in Chapter~\ref{chapter_3}, we study the Cora, Citeseer, Pubmed, Google, and Patent graphs, presented in Section~\ref{c3s331}. In the paper associated with this work~\cite{salha2021fastgae}, we also considered two other large graphs:
\begin{itemize}
    \item the Youtube social network of users (edges are friendship connections), available on Konect\footnote{\href{http://konect.cc/networks/}{https://konect.cc/networks/}}, and with $n =$ 3 223 589 nodes and $m =$ 9 375 374 edges;
    \item  a synthetic graph, denoted SBM, generated from a \textit{stochastic block model} which is a generative model for random graphs \cite{abbe2017community}. In this last graph, by design, nodes are clustered in 100 groups of 1000 nodes, acting as ground truth communities. Two nodes from the same community (resp. from different communities) are connected by an edge with probability $2 \times 10^{-2}$ (resp. $2 \times 10^{-4}$). The SBM graph has $n =$ 100 000 nodes and $m =$~1~498~844~edges.
\end{itemize}

Our evaluation therefore includes graphs with various characteristics, sizes, and from four different families (citation networks, social networks, web graphs, and stochastic block model graphs). As in Chapter~\ref{chapter_3}, we consider undirected versions of these graphs. We refer to Chapter~\ref{chapter_5} for an extension of GAE and VGAE models to directed graphs.

\paragraph{Tasks} We consider the \textit{link prediction} and \textit{community detection} tasks (for nodes with ground truth communities), already described in Section~\ref{c3s331}, with a completely similar setting and with the same evaluation metrics.

\paragraph{Models: Standard and FastGAE-based GAE/VGAE}

In the upcoming experiments, for the seven graphs and the two evaluation tasks, we compare standard GAE and VGAE models (when they are tractable) to FastGAE-based versions of these models. All GAE and VGAE models, with and without FastGAE, were optimized for the \textit{link prediction} task. More specifically, we selected the best sets of hyperparameters in terms of mean AUC scores on validation sets. Instructions to easily run a similar validation are provided in our source code.

We trained models for 200 iterations (resp. 300) for graphs with $n <$ 100 000 (resp. $n \geq$ 100 000), and thoroughly checked the convergence of all models for these values (in terms of loss stability in the validation set). Other hyperparameters for these models are described thereafter. 

Our encoders are 2-layer GCNs\footnote{Experiments for Chapter~\ref{chapter_3} considered several different encoders but reported few to no empirical difference w.r.t. 2-layer GCNs. For the sake of brevity and clarity, experiments from this Chapter~\ref{chapter_4} will only report results obtained from 2-layer GCN encoders.} (we tested models with 1 to 3 layers). They include 32-dimensional hidden layers, and 16-dimensional output layer, which means that the dimension of embedding vectors is equal to $d = 16$. We emphasize that we also tested models with $d \in \{32,64,128\}$, reaching similar conclusions w.r.t. $d =16$ (the impact of $d$ is further discussed in Section~\ref{c4s434}). 

Besides, for all models, we used the Adam optimizer \cite{kingma2014adam}, without dropout (we tested models with dropout values in $\{0,0.1,0.2,0.3,0.4,0.5\}$). Regarding learning rates for such an optimizer, we tested values from the grid $\{0.0001,0.0005,0.001,0.005,0.01,0.05,0.1,0.2\}$. We eventually picked a learning rate of 0.1 for Patent with uniform sampling, and of 0.01 otherwise as, once again, these values returned the best mean AUC scores on validation sets. Once again, we mainly used TensorFlow~\cite{abadi2016tensorflow}, training models on an NVIDIA GTX 1080 GPU, and running other operations on a double Intel Xeon Gold 6134 CPU.

\paragraph{Models: Other Baselines} For completeness, we also compare standard GAE/VGAE and FastGAE-based models to the few other existing methods to scale GAE/VGAE:
\begin{itemize}
    \item we consider a simple \textit{negative sampling} strategy, briefly mentioned by Kipf and Welling~\cite{kipf2016-2}. We reconstruct all edges but only $|\mathcal{E}|$ randomly picked unconnected node pairs to compute losses. We leveraged methods made available in PyTorch Geometric~\cite{pytorchgeometric} to estimate losses, with consistent dropout values, learning rates, and architectures w.r.t. the above models;
    \item we also compare to the degeneracy framework from Chapter~\ref{chapter_3}, denoted as \textit{Core-GAE} in the next tables. We used our own implementation~\cite{salha2019-1} with optimal values (regarding mean AUC scores on validation sets) for the hyperparameter $k$ detailed in next tables, and with consistent dropout values, learning rates, and architectures w.r.t. the above models;
    \item besides, while some other sampling ideas were briefly mentioned (as possible extensions) in the recent literature on graph autoencoders \cite{grover2019graphite,salha2020simple}, they actually consist in particular cases of FastGAE, namely with \textit{uniform} sampling. 
\end{itemize}
Lastly, in addition to an extensive comparison between the different GAE/VGAE models, we also report results obtained with three non GAE/VGAE-based baselines: the Louvain method (for community detection)~\cite{blondel2008louvain}, node2vec\footnote{We omit comparison to other random walk-based methods DeepWalk \cite{perozzi2014deepwalk} and LINE \cite{tang2015line} in this chapter, due to quite similar performances w.r.t. node2vec on some of our preliminary tests.}~\cite{grover2016node2vec} and Laplacian eigenmaps~\cite{pedregosa2011scikit,von2007tutorial}. We adopt similar hyperparameters w.r.t. experiments from the previous chapter, with the notable exception that we also set $d=16$ for Laplacian eigenmaps and node2vec (again, the impact of $d$ is further discussed in Section~\ref{c4s434}).

\subsection{Preliminary Results on High Degree/Core Nodes}
\label{c4s432}

Before studying FastGAE we report important insights from preliminary experiments on standard GAE and VGAE models. They motivated the design of our framework and emphasize the relevance of sampling high-degree/core nodes. On the medium-size Cora, Citeseer, and Pubmed graphs, we trained standard GAE and VGAE models, but \textit{tried to mask $k$ nodes and their edges} from the computation of reconstruction losses, for different values of $k$. Such a masking procedure is expected to lower performances, as the model leverages less information about the quality of the reconstruction for learning.

\begin{figure}[!ht]
\centering
  \subfigure[Cora - Degree masking]{
  \scalebox{0.38}{\includegraphics{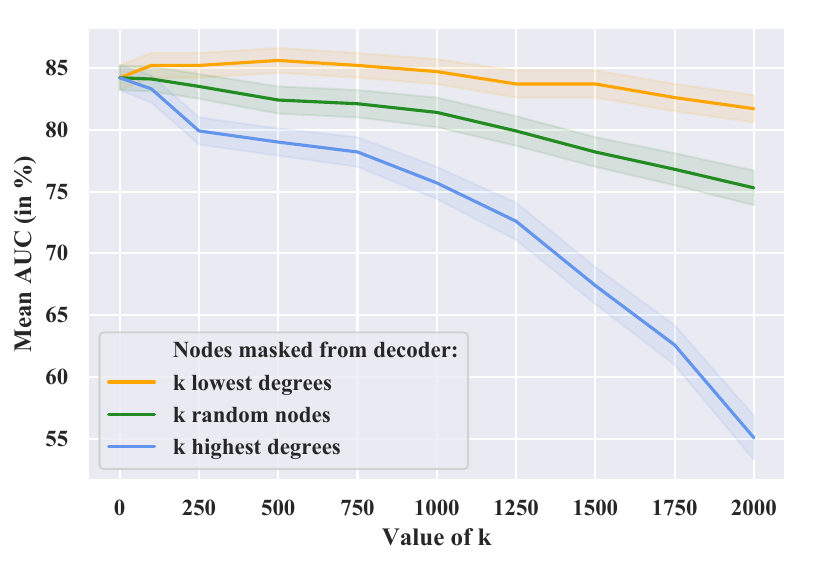}}}\subfigure[Citeseer - Degree masking]{
  \scalebox{0.38}{\includegraphics{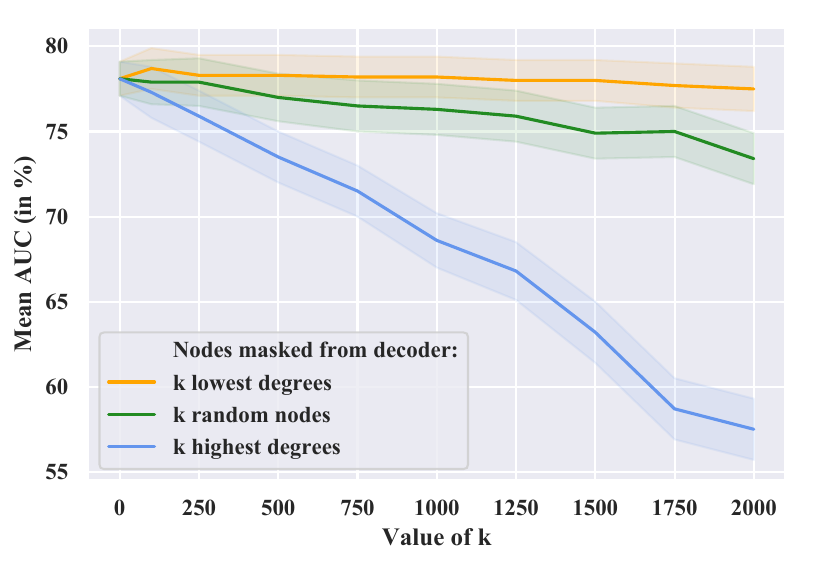}}}\subfigure[Pubmed - Degree masking]{
  \scalebox{0.38}{\includegraphics{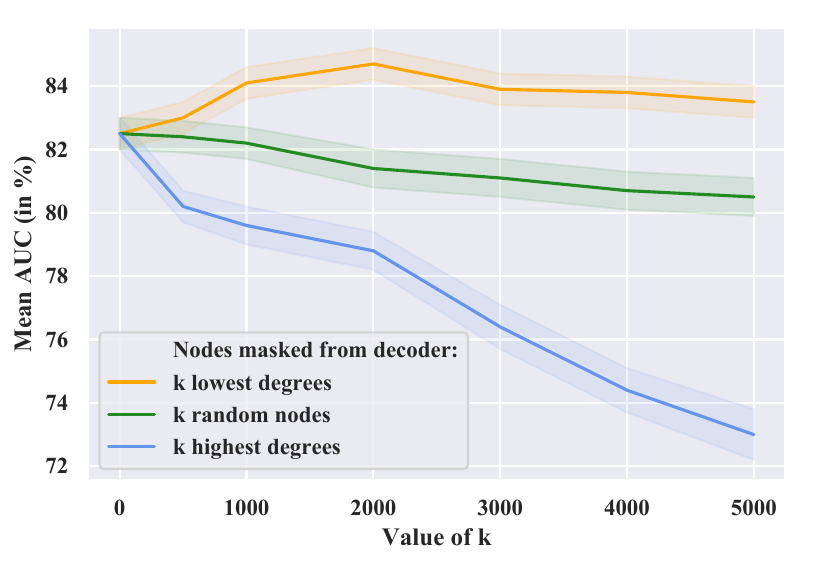}}}
    \subfigure[Cora - Core masking]{
  \scalebox{0.38}{\includegraphics{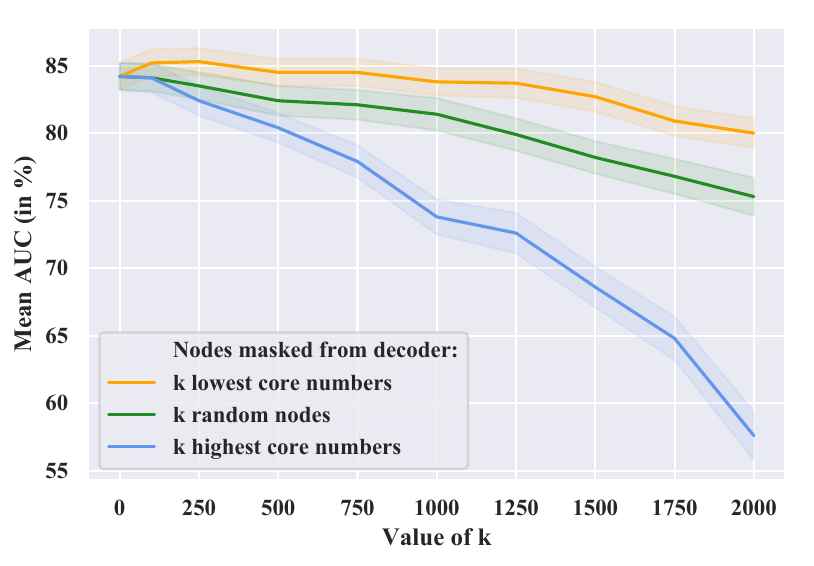}}}\subfigure[Citeseer - Core masking]{
  \scalebox{0.38}{\includegraphics{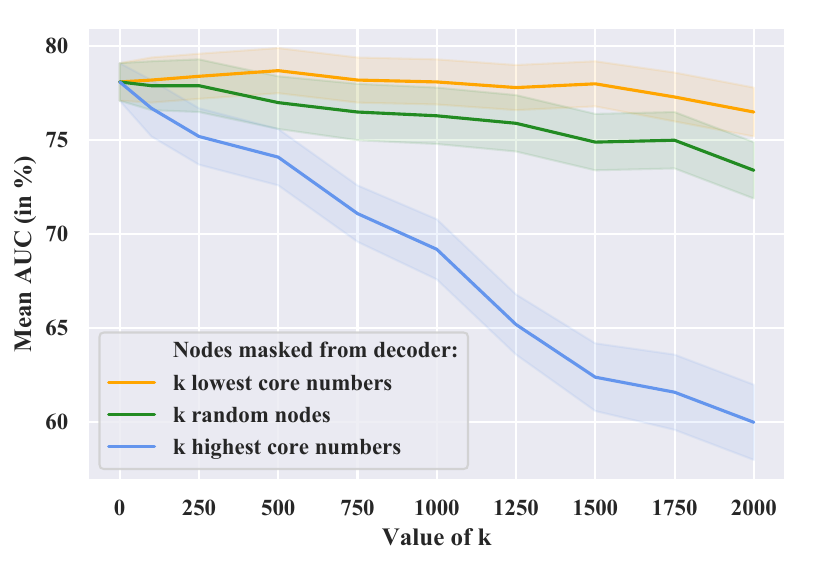}}}\subfigure[Pubmed - Core masking]{
  \scalebox{0.38}{\includegraphics{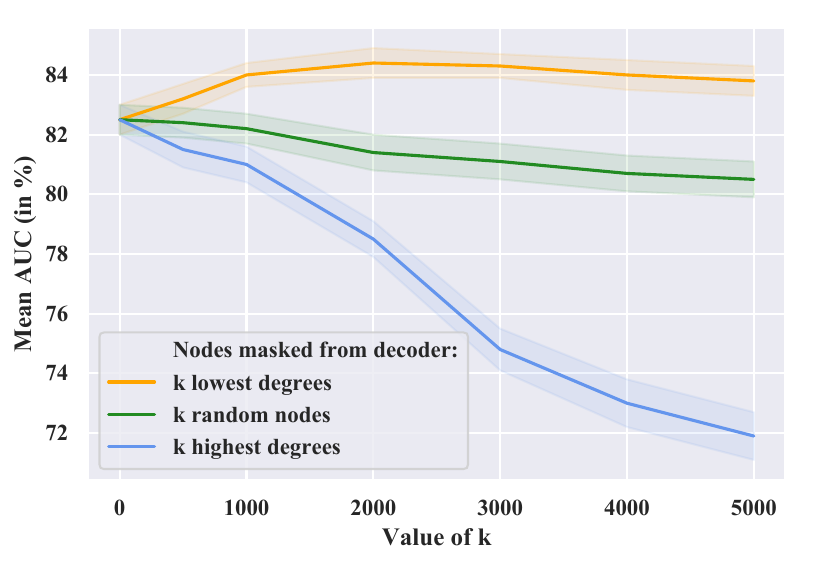}}}
      \caption[Masking $k$ nodes at the decoding phase: impact on link prediction]{Link prediction on the featureless Cora, Citesser and Pubmed graph using standard VGAE models, but trained while masking $k$ nodes and their connections from the decoder/reconstruction loss. AUC scores are averaged over 100 runs with random train/test splits.} 
      \label{cool1}
      \vspace{5pt}
  \subfigure[Cora - Degree masking]{
  \scalebox{0.38}{\includegraphics{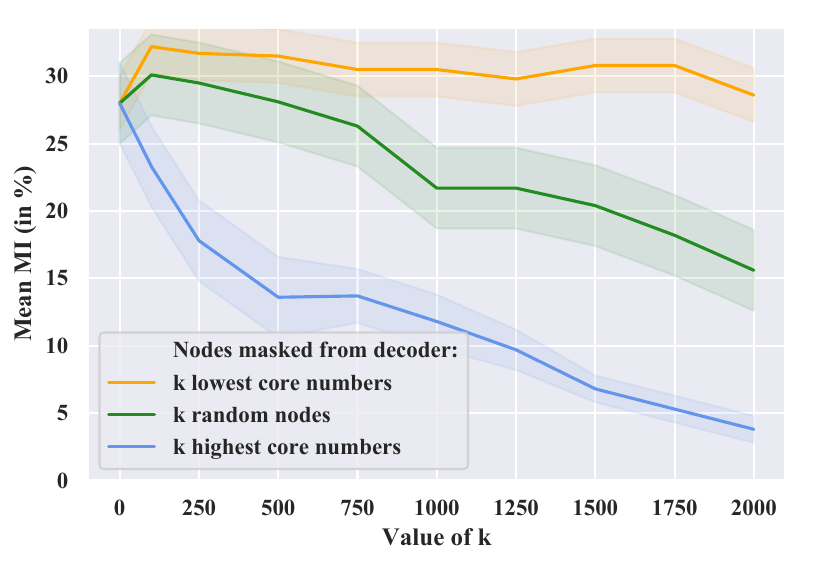}}}\subfigure[Citeseer - Degree masking]{
  \scalebox{0.38}{\includegraphics{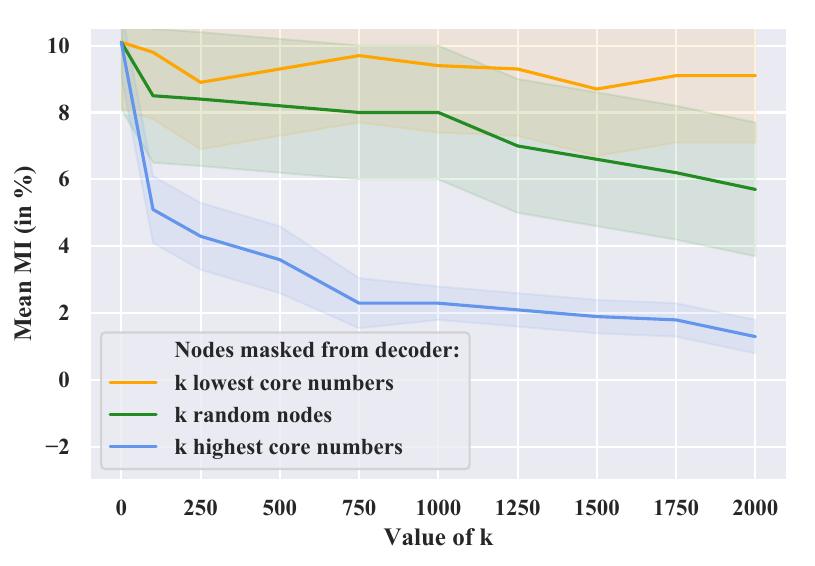}}}\subfigure[Pubmed - Degree masking]{
  \scalebox{0.38}{\includegraphics{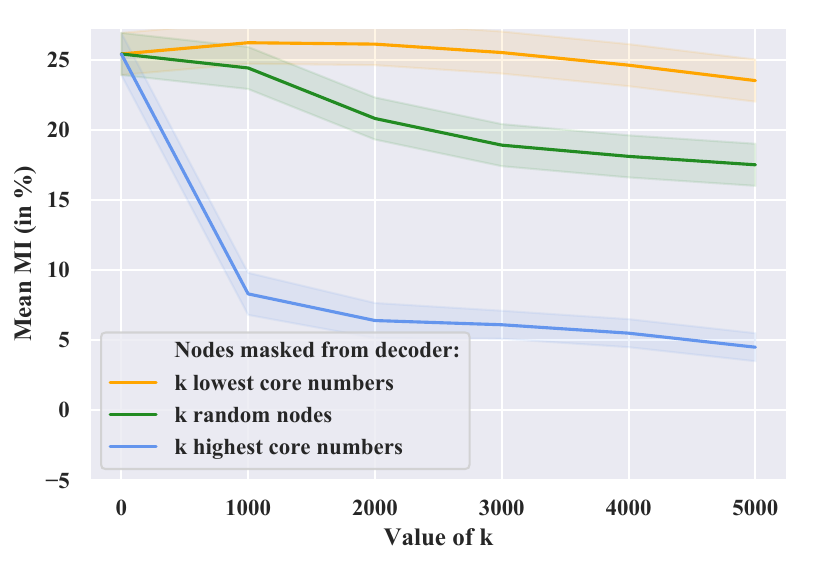}}}
    \subfigure[Cora - Core masking]{
  \scalebox{0.38}{\includegraphics{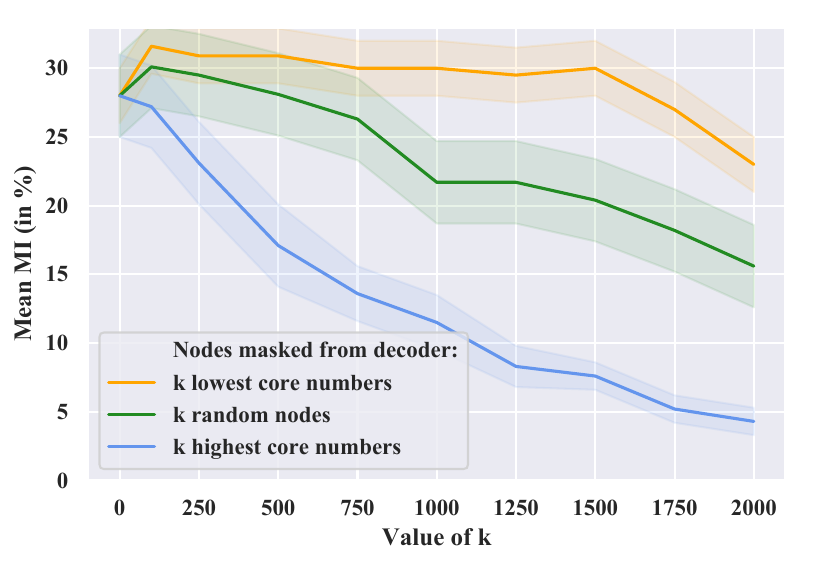}}}\subfigure[Citeseer - Core masking]{
  \scalebox{0.38}{\includegraphics{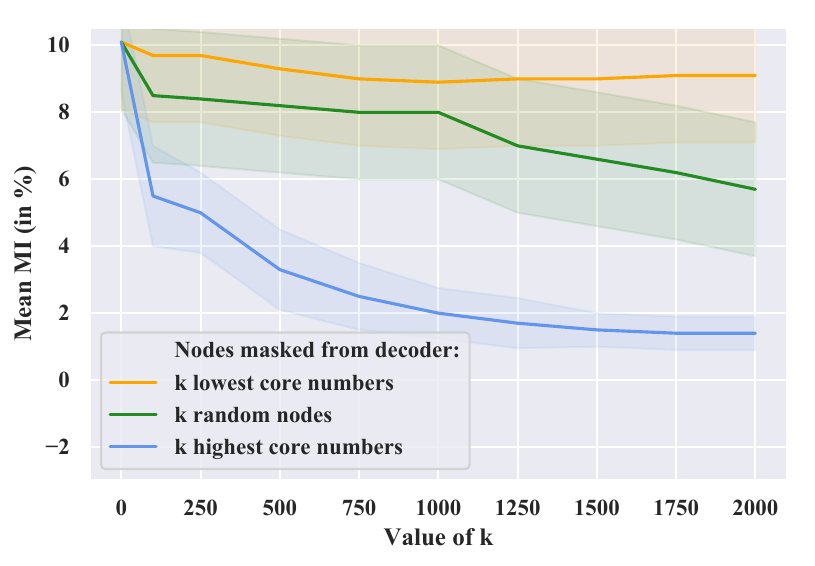}}}\subfigure[Pubmed - Core masking]{
  \scalebox{0.38}{\includegraphics{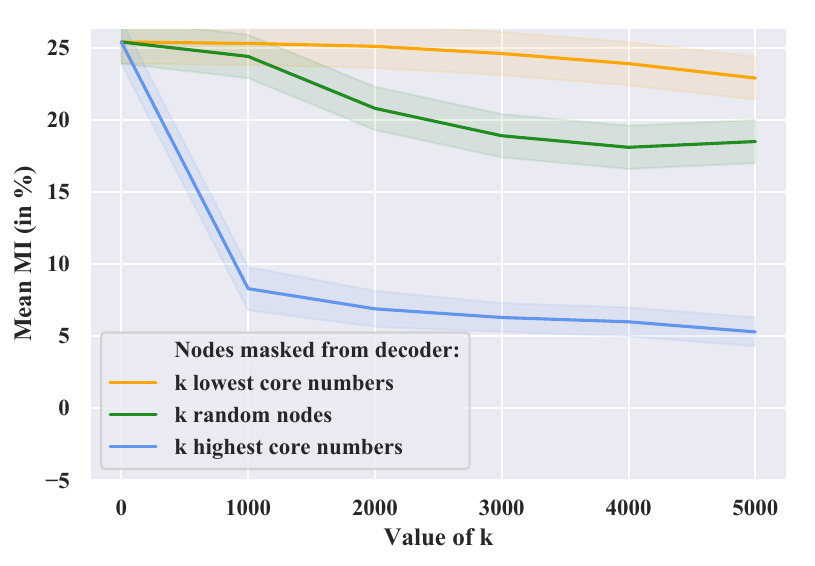}}}
  \caption[Masking $k$ nodes at the decoding phase: impact on community detection]{Community detection on the featureless Cora, Citesser and Pubmed graphs using standard VGAE models, but trained while masking $k$ nodes and their connections from the decoder/reconstruction loss. AMI scores are averaged over 100 runs with random train/test splits.}
  \label{cool2}
\end{figure}

Figure~\ref{cool1} shows that, when these $k$ removed nodes are the \textit{top-$k$ highest degrees/cores nodes}, performances on the link prediction task tumble down. On the contrary, removing the \textit{$k$ nodes with minimal degrees or core numbers} from the loss leads to almost no drop, and even slightly better results on Pubmed, which suggests that removing non-informative nodes might even be beneficial for learning.  In Figure~\ref{cool2}, we report similar results for community detection. These ablation studies suggest that, when implementing stochastic subgraph decoding strategies for scalability, sampling high-degree/core nodes is indeed crucial to learn meaningful embeddings. FastGAE, which explicitly exploits these structural node properties, and optimizes a reconstruction loss that re-weights high degrees/cores node pairs, behaves consistently~w.r.t.~such~insights.

\subsection{Results on Medium-Size Graphs}
\label{c4s433}

\begin{figure*}[t]
\centering
  \subfigure[Pubmed]{
  \scalebox{0.4}{\includegraphics{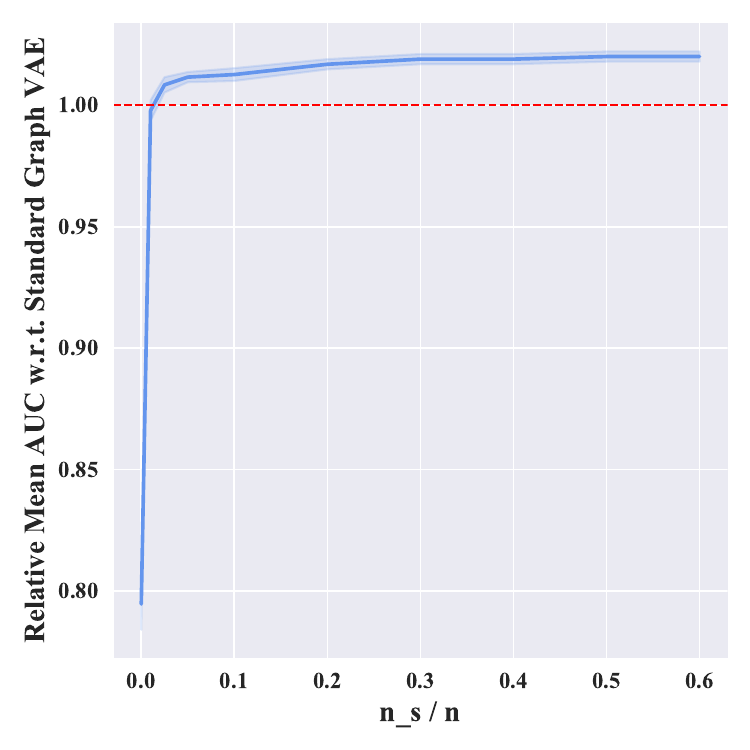}}}\subfigure[Google]{
  \scalebox{0.4}{\includegraphics{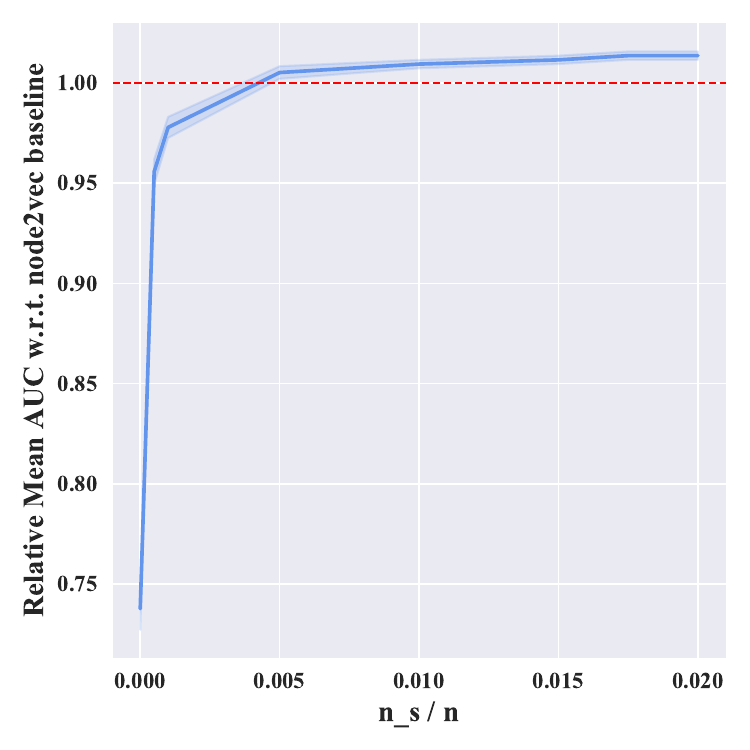}}}\subfigure[Youtube]{
  \scalebox{0.4}{\includegraphics{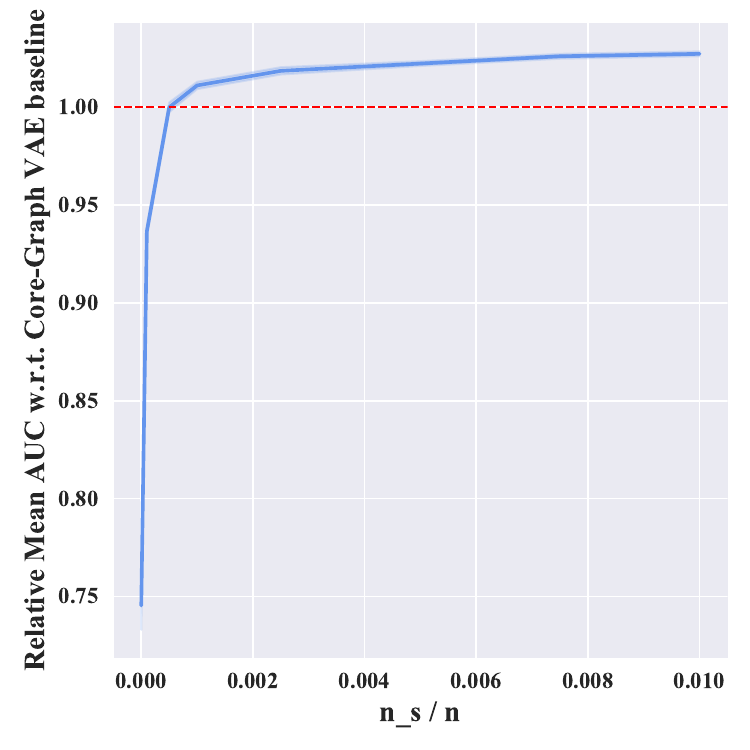}}}
  \caption[Relative AUC depending on the proportion of sampled nodes ]{Summarized results for link prediction on the featureless Pubmed, Google and Youtube graphs: relative mean AUC scores of degree-based Variational FastGAE models w.r.t. standard VGAE (for Pubmed) or w.r.t. the best scalable baseline (for Google and Youtube) depending on the proportion of sampled nodes $n_{(S)}/n$ in decoders. Even for relatively low $n_{(S)}/n$ proportions, Variational FastGAE achieves comparable or slightly better performances w.r.t. a standard VGAE or to the best scalable baseline (results above the red line). The FastGAE paper~\cite{salha2021fastgae} provides similar figures for all other graphs, reaching comparable conclusions.}
  \label{figurefastgae}
\end{figure*}

We now evaluate FastGAE and its variational FastGAE variant. Firstly, we focus on \textit{medium-size graphs}. For Cora, Citeseer, and Pubmed, we can compare FastGAE to standard graph autoencoders. The next Table~\ref{fastgaetable2} details mean AUC and AP scores and standard errors over 100 runs with different train/test splits for link prediction on the (featureless) Pubmed graph with GAE models. For the sake of brevity, we report more summarized results for other medium-size graphs, for VGAE and for community detection, in Table~\ref{fastgaetable3} and Figure~\ref{figurefastgae} (for link prediction) as well as in Table~\ref{fastgaetable4} (for community detection).

\paragraph{FastGAE vs Standard GAE/VGAE} In Table~\ref{fastgaetable2}, we observe that, for sample sizes roughly 20 times smaller than $n$, FastGAE models with \textit{degree} and \textit{core} sampling both achieve competitive or even outperforming\footnote{At first glance, the fact that FastGAE sometimes even slightly \textit{outperforms} standard GAE or VGAE models might be surprising. This improvement is actually consistent with recent research on the benefits of mini-batch-based GNNs \cite{hu2020open,rong2020dropedge}. It comes from the relevance of the two core and degree-based sampling schemes that we consider and from the stochastic nature of the training, which might tend to avoid local minima more easily \cite{kleinberg2018alternative}.} results w.r.t. standard GAE on Pubmed (e.g., +2.31 AUC points for FastGAE with degree sampling and $n_{(S)} =$ 5 000). Furthermore, FastGAE models are also significantly \textit{faster}: in Table~\ref{fastgaetable2} for instance, our approach with degree sampling is up to $\times$~252.78 faster without performance degradation. The additional operation required by our framework, i.e., computing the $p_i$ distribution, is efficient in practice, especially for degree sampling. By further reducing the subgraph size $n_{(S)}$, one can achieve even faster results, while only losing a few AUC/AP points in performance.

\begin{table}[t]
\centering
\caption[Link prediction on Pubmed using FastGAE]{Link prediction on the featureless Pubmed graph ($n=$ 19 717, $m =$ 44 338) using a standard GAE, FastGAE with degree, core and uniform sampling, and other baselines. For degree and core sampling, values of the hyperparameter $\alpha$ (as defined in Equation~\eqref{eq:alphafastgae}) were tuned as illustrated in Figure~\ref{figurealphafastgae}. All GAE models learn embedding vectors of dimension $d=16$. Scores are averaged over 100 runs. \textbf{Bold} numbers correspond to the best scores and best running time. Scores \textit{in italic} are within one standard deviation range from the best ones. Subgraphs sizes annotated with $^*$ correspond to the $n^*_{(S)}$ threshold, as introduced in Equation~\eqref{eq:threshold}.}
\label{fastgaetable2}
\resizebox{1.0\textwidth}{!}{
\begin{tabular}{c|c|cc|ccc|c}
\toprule
\textbf{Model}  & \textbf{Subgraphs} & \multicolumn{2}{c}{\textbf{Average Perf. on Test Set}} & \multicolumn{4}{c}{\textbf{Average Running Times (in seconds)}}\\
& \textbf{size $n_{(S)}$} & \footnotesize \textbf{AUC (in \%)} & \footnotesize \textbf{AP (in \%)} & \footnotesize Compute & \footnotesize Train & \footnotesize \textbf{Total} & \footnotesize Speed gain \\ &  &  & & \footnotesize $p_i$ & \footnotesize model &  & \footnotesize w.r.t. GAE \\
\midrule
\midrule
Standard GAE  & - & 82.51 $\pm$ 0.64 & 87.42 $\pm$ 0.38 & - & 811.43 & 811.43 & - \\
\midrule
FastGAE with & 5~000 & \textbf{84.82} $\pm$ \textbf{0.32} & \textbf{88.19} $\pm$ \textbf{0.23} & 0.01 & 14.41 & 14.42 & $\times$ 56.27 \\
\textbf{degree} sampling & 2~500 & 84.12 $\pm$ 0.40 & 87.56 $\pm$ 0.30 & 0.01 & 5.72 & 5.73 & $\times$ 141.61 \\
($\alpha = 1$)& 1~187$^*$ & 83.67 $\pm$ 0.42 & 87.01 $\pm$ 0.31 & 0.01 & 3.20 & 3.21 & $\times$ 252.78\\
& 500 & 82.68 $\pm$ 0.51 & 85.89 $\pm$ 0.47 & 0.01 & 2.98 & 2.99 & $\times$ 271.38 \\
& 250 & 80.77 $\pm$ 0.55 & 84.05 $\pm$ 0.51 & 0.01 & 2.83 & 2.84 & $\times$ 285.71 \\
\midrule
FastGAE with & 5~000 & \textit{84.62} $\pm$ \textit{0.24} & \textit{88.09} $\pm$ \textit{0.16} & 1.75 & 15.98 & 17.73 & $\times$ 45.77 \\
\textbf{core} sampling & 2~500 & 83.69 $\pm$ 0.34 & 87.28 $\pm$ 0.31 & 1.75 & 7.51 & 9.26 & $\times$ 87.63 \\
($\alpha = 2$) & 1~187$^*$ & 82.53 $\pm$ 0.46 & 86.28 $\pm$ 0.37 & 1.75 & 4.81 & 6.56 & $\times$ 123.69 \\
& 500 & 80.96 $\pm$ 0.52 & 84.86 $\pm$ 0.46 & 1.75 & 4.57 & 6.32 & $\times$ 128.39 \\
& 250 & 79.53 $\pm$ 0.53 & 83.10 $\pm$ 0.50 & 1.75 & 4.44 & 6.19 & $\times$ 131.08 \\
\midrule
FastGAE with & 5~000 & 81.08 $\pm$ 0.48 & 85.90 $\pm$ 0.60 & - & 13.90 & 13.90 & $\times$ 58.37 \\
\textbf{uniform} sampling & 2~500 & 78.72 $\pm$ 0.74 & 83.50 $\pm$ 0.75 & - & 5.48 & 5.48 & $\times$ 148.07 \\
& 1~187$^*$ & 77.28 $\pm$ 0.89 & 81.89 $\pm$ 0.91 & - & 3.10 & 3.10 & $\times$ 261.75 \\
& 500 & 75.09 $\pm$ 2.05 & 78.53 $\pm$ 2.04 & - & 2.98 & 2.98 & $\times$ 271.29\\
& 250 & 74.12 $\pm$ 2.07 & 77.72 $\pm$ 1.22 & - & 2.82 & \textbf{2.82} & $\times$ \textbf{287.74} \\
\midrule
\midrule
Core-GAE, $k=2$ (best choice) & - & 84.30 $\pm$ 0.27 & 86.11 $\pm$ 0.43 & - & 168.91 & 168.91 & $\times$ 4.80 \\
Core-GAE, $k=9$ (fastest choice) & - & 61.65 $\pm$ 0.94 & 64.82 $\pm$ 0.72 & - & 2.92 & 2.92 & $\times$ 277.89 \\
Negative sampling GAE & - & 81.19 $\pm$ 0.68 & 83.21 $\pm$ 0.40 & - & 111.79 & 111.79  & $\times$ 7.28\\
node2vec & - & 81.25 $\pm$ 0.26 & 85.55 $\pm$ 0.26 & - & 48.91 & 48.91 & $\times$ 16.59\\
Laplacian eigenmaps & - & 83.14 $\pm$ 0.42 & 86.55 $\pm$ 0.41 & - & 31.71 & 31.71 & $\times$ 25.59 \\
\bottomrule
\end{tabular}
}
\end{table}

In Table~\ref{fastgaetable3}, Table~\ref{fastgaetable4} and Figure~\ref{figurefastgae}, we consolidate our results by reaching similar conclusions on VGAE, on other medium-size graphs (with and without features) and on community detection. In Figure~\ref{figurefastgae}, we also illustrate that, even for relatively low $n_{(S)}/n$ proportions, our proposed method achieves comparable performances w.r.t. baselines.

\paragraph{Comparison of Uniform, Core-based, and Degree-based FastGAE} In all our experiments, we observe that FastGAE with core and degree sampling both outperform FastGAE (and variational FastGAE) with uniform sampling. Furthermore, core and degree sampling also return more stable scores, i.e., with lower standard errors, especially when the number of samples $n_{(S)}$ is relatively small. Such results confirm the empirical superiority of strategies that leverage the graph structure w.r.t. pure random strategies.

\paragraph{FastGAE vs Baselines} In Table~\ref{fastgaetable2}, Table~\ref{fastgaetable3} and Table~\ref{fastgaetable4}, these models also outperform the other few existing methods to scale GAE and VGAE, usually by a wide margin. For instance, in Table~\ref{fastgaetable2}, we show that, to achieve (almost) comparable link prediction performances w.r.t. FastGAE on Pubmed, our degeneracy framework \textit{Core-GAE} \cite{salha2019-1} requires longer running times (see \textit{Core-GAE} with $k=2$), and that faster variants significantly underperform (almost -20 AUC points for \textit{Core-GAE} with $k=9$ w.r.t. FastGAE with degree sampling). As previously explained, FastGAE is also conceptually simpler than Core-GAE, which we consider to be another advantage of our approach. 

Besides, FastGAE-based models are faster and more effective than the ones leveraging negative sampling~\cite{pytorchgeometric} (e.g., +3.63 AUC points for FastGAE with degree sampling and $n_{(S)} =$ 20 000 w.r.t. \textit{Negative sampling GAE} in Table~\ref{fastgaetable2}). This performance gain might be explained by the more systematic inclusion of \textit{unconnected pairs of important nodes}\footnote{Indeed, when performing negative sampling for GAE, we only reconstruct a few random unconnected node pairs, and ignore the others. However, reconstructing some of these neglected pairs might actually be crucial. Let us consider two nodes with high core number or centrality: knowing that these two important nodes are \textit{not} connected is critical to learn meaningful embeddings. The FastGAE sampling scheme ensures a more systematic inclusion of these important "negative pairs" in the decoding step than negative sampling.} in the losses of FastGAE-based models. Last, but not least, our proposed framework is also competitive w.r.t. the popular non GAE/VGAE-based baselines in most cases. The only exception concerns the community detection experiments on Cora and Citeseer (see Table~\ref{fastgaetable4}) where the Louvain baseline~\cite{blondel2008louvain} outperforms GAE/VGAE models, which we discuss thereafter.

\paragraph{On the hyperparameter $\alpha$} In the ``supplementary'' Section~\ref{c4s45}, we report optimal values of $\alpha$ for all graphs. We recall that $\alpha \in \mathbb{R}^+$ is the hyperparameter introduced in Equation~\eqref{eq:alphafastgae}, which helps balance important and ``less important'' nodes during sampling. Setting $\alpha = 0$ leads to the uniform sampling setting where all nodes are sampled with an equal probability. On the contrary, by setting $\alpha \rightarrow \infty$ we would always sample the most important nodes. Experiments from Figure~\ref{figurealphafastgae} in Section~\ref{c4s45} show that these two extreme cases are usually suboptimal, and that a careful tuning of $\alpha$ (e.g., $\alpha = 2$ for core sampling in Table~\ref{fastgaetable2}) improves performances.

\paragraph{On the threshold $n^*_{(S)}$} In Section~\ref{c4s423}, we introduced a theoretically-grounded threshold $n^*_{(S)} = C \sqrt{n}$ to select the subgraph size. Overall, in all our tables, selecting the proposed $n^*_{(S)}$ provided interesting performance/speed trade-offs, leading to fairly competitive results w.r.t. standard GAE/VGAE models and the best baselines, while being significantly faster.

\subsection{Results on Large Graphs}
\label{c4s434}

\begin{table}[t]
\centering
\caption[Link prediction on Patent using FastGAE]{Link prediction on the Patent graph ($n =$ 2 745 762, $m =$ 13 965 410), using FastGAE with degree, core and uniform sampling, and other baselines. The standard GAE is intractable for this graph. For degree and core sampling, values of the hyperparameter $\alpha$ (as defined in Equation~\eqref{eq:alphafastgae}) were tuned as illustrated in Figure~\ref{figurealphafastgae}. All GAE models learn embedding vectors of dimension $d=16$. Scores are averaged over 10 runs. \textbf{Bold} numbers correspond to the best scores and best running time. Scores \textit{in italic} are within one standard deviation range from the best one. Subgraphs sizes annotated with $^*$ correspond to the $n^*_{(S)}$ threshold, as introduced in Equation~\eqref{eq:threshold}.}
\label{fastgaetable5}
\resizebox{1.0\textwidth}{!}{
\begin{tabular}{c|c|cc|ccc}
\toprule
\textbf{Model}  & \textbf{Subgraphs} & \multicolumn{2}{c}{\textbf{Average Perf. on Test Set}} & \multicolumn{3}{c}{\textbf{Average Running Times (in seconds)}}\\
 & \textbf{size $n_{(S)}$} & \footnotesize \textbf{AUC (in \%)} & \footnotesize \textbf{AP (in \%)} & \footnotesize Compute  & \footnotesize Train& \footnotesize \textbf{Total} \\ 
 &  & \footnotesize & \footnotesize \ & \footnotesize $p_i$ & \footnotesize model &  \\ 
\midrule
\midrule
Standard GAE  & - &  \multicolumn{2}{c|}{\textit{(intractable)}}  & \multicolumn{3}{c}{\textit{(intractable)}} \\
\midrule
FastGAE with & 20~000 & \textit{92.91} $\pm$ \textit{0.22} & \textit{93.35} $\pm$ \textit{0.21} & 0.30 & 4~401.67 & 4~401.97 (1h13)  \\
\textbf{degree} sampling   & 16~425$^{*}$ & \textbf{93.02} $\pm$ \textbf{0.23} & \textbf{93.39} $\pm$ \textbf{0.23} & 0.30 & 3~693.32 & 3~693.62 (1h02)  \\
($\alpha = 2$)& 10~000 & 91.76 $\pm$ 0.23 & 91.74 $\pm$ 0.21 & 0.30 & 1~164.22 & 1~164.52 (19 min) \\
& 2~500 & 87.53 $\pm$ 0.50 & 87.42 $\pm$ 0.51 & 0.30 & 537.99 & 538.29 (9 min)  \\
& 1~000 & 85.55 $\pm$ 0.62 & 85.96 $\pm$ 0.55 & 0.30 & 500.12 & 500.42 (8 min) \\
\midrule
FastGAE with & 20~000 & 90.71 $\pm$ 0.21 & 91.70 $\pm$ 0.19 & 668.05 & 4~800.58 & 5~468.63 (1h31)  \\
\textbf{core} sampling & 16~425$^{*}$ & 90.48 $\pm$ 0.21 & 90.85 $\pm$ 0.23 & 668.05 & 4~027.90 & 4~695.95 (1h18) \\
($\alpha = 2$) & 10~000 & 89.08 $\pm$ 0.25 & 88.65 $\pm$ 0.24 & 668.05 & 1~232.03 & 1~900.08 (32 min)  \\
& 2~500 & 82.50 $\pm$ 0.51 & 81.42 $\pm$ 0.60 & 668.05 & 544.64 & 1~222.69 (20 min)  \\
 & 1~000 & 73.99 $\pm$ 0.70 & 75.24 $\pm$ 0.74 & 668.05 & 503.88 & 1~171.93 (19 min)  \\
\midrule
FastGAE with & 20~000 & 85.97 $\pm$ 0.26 & 87.71 $\pm$ 0.25 & - & 4~397.89 & 4~387.89 (1h13)  \\
\textbf{uniform} sampling & 16~425$^{*}$ & 84.40 $\pm$ 0.25 & 86.11 $\pm$ 0.25 & - & 3~602.66 & 3~602.66 (1h00) \\
 & 10~000 & 83.77 $\pm$ 0.28 & 83.37 $\pm$ 0.26 & - & 1~106.01 & 1~106.01 (18 min)  \\
& 2~500 & 70.66 $\pm$ 0.35 & 71.16 $\pm$ 0.38 & - & 485.03 & 485.03 (8 min)  \\
& 1~000 & 59.34 $\pm$ 0.83 & 58.83 $\pm$ 1.30 & - & 438.02 & \textbf{438.02 (7 min)} \\
\midrule
\midrule
Core-GAE, $k=14$ (best choice) & - & 88.06 $\pm$ 0.27 & 88.94 $\pm$ 0.23 & - & 4~805.11 & 4~805.11 (1h20) \\
Core-GAE, $k=21$ (fastest choice) & - & 86.94 $\pm$ 0.69 & 87.23 $\pm$ 0.71 & - & 619.01 & 619.01 (10 min)  \\
Negative sampling GAE & - & 86.11 $\pm$ 0.48 & 86.70 $\pm$ 0.49 & - & 2~392.96 & 2~392.96 (40 min)  \\
node2vec & - & \textit{92.96} $\pm$ \textit{0.23} & \textit{93.36} $\pm$ \textit{0.20} & - & 25~851.39 & 25~851.39 (7h11)  \\
Laplacian eigenmaps & - & \multicolumn{2}{c|}{\textit{(intractable)}}   & \multicolumn{3}{c}{\textit{(intractable)}} \\
\bottomrule
\end{tabular}
}
\end{table}

We now report the evaluation of FastGAE and variational FastGAE on the four large graphs from our experiments: SBM, Google, Youtube, and Patent. The above Table~\ref{fastgaetable5} details mean AUC  and  AP scores and standard errors over 10 runs with different train/test splits for link prediction on the Patent graph with FastGAE. We also report more summarized results (for the sake of brevity) for link prediction on SBM, Google, Youtube, and Patent in Table~\ref{fastgaetable6}, and summarized results for community detection on SBM in Table~\ref{fastgaetable4}. As in Table~\ref{fastgaetable5}, all scores are averaged over 10 runs with different train/test splits for link prediction.

\paragraph{FastGAE vs Scalable GAE/VGAE Baselines} On large graphs, a direct comparison with standard GAEs and VGAEs is unfortunately impossible. However, our FastGAE and variational FastGAE models almost always outperform the other existing approaches to scale GAE and VGAE models, usually by a wide margin. For instance, for link prediction on Patent (Table~\ref{fastgaetable5}), degree-based and core-based FastGAE models with $n_{(S)} =$ 20~000, 16~425 and 10~000 all outperform the best \textit{Core-GAE} by up to roughly 5 AUC points (for degree-based FastGAE with $n_{(S)} =$ 20~000) and with comparable or better running times.
Regarding the \textit{Core-GAE} baseline \cite{salha2019-1}, we also point out that, in one of our large graphs, namely on the SBM one, this method was even \textit{intractable} due to the lack of \textit{size decreasing} core structure on this graph. Indeed, the $21$-core of SBM includes 95~200 nodes, which is too large to train a graph AE or VAE on our machines, and the $22$-core is empty. Requiring a size decreasing core structure is a drawback of \textit{Core-GAE} w.r.t. the more flexible FastGAE approach, which we already mentioned in Section~\ref{c4s424}.

Moreover, as for medium-size graphs, we also observe that core-based and degree-based FastGAE tend to significantly outperform negative sampling (e.g., up to +6.8 AUC points for link prediction on Patent in Table~\ref{fastgaetable5}. Also, \textit{Negative sampling GAE} never appears as the best baseline in Table~\ref{fastgaetable4} nor in Table~\ref{fastgaetable6}), consolidating our previous conclusions.
Besides, as before, the proposed $n^*_{(S)}$ provides quite effective performance/speed trade-offs and will constitute an interesting heuristic to help future FastGAE users select subgraph sizes.

\paragraph{Comparison of Uniform, Core-based and Degree-based FastGAE} As for medium-size graphs, core-based sampling and degree-based sampling are empirically more effective than uniform sampling (e.g., in Table~\ref{fastgaetable5}, +6.94 AUC points for FastGAE with degree sampling on Patent, with $n_{(S)} =$ 20~000), and associated with lower standard errors. We observe that computing the $p_i$ probabilities through core-based sampling is longer on large graphs, but brings no empirical benefit w.r.t. degree-based sampling: we therefore recommend using degree-based sampling for large graphs.

\paragraph{FastGAE vs Non-GAE/VGAE Baselines, and the Case of Community Detection} For the link prediction task, the best FastGAE models usually reach competitive results w.r.t. node2vec while being significantly faster. However, regarding community detection, we observe in Table~\ref{fastgaetable4} that the \textit{Louvain} baseline outperforms GAE and VGAE models on SBM, a phenomenon that we also noted on the Cora and Citeseer graphs in the previous section. We conjecture that current graph autoencoders models might be suboptimal to effectively reconstruct communities in graph data. This claim is consistent with some of our experiments from Chapter~\ref{chapter_3} and elsewhere in the scientific literature~\cite{choong2018learning}.  As our objective, in the paper associated with FastGAE~\cite{salha2021fastgae}, was to scale existing GAE and VGAE models, but not to ensure nor to claim their superiority over all other methods for community detection, we did not further investigate this limit in this paper. Nonetheless, in Chapter~\ref{chapter_7}, we will propose a method, referred to as Modularity-Aware Graph Autoencoders~\cite{salhagalvan2022modularity}, to improve the reconstruction of communities from GAE/VGAE-based node embedding spaces. 

\paragraph{On the embedding dimension $d$} Our tables present results for $d=16$ for all methods and all graphs. Nonetheless, we reached similar conclusions for $d = 32$, $64$ and $128$: although performances sometimes slightly improved by increasing $d$, the \textit{ranking} of the different models remained unchanged. We also considered re-optimizing $d$ individually for each model as in Chapter~\ref{chapter_3} (to cover potential cases where the impact of $d$ on the performance of each model would have been different) but, again, it did not modify the ranking of these models in terms of AUC, AP and AMI scores.

\paragraph{On the number of training iterations} 
As detailed in Section~\ref{c4s431}, all GAE and VGAE models, with or without our FastGAE framework, were trained for 200 iterations (resp. 300) for graphs with $n <$ 100~000 (resp. $n \geq$ 100~000). We thoroughly checked the convergence of all models, by assessing the stabilization of performances in terms of AUC scores on validation sets. Using a fixed number of iterations is common in recent research on GAE and VGAE \cite{berg2018matrixcomp,semiimplicit2019,kipf2016-2}. We nonetheless think that \textit{early-stopping} \cite{girosi1995regularization} would also be a relevant alternative strategy, that could lead to additional speed-ups, and might deserve further investigations in future works. Besides, we observed that, for very small values of $n_{(S)}$, increasing the number of training iterations did not significantly improve our results. To improve scores on such settings, increasing the sampling size $n_{(S)}$ was overall more effective than increasing the number of training iterations.

\begin{table}[t]
\centering
\caption[Link prediction on all medium-size graphs using FastGAE]{Link prediction on all medium-size graphs. For each graph, for brevity, we only report the \textbf{best} GAE \textbf{or} VGAE model in terms of AUC and AP scores, a few representative degree-based FastGAE versions of this model, and the best baseline (among Core-GAE/VGAE, Negative sampling GAE/VGAE, node2vec and Laplacian eigenmaps). Scores are averaged over 100 runs. For degree sampling, values of the hyperparameter $\alpha$ (as defined in Equation~\eqref{eq:alphafastgae}) were tuned as illustrated in Figure~\ref{figurealphafastgae}. All GAE/VGAE models learn embedding vectors of dimension $d=16$. \textbf{Bold} numbers correspond to the best scores and best running time. Scores \textit{in italic} are within one standard deviation range from the best score.}
\label{fastgaetable3}
 \resizebox{1.0\textwidth}{!}{
\begin{tabular}{c|l|cc|ccc|c}
\toprule
\textbf{Dataset}  & \textbf{Model}  & \multicolumn{2}{c}{\textbf{Average Perf. on Test Set}} & \multicolumn{3}{c}{\textbf{Avg. Run. Times (in sec.)}} & \textbf{Speed}\\
&  & \footnotesize \textbf{AUC (in \%)} & \footnotesize \textbf{AP (in \%)} & \footnotesize Comp. & \footnotesize Train & \footnotesize \textbf{Total} &  \textbf{Gain} \\
&  &  &  & \footnotesize $p_i$ & \footnotesize model & \footnotesize  &  \\ 
\midrule
\midrule
 & Standard GAE & 84.79 $\pm$ 1.10 & \textbf{88.45} $\pm$ \textbf{0.82} & - & 3.87 & 3.87 & - \\
 & \underline{FastGAE (degree, $\alpha = 2$)} &  &  &  & & \\
\textbf{Cora} & - with $n_{(S)} =$ 250  & 84.13 $\pm$ 1.20 & 86.65 $\pm$ 1.23 & 0.002 & 1.46 & \textbf{1.462} & $\times$ \textbf{2.65} \\
 & -  with $n_{(S)} = n^*_{(S)} =$ 440  & 84.74 $\pm$ 0.81 & 87.42 $\pm$ 0.75 & 0.002 & 1.56 & 1.562 & $\times$ 2.48 \\
 &  - with  $n_{(S)} =$ 1~000& 84.75 $\pm$ 0.84 & \textit{87.77} $\pm$ \textit{0.81} & 0.002 & 1.65 & 1.652 & $\times$ 2.34 \\
& \underline{Best baseline}  &  &  &  & & \\
& Laplacian eigenmaps & \textbf{86.49} $\pm$ \textbf{0.98} & \textit{87.42} $\pm$ \textit{1.04} & - & 2.49 & 2.49 & $\times$ 1.55 \\
\midrule

 & Standard VGAE & \textit{91.64} $\pm$ \textit{0.92} & \textbf{92.66} $\pm$ \textbf{0.91} & - & 4.25 & 4.25 & - \\
 & \underline{Var. FastGAE (degree, $\alpha = 2$)} &  &  &  & & \\
\textbf{Cora} & - with $n_{(S)} =$ 250 & 90.50 $\pm$ 1.10 & 91.10 $\pm$ 1.08 & 0.002 & 2.30 & \textbf{2.302} & $\times$ \textbf{1.85} \\
\textbf{with} & - with $n_{(S)} = n^*_{(S)} =$ 440  & 90.82 $\pm$ 1.07 & 91.44 $\pm$ 1.13 & 0.002 & 2.52 & 2.522 & $\times$ 1.69 \\
\textbf{features}& - with $n_{(S)} =$ 1~000  & \textbf{91.72} $\pm$ \textbf{0.98} & \textit{92.36} $\pm$ \textit{1.11} & 0.002 & 2.87 & 2.872 & $\times$ 1.48 \\
& \underline{Best baseline}  &   &  &  & & \\
& Core-VGAE, $k=2$ & 87.94 $\pm$ 1.12 & 89.00 $\pm$ 1.11 & - & 3.09 & 3.09 & $\times$ 1.38 \\
\midrule

 & Standard GAE & 78.25 $\pm$ 1.69 & \textbf{83.79} $\pm$ \textbf{1.24} & - & 5.25 & 5.25 & - \\
 & \underline{FastGAE (degree, $\alpha = 1$)}  &  &  &  & & \\
\textbf{Citeseer} & - with $n_{(S)} =$ 250 & 77.28 $\pm$ 1.11 & 81.29 $\pm$ 0.92 & 0.002 & 1.47 & \textbf{1.472} & $\times$ \textbf{3.57}\\
 & - with $n_{(S)} = n^*_{(S)} =$ 488 & 78.30 $\pm$ 1.30 & \textit{82.42} $\pm$ \textit{1.09} & 0.002 & 1.58 & 1.582 & $\times$ 3.32 \\
 & - with $n_{(S)} =$ 1~000 & 78.31 $\pm$ 1.25 & \textit{82.40} $\pm$ \textit{0.99} & 0.002 & 1.61 & 1.612 & $\times$ 3.26  \\
& \underline{Best baseline} &  &  &  & & \\
& Laplacian eigenmaps  & \textbf{80.42} $\pm$ \textbf{1.38} & \textit{83.75} $\pm$ \textit{1.12} & - & 3.50 & 3.50 & $\times$ 1.50 \\
\midrule

 & Standard VGAE & \textbf{90.72} $\pm$ \textbf{1.01} & \textbf{92.05} $\pm$ \textbf{0.97} & - & 6.28 & 6.28 & - \\
 & \underline{Var. FastGAE (degree, $\alpha = 1$)}   &   &  &  & & \\
\textbf{Citeseer}  & - with $n_{(S)} =$ 250  & 89.37 $\pm$ 1.69 & 89.63 $\pm$ 1.83 & 0.002 & 2.32 & \textbf{2.322} & $\times$ \textbf{2.70} \\
\textbf{with} &  - with $n_{(S)} = n^*_{(S)} =$ 488  & \textit{90.10} $\pm$ \textit{1.33} & 90.15 $\pm$ 1.50 & 0.002 & 2.62 & 2.622 & $\times$ 2.40\\
\textbf{features} & - with $n_{(S)} =$ 1~000 & \textit{90.22} $\pm$ \textit{1.14} & 90.16 $\pm$ 1.20 & 0.002 & 2.89 & 2.892 & $\times$ 2.17 \\
& \underline{Best baseline}   &   &  &  & & \\
& Core-VGAE, $k=2$ & 81.85 $\pm$ 1.72 & 83.65 $\pm$ 1.64 & - & 2.55 & 2.55 & $\times$ 2.46 \\
\midrule

 & Standard GAE & 82.51 $\pm$ 0.64 & 87.42 $\pm$ 0.38 & - & 811.43 & 811.43& - \\
& \underline{FastGAE (degree, $\alpha = 1$)} &  &   &  &  & & \\ 
\textbf{Pubmed} & - with $n_{(S)} =$ 500  & 82.68 $\pm$ 0.51 & 85.89 $\pm$ 0.47 & 0.01 & 2.98 & \textbf{2.99} & $\times$ \textbf{271.38} \\ 
 & - with $n_{(S)} = n^*_{(S)} =$ 1~187 & 83.67 $\pm$ 0.42 & 87.01 $\pm$ 0.31 & 0.01 & 3.20& 3.21 & $\times$ 252.78 \\ 
& - with $n_{(S)} =$ 5~000 & \textbf{84.82} $\pm$ \textbf{0.32} & \textbf{88.19} $\pm$ \textbf{0.23} & 0.01 & 14.41 & 14.42 & $\times$ 56.27 \\
& \underline{Best baseline} &  &   &  &  & & \\ 
& Core-GAE, $k=2$  & 84.30 $\pm$ 0.27 & 86.11 $\pm$ 0.43 & - & 168.91 & 168.91 & $\times$ 4.80 \\
\midrule

 & Standard GAE & \textbf{96.28} $\pm$ \textbf{0.36} & \textit{96.29} $\pm$ \textit{0.25} & - & 952.63 & 952.63 & - \\
 & \underline{FastGAE (degree, $\alpha = 1$)} &  &   &  &  & \\ 
\textbf{Pubmed} & - with $n_{(S)} =$ 500 & 95.08 $\pm$ 0.45 & 95.24 $\pm$ 0.46 & 0.01 & 3.53 & \textbf{3.54} & $\times$ \textbf{269.10}\\ 
\textbf{with} & - with $n_{(S)} = n^*_{(S)} =$ 1~187 & 95.45 $\pm$ 0.26 & 95.70 $\pm$ 0.30 & 0.01 & 4.01 & 4.02 & $\times$ 237.56\\ 
\textbf{features} & - with $n_{(S)} =$ 5~000 & \textit{96.12} $\pm$ \textit{0.20} & \textbf{96.35} $\pm$ \textbf{0.19} & 0.01 & 19.74 & 19.75 & $\times$ 48.23 \\
& \underline{Best baseline} &   &   &  &  & & \\ 
& Core-GAE, $k=2$ & 85.34 $\pm$ 0.33 & 86.06 $\pm$ 0.24 & - & 40.22 & 40.22 & $\times$ 23.69 \\
\bottomrule
\end{tabular}
}
\end{table}

\begin{table}[t]
\centering
\caption[Link prediction on all large graphs using FastGAE]{Link prediction on all large graphs. For each graph, for brevity, we only report the \textbf{best} GAE \textbf{or} VGAE model in terms of AUC and AP scores, a few representative degree-based FastGAE versions of this model, and the best baseline (among Core-GAE/VGAE, Negative sampling GAE/VGAE and node2vec). Scores are averaged over 10 runs. For degree sampling, values of the hyperparameter $\alpha$ (as defined in Equation~\eqref{eq:alphafastgae}) were tuned as illustrated in Figure~\ref{figurealphafastgae}. All GAE/VGAE models learn embedding vectors of dimension $d=16$. \textbf{Bold} numbers correspond to the best scores and best running times. Scores \textit{in italic} are within one standard deviation range from the best score.}
\label{fastgaetable6}
 \resizebox{1.0\textwidth}{!}{
\begin{tabular}{c|l|cc|ccc}
\toprule
\textbf{Dataset}  & \textbf{Model}  & \multicolumn{2}{c}{\textbf{Average Perf. on Test Set}} & \multicolumn{3}{c}{\textbf{Average Running Times (in sec.)}} \\
& & \footnotesize \textbf{AUC (in \%)} & \footnotesize \textbf{AP (in \%)} & \footnotesize Compute & \footnotesize Train & \footnotesize \textbf{Total} \\ 
&  &  & & \footnotesize $p_i$ & \footnotesize model & \\ 
\midrule
\midrule

 & Standard VGAE  & \multicolumn{2}{c|}{\textit{(intractable)}}  & \multicolumn{3}{c}{\textit{(intractable)}}  \\
 & \underline{Var. FastGAE (degree, $\alpha = 2$)}&  &  &  &  & \\
\textbf{SBM} & - with $n_{(S)} =$ 2~000 & 79.37 $\pm$ 0.52 & 80.68 $\pm$ 0.84 & 0.03 & 27.36 & \textbf{27.39} \\
 & - with $n_{(S)} = n^*_{(S)} =$ 2~673 & 80.96 $\pm$ 0.35 & 83.69 $\pm$ 0.60 & 0.03 & 30.66 & 30.69\\
 & - with $n_{(S)} =$ 5~000 & \textbf{81.45} $\pm$ \textbf{0.39} & \textbf{84.30} $\pm$ \textbf{0.82} & 0.03 & 43.86 & 43.89 \\
& \underline{Best baseline} &   &  &  &  & \\
& node2vec & 80.89 $\pm$ 0.32 & 83.51 $\pm$ 0.29 & - & 1~328.82 & 1~328.82 (22 min) \\
\midrule

 & Standard GAE  & \multicolumn{2}{c|}{\textit{(intractable)}}  & \multicolumn{3}{c}{\textit{(intractable)}}\\
 & \underline{FastGAE (degree, $\alpha = 1$)}& &  &   &  & \\
\textbf{Google} & - with $n_{(S)} =$ 2~500 & 94.52 $\pm$ 0.26 & 95.50 $\pm$ 0.11 & 0.14 & 122.53 & \textbf{122.67} \\
 & - with $n_{(S)} = n^*_{(S)} =$ 7~911 & \textit{95.75} $\pm$ \textit{0.24} & \textit{96.62} $\pm$ \textit{0.09} & 0.14  & 158.63 & 158.77 \\
 & - with $n_{(S)} =$ 10~000 & \textbf{95.91} $\pm$ \textbf{0.19} & \textit{96.64} $\pm$ \textit{0.12} & 0.14 & 168.10 & 168.24  \\
& \underline{Best baseline} &   &   &  &  & \\
& node2vec  & 94.89 $\pm$ 0.63 & \textbf{96.82} $\pm$ \textbf{0.72} & - & 14~762.78 & 14~762.78 (4h06)  \\
\midrule

 & Standard VGAE  & \multicolumn{2}{c|}{\textit{(intractable)}}  & \multicolumn{3}{c}{\textit{(intractable)}} \\
& \underline{Var. FastGAE (degree, $\alpha = 5$)} & &  &  &  & \\
\textbf{Youtube} & - with $n_{(S)} =$ 3~000 & 81.14 $\pm$ 0.19 & 86.61 $\pm$ 0.16 & 0.28 & 453.22 & \textbf{453.50 (8min)}\\
 & - with $n_{(S)} = n^*_{(S)} =$ 15~179 & 81.83 $\pm$ 0.15 & \textit{87.21} $\pm$ \textit{0.15} & 0.28 & 2~964.51 & 2~964.79 (49min) \\
 & - with $n_{(S)} =$ 20~000 & \textbf{82.31} $\pm$ \textbf{0.18} & \textbf{87.36} $\pm$ \textbf{0.15} & 0.28 & 3~596.03 & 3~596.31 (1h00) \\
& \underline{Best baseline} & &  &  &  & \\
& Core-VGAE, $k=40$  & 80.53 $\pm$ 0.23 & 82.45 $\pm$ 0.20 & - & 12~433.51 & 12~433.51 (3h27) \\
\midrule

 & Standard GAE  & \multicolumn{2}{c|}{\textit{(intractable)}}  & \multicolumn{3}{c}{\textit{(intractable)}}\\
 & \underline{FastGAE with (degree, $\alpha = 2$)} &  &  &  & \\
\textbf{Patent} & - with $n_{(S)} =$ 5~000 & 90.66 $\pm$ 0.25 & 90.76 $\pm$ 0.22 & 0.30 & 605.75 & \textbf{606.05 (10min)}\\
 & - with $n_{(S)} = n^*_{(S)} =$ 16~425 &  \textbf{93.02} $\pm$ \textbf{0.23} & \textbf{93.39} $\pm$ \textbf{0.23} & 0.30 & 3~693.32 & 3~693.62 (1h02)\\
 & - with $n_{(S)} =$ 20~000 & \textit{92.91} $\pm$ \textit{0.22} & \textit{93.35} $\pm$ \textit{0.21} & 0.30 & 4~401.67 & 4~401.67 (1h13) \\
& \underline{Best baseline} &  &  &  & \\
& node2vec  & \textit{92.96} $\pm$ \textit{0.23} & \textit{93.36} $\pm$ \textit{0.20} & - & 25~851.39 & 25~851.39 (7h11) \\
\bottomrule
\end{tabular}
}
\end{table}

\begin{table}[t]
\centering
\caption[Community detection on all graphs with communities using FastGAE]{Community detection on all graphs with communities. For each graph, for brevity, we only report the \textbf{best} GAE \textbf{or} VGAE model in terms of AMI, a few representative degree-based FastGAE versions of this model, and the best baseline. Scores are averaged over 100 runs (except SBM: 10 runs). For degree sampling, values of the hyperparameter $\alpha$ (as defined in Equation~\eqref{eq:alphafastgae}) were tuned as illustrated in Figure~\ref{figurealphafastgae}. All GAE/VGAE models learn embedding vectors of dimension $d=16$. \textbf{Bold} numbers correspond to the best scores and best running times. Scores \textit{in italic} are within one standard deviation range from the best ones.}
\label{fastgaetable4}
 \resizebox{1.0\textwidth}{!}{
\begin{tabular}{c|l|c|ccc|c}
\toprule
\textbf{Dataset}  & \textbf{Model}& \textbf{Average Performance} & \multicolumn{3}{c}{\textbf{Average Running Times (in sec.)}} & \textbf{Speed}\\
&   & \footnotesize \textbf{AMI (in \%)} & \footnotesize Compute & \footnotesize Train & \footnotesize \textbf{Total} & \textbf{Gain} \\
& &  & \footnotesize $p_i$ & \footnotesize model &  &  \\ 
\midrule
\midrule

 & Standard GAE & 30.88 $\pm$ 2.56 & - & 3.90 & 3.90 & - \\
 & \underline{FastGAE (degree, $\alpha = 2$)} &  &  &  & & \\
\textbf{Cora}  & - with $n_{(S)} =$ 250 & 33.32 $\pm$ 2.61 & 0.002 & 1.51 & \textbf{1.512} & $\times$ \textbf{2.58}\\
 & - with $n_{(S)} = n^*_{(S)} =$ 440 & 34.64 $\pm$ 2.45 & 0.002 & 1.59 & 1.592 & $\times$ 2.45\\
 & - with $n_{(S)} =$ 1~000 & 35.56 $\pm$ 2.80 & 0.002 & 1.67 & 1.672 & $\times$ 2.33\\
& \underline{Best baseline} &  &  &  & & \\
& Louvain & \textbf{46.72} $\pm$ \textbf{0.85} & - & 1.79 & 1.79 & $\times$ 2.18 \\
\midrule

  & Standard VGAE & \textit{44.84} $\pm$ \textit{2.63} & - & 4.32 & 4.32 & - \\
 & \underline{Var. FastGAE (degree, $\alpha = 2$)} &  &  &  & & \\
\textbf{Cora} & - with $n_{(S)} =$ 250 & 41.35 $\pm$ 3.49 & 0.002 & 2.40 & 2.402 & $\times$ 1.80\\
\textbf{with} & - with $n_{(S)} = n^*_{(S)} =$ 440 & 42.89 $\pm$ 2.72 & 0.002 & 2.67 & 2.672 & $\times$ 1.62\\
\textbf{features}& - with $n_{(S)} =$ 1~000 & \textit{45.02} $\pm$ \textit{2.81} & 0.002 & 2.92 & 2.922 & $\times$ 1.48 \\
& \underline{Best baseline} &  &  &  & & \\
& Louvain & \textbf{46.72} $\pm$ \textbf{0.85} & - & 1.79 & \textbf{1.79} & $\times$ \textbf{2.41} \\
\midrule

 & Standard VGAE & 9.85 $\pm$ 1.24  & - & 5.44 & 5.44 & - \\
 & \underline{Var. FastGAE (degree, $\alpha = 1$)} &  &  &  & & \\
\textbf{Citeseer} & - with $n_{(S)} =$ 250 & 9.34 $\pm$ 1.48 & 0.002 & 1.77 & \textbf{1.772} & $\times$ \textbf{3.07}\\
 & - with $n_{(S)} = n^*_{(S)} =$ 488 & 10.02 $\pm$ 1.42 & 0.002 &  2.02 & 2.022 & $\times$ 2.69\\
 & - with $n_{(S)} =$ 1~000 & 10.16 $\pm$ 1.41 & 0.002 & 2.19 & 2.192 & $\times$ 2.48\\
& \underline{Best baseline} &  &  &  & & \\
& Louvain & \textbf{16.39} $\pm$ \textbf{1.45} & - & 2.41 & 2.41 & $\times$ 2.26 \\
\midrule

 & Standard VGAE & \textit{20.17} $\pm$ \textit{3.07} & - & 6.45 & 6.45 & - \\
 & \underline{Var. FastGAE (degree, $\alpha = 1$)} & &   &  &  & \\
\textbf{Citeseer} & - with $n_{(S)} =$ 250  & \textit{20.49} $\pm$ \textit{3.74} & 0.002 & 2.80 & 2.802 & $\times$ 2.30 \\
\textbf{with} & - with $n_{(S)} = n^*_{(S)} =$ 488  & \textit{20.53} $\pm$ \textit{3.45}  & 0.002 & 2.88 & 2.882 & $\times$ 2.24\\
\textbf{features}&  - with $n_{(S)} =$ 1~000 & \textbf{20.94} $\pm$ \textbf{3.21} & 0.002 & 3.11 & 3.112 & $\times$ 2.07 \\
 & \underline{Best baseline} &   &   &  &  & \\
 & Cora-Graph VAE, $k=2$  & 16.53 $\pm$ 1.95 & - & 2.76 & \textbf{2.76} & $\times$ \textbf{2.33} \\
\midrule

 & Standard VGAE & 20.52 $\pm$  2.97  & - & 856.05 & 856.05 & - \\
 & \underline{Var. FastGAE (degree, $\alpha = 1$)} &  &   &   &  &  \\
\textbf{Pubmed} & - with $n_{(S)} =$ 500 & 16.86 $\pm$ 4.84 & 0.01  & 3.17 & \textbf{3.18} & $\times$ \textbf{269.20}\\
 & - with $n_{(S)} = n^*_{(S)} =$ 1~187 &  18.84 $\pm$ 4.78 & 0.01  & 3.61  & 3.62 & $\times$ 236.49 \\
& - with $n_{(S)} =$ 5~000 & \textit{22.81} $\pm$ \textit{4.80} & 0.01 & 14.95 & 14.96 & $\times$ 57.22 \\
& \underline{Best baseline} &   &   &  &  & \\
& Core-VGAE, $k=2$  & \textbf{23.56} $\pm$ \textbf{3.12} & - & 50.11 & 50.11 & $\times$ 17.08 \\
\midrule

 & Standard VGAE  & 25.43 $\pm$ 1.47  & - & 970.67 & 970.67 & - \\
 & \underline{Var. FastGAE (degree, $\alpha = 1$)} &  &  &  & & \\
\textbf{Pubmed} & - with $n_{(S)} =$ 500 & 29.04 $\pm$ 4.17 & 0.01 & 4.03 & \textbf{4.04} & $\times$ \textbf{240.26}\\
\textbf{with} & - with $n_{(S)} = n^*_{(S)} =$ 1~187 & \textbf{31.11} $\pm$ \textbf{3.27} & 0.01 & 4.65 & 4.66 & $\times$ 208.30\\
\textbf{features}& - with $n_{(S)} =$ 5~000 & \textit{30.89} $\pm$  \textit{3.01} & 0.01 & 20.01 & 20.02 & $\times$ 48.49\\
 & \underline{Best baseline}&  &  &  & & \\
 & Core-VGAE, $k=2$ & 24.35 $\pm$ 1.55 & - & 57.09 & 57.09 & $\times$ 17.00 \\
 \midrule

 & Standard VGAE & \textit{(intractable)}  & \multicolumn{3}{c|}{\textit{(intractable)}} & - \\
& \underline{Var. FastGAE (degree, $\alpha = 2$)} &  &  &  & & \\
\textbf{SBM} & - with $n_{(S)} =$ 2~500 & 30.77 $\pm$ 0.32 & 0.03 & 52.01 & \textbf{52.04} & - \\
& - with $n_{(S)} = n^*_{(S)} =$ 2~673 & 30.89 $\pm$ 0.30 & 0.03 & 53.98 & 54.01 & - \\
& -with $n_{(S)} =$ 5~000 & 32.28 $\pm$ 0.26 & 0.03 & 61.96 & 61.69 & - \\
& \underline{Best baseline} &  &  &  & & \\
& Louvain & \textbf{35.90} $\pm$ \textbf{0.14} & - & 464.11 & 464.11 & - \\
\bottomrule
\end{tabular}
}
\end{table}

\section{Conclusion}
\label{c4s44}

In this chapter, we introduced FastGAE, a stochastic method to scale GAE and VGAE models. We publicly released our Python/TensorFlow implementation of this method along with the paper associated with this work~\cite{salha2021fastgae}. We demonstrated its effectiveness on several large graphs with up to millions of nodes and edges, both in terms of speed, scalability, and performance. We outperformed the few existing approaches to scale GAE and VGAE models, including our degeneracy framework from Chapter~\ref{chapter_3}. We also showed that FastGAE addresses some of the limitations of this framework. As a consequence, we consider that FastGAE constitutes an improvement of our previous efforts and, in the remainder of this thesis, we will rather resort to FastGAE when dealing with large graphs.

Besides its empirical superiority over the degeneracy framework from Chapter~\ref{chapter_3} in a majority of our experiments, FastGAE is also conceptually simpler. We believe that simple solutions have the most impact. At the time of writing, the FastGAE method has already been explicitly mentioned and used in other research experiments, e.g., in~\cite{huang2020edge,ray2020predicting,xiang2021general}. Xiang Sheng (\texttt{xiangsheng1325} on GitHub) also recently developed a PyTorch implementation of FastGAE\footnote{\href{https://github.com/xiangsheng1325/fastgae_pytorch}{https://github.com/xiangsheng1325/fastgae\_pytorch}}.

In addition, we emphasize that our method easily extends to GAE and VGAE models with alternative GNN encoders. In our experiments, the GCN encoders of standard GAE and VGAE models~\cite{kipf2016-2} and of FastGAE-based models could easily be replaced by any alternative architecture learning the embedding matrix $Z$ in another way, e.g., by a FastGCN~\cite{chen2018fastgcn}, a Cluster-GCN~\cite{chiang2019cluster} or a GraphSAGE~\cite{hamilton2017inductive} encoder. Besides, FastGAE easily extends to alternative \textit{decoders}. For instance, one could replace the symmetric inner product decoder from our experiments with some more elaborated decoders \cite{grover2019graphite,aaai20}, including the gravity-inspired decoder that we will ourselves propose in Chapter~\ref{chapter_5} to reconstruct \textit{directed} graphs.

Simultaneously, we identify several future research directions for improvements. Apart from the aforestated limit (see Section~\ref{c4s433}) of current GAE and VGAE models on community detection, which we will investigate in Chapter~\ref{chapter_7}, we underline that the proposed FastGAE method could underperform on very sparse graphs. Indeed, in such a scenario, the subgraphs to reconstruct might include a large proportion of isolated nodes, which would negatively impact the learning process. Moreover, in the case of large graphs with a lot of sparsely connected components, we recommend applying FastGAE separately on each component. Lastly, in this chapter, we assumed that the graph was fixed. This aspect might appear as a limit, that could motivate future interesting studies on extensions of FastGAE for scalable \textit{dynamic} graph embeddings, potentially with a dynamic selection of $n_{(S)}$.

\clearpage

\section{Appendices}
\label{c4s45}

This supplementary section provides the theoretical analyses announced in Section~\ref{c4s423} as well as an additional figure presenting the optimal values of the hyperparameter $\alpha$ in our experiments. They were placed out of the main content of Chapter~\ref{chapter_4} for the sake of brevity and readability.

\subsection*{On Approximate Losses}

Let us recall that, in our FastGAE framework, at each training iteration we run a full GCN forward pass and sample a subgraph $\mathcal{G}_{(S)} = (\mathcal{V}_{(S)},\mathcal{E}_{(S)})$. Then, we evaluate reconstruction losses only on this subgraph, which involves fewer operations w.r.t. standard decoders, and we use the resulting approximate loss for GCN weights updates via gradient descent. 
More precisely, in standard implementations of GAE/VGAE and assuming an unweighted graph, the cross entropy loss and the negative of the ELBO's expectation part are empirically derived by computing the following node pairs average at each training iteration:
\begin{equation}\mathcal{L} = \frac{1}{n^2}\sum_{(i,j) \in \mathcal{V}^2} \mathcal{L}_{ij}(A_{ij},\hat{A}_{ij}), 
\label{eq:lossfastgae} \end{equation}
with\footnote{In most implementations, as explained in Section~\ref{c2s241}, the terms with $A_{ij} = 1$ are re-weighted in the loss, in case of sparse $A$. They are multiplied by $w_{\text{pos}} \geq 1$, a positive links re-weighting scalar parameter which is usually inversely proportional to the graph sparsity. In our analyses, to clarify notations, we omit this scalar multiplication, which is equivalent to implicitly assuming that $w_{\text{pos}} =1$. This simplification is made without loss of generality and all results remain valid for any $w_{\text{pos}} >1$.}:
$\mathcal{L}_{ij}(A_{ij},\hat{A}_{ij}) = - [A_{ij}\log(\hat{A}_{ij}) + (1-A_{ij})\log(1 - \hat{A}_{ij})].$
In the FastGAE framework, we instead compute:
\begin{equation}\mathcal{L}_{\text{\tiny FastGAE}} = \frac{1}{n_{(S)}^2}\sum_{(i,j) \in \mathcal{V}^{2}} \mathds{1}_{((i,j) \in \mathcal{V}_{(S)}^2)}\mathcal{L}_{ij}(A_{ij},\hat{A}_{ij}),
\label{eq:newfastgaeloss} \end{equation}
with $\mathds{1}_{((i,j) \in \mathcal{V}_{(S)}^2)} = 1$ if $(i,j) \in \mathcal{V}_{(S)}^2$ and $0$ otherwise. We recall that, for variational FastGAE, we need to substract the Kullback-Leibler (KL) divergence~\cite{kullback1951information}, as in the ELBO of standard VGAE, to obtain our actual objective function. At this stage, two options are possible:
\begin{itemize}
    \item computing the KL term only on the nodes in the subgraph; 
    \item or, computing the KL term on all nodes.
\end{itemize}
We consider that the two options are valid. The first one ensures that the resulting loss is a proper lower bound of the likelihood computed on this subgraph. The second one, despite violating this property, can nonetheless be empirically convenient and interpreted as the addition of a \textit{regularization term} on all node embedding vectors (penalizing large deviations w.r.t. a $\mathcal{N}(0,I_d)$ prior distribution on these vectors) to the \textit{performance} term $\mathcal{L}_{\text{\tiny FastGAE}}$. In our experiments, both options returned similar results. In the following propositions, we assume that the KL term is computed on \textit{all} nodes for simplicity, and we therefore only approximate the performance term $\mathcal{L}$, both in the GAE and in the VGAE settings.

Propositions~\ref{fastgaeprop1} and \ref{fastgaeprop2} detail the formal probabilities to sample a given node or a given node pair at each training iteration. We consider both sampling variants \textit{with} and \textit{without} replacement (see Section 3.2) for this analysis, as the former significantly simplifies results w.r.t. the latter. 

\begin{theorem}
Let $\mathcal{G}_{(S)} = (\mathcal{V}_{(S)},\mathcal{E}_{(S)})$ be a subgraph of $\mathcal{G}$ obtained from sampling $n_{(S)}$ nodes \textbf{with} replacement using the node sampling strategy of FastGAE. Let $i$ and $j$ denote two distinct nodes from the original graph $\mathcal{G}$: $(i,j) \in \mathcal{V}^2$. Then:
\begin{equation}\mathbb{P}\Big(i \in \mathcal{V}_{(S)}\Big) = 1 - (1 - p_i)^{n_{(S)}}.\end{equation}
Also:
\begin{equation}
\mathbb{P}\Big((i,j) \in \mathcal{V}_{(S)}^2\Big) = 1 - \Big[(1 - p_i)^{n_{(S)}} + (1 - p_j)^{n_{(S)}} - (1 - p_i - p_j)^{n_{(S)}}\Big].
\end{equation}
\label{fastgaeprop1}
\end{theorem}

\begin{proof}
In this setting, sampling probabilities are independent of previous sampling steps, and remain fixed to $p_i$.  Therefore, for node $i \in \mathcal{V}$, we have:
$$\mathbb{P}\Big(i \notin \mathcal{V}_{(S)}\Big) = (1 - p_i)^{n_{(S)}}.$$
Indeed, for $i$ \textit{not} to belong to $\mathcal{V}_{(S)}$, it must not be selected at any of the $n_{(S)}$ draws, which happens with probability $1 - p_i$ for each draw. Therefore:
$$\mathbb{P}\Big(i \in \mathcal{V}_{(S)}\Big) = 1 - (1 - p_i)^{n_{(S)}}.$$

Moreover, let $i$ and $j$ denote two distinct nodes from the original graph $\mathcal{G}$: $(i,j) \in \mathcal{V}^2$. We have:
\begin{align*}
\mathbb{P}\Big((i,j) \notin \mathcal{V}_{(S)}^2\Big) &=
\mathbb{P}\Big(i \notin \mathcal{V}_{(S)} \text{ or } j \notin \mathcal{V}_{(S)}\Big) \\
&= \mathbb{P}\Big(i \notin \mathcal{V}_{(S)}\Big) + \mathbb{P}\Big(j \notin \mathcal{V}_{(S)}\Big) - \mathbb{P}\Big(i \notin \mathcal{V}_{(S)}, j \notin \mathcal{V}_{(S)}\Big)
\end{align*}
with, using the previous result, $\mathbb{P}(i \notin \mathcal{V}_{(S)}) = (1 - p_i)^{n_{(S)}}$ and $\mathbb{P}(j \notin \mathcal{V}_{(S)}) = (1 - p_j)^{n_{(S)}}$. Using a similar argument, we also obtain:
$$\mathbb{P}\Big(i \notin \mathcal{V}_{(S)}, j \notin \mathcal{V}_{(S)}\Big) = \Big(1 - (p_i + p_j)\Big)^{n_{(S)}}.$$
Therefore:
\begin{equation*}
\mathbb{P}\Big((i,j) \notin \mathcal{V}_{(S)}^2\Big) = \Big[(1 - p_i)^{n_{(S)}} + (1 - p_j)^{n_{(S)}} - (1 - p_i - p_j)^{n_{(S)}}\Big].
\end{equation*}
And:
\begin{equation*}
\mathbb{P}\Big((i,j) \in \mathcal{V}_{(S)}^2\Big) = 1 - \mathbb{P}\Big((i,j) \notin \mathcal{V}_{(S)}^2\Big) = 1 - \Big[(1 - p_i)^{n_{(S)}} + (1 - p_j)^{n_{(S)}} - (1 - p_i - p_j)^{n_{(S)}}\Big].
\end{equation*}
Lastly, for self-loops:
$$\mathbb{P}\Big((i,i) \in \mathcal{V}_{(S)}^2\Big) = \mathbb{P}\Big(i \in \mathcal{V}_{(S)}\Big) = 1 - (1 - p_i)^{n_{(S)}}.$$
\end{proof}

\begin{theorem}
Let $\mathcal{G}_{(S)} = (\mathcal{V}_{(S)},\mathcal{E}_{(S)})$ be a subgraph of $\mathcal{G}$ obtained from sampling $n_{(S)}$ nodes \textbf{without} replacement using the node sampling strategy of FastGAE. Let $i$ and $j$ denote two distinct nodes from $\mathcal{G}$: $(i,j) \in \mathcal{V}^2$. Then:
\begin{equation}\mathbb{P}\Big(i \in \mathcal{V}_{(S)}\Big) = \sum_{\mathcal{U} \in \mathcal{\textbf{U}}(i)} p_{u_1} \prod_{k=2}^{n_{(S)}} \frac{p_{u_k}}{1 - \sum_{k'=1}^{k-1} p_{u_{k'}}},\end{equation}
where $\mathcal{\textbf{U}}(i) = \{\mathcal{U} \subset \mathcal{V}, |\mathcal{U}| = n_{(S)} \text{ and } i \in \mathcal{U}\}$ is the set of all ordered subsets of $n_{(S)}$ distinct nodes including node $i$. For a given set $\mathcal{U} \in \mathcal{\textbf{U}}(i)$, we denote by $(u_1, u_2,...,u_{n_{(S)}})$ its ordered elements. Also,
\begin{equation}\mathbb{P}\Big((i,j) \in \mathcal{V}_{(S)}^2\Big) = 
\sum_{\mathcal{U} \in \mathcal{\textbf{U}}(i) \cap \textbf{U}(j)} p_{u_1} \prod_{k=2}^{n_{(S)}} \frac{p_{u_k}}{1 - \sum_{k'=1}^{k-1} p_{u_{k'}}}.\end{equation}
\label{fastgaeprop2}
\end{theorem}

\begin{proof}
We are looking for the probability that a node $i  \in \mathcal{V}$ from the graph belongs to a drawn subset $\mathcal{V}_{(S)}$, that contains $n_{(S)}$ distinct nodes. For $\mathcal{V}_{(S)}$ to include $i$, $\mathcal{V}_{(S)}$ should match any of the possible ordered subsets of $n_{(S)}$ nodes that include node $i$. In this setting where we sample without replacement, the probability to draw node $i$ \textit{depends on nodes previously drawn}. All possible orders of sampling the nodes should be considered. Let:
$$\mathcal{\textbf{U}}(i) = \Big\{\mathcal{U} \subset \mathcal{V}, |\mathcal{U}| = n_{(S)} \text{ and } i \in \mathcal{U}\Big\}$$
denote the set of all \textbf{ordered} subsets of $n_{(S)}$ distinct nodes that include node $i$. With such a notation, we have:
$$\mathbb{P}\Big(i \in \mathcal{V}_{(S)}\Big) = \mathbb{P}\Big(\mathcal{V}_{(S)} \in \mathcal{\textbf{U}}(i) \Big) = \sum_{\mathcal{U} \in \mathcal{\textbf{U}}(i)} \mathbb{P}\Big(\mathcal{V}_{(S)} = \mathcal{U}\Big).$$

The summation comes from the fact that events are \textit{disjoint} ($\mathcal{V}_{(S)}$ can not match two of these ordered subsets simultaneously).

Now, for a given set $\mathcal{U} \in \mathcal{\textbf{U}}(i)$, let us denote by $(u_1, u_2,...,u_{n_{(S)}})$ its \textbf{ordered} elements. Also, let $(\mathcal{V}_{(S)1}, \mathcal{V}_{(S) 2},...,\mathcal{V}_{(S) n_{(S)}})$ be the $n_{(S)}$ ordered nodes of set $\mathcal{G}_{(S)}$ (i.e., $\mathcal{V}_{(S) 1}$ is the first drawn node, $\mathcal{V}_{(S) 2}$ is the second one, etc). We have:

\begin{align*}
\mathbb{P}\Big(\mathcal{V}_{(S)} = \mathcal{U} \Big) &= \mathbb{P}\Big(\mathcal{V}_{(S) 1} = u_1, \mathcal{V}_{(S)2} = u_2,..., \mathcal{V}_{(S) n_{(S)}} = u_{n_{(S)}} \Big)\\
&= \mathbb{P}(\mathcal{V}_{(S) 1} = u_1) \prod_{k=2}^{n_{(S)}}  \mathbb{P}(\mathcal{V}_{(S) k} = u_k | \mathcal{V}_{(S) k-1} = u_{k-1},..., \mathcal{V}_{(S) 1} = u_1) \\
&= p_{u_1} \prod_{k=2}^{n_{(S)}} \frac{p_{u_k}}{1 - \sum_{k'=1}^{k-1} p_{u_{k'}}}.
\end{align*}
Therefore, by summing elements to come back to $\mathbb{P}(i \in \mathcal{V}_{(S)})$:
$$\mathbb{P}\Big(i \in \mathcal{V}_{(S)}\Big) = \sum_{\mathcal{U} \in \mathcal{\textbf{U}}(i)} p_{u_1} \prod_{k=2}^{n_{(S)}} \frac{p_{u_k}}{1 - \sum_{k'=1}^{k-1} p_{u_{k'}}}.$$

Moreover, let $i$ and $j$ denote two distinct nodes from the original graph $\mathcal{G}$: $(i,j) \in \mathcal{V}^2$. Using a similar notation and reasoning, we get:
\begin{equation*}
\mathbb{P}\Big((i,j) \in \mathcal{V}_{(S)}^2\Big) = \mathbb{P}\Big(i \in \mathcal{V}_{(S)},j \in \mathcal{V}_{(S)} \Big) = \sum_{\mathcal{U} \in \mathcal{\textbf{U}}(i) \cap \textbf{U}(j)} \mathbb{P}\Big(\mathcal{V}_{(S)} = \mathcal{U} \Big).
\end{equation*}
Therefore:
$$\mathbb{P}\Big((i,j) \in \mathcal{V}_{(S)}^2\Big) = 
\sum_{\mathcal{U} \in \mathcal{\textbf{U}}(i) \cap \textbf{U}(j)} p_{u_1} \prod_{k=2}^{n_{(S)}} \frac{p_{u_k}}{1 - \sum_{k'=1}^{k-1} p_{u_{k'}}}.$$
And, for self-loops, $\mathbb{P}((i,i) \in \mathcal{V}_{(S)}^2) = \mathbb{P}(i \in \mathcal{V}_{(S)})$.
\end{proof}

Despite different formulations, both variants share a similar behavior in practice on most real-world graphs. In this paper, as explained in Section~\ref{c4s422}, we sample nodes \textit{without replacement}. One can derive from the above expressions that the probability to draw a node $i$, or an edge incident to $i$, increases with $n_{(S)}$, with $p_i$ and with $f(i)$ for $\alpha > 0$. This also leads to the following formulation of the expected (re-weighted) loss that FastGAE stochastically optimizes.

\begin{theorem}
Using the expressions of Proposition~\ref{fastgaeprop1} (when sampling with replacement) or Proposition~\ref{fastgaeprop2} (when sampling without replacement):
\begin{equation}\mathbb{E}\Big[\mathcal{L}_{\text{\tiny FastGAE}}\Big] = \frac{1}{n_{(S)}^2}\sum_{(i,j) \in \mathcal{V}^2} \mathbb{P}\Big((i,j) \in \mathcal{V}_{(S)}^2\Big) \mathcal{L}_{ij}(A_{ij},\hat{A}_{ij}).
\label{eq:expectfastgae}\end{equation}
\label{fastgaeprop3}
\end{theorem}

\begin{proof}
We have:
\begin{align*}
\mathbb{E}\Big[\mathcal{L}_{\text{\tiny FastGAE}}\Big] &= \mathbb{E}\Big[\frac{1}{n_{(S)}^2}\sum_{(i,j) \in \mathcal{V}^{2}} \mathds{1}_{((i,j) \in \mathcal{V}_{(S)}^2)}\mathcal{L}_{ij}(A_{ij},\hat{A}_{ij})\Big] \\
&= \frac{1}{n_{(S)}^2}\sum_{(i,j) \in \mathcal{V}^{2}} \mathbb{E}\Big[\mathds{1}_{((i,j) \in \mathcal{V}_{(S)}^2)}\Big]\mathcal{L}_{ij}(A_{ij},\hat{A}_{ij}) \\
&= \frac{1}{n_{(S)}^2}\sum_{(i,j) \in \mathcal{V}^2} \mathbb{P}\Big((i,j) \in \mathcal{V}_{(S)}^2\Big) \mathcal{L}_{ij}(A_{ij},\hat{A}_{ij}).
\end{align*}
By replacing $\mathbb{P}((i,j) \in \mathcal{V}_{(S)}^2)$ by the expressions of Proposition~\ref{fastgaeprop1} (with replacement) or Proposition~\ref{fastgaeprop2} (without replacement), we obtain an explicit formulation for $\mathbb{E}\Big[\mathcal{L}_{\text{\tiny FastGAE}}\Big]$.
\end{proof}

\subsection*{On the Selection of $n_{(S)}$}

While our experiments will tend to show that stochastically minimizing $\mathbb{E}(\mathcal{L}_{\text{\tiny FastGAE}})$ (Equation~\eqref{eq:expectfastgae}) instead of $\mathcal{L}$ (Equation~\eqref{eq:lossfastgae}) might be beneficial, we also acknowledge that, for small values of $n_{(S)}$, the actual loss $\mathcal{L}_{\text{\tiny FastGAE}}$ computed at a given training iteration (Equation~\eqref{eq:newfastgaeloss}) might significantly deviate from its expectation.

We propose to use these deviations as a criterion to automatically select a relevant subgraph size. More precisely, let us rewrite $\mathcal{L}_{\text{\tiny FastGAE}}$ from Equation~\eqref{eq:newfastgaeloss} as follows:
\begin{equation}
    \mathcal{L}_{\text{\tiny FastGAE}} = \frac{1}{n_{(S)}} \sum_{i \in \mathcal{V}} \mathds{1}_{(i \in \mathcal{V}_{(S)})} \mathcal{L}_{\text{\tiny FastGAE}}(i),\end{equation}
where the node-level terms $\mathcal{L}_{\text{\tiny FastGAE}}(i)$ are defined as:
\begin{equation}\mathcal{L}_{\text{\tiny FastGAE}}(i) =  \frac{1}{n_{(S)}} \sum_{j \in \mathcal{V}} \mathds{1}_{(j \in \mathcal{V}_{(S)})} \mathcal{L}_{ij}(A_{ij},\hat{A}_{ij}),
\label{eq12}\end{equation}
and where $\mathcal{L}_{ij}$ denotes the cross entropy loss as in Equation~\eqref{eq:lossfastgae}.
In the following, we leverage concentration inequalities \cite{hoeffding1963probability} to derive a theoretically-grounded threshold size, denoted  $n^*_{(S)}$ in the following, for which, under mild assumptions, the (random) node-level deviation
$|\mathcal{L}_{\text{\tiny FastGAE}}(i) - \mathbb{E}[\mathcal{L}_{\text{\tiny FastGAE}}(i)]|$
at each training iteration is proven to be bounded with a high probability, for any node $i$. This proposed subgraph size is of the form:
\begin{equation}n^*_{(S)} = C \sqrt{n},\end{equation}
where the constant $C >0$ depends on the deviation magnitude and probability, and is explicitly presented in Proposition~\ref{fastgaeprop5}. In our empirical analysis, this criterion will allow us to significantly improve the scalability and training speed of GAE and VGAE models (see discussion on complexity in Section~\ref{c4s424}), while reaching fairly competitive performances in a majority of experiments (see Section~\ref{c4s43}).
To prove our bounds, we require a technical assumption on $\hat{A}$:

\begin{assumption}
Let $(i,j) \in \mathcal{V}^2$. We thereafter assume that $\hat{A}_{ij} = \sigma(z^T_i z_j)$ can actually be \textit{capped}, and that $\hat{A}_{ij} \in [\varepsilon,1-\varepsilon]$, where $0 < \varepsilon < 1$ is a constant that can be arbitrarily~close~to~0.
\label{assumption}
\end{assumption}
Under this assumption, we derive Propositions~\ref{fastgaeprop4} and \ref{fastgaeprop5}.
\begin{theorem}
Let us consider a training iteration of the FastGAE framework, a sampled subgraph $\mathcal{G}_{(S)} = (\mathcal{V}_{(S)},\mathcal{E}_{(S)})$, with $|\mathcal{V}_{(S)}| = n_{(S)} < n$ nodes sampled without replacement, and the corresponding node-level approximate reconstruction computed for a given node $i$: $\mathcal{L}_{\text{\tiny FastGAE}}(i)$ from Equation~\eqref{eq12}, with $\mathcal{L}_{ij}(A_{ij},\hat{A}_{ij}) = - [A_{ij}\log(\hat{A}_{ij}) + (1-A_{ij})\log(1 - \hat{A}_{ij})]$ in this same equation. Then, under Assumption~\ref{assumption}, for any $\gamma \geq 0$, we have:
\begin{equation}
\mathbb{P}\Big(|\mathcal{L}_{\text{\tiny FastGAE}}(i) - \mathbb{E}[\mathcal{L}_{\text{\tiny FastGAE}}(i)]| \geq \gamma \Big) \leq 2~\text{exp}\Big( - 2(\frac{\gamma}{\log(\varepsilon)})^2 \frac{n_{(S)}^2}{n}\Big).
\end{equation}
\label{fastgaeprop4}
\end{theorem}

\begin{proof}
As a preliminary, let us recall Hoeffding's inequality \cite{hoeffding1963probability}. Let $X_1, X_2..., X_n$ be real independent random variables verifying, for some $(a_k)_{1 \leq k \leq n}$ and $(b_k)_{1 \leq k \leq n}$ with $a_k < b_k$: $\forall k, \mathbb{P}(a_k \leq X_k \leq b_k) = 1.$ Let $S_n = \sum_{i=1}^n X_i.$ Then, for all $\gamma>0$, we have:
$$\mathbb{P}\Big(|S_n - \mathbb{E}(S_n)| \geq t\Big) \leq 2~\text{exp}\Big( - \frac{2\gamma^2}{\sum_{i=1}^n (b_i - a_i)^2} \Big).$$
Hoeffding~\cite{hoeffding1963probability} also proves that the inequality holds when the $X_i$ are samples without replacement from a finite population (and therefore not independent). In the setting of Proposition~\ref{fastgaeprop4}, that falls into this second case, we have $\mathcal{L}_{\text{\tiny FastGAE}}(i) = \sum_{j \in \mathcal{V}} X_{ij}$, where, under Assumption~\ref{assumption}:
\begin{small}
\begin{equation*}
X_{ij} = \frac{1}{n_{(S)}} \mathds{1}_{(j \in \mathcal{V}_{(S)})}\mathcal{L}_{ij}(A_{ij},\hat{A}_{ij}) =  \underbrace{\mathds{1}_{(j \in \mathcal{V}_{(S)})}}_{\in \{0,1\}} \underbrace{\frac{-1}{n_{(S)}} \underbrace{[A_{ij}\log(\hat{A}_{ij}) + (1-A_{ij})\log(1 - \hat{A}_{ij})]}_{\in [\log(\varepsilon),\log(1-\varepsilon)]}}_{\in [-\log(1-\varepsilon)/n_{(S)},-\log(\varepsilon)/n_{(S)}]} \in \Big[0, \frac{-\log (\varepsilon)}{n_{(S)}}\Big].
\end{equation*}
\end{small}
We note that $\frac{-\log (\varepsilon)}{n_{(S)}} >0$, as $0 < \varepsilon < 1$. Applying \cite{hoeffding1963probability}, at each sampling step and for all $\gamma >0$:

\begin{align*}
    \mathbb{P}\Big(|\mathcal{L}_{\text{\tiny FastGAE}}(i) - \mathbb{E}[\mathcal{L}_{\text{\tiny FastGAE}}(i)]| \geq \gamma\Big) &\leq 2~\text{exp}\Big( - \frac{2\gamma^2}{\sum_{j \in \mathcal{V}} (\frac{-\log (\varepsilon)}{n_{(S)}})^2} \Big) \\
    &= 2~\text{exp}\Big( \frac{- 2\gamma^2}{n \frac{(-\log (\varepsilon))^2}{n_{(S)}^2}}\Big) 
    = 2~\text{exp}\Big( - 2(\frac{\gamma}{\log(\varepsilon)})^2 \frac{n_{(S)}^2}{n}\Big).
\end{align*}
\end{proof}

We note that it exhibits the link between the loss deviation  and the $\frac{n_{(S)}^2}{n}$ ratio. Also, the right-hand side term tends to 0 exponentially fast w.r.t. $\gamma$ and $n_{(S)}$.

\begin{theorem}
For any confidence level $\alpha \in ]0,1[$ and node $i \in \mathcal{V}$, selecting a subgraph size $n_{(S)}$ such that:
\begin{equation}
n_{(S)} \geq n^*_{(S)} = \sqrt{n} \underbrace{\sqrt{\frac{-\log (\frac{\alpha}{2})\log(\varepsilon)^2}{ 2\gamma^2}}}_{\text{denoted $C$}},
\label{eq:Cexplicit}
\end{equation}
guarantees that:
\begin{equation}
\mathbb{P}\Big(|\mathcal{L}_{\text{\tiny FastGAE}}(i) - \mathbb{E}[\mathcal{L}_{\text{\tiny FastGAE}}(i)]| \geq \gamma\Big) \leq \alpha.\end{equation}
\label{fastgaeprop5}
\end{theorem}

\begin{proof}
This is a corollary of Proposition~\ref{fastgaeprop4}, from which we derive that, for any $\alpha \in ]0,1[$:
\begin{equation*}
2~\text{exp}\Big( - 2(\frac{\gamma}{\log(\varepsilon)})^2 \frac{n_{(S)}^2}{n}\Big) \leq \alpha 
\Rightarrow \mathbb{P}(|\mathcal{L}_{\text{\tiny FastGAE}} - \mathbb{E}[\mathcal{L}_{\text{\tiny FastGAE}}]| \geq \gamma) \leq \alpha.
\end{equation*}
Then:
\begin{equation*}
2~\text{exp}\Big( - 2(\frac{\gamma}{\log(\varepsilon)})^2 \frac{n_{(S)}^2}{n}\Big) \leq \alpha 
\Leftrightarrow~ n_{(S)} \geq \sqrt{n} \sqrt{\frac{-\log (\frac{\alpha}{2})\log(\varepsilon)^2}{ 2\gamma^2}}.
\end{equation*}
\end{proof}
As an opening, we note that, while the current bounds are empirically effective (see Section~\ref{c4s43}), future research will aim to directly bound the deviation of $\mathcal{L}_{\text{\tiny FastGAE}}$ instead of the node-level terms $\mathcal{L}_{\text{\tiny FastGAE}}(i)$, which would be more ambitious and challenging due to the inherent dependencies among sampled node pairs in FastGAE. Also, while Propositions~\ref{fastgaeprop4}~and~\ref{fastgaeprop5} focus on the case of the \textit{cross entropy loss} for consistency w.r.t. the content presented in this chapter, a similar analysis (omitted here) could be performed to obtain comparable bounds for other \textit{bounded} reconstruction losses. For instance, in the case of the Frobenius loss, where $\mathcal{L}_{ij}(A_{ij},\hat{A}_{ij}) = (A_{ij} - \hat{A}_{ij})^2$, and without Assumption~\ref{assumption}, one can obtain similar concentration guarantees as Proposition 5, with $C$ being replaced by the constant $C' = \sqrt{\frac{-\log(\alpha/2)}{2 \gamma^2}}$.

\paragraph{Numerical Application}

In our experiments, all $n^*_{(S)}$ thresholds are computed by evaluating Equation~\eqref{eq:Cexplicit}, setting $\gamma = 1$, $\alpha = 0.1$ and $\varepsilon = 0.001$.

\subsection*{On the hyperparameter $\alpha$}

\begin{figure}[h]
\centering
  \subfigure[Cora - Degree Sampling]{
  \scalebox{0.35}{\includegraphics{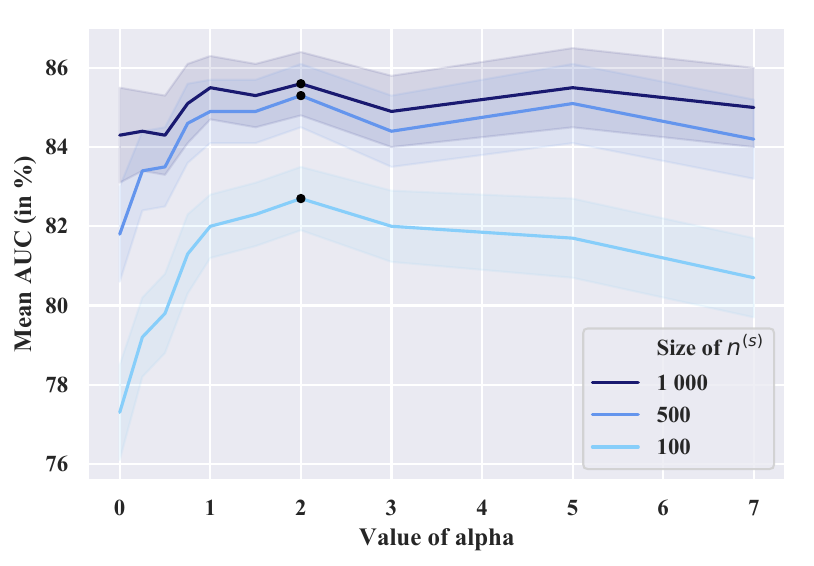}}}\subfigure[Citeseer - Degree Sampling]{
  \scalebox{0.35}{\includegraphics{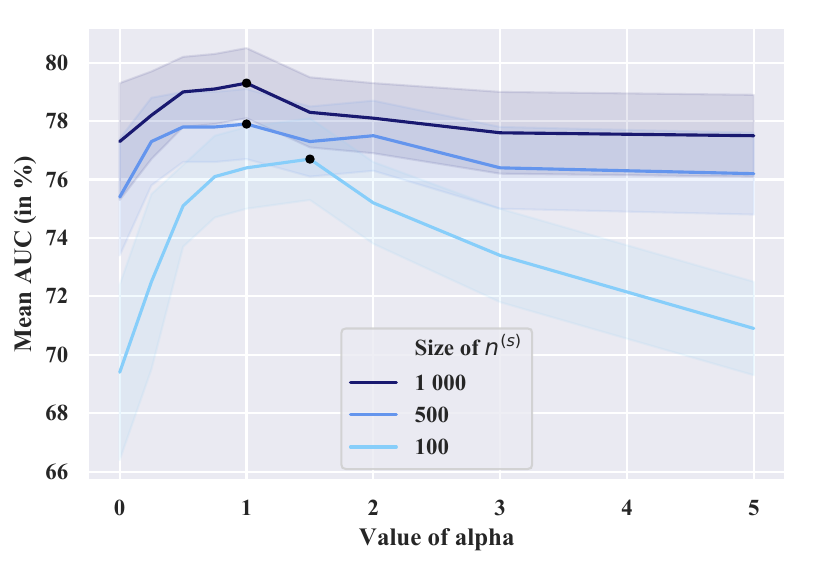}}}\subfigure[Pubmed - Degree Sampling]{
  \scalebox{0.35}{\includegraphics{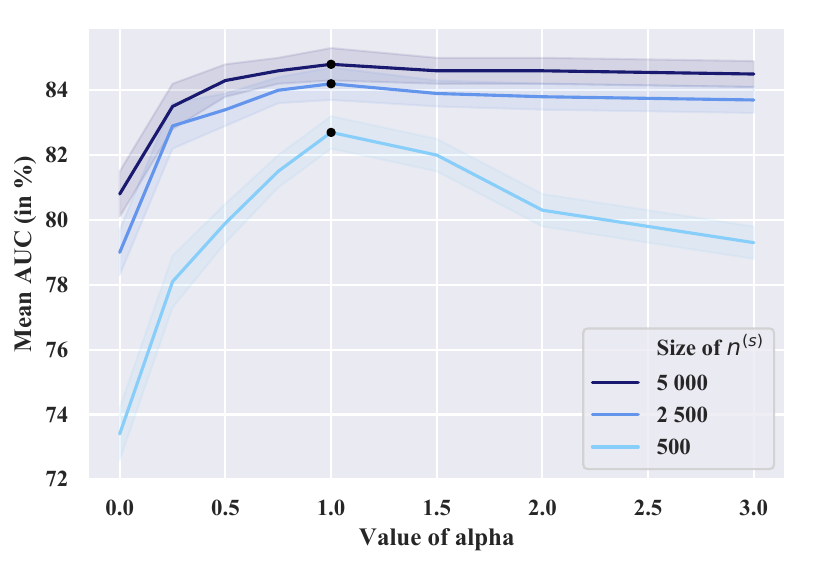}}}
   \subfigure[Cora - Core Sampling]{
  \scalebox{0.35}{\includegraphics{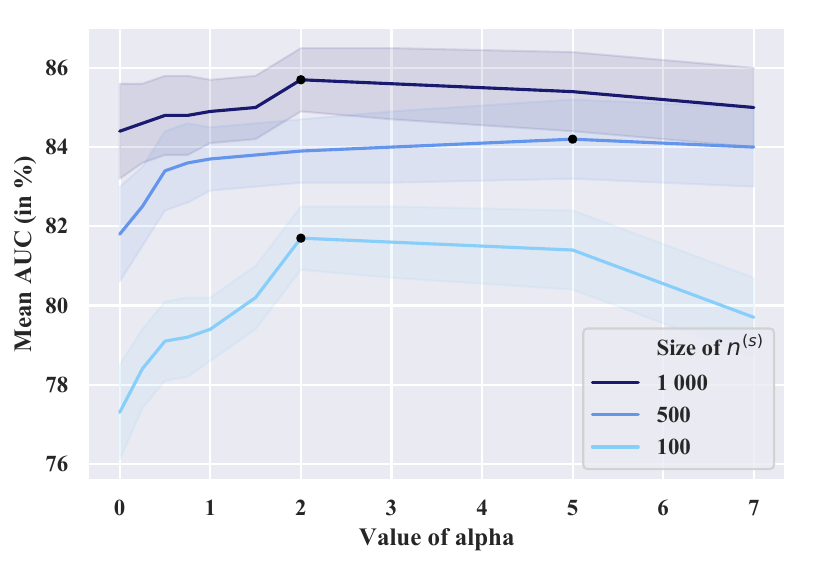}}}\subfigure[Citeseer - Core Sampling]{
  \scalebox{0.35}{\includegraphics{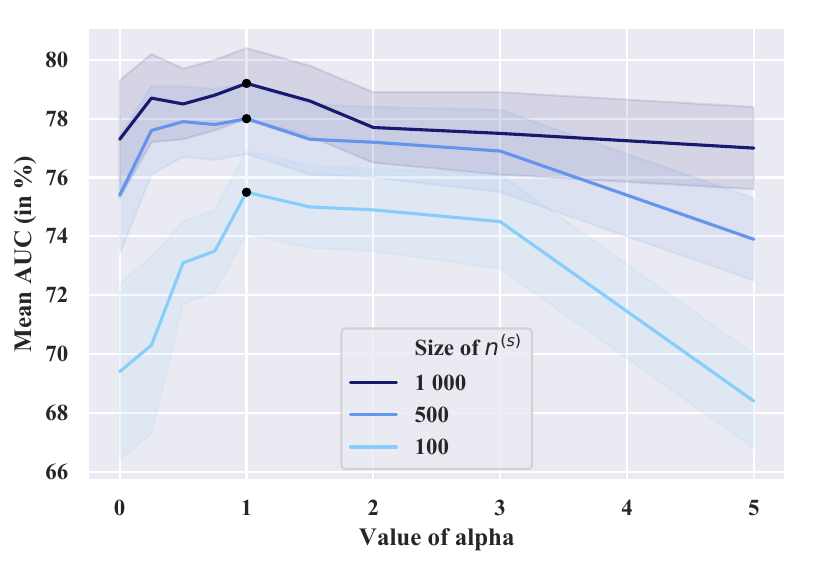}}}\subfigure[Pubmed - Core Sampling]{
  \scalebox{0.35}{\includegraphics{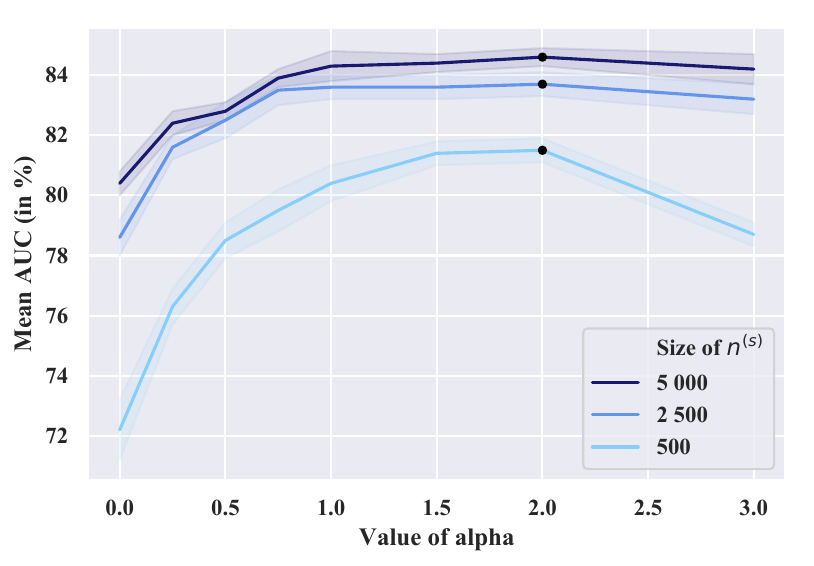}}}
     \subfigure[SBM - Degree Sampling]{
  \scalebox{0.35}{\includegraphics{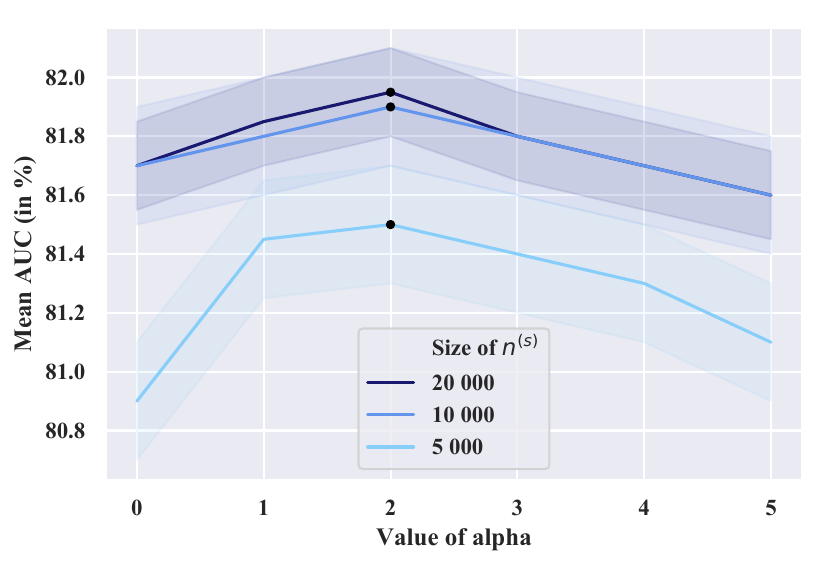}}}\subfigure[Google - Degree Sampling]{
  \scalebox{0.35}{\includegraphics{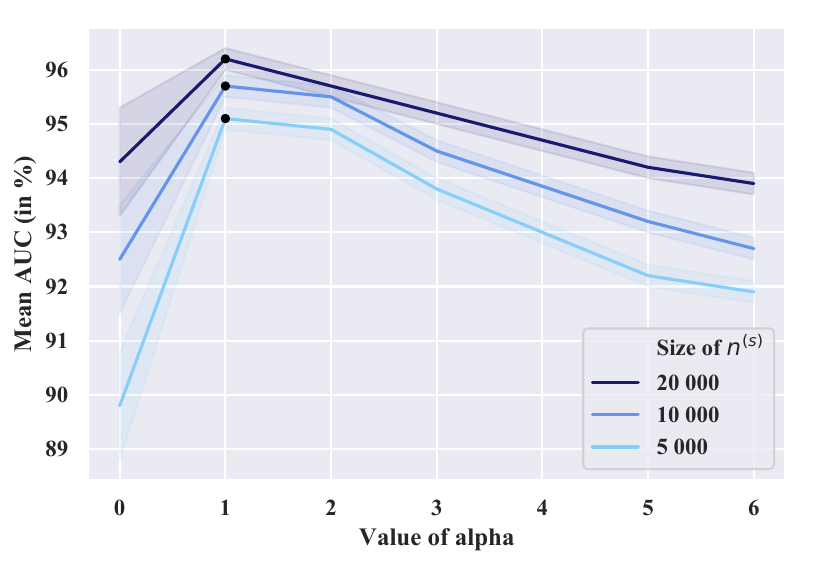}}}\subfigure[Youtube - Degree Sampling]{
  \scalebox{0.35}{\includegraphics{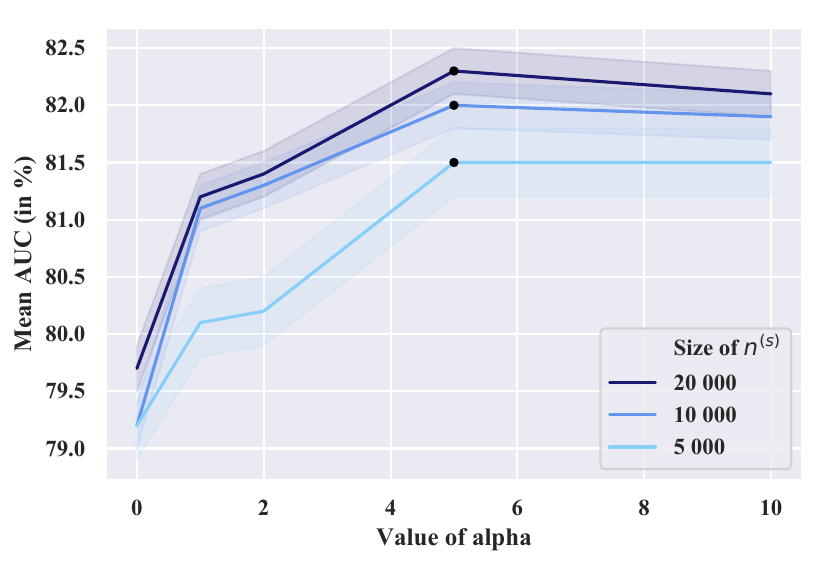}}}
     \subfigure[SBM - Core Sampling]{
  \scalebox{0.35}{\includegraphics{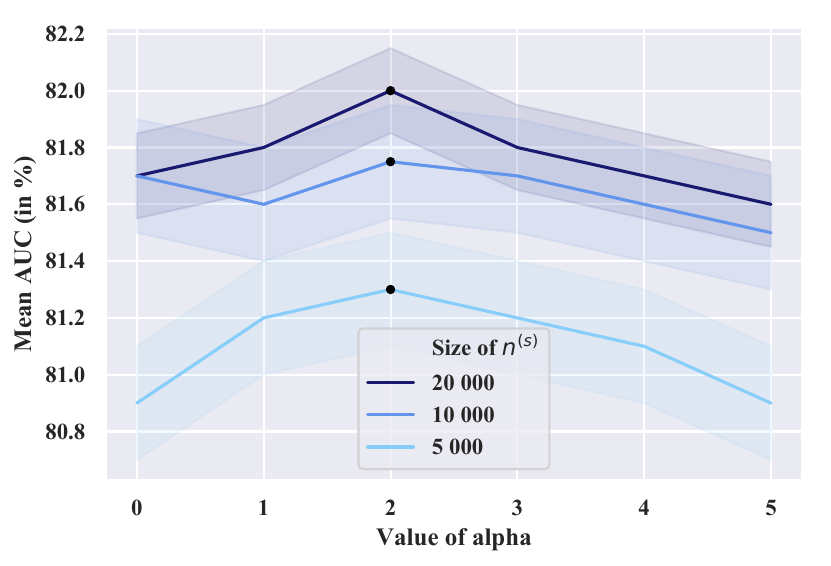}}}\subfigure[Google - Core Sampling]{
  \scalebox{0.35}{\includegraphics{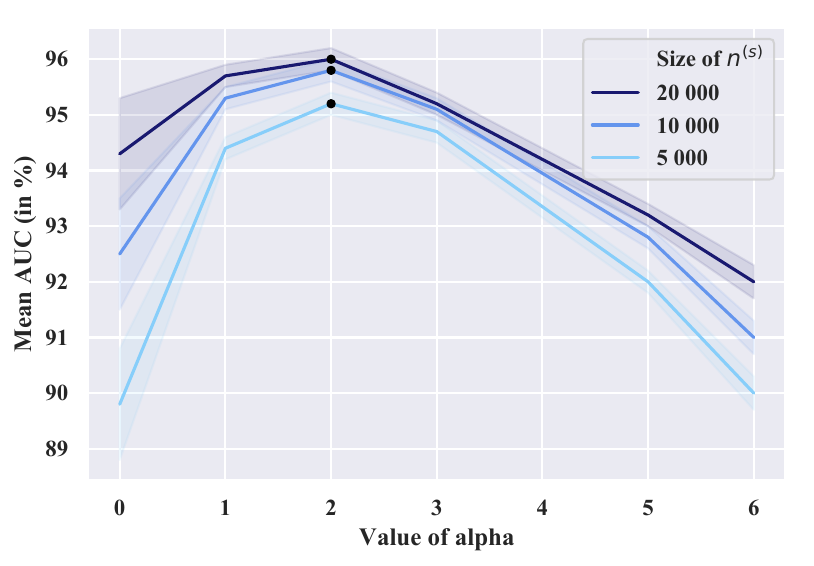}}}\subfigure[Youtube - Core Sampling]{
  \scalebox{0.35}{\includegraphics{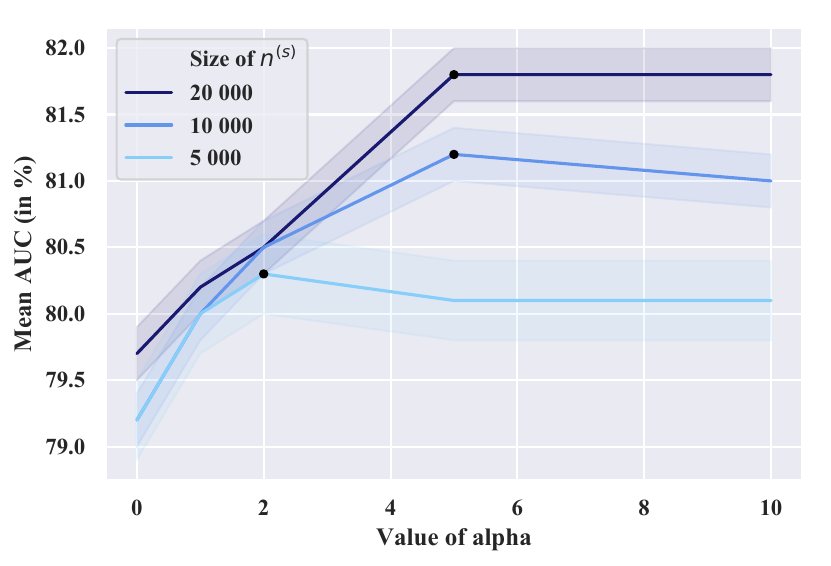}}}
    \subfigure[Patent - Degree Sampling]{
  \scalebox{0.35}{\includegraphics{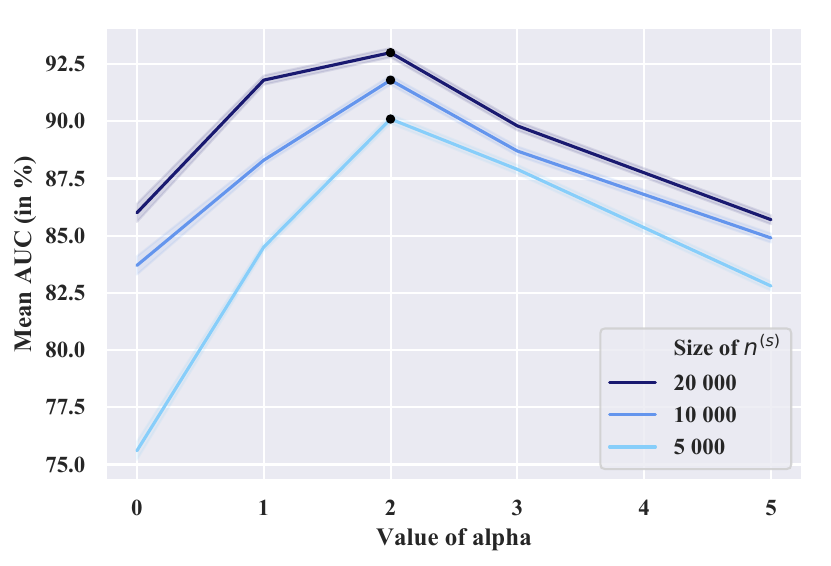}}}\subfigure[Patent - Core Sampling]{
  \scalebox{0.35}{\includegraphics{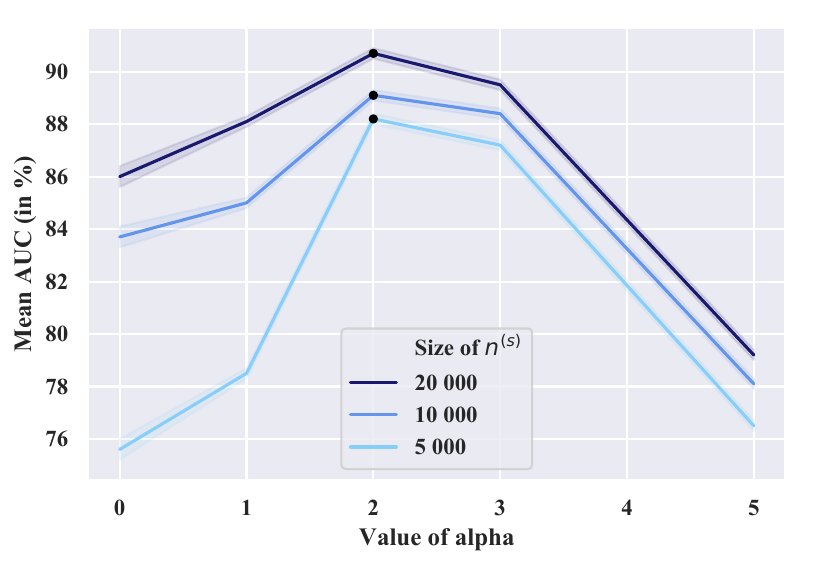}}}
  \caption[Optimal values of $\alpha$ for degree-based and core-based node sampling]{Optimal values of the hyperparameter $\alpha$ for degree-based and core-based node sampling w.r.t. mean AUC scores on validation sets, for Variational FastGAE models and for all graphs.}
  \label{figurealphafastgae}
\end{figure}

To finish, we report the additional Figure~\ref{figurealphafastgae} presenting the optimal values of the hyperparameter $\alpha$, for all graphs, and for both core-based and degree-based sampling.

\chapter[Gravity-Inspired Graph Autoencoders for Directed Link Prediction]{Gravity-Inspired Graph Autoencoders for~Directed~Link~Prediction}\label{chapter_5}
\chaptermark{Gravity-Inspired Graph Autoencoders for Directed Link Prediction}

\textit{This chapter presents research conducted with Stratis Limnios, Romain Hennequin, Viet-Anh Tran, and Michalis Vazirgiannis, and published in the proceedings of the 28\up{th} ACM International Conference on Information and Knowledge Management (CIKM 2019)~\cite{salha2019-2}.}

\section{Introduction}
\label{c5s51}

While GAEs and VGAEs emerged as powerful node embedding models, their original versions were designed for \textit{undirected} graphs~\cite{kipf2016-2} and ignore the potential direction of edges during the decoding step. As we will explain in Section~\ref{c5s52}, if a standard GAE/VGAE model predicts that a node $i \in \mathcal{V}$ is connected to a node $j \in \mathcal{V}$, then the same model will also predict that $j$ is connected to $i$ with the same probability. As a consequence, in our own experiments from Chapters~\ref{chapter_3}~and~\ref{chapter_4}, we also only analyzed undirected versions of the~graphs~under~consideration.

This is limiting for numerous real-world applications, as \textit{directed} graphs are ubiquitous. For instance, web graphs are made up of directed hyperlinks. In social networks such as Twitter, opinion leaders are usually followed by many users, but only few of these connections are reciprocal. Moreover, directed graphs are efficient abstractions in many domains where data are not explicitly structured as graphs. For instance, in Chapter~\ref{chapter_8}, we will study top-$k$ similar artists graphs, that artificially connect Deezer artists to their top-$k$ most similar ones according to usage-based similarity metrics, e.g., the proportion of the artist's fans that also listened to these other artists. By nature, such a graph is directed. While most fans of a little known reggae band might listen to Bob Marley (Marley thus appearing among their top-$k$ similar artists according to the aforestated criterium), Bob Marley’s fans will rarely listen to this band, which is unlikely to appear back among Bob Marley’s own top-$k$ most similar artists.

In this chapter, we aim to extend GAEs and VGAEs to directed graphs, with a particular focus on applications to \textit{directed link prediction} tasks. Predicting the location of directed links has been historically performed by leveraging graph mining-based asymmetric measures \cite{garcia2015phd,schall2015link,yu2014link} and, recently, a few attempts at capturing asymmetric proximity when learning node embedding spaces were proposed \cite{miller2009nonparametric,ou2016asymmetric,zhou2017scalable}. However, the question of how to reconstruct directed graphs from vector space representations to effectively perform directed link prediction remains widely open. In particular, at the time of this work, i.e., in 2019~\cite{salha2019-2}, it was unclear how to extend GAE and VGAE models to directed graphs, and to which extent the promising performances of these models on undirected graphs could also be achieved on directed link prediction tasks. 

In this chapter, we address these research questions. We present a new method, referred to as \textit{Gravity-Inspired GAE or VGAE}, to effectively learn node embedding representations from directed graphs using the GAE and VGAE frameworks. We draw inspiration from Newton's theory of universal gravitation~\cite{newton1687philosophiae} to introduce a new decoding scheme, able to reconstruct asymmetric relations from embedding vectors. We empirically evaluate our method on three different directed link prediction tasks, for which standard GAE and VGAE models perform poorly. We achieve competitive results on three real-world datasets, outperforming popular baselines. To the best of our knowledge, our work provided the first GAE/VGAE experiments on directed graphs.  Our implementation of Gravity-Inspired GAE and VGAE is publicly available~\cite{salha2019-2}.

This chapter is organized as follows. In Section~\ref{c5s52}, we explain why standard GAEs and VGAEs are not suitable for directed link prediction. In Section~\ref{c5s53}, we introduce our proposed Gravity-Inspired GAE and VGAE models. We present and discuss our experimental analysis in Section~\ref{c5s54}, and we conclude in Section~\ref{c5s55}.

\section{Extending Graph Autoencoders to Directed Graphs}
\label{c5s52}

In this section, we present the limitations of standard GAE and VGAE models in the presence of directed graphs. We also mention the ``source-target'' paradigm, a strategy previously used in the scientific literature to reconstruct asymmetric links from an embedding space~\cite{ou2016asymmetric,zhou2017scalable}.

\subsection{On the Limitations of Standard Decoders for Directed Graphs}
\label{c5s521}

The GAE and VGAE models presented in Section~\ref{c2s241} and~\ref{c2s242} from Chapter~\ref{chapter_2} respectively, as well as the extensions of these models cited in Section~\ref{c2s243} from this same chapter, all assume that the input graph $\mathcal{G}$ is \textit{undirected}. By design, these GAEs and VGAEs are \textit{not} suitable for directed graphs, as they are ignoring directions when reconstructing the adjacency matrix from an embedding space. In particular, due to the symmetry of the inner product decoder, we have:
\begin{equation}
\hat{A}_{ij} = \sigma(z_i^Tz_j) = \sigma(z_j^Tz_i) = \hat{A}_{ji}.
\end{equation}
In other words, if the inner product decoder predicts the existence of an edge $(i,j)$ from a node $i \in \mathcal{V}$ to a node $j \in \mathcal{V}$, then it also necessarily predicts the existence of the reverse edge $(j,i)$, \textit{with the same probability}. This is undesirable for directed graphs, where $A_{ij} \neq A_{ji}$ in general. Replacing inner product decoders by any $L_p$ distance in the embedding (e.g., the Euclidean distance, if $p=2$) or by existing more refined decoders \cite{grover2019graphite,aaai20} would lead to the same problem, as they are symmetric functions as well.

Therefore, as we will empirically show in Section~\ref{c5s54}, standard GAE and VGAE models significantly underperform on link prediction tasks in directed graphs. In 2019, Zhang et al.~\cite{zhang2019d} proposed D-VAE, a variational autoencoder for small Directed Acyclic Graphs (DAG) such as neural networks architectures or Bayesian networks, focusing on neural architecture search and structure learning. However, at the time of this work, the question of how to extend GAE and VGAE to general directed graphs, such as citation graphs or web graphs,~remained~open.

\subsection{The Source-Target Paradigm}
\label{c5s522}

Out of the GAE and VGAE frameworks, a few recent studies did tackle directed link prediction tasks using node embedding methods~\cite{ou2016asymmetric,zhou2017scalable}. These studies actually proposed to learn, not one, but \textit{two embedding vectors} for each node $i \in \mathcal{V}$ of the graph: a \textit{source} vector $z^{(s)}_i \in \mathbb{R}^d$ and a target vector $z^{(t)}_i \in \mathbb{R}^d$. More precisely: 
\begin{itemize}
    \item HOPE, short for \textit{High-Order Proximity preserved Embedding} \cite{ou2016asymmetric}, aims to preserve high-order node-level proximity and to capture asymmetric transitivity. Nodes are represented by source vectors, stacked up in an $n \times d$ matrix $Z^{(s)}$, and by target vectors stacked up in another $n \times d$ matrix $Z^{(t)}$. For a given $n \times n$ node-level similarity matrix $S$, Ou~et~al.~\cite{ou2016asymmetric} learn these vectors by approximately minimizing:
    \begin{equation}
\|S - Z^{(s)} Z^{(t) T} \|_F
    \end{equation} using a generalized SVD (we refer to Section~\ref{c2s222} from Chapter~\ref{chapter_2} for an introduction to node embeddings from matrix factorization). For directed graphs, an usual choice for $S$ is the Katz matrix \cite{katz1953new}:
    \begin{equation}
       S^{\text{Katz}} = \sum_{i=1}^{\infty} \beta^ i A^i,
    \end{equation}with $S^{\text{Katz}} = (I - \beta A)^{-1} \beta A$ if the parameter $\beta > 0$ is smaller than the spectral radius of $A$ \cite{katz1953new}. It computes the number of paths from a node to another one, these paths being exponentially weighted according to their length. Then, for directed link prediction, one can assess the likelihood of a missing edge from node $i$ to node $j$ using the asymmetric reconstruction:
    \begin{equation}
        \hat{A}_{ij} = \sigma(z^{(s)T}_i z^{(t)}_j),
    \end{equation}
    with verifies $\hat{A}_{ij} \neq \hat{A}_{ji} = \sigma(z^{(s)T}_j z^{(t)}_i)$ in general;
    \item APP, for \textit{Asymmetric Proximity Preserving}, is a related node embedding method~\cite{zhou2017scalable}, that instead aims to conserve the Rooted PageRank score~\cite{page1999pagerank} for any node pair. APP leverages \textit{random walk} with restart strategies to learn, as HOPE, a source vector and a target vector for each node (we refer to Section~\ref{c2s223} from Chapter~\ref{chapter_2} for an introduction to node embeddings from random walks). As above, one can predict whether node $i$ is connected to node $j$ through a directed edge by computing the inner product of the source vector of $i$ with the target vector of $j$, with a sigmoid activation.
\end{itemize}

We can derive a straightforward extension of this \textit{source/target vectors} paradigm for GAE and VGAE models. Indeed, considering GCN encoders returning $d$-dimensional embedding vectors $z_i$, with $d$ being even, we can assume that the $d/2$ first dimensions (respectively the $d/2$ last dimensions) of $z_i$ actually correspond to the source (resp. to the target) vector of node $i$, i.e.:
\begin{equation}
z^{(s)}_i = z_{i} [1:\frac{d}{2}]~\text{ and }~z^{(t)}_i = z_{i} [(\frac{d}{2}+1):d].
\end{equation} Then, we can replace the symmetric decoder $\hat{A}_{ij} = \hat{A}_{ji} = \sigma(z^T_i z_j)$ by $\hat{A}_{ij} = \sigma(z^{(s)T}_i z^{(t)}_j)$ and $\hat{A}_{ji} = \sigma(z^{(s)T}_j z^{(t)}_i)$ as in APP and HOPE,  to reconstruct directed links from GAE/VGAE-based encoded representations. In Section~\ref{c5s54}, we will provide an experimental evaluation of such an approach, which we will refer to as \textit{Source/Target GAE (or VGAE)}.

Nonetheless, in most of this chapter, we will consider a different approach. We will come back to the original idea consisting in learning a \textit{single} node embedding space, and therefore represent each node by a single embedding vector. Such an approach has a stronger interpretability power. It permits to better visualize representations, and is preferable for other tasks than link prediction (e.g., for community detection, as one can directly run a $k$-means algorithm on such representations). Besides, we will show in Section~\ref{c5s54} that this single vector approach significantly outperforms Source/Target GAE and VGAE on various directed link~prediction~tasks.

\section{Gravity-Inspired GAE and VGAE}
\label{c5s53}

In this section, we introduce our proposed method to learn node embedding spaces from directed graphs, and subsequently perform directed link prediction, using the GAE and VGAE frameworks. Our main challenge is the following:\textit{ how to effectively reconstruct asymmetric relations from encoded representations that are (unique) vectors in a node embedding where inner product and common distances are symmetric?}

To overcome this challenge, we resort to classical mechanics and especially to Newton's theory of universal gravitation~\cite{newton1687philosophiae}. We propose an analogy between vectors in a node embedding space and celestial objects in space. Specifically, even if the Earth-Moon distance is symmetric, the \textit{acceleration} of the Moon towards the Earth due to gravity is larger than the acceleration of the Earth towards the Moon. As explained below, this is due to the fact that the Earth is more massive. In the remainder of this section, we transpose these notions of \textit{mass} and \textit{acceleration} to node embedding spaces, to build up our asymmetric graph decoding scheme.

\subsection{From Physics to Node Representations}
\label{c5s531}

\paragraph{Newton's Theory of Universal Gravitation}

According to Newton's theory of universal gravitation \cite{newton1687philosophiae}, each particle in the universe attracts the other particles through a force called \textit{gravity}. This force is proportional to the product of the masses of the particles, and inversely proportional to the squared distance between their centers. More formally, let us denote by $m_1$ and $m_2$ the positive masses of two objects $1$ and $2$ and by $r$ the distance between their centers. Then, the gravitational force $F$ attracting the two objects is:
\begin{equation}F = \frac{Gm_1m_2}{r^2},\end{equation}
where $G$ is the gravitational constant \cite{cavendish1798xxi}. Then, using Newton's second law of motion \cite{newton1687philosophiae}, we derive $a_{1 \rightarrow 2}$, the acceleration of object $1$ towards object $2$ due to gravity:
\begin{equation}a_{1 \rightarrow 2} =  \frac{F}{m_1} = \frac{Gm_2}{r^2}.\end{equation}
Likewise, the acceleration $a_{2 \rightarrow 1}$ of $2$ towards $1$ due to gravity is:
\begin{equation}a_{2 \rightarrow 1} = \frac{F}{m_2} = \frac{Gm_1}{r^2}.\end{equation}
We note that $a_{1 \rightarrow 2} \neq a_{2 \rightarrow 1}$ when $m_1 \neq m_2$. More precisely, we have $a_{1 \rightarrow 2} > a_{2 \rightarrow 1}$ when $m_2 > m_1$ and conversely, i.e., the acceleration of the least massive object towards the most massive object due to gravity is higher.

Despite being superseded in modern physics by Einstein's theory of general relativity \cite{einstein1915erklarung}, describing gravity not as a force but as a consequence of spacetime curvature, Newton's law of universal gravitation is still used in many applications, as the theory provides precise approximations of the effect of gravity when gravitational fields are not extreme. In this chapter, we draw inspiration from this theory, notably from the formulation of acceleration, to build our proposed autoencoders. We highlight that Newtonian gravity concepts were already successfully leveraged for graph visualization~\cite{gravity1} and to build symmetric node similarity scores~\cite{gravity2}.

\paragraph{From Physics to Node Embedding Spaces} We come back to our initial analogy between celestial objects in space and node embedding representations. In this paragraph, let us assume that we have at our disposal a model that is able to learn, for each node $i \in \mathcal{V}$~of~a~directed~graph:
\begin{itemize}
    \item a node embedding vector $z_i \in \mathbb{R}^d$, of dimension $d \ll n$, as before;
    \item but also a new \textit{mass parameter} $m_i \in \mathbb{R}^+$.
\end{itemize}.
We explain how to learn $m_i$ in the next sections. Such a parameter $m_i$ would capture the propensity of $i$ to attract other nodes from its neighborhood in this graph, i.e., to make them point towards $i$ through a directed edge. From such an augmented model, we could apply Newton's equations in the resulting embedding. Specifically, we propose to use the \textit{acceleration} $a_{i \rightarrow j} = \frac{G m_j}{r^2}$ of a node $i$ towards a node $j$ due to gravity in the embedding as an indicator of the likelihood that $i$ is connected to $j$ in the directed graph, with $r^2 = \|z_i - z_j\|_2^2$. In a nutshell:
\begin{itemize}
    \item the numerator captures the fact that some nodes are more influential than others in the graph. For instance, in a citation network of scientific publications, seminal groundbreaking articles are more influential and should be more cited than others. Here, the larger $m_j$ the more likely $i$ will be connected to $j$ via the $(i,j)$ directed edge;
    \item the denominator highlights that nodes with structural proximity in the graph, typically with a common neighborhood, are more likely to be connected, provided that the model effectively manages to embed these nodes close to each other in the embedding space. For instance, in a scientific publications citation network, an article $i$ will more likely cite an article $j$ if it comes from a similar field of study.
\end{itemize}

More precisely, instead of directly dealing with $a_{i \rightarrow j}$, we use $\log a_{i \rightarrow j}$ in the remainder of this chapter. Using the logarithm has two advantages. Firstly, thanks to its concavity we limit the potentially large values resulting from the acceleration towards very central nodes. Also, $\log a_{i \rightarrow j}$ can be negative, which is more convenient to reconstruct an unweighted edge using a sigmoid activation function, as follows:
\begin{equation}
\hat{A}_{ij} = \sigma(\log a_{i \rightarrow j}) = \sigma(\underbrace{\log G m_j}_{\text{denoted~} \tilde{m}_j} - \log \|z_i - z_j\|_2^2).
\end{equation}

\subsection{Gravity-Inspired GAE}
\label{c5s532}

For pedagogical purposes, we assumed that we had, at our disposal, a model providing the $m_i$ masses. In this section, we detail how to actually learn them,~using~graph~autoencoders.

\paragraph{Encoder}

For the GAE encoder, we still leverage a multi-layer GCN, as defined in Definition~\ref{def:gcn} and processing an adjacency matrix $A$ potentially combined with a node features matrix $X$. However, such a GCN will now assign a vector of size $(d +1)$ to each node of the graph, instead of $d$ as in standard GAE models. The first $d$ dimensions correspond to the embedding vector of the node, i.e., $z_i$, where $d \ll n$ still denotes the dimension of the node embedding space. The last dimension will correspond to the model's estimate of $\tilde{m}_i = \log G m_i$. To~sum~up,~we~have:
\begin{equation}
\tilde{Z} = (Z,\tilde{M}) = \text{GCN}(A,X),
\end{equation}
where, as in previous chapters, $Z$ is the $n \times d$ node embedding matrix, and where $\tilde{M}$ is the $n$-dimensional vector of all values of $\tilde{m}_i$. $\tilde{Z} = (Z,\tilde{M})$ denotes the $n \times (d+1)$ matrix row-concatenating $Z$ and $\tilde{M}$. We note that learning $\tilde{m}_i$ is equivalent to learning $m_i$, but is also more convenient since we get rid of the gravitational constant $G$ and of the logarithm.

In this GCN encoder, as we process directed graphs, we need to differentiate between incoming and outcoming edges during message passing. In this chapter, we therefore replace the usual symmetric normalization of $A$ from Definition~\ref{def:norm_c2} by the out-degree normalization $\tilde{A}_{\text{out}}$:

\begin{definition}
The \textit{out-degree normalization} of the adjacency matrix $A$ of a graph $\mathcal{G} = (\mathcal{V}, \mathcal{E})$ with diagonal out-degree matrix $D_{\text{out}}$ (as defined in Definition~\ref{def:degree_mat}) is:
\begin{equation}
\tilde{A}_{\text{out}} = (D_{\text{out}}+I_n)^{-1}(A+I_n).
\end{equation}
\label{def:norm_c6}
\end{definition}
Therefore, at each layer of the GCN, the vector of a node becomes a weighted average of hidden vectors from the previous layer of the neighbors \textit{to which it points}, together with its own vector.

\paragraph{Decoder and Optimization} We leverage the previously defined logarithmic version of acceleration, together with a sigmoid activation, to reconstruct an estimation of the adjacency matrix $A$ from $Z$ and $\tilde{M}$. Denoting $\hat{A}$ the reconstruction of $A$, we have:
\begin{equation}\hat{A}_{ij} = \sigma(\tilde{m}_j - \log \|z_i - z_j\|^2_2).\end{equation}
For weighted graphs, $\hat{A}_{ij}$ is an estimation
of some true normalized weight on the edge connecting $i$ to $j$. For unweighted graphs, it corresponds to the probability of a missing edge from $i$ to $j$. Contrary to the inner product decoder, we generally have $\hat{A}_{ij} \neq \hat{A}_{ji}$. This approach is therefore suitable for directed graph reconstruction. During the training phase, we tune the GCN weights of this model in a similar fashion w.r.t. standard GAEs (see Section~\ref{c2s241}), i.e., we iteratively minimize a weighted cross entropy reconstruction loss \cite{kipf2016-2}, by gradient descent~\cite{goodfellow2016deep}.

\subsection{Gravity-Inspired VGAE}
\label{c5s533}

In this chapter, we also extend our proposed decoder to variational graph autoencoders.

\paragraph{Encoder} We extend the VGAE inference model from Kipf and Welling~\cite{kipf2016-2}. Formally, using the same notation as in  Section~\ref{c2s242} from Chapter~\ref{chapter_2}, we set:
\begin{equation}
q(\tilde{Z}|A,X) = \prod_{i=1}^n q(\tilde{z}_i|A,X), \text{~with~}
q(\tilde{z}_i|A,X) = \mathcal{N}(\tilde{z}_i|\mu_i, \text{diag}(\sigma_i^2)),
\end{equation}
where $\tilde{z}_i = (z_i, \tilde{m}_i)$ is the $(d+1)$-dimensional vector concatening the $d$-dimensional embedding vector $z_i$ and the scalar $\tilde{m}_i$. As is the case for standard VGAE, Gaussian parameters are learned from two GCNs, i.e., $\mu = \text{GCN}_{\mu}(A,X)$, with $\mu$ denoting the $n \times (d+1)$ matrix stacking up mean vectors $\mu_i$ for each node. Similarly, $\log \sigma = \text{GCN}_{\sigma}(A,X)$.

\paragraph{Decoder and Optimization} As in Section~\ref{c2s242}, the actual embedding vectors $\tilde{z}_i$ are sampled from the above distributions.
Then, we incorporate our gravity-inspired decoding scheme into the generative model attempting to reconstruct $A$:
\begin{equation}p(A|\tilde{Z}) = \prod_{i=1}^n \prod_{j=1}^n p(A_{ij}|\tilde{z}_i, \tilde{z}_j), \end{equation}
with: 
\begin{equation}
p(A_{ij}|\tilde{z}_i, \tilde{z}_j) = \hat{A}_{ij} = \sigma(\tilde{m}_{j} - \log \|z_i - z_j\|_2^2).
\end{equation}

During training, weights of the two GCN encoders are tuned by iteratively maximizing, by gradient ascent, the corresponding ELBO objective acting as a reconstruction quality measure.

\subsection{Generalization of the Decoding Scheme}
\label{c5s534}

One can improve the flexibility of our decoders, both in the GAE and in the VGAE settings, by introducing an additional parameter $\lambda \in \mathbb{R}^+$ and by reconstructing $\hat{A}_{ij}$ as follows:
\begin{equation}\hat{A}_{ij} = \sigma(\tilde{m}_j - \lambda \log \|z_i - z_j\|^2_2).
\label{eq:lambdac5}
\end{equation}
Decoders from Sections~\ref{c5s532}~and~\ref{c5s533} correspond to special cases of Equation~\eqref{eq:lambdac5} where we set $\lambda~=~1$. Such a parameter can be tuned by cross-validation on link prediction tasks (See section~\ref{c5s54}).
Our interpretation of $\lambda$ is twofold. Firstly, it balances the relative importance of distances in the embedding for reconstruction w.r.t. the mass attraction parameter. Then, from a ``physics'' point of view, it is equivalent to replacing the squared distance in Newton's formula with a distance to the power of $2\lambda$. In our experimental analysis on link prediction, we will provide insights on when and why deviating from Newton's actual theory (i.e., $\lambda = 1$) is relevant.

\subsection{On Complexity and Scalability}
\label{c5s535}

Assuming featureless nodes, a sparse representation of the adjacency matrix $A$ with $m$ non-zero entries, and considering that our models return a dense $n \times (d+1)$ embedding matrix $\tilde{Z}$, then the space complexity of our approach is $O(m + n(d+1))$, both in the GAE and VGAE frameworks. If nodes also have features summarized in the $n \times f$ matrix $X$, then the space complexity becomes $O(m + n(f+d+1))$, with $d \ll n$ and $f \ll n$ in practice. Therefore, as is the case for standard GAE and VGAE models \cite{kipf2016-2}, space complexity increases linearly w.r.t. the size of the graph.

Moreover, due to the pairwise computations of $L_2$ distances between all  $d$-dimensional vectors $z_i$ and $z_j$ involved in our gravity-inspired decoding scheme, our models have a quadratic time complexity $O(dn^2)$ w.r.t. the number of nodes in the graph, similarly to standard GAE and VGAE models. As our experiments from Section~\ref{c5s54} will focus on medium-size graphs, such a quadratic complexity will be computationally affordable.

One can extend our method to larger graphs, e.g., with millions of nodes and edges, by applying the degeneracy framework proposed in Chapter~\ref{chapter_3}, or a variant of this approach involving \textit{directed graph degeneracy} a.k.a. ``D-Cores'' concepts \cite{giatsidis2013d}. One can also apply the FastGAE method from Chapter~\ref{chapter_4}, either with degree-based or core-based sampling, or alternatively with a refined sampling strategy taking into account the directionality of edges. Floyd Everest (\texttt{fleverest} on GitHub) recently combined FastGAE with Gravity-Inspired GAE/VGAE models in one of his own repositories\footnote{\href{https://github.com/fleverest/Gravity_FastGAE}{https://github.com/fleverest/Gravity\_FastGAE}}. In Chapter~\ref{chapter_8}, we will also mention a similar combination when leveraging Gravity-Inspired GAE and VGAE models for graph-based similar artists ranking on Deezer.

\section{Experimental Analysis}
\label{c5s54}


\subsection{Three Directed Link Prediction Tasks}
\label{c5s541}

We consider the following three learning tasks for our experimental evaluation of our models.

\paragraph{Task 1: General Directed Link Prediction}

The first task is referred to as \textit{general directed link prediction}. Similarly to experiments from Chapter~\ref{chapter_3}~and~\ref{chapter_4}, we train models on incomplete versions of graphs where $15\%$ of edges were randomly removed. We take directionality into account in the masking process. In other words, if a link between node $i$ and $j$ is reciprocal, we can possibly remove the $(i,j)$ edge but still observe the reverse $(j,i)$ edge in the training graph. Then, we create validation and test sets from removed edges and from the same number of randomly sampled pairs of unconnected nodes. In the following, the validation set contains $5\%$
of edges, and the test set contains $10\%$ of edges. As in previous chapters, we evaluate the performance of our models on a binary classification task consisting in discriminating the actually removed edges from the fake ones, and we compare results using the~AUC~and~AP~scores.

This setting corresponds to the most general formulation of link prediction. However, due to the large number of unconnected pairs of nodes in numerous real-world graphs, we expect the impact of directionality on performances to be limited. Indeed, for each actual unidirectional edge $(i,j)$ from the graph, it is unlikely to retrieve the reverse (unconnected) pair $(j,i)$ among negative samples in the test set. As a consequence, models focusing on graph proximity and ignoring the direction of the link, such as standard GAEs and VGAEs, might still perform fairly well on such a task. For this reason, in the remainder of this section, we also propose and study two additional learning tasks, designed to reinforce the importance of directionality learning.

\paragraph{Task 2: Biased Negative Samples (B.N.S.) Link Prediction}

For the second task, we also train models on incomplete versions of graphs where $15\%$ of edges were removed: $5\%$ initially extracted as a validation set, and $10\%$ acting as a test set. However, the removed edges are all \textit{unidirectional}, i.e., $(i,j)$ exists but not $(j,i)$. In this task, the reverse node pairs are included in validation and test sets and constitute negative samples. In other words, all node pairs from validation and test sets are included in \textit{both} directions. As for Task~1, we evaluate the performance of our models on a binary classification task consisting in discriminating actual edges from fake ones, and therefore evaluate the ability of our models to correctly reconstruct $A_{ij} = 1$ and $A_{ji} = 0$ \textit{simultaneously}.

This task has been presented by Zhou~et~al.~\cite{zhou2017scalable} under the name \textit{biased negative samples link prediction}. It is more challenging than Task~1, as the ability to reconstruct asymmetric relations is more crucial. Models ignoring directionality, such as standard GAEs and VGAEs who always predict $\tilde{A}_{ij} = \tilde{A}_{ji}$, will fail in~such~a~setting.

\paragraph{Task 3: Bidirectionality Prediction}

As a third task, we evaluate the ability of our models to discriminate \textit{bidirectional} edges, i.e., reciprocal connections, from \textit{unidirectional} edges. We create a training graph by removing at random one of the two directions of all bidirectional edges for an initial graph. Therefore, the training graph only has unidirectional connections. Then, we once again consider a binary classification problem, aiming to retrieve bidirectional edges in a test set composed of their removed direction and of the same number of reverse directions from unidirectional edges (that are therefore fake edges a.k.a. negative samples). In other words, for each pair of nodes $(i,j)$ from the test set, we observe a connection from $j$ to $i$ in the incomplete training graph, but only half of them are reciprocal. This third evaluation task, referred to as \textit{bidirectionality prediction} in this chapter, also strongly relies on directionality learning. As a consequence, as for Task 2, standard GAEs and VGAEs are expected to~perform~poorly.

\subsection{Experimental Setting}
\label{c5s542}

\paragraph{Datasets} We provide experiments on three publicly available real-world directed graphs. Firstly, we consider the Cora and Citeseer \textit{citation graphs} already described in Chapter~\ref{chapter_3}, and consisting of scientific publications citing one another. Contrary to experiments from Chapters~\ref{chapter_3}~and~\ref{chapter_4}, we deal with the original \textit{directed} versions of these graphs.

We also consider the Google directed web graph from Konect\footnote{\href{http://konect.cc/networks/}{https://konect.cc/networks/}}. The 15~763 nodes of this graph are web pages, and its 171~206 directed edges represent hyperlinks between these pages. We point out that this graph is different from the large Google web graph from SNAP, that we mentioned in Chapter~\ref{chapter_3}~and~\ref{chapter_4}. To avoid confusion, we refer to it as ``Google-Medium'' in the remainder of this chapter, due to its smaller size w.r.t. the Google graph from previous chapters.

The Google-Medium graph is denser than Cora and Citeseer. It also has a higher proportion of bidirectional edges. Specifically, we have 2.86\%, 1.20\%, and 14.55\% of bidirectional edges in Cora, Citeseer, and Google-Medium respectively. The three graphs are unweighted and featureless.

\paragraph{Models: Standard and Gravity-Inspired GAE and VGAE}

We train Gravity-Inspired GAE and VGAE models for each graph. For comparison purposes, we also train standard GAE and VGAE from Kipf and Welling~\cite{kipf2016-2}. Each of these four models includes a two-layer GCN encoder with a 64-dimensional hidden layer and with out-degree left normalization of $A$ as defined in mentioned in Section~\ref{c5s532}. All models are trained for 200 epochs and return 32-dimensional embedding vectors (i.e., $d = 32$). We used the Adam optimizer \cite{kingma2014adam}, apply a learning rate of 0.1 for Cora and Citeseer and 0.2 for Google-Medium, trained models without dropout, performing full-batch gradient descent (or ascent for VGAE), and using the reparameterization trick \cite{kingma2013vae} in the case of VGAEs. Also, for Task~1 and Task~3 we picked $\lambda = 1$ (respectively $\lambda = 10$)  for Cora and Citeseer (resp. for Google-Medium). For Task~2 we picked $\lambda = 0.05$ for all three graphs, which we interpret and discuss in the next subsections. All hyperparameters were tuned from validation AUC scores on Task 1, i.e., on the general directed link prediction task.

\paragraph{Models: Other Baselines}

Besides standard GAEs and VGAEs, we also compare the performance of our models to the graph embedding methods introduced in Section~\ref{c5s522}:
\begin{itemize}
    \item our own Source/Target GAE and VGAE, extending the source/target paradigm to GAE and VGAE, and trained with similar settings w.r.t. standard and gravity-inspired models;
    \item HOPE \cite{ou2016asymmetric}, setting $\beta = 0.01$ and with source and target vectors of dimension 16, in order to learn 32-dimensional node representations.
    \item APP \cite{zhou2017scalable}, training models over 100 iterations to learn 16-dimensional source and target vectors, i.e., 32-dimensional node representations. We adopted a similar setting and similar hyperparameters w.r.t. Zhou~et~al.~\cite{zhou2017scalable}'s public implementation.
    \item for comparison purposes, in our experiments we also train \textit{node2vec} models \cite{grover2016node2vec} that, while dealing with directionality in random walks, only return one 32-dimensional embedding vector per node. We rely on symmetric inner products with sigmoid activation for link prediction, and we therefore expect node2vec to underperform w.r.t. APP and HOPE on Tasks~2~and~3. We trained models with consistent hyperparameters w.r.t. Chapter~\ref{chapter_3}.
\end{itemize}
We used Python and especially the TensorFlow library~\cite{abadi2016tensorflow}, except for APP where we used the authors' Java implementation \cite{zhou2017scalable}. We trained models on an NVIDIA GTX 1080 GPU and ran other operations on a double Intel Xeon Gold 6134 CPU.

\subsection{Results and Discussion}
\label{c5s543}

We now present our experimental results. Our source code is publicly available on GitHub\footnote{\href{https://github.com/deezer/gravity_graph_autoencoders}{https://github.com/deezer/gravity\_graph\_autoencoders}}.

\paragraph{Results} Table~\ref{c5_maintable} reports mean AUC and AP scores, along with standard errors over 100 runs with different test sets, for each dataset and the three tasks under consideration. Overall, our Gravity-Inspired GAE and VGAE models achieve very competitive results.

On Task 1, standard GAE and VGAE models, despite ignoring directionality for graph reconstruction, still perform fairly well (e.g., with an $82.79\%$ AUC score for the standard VGAE on Cora). This emphasizes the limited impact of directionality on performances for such a task, as expected in Section~\ref{c5s541}. Nonetheless, our gravity-inspired models significantly outperform their standard counterparts (e.g., with a $91.92\%$ AUC score for our Gravity-Inspired VGAE on Cora), confirming the relevance of capturing both proximity \textit{and} directionality for general directed link prediction. Moreover, our models often reach comparable results w.r.t. the other baselines specifically designed for directed graphs. Among them, APP is the best on the three datasets, together with the Source/Target GAE on the Google-Medium graph.

\begin{table}[t]
  \centering
  \caption[Directed link prediction on all graphs using Gravity-Inspired GAE and VGAE]{Directed link prediction on the Cora, Citeseer, and Google-Medium graphs, using our Gravity-Inspired GAE/VGAE, standard GAE/VGAE, and other baselines. Scores are reported for the three tasks described in Section~\ref{c5s541}, and are averaged over 100 runs. All models learn embedding vectors of dimension $d = 32$. \textbf{Bold} numbers correspond to the best scores. Scores \textit{in italic} are within one standard deviation range from~the~best~ones.}
  \label{c5_maintable}
 \resizebox{1.0\textwidth}{!}{
  \begin{tabular}{c|c|cc|cc|cc}
    \toprule
    \textbf{Dataset}  & \textbf{Model} & \multicolumn{2}{c}{\textbf{Task 1: General Link Prediction}} & \multicolumn{2}{c}{\textbf{Task 2: B.N.S. Link Prediction}} & \multicolumn{2}{c}{\textbf{Task 3: Bidirectionality Prediction}} \\
    &  & \footnotesize \textbf{AUC (in \%)} & \footnotesize \textbf{AP (in \%)} & \footnotesize \textbf{AUC (in \%)} & \footnotesize \textbf{AP (in \%)} & \footnotesize \textbf{AUC (in \%)} & \footnotesize \textbf{AP (in \%)}\\
    \midrule
    \midrule
    \textbf{Cora} & \textit{Gravity-Inspired VGAE (ours)} & $91.92 \pm 0.75$ & $\textit{92.46} \pm \textit{0.64}$ & $\textbf{83.33} \pm \textbf{1.11}$ & $\textbf{84.50} \pm \textbf{1.24}$ & $\textit{75.00} \pm \textit{2.10}$ & $\textbf{73.87} \pm \textbf{2.82}$ \\
     & \textit{Gravity-Inspired GAE (ours)} & $87.79 \pm 1.07$ & $90.78 \pm 0.82$ & $\textit{83.18} \pm \textit{1.12}$ & $\textit{84.09} \pm \textit{1.16}$ & $\textbf{75.57} \pm \textbf{1.90}$ & $\textit{73.40} \pm \textit{2.53}$ \\
     \cmidrule{2-8}
     & Standard VGAE & $82.79 \pm 1.20$ & $86.69 \pm 1.08$ & $50.00 \pm 0.00$ & $50.00 \pm 0.00$ & $58.12 \pm 2.62$ & $59.70 \pm 2.08$ \\
     & Standard GAE & $81.34 \pm 1.47$ & $82.10 \pm 1.46$ & $50.00 \pm 0.00$ & $50.00 \pm 0.00$ & $53.07 \pm 3.09$ & $54.60 \pm 3.13$ \\
     & Source/Target VGAE & $85.34 \pm 1.29$ & $88.35 \pm 0.99$ & $63.00 \pm 1.05$ & $64.62 \pm 1.37$ & $\textit{75.20} \pm \textit{2.62}$ & $\textit{73.86} \pm \textit{3.04}$ \\
     & Source/Target GAE & $82.67 \pm 1.42$ & $83.25 \pm 1.51$ & $57.81 \pm 2.64$ & $57.66 \pm 3.35$ & $65.83 \pm 3.87$ & $63.15 \pm 4.58$ \\
     & APP & $\textbf{93.92} \pm \textbf{1.01}$ & $\textbf{93.26} \pm \textbf{0.60}$ & $69.20 \pm 0.65$ & $67.93 \pm 1.09$ & $72.85 \pm 1.91$ & $70.97 \pm 2.60$ \\
     & HOPE & $80.82 \pm 1.63$ & $81.61 \pm 1.08$  & $61.84 \pm 1.84$ & $63.73 \pm 1.12$ & $65.11 \pm 1.40$ & $64.24 \pm 1.18$ \\
     & node2vec & $79.01 \pm 2.00$ & $84.20 \pm 1.62$ & $50.00 \pm 0.00$ & $50.00 \pm 0.00$ & $66.97 \pm 1.41$ & $67.61 \pm 1.80$ \\
    \midrule
    \midrule
    \textbf{Citeseer} & \textit{Gravity-Inspired VGAE (ours)} & $\textit{87.67} \pm \textit{1.07}$ & $\textit{89.79} \pm \textit{1.01}$ & $\textbf{76.19} \pm \textbf{1.35}$ & $\textbf{79.27} \pm \textbf{1.24}$ & $\textbf{71.61} \pm \textbf{3.20}$ & $\textbf{71.87} \pm \textbf{3.87}$ \\
     & \textit{Gravity-Inspired GAE  (ours)} & $78.36 \pm 1.55$ & $84.75 \pm 1.10$ & $\textit{75.32} \pm \textit{1.53}$ & $\textit{78.47} \pm \textit{1.27}$ & $\textit{71.48} \pm \textit{3.64}$ & $\textit{71.50} \pm \textit{3.62}$ \\
     \cmidrule{2-8}
     & Standard VGAE & $78.56 \pm 1.43$ & $83.66 \pm 1.09$ & $50.00 \pm 0.00$ & $50.00 \pm 0.00$ & $47.66 \pm 3.73$ & $50.31 \pm 3.27$ \\
     & Standard GAE & $75.23 \pm 2.13$ & $75.16 \pm 2.04$ & $50.00 \pm 0.00$ & $50.00 \pm 0.00$ & $45.01 \pm 3.75$ & $49.79 \pm 3.71$ \\
     & Source/Target VGAE & $79.45 \pm 1.75$ & $83.66 \pm 1.32$ & $57.32 \pm 0.92$ & $61.02 \pm 1.37$ & $69.67 \pm 3.12$ & $67.05 \pm 4.10$ \\
     & Source/Target GAE & $73.97 \pm 3.11$ & $75.03 \pm 3.37$ & $56.97 \pm 1.33$ & $57.62 \pm 2.62$ & $54.88 \pm 6.02$ & $55.81 \pm 4.93$ \\
     & APP & $\textbf{88.70} \pm \textbf{0.92}$ & $\textbf{90.29} \pm \textbf{0.71}$ & $64.35 \pm 0.45$ & $63.70 \pm 0.51$ & $64.16 \pm 1.90$ & $63.77 \pm 3.28$ \\
     & HOPE & $72.91 \pm 0.59$ & $71.29 \pm 0.52$ & $60.24 \pm 0.51$ & $61.28 \pm 0.57$ & $52.65 \pm 3.05$ & $54.87 \pm 1.67$ \\
     & node2vec & $71.02 \pm 1.78$  & $77.70 \pm 1.22$ & $50.00 \pm 0.00$ & $50.00 \pm 0.00$ & $61.08 \pm 1.88$ & $63.63 \pm 2.77$ \\
    \midrule
    \midrule
    \textbf{Google} & \textit{Gravity-Inspired VGAE (ours)} & $\textbf{97.84} \pm \textbf{0.25}$ & $\textit{98.18} \pm \textit{0.14}$ & $\textbf{88.03} \pm \textbf{0.25}$ & $\textbf{91.04} \pm \textbf{0.14}$ & $84.69 \pm 0.31$ & $84.89 \pm 0.30$ \\
    \textbf{Medium} & \textit{Gravity-Inspired GAE (ours)} & $\textit{97.77} \pm \textit{0.10}$ & $\textbf{98.43} \pm \textbf{0.10}$ & $\textit{87.71} \pm \textit{0.29}$ & $\textit{90.84} \pm \textit{0.16}$ & $\textbf{85.82} \pm \textbf{0.63}$ & $\textbf{85.91} \pm \textbf{0.50}$ \\
     \cmidrule{2-8}
     & Standard VGAE & $87.14 \pm 1.20$ & $88.14 \pm 0.98$ & $50.00 \pm 0.00$ & $50.00 \pm 0.00$ & $40.03 \pm 4.98$ & $44.69 \pm 3.52$ \\
     & Standard GAE & $91.34 \pm 1.13$ & $92.61 \pm 1.14$ & $50.00 \pm 0.00$ & $50.00 \pm 0.00$ & $41.35 \pm 1.92$ & $41.92 \pm 0.81$ \\
     & Source/Target VGAE & $96.33 \pm 1.04$ & $96.24 \pm 1.06$ & $85.30 \pm 3.18$ & $84.69 \pm 4.42$ & $75.11 \pm 2.07$ & $73.63 \pm 2.06$ \\
     & Source/Target GAE & $\textit{97.76} \pm \textit{0.41}$ & $\textit{97.74} \pm \textit{0.40}$ & $86.16 \pm 2.95$ & $86.26 \pm 3.33$ & $82.27 \pm 1.29$ & $80.10 \pm 1.80$ \\
     & APP & $97.04 \pm 0.10$ & $96.97 \pm 0.11$ & $83.06 \pm 0.46$ & $85.15 \pm 0.42$ & $73.43 \pm 0.16$ & $68.74 \pm 0.19$ \\
     & HOPE & $81.16 \pm 0.67$ & $83.02 \pm 0.35$ & $74.23 \pm 0.80$ & $72.70 \pm 0.79$ & $70.45 \pm 0.18$ & $70.84 \pm 0.22$ \\
     & node2vec & $83.11 \pm 0.27$ & $85.79 \pm 0.30$ & $50.00 \pm 0.00$ & $50.00 \pm 0.00$ & $78.99 \pm 0.35$ & $76.72 \pm 0.53$ \\
    \bottomrule
  \end{tabular}
  }
\end{table}

On Task 2, our models consistently achieve the best performances (e.g., with a top $76.19\%$ AUC score on Citeseer, $11+$ points above the best baseline). Models ignoring directionality for prediction, i.e., node2vec and standard GAE/VGAE, totally fail ($50.00\%$ AUC and AP scores on all graphs, corresponding to the random classifier level) which was expected since test sets include both directions of each node pair. Experiments on Task~3 confirm the superiority of our approach when dealing with challenging tasks where directionality learning is crucial. On this last task, gravity-inspired models also outperform alternative approaches (e.g., with a top $85.82\%$ AUC score for Gravity-Inspired GAE on Google-Medium).

We hardly found any consistent and significant performance gap between \textit{determinic} GAEs and their \textit{variational} counterparts. This result is in phase with previous empirical results in the literature \cite{kipf2016-2,pan2018arga} on undirected graphs, as well as in our own previous insights from Chapters~\ref{chapter_3}~and~\ref{chapter_4}. In futures studies, we might investigate alternative prior distributions for VGAE, to challenge the Gaussian hypothesis that, despite being convenient for computations, might not be optimal \cite{kipf2016-2}. 
Last, we note that all GAE and VGAE models required a comparable training time of roughly 7~seconds (respectively 8~seconds, 5~minutes) for Cora (resp. for Citeseer, for Google-Medium) on our machine. Baselines were faster: for instance, on Google-Medium, 1~minute (resp. 1.30~minutes, 2~minutes) were required to train HOPE (resp. APP, node2vec).

\begin{figure}[H]
    \centering
    \includegraphics[width=1.0\textwidth]{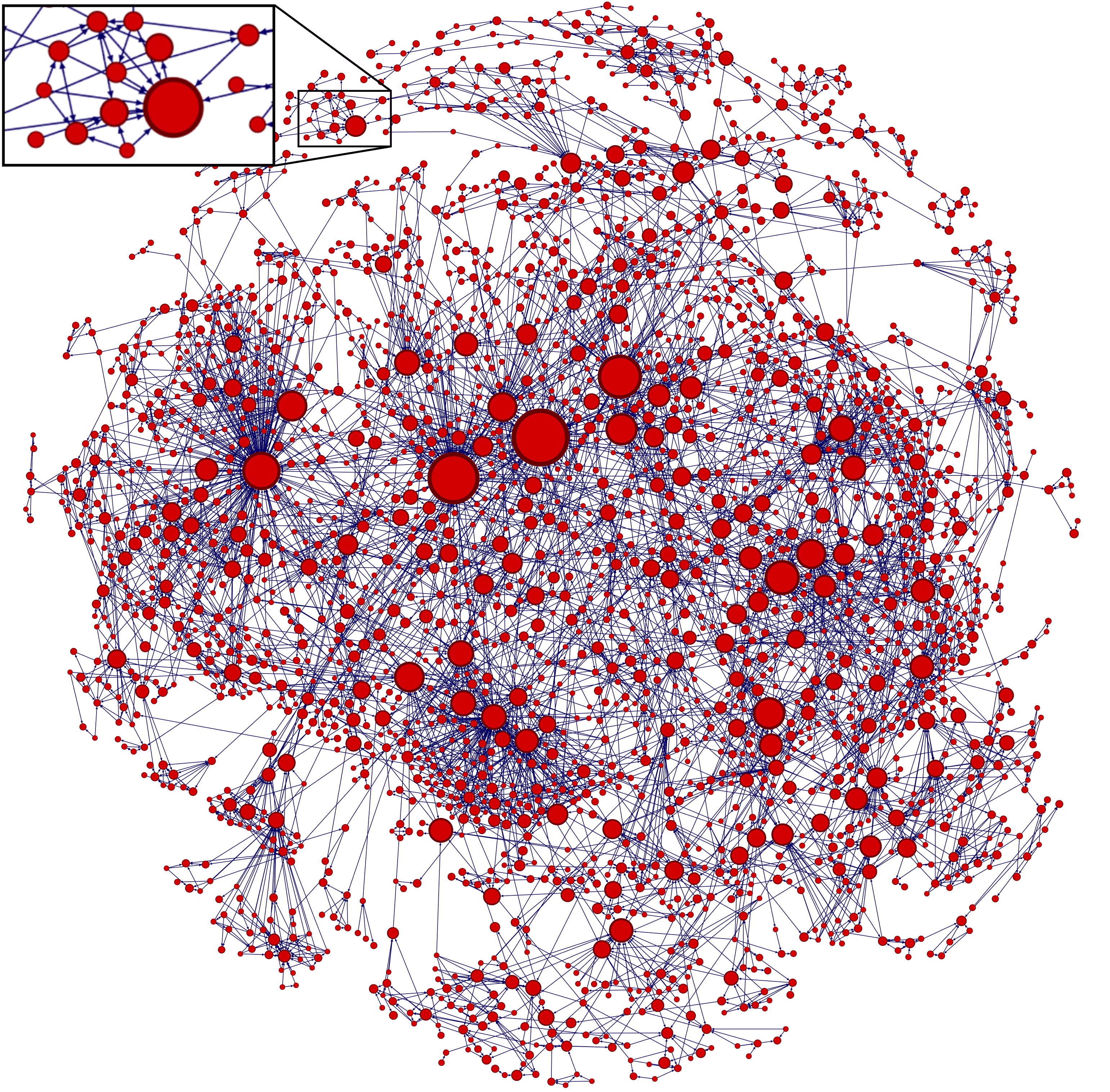}
     \caption[Visualization of node embedding representations from Gravity-Inspired VGAE]{Visualization of the Cora graph based on node embedding representations learned from our Gravity-Inspired VGAE model. In this graph, nodes are scaled using the mass parameters $\tilde{m}_i$. The node separation is based on distances in the embedding, using the method described by Fruchterman and Reingold~\cite{fruchterman1991graph} on Gephi.}   
    \label{fig:cora_visu_c5}
\end{figure}

\paragraph{More insights on $\tilde{m}_i$} 
To pursue our experimental analysis, we propose a discussion on the nature of $\tilde{m}_i$ (in this paragraph), followed by discussions on the role of $\lambda$ to balance node proximity and influence and on some possible extensions of our work (in the next paragraphs).

Figure~\ref{fig:cora_visu_c5} provides a visualization of the Cora graph, using node embedding representations and $\tilde{m}_i$ masses learned by our Gravity-Inspired VGAE model. In such a visualization, we observe that nodes with smaller masses tend to be connected to nodes with larger masses from their embedding neighborhood, which was expected by the design of our decoding scheme. 

From Figure~\ref{fig:cora_visu_c5}, one might argue that $\tilde{m}_i$ tend to reflect the node \textit{centrality} in the graph. In this direction, we compared $\tilde{m}_i$ to the most common graph centrality measures. Specifically, in Table~\ref{tab:correlation_metrics} we report Pearson correlation coefficients of $\tilde{m}_i$ w.r.t. the following measures, computed using the NetworkX~\cite{networkx} Python library:
\begin{itemize}
    \item the \textit{in-degree} and \textit{out-degree} of the node, i.e., the number of edges coming into and going out of the node respectively;
    \item the \textit{betweenness centrality}, which is, for a node $i \in \mathcal{V}$, the sum of the fraction of shortest paths between node pairs going through $i$, i.e.,
    $c_B(i) = \sum_{(s,t) \in \mathcal{V}^2}\frac{\text{sp}(s,t|i)}{\text{sp}(s,t)}$,where $\text{sp}(s,t)$ is the number of shortest paths from node $s$ to node $t$, and $\text{sp}(s,t|i)$ is the number of those paths going through the node $i$ \cite{brandes2008variants};
    \item the \textit{PageRank} \cite{page1999pagerank}, computing a node ``importance'' ranking based on the structure of incoming links. It was originally designed to rank web pages;
    \item the \textit{Katz centrality}, a generalization of the eigenvector centrality~\cite{katz1953new}. The Katz centrality of a node~$i \in \mathcal{V}$ is: $c_{K}(i) = \alpha \sum_{j \in \mathcal{V}} A_{ij} c_{K}(j) + \beta$, where $A$ is the adjacency matrix with largest eigenvalue $\lambda_{\text{max}}$. Usually, $\beta = 1$ and $\alpha < \frac{1}{\lambda_{\text{max}}}$ \cite{katz1953new}.
\end{itemize}

\begin{multicols}{2}

We observe in Table~\ref{tab:correlation_metrics} that $\tilde{m}_i$ is positively correlated to all of these centrality measures, except for the out-degree where the correlation is negative (or almost null for Google-Medium), meaning that nodes with few edges going out of them tend to have larger values of $\tilde{m}_i$. Correlations are not perfect, which emphasizes that our models do \textit{not} exactly learn one of these measures. We also note that centralities are lower for Google-Medium, which might be due to the structure of this graph and especially to its density.
In our experiments, we tried to replace $\tilde{m}_i$ by any of these (normalized) centrality measures when performing link prediction. 
\columnbreak

\vspace{-0.5cm}
\captionof{table}[Pearson correlation coefficients of $\tilde{m}_i$ w.r.t. common centrality measures]{Pearson correlation coefficients of $\tilde{m}_i$ w.r.t. common centrality measures, for our Gravity-Inspired VGAE model. Correlation w.r.t. the Katz score is not reported on the Google-Medium graph due to its computational complexity.}\label{tab:correlation_metrics}
\vspace{0.2cm}
\resizebox{0.5\textwidth}{!}{
\begin{tabular}{c|c|c|c}
    \toprule
\textbf{Centrality} & \textbf{Cora} & \textbf{Citeseer} & \textbf{Google} \\
\textbf{Measures} &  &  & \textbf{Medium} \\
\midrule
\midrule
\textbf{In-degree} & $0.5960$ &  $0.6557$ & $0.1571$\\
\textbf{Out-degree} & $-0.2662$ & $-0.1994$ & $0.0559$ \\
\textbf{Betweenness} & $0.5370$ &  $0.4945$ & $0.2223$\\
\textbf{Pagerank} & $0.4143$ & $0.3715$ & $0.1831$\\
\textbf{Katz} & $0.5886$ & $ 0.6428$ & - \\
    \bottomrule
\end{tabular}
}
\end{multicols}
\vspace{-0.5cm}
Specifically, we tried to learn optimal vectors $z_i$ while these scores were fixed as masses values, achieving underperforming results.
For instance, we reached an $89.05\%$ main AUC score by using betweenness centrality on Cora instead of the actual $\tilde{m}_i$ learned by the VGAE, which is above standard VGAE ($82.79\%$ AUC) but below the Gravity-Inspired VGAE with optimal $\tilde{m}_i$ ($91.92\%$ AUC). Also, using centrality measures as initial values for $\tilde{m}_i$ before model training did not significantly impact performances in experiments.

\begin{figure}[t]
  \centering
  \subfigure[Task 1]{
  \scalebox{0.55}{\includegraphics{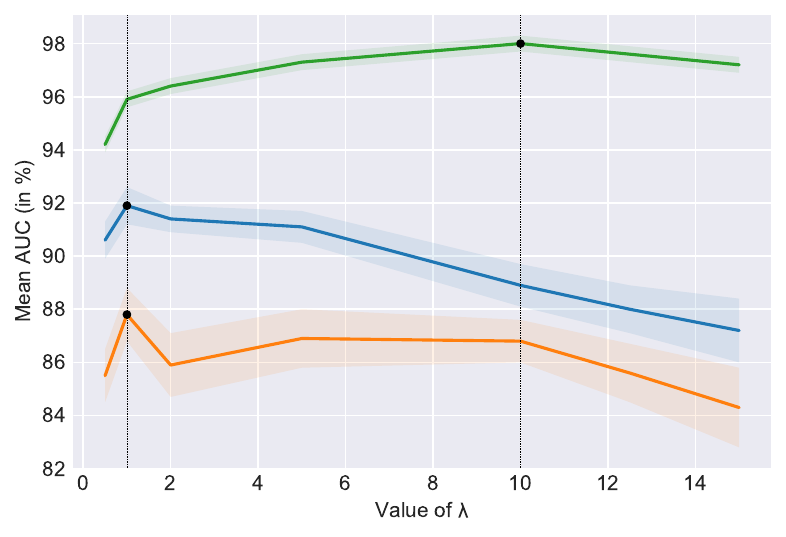}}}
  \subfigure[Task 2]{
  \scalebox{0.55}{\includegraphics{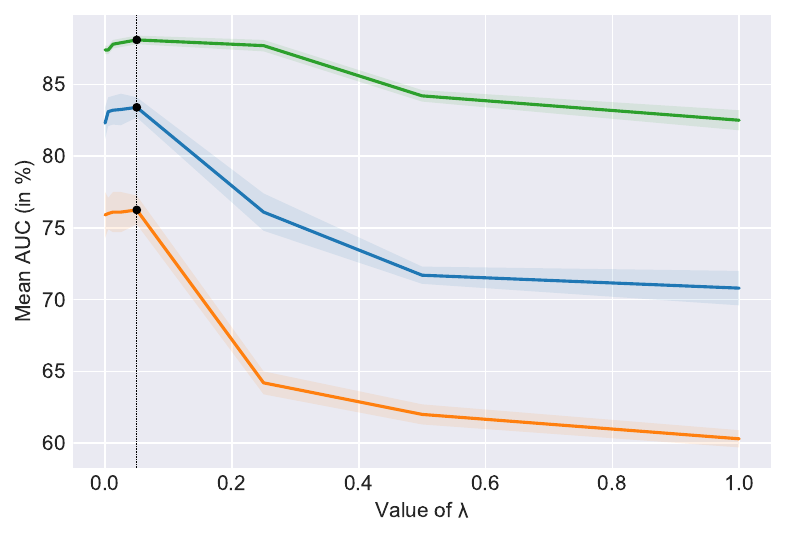}}}
  \subfigure[Task 3]{
  \scalebox{0.55}{\includegraphics{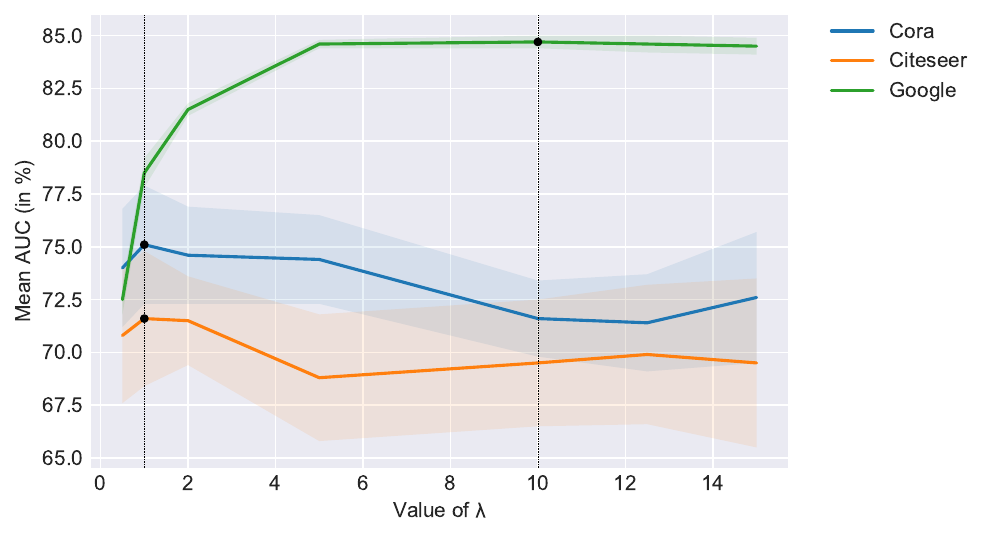}}}
  \caption[Optimal values of $\lambda$ for Gravity-Inspired VGAE]{Optimal values of the hyperparameter $\lambda$ w.r.t. validation AUC scores, for the Gravity-Inspired VGAE model, for all three tasks and for all three graphs under consideration.}
  \label{c5_parameterfigure}
\end{figure}

\paragraph{Impact of $\lambda$}

In Equation~\eqref{eq:lambdac5}, we introduced the hyperparameter $\lambda \in \mathbb{R}^+$ to tune the relative importance of node proximity in the embedding space w.r.t. mass attraction, leading to the reconstruction scheme $\hat{A}_{ij} = \sigma(\tilde{m}_j - \lambda \log \|z_i - z_j\|^2_2)$. In Figure~\ref{c5_parameterfigure}, we show the impact of $\lambda$ on mean AUC scores for the Gravity-Inspired VGAE model and for all three graphs. For Cora and Citeseer, on Task~1 and Task~3, $\lambda = 1$ is an optimal choice, consistently with Newton's formula. However, for Google-Medium, on Task~1 and Task~3, we obtained better performances for higher values of $\lambda$, notably for $\lambda = 10$ that we used in our experiments. Increasing $\lambda$ reinforces the relative importance of the node pair's (symmetric) distance, measured by $\log \|z_i - z_j\|^2_2$, in the decoder w.r.t. parameter $\tilde{m}_j$ capturing the global influence of a node on its neighbors and therefore asymmetries in links. Since the Google-Medium graph is denser than Cora and Citesser, and has a higher proportion of symmetric relations (see Section~\ref{c5s542}), putting the emphasis on node proximity appears as a relevant strategy.

On a contrary, on Task~2 we achieved optimal performances by setting $\lambda = 0.05$, for all three graphs. As $\lambda < 1$, we improved scores by assigning more relative importance to the mass parameter $\tilde{m}_j$. Such a result is not surprising as, for \textit{biased negative samples link prediction} task, capturing the directionality is more crucial than the proximity, as nodes pairs from test sets are all included in both directions. As illustrated in Figure~\ref{c5_parameterfigure}(b), increasing $\lambda$ significantly deteriorates performances.

\paragraph{Extensions} 
Throughout these experiences, we focused on featureless graphs, to fairly compete with the HOPE, APP, and node2vec methods (see Section~\ref{c2s22}). However, as explained in Section~\ref{c5s532}, our models can easily leverage node features $X$, in addition to the graph structure summarized in $A$. Moreover, we recall that the Gravity-Inspired GAE and VGAE models are not limited to GCN encoders, and can be generalized to any alternative GNN models. However, so far, we assumed a fixed graph structure. Future research on directed link prediction in \textit{dynamic} graphs~\cite{kazemi2020representation} could definitely improve our approach. Also, while we focused on directed link prediction, we neglected the community detection problem in this chapter. Community detection in directed graphs is a challenging problem~\cite{malliaros2013clustering}, which could deserve further attention in our future research studies.

\section{Conclusion}
\label{c5s55}

In this chapter, we extended GAEs and VGAEs to directed graphs. We drew inspiration from physics to introduce a new gravity-inspired decoder, that can effectively reconstruct asymmetric relations from node embedding spaces. We achieved competitive experimental results on three different directed link prediction tasks, for which standard GAE and VGAE models perform poorly. We also pinpointed several research directions that, in the future, should lead to the improvement of our approach.

Since the publication of this work in late 2019~\cite{salha2019-2}, various recent papers explicitly mentioned our Gravity-Inspired GAE and VGAE models. Several of them included these models into their own experiments, including \cite{hibshman2021joint,rennard2020graph,roy2021adversarial,yoo2021evaluation,zhang2020hyper}. In Chapter~\ref{chapter_8}, we will ourselves leverage these models once again, to address music recommendation problems.
Specifically, in this chapter, we will model the \textit{cold start similar artists ranking} problem~\cite{salha2021cold} as a link prediction task in a directed and attributed
graph, connecting music artists to their top-$k$ most similar neighbors and incorporating side musical information as node features. We will show how the gravity-inspired decoder can be used to automatically rank the top-$k$ most similar neighbors of new artists. Such an application will also emphasize how Gravity-Inspired GAE/VGAE models, when equipped with node features, can be used in \textit{inductive} settings that involve generalizing representations to new unseen nodes after training.



\chapter[Simplifying Graph Autoencoders with One-Hop Linear Models]{Simplifying Graph Autoencoders with~One-Hop~Linear~Models}\label{chapter_6}
\chaptermark{Simplifying Graph Autoencoders with One-Hop Linear Models}

\textit{This chapter presents research conducted with Romain Hennequin and Michalis Vazirgiannis, and published in the proceedings of the 2020 European Conference on Machine Learning and Principles and Practice of Knowledge Discovery in Databases (ECML-PKDD 2020)~\cite{salha2020simple}. A~preliminary version of this work has also been presented at the  ``Graph Representation Learning'' workshop of the 33\up{rd} Conference on Neural Information Processing Systems (NeurIPS 2019)~\cite{salha2019keep}.}

\section{Introduction}
\label{c6s61}
 Despite the prevalent use of multi-layer GCN encoders in the recent literature (see Section~\ref{c6s62}), at the time of this work the relevance of this architecture choice had never been thoroughly studied nor challenged. The actual benefit of incorporating multi-layer GCNs, or even more complex GNNs, in GAE and VGAE models remained unclear. This is an important question, as simpler encoding strategies are easier to understand, to train, to deploy in production, and to debug, and might therefore be preferred for real-world industrial applications.

In this chapter, we propose to tackle this important aspect, showing that GCN-based GAE and VGAE models are often unnecessarily complex for numerous applications. Our work falls into a family of recent efforts questioning the systematic use of complex deep learning methods without a clear comparison to less fancy but simpler baselines \cite{recsys,lin2019neural,shchur2018pitfalls}. 

More precisely, in this chapter, we introduce simpler versions of the GAE and VGAE models, referred to as \textit{Linear GAE} and \textit{Linear VGAE}. We propose to replace multi-layer GCN encoders with linear models w.r.t. the direct neighborhood (one-hop) adjacency matrix of the graph, involving a unique weight matrix to tune, fewer operations, and no activation function.

Then, through an extensive empirical analysis on 17 real-world graphs with various sizes and characteristics, we show that these simplified models consistently reach competitive performances w.r.t. GCN-based GAE and VGAE models on link prediction and community detection tasks. We identify the settings where simple linear encoders appear as an effective alternative to GCNs, and as a first relevant baseline to implement before considering more complex models. In this chapter, we also question the relevance of the current benchmark datasets (Cora, Citeseer, Pubmed) commonly used in the literature to evaluate GAE and VGAE models.

This chapter is organized as follows. In Section~\ref{c6s62} we introduce our proposed Linear GAE and VGAE models, and mention some related work aiming to simplify GCNs. In Section~\ref{c6s63}, we present our experimental results, and provide discussions on the simplification and the evaluation of GAE and VGAE models. We conclude in Section~\ref{c6s64}.

\section{Simplifying Graph Autoencoders}
\label{c6s62}

We adopt a consistent notation with respect to previous chapters but, for the sake of simplicity, we assume in this section that the graph $\mathcal{G}$ is \textit{undirected}. $\tilde{A}$ therefore denotes the symmetric normalization of $A$, consistently with Definition~\ref{def:norm_c2} from Chapter~\ref{chapter_2}.

Nonetheless, our models could be straightforwardly extended to \textit{directed} graphs by replacing $\tilde{A}$ by the out-degree normalization $\tilde{A}_{\text{out}}$, as in Chapter~\ref{chapter_5}, in all equations of this section. Our analysis from Section~\ref{c6s63} will include experiments related to such an extension, in order to simplify as well our Gravity-Inspired GAE and VGAE models from Chapter~\ref{chapter_5}.

\subsection{One-Hop Linear Encoders}
\label{c6s621}

To this day, multi-layer GCNs remain the most popular encoders for GAE and VGAE models building upon the seminal work of Kipf~and~Welling~\cite{kipf2016-2}, including (but not limited to) the recent research of \cite{choong2018learning,do2019matrix,grover2019graphite,semiimplicit2019,pei2021generalization,kipf2016-2,pan2018arga,rennard2020graph,aaai20,huang2019rwr}. We ourselves mainly considered  multi-layer GCN encoders in our previous experiments from Chapters~\ref{chapter_3},~\ref{chapter_4},~and~\ref{chapter_5}, although our contributions are not restricted to this architecture choice.

We recall that, in a multi-layer GCN~\cite{kipf2016-1} with $L \geq 2$ layers, with an input layer $H^{(0)} = I_n$ (in the absence of node features) or $H^{(0)} = X$, and with an output layer $H^{(L)}$ (with $H^{(L)} = Z$ for a GAE, and $H^{(L)} = \mu$ or $\log \sigma$ for a VGAE), embedding vectors are computed as follows:
\begin{equation} 
\begin{cases}
H^{(0)} = X \text{ (or $I_n$)} \\
H^{(l)} = \text{ReLU} (\tilde{A} H^{(l-1)} W^{(l-1)}), \hspace{5pt} \text{for } l \in \{1,...,L-1\} \\
H^{(L)} = \tilde{A} H^{(L-1)} W^{(L-1)},
\end{cases}
\label{eq:gcnc6}
\end{equation}
using the notation from Definition~\ref{def:gcn}. In this chapter, we consider a simpler linear model w.r.t. the normalized one-hop (i.e., direct neighborhood) adjacency matrix of the graph. More precisely, the term \textit{linear encoder} will refer to the following function.

\begin{definition}
A \textit{linear encoder} is a function taking as input an adjacency matrix $A$, potentially equipped with a node feature matrix $X \in \mathbb{R}^{n \times f}$, and returning an output matrix $Z \in \mathbb{R}^{n \times d}$ computed as follows:
\begin{equation}
Z = \text{Linear}(A, X) = \tilde{A} X W,  
\label{yesfeat}
\end{equation}
for some trainable weight matrix $W \in \mathbb{R}^{f \times d}$. In particular, in the absence of node features (i.e., $X = I_n$), we have:
\begin{equation}
Z = \text{Linear}(A,I_n) = \tilde{A}  W, 
\label{nofeat}
\end{equation}
and the weight matrix $W$ is then of dimension $n \times d$.
\label{def:linearencoder}
\end{definition}

In Sections~\ref{c6s622} and \ref{c6s623}, we incorporate this straightforward one-hop linear encoder in the GAE and VGAE frameworks, and discuss the implications of such a modification.

\subsection{Linear GAE}
\label{c6s622}

In this chapter, we propose to replace the multi-layer GCN encoder of standard GAE and VGAE models with the linear encoder from Definition~\ref{def:linearencoder}. Firstly, in the GAE~framework,~we~set:
\begin{equation}
Z = \text{Linear}(A,X) = \tilde{A}  X W, \text{ then } \hat{A} = \sigma(ZZ^T).
\end{equation}
In the absence of node features $X$, the model is simplified as follows:
\begin{equation}
Z = \tilde{A}  W, \text{ then } \hat{A} = \sigma(ZZ^T).
\end{equation}
We refer to this model as \textit{Linear Graph Autoencoder (Linear GAE)}. In the absence of node features, embedding vectors are obtained by multiplying the $n \times n$ normalized adjacency matrix $\tilde{A}$ by a \textit{unique} $n \times d$ weight matrix $W$. In the presence of node features, the linear encoder also simply consists in multiplying $\tilde{A}  X$ with a unique $f \times d$ weight matrix $W$.

We tune this matrix $W$ in a similar fashion w.r.t. standard GAE models considered in the previous chapters, i.e., we iteratively minimize a weighted cross-entropy loss capturing the quality of the reconstruction $\hat{A}$ w.r.t. the original matrix $A$, by gradient descent~\cite{kipf2016-2}.

\begin{figure}[t]
    \centering
\resizebox{1.0\textwidth}{!}{
    \tikzstyle{block} = [draw, fill=ImperialColor, rectangle, 
    minimum height=4em, minimum width=8em]
\tikzstyle{sum} = [draw, fill=ImperialColor, circle, node distance=1cm]
\tikzstyle{input} = [coordinate]
\tikzstyle{output} = [coordinate]
\tikzstyle{pinstyle} = [pin edge={to-,thin,black}]
\begin{tikzpicture}[auto, node distance=8cm,>=latex']
    \node  (A) {$A, X$};
    \node [block, right of=A, node distance=3cm, align=center] (encoder){\textcolor{white}{Linear} \\ \textcolor{white}{encoder}};
    \node [node distance=3cm, right of = encoder] (Z) {$Z$};
    \node [block, right of=Z, node distance=3cm, align=center] (decoder) {\textcolor{white}{Inner product} \\ \textcolor{white}{decoder}};
    \node [node distance=3cm, right of = decoder] (Arec) {$\hat{A}$};
    \draw [->] (A) -- (encoder);
    \draw [->] (encoder) -- (Z);
    \draw [->] (Z) -- (decoder);
    \draw [->] (decoder) -- (Arec);
\end{tikzpicture}}
\caption[Schematic representation of a Linear GAE]{Schematic representation of a Linear GAE model.}
    \label{fig:c6_gae}
\end{figure}
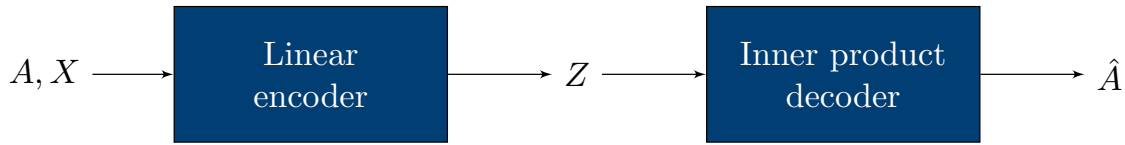

This encoder is a straightforward linear mapping. It can be interpreted as the simplest possible GCN (with a single layer). Each element of the $z_i$ embedding vector is a weighted average from node $i$'s direct one-hop connections. Contrary to multi-layer GCN encoders (with $L\geq 2$ layers):
\begin{itemize}
    \item we \textit{ignore the higher-order information} from the $k$-hop neighborhood with $k>1$;
    \item also, we do not include any \textit{non-linear activation function}. 
\end{itemize}In Section~\ref{c6s63}, we will highlight the very limited impact of these two simplifications on empirical performances. The above Figure~\ref{fig:c6_gae} provides an illustration of a Linear GAE model.

Our proposed linear encoder runs in a linear time w.r.t. the number of edges $m$ of the graph using a sparse representation for $\tilde{A}$, and involves fewer matrix operations than a multi-layer GCN. It includes $n d$ parameters in the absence of node features (resp. $f  d$ parameters with node features), i.e., fewer than the $n d + (L - 1) d^2$ parameters (resp. $f d + (L - 1) d^2$ parameters) required by a $L$-layer GCN where all layers are of dimension $d$.

As is the case for standard GAEs, the inner product decoder has a $O(dn^2)$ complexity, as it involves the multiplication of the two dense matrices $Z$ and $Z^T$. Such a complexity can be significantly reduced by leveraging the scalable methods we proposed in Chapter~\ref{chapter_3}~and~\ref{chapter_4}.
Our evaluation from Section~\ref{c6s63} will include experiments on large graphs requiring the~use~of~such~methods. 

To emphasize that the Linear GAE model is not restricted to inner product decoders, the evaluation from Section~\ref{c6s63} will also consider two alternative decoders to replace inner products~\cite{grover2019graphite,salha2019-2}, including our proposed gravity-inspired decoder from Chapter~\ref{chapter_5}.

\subsection{Linear VGAE}
\label{c6s623}

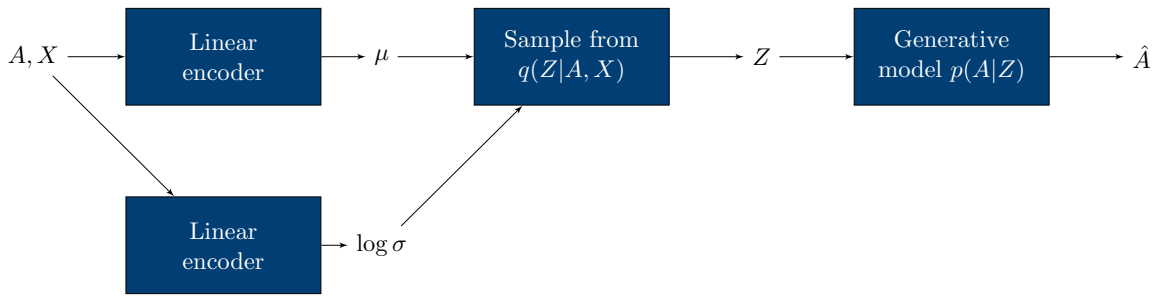
\begin{figure}[t]
    \centering
\resizebox{1.0\textwidth}{!}{
    \tikzstyle{block} = [draw, fill=ImperialColor, rectangle, 
    minimum height=4em, minimum width=8em]
\tikzstyle{sum} = [draw, fill=ImperialColor, circle, node distance=1cm]
\tikzstyle{input} = [coordinate]
\tikzstyle{output} = [coordinate]
\tikzstyle{pinstyle} = [pin edge={to-,thin,black}]
\begin{tikzpicture}[auto, node distance=2cm,>=latex']
    \node  (A) {$A, X$};
    \node [block, right of=A, node distance=3cm, align=center] (encoder) {\textcolor{white}{Linear} \\ \textcolor{white}{encoder}};
    \node [block, right of=A, below of=A, node distance=3cm, align=center] (encoder2) {\textcolor{white}{Linear} \\ \textcolor{white}{encoder}};
    \node [node distance=2.5cm, right of = encoder] (mu) {$\mu$};
    \node [node distance=3cm, below of = mu] (sigma) {$\log \sigma$};
    \node [block, right of=mu, node distance=3cm, align=center] (sampler) {\textcolor{white}{Sample from}\\ \textcolor{white}{$q(Z|A, X)$}};
    \node [node distance=3cm, right of = sampler] (Z) {$Z$};
    \node [block, node distance=3cm, right of = Z, align=center] (decoder) {\textcolor{white}{Generative} \\ \textcolor{white}{model} \textcolor{white}{$p(A|Z)$}};
    \node [node distance=3cm, right of = decoder] (Ahat) {$\hat{A}$};

    \draw [->] (A) -- (encoder);
    \draw [->] (A) -- (encoder2);
    \draw [->] (encoder) -- (mu);
    \draw [->] (encoder2) -- (sigma);
    \draw [->] (mu) -- (sampler);
    \draw [->] (sigma) -- (sampler);
    \draw [->] (sampler) -- (Z);
    \draw [->] (Z) -- (decoder);
    \draw [->] (decoder) -- (Ahat);
\end{tikzpicture}}
\caption[Schematic representation of a Linear VGAE]{Schematic representation of a Linear VGAE model.}
    \label{fig:c6_vgae}
\end{figure}

We adopt a similar approach to replace the two multi-layer GCN encoders of standard VGAE models~\cite{kipf2016-2} by:
\begin{equation}
\mu = \text{Linear}_{\mu}(A,X) = \tilde{A} X W_{\mu} \text{ and } \log \sigma = \text{Linear}_{\sigma}(A,X) = \tilde{A} X W_{\sigma},
\label{linvgaec6equation}
\end{equation}
with $f \times d$ weight matrices $W_{\mu}$ and $W_{\sigma}$.
In the absence of node features $X$, the encoding step is simplified as follows:
\begin{equation}
\mu = \tilde{A} W_{\mu} \text{ and } \log \sigma = \tilde{A} W_{\sigma},
\end{equation}
with $n \times d$ weight matrices $W_{\mu}$ and $W_{\sigma}$. Then:
\begin{equation}
\forall i \in \mathcal{V}, z_i \sim \mathcal{N}(\mu_i, \text{diag}(\sigma_i^2)),
\end{equation}
with a similar decoder a.k.a. generative process w.r.t. standard VGAEs (see Section~\ref{c2s242} from Chapter~\ref{chapter_2}). We refer to this model as \textit{Linear Variational Graph Autoencoder (Linear VGAE)}. During the learning phase, as is the case for standard VGAEs, we iteratively optimize the ELBO variational lower bound from Equation~\eqref{elbo}, w.r.t. $W_{\mu}$ and $W_{\sigma}$, by gradient ascent~\cite{kipf2016-2}. Figure~\ref{fig:c6_vgae} provides a schematic representation of the proposed Linear VGAE model.

\subsection{Related Work}
\label{c6s624}

Our work falls into a family of research efforts aiming to challenge and question the prevalent use of complex deep learning methods without a clear comparison to simpler baselines \cite{recsys,lin2019neural,shchur2018pitfalls}. In particular, in a concurrent study, Wu~et~al.~\cite{wu2019simplifying} also recently proposed to simplify GCN models, notably by removing non-linearities between layers and collapsing some
weight matrices during training. Their simplified model, referred to as Simple Graph Convolution (SGC), empirically rivals standard GCNs on several large-scale classification tasks~\cite{wu2019simplifying}. While our work also focuses on simplifications of GCNs, we argue that the two research studies actually tackle different and complementary problems:
\begin{itemize}
    \item Wu~et~al.~\cite{wu2019simplifying} focus on supervised and semi-supervised settings. They consider the GCN as the model itself, optimized to classify node-level labels. On the contrary, we consider two \textit{unsupervised} settings, in which GCNs are only a building part (the encoder) of a larger framework (the GAE or the VGAE), and where we optimize reconstruction losses/objectives from GCN-based embedding vectors (for GAE) or from vectors drawn from distributions learned through two GCNs (for VGAE);
    \item besides, our encoders only capture one-hop interactions, i.e., nodes only aggregate information from their direct neighbors during message passing. On the contrary, Wu~et~al.~\cite{wu2019simplifying} still rely on a stacked layers design that, although simplified, permits learning from higher-order interactions. Contrary to us, considering such relations is crucial in their model for good performances~\cite{wu2019simplifying}. We will show in Section~\ref{c6s63} that, in our setting, it would mainly increase running times while bringing few to no improvement.
\end{itemize}

\section{Experimental Analysis}
\label{c6s63}

In this section, we propose an in-depth empirical evaluation of our Linear GAE~and~VGAE~models. Our Python/TensorFlow implementation of these models is publicly available on GitHub\footnote{\href{https://github.com/deezer/linear_graph_autoencoders}{https://github.com/deezer/linear\_graph\_autoencoders}}.

\subsection{Experimental Setting}
\label{c6s631}

\paragraph{Tasks}
We consider two tasks. Firstly, we focus on \textit{link prediction}, similarly to Chapters~\ref{chapter_3}~to~\ref{chapter_5}. As in these chapters, we train models on incomplete versions of graphs where $15\%$ of edges were randomly removed, and used for validation ($5\%$) and test ($10\%$), together with the same number of randomly sampled pairs of unconnected nodes. We evaluate the ability of our models to classify edges from non-edges in these sets using, as in previous chapters, the mean AUC and AP scores averaged over 100 runs (resp. 10 runs for datasets with $n >$ 100 000). Models were trained on 100 (resp. 10) different random train/test splits.

As a second task, we perform \textit{community detection}. When datasets include node-level ground truth communities. Similarly to experiments from Chapters~\ref{chapter_3}~and~\ref{chapter_4}, we train models on complete graphs, then run $k$-means algorithms in node embedding spaces. We compare the resulting clusters to ground truth communities via mean AMI scores.

\paragraph{Datasets} We provide experiments on 17 publicly available real-world graphs. For each graph, Tables~\ref{t1}~and~\ref{t2} report the number of nodes $n$, the number of edges $m$, and the dimension $f$ of node feature vectors, when such vectors are available:
\begin{itemize}
\item we first consider the Cora, Citeseer, and Pubmed citation graphs\footnote{\label{linqsc6}\href{https://linqs.soe.ucsc.edu/data}{https://linqs.soe.ucsc.edu/data}}~\cite{sen2008collective}, with and without node features corresponding to $f$-dimensional bag-of-words vectors. These three graphs were used in the original experiments of Kipf~and~Welling~\cite{kipf2016-2}, and subsequently in the wide majority of recent works on GAE and VGAE models including (but not limited to) \cite{grover2019graphite,semiimplicit2019,pan2018arga,park2019symmetric,salha2019-1,aaai20,tran2018multi,huang2019rwr,wang2017mgae}, becoming the \textit{de facto} benchmark datasets for evaluating GAE and VGAE. We ourselves considered these citation networks in our own experiments from Chapters~\ref{chapter_3}~to~\ref{chapter_4}. Therefore, we saw value in comparing Linear and GCN-based models on these graphs;

\item we also report results on 14 alternative graphs, including some of those from the previous chapters. We consider four other citations networks:
Patent\footnote{\label{snap}\href{https://snap.stanford.edu/data/index.html}{https://snap.stanford.edu/data/index.html}} from Chapters~\ref{chapter_3}~and~\ref{chapter_4}, DBLP\textsuperscript{\ref{konect}}, Arxiv-HepTh\textsuperscript{\ref{snap}}, and a larger version of Cora\textsuperscript{\ref{konect}}, that we denote Cora-larger. We add the WebKD\textsuperscript{\ref{linqsc6}}, Blogs\textsuperscript{\ref{konect}} and Stanford\footnote{\label{konect}\href{http://konect.cc/networks/}{https://konect.cc/networks/}} web graphs, where hyperlinks connect web pages, as well as two Google web graphs: the medium-size one\textsuperscript{\ref{konect}} from Chapter~\ref{chapter_5}, denoted Google-Medium, and the larger one\textsuperscript{\ref{snap}} from Chapters~\ref{chapter_3}~and~\ref{chapter_4}, denoted Google. We complete the list with two social networks (Hamsterster\textsuperscript{\ref{konect}} and LiveMocha\textsuperscript{\ref{konect}}), the Flickr\textsuperscript{\ref{konect}} image graph (nodes represent images, connected when sharing metadata), the Proteins\textsuperscript{\ref{konect}} network of proteins interactions and the Amazon\textsuperscript{\ref{konect}} products co-purchase network. Therefore, in these experiments, we consider a wide variety of real-world graphs of various origins, characteristics and sizes (from 877 to 2.7 million nodes, from 1 608 to 13.9 million edges).
\end{itemize}

\paragraph{Models} In all experiments, we compare the Linear GAE and VGAE models to 2-layer and 3-layer GCN-based graph GAE and VGAE models. We do not report performances of deeper models, due to a significant deterioration of all scores. For a comparison to other methods and notably to non-GAE/VGAE methods, which is out of the scope of this study, we refer to experiments from the previous chapters, as well as to experiments from the next Chapter~\ref{chapter_7} for a stronger emphasis on community detection.

All models were trained for $200$ epochs (resp. $300$ epochs) for graphs with $n <$ 100 000 (resp. $n \geq$ 100  000). We thoroughly checked the convergence of all models, in terms of mean AUC performances on validation sets, for these epochs numbers. As in Chapters~\ref{chapter_3}~and~\ref{chapter_4}, we ignored edges directions when initial graphs were directed, as we focus on symmetric inner product decoders in most of this chapter\footnote{We will nonetheless relax this restriction and consider directed edges during training and decoding later in Section~\ref{c6s632}, when we will extend our linear encoders to Gravity-Inspired GAE and VGAE models.}. For Cora, Citeseer, and Pubmed, we set identical hyperparameters w.r.t. Kipf~and~Welling~\cite{kipf2016-2} to reproduce their results, i.e., we had $d = 16$, 32-dimensional hidden layer(s) for GCNs, and we used the Adam optimizer~\cite{kingma2014adam} with a learning rate of 0.01.

For other datasets, we tuned hyperparameters by performing a grid search on the validation set. We adopted a learning rate of 0.1 for Arxiv-HepTh, Patent, and Stanford; of 0.05 for Amazon, Flickr, LiveMocha, and Google; of 0.01 for Blogs, Cora-larger, DBLP, Google-Medium, Hamsterster, and Proteins (GAE models); of 0.005 for WebKD (except Linear GAE and VGAE where we used 0.001 and 0.01) and Proteins (VGAE models). We set $d =16$ (but we reached similar conclusions with $d =32$ and $64$), with 32-dimensional hidden layer(s) and without dropout.

Lastly, due to the prohibitive cost of decoding the entire matrix $\hat{A}$ for large graphs (i.e., Amazon, Flickr, Google, LiveMocha, Patent, and Stanford that all verify with $n >$ 100 000), we adopted a stochastic sampling strategy for these graphs. At each training iteration, we estimated losses by only reconstructing a subgraph of 10 000 nodes from the original graph. These 10~000 nodes were randomly picked during training at each iteration. This strategy corresponds to a special case of the FastGAE method from Chapter~\ref{chapter_4}, with $n_{(S)}=$ 10 000 and with \textit{uniform} sampling\footnote{Experiments from this chapter and from the papers associated with this work~\cite{salha2019keep,salha2020simple} were actually done \textit{before} our study of FastGAE from Chapter~\ref{chapter_4}. We acknowledge that using FastGAE with a non-uniform sampling, e.g., with degree-based sampling as in Section~\ref{c4s422}, should improve results w.r.t. a uniform sampling.}.

\subsection{Results and Discussion}
\label{c6s632}

\begin{table}[t]
  \centering
\caption[Link prediction on common benchmark graphs using Linear GAE and VGAE]{Link prediction on the Cora, Citeseer, and Pubmed common benchmark datasets, with Linear GAE and VGAE models, and with their multi-layer GCN-based counterparts. We report details on hyperparameters for each model and dataset in Section~\ref{c6s631}. Cells are colored in \textcolor{ImperialColor}{blue} when Linear GAE/VGAE models are reaching competitive results w.r.t. standard GCN-based models, i.e., results that are at least as good as GCN-based models $\pm$ 1 standard deviation. Note: as we aim to evaluate whether linear models are \textit{as good} as others, and not necessarily \textit{better}, we do not report \textbf{bold} numbers in this table contrary to tables from previous chapters.}
\label{t1}
\begin{center}
 \resizebox{1.0\textwidth}{!}{
\begin{tabular}{c|cc|cc|cc}
    \toprule
    & \multicolumn{2}{c}{\textbf{Cora}} & \multicolumn{2}{c}{\textbf{Citeseer}} & \multicolumn{2}{c}{\textbf{Pubmed}} \\
     \textbf{Model} & \multicolumn{2}{c}{\footnotesize (n = 2 708, m = 5 429)} & \multicolumn{2}{c}{\footnotesize (n = 3 327, m = 4 732)} & \multicolumn{2}{c}{\footnotesize (n = 19 717, m = 44 338)} \\
     \cmidrule{2-7}
      & \footnotesize \textbf{AUC (in \%)} & \footnotesize \textbf{AP (in \%)} & \footnotesize \textbf{AUC (in \%)} & \footnotesize \textbf{AP (in \%)} & \footnotesize \textbf{AUC (in \%)} & \footnotesize \textbf{AP (in \%)}\\
    \midrule
    \midrule
     Linear GAE (ours) & \cellcolor{ImperialColor!15} 83.19 $\pm$ 1.13 &  \cellcolor{ImperialColor!15} 87.57 $\pm$ 0.95 & \cellcolor{ImperialColor!15} 77.06 $\pm$ 1.81 & \cellcolor{ImperialColor!15} 83.05 $\pm$ 1.25 & 81.85 $\pm$ 0.32 & \cellcolor{ImperialColor!15} 87.54 $\pm$ 0.28 \\
     2-layer GCN-based GAE & \cellcolor{ImperialColor!15} 84.79 $\pm$ 1.10 & \cellcolor{ImperialColor!15} 88.45 $\pm$ 0.82 & \cellcolor{ImperialColor!15} 78.25 $\pm$ 1.69 & \cellcolor{ImperialColor!15} 83.79 $\pm$ 1.24 & 82.51 $\pm$ 0.64 & \cellcolor{ImperialColor!15} 87.42 $\pm$ 0.38 \\
     3-layer GCN-based GAE & \cellcolor{ImperialColor!15} 84.61 $\pm$ 1.22 & \cellcolor{ImperialColor!15} 87.65 $\pm$ 1.11 & \cellcolor{ImperialColor!15} 78.62 $\pm$ 1.74 & \cellcolor{ImperialColor!15} 82.81 $\pm$ 1.43 & 83.37 $\pm$ 0.98 & \cellcolor{ImperialColor!15} 87.62 $\pm$ 0.68 \\
    \midrule
    Linear VGAE (ours) & \cellcolor{ImperialColor!15} 84.70 $\pm$ 1.24 & \cellcolor{ImperialColor!15} 88.24 $\pm$ 1.02 & \cellcolor{ImperialColor!15} 78.87 $\pm$ 1.34 & \cellcolor{ImperialColor!15} 83.34 $\pm$ 0.99 & \cellcolor{ImperialColor!15} 84.03 $\pm$ 0.28 & \cellcolor{ImperialColor!15} 87.98 $\pm$ 0.25 \\
    2-layer GCN-based VGAE & \cellcolor{ImperialColor!15} 84.19 $\pm$ 1.07 & \cellcolor{ImperialColor!15} 87.68 $\pm$ 0.93 & \cellcolor{ImperialColor!15} 78.08 $\pm$ 1.40 & \cellcolor{ImperialColor!15} 83.31 $\pm$ 1.31 & \cellcolor{ImperialColor!15} 82.63 $\pm$ 0.45 & \cellcolor{ImperialColor!15} 87.45 $\pm$ 0.34 \\
     3-layer GCN-based VGAE & \cellcolor{ImperialColor!15} 84.48 $\pm$ 1.42 & \cellcolor{ImperialColor!15} 87.61 $\pm$ 1.08 & \cellcolor{ImperialColor!15} 79.27 $\pm$ 1.78 & \cellcolor{ImperialColor!15} 83.73 $\pm$ 1.13 & \cellcolor{ImperialColor!15} 84.07 $\pm$ 0.47 & \cellcolor{ImperialColor!15} 88.18 $\pm$ 0.31 \\
    \midrule
    \midrule
       & \multicolumn{2}{c}{\textbf{Cora, with features}} & \multicolumn{2}{c}{\textbf{Citeseer, with features}} & \multicolumn{2}{c}{\textbf{Pubmed, with features}} \\
            \textbf{Model}  & \multicolumn{2}{c}{\footnotesize (n = 2 708, m = 5 429,} & \multicolumn{2}{c}{\footnotesize (n = 3 327, m = 4 732,} & \multicolumn{2}{c}{\footnotesize (n = 19 717, m = 44 338,} \\
            & \multicolumn{2}{c}{\footnotesize f = 1 433)} & \multicolumn{2}{c}{\footnotesize f = 3 703)} & \multicolumn{2}{c}{\footnotesize f = 500)} \\
     \cmidrule{2-7}
      & \footnotesize \textbf{AUC (in \%)} & \footnotesize \textbf{AP (in \%)} & \footnotesize \textbf{AUC (in \%)} & \footnotesize \textbf{AP (in \%)} & \footnotesize \textbf{AUC (in \%)} & \footnotesize\textbf{AP (in \%)}\\
    \midrule
    \midrule
    Linear GAE (ours) & \cellcolor{ImperialColor!15} 92.05 $\pm$ 0.93 & \cellcolor{ImperialColor!15} 93.32 $\pm$ 0.86 & \cellcolor{ImperialColor!15} 91.50 $\pm$ 1.17 & \cellcolor{ImperialColor!15} 92.99 $\pm$ 0.97 & \cellcolor{ImperialColor!15} 95.88 $\pm$ 0.20 & \cellcolor{ImperialColor!15} 95.89 $\pm$ 0.17 \\
     2-layer GCN-based GAE & \cellcolor{ImperialColor!15} 91.27 $\pm$ 0.78 & \cellcolor{ImperialColor!15} 92.47 $\pm$ 0.71 & \cellcolor{ImperialColor!15} 89.76 $\pm$ 1.39 & \cellcolor{ImperialColor!15} 90.32 $\pm$ 1.62 & \cellcolor{ImperialColor!15} 96.28 $\pm$ 0.36 & \cellcolor{ImperialColor!15} 96.29 $\pm$ 0.25 \\
     3-layer GCN-based GAE & \cellcolor{ImperialColor!15} 89.16 $\pm$ 1.18 & \cellcolor{ImperialColor!15} 90.98 $\pm$ 1.01 & \cellcolor{ImperialColor!15} 87.31 $\pm$ 1.74 & \cellcolor{ImperialColor!15} 89.60 $\pm$ 1.52 & \cellcolor{ImperialColor!15} 94.82 $\pm$ 0.41 & \cellcolor{ImperialColor!15} 95.42 $\pm$ 0.26 \\
    \midrule
    Linear VGAE (ours)& \cellcolor{ImperialColor!15} 92.55 $\pm$ 0.97 & \cellcolor{ImperialColor!15} 93.68 $\pm$ 0.68& \cellcolor{ImperialColor!15} 91.60 $\pm$ 0.90 & \cellcolor{ImperialColor!15} 93.08 $\pm$ 0.77 & \cellcolor{ImperialColor!15} 95.91 $\pm$ 0.13 & \cellcolor{ImperialColor!15} 95.80 $\pm$ 0.17 \\
     2-layer GCN-based VGAE& \cellcolor{ImperialColor!15} 91.64 $\pm$ 0.92 & \cellcolor{ImperialColor!15} 92.66 $\pm$ 0.91 & \cellcolor{ImperialColor!15} 90.72 $\pm$ 1.01 & \cellcolor{ImperialColor!15} 92.05 $\pm$ 0.97 & \cellcolor{ImperialColor!15} 94.66 $\pm$ 0.51 & \cellcolor{ImperialColor!15} 94.84 $\pm$ 0.42 \\
     3-layer GCN-based VGAE & \cellcolor{ImperialColor!15} 90.53 $\pm$ 0.94 & \cellcolor{ImperialColor!15} 91.71 $\pm$ 0.88 & \cellcolor{ImperialColor!15} 88.63 $\pm$ 0.95 & \cellcolor{ImperialColor!15} 90.20 $\pm$ 0.81 & \cellcolor{ImperialColor!15} 92.78 $\pm$ 1.02 & \cellcolor{ImperialColor!15} 93.33 $\pm$ 0.91 \\
    \bottomrule
  \end{tabular} }
  \end{center}

\end{table}

\paragraph{Cora, Citeseer, and Pubmed benchmarks} Table~\ref{t1} reports link prediction results for Cora, Citeseer, and Pubmed. For standard 2-layer GAEs and VGAEs, we reproduce similar performances w.r.t. Kipf and Welling~\cite{kipf2016-2}. We show that Linear GAE and VGAE models consistently reach competitive performances w.r.t. 2-layer and 3-layer GCN-based models, i.e., they are at least as good ($\pm$ 1 standard deviation). In some settings, Linear GAE and VGAE models are even slightly better (e.g., $+1.25$ points in AUC for the Linear VGAE on Pubmed with features, w.r.t. a 2-layer GCN-based VGAE). These results emphasize the effectiveness of our proposed simple encoder on these datasets, where the empirical benefit of multi-layer GCNs is very limited. In Table~\ref{t3}, we consolidate our results by reaching similar conclusions on the community detection task. As explained in Chapter~\ref{chapter_3}, in these graphs nodes are documents clustered in respectively 6, 7, and 3 topic classes, acting as communities. In almost all settings, Linear GAE and VGAE models rival their multi-layer GCN-based counterparts (e.g., $+4.31$ AMI points for the Linear VGAE on Pubmed with features, w.r.t. 2-layer GCN-based VGAE).

\paragraph{Alternative graph datasets}
Table~\ref{t2} reports link prediction results for all other graphs. Linear GAE models are competitive in 13 cases out of 15, and sometimes even achieve better performances (e.g., $+1.72$ AUC points for Linear GAE on the largest dataset, Patent, w.r.t. 3-layer GCN-based GAE). Moreover, Linear VGAE models rival or outperform GCN-based models in 10 cases out of 15. Overall, Linear GAE and VGAE models also achieve very close results w.r.t. GCN-based models in all remaining datasets (e.g., on Google-Medium, with a mean AUC score of 96.02\% $\pm$~0.14 for Linear GAE, only slightly below the mean AUC score of 96.66\% $\pm$~0.24 of 2-layer GCN-based GAE). This confirms the empirical effectiveness of simple node encoding schemes, that appear as a suitable alternative to more complex multi-layer encoders for many real-world applications. Regarding community detection in Table~\ref{t3}, Linear GAE and VGAE models are competitive on the Cora-larger graph, in which nodes are documents clustered in 70 topic classes. However, 2-layer and 3-layer GCN-based models are significantly outperforming on the Blogs graph, where political blogs are classified as either left-leaning or right-leaning (e.g., $-23.42$ AMI points for Linear VGAE w.r.t. 2-layer GCN-based VGAE).

\paragraph{On running times and extensions to $k$-hop linear encoders}
While this work puts the emphasis on performance and not on training speed, we also note that Linear GAE and VGAE models are 10\% to 15\% faster than their GCN-based counterparts. For instance, on our NVIDIA GTX 1080 GPU, we report a 6.03 seconds (vs 6.73 seconds) mean running time for training our Linear VGAE (vs a 2-layer GCN-based VGAE) on the featureless Citeseer dataset, and roughly 800 seconds (vs 900+ seconds) on the Patent dataset, using our sampling strategy from Section~\ref{c6s631}. This gain comes from the fewer parameters and matrix operations required by one-hop linear encoders and from the sparsity of the one-hop matrix $\tilde{A}$ in our graphs. 

Besides, while in this work we only learn embedding vectors from direct neighbors, variants of our models could capture higher-order links by considering polynomials of the matrix $A$. For instance, we could learn node embedding representations from one-hop and two-hop links by replacing $\tilde{A}$ by the normalized version of $A + \alpha A^2$ (with $\alpha > 0$), or simply $A^2$, in the linear encoders of Section~\ref{c6s62}. However, we observed few to no improvement for our graphs in some preliminary tests, consistently with our claim on the effectiveness of simple one-hop strategies. Such variants also tend to increase running times, as $A^2$ is usually denser than $A$.
\clearpage

\begin{table}[ht]
  \centering
\caption[Link prediction on alternative graphs using Linear GAE and VGAE]{Link prediction on alternative datasets, with Linear GAE and VGAE models, and with their multi-layer GCN-based counterparts. We report details on hyperparameters for each model and dataset in Section~\ref{c6s631}. Cells are colored in \textcolor{ImperialColor}{blue} when Linear GAE/VGAE models are reaching competitive results w.r.t. standard GCN-based models, i.e., results that are at least as good as GCN-based models $\pm$ 1 standard deviation. Note: as we aim to evaluate whether linear models are \textit{as good} as others, and not necessarily \textit{better}, we do not report \textbf{bold} numbers in this table contrary to tables from previous chapters.}
\label{t2}
 \resizebox{1.0\textwidth}{!}{
  \begin{tabular}{c|cc|cc|cc}
    \toprule
   & \multicolumn{2}{c}{\textbf{WebKD}} & \multicolumn{2}{c}{\textbf{WebKD, with features}} & \multicolumn{2}{c}{\textbf{Hamsterster}} \\
       \textbf{Model}  & \multicolumn{2}{c}{\footnotesize (n = 877, m = 1 608)} & \multicolumn{2}{c}{\footnotesize (n = 877, m = 1 608, f = 1 703)} & \multicolumn{2}{c}{\footnotesize (n = 1 858, m = 12 534)}  \\
     \cmidrule{2-7}
       & \footnotesize \textbf{AUC (in \%)} & \footnotesize \textbf{AP (in \%)} & \footnotesize \textbf{AUC (in \%)} & \footnotesize \textbf{AP (in \%)} & \footnotesize \textbf{AUC (in \%)} & \footnotesize \textbf{AP (in \%)} \\
    \midrule
    \midrule
    Linear GAE (ours) & \cellcolor{ImperialColor!15} 77.20 $\pm$ 2.35 & \cellcolor{ImperialColor!15} 83.55 $\pm$ 1.81 & \cellcolor{ImperialColor!15} 84.15 $\pm$ 1.64 & \cellcolor{ImperialColor!15} 87.01 $\pm$ 1.48 & \cellcolor{ImperialColor!15} 93.07 $\pm$ 0.67 & \cellcolor{ImperialColor!15} 94.20 $\pm$ 0.58 \\
     2-layer GCN-based GAE & \cellcolor{ImperialColor!15} 77.88 $\pm$ 2.57 & \cellcolor{ImperialColor!15} 84.12 $\pm$ 2.18 & \cellcolor{ImperialColor!15} 86.03 $\pm$ 3.97 & \cellcolor{ImperialColor!15} 87.97 $\pm$ 2.76 & \cellcolor{ImperialColor!15} 92.07 $\pm$ 0.63 & \cellcolor{ImperialColor!15} 93.01 $\pm$ 0.69 \\
     3-layer GCN-based GAE & \cellcolor{ImperialColor!15} 78.20 $\pm$ 3.69 & \cellcolor{ImperialColor!15} 83.13 $\pm$ 2.58 & \cellcolor{ImperialColor!15} 81.39 $\pm$ 3.93 & \cellcolor{ImperialColor!15} 85.34 $\pm$ 2.92 & \cellcolor{ImperialColor!15} 91.40 $\pm$ 0.79 & \cellcolor{ImperialColor!15} 92.22 $\pm$ 0.85 \\
    \midrule
    Linear VGAE (ours)& \cellcolor{ImperialColor!15} 83.50 $\pm$ 1.98 & \cellcolor{ImperialColor!15} 86.70 $\pm$ 1.53 & \cellcolor[gray]{1.0} 85.57 $\pm$ 2.18 & \cellcolor{ImperialColor!15} 88.08 $\pm$ 1.76 & \cellcolor{ImperialColor!15} 91.08 $\pm$ 0.70 & \cellcolor{ImperialColor!15} 91.85 $\pm$ 0.64 \\
     2-layer GCN-based VGAE & \cellcolor{ImperialColor!15} 82.31 $\pm$ 2.55 & \cellcolor{ImperialColor!15} 86.15 $\pm$ 2.03 & \cellcolor[gray]{1.0} 87.87 $\pm$ 2.48 & \cellcolor{ImperialColor!15} 88.97 $\pm$ 2.17 & \cellcolor{ImperialColor!15} 91.62 $\pm$ 0.60 & \cellcolor{ImperialColor!15} 92.43 $\pm$ 0.64 \\
     3-layer GCN-based VGAE& \cellcolor{ImperialColor!15} 82.17 $\pm$ 2.70 & \cellcolor{ImperialColor!15} 85.35 $\pm$ 2.25 & \cellcolor[gray]{1.0} 89.69 $\pm$ 1.80 & \cellcolor{ImperialColor!15} 89.90 $\pm$ 1.58 & \cellcolor{ImperialColor!15} 91.06 $\pm$ 0.71 & \cellcolor{ImperialColor!15} 91.85 $\pm$ 0.77 \\
    \midrule
    \midrule
    & \multicolumn{2}{c}{\textbf{DBLP}} & \multicolumn{2}{c}{\textbf{Cora-larger}} & \multicolumn{2}{c}{\textbf{Arxiv-HepTh}}  \\
        \textbf{Model} & \multicolumn{2}{c}{\footnotesize (n = 12 591, m = 49 743)} & \multicolumn{2}{c}{\footnotesize (n = 23 166, m = 91 500)} & \multicolumn{2}{c}{\footnotesize (n = 27 770, m = 352 807)}\\
     \cmidrule{2-7}
      & \footnotesize \textbf{AUC (in \%)} & \footnotesize \textbf{AP (in \%)} & \footnotesize \textbf{AUC (in \%)} & \footnotesize \textbf{AP (in \%)} & \footnotesize \textbf{AUC (in \%)} & \footnotesize \textbf{AP (in \%)} \\
    \midrule
    \midrule
    Linear GAE (ours) & \cellcolor{ImperialColor!15} 90.11 $\pm$ 0.40 & \cellcolor{ImperialColor!15} 93.15 $\pm$ 0.28 & \cellcolor{ImperialColor!15} 94.64 $\pm$ 0.08 & \cellcolor{ImperialColor!15} 95.96 $\pm$ 0.10 & \cellcolor{ImperialColor!15} 98.34 $\pm$ 0.03 & \cellcolor{ImperialColor!15} 98.46 $\pm$ 0.03 \\
     2-layer GCN-based GAE & \cellcolor{ImperialColor!15} 90.29 $\pm$ 0.39 & \cellcolor{ImperialColor!15} 93.01 $\pm$ 0.33 & \cellcolor{ImperialColor!15} 94.80 $\pm$ 0.08 & \cellcolor{ImperialColor!15} 95.72 $\pm$ 0.05 & \cellcolor{ImperialColor!15} 97.97 $\pm$ 0.09 & \cellcolor{ImperialColor!15} 98.12 $\pm$ 0.09 \\
     3-layer GCN-based GAE & \cellcolor{ImperialColor!15} 89.91 $\pm$ 0.61 & \cellcolor{ImperialColor!15} 92.24 $\pm$ 0.67 & \cellcolor{ImperialColor!15} 94.51 $\pm$ 0.31 & \cellcolor{ImperialColor!15} 95.11 $\pm$ 0.28 & \cellcolor{ImperialColor!15} 94.35 $\pm$ 1.30 & \cellcolor{ImperialColor!15} 94.46 $\pm$ 1.31\\
    \midrule
    Linear VGAE (ours) & \cellcolor{ImperialColor!15} 90.62 $\pm$ 0.30 & \cellcolor{ImperialColor!15} 93.25 $\pm$ 0.22 & \cellcolor{ImperialColor!15} 95.20 $\pm$ 0.16 & \cellcolor{ImperialColor!15} 95.99 $\pm$ 0.12 & \cellcolor{ImperialColor!15} 98.35 $\pm$ 0.05 & \cellcolor{ImperialColor!15} 98.46 $\pm$ 0.05\\
     2-layer GCN-based VGAE & \cellcolor{ImperialColor!15} 90.40 $\pm$ 0.43 & \cellcolor{ImperialColor!15} 93.09 $\pm$ 0.35 & \cellcolor{ImperialColor!15} 94.60 $\pm$ 0.20 & \cellcolor{ImperialColor!15} 95.74 $\pm$ 0.13 & \cellcolor{ImperialColor!15} 97.75 $\pm$ 0.08 & \cellcolor{ImperialColor!15} 97.91 $\pm$ 0.06 \\
     3-layer GCN-based VGAE & \cellcolor{ImperialColor!15} 89.92 $\pm$ 0.59 & \cellcolor{ImperialColor!15} 92.52 $\pm$ 0.48 & \cellcolor{ImperialColor!15} 94.48 $\pm$ 0.28 & \cellcolor{ImperialColor!15} 95.30 $\pm$ 0.22 & \cellcolor{ImperialColor!15} 94.57 $\pm$ 1.14 & \cellcolor{ImperialColor!15} 94.73 $\pm$ 1.12 \\
    \midrule
    \midrule
    & \multicolumn{2}{c}{\textbf{LiveMocha}} & \multicolumn{2}{c}{\textbf{Flickr}} & \multicolumn{2}{c}{\textbf{Patent}} \\
        \textbf{Model}  & \multicolumn{2}{c}{\footnotesize (n = 104 103, m = 2 193 083)} & \multicolumn{2}{c}{\footnotesize (n = 105 938, m = 2 316 948)} &\multicolumn{2}{c}{\footnotesize (n = 2 745 762, m = 13 965 410)} \\
     \cmidrule{2-7} 
     & \footnotesize \textbf{AUC (in \%)} & \footnotesize \textbf{AP (in \%)} & \footnotesize \textbf{AUC (in \%)} & \footnotesize \textbf{AP (in \%)} & \footnotesize \textbf{AUC (in \%)} & \footnotesize \textbf{AP (in \%)} \\
    \midrule
    \midrule
    Linear GAE (ours) &\cellcolor{ImperialColor!15} 93.35 $\pm$ 0.10 & \cellcolor{ImperialColor!15} 94.83 $\pm$ 0.08 & \cellcolor{ImperialColor!15} 96.38 $\pm$ 0.05 & \cellcolor{ImperialColor!15} 97.27 $\pm$ 0.04 & \cellcolor{ImperialColor!15} 85.49 $\pm$ 0.09 & \cellcolor{ImperialColor!15} 87.17 $\pm$ 0.07\\
     2-layer GCN-based GAE & \cellcolor{ImperialColor!15} 92.79 $\pm$ 0.17 & \cellcolor{ImperialColor!15} 94.33 $\pm$ 0.13 & \cellcolor{ImperialColor!15} 96.34 $\pm$ 0.05 & \cellcolor{ImperialColor!15} 97.22 $\pm$ 0.04 & \cellcolor{ImperialColor!15} 82.86 $\pm$ 0.20 & \cellcolor{ImperialColor!15} 84.52 $\pm$ 0.24\\
     3-layer GCN-based GAE & \cellcolor{ImperialColor!15} 92.22 $\pm$ 0.73 & \cellcolor{ImperialColor!15} 93.67 $\pm$ 0.57 & \cellcolor{ImperialColor!15} 96.06 $\pm$ 0.08 & \cellcolor{ImperialColor!15} 97.01 $\pm$ 0.05 & \cellcolor{ImperialColor!15} 83.77 $\pm$ 0.41 & \cellcolor{ImperialColor!15} 84.73 $\pm$ 0.42  \\
    \midrule
    Linear VGAE (ours) & \cellcolor{ImperialColor!15} 93.23 $\pm$ 0.06 & \cellcolor{ImperialColor!15} 94.61 $\pm$ 0.05 & \cellcolor[gray]{1.0} 96.05 $\pm$ 0.08 & \cellcolor{ImperialColor!15} 97.12 $\pm$ 0.06 & \cellcolor{ImperialColor!15} 84.57 $\pm$ 0.27 & \cellcolor{ImperialColor!15} 85.46 $\pm$ 0.30\\
     2-layer GCN-based VGAE & \cellcolor{ImperialColor!15} 92.68 $\pm$ 0.21 & \cellcolor{ImperialColor!15} 94.23 $\pm$ 0.15 & \cellcolor[gray]{1.0} 96.35 $\pm$ 0.07 & \cellcolor{ImperialColor!15} 97.20 $\pm$ 0.06 & \cellcolor{ImperialColor!15} 83.77 $\pm$ 0.28 & \cellcolor{ImperialColor!15} 83.37 $\pm$ 0.26  \\
     3-layer GCN-based VGAE & \cellcolor{ImperialColor!15} 92.71 $\pm$ 0.37 & \cellcolor{ImperialColor!15} 94.01 $\pm$ 0.26 & \cellcolor[gray]{1.0} 96.39 $\pm$ 0.13 & \cellcolor{ImperialColor!15} 97.16 $\pm$ 0.08 & \cellcolor{ImperialColor!15} 85.30 $\pm$ 0.51 & \cellcolor{ImperialColor!15} 86.14 $\pm$ 0.49  \\
    \midrule
    \midrule
    & \multicolumn{2}{c}{\textbf{Blogs}} & \multicolumn{2}{c}{\textbf{Amazon}} & \multicolumn{2}{c}{\textbf{Google}} \\
        \textbf{Model} & \multicolumn{2}{c}{\footnotesize (n = 1 224, m = 19 025)}  & \multicolumn{2}{c}{\footnotesize (n = 334 863, m = 925 872)} & \multicolumn{2}{c}{\footnotesize (n = 875 713, m = 5 105 039)} \\
     \cmidrule{2-7} 
     & \footnotesize \textbf{AUC (in \%)} & \footnotesize \textbf{AP (in \%)} & \footnotesize \textbf{AUC (in \%)} & \footnotesize \textbf{AP (in \%)} & \footnotesize \textbf{AUC (in \%)} & \footnotesize \textbf{AP (in \%)} \\
    \midrule
    \midrule
    Linear GAE (ours) & \cellcolor{ImperialColor!15} 91.71 $\pm$ 0.39 & \cellcolor{ImperialColor!15} 92.53 $\pm$ 0.44 & \cellcolor{ImperialColor!15} 90.70 $\pm$ 0.09 & \cellcolor{ImperialColor!15} 93.46 $\pm$ 0.08  & \cellcolor{ImperialColor!15} 95.37 $\pm$ 0.05 & \cellcolor{ImperialColor!15} 96.93 $\pm$ 0.05 \\
     2-layer GCN-based GAE & \cellcolor{ImperialColor!15} 91.57 $\pm$ 0.34 & \cellcolor{ImperialColor!15} 92.51 $\pm$ 0.29 & \cellcolor{ImperialColor!15} 90.15 $\pm$ 0.15 & \cellcolor{ImperialColor!15} 92.33 $\pm$ 0.14 & \cellcolor{ImperialColor!15} 95.06 $\pm$ 0.08 & \cellcolor{ImperialColor!15} 96.40 $\pm$ 0.07 \\
     3-layer GCN-based GAE & \cellcolor{ImperialColor!15} 91.74 $\pm$ 0.37 & \cellcolor{ImperialColor!15} 92.62 $\pm$ 0.31  & \cellcolor{ImperialColor!15} 88.54 $\pm$ 0.37 & \cellcolor{ImperialColor!15} 90.47 $\pm$ 0.38 & \cellcolor{ImperialColor!15} 93.68 $\pm$ 0.15 & \cellcolor{ImperialColor!15} 94.99 $\pm$ 0.14  \\
    \midrule
    Linear VGAE (ours) & 91.34 $\pm$ 0.24 & 92.10 $\pm$ 0.24 & \cellcolor[gray]{1.0} 84.53 $\pm$ 0.08 & \cellcolor[gray]{1.0} 87.79 $\pm$ 0.06 & 91.13 $\pm$ 0.14 & 93.79 $\pm$ 0.10 \\
     2-layer GCN-based VGAE & 91.85 $\pm$ 0.22 & 92.60 $\pm$ 0.25 & \cellcolor[gray]{1.0} 90.14 $\pm$ 0.22 & \cellcolor[gray]{1.0} 92.33 $\pm$ 0.23 & 95.04 $\pm$ 0.09 & 96.38 $\pm$ 0.07 \\
     3-layer GCN-based VGAE & 91.83 $\pm$ 0.48 & 92.65 $\pm$ 0.35 & \cellcolor[gray]{1.0} 89.44 $\pm$ 0.25 & \cellcolor[gray]{1.0} 91.23 $\pm$ 0.23 & 93.79 $\pm$ 0.22 & 95.12 $\pm$ 0.21 \\
    \midrule
    \midrule
    & \multicolumn{2}{c}{\textbf{Stanford}} & \multicolumn{2}{c}{\textbf{Proteins}} & \multicolumn{2}{c}{\textbf{Google-Medium}} \\
        \textbf{Model}  & \multicolumn{2}{c}{\footnotesize (n = 281 903, m = 2 312 497)} & \multicolumn{2}{c}{\footnotesize (n = 6 327, m = 147 547)} & \multicolumn{2}{c}{\footnotesize (n = 15 763, m = 171 206)} \\
     \cmidrule{2-7} 
     & \footnotesize \textbf{AUC (in \%)} & \footnotesize \textbf{AP (in \%)} & \footnotesize \textbf{AUC (in \%)} & \footnotesize \textbf{AP (in \%)} & \footnotesize \textbf{AUC (in \%)} & \footnotesize \textbf{AP (in \%)} \\
    \midrule
    \midrule
    Linear GAE (ours) & \cellcolor{ImperialColor!15} 97.73 $\pm$ 0.10 & \cellcolor{ImperialColor!15} 98.37 $\pm$ 0.10 & \cellcolor[gray]{1.0} 94.09 $\pm$ 0.23 & \cellcolor[gray]{1.0} 96.01 $\pm$ 0.16 & \cellcolor[gray]{1.0} 96.02 $\pm$ 0.14 & \cellcolor[gray]{1.0} 97.09 $\pm$ 0.08 \\
     2-layer GCN-based GAE & \cellcolor{ImperialColor!15} 97.05 $\pm$ 0.63 & \cellcolor{ImperialColor!15} 97.56 $\pm$ 0.55  & \cellcolor[gray]{1.0} 94.55 $\pm$ 0.20 & \cellcolor[gray]{1.0} 96.39 $\pm$ 0.16 & \cellcolor[gray]{1.0} 96.66 $\pm$ 0.24 & \cellcolor[gray]{1.0} 97.45 $\pm$ 0.25 \\
     3-layer GCN-based GAE & \cellcolor{ImperialColor!15} 92.19 $\pm$ 1.49 & \cellcolor{ImperialColor!15} 92.58 $\pm$ 1.50 & \cellcolor[gray]{1.0} 94.30 $\pm$ 0.19 & \cellcolor[gray]{1.0} 96.08 $\pm$ 0.15 & \cellcolor[gray]{1.0} 95.10 $\pm$ 0.27 & \cellcolor[gray]{1.0} 95.94 $\pm$ 0.20  \\
    \midrule
    Linear VGAE (ours) & \cellcolor[gray]{1.0} 94.96 $\pm$ 0.25 & \cellcolor[gray]{1.0} 96.64 $\pm$ 0.15 & 93.99 $\pm$ 0.10 & \cellcolor{ImperialColor!15} 95.94 $\pm$ 0.16 & 91.11 $\pm$ 0.31 & 92.91 $\pm$ 0.18\\
     2-layer GCN-based VGAE & \cellcolor[gray]{1.0} 97.60 $\pm$ 0.11 & \cellcolor[gray]{1.0} 98.02 $\pm$ 0.10 & 94.57 $\pm$ 0.18 & \cellcolor{ImperialColor!15} 96.18 $\pm$ 0.33 & 96.11 $\pm$ 0.59 & 96.84 $\pm$ 0.51 \\
     3-layer GCN-based VGAE &\cellcolor[gray]{1.0} 97.53 $\pm$ 0.13 & \cellcolor[gray]{1.0} 98.01 $\pm$ 0.10& 94.27 $\pm$ 0.25 & \cellcolor{ImperialColor!15} 95.71 $\pm$ 0.28 & 95.10 $\pm$ 0.54 & 96.00 $\pm$ 0.44 \\
    \bottomrule
  \end{tabular}
}
\end{table} 
\clearpage

\begin{table}[t]
  \centering
\caption[Community detection on all graphs using Linear GAE and VGAE]{Community detection on graphs with communities, with Linear GAE and VGAE models, and with their multi-layer GCN-based counterparts. We report details on hyperparameters for each model and dataset in Section~\ref{c6s631}. Cells are colored in \textcolor{ImperialColor}{blue} when Linear GAE/VGAE models are reaching competitive results w.r.t. standard GCN-based models, i.e., results that are at least as good as GCN-based models $\pm$ 1 standard deviation. Note: as we aim to evaluate whether linear models are \textit{as good} as others, and not necessarily \textit{better}, we do not report \textbf{bold} numbers in this table contrary to tables from previous chapters.}
\label{t3}
 \resizebox{1.0\textwidth}{!}{
  \begin{tabular}{c|c|c|c|c}
    \toprule
    & \textbf{Cora} & \textbf{Cora with features} & \textbf{Citeseer} & \textbf{Citeseer with features}\\
      \textbf{Model}  & \footnotesize (n = 2 708, & \footnotesize (n = 2 708,  m = 5 429, & \footnotesize (n = 3 327, & \footnotesize (n = 3 327, m = 4 732,\\
      &  \footnotesize m = 5 429) & \footnotesize f = 1 433) & \footnotesize m = 4 732) & \footnotesize f = 3 703)\\
     \cmidrule{2-5}
      & \footnotesize \textbf{AMI (in \%)} & \footnotesize \textbf{AMI (in \%)} & \footnotesize \textbf{AMI (in \%)} & \footnotesize \textbf{AMI (in \%)}\\
    \midrule
    \midrule
    Linear GAE (ours) & 26.31 $\pm$ 2.85 & \cellcolor{ImperialColor!15} 47.02 $\pm$ 2.09 & \cellcolor{ImperialColor!15} 8.56 $\pm$ 1.28 & \cellcolor{ImperialColor!15} 20.23 $\pm$ 1.36 \\
     2-layer GCN-based GAE & 30.88 $\pm$ 2.56 & \cellcolor{ImperialColor!15} 43.04 $\pm$ 3.28 & \cellcolor{ImperialColor!15} 9.46 $\pm$ 1.06 & \cellcolor{ImperialColor!15} 19.38 $\pm$ 3.15 \\
     3-layer GCN-based GAE & 33.06 $\pm$ 3.10 & \cellcolor{ImperialColor!15} 44.12 $\pm$ 2.48 & \cellcolor{ImperialColor!15} 10.69 $\pm$ 1.98 & \cellcolor{ImperialColor!15} 19.71 $\pm$ 2.55 \\
     \midrule
         Linear VGAE (ours) & \cellcolor{ImperialColor!15} 34.35 $\pm$ 1.42 & \cellcolor{ImperialColor!15} 48.12 $\pm$ 1.96 & \cellcolor{ImperialColor!15} 12.67 $\pm$ 1.27 & \cellcolor{ImperialColor!15} 20.71 $\pm$ 1.95  \\
     2-layer GCN-based VGAE & \cellcolor{ImperialColor!15} 26.66 $\pm$ 3.94 & \cellcolor{ImperialColor!15} 44.84 $\pm$ 2.63 & \cellcolor{ImperialColor!15} 9.85 $\pm$ 1.24 & \cellcolor{ImperialColor!15} 20.17 $\pm$ 3.07 \\
     3-layer GCN-based VGAE & \cellcolor{ImperialColor!15} 28.43 $\pm$ 2.83 & \cellcolor{ImperialColor!15} 44.29 $\pm$ 2.54 & \cellcolor{ImperialColor!15} 10.64 $\pm$ 1.47 & \cellcolor{ImperialColor!15} 19.94 $\pm$ 2.50  \\
    \midrule
    \midrule
    & \textbf{Pubmed} & \textbf{Pubmed with features} & \textbf{Cora-larger} & \textbf{Blogs}\\
       \textbf{Model}  & \footnotesize (n = 19 717, & \footnotesize (n = 19 717, m = 44 338,  & \footnotesize (n = 23 166, & \footnotesize (n = 1 224,  m = 19 025)\\
     &  \footnotesize m = 44 338) & \footnotesize f = 500) & \footnotesize  m = 91 500) & \\
     \cmidrule{2-5}
      & \footnotesize \textbf{AMI (in \%)} & \footnotesize \textbf{AMI (in \%)} & \footnotesize \textbf{AMI (in \%)} & \footnotesize \textbf{AMI (in \%)}\\
    \midrule
    \midrule
    Linear GAE (ours) & 10.76 $\pm$ 3.70 & \cellcolor{ImperialColor!15} 26.12 $\pm$ 1.94 & \cellcolor{ImperialColor!15} 40.34 $\pm$ 0.51 & 46.84 $\pm$ 1.79 \\
     2-layer GCN-based GAE & 16.41 $\pm$ 3.15 & \cellcolor{ImperialColor!15} 23.08 $\pm$ 3.35 & \cellcolor{ImperialColor!15} 39.75 $\pm$ 0.79 & 72.58 $\pm$ 4.54 \\
     3-layer GCN-based GAE & 23.11 $\pm$ 2.58 & \cellcolor{ImperialColor!15} 25.94 $\pm$ 3.09 & \cellcolor{ImperialColor!15} 35.67 $\pm$ 1.76 & 72.72 $\pm$ 1.80 \\
     \midrule
         Linear VGAE (ours) & \cellcolor{ImperialColor!15} 25.14 $\pm$ 2.83 & \cellcolor{ImperialColor!15} 29.74 $\pm$ 0.64 & \cellcolor{ImperialColor!15} 43.32 $\pm$ 0.52 & \cellcolor[gray]{1.0} 49.70 $\pm$ 1.08 \\
     2-layer GCN-based VGAE & \cellcolor{ImperialColor!15} 20.52 $\pm$ 2.97 & \cellcolor{ImperialColor!15} 25.43 $\pm$ 1.47 & \cellcolor{ImperialColor!15} 38.34 $\pm$ 0.64 & \cellcolor[gray]{1.0} 73.12 $\pm$ 0.83 \\
     3-layer GCN-based VGAE & \cellcolor{ImperialColor!15} 21.32 $\pm$ 3.70 & \cellcolor{ImperialColor!15} 24.91 $\pm$ 3.09 & \cellcolor{ImperialColor!15} 37.30 $\pm$ 1.07 & \cellcolor[gray]{1.0} 70.56 $\pm$ 5.43 \\
    \bottomrule
  \end{tabular}
}
\end{table}

\paragraph{Experiments on more complex decoders}
So far, we compared different encoders but the standard inner product decoder was fixed. As a robustness check, in the next Table~\ref{t4}, we report complementary link prediction experiments, on variants of GAE and VGAE models with two more complex decoders:
\begin{itemize}
    \item the ``Graphite'' GAE and VGAE models from Grover~et~al.~\cite{grover2019graphite} for iterative generative modeling of graphs, and already mentioned in our experiments from Chapter~\ref{chapter_3}. Graphite models still process undirected graphs only. They rely on an iterative graph refinement strategy inspired by low-rank approximations for decoding, instead of a simple inner product with a sigmoid activation as in Section~\ref{c6s62};
\item our own Gravity-Inspired GAE and VGAE models from Chapter~\ref{chapter_5} that, contrary to the other models, incorporate an asymmetric decoder, i.e., $\hat{A}_{ij} \neq \hat{A}_{ji}$. As explained in Chapter~\ref{chapter_5}, this model is designed for directed link prediction. Therefore, contrary to previous experiments, we do not ignore edges directions when initial graphs were directed. The models from Table~\ref{chapter_4} all process the out-degree normalization of $A$ a.k.a.  $\tilde{A}_{\text{out}}$ instead of the symmetric normalization $\tilde{A}$.
\end{itemize}

\begin{table}[ht]
  \centering
  
\caption[Link prediction using Linear GAE and VGAE with alternative decoders]{Link prediction with Linear GAE and VGAE models incorporating the ``Graphite''~\cite{grover2019graphite} and Gravity-Inspired~\cite{salha2019-2} decoders, and with their multi-layer GCN-based counterparts. We report details on hyperparameters for each model and dataset in Section~\ref{c6s631}. Cells are colored in \textcolor{ImperialColor}{blue} when Linear GAE/VGAE models are reaching competitive results w.r.t. standard GCN-based models, i.e., results that are at least as good as GCN-based models $\pm$ 1 standard deviation. Note: as we aim to evaluate whether linear models are \textit{as good} as others, and not necessarily \textit{better}, we do not report \textbf{bold} numbers in this table contrary to tables from previous chapters.}
\label{t4}
 \resizebox{0.85\textwidth}{!}{
  \begin{tabular}{c|cc|cc}
    \toprule
    & \multicolumn{2}{c}{\textbf{Cora}} & \multicolumn{2}{c}{\textbf{Citeseer}} \\
     \textbf{Model} & \multicolumn{2}{c}{\footnotesize (n = 2 708, m = 5 429)} & \multicolumn{2}{c}{\footnotesize (n = 3 327, m = 4 732)} \\
     \cmidrule{2-5}
      & \footnotesize \textbf{AUC (in \%)} & \footnotesize \textbf{AP (in \%)} & \footnotesize \textbf{AUC (in \%)} & \footnotesize \textbf{AP (in \%)}\\
    \midrule
    \midrule
    Linear Graphite GAE (ours) & \cellcolor{ImperialColor!15} 83.42 $\pm$ 1.76 & \cellcolor{ImperialColor!15} 87.32 $\pm$ 1.53 & \cellcolor{ImperialColor!15} 77.56 $\pm$ 1.41 & \cellcolor{ImperialColor!15} 82.88 $\pm$ 1.15 \\
     2-layer Graphite GAE & \cellcolor{ImperialColor!15} 81.20 $\pm$ 2.21 & \cellcolor{ImperialColor!15} 85.11 $\pm$ 1.91 & \cellcolor{ImperialColor!15} 73.80 $\pm$ 2.24 & \cellcolor{ImperialColor!15} 79.32 $\pm$ 1.83 \\
     3-layer Graphite GAE & \cellcolor{ImperialColor!15} 79.06 $\pm$ 1.70 & \cellcolor{ImperialColor!15} 81.79 $\pm$ 1.62 & \cellcolor{ImperialColor!15} 72.24 $\pm$ 2.29 & \cellcolor{ImperialColor!15} 76.60 $\pm$ 1.95 \\
    \midrule
    Linear Graphite VGAE (ours) & \cellcolor{ImperialColor!15} 83.68 $\pm$ 1.42 & \cellcolor{ImperialColor!15} 87.57 $\pm$ 1.16 & \cellcolor{ImperialColor!15} 78.90 $\pm$ 1.08 & \cellcolor{ImperialColor!15} 83.51 $\pm$ 0.89 \\
     2-layer Graphite VGAE  & \cellcolor{ImperialColor!15} 84.89 $\pm$ 1.48 & \cellcolor{ImperialColor!15} 88.10 $\pm$ 1.22 & \cellcolor{ImperialColor!15} 77.92 $\pm$ 1.57 & \cellcolor{ImperialColor!15} 82.56 $\pm$ 1.31 \\
     3-layer Graphite VGAE  & \cellcolor{ImperialColor!15} 85.33 $\pm$ 1.19 & \cellcolor{ImperialColor!15} 87.98 $\pm$ 1.09 & \cellcolor{ImperialColor!15} 77.46 $\pm$ 2.34 & \cellcolor{ImperialColor!15} 81.95 $\pm$ 1.71\\
    \midrule
    \midrule
    Linear Gravity-Inspired GAE (ours) & \cellcolor{ImperialColor!15} 90.71 $\pm$ 0.95 & \cellcolor{ImperialColor!15} 92.95 $\pm$ 0.88 & \cellcolor{ImperialColor!15} 80.52 $\pm$ 1.37 & \cellcolor{ImperialColor!15} 86.29 $\pm$ 1.03 \\
     2-layer Gravity-Inspired GAE & \cellcolor{ImperialColor!15} 87.79 $\pm$ 1.07 & \cellcolor{ImperialColor!15} 90.78 $\pm$ 0.82 & \cellcolor{ImperialColor!15} 78.36 $\pm$ 1.55 & \cellcolor{ImperialColor!15} 84.75 $\pm$ 1.10 \\
     3-layer Gravity-Inspired GAE & \cellcolor{ImperialColor!15} 87.76 $\pm$ 1.32 & \cellcolor{ImperialColor!15} 90.15 $\pm$ 1.45 & \cellcolor{ImperialColor!15} 78.32 $\pm$ 1.92 & \cellcolor{ImperialColor!15} 84.88 $\pm$ 1.36 \\
    \midrule
    Linear Gravity-Inspired VGAE (ours) & \cellcolor{ImperialColor!15} 91.29 $\pm$ 0.70 & \cellcolor{ImperialColor!15} 93.01 $\pm$ 0.57 & \cellcolor{ImperialColor!15} 86.65 $\pm$ 0.95 & \cellcolor{ImperialColor!15} 89.49 $\pm$ 0.69 \\
     2-layer Gravity-Inspired VGAE  & \cellcolor{ImperialColor!15} 91.92 $\pm$ 0.75 & \cellcolor{ImperialColor!15} 92.46 $\pm$ 0.64 & \cellcolor{ImperialColor!15} 87.67 $\pm$ 1.07 & \cellcolor{ImperialColor!15} 89.79 $\pm$ 1.01\\
     3-layer Gravity-Inspired VGAE  & \cellcolor{ImperialColor!15} 90.80 $\pm$ 1.28 & \cellcolor{ImperialColor!15} 92.01 $\pm$ 1.19 & \cellcolor{ImperialColor!15} 85.28 $\pm$ 1.33 & \cellcolor{ImperialColor!15} 87.54 $\pm$ 1.21 \\
    \midrule
    \midrule
    & \multicolumn{2}{c}{\textbf{Pubmed}} & \multicolumn{2}{c}{\textbf{Google-Medium}} \\
     \textbf{Model} & \multicolumn{2}{c}{\footnotesize (n = 19 717, m = 44 338)} & \multicolumn{2}{c}{\footnotesize (n = 15 763, m = 171 206)} \\
     \cmidrule{2-5}
      & \footnotesize \textbf{AUC (in \%)} & \footnotesize \textbf{AP (in \%)} & \footnotesize \textbf{AUC (in \%)} & \footnotesize \textbf{AP (in \%)}\\
    \midrule
    \midrule
    Linear Graphite GAE (ours) & \cellcolor{ImperialColor!15} 80.28 $\pm$ 0.86 & \cellcolor{ImperialColor!15} 85.81 $\pm$ 0.67 & 94.30 $\pm$ 0.22  & 95.09 $\pm$ 0.16 \\
     2-layer Graphite GAE & \cellcolor{ImperialColor!15} 79.98 $\pm$ 0.66 & \cellcolor{ImperialColor!15} 85.33 $\pm$ 0.41 & 95.54 $\pm$ 0.42  & 95.99 $\pm$ 0.39 \\
     3-layer Graphite GAE & \cellcolor{ImperialColor!15} 79.96 $\pm$ 1.40 & \cellcolor{ImperialColor!15} 84.88 $\pm$ 0.89 & 93.99 $\pm$ 0.54 & 94.74 $\pm$ 0.49 \\
    \midrule
    Linear Graphite VGAE (ours) & 79.59 $\pm$ 0.33 & 86.17 $\pm$ 0.31 & 92.71 $\pm$ 0.38  & 94.41 $\pm$ 0.25 \\
     2-layer Graphite VGAE  & 82.74 $\pm$ 0.30 & 87.19 $\pm$ 0.36 & 96.49 $\pm$ 0.22  & 96.91 $\pm$ 0.17 \\
     3-layer Graphite VGAE  & 84.56 $\pm$ 0.42 & 88.01 $\pm$ 0.39 & 96.32 $\pm$ 0.24  & 96.62 $\pm$ 0.20 \\
    \midrule
    \midrule
    Linear Gravity-Inspired GAE (ours) & \cellcolor{ImperialColor!15} 76.78 $\pm$ 0.38 & \cellcolor{ImperialColor!15} 84.50 $\pm$ 0.32 & 97.46 $\pm$ 0.07  & 98.30 $\pm$ 0.04 \\
     2-layer Gravity-Inspired GAE & \cellcolor{ImperialColor!15} 75.84 $\pm$ 0.42 & \cellcolor{ImperialColor!15} 83.03 $\pm$ 0.22 & 97.77 $\pm$ 0.10  & 98.43 $\pm$ 0.10 \\
     3-layer Gravity-Inspired GAE & \cellcolor{ImperialColor!15} 74.61 $\pm$ 0.30 & \cellcolor{ImperialColor!15} 81.68 $\pm$ 0.26 & 97.58 $\pm$ 0.12  & 98.28 $\pm$ 0.11 \\
    \midrule   
    Linear Gravity-Inspired VGAE (ours) & \cellcolor{ImperialColor!15} 79.68 $\pm$ 0.36 & \cellcolor{ImperialColor!15} 85.00 $\pm$ 0.21 & 97.32 $\pm$ 0.06  & \cellcolor{ImperialColor!15} 98.26 $\pm$ 0.05 \\
     2-layer Gravity-Inspired VGAE  & \cellcolor{ImperialColor!15} 77.30 $\pm$ 0.81 & \cellcolor{ImperialColor!15} 82.64 $\pm$ 0.27 & 97.84 $\pm$ 0.25  & \cellcolor{ImperialColor!15} 98.18 $\pm$ 0.14 \\
     3-layer Gravity-Inspired VGAE  & \cellcolor{ImperialColor!15} 76.52 $\pm$ 0.61 & \cellcolor{ImperialColor!15} 80.73 $\pm$ 0.63 & 97.32 $\pm$ 0.23 & \cellcolor{ImperialColor!15} 97.81 $\pm$ 0.20 \\
    \bottomrule
  \end{tabular}
}
\end{table}

Overall, we draw similar conclusions w.r.t. Tables~\ref{t1} and \ref{t2}, consolidating our findings. For brevity, in Tables~\ref{t4} we only report results for the Cora, Citeseer, and Pubmed graphs, where linear models are competitive, and for the Google-Medium graph, where GCN-based GAE and VGAE models slightly outperform their linear counterparts. We stress out that scores from Graphite and Gravity-Inspired GAEs/VGAEs are \textit{not} directly comparable, as the former ignores the directionality of edges while the latter processes directed graphs, i.e., the learning task is a \textit{directed} link prediction problem for Gravity-Inspired GAEs/VGAEs.

\paragraph{When (not) to use multi-layer GCN encoders?}

Linear GAE and VGAE models reached strong empirical results on all graphs under consideration in this work despite their various sizes and characteristics, and rival or outperform GCN-based GAE and VGAE models in a majority of experiments. These models are also simpler and more interpretable, each element of $z_i$ being interpreted as a weighted average from node $i$'s direct neighborhood. We recall that, in Chapter~\ref{chapter_3}, we also showed that multi-layer GCNs are themselves often competitive (or very close) to more complex GNN encoders. Therefore, we recommend the systematic use of Linear GAE and VGAE models as a first baseline, before diving into more complex encoding schemes whose actual benefit might be unclear.

From our experiments, we also conjecture that multi-layer GCN encoders \textit{can} bring an empirical advantage in the presence of graphs with \textit{intrinsic non-trivial high-order interactions}. Notable examples of such graphs include the Amazon co-purchase graph (+5.61 AUC points for 2-layer GCN-based VGAE) and web graphs such as Blogs, Google, and Stanford, in which two-hop hyperlinks connections of pages usually include relevant information on the global network structure. On such graphs, capturing this information tends to improve results, especially 1) for the probabilistic VGAE framework, and 2) when evaluating embeddings via the community detection task (e.g., 20+ AMI points on Blogs for 2-layer GCN-based GAE and VGAE) which is, by design, a more \textit{global} learning task than the quite \textit{local} link prediction problem.

On the contrary, in citation graphs such as Cora, Citeseer, and Pubmed, the relevance of two-hop links is more limited. Indeed, if a reference A in an article B cited by some authors is relevant to their work, authors will likely also cite this reference A, thus creating a one-hop link. Lastly, while the impact of the graph \textit{size} is unclear in our experiments (linear models achieve strong results even on large graphs, such as Patent), we note that graphs where multi-layer GCN encoders tend to outperform linear models are all relatively \textit{dense}.

As a consequence, we conjecture that denser graphs with intrinsic high-order interactions (e.g., the aforestated web graphs) should be better suited than the sparse Cora, Citeseer, and Pubmed citation networks to evaluate and compare complex GAE and VGAE models. Our recommendation to the scientific community is not to completely avoid these three graphs, but rather to stop using them \textit{exclusively} in experiments.

\section{Conclusion}
\label{c6s64}

In this chapter, we proposed to simplify GAE and VGAE models.  While most existing variants of these methods rely on multi-layer GCNs encoders to learn embedding vectors, we emphasized that these encoders are unnecessarily complex for many applications. We suggested replacing them with simpler linear models w.r.t. the (one-hop) normalized adjacency matrix of the graph, involving fewer operations, fewer parameters, and no activation function. We provided an in-depth experimental evaluation of such an approach, showing that our simplified models consistently reach competitive or very close performances w.r.t. multi-layer GCN-based GAE and VGAE models on link prediction and community detection, and on numerous real-world graphs. We also aimed to identify the settings where simple one-hop linear encoders appear as an effective alternative to multi-layer GCNs. 

Based on these experiments, and on previous results from Chapter~\ref{chapter_3} showing that multi-layer GCNs can themselves be empirically close to more complex GNN encoders, we recommended the systematic use of Linear GAE and VGAE models as a first baseline, before diving into more complex encoding schemes whose actual benefit might be unclear. Since the publication of this research in late 2019 (for the NeurIPS workshop paper~\cite{salha2019keep}) and then in mid-2020 (for the ECML-PKDD conference paper~\cite{salha2020simple}), several recent articles including \cite{ahn2021variational,choong2020optimizing,do2021improving,hibshman2021joint,shin2020bipartite,wu2021deepened} considered Linear GAE and VGAE models in their own experiments, either as a baseline or as a component of a larger model. These experiments confirmed that one-hop linear models often reach comparable results w.r.t. multi-layer GCNs.

Besides research papers, we emphasize that such a result is valuable in the context of an industrial PhD thesis. Indeed, in the industry, e.g., for music streaming services, simple models are often preferred in production environments, as they are easier to understand, to deploy, and to debug. In Chapters~\ref{chapter_9}~and~\ref{chapter_10}, we will provide a broader overview  of  how  Deezer  internally leverages similarity graphs and graph ontologies for music  recommendation, involving Linear GAE and VGAE models. Moreover, to encourage the future usage of our method, we also recently implemented a PyTorch version of Linear GAE and VGAE, now available in the popular PyTorch-Geometric\footnote{\href{https://github.com/pyg-team/pytorch_geometric}{https://github.com/pyg-team/pytorch\_geometric}} library~\cite{pytorchgeometric}, in addition to our initial TensorFlow implementation\footnote{\href{https://github.com/deezer/linear_graph_autoencoders}{https://github.com/deezer/linear\_graph\_autoencoders}}.

Last, but not least, in this chapter we also questioned the relevance of repeatedly using the same sparse medium-size citation networks (Cora, Citeseer, Pubmed) to evaluate and compare complex GAE and VGAE models. In Section~\ref{c6s632}, we recommended to stop using these three graphs \textit{exclusively} in experiments. While we admit that finding ``challenging'' alternative datasets might be difficult,
we also praise the recent efforts from the graph learning community to release large, various, and realistic benchmark datasets. This includes the Open Graph Benchmark~\cite{hu2020open} initiative, aiming to provide challenging datasets for various graph-based machine learning problems, together with standardized dataset splits and evaluators to properly compare the performances of different models.

\chapter[Improving Community Detection with Graph Autoencoders]{Improving Community Detection with~Graph~Autoencoders }\label{chapter_7}
\chaptermark{Improving Community Detection with Graph Autoencoders}

\newcommand\todo[1]{\textcolor{red}{#1}}
\newcommand{\weight}{\lambda}
\newcommand{\Ac}{A_c}
\newcommand{\agcn}[1]{\mathcal{F}(#1)} 
\newcommand{\ncluster}{K}
\newcommand{\clustersubscript}{k}
\newcommand{\dreg}{k} 
\newcommand{\dregc}{s}
\newcommand{\Ao}{A_\dregc}
\newcommand{\Acomp}{A'}
\newcommand{\ncomp}{n'}
\newcommand{\Accomp}{\Ac'}
\newcommand{\dregconec}{o_\clustersubscript}
\newcommand{\GCNFirstEval}{\theta_1}
\newcommand{\GCNSecondEval}{\theta_2}
\newcommand{\GCNAoSecondEval}{\gamma_2}

\newcommand{\lossGAE}{\mathcal{L}_{\text{GAE}}}
\newcommand{\lossVGAE}{\mathcal{L}_{\text{VGAE}}}
\newcommand{\lossMAGAE}{\tilde{\mathcal{L}}_{\text{GAE}}}
\newcommand{\lossMAVGAE}{\tilde{\mathcal{L}}_{\text{VGAE}}}

\textit{This chapter presents research conducted with Johannes F. Lutzeyer, George Dasoulas, Romain Hennequin, and Michalis Vazirgiannis, and currently under review for publication in Elsevier's Neural Networks journal in 2022~\cite{salhagalvan2022modularity}.}

\section{Introduction}
\label{c7s71}

GAE and VGAE models were originally mainly designed for link prediction, at least in their modern formulation~\cite{kipf2016-2}. The overall effectiveness of GAE and VGAE models and of their extensions on this specific task has been widely experimentally confirmed over the past few years \cite{grover2019graphite,semiimplicit2019,pan2018arga,aaai20,tran2018multi,huang2019rwr}, including in our own experiments from Chapters~\ref{chapter_3}~to~\ref{chapter_6}.

On the other hand, several concurring studies \cite{choong2018learning,choong2020optimizing,salha2021fastgae} have simultaneously pointed out the limitations of these models on community detection. They emphasized that standard GAEs and VGAEs are often outperformed by simpler clustering alternatives, such as the popular Louvain method \cite{blondel2008louvain}. We ourselves observed and discussed this limitation in some of our community detection experiments from Chapters~\ref{chapter_3},~\ref{chapter_4}~and~\ref{chapter_6}.

While some recent studies worked on this issue (see Section~\ref{c7s72} for an overview), their solutions strongly relied on \textit{clustering-oriented probabilistic} priors that only fit the VGAE setting and can not be directly transposed to deterministic GAEs. They also benefited greatly from the presence of \textit{node features} complementing the graph structure, but provided only little to no empirical gain on featureless graphs that are nonetheless ubiquitous. 
Thirdly, they did not explicitly try to preserve the good performances of GAE and VGAE models on link prediction. 
In practice, as we will argue in this chapter, learning node embedding spaces that jointly enable good link prediction and community detection performances is often desirable, both for real-world applications and in pursuit of learning accurate and general representations~of~a~graph~structure.

In summary, the question of how to improve community detection with GAE and VGAE models remains incompletely addressed, especially in the absence of node features, and it is still unclear to which extent one can improve community detection with these models without simultaneously deteriorating link prediction. In this chapter, we propose to tackle these important problems by investigating the following two research questions:
\begin{itemize}
    \item \textbf{Question 1:} can we improve community detection for \textit{both} the GAE \textit{and} VGAE settings? And does this improvement persist for \textit{featureless} graphs?
    \item \textbf{Question 2:} do improvements on the community detection task necessarily incur a loss in the link prediction performance or can they be \textit{jointly} addressed with high accuracy? 
\end{itemize}

In this chapter we propose several novel contributions to both the GAE and VGAE frameworks, which allow us to answer both of these research questions positively. More precisely, our contributions are listed as follows. We first diagnose the reasons why GAE and VGAE models tend to perform well on link prediction but to underperform on community detection. Then, based on insights from this diagnosis, we improve GAE and VGAE models for graph-based community detection while preserving their ability to identify missing edges. Our strategy leverages concepts inspired by \textit{modularity-based} clustering \cite{blondel2008louvain,brandes2007modularity,shiokawa2013fast}. Specifically, we first present and theoretically study a novel \textit{community-preserving message passing scheme}, doping our GAE and VGAE encoders by considering both the initial graph structure and modularity-based prior communities when computing embedding spaces. We also introduce revised training and optimization strategies w.r.t. current practices in the scientific literature, including the introduction of modularity-inspired losses complementing the existing reconstruction losses with the aim of jointly ensuring good performances on link prediction and community detection. Backed by in-depth experiments on several real-world graphs, we demonstrate the empirical effectiveness of our approach at addressing 1)~pure community detection problems, and 2) joint community detection and link prediction problems. We publicly release our source code on GitHub. 

The remainder of this chapter is organized as follows. In Section \ref{c7s72}, we point out the limitations of current GAE and VGAE models on community detection. In Section \ref{c7s73}, we diagnose the reasons explaining these limits. We subsequently introduce and theoretically study our proposed solution, referred to as \textit{Modularity-Aware GAE and VGAE}, to overcome these limitations. We report and discuss our experimental evaluation in Section \ref{c7s74}, and we conclude in Section \ref{c7s75}. We provide additional proofs in Section~\ref{c7s76}, placed out of the ``main'' chapter for the sake of brevity and readability.

\section{On the Limitations of GAE/VGAE-Based Community Detection}
\label{c7s72}

While link prediction remains the most prominent evaluation task for GAE and VGAE models, they have also shown promising results on (semi-supervised) node classification \cite{semiimplicit2019,tran2018multi}, canonical correlation analysis \cite{kaloga2021multiview} and, in the case of VGAE, graph generation especially in the context of molecular graph data \cite{molecule1,molecule2,simonovsky2018graphvae}. However, their performances are less impressive on \textit{community detection} \cite{choong2018learning,choong2020optimizing}, on which we focus in this section.

\subsection{Community Detection with GAE and VGAE}
\label{c7s721}

Throughout this chapter, we consider the community detection problem formulated in Section~\ref{c2s212} from Chapter~\ref{chapter_2}, which we briefly remind in this paragraph. Among the fundamental problems in graph-based machine learning, community detection consists in identifying $\ncluster < n$ clusters a.k.a. \textit{communities} of nodes that, in some sense, are more similar to each other than to the other nodes \cite{choong2018learning,malliaros2013clustering}. More formally, we aim to obtain a partition of $\mathcal{V}$ into $\ncluster$ sets:
\begin{equation}
C_1 \subseteq \mathcal{V}, \ldots, C_\ncluster \subseteq \mathcal{V},
\end{equation}
with cardinality $|C_\clustersubscript| = n_\clustersubscript \leq n $ for $\clustersubscript \in \{1, \ldots, \ncluster\}$. 
The quality of such a partition is usually assessed through some predefined similarity metrics, e.g., unsupervised density-based metrics calculated from the intra- and inter-cluster edge density
\cite{malliaros2013clustering}, or scores such as the \textit{Adjusted Mutual Information} (AMI) \cite{vinh2010information} and \textit{Adjusted Rand Index} (ARI) \cite{hubert1985comparing} scores, that compare the partition to some ground truth node labels hidden during training. 

In the presence of node embedding representations, community detection boils down to the more standard problem of clustering $n$ vectors in a $d$-dimensional Euclidean space into $\ncluster$ groups \cite{macqueen1967some}. With this goal in mind, several studies specifically tried to perform community detection with GAE and VGAE by:
\begin{itemize}
    \item learning an embedding vector $z_i$ for each $i \in \mathcal{V}$, as described in the previous chapters;
    \item clustering the resulting vectors $z_i$ into $\ncluster$ groups, through one of the numerous clustering methods for Euclidean data, such as the popular $k$-means algorithm \cite{macqueen1967some}.
\end{itemize} 

However, concurring experimental evaluations \cite{choong2018learning,choong2020optimizing,salha2021fastgae,salha2019-1} recently pointed out the limitations of such an approach. They emphasized its lower performance w.r.t. simpler community detection alternatives, that sometimes even directly operate on the graph structure without considering node features, such as the popular Louvain method~\cite{blondel2008louvain}. 

For instance, Choong et~al.~\cite{choong2020optimizing} show that, on the (featureless) Cora citation network~\cite{sen2008collective}, a VGAE+$k$-means strategy reaches a mean normalized mutual information score of 23.84\%, way below the Louvain method (43.36\%). In the previous Chapter~\ref{chapter_4} on FastGAE, we showed that, on the same graph, a GAE+$k$-means also reaches an underwhelming 30.88\% mean AMI score. We reached comparable conclusions from experiments on several other graphs, such as the featureless versions \cite{sen2008collective} of Citeseer (9.85\% AMI for VGAE+$k$-means vs 16.39\% for Louvain, in our Chapter~\ref{chapter_4}) and Pubmed (20.41\% for VGAE+$k$-means in the study of Choong et~al.~\cite{choong2020optimizing}, which is comparable to Louvain, but significantly below the 29.46\% obtained by running a $k$-means on node embedding vectors learned via DeepWalk \cite{perozzi2014deepwalk}).

\subsection{Community Detection with Extensions of GAE and VGAE}
\label{c7s722}

Several studies have worked on the issue of the underwhelming performance of GAE and VGAE models in the community detection task \cite{choong2018learning,choong2020optimizing,li2020dirichlet,mrabah2021rethinking}. Choong~et~al.~\cite{choong2018learning} introduced VGAECD, a \textit{VGAE for Community Detection (CD)} model that replaces Gaussian priors by learnable \textit{Gaussian mixtures}. Such a choice permits recovering communities from node embedding spaces without relying on an additional $k$-means step. In a subsequent study \cite{choong2020optimizing}, the same authors proposed VGAECD-OPT, an improved version of VGAECD. Specifically, VGAECD-OPT replaces GCN encoders with the simpler linear models we proposed in Chapter~\ref{chapter_6}. It also adopts a different optimization procedure based on neural expectation-maximization \cite{greff2017neural}, which guarantees that communities do not collapse during training \cite{choong2020optimizing} and experimentally leads to better performances.

More recently, Li et al. \cite{li2020dirichlet} introduced \textit{Dirichlet Variational Graph Autoencoder} (DVGAE), another extension of VGAE which uses Dirichlet distributions as priors on latent vectors, acting as indicators of community memberships. The \textit{Marginalized GAE} (MGAE) model from Wang~et~al.~\cite{wang2017mgae} is also evaluated on community detection. However, the MGAE model does not explicitly leverage embedding representations for this task; instead the \textit{spectral clustering} \cite{von2007tutorial} is applied to the decoded~graphs.
Last, while community detection was not the main focus in \cite{pei2021generalization,pan2018arga,park2019symmetric,aaai20,huang2019rwr}, these works all proposed various encoding-decoding methods that, to different extents, seem to outperform standard GAE and VGAE models on the community detection task, in the reported evaluations. They consider alternatives encoders or training choices, which we further discuss and investigate~in~Section~\ref{c7s73}.

\subsection{Current Limitations}
\label{c7s723}

While the models discussed in Section \ref{c7s722} will constitute relevant baselines in our experiments (see Section~\ref{c7s74}), they still suffer from several fundamental limitations that motivate our work:
\begin{itemize}
    \item firstly, all extensions explicitly designed for community detection \cite{choong2018learning,choong2020optimizing,li2020dirichlet} \textit{rely on clustering-oriented probabilistic priors}. They are only applicable in the VGAE paradigm, and cannot be directly transposed to the deterministic GAE setting. The question of how to design clustering-efficient GAE models thus remains widely open;
    
    \item more importantly, a closer look at these models reveals that their \textit{empirical gains often mostly stem from the addition of node features} to the graph. As an illustration, Table~\ref{tab:vgaecd} displays the reported performances of VGAECD and VGAECD-OPT on the \textit{featureless} versions of two graphs \cite{choong2020optimizing}. We observe that they offer little to no empirical advantage when features are absent. This draws into question the extent to which these models are able to capture communities from~(only)~graph~data.

  \begin{table}
      \caption[Community detection with VGAECD and VGAECD-OPT]{Normalized mutual information scores (in \%) for community detection on the Cora and Pubmed citation networks, \textit{with} and \textit{without} node features. Results are directly taken from the evaluation of Choong et al.~\cite{choong2020optimizing}. \textbf{Bold} numbers correspond to the best score for each graph. This table emphasizes that, in the absence of node features, VGAECD and VGAECD-OPT bring little (to no) advantage w.r.t. standard VGAE, and remain below the Deepwalk and/or Louvain baselines. Scores of VGAECD and VGAECD-OPT significantly increase when adding features to the graph. Recall: in this table, Deepwalk and Louvain both ignore node features.}
       \vspace{0.2cm}
    \label{tab:vgaecd}
    \centering
   \resizebox{\textwidth}{!}{
  \begin{tabular}{c|c|cc|ccc|cc}
     \textbf{Dataset} & \textbf{VGAE} & \textbf{VGAECD} & \textbf{VGAECD-OPT} & \textbf{DeepWalk} & \textbf{Louvain} \\
    \midrule \midrule
    Cora \textit{without} node features & 23.84 & 28.22 & 37.35 & 37.96 & \textbf{43.36} \\

    Pubmed \textit{without} node features & 20.41 & 16.42 & 25.05 & \textbf{29.46} & 19.83 \\
\midrule
    Cora \textit{with} node features& 31.73 & 50.72 & \textbf{54.37} & 37.96 & 43.36\\
    Pubmed \textit{with} node features & 19.81 & 32.53 & \textbf{35.52} & 29.46 & 19.83\\
    
    \bottomrule
  \end{tabular}}

    \vspace{-0.1cm}
\end{table}

    The important role of node features has subsequently been confirmed (e.g., Park et al.~\cite{park2019symmetric} show that, on the Pubmed dataset, a straightforward $k$-means on the node features alone reaches comparable AMI scores w.r.t. VGAE and MGAE). On the other hand, most of the aforementioned other studies with empirical improvements \cite{pei2021generalization,mrabah2021rethinking,pan2018arga,park2019symmetric,aaai20,huang2019rwr} only reported results on graphs equipped with node features. This motivates the need for a proper investigation of the \textit{featureless} case where models cannot rely on the additional node feature information; 

    \item lastly, previous studies centered around community detection \cite{choong2018learning,choong2020optimizing,li2020dirichlet,wang2017mgae} \textit{did not explicitly try to preserve the good performances of GAE and VGAE models on link prediction}. Overall, most of the aforementioned existing works learn node representations specific to a particular learning task. Therefore, it is still unclear whether one can improve community detection with these models without simultaneously deteriorating link prediction.
    With the general aim of learning high-quality node embedding representations, one can wonder to which extent these models can learn representations that are \textit{jointly} useful for several tasks. Besides providing a more accurate summary of the graph structure under consideration, such representations could also lead to significant resource savings in real-world applications, as we will illustrate in Section~\ref{c7s74}.
\end{itemize}
In conclusion to this section, the question of how to effectively improve community detection with GAE and VGAE remains incompletely addressed.

\section{Modularity-Aware Graph Autoencoders}
\label{c7s73}

\begin{figure}[t]
    \centering
    \includegraphics[width=1.0\textwidth]{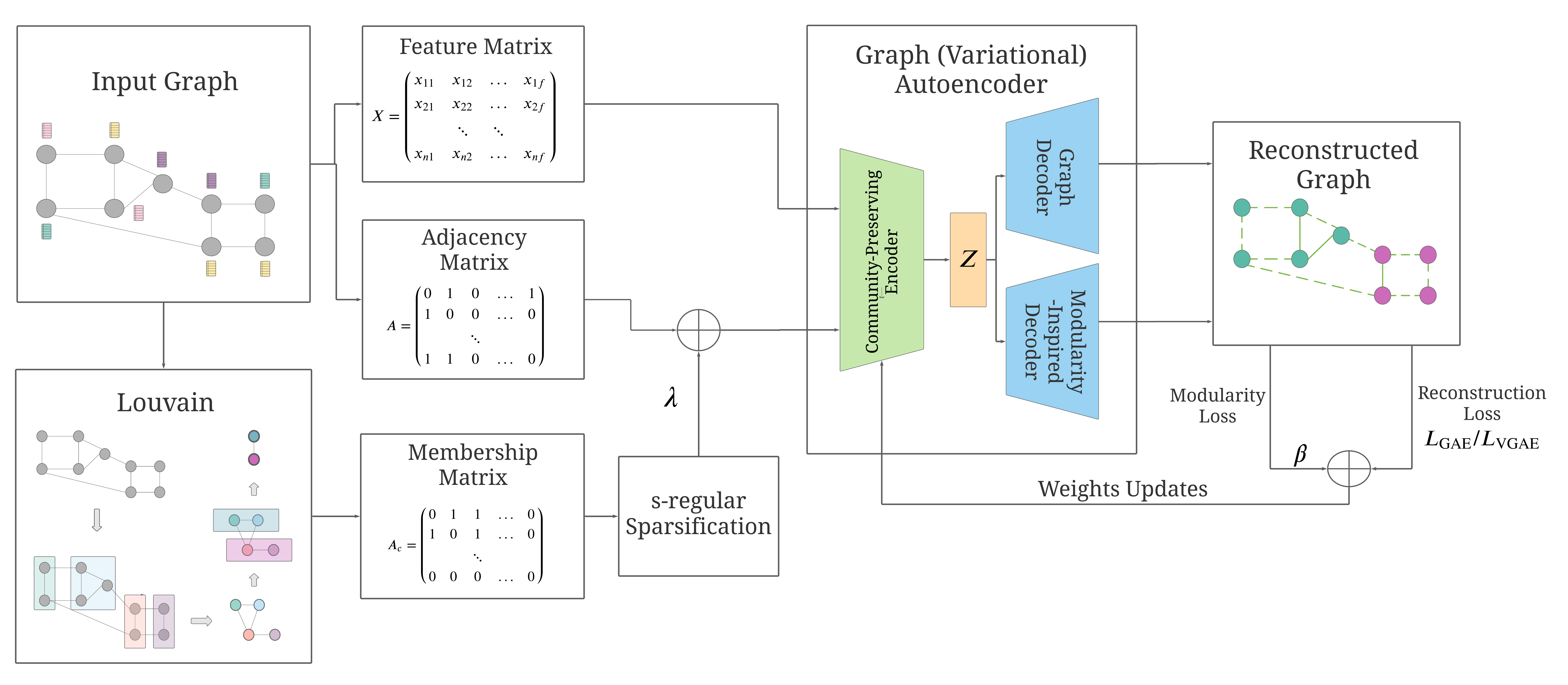}
    \caption[Overview of our Modularity-Aware GAE and VGAE]{Overview of our proposed Modularity-Aware GAE and VGAE. Firstly, input graph data $A$ and $X$ are combined with the $\dregc$-regular sparsified prior community membership matrix $\Ao$, derived through iterative modularity maximization via the Louvain algorithm, as described in Section~\ref{c7s732}. Then, they are processed by our revised community-preserving (linear or multi-layer GCN) encoders, encoding each node $i \in \mathcal{V}$ as an embedding vector $z_i$ of dimension $d \ll n$. Neural weights of encoders are optimized through a procedure combining reconstruction and modularity-inspired losses, and described in Section~\ref{c7s733}. Furthermore, other hyperparameters from this model are tuned via the method described in Section~\ref{c7s733} and designed for joint link prediction and community detection applications.}
    \label{fig:magae_c7}
\end{figure}

We now introduce our approach, referred to as \textit{Modularity-Aware GAE and VGAE} in the following, to address the aforementioned limitations.
In Section~\ref{c7s731}, we first provide a general overview of the key components of our solution. They transpose concepts from \textit{modularity-based} clustering \cite{blondel2008louvain,brandes2007modularity,shiokawa2013fast} to GAEs and VGAEs, and are illustrated in Figure~\ref{fig:magae_c7}. 
We subsequently detail these solution components in Sections~\ref{c7s732}~and~\ref{c7s733}.

\subsection{Diagnostic and Overview of our Framework}
\label{c7s731}

Based on our literature review, we diagnose three main reasons that can explain why previous GAE and VGAE models still suffer from the limitations described in Section~\ref{c7s723}:
\begin{itemize}
    \item firstly, they leveraged \textit{encoders} that were not specifically designed to preserve the intrinsic communities from the graph structure under consideration in the node embedding space. This includes the popular GCN, as well as refined neural models that rather aimed to preserve clusters from node features (but not necessarily the actual communities from the graph under consideration). In Modularity-Aware GAE and VGAE, we overcome this issue by incorporating a novel encoding scheme for graph community-preserving representation learning. It consists in an improvement of the GCN \textit{message passing operator}, boosting both GAE and VGAE models by simultaneously considering the initial graph structure and \textit{modularity-based node communities} when computing node embedding spaces. We present and theoretically study this encoder in Section~\ref{c7s732};
    
    \item besides the encoder's architecture, previous models were often \textit{optimized} in a fashion that, by design, favors link prediction over community detection. In particular, the standard cross-entropy and ELBO losses, used to learn neural weight matrices, directly involve the reconstruction of \textit{node pairs} from the embedding space\footnote{In the case of the probabilistic VGAE paradigm, another limitation of the ELBO loss - and of the underlying generative decoder - lies in the use of standard Gaussian priors. Replacing these priors by for example \textit{Gaussian mixtures} as in \cite{choong2018learning,choong2020optimizing}, appears to be an intuitive approach for community-based learning. However, as this approach 1) does not extend to deterministic GAE, and 2) has been extensively studied in \cite{choong2018learning,choong2020optimizing}, we do not further develop it in this work. We will nonetheless compare to \cite{choong2018learning,choong2020optimizing} in experiments, and will argue in Section~\ref{c7s733} that Gaussian mixtures could straightforwardly be incorporated in our proposed~Modularity-Aware~VGAE.}. However, as we will detail, a good reconstruction of \textit{local} pairwise connections does not necessarily imply a good reconstruction of the \textit{global}~community~structure. 
    In Modularity-Aware GAE and VGAE, we instead optimize an alternative loss inspired by the \textit{modularity} \cite{blondel2008louvain}. Such a loss acts as a simple yet effective global regularization over pairwise reconstruction losses, with desirable properties for joint link prediction and community detection. It will empirically enable a refined optimization of the weight matrices from our encoders. We present this aspect in detail~in~Section~\ref{c7s733};
    \item lastly, in addition to these weight matrices, GAE and VGAE models involve several other hyperparameters, ranging from the number of training iterations to the learning rate \cite{kipf2016-2}. While they also impact the model's performance, the selection procedure for such hyperparameters was sometimes omitted in previous works \cite{choong2020optimizing} or based on link prediction validation sets, as in experiments from our previous chapters (while, intuitively, the best hyperparameters for community detection might differ from those for link prediction).   
    For the Modularity-Aware GAE and VGAE, we adopt an alternative graph-based model selection procedure. It completes the previous two aspects, by providing the most relevant GAE/VGAE hyperparameters for joint link prediction and community selection. We present and discuss this procedure in Section~\ref{c7s733} as well.
    \end{itemize}

\subsection{Community-Aware Encoders for GAE and VGAE}
\label{c7s732}

Following this diagnosis and overview, we now detail the first of the three bullet points from Section \ref{c7s731}, i.e., our proposed revised \textit{encoding}~strategy. We recall that our proposed solution aims to encode nodes as embedding vectors $z_i$  \textit{more suitable for community detection}. Essentially, intrinsic communities in the graph under consideration should be easily retrievable from these representations, e.g., from their $L_2$ distances via a straightforward $k$-means clustering. These vectors should also simultaneously remain relevant for \textit{link prediction}, i.e., as for existing GAE and VGAE models, the likelihood of a missing edge between two nodes should also be inferred from the learned representations $z_i$. 

\paragraph{Preliminary remark on the notation} In the following, for simplicity, we mainly consider \textit{undirected} graphs and continue using the symmetric inner product $\hat{A}_{ij} = \sigma(z^T_i z_j)$ as the probability of an edge between nodes $i$ and $j$.
Also, in this chapter, we will use $\agcn{\cdot}$ to denote the symmetrically normalized adjacency matrix of a graph, i.e.:
\begin{equation}
\agcn{A} = (D+I_n)^{-\frac{1}{2}}(A+I_n)(D+I_n)^{-\frac{1}{2}
\label{eq:norm}
}\end{equation}
instead of $\tilde{A}$ as in Definition~\ref{def:norm_c2} and other chapters of the thesis. This slight discrepancy is due to the fact that, in this chapter, the symmetric normalization will be simultaneously applied to several matrices and to sums of matrices, which makes the \textit{``tilde''} notation too heavy.

\subsubsection{Revising the Message Passing Operator}
Existing graph encoders usually involve normalized versions of the adjacency matrix~$A$, or some generalized \textit{message passing operator} matrix that also captures each node's direct connections in the graph under consideration \cite{Dasoulas2021}. For instance, in the popular multi-layer GCN from Definition~\ref{def:gcn}, the symmetric normalization $\tilde{A}$ a.k.a. $\agcn{A}$ from Definition~\ref{def:norm_c2} is used such that at each layer $l$ a vectorial representation for each node is computed by averaging the representations from layer
$l - 1$ of its direct neighbors and of itself. In this chapter, we adopt an alternative strategy that consists in averaging, at each layer:
\begin{itemize}
    \item representations from the direct neighbors of each node, as above;
    \item but also representations from other \textit{unconnected nodes} that, according to some prior available knowledge and criteria, belong to the same graph community. 
\end{itemize}
More precisely, let us assume that we have, at our disposal, a pre-processing \textit{graph mining} technique that, based on the graph structure and on some fixed criteria, learns an initial \textit{prior partition of the node set} $\mathcal{V}$ into $\ncluster$ sets $C_1, \ldots, C_\ncluster,$ with $|C_\clustersubscript| = n_\clustersubscript$ for $\clustersubscript \in \{1, \ldots, \ncluster\}.$ Here, $\ncluster$ acts as a hyperparameter, that can differ from the actual number of communities eventually used for the community detection downstream evaluation task (i.e., the number of clusters in the $k$-means operated on the final vectors $z_i$). A concrete example of such a technique will be provided in the upcoming Subsection~\ref{s323}. We simply assume its availability at this stage.

We propose to leverage such an initial partition as a \textit{prior node clustering signal} from which the GAE/VGAE encoder should benefit, but also can deviate during training, when learning the embedding space. Specifically, we propose to replace the standard input adjacency matrix $A$~by:
\begin{equation}
A + \lambda \Ac,   
\end{equation}
where $\lambda \geq 0$ is a scalar hyperparameter, and where $\Ac$ is the community membership matrix defined as follows:

\begin{definition}\label{def:Ac}
Let us consider a partition of the node set $\mathcal{V}$ into $\ncluster$ sets $C_1, \ldots, C_\ncluster$. The corresponding \textit{community membership matrix} is defined as:
\begin{equation}
A_c=MM^T-I_n,
\label{eqAc}
\end{equation}
 with $M\in\{0,1\}^{n\times \ncluster}$ denoting the $n\times \ncluster$ matrix where elements $M_{i\clustersubscript}=1$ if and only if $i\in C_\clustersubscript$ according to the prior clustering.
\end{definition}
We interpret $\Ac$ as the adjacency matrix of an alternative graph in which each cluster of our prior partition is represented by a fully connected graph, without self-loops. Since nodes are only allocated to one cluster, there exists a node ordering such that the matrix $\Ac$ is block-diagonal. In essence, $A + \lambda \Ac$ aims to capture refined node similarities, by simultaneously considering some \textit{local} information from direct neighborhoods, and some \textit{global} information from prior node communities. The hyperparameter $\lambda$ helps to balance these two aspects. In particular, setting $\lambda = 0$ results in the standard adjacency matrix.

\subsubsection{From Message Passing Operators to Encoding Schemes}
At first glance, $A + \lambda \Ac$ could straightforwardly be incorporated as a refined message passing operator in popular GAE and VGAE encoders. For instance, one could consider its direct incorporation in:
\begin{itemize}
    \item variants of \textit{2-layer GCN encoders}, as initially proposed by Kipf and Welling \cite{kipf2016-2}, since this neural architecture remains the most popular GAE/VGAE encoder in the literature (see Chapters~\ref{chapter_2} and~\ref{chapter_6} for a review). Specifically, one could consider:
    \begin{itemize}
        \item a version incorporating $A + \lambda \Ac$ in both layers. Then, in the case of a GAE\footnote{For clarity of exposition we discuss the deterministic GAE framework (Section~\ref{c2s241}). However, the changes are equally applicable to the VGAE framework (Section~\ref{c2s242}), for which $Z$ has to be replaced by $\mu$ and $\log \sigma$ as in Equation~\eqref{vaeencoder}.} as described in Section~\ref{c2s241}, we would have: $Z = \text{GCN}^{(1)}(A+\weight\Ac, X) = \agcn{A+\weight\Ac} \text{ReLU} (\agcn{A+\weight\Ac} X W^{(0)}) W^{(1)}$;
        \item a version incorporating the prior communities only on the first layer, i.e.,  $Z = \text{GCN}^{(2)}(A+\weight\Ac, X) = \agcn{A} \text{ReLU} (\agcn{A+\weight\Ac} X W^{(0)}) W^{(1)}$;
    \end{itemize}
    \item or, a variant of the \textit{linear encoder} proposed in Chapter~\ref{chapter_6}, as this simplified one-hop GCN without activation reached competitive performances w.r.t. multi-layer GCNs for GAE/VGAE-based community detection in this previous Chapter~\ref{chapter_6} as well as in the study of Choong et al.~\cite{choong2020optimizing}. In this case: $Z = \text{Linear}(A+\weight\Ac, X) =  \agcn{A+\weight\Ac} X W^{(0)}$.
\end{itemize}

However, the computational cost of evaluating each layer of a GCN or a linear encoder depends linearly on the number of edges $|\mathcal{E}| = m$ in the message passing operator \cite{kipf2016-1}. As the graph represented by $A+\weight \Ac$ contains at least $\sum_{\clustersubscript=1}^\ncluster n_\clustersubscript^2$ edges, such a direct incorporation of $A+\weight \Ac$ in encoders could incur a large computational expense. 

To alleviate this cost, we will instead consider a \textit{$\dregc$-regular sparsification of $\Ac$}, denoted by $\Ao$ in the following. In $\Ao$, each node $i \in C_\clustersubscript$ is only connected to $\dregc < n_\clustersubscript$ randomly selected nodes in $C_\clustersubscript$ (instead of all other nodes in $C_\clustersubscript$). Therefore, the $A + \lambda \Ao$ message passing operator still contains some of the prior clustering information without necessarily incurring 
the cost implied by the use of $\Ac$. In particular, selecting $\dregc \approx \frac{m}{n}$ ensures that $A + \lambda \Ao$ has $O(2m)$ non-null elements, preserving the linear complexity w.r.t. $m$ of the aforementioned encoders. Note that we only sample $\Ao$ once at the beginning of the model training and then keep it fixed throughout training and testing. To sum up, in our upcoming experiments in Section~\ref{c7s74} we will instead consider the following two\footnote{We will favor  $\text{GCN}^{(2)}$ over $\text{GCN}^{(1)}$ in the remainder of this work, as the former outperformed the latter in our experiments. To simplify the notation $\text{GCN}^{(2)}$ will be referred to as GCN~in~experiments.} encoding schemes:
\begin{itemize}
    \item $Z = \text{GCN}^{(2)}(A+\weight \Ao, X) = \agcn{A} \text{ReLU} (\agcn{A+\weight \Ao} X W^{(0)}) W^{(1)}$;
    \item $Z = \text{Linear}(A+\weight \Ao, X) = \agcn{A+\weight \Ao} X W^{(0)}$.
\end{itemize}

In these encoders, our altered message passing scheme allows practitioners to incorporate information from prior communities in the resulting node embedding space. A given node $i \in \mathcal{C}_\clustersubscript \subset \mathcal{V}$, for $\clustersubscript \in \{1,...,\ncluster\},$ will aggregate information from its direct neighbors and from some nodes in  $\mathcal{C}_\clustersubscript$. By design, $i$ will thus have an embedding vector $z_i$  more similar to the embedding vectors of the other nodes in $\mathcal{C}_\clustersubscript$ than would be the case for the standard encoders based on $\agcn{A}$.
We recall that the choice of linear and 2-layer GCN encoders is made without loss of generality. $A + \lambda  \Ao$ could be incorporated into other encoders including deeper GCNs, ChebNets \cite{defferrard2016} or graph attention networks \cite{velivckovic2019graph}.

The remainder of this Section~\ref{c7s732} on encoders is organized as follows. In Subsection~\ref{s323} we detail how we derive the matrix $\Ac$ (that has loosely been assumed to be ``available'' so far) in our work. Then, in Subsection~\ref{s324}, we provide a theoretical analysis of our novel encoding strategy. It notably aims to better understand our newly introduced operators $\Ac$ and $\Ao$ in terms of the spectral filtering they induce, as well as to assess the impact of the $\dregc$-regular sparsification of $\Ac.$

\subsubsection{Learning $\Ac$ and $\Ao$ with Modularity-Based Clustering}
\label{s323}

So far, for pedagogical purposes, we loosely assumed the availability of the $\Ac$ and $\Ao$ prior community membership matrices. In practice, how these matrices are \textit{learned} plays an important role, as the empirical performance of our strategy will directly depend on the quality of the underlying prior node clusters. Throughout this chapter, we will rely on \textit{modularity} concepts to learn $\Ac$ - hence the name \textit{Modularity-Aware GAE and VGAE}. Specifically, we will leverage the popular \textit{Louvain}~algorithm~\cite{blondel2008louvain}.

In the absence of node feature information, the Louvain greedy algorithm remains a popular and powerful approach for community detection \cite{blondel2008louvain}.
It iteratively aims to maximize the \textit{modularity} value \cite{Newman8577}, defined as follows:

\begin{definition}\label{def:modularity}
Let us consider a graph $\mathcal{G} = (\mathcal{V}, \mathcal{E})$ with adjacency matrix $A$ and nodes $i\in \mathcal{V}$ of degree $d_i = D_{ii} = \sum_{j=1}^n A_{ij}$. We denote a partition of these nodes into $\ncluster \leq |\mathcal{V}|$ communities by $\{C_1,...,C_\ncluster\}.$ Then, the \textit{modularity} associated with this partition is:
\begin{equation}\label{eq:modularity}
Q = \frac{1}{2m} \sum_{i,j=1}^n \left[A_{ij} - \frac{d_id_j}{2m}\right] \delta(i,j),
\end{equation}
where $m$ is the sum of all edge weights in the graph, i.e., the number of edges for unweighted graphs, and where $\delta(i,j) = 1$ if nodes $i$ and $j$ belong to the same community and $0$ otherwise. 
\end{definition}
In essence, the modularity compares the density of connections inside communities to connections between communities. More specifically, Equation \eqnref{eq:modularity} returns a scalar $Q$ in the range $[-\frac{1}{2}, 1]$, that measures the difference between the observed fraction of (potentially weighted) edges that occur within the same community and the expected fraction of edges in a configuration model graph, which matches our observed degree distribution but allocates edges randomly without any specified community structure.

\paragraph{The Louvain algorithm} In the Louvain greedy algorithm, aiming to maximize the modularity of a graph over the set of possible cluster assignments, each node is initialized in its own community. Then, the algorithm iteratively completes two phases:

\begin{itemize}
    \item in phase 1, for each node $i \in \mathcal{V}$, one computes changes in modularity occurring from placing $i$ into the community of each of its neighbors. Then, $i$ is either placed into the community leading to the greatest modularity increase, or remains in its original group if no increase is possible;
\item in phase 2, a new graph is constructed. Nodes correspond to communities obtained in phase 1 and edges are formed by summing edge weights occurring between communities. Edges within a community are represented by self-loops in this new graph. One repeats phase 1 on this new graph until no further modularity improvement is possible.
\end{itemize}

Our justification for the use of this method to derive $\Ac$ is threefold. First and foremost, it automatically selects the relevant number of prior communities $\ncluster$, by iteratively maximizing the modularity value. Secondly, it runs in $O(n\log n)$ time  \cite{blondel2008louvain}. It therefore scales to large graphs with millions of nodes. Thirdly, such a modularity criterion complements the encoding-decoding paradigm of standard GAE and VGAE models. We argue that learning node embedding spaces from complementary criteria is beneficial. Our experiments will confirm that leveraging prior modularity-based node clusters in the GAE/VGAE outperforms the individual use of the Louvain or of the GAE/VGAE \textit{alone}. 

Note that, the use of the Louvain method is made without loss of generality as our framework remains valid for alternative graph mining methods deriving $\Ac$ and $\Ao$.

\subsubsection{Theoretical Analysis of the Encoder's Message Passing Operator}
\label{s324}

We now conduct a theoretical analysis of our newly introduced message passing operator, which we begin by motivating the spectral analysis of the matrices involved.
We recall that the computations performed by a GCN at a given layer are the following:
\begin{equation} \label{eq:GCN_layer}
    \text{ReLU} (\agcn{A} H^{(l-1)} W^{(l-1)}).
\end{equation}
If we consider the spectral decomposition of the message passing operator that is used in Equation \eqnref{eq:GCN_layer}, $\agcn{A} = U\Theta U^T,$ where $U=[u_1, \ldots, u_n]^T$ denotes the matrix containing the eigenvectors $u_i$ of $\agcn{A}$ and $\Theta$ is a diagonal matrix containing the eigenvalues $\theta_i$ of $\agcn{A}.$ Then, the computation performed in Equation \eqnref{eq:GCN_layer} can be reformulated as follows:
\begin{equation} \label{eq:GCN_spectralc7}
    \text{ReLU} (U\Theta U^T H^{(l-1)} W^{(l-1)}) = \text{ReLU} \left(\sum_{i=1}^n \theta_i u_iu_i^T H^{(l-1)} W^{(l-1)}\right).
\end{equation}
Therefore, performing one message passing step of the hidden states $H$ on a graph given by $\agcn{A},$ i.e., $\agcn{A} H^{(l-1)},$ can be interpreted as a Fourier transform of $H,$ called \textit{graph Fourier transform} \cite{Shuman2013}, where the eigenvectors of $\agcn{A}$ act as a Fourier basis and the eigenvalues of $\agcn{A}$ define the Fourier coefficients. 

When trying to perform a theoretical analysis of the message passing step in Equation~\eqnref{eq:GCN_layer} it often turns out to be more insightful to consider Equation \eqref{eq:GCN_spectralc7} instead and analyze the eigenvalues and eigenvectors of the used message passing operator. Such a spectral perspective has 
given rise to a variety of architectures proposing learnable functions applied to the diagonal terms of $\Theta$ \cite{bruna2013spectral,defferrard2016,Levie2019}. 
Historically, the study of spectral graph theory \cite{Chung1997,spielman2007spectral}, and in particular the area of graph signal processing \cite{Ortega2018,Sandryhaila2014}, has yielded much insight in the study of graphs.  Therefore, it is somewhat unsurprising that also in the study of the GNNs the spectral analysis of these architectures is a promising avenue of analysis \cite{Balcilar2021,Dasoulas2021,gama2020}. 

Therefore, we now provide spectral results allowing us to gain a better understanding of our proposed message passing operator and compare our proposed message passing operator to the standard message passing operators.  To characterize the eigenvectors of our newly introduced $\agcn{\Ac}$ we rely on the concept of 2-sparse eigenvectors. 

\begin{definition} (from Teke and Vaidyanathan~\cite{Teke2017})
The entries of \textit{2-sparse eigenvectors} are all equal to $0$ except for the $i^{\mathrm{th}}$ and $j^{\mathrm{th}}$ entry which equal to $1$ and $-1,$ where $i$ and  $j$ denote two nodes which share all their neighbors, i.e., $A_{ih}=A_{jh}$ for $h\in\{1,\ldots,n\}\backslash\{i,j\}.$ 

\end{definition}
An extended discussion of the literature related to such 2-sparse eigenvectors and their corresponding vertices, which are sometimes referred to as twin vertices, can be found in the thesis of Lutzeyer~\cite{Lutzeyer2020}. 
We are now able to characterize the spectrum and eigenvectors of $\agcn{\Ac}.$

\begin{proposition} \label{thm:Acspectrum}
The matrix $\agcn{\Ac}$ has eigenvalues $\{\{1\}^{\ncluster}, \{0\}^{n-\ncluster}\},$ where we denote the multiset containing a given element $x,$ $y$ times, by $\{x\}^y.$  Each non-zero eigenvalue has an associated eigenvector $v_\clustersubscript,$ with $\clustersubscript \in \{1, \ldots, \ncluster\},$  with entries $(v_\clustersubscript)_i=1$ for $i \in C_\clustersubscript$ and $(v_\clustersubscript)_i=0$ for $i \not\in C_\clustersubscript.$ 
The eigenspace corresponding to the zero eigenvalue has dimension $n-\ncluster$ and is spanned by, for example, a set of two-sparse eigenvectors on each of the connected components in the graph.

\end{proposition}


The proof of Proposition \ref{thm:Acspectrum} can be found in Section~\ref{c7s76}. 
The informal take-away from Proposition \ref{thm:Acspectrum} is that \textit{the cluster membership of nodes is encoded clearly and compactly in the spectrum and eigenvectors of $\agcn{\Ac}.$ }
More formally, in Proposition \ref{thm:Acspectrum} we observe that in the spectral domain the operator $\agcn{A_c},$ which we introduce to the encoder's message passing scheme, directly encodes the cluster membership of the different nodes and all other signals are filtered out by the $0$ eigenvalues. Also in the graph domain the matrix $\agcn{\Ac}$ encodes the cluster structure by representing each cluster by a fully connected component of the graph. Therefore, the matrix $\agcn{\Ac}$ is an appropriate choice to introduce cluster information into the message passing scheme and does so clearly in both the graph and spectral domains.

In general, the spectral filtering steps performed by our message passing operator, $\agcn{A +\weight \Ac},$ and the standard message passing operator, $\agcn{A},$ are different. $\agcn{A +\weight \Ac}$ accounts for the clustering information which we introduce. In the following theorem, we provide a result allowing us to establish under which conditions the spectral filtering performed by $\agcn{A}$ and $\agcn{A +\weight \Ac}$ are equal, to gain a better understanding of the action of $\agcn{A +\weight \Ac}.$

\begin{proposition}\label{thm:spectral_relation_A_Ac_AAc}
If $\mathcal{G}$ is composed of regular connected components, i.e., connected components containing only vertices of equal degree, and the partition of the node set defining these regular components equals the partition defining $\Ac,$ then the matrices $\agcn{\Ac}$ and $\agcn{A+\weight \Ac}$ have a shared set of eigenvectors and the spectrum of $\agcn{A+\weight \Ac},$ denoted by $\mathcal{S}(\agcn{A+\weight \Ac}),$ can be expressed in terms of the eigenvalues of $\agcn{A}$ and $\agcn{\Ac},$ denoted by $\theta$ and $\eta$, respectively,~as: 
\begin{equation}
\mathcal{S}(\agcn{A+\weight \Ac}) = \{g_1(\theta_1) + g_2(\eta_{s(1)}), \ldots, g_1(\theta_n) + g_2(\eta_{s(n)})\},
\end{equation}
for affine functions $g_1, g_2$ parameterized by the node degrees and some permutation $s(\cdot)$ defined on the set $\{1, \ldots, n\}.$
\end{proposition}


The proof of Proposition \ref{thm:spectral_relation_A_Ac_AAc} can be found in Section~\ref{c7s76} as well. 
Hence, we observe that for graphs consisting of regular connected components the spectral filtering performed by our proposed message passing operator $\agcn{A+\weight \Ac}$ is equal to that of the standard operator $\agcn{A}.$ For graphs consisting of regular connected components the clustering information is already contained in the spectrum of $\agcn{A}$ and therefore its further addition does not affect the eigenvectors of our proposed message passing operator. 

Note that in general the spectrum of the sum of two matrices cannot be characterized by the individual spectra of the two matrices, meaning that, in general, there does not exist an exact relation between the spectra of $\agcn{A}$ and $\agcn{\Ac}$ to $\agcn{A+\lambda\Ac}.$ We can, however, make direct use of existing results such as Weyl's inequality \cite{Weyl1912} and the extended Davis--Kahan theorem \cite{Lutzeyer2019}, which, respectively, upper bound the distance of the eigenvalues and spaces spanned by the eigenvectors of the sum of matrices and the individual matrices. 


\paragraph{$\dregc$-regular sparsification}

We now turn to the analysis of our sparsified message passing operator  
$\Ao,$ which, as we will see now, still contains the external cluster information without incurring the large computation cost implied by the use of $\Ac.$ 

\begin{proposition} \label{thm:Aospectrum}
If the partition defining the connected components of $\Ao$ is a refinement of the partition defining the components of $\Ac$,
then the multiplicity of the largest eigenvalue of $\agcn{\Ao}$ is greater or equal to the multiplicity of the largest eigenvalue of $\agcn{\Ac}.$ Further, the largest eigenvalue of both $\agcn{\Ao}$ and $\agcn{\Ac}$ equals 1 and the eigenvectors corresponding to the eigenvalue 1 of $\agcn{\Ac}$ are also eigenvectors corresponding to the eigenvalue 1 of $\agcn{\Ao}.$

\end{proposition}


The proof of Proposition \ref{thm:Aospectrum} can be found in Section~\ref{c7s76}. 
Informally, Proposition \ref{thm:Aospectrum} can be interpreted to show that \textit{the sparsification of $\Ac$ producing $\Ao$ does not impact the ``informative'' part of the spectrum. } 
We recall that the eigenvectors corresponding to the largest eigenvalue of $\agcn{\Ac}$ and $\agcn{\Ao}$ are indicator vectors of our introduced cluster membership. Since the remaining eigenvectors are orthogonal to these indicator vectors we know that none of them encode our cluster membership as compactly as the eigenvectors corresponding to the~largest~eigenvalue. 

For $\agcn{\Ao}$ the eigenvalues corresponding to the less informative eigenvectors correspond to nonzero eigenvalues in general and we expect the choice of $\dregc$ to influence the impact of this uninformative part of the spectrum. This uninformative part of the spectrum can be upper bounded by adapting the bound by Friedman~\cite{Friedman2003} to our message passing operator. However, in our work, we choose a more practice-oriented approach by treating $\dregc$ as a hyperparameter of our model and find its optimal values using the procedure which is described in Section~\ref{c7s7332}.

\subsection{Modularity-Inpired Loss and Training Strategy}
\label{c7s733}

So far, our work considered improvements of the encoder's \textit{architecture}. While this aspect is crucial, we also argue that previously proposed models were often \textit{optimized} in a fashion that, by design, favors link prediction over community detection. With this in mind, this Section~\ref{c7s733} now complements our contributions from Section~\ref{c7s732} with revised training and optimization~strategies.

\subsubsection{Modularity-Inspired Losses for GAE and VGAE}
\label{c7s7331}

As explained in Section~\ref{c2s24} from Chapter~\ref{chapter_2}, neural weight matrices of standard GAE and VGAE encoders were tuned by optimizing \textit{reconstruction losses/objectives}, capturing the similarity between the decoded graph and the original one. Usually, these losses directly evaluate the quality of reconstructed node pairs $\hat{A}_{ij}$ w.r.t. their ground truth counterpart $A_{ij}$. This includes the cross-entropy loss $\lossGAE$ from Equation~\eqref{lossGAE} and the ELBO loss $\lossVGAE$ from Equation~\eqref{elbo}. We argue that this optimization strategy also contributes to explaining the underwhelming performance of some GAE and VGAE models on community detection~tasks:
\begin{itemize}
    \item by design, existing optimization strategies favor good performances on link prediction tasks, that precisely consist in accurately reconstructing connected/unconnected node pairs. However, some recent studies emphasized that a good reconstruction of \textit{local} pairwise connections does not always imply a good reconstruction of the \textit{global} community structure from the graph under consideration \cite{liu2019much,wang2017community}. This motivates the need for a revised loss function capturing some global community information;
    \item besides, GAE/VGAE-based community detection experiments often consisted in running $k$-means algorithms in the final node embedding space (and, as stated at the beginning of Section~\ref{c7s732}, we also adopt this strategy). However, this results in clustering embedding vectors based on their $L_2$ \textit{distances} $\Vert z_i - z_j \Vert_2 $, whereas the aforementioned reconstruction losses instead often involve \textit{inner products} ($\hat{A}_{ij} = \sigma(z^T_i z_j)$). There is thus a discrepancy between the criterion ultimately used for $k$-means clustering, and the one used during training to assess node similarities.
\end{itemize}
To address these issues, we propose to complement standard GAE and VGAE losses with an additional loss term, involving $L_2$ distances and inspired by the \textit{modularity}\footnote{We emphasize that this new term does \textit{not} involve the prior Louvain clusters used in $\Ac$.} from Equation~\eqref{eq:modularity}. In the case of the GAE, we will iteratively minimize by gradient descent:
\begin{equation}
 \lossMAGAE = \lossGAE - \frac{\beta}{2m} \sum_{i,j=1}^n \left[A_{ij} - \frac{d_id_j}{2m}\right]  e^{-\gamma \Vert z_i - z_j \Vert^2_2},
 \label{eq:tildeAE}
 \end{equation}
with hyperparameters $\beta \geq 0$, $\gamma \geq 0.$ Also, for VGAE we will maximize\footnote{We recall that $\lossGAE$ is \textit{minimized} while $\lossVGAE$ is \textit{maximized}, hence the occurrence of a \textit{minus} term in Equation \eqref{eq:tildeAE} but a \textit{plus} term in Equation \eqref{eq:tildeVAE}.}:
\begin{equation}
\lossMAVGAE = \lossVGAE + \frac{\beta}{2m} \sum_{i,j=1}^n \left[A_{ij} - \frac{d_id_j}{2m}\right] e^{-\gamma \Vert z_i - z_j \Vert^2_2}. \label{eq:tildeVAE}
\end{equation}

In Equations \eqref{eq:tildeAE} and \eqnref{eq:tildeVAE}, the exponential term (taking values in $[0,1]$) acts as a \textit{soft counterpart of the common community indicator} $\delta(i,j) \in \{0,1\}$ from Equation~\eqref{eq:modularity}. It tends to 1 when nodes $i$ and $j$ get closer in the embedding space, and tends to 0 when they move apart. 

In essence, we expect the addition of such a \textit{global regularizer} to $\lossGAE$ and $\lossVGAE$ to encourage closer embedding vectors (in the $L_2$ distance) of densely connected parts of the original graph, and therefore to permit a  $k$-means-based \textit{detection of communities with higher modularity values}. On the other hand, the remaining presence of the original $\lossGAE$ or $\lossVGAE$ term\footnote{Our experiments will consider the original $\lossGAE$ or $\lossVGAE$ from Equations \eqref{lossGAE} and \eqref{elbo} and originally formulated by Kipf and Welling \cite{kipf2016-2}. Nonetheless, one can observe that our modularity-inspired global regularizer term could be optimized \textit{in conjunction with other reconstruction losses.} For instance, modularity-inspired terms could be added to the variant formulation of ELBO loss from Choong et al. \cite{choong2018learning}, that incorporates \textit{Gaussian mixtures} in the Kullback-Leibler divergence.} in the loss aims to \textit{preserve good performances on link prediction}. The hyperparameter $\beta$ balances the relative importance of the modularity regularizer w.r.t. the pairwise node pairs reconstruction loss, while the hyperparameter $\gamma$ regulates the magnitude of $\Vert z_i - z_j \Vert^2_2$ in the exponential term. Our experiments will show that proper tuning of $\beta$ and $\gamma$ permits us to improve community detection while jointly preserving performances on link prediction.

The use of a modularity-inspired regularizer in the loss of the Modularity-Aware GAE and VGAE builds upon several studies, which were not studying the GAE/VGAE frameworks but emphasized the benefits of various modularity-inspired losses for learning community-preserving node embedding representations \cite{lobov2019unsupervised,wang2017community,yang2016modularity}. In our setting, we favor the use of a \textit{soft} modularity instead of the term in Equation~\eqref{eq:modularity}, as it permits 1) to obtain a differentiable loss, and 2) to avoid the actual reconstruction of node communities at each training iteration, which would incur a larger computational expense.

To conclude we note that, as for a complete evaluation of $\lossGAE$ and $\lossVGAE$, computing the modularity-inspired terms in Equations~\eqref{eq:tildeAE}~and~\eqref{eq:tildeVAE} \textit{on the entire graph} would be of quadratic complexity w.r.t. the number of nodes in the graph. In some of our experiments where such a complexity would be unaffordable (roughly, when $n \geq$ 30 000), we will rely on the FastGAE method from Chapter~\ref{chapter_4} to approximate modularity-inspired terms on random subgraphs and scale our method to larger graphs with up to millions of nodes and edges. 

\subsubsection{On the Selection of Hyperparameters}
\label{c7s7332}

We expect our modularity-inspired losses to improve the training of our Linear/GCN encoders for community detection, i.e., the tuning of their \textit{weight matrices}. However, in addition to these weight matrices, our Modularity-Aware GAE and VGAE models involve several other hyperparameters, that also play a key role. This includes the standard hyperparameters of GAE and VGAE models (e.g., the number of training iterations, the learning rate, the dimensions of encoding layers, and, potentially, the dropout rate \cite{srivastava2014dropout}), but also our newly introduced hyperparameters:  $\lambda$ and $\dregc$ from our encoders, as well as $\beta$ and $\gamma$ from our losses.

In previous research, the selection procedure for such important hyperparameters was sometimes omitted~\cite{choong2018learning,choong2020optimizing}. In our previous community detection experiments from Chapters~\ref{chapter_3},~\ref{chapter_4}~and~\ref{chapter_6}, it was solely based on the optimization of AUC scores on link prediction \textit{validation} sets, following the train/validation/test splitting procedure initially adopted by Kipf and Welling \cite{kipf2016-2} and described in Section~\ref{c2s212} from Chapter~\ref{chapter_2}. However, intuitively, the best hyperparameters for community detection might differ from the best ones for link prediction. Such a selection procedure might therefore be suboptimal for community detection problems.

To tackle this issue, and to complement our novel encoders (Section~\ref{c7s732}) and losses (Subsection~\ref{c7s7331}), we propose an alternative hyperparameters selection procedure w.r.t. previous practices. As community detection is an \textit{unsupervised} downstream task, we cannot rely on train/validation/test splits as for the \textit{supervised} link prediction binary classification task\footnote{We recall that the ground truth communities of each node will be \textit{unavailable} during training. They will only be ultimately revealed for model evaluation, to compare the agreement of the node partition proposed by our GAE or VGAE model to the ground truth partition.}. Consistently with our already described contributions, we rather propose to rely on \textit{modularity} scores, as it is an unsupervised criterion computed independently of the unobserved ground truth clusters.
More precisely, to select relevant hyperparameters, we will:
\begin{itemize}
        \item firstly, construct link prediction train/validation/test sets, as in Section~\ref{c2s212};
        \item then, select hyperparameters that maximize the average of:
        \begin{itemize}
            \item the \textit{AUC score} obtained for link prediction on the validation set; 
            \item the \textit{modularity score} $Q$ defined in Equation~\eqref{eq:modularity}. This score is obtained from the communities extracted by running a $k$-means on the final vectors $z_i,$ learned from the train graph (all nodes are visible but edges from validation and test sets~are~masked).
        \end{itemize}
    \end{itemize} 
We expect this dual criterion 
to facilitate the identification of hyperparameters that will be jointly relevant for link prediction and community detection downstream applications.

\section{Experimental Analysis}
\label{c7s74}

We now present an in-depth experimental evaluation of our proposed Modularity-Aware GAE and VGAE models together with relevant baselines. In Section~\ref{c7s741} we first describe our experimental setting. Then in Section~\ref{c7s742}, we report and discuss our results.

\subsection{Experimental Setting}
\label{c7s741}

\paragraph{Datasets} In the following, we provide an experimental evaluation on seven graphs of various origins, characteristics, and sizes. First and foremost, we consider the Cora ($n =$ 2 708 and $m =$ 5 429), Citeseer ($n =$ 3~327 and $m =$ 4~732), and Pubmed ($n =$ 19 717 and $m  =$ 44 338) citation networks already used in previous chapters and described in Chapter~\ref{chapter_3}. As in our previous experiments, we study two versions of each of these datasets, with and without node features that correspond to bag-of-words vectors of dimensions $f =$ 1433, 3703, and 500, respectively. We recall that, in these datasets, nodes are clustered in 6, 7, and 3 topic classes. As these three citations networks remain the most commonly used graph datasets to evaluate GAE and VGAE models (see our review in Chapter~\ref{chapter_6}), we see value in studying them as well, especially in their \textit{featureless} version where, as explained in Section~\ref{c7s723}, previous GAE and VGAE extensions fall short on community detection.

Consistently with our concluding recommendation from Chapter~\ref{chapter_6}, we complete our experimental evaluation with four other datasets. Firstly, we consider the ten times larger version of Cora already used in Chapter~\ref{chapter_6} for community detection, and referred to as Cora-Larger in the following ($n =$ 23~166 and $m =$ 91~500). Nodes are documents clustered in 70 topic-related communities. Additionally, we consider the Blogs web graph ($n =$ 1~224 and $m =$ 19~025) also used in Chapter~\ref{chapter_6}, where nodes correspond to webpages of political blogs connected through hyperlinks. The blogs are clustered in two communities corresponding to politically left-leaning or right-leaning blogs.
Thirdly, we examine the SBM graph ($n =$ 100~000 and $m =$ 1~498~844) presented in Chapter~\ref{chapter_4}, and generated from a stochastic block model~\cite{abbe2017community}. We recall that nodes from this graph are clustered in 100 ground truth communities of 1~000 nodes each.

Lastly, in the paper associated with this work~\cite{salhagalvan2022modularity}, we also considered an industrial-scale private graph provided by Deezer. As we will further emphasize in Part~\ref{partIII} of this thesis, graph-based methods are at the core of Deezer's recommender systems. In the graph under consideration in this study, denoted Albums ($n =$ 2~503~985 and $m =$ 25~039~155) nodes correspond to \textit{music albums} available on the service. They are connected through an undirected edge when they are regularly \textit{co-listened} by Deezer users (as assessed by internal usage metrics computed from millions of users,  but undisclosed in this work for privacy reasons). Deezer is jointly interested in 1) predicting new connections in the graph corresponding to new albums pairs that users would enjoy listening to together, and is achieved by performing the \textit{link prediction} task; and 2) learning groups of similar albums, with the aim of providing usage-based recommendations (i.e., if users listen to several albums from a community, other unlistened albums from this same community could be recommended to them), which is achieved by performing the \textit{community detection} task. In such an industrial application, learning high-quality album representations that would jointly enable effective link prediction and community detection would therefore be desirable. For evaluation, node communities will be compared to a ground~truth clustering of albums in 20 groups defined by their main \textit{music genre}, allowing us to assess the musical homogeneity of the node~communities~proposed~by~each~model.

\paragraph{Tasks} For each of these seven graphs, we assess the performance of our models on two tasks:
\begin{itemize}
    \item \textbf{Task 1:} we first consider a pure \textit{community detection} task, consisting in the extraction of a partition of the node set $\mathcal{V}$ which ideally agrees with the ground~truth communities of each graph. This task corresponds to the community detection problem considered in experiments from the previous Chapters~\ref{chapter_3},~\ref{chapter_4}~and~\ref{chapter_6}. Communities are retrieved by running the $k$-means algorithm (with $k$-means++ initialization \cite{arthur2007kmeansplus}) in the final embedding space of each model to cluster the vectors $z_i$ (with $k$ matching the true number of communities); except for some baseline methods that explicitly incorporate another strategy to partition nodes (see thereafter). We compare the obtained partitions to the ground truth using the AMI score already used in previous chapters, that we complete in this study (more focused on community detection than others) by the Adjusted Rand Index (ARI). 

\item \textbf{Task 2:} we also consider a \textit{joint link prediction and community detection} task. In such a setting, we learn all node embedding spaces from \textit{incomplete} versions of the seven graphs, where 15\% of edges were randomly masked. We create a validation and
a test set from these masked edges (resp. from 5\% and 10\% of edges, as in Chapters~\ref{chapter_3}~to~\ref{chapter_6}) and the same number of randomly picked unconnected node pairs acting as ``non-edge'' negative pairs. Then, as in previous chapters, we evaluate the ability to distinguish edges from non-edges, i.e., \textit{link prediction}, from the embedding space, using once again the AUC and AP scores on test sets. Jointly, we also evaluate the community detection performance obtained from such incomplete graphs, using the same methodology and AMI/ARI scores as in Task~1.
\end{itemize}

In the case of Task~2, we expect AMI and ARI scores to slightly decrease w.r.t. Task~1, as models will only observe \textit{incomplete} versions of the graphs when learning embedding spaces. Task~2 will further assess whether empirically improving community detection inevitably leads to deteriorating the original good performances of GAE and VGAE models on link prediction. As our proposed modularity-inspired GAE and VGAE models are designed for \textit{joint link prediction and community detection}, we expect them to 1) reach comparable (or, ideally, identical) link prediction scores w.r.t. standard GAE and VGAE models, while 2) reaching better community~detection~scores.

\paragraph{Details on Models}
For the aforementioned evaluation tasks and graphs, we will compare the performances of our proposed \textit{Modularity-Aware GAE and VGAE} models to standard GAE and VGAE and to several other baselines. All results reported below will verify $d = 16,$ i.e., all node embedding models will learn embedding vectors $z_i$ of dimension 16. We also tested models with $d \in \{32, 64\}$ by including them in our grid search space and reached similar conclusions to the $d = 16$ setting (we report and further discuss the impact of $d$ in Section~\ref{c7s742}):

\begin{table}[t]
\centering
\caption[Optimal hyperparameters of all Modularity-Aware GAE and VGAE models]{Complete list of optimal hyperparameters of Modularity-Aware GAE and VGAE models.}
 \vspace{0.2cm}
    \label{tab:hyperparameterstable}
   \resizebox{\textwidth}{!}{
\begin{tabular}{c|cccccccc}
\toprule
\textbf{Dataset} & \textbf{Learning} & \textbf{Number of} & \textbf{Dropout} & \textbf{Use of FastGAE} & $\lambda$ & $\beta$ & $\gamma$ & $\dregc$ \\
& \textbf{rate} & \textbf{iterations} & \textbf{rate} & \textbf{(if yes: subgraphs size)} & & & \\
\midrule
\midrule
\textbf{Blogs} & 0.01 & 200 & 0.0 & No & 0.5 & 0.75 & 2 & 10\\
\textbf{Cora (featureless)}  & 0.01 & 500 & 0.0 & No & 0.25 & 1.0 & 0.25 & 1\\
\textbf{Cora (with features)}  & 0.01 & 300 & 0.0 & No & 0.001 & 0.01 & 1 & 1\\
\textbf{Citeseer (featureless)}  & 0.01 & 500 & 0.0 & No & 0.75 & 0.5 & 0.5 & 2\\
\textbf{Citeseer (with features)}  & 0.01 & 500 & 0.0 & No & 0.75 & 0.5 & 0.5 & 2\\
\textbf{Pubmed (featureless)}  & 0.01 & 500 & 0.0 & No & 0.1 & 0.5 & 0.1 & 5\\ 
\textbf{Pubmed (with features)}  & 0.01 & 700 & 0.0 & No & 0.1 & 0.5 & 10 & 2\\
\textbf{Cora-Large}  & 0.01 & 500 & 0.0 & No & 0.001 & 0.1 & 0.1 & 10 \\
\textbf{SBM}  & 0.01 & 300 & 0.0 & Yes (10 000) & 0.5 & 0.1 & 2 & 10 \\
\textbf{Albums} & 0.005 & 600 & 0.0 & Yes (10 000) & 0.25 & 0.25 & 1 & 5\\
\bottomrule
\end{tabular}}
\end{table}

\begin{itemize}
    \item \textbf{Modularity-Aware GAE and VGAE models}: we trained two versions of Modularity-Aware GAE and VGAE: one with the \textit{linear encoder} described in Section~\ref{c7s732}, and one with the \textit{2-layer GCN encoder} ($\text{GCN}^{(2)}$). The latter encoder includes a 32-dimensional hidden layer. We recall that link prediction is performed from inner product decoding $\hat{A}_{ij} = \sigma(z^T_i z_j)$, and that community detection is performed via a $k$-means on the final vectors $z_i$ learned by each model. During training, as in previous chapters, we used the Adam optimizer~\cite{kingma2014adam}, without dropout (but we tested models with dropout values in $\{0,0.1,0.2\}$ in our grid search optimization). All hyperparameters were carefully tuned following the procedure described in Section~\ref{c7s7332}. For each graph, we tested learning rates from the grid $\{0.001,0.005,0.01,0.05,0.1,0.2\}$, number of training iterations in $\{100, 200, 300, ..., 800\}$, with $\lambda \in \{0, 0.01,0.05, 0.1, 0.2, 0.3, ..., 1.0\}$,  $\beta \in \{0, 0.01,0.05, 0.1, 0.25, 0.5, 1.0, 1.5, 2.0\}$, $\gamma \in \{0.1, 0.2, 0.5, 1.0, 2, 5, 10\}$ and $\dregc \in \{1, 2, 5, 10\}$. The best hyperparameters for each graph are reported in Table~\ref{tab:hyperparameterstable}. We adopted the same optimal hyperparameters for GAE \textit{and} VGAE variants (a result which is consistent with the literature \cite{kipf2016-2}). Lastly, as exact loss computation was computationally unaffordable for our two largest graphs, SBM and Albums, their corresponding models were trained by using the FastGAE method from Chapter~\ref{chapter_4}, approximating losses by reconstructing degree-based sampled subgraphs of $n_{(S)} =$ 10 000 nodes, with $\alpha = 1$.
    
    As in previous chapters, we used Tensorflow~\cite{abadi2016tensorflow}, training our models (as well as GAE/VGAE baselines described below) on an NVIDIA GTX 1080 GPU, and running other operations on a double Intel Xeon Gold 6134 CPU\footnote{On our machines, running times of the Modularity-Aware GAE and VGAE models were comparable to running times of their standard GAE and VGAE counterparts. For example, training each variant of VGAE on the Pubmed graph for 500 training iterations and with $\dregc = 5$ approximately takes 25 minutes on a single GPU (without the FastGAE method which significantly speeds up training \cite{salha2021fastgae}). This is consistent with our claims on the comparable complexity of Modularity-Aware and standard models.}. Along with the publication of our paper~\cite{salhagalvan2022modularity} (which is still under review), we will publicly release our source code on GitHub.

\item \textbf{Standard GAE and VGAE:} we compare these models to two variants of GAEs and VGAEs: one with 2-layer GCN encoders with a 32-dimensional hidden layer (which is equal to the standard GAE and VGAE models from Kipf and Welling~\cite{kipf2016-2}) and one with a linear encoder (corresponding to our models from Chapter~\ref{chapter_6}). We note that these are particular cases of our Modularity-Aware GAE/VGAE with GCN or linear encoder and with $\lambda = 0$ and $\beta = 0$. As for our Modularity-Aware models, link prediction is performed from inner product decoding, and community detection via a $k$-means on vectors $z_i.$ We also adopt a similar model selection procedure as for our Modularity-Aware GAE and VGAE to select hyperparameters (see Section~\ref{c7s7332}). We selected similar learning rates and number of iterations to the values reported~in~Table~\ref{tab:hyperparameterstable}.

\item \textbf{Other baselines:} for completeness, we also compare the standard and Modularity-Aware GAE/VGAE to several other relevant baselines. First and foremost, we report experiments on the VGAECD \cite{choong2018learning} and VGAECD-OPT \cite{choong2020optimizing} models, designed for community detection and discussed in Section~\ref{c7s722}. We use our own Tensorflow implementation of these models\footnote{Authors of VGAECD/VGAECD-OPT did not release any public implementation of their models, and we were unable to reach them by e-mail. We note that we obtained some inconsistent results w.r.t. their original performances (specifically, we reached better performances on featureless graphs, and lower performances on graphs with node features), even when adopting their set of hyperparameters. Authors followed an experimental setting and pure community detection task similar~to~ours.}. We set similar hyperparameters to the above other GAE/VGAE-based models. In all models, the number of Gaussian mixtures matches the ground~truth number of communities of each graph. 
Besides, we also report experiments on the DVGAE \cite{li2020dirichlet} model also discussed in Section \ref{c7s722}, setting similar learning rates and layer dimensions to the above GAE/VGAE-based models, and using the authors' public implementation. In the case of DVGAE, we use 2-layer GCN encoders for consistency with other models of our experiments; we nonetheless acknowledge that Li~et~al.~\cite{li2020dirichlet} also proposed another encoding scheme, denoted Heatts in their paper (but unavailable in their public code at the time of writing) that could replace GCNs both in DVGAE and in Modularity-Aware GAE and VGAE. We also report experiments on the ARGA and ARVGA models from Pan et al.~\cite{pan2018arga} that incorporate an adversarial regularization scheme, with similar hyperparameters, and using the authors' implementation. We already considered these models in Chapter~\ref{chapter_3}. While they were not specifically introduced for community detection, Pan et al. \cite{pan2018arga} reported empirical gains on this task w.r.t.~standard~GAE/VGAE (on graphs with node features).

We furthermore consider three additional baselines not utilizing the autoencoder paradigm. Firstly, we report results from node2vec~\cite{grover2016node2vec} and DeepWalk \cite{perozzi2014deepwalk}, using similar settings w.r.t. Chapter~\ref{chapter_3}. We use a similar strategy to our aforementioned GAE/VGAE models ($k$-means/inner products) for community detection and link prediction from embedding spaces. Lastly, we also compare to the \textit{Louvain} community detection method, using the authors' implementation~\cite{blondel2008louvain}. 
We see value in comparing our methods to a direct use of Louvain, as this method 1) often emerged as a simple but competitive alternative to GAE/VGAE for community detection (see Section~\ref{c7s72}), and 2) is directly leveraged in our proposed Modularity-Aware GAE and VGAE models as a pre-processing step for the computation of $A_c$ and $\Ao$ (see Section~\ref{c7s732}).

\end{itemize}

\subsection{Results and Discussion}
\label{c7s742}

We now present our experimental results. Firstly, we analyze the impact of our proposed hyperparameter selection procedure. Then, we discuss results on Task~1 and then on Task~2. Finally, we mention the limitations and possible extensions of our approach.

\paragraph{On the Selection of Hyperparameters}
In Subsection~\ref{c7s7332}, we proposed an alternative hyperparameter selection procedure w.r.t. previous practices in the literature. Based on the joint maximization of AUC validation scores for link prediction and modularity scores $Q$, it aims to identify more relevant GAE/VGAE hyperparameters for joint link prediction and community detection. We recall that the resulting optimal hyperparameters are displayed in Table \ref{tab:hyperparameterstable}.

In our experiments, this procedure did not modify our choices of learning rates and dropout rates for the different GAE/VGAE models under consideration, w.r.t. a standard selection solely relying on AUC validation scores. It had a more noticeable impact on the choices of clustering-related hyperparameters in Modularity-Aware GAE and VGAE (i.e., $\lambda$, $\beta$, $\gamma$, and $\dregc$) as well as on the required number of training iterations in~the~gradient~descent/ascent.

\begin{figure*}[t]
\centering
\resizebox{0.86\textwidth}{!}{
  \subfigure[Cora]{
  \scalebox{0.49}{\includegraphics{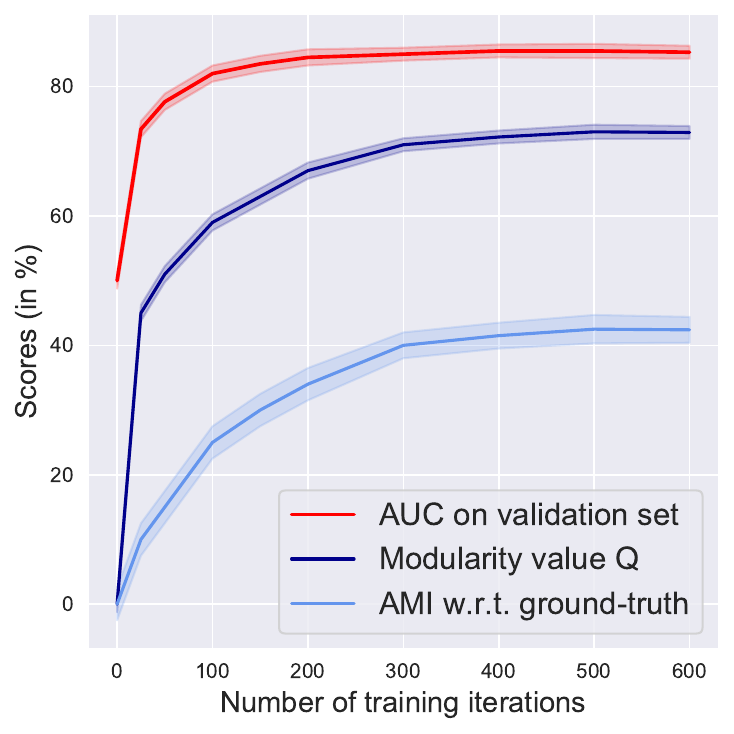}}}\subfigure[Pubmed]{
  \scalebox{0.49}{\includegraphics{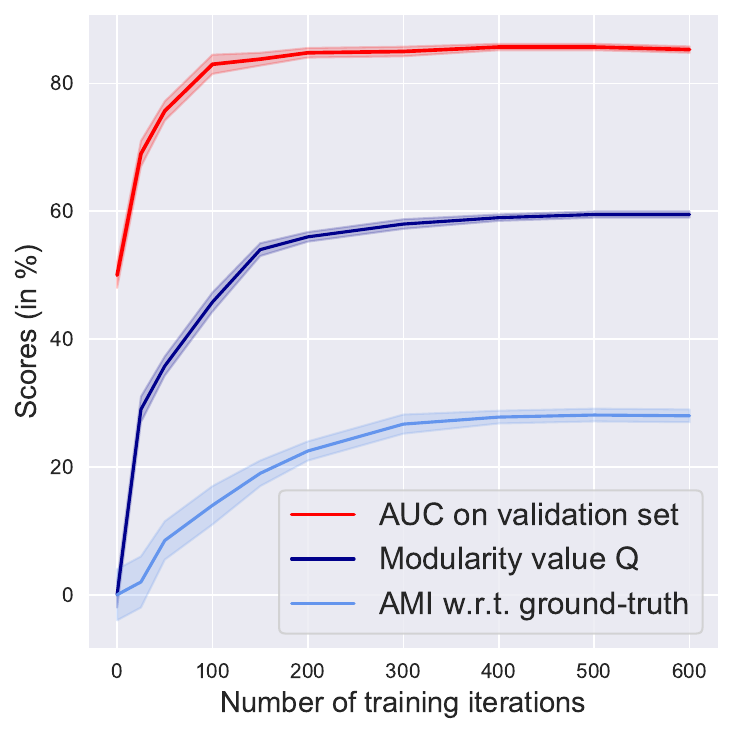}}}}
  \caption[Evolution of AUC and modularity w.r.t. the number of training iterations]{Identification of the required number of training iterations, for Modularity-Aware VGAE with linear encoders trained on the featureless (a) Cora, and (b) Pubmed graphs. The plots report the evolution of the modularity $Q$ (\textcolor{lightblue4}{dark blue}) and AUC link prediction scores on validation sets (\textcolor{red}{red}) w.r.t. the number of model training iterations in gradient descent. By looking at the red curves only, one might choose to stop training models after 200 iterations as Kipf and Welling~\cite{kipf2016-2}, as the AUC validation scores have almost stabilized. However, the dark blue curves emphasize that $Q$ still increases up to 400-500 training iterations for both graphs. By also using $Q$ for hyperparameter selection (as we proposed), one will therefore continue training VGAE models up to 400-500 iterations. The \textcolor{ImperialColor!90}{light blue} curves confirm that such a strategy eventually leads to better AMI final scores w.r.t. ground~truth communities. Note, that the light blue curves could \textit{not} be directly used for tuning, as ground~truth communities are assumed to be unavailable at training time.}
  \label{fig:optimization}
\end{figure*}

Figure~\ref{fig:optimization} provides an example of this phenomenon, for the number of training iterations required to train Modularity-Aware VGAE models on the featureless Cora and Pubmed graphs. The figure shows that, unlike our proposed procedure jointly based on AUC and $Q$, a hyperparameter selection based solely on AUC validation scores leads to earlier stopping of the model training and suboptimal performances on community detection. This reaffirms the empirical relevance of our proposed procedure, and that optimal hyperparameters for joint link prediction and community detection might differ from those for link prediction only. Moreover, we note that, while Figure~\ref{fig:optimization} focuses on two Modularity-Aware VGAE models, our procedure also leads to the selection of a larger number of training iterations for the other GAE/VGAE-based methods under consideration in this work (values are similar to those in Table~\ref{tab:hyperparameterstable}), which explains why, on some occasions, we will report slightly improved results w.r.t. those obtained in the original papers for these models (for instance, some results on Linear GAE and VGAE models will be slightly better than those reported in Chapter~\ref{chapter_6}).

\paragraph{Results for Community Detection on Original Graphs (Task 1)}

\begin{table}[t]
\begin{center}
\begin{small}
\centering
\caption[Task 1 and Task 2 on Cora using Modularity-Aware GAE and VGAE]{Results for Task 1 and Task 2 on the featureless Cora graph, using Modularity-Aware GAE and VGAE with Linear and GCN encoders, their standard GAE and VGAE counterparts, and other baselines. All node embedding models learn embedding vectors of dimension $d =16$, with other hyperparameters set as described in Section~\ref{c7s741}. Scores are averaged over 100 runs. For Task 2, link prediction results are reported from test sets (edges masked for the original graph in addition to the same number of randomly picked unconnected node pairs). \textbf{Bold} numbers correspond to the best performance for each score. Scores \textit{in italic} are within one standard deviation range from the best score.}
    \label{tab:coraresults}
   \resizebox{1.0\textwidth}{!}{
\begin{tabular}{r||cc||cc|cc}
\toprule
\textbf{Model} &  \multicolumn{2}{c}{\textbf{Task 1: Community Detection}} & \multicolumn{4}{c}{\textbf{Task 2: Joint Link Prediction and Community Detection}}\\
 & \multicolumn{2}{c}{\textbf{on complete graph}} & \multicolumn{4}{c}{\textbf{on graph with 15\% of edges being masked}}\\
\midrule
 & \textbf{AMI (in \%)} & \textbf{ARI (in \%)} &  \textbf{AMI (in \%)} &  \textbf{ARI (in \%)} &  \textbf{AUC (in \%)} & \textbf{AP (in \%)} \\ 
\midrule
\midrule
\underline{\textit{Modularity-Aware GAE/VGAE Models}} &  &  &  &  &  &  \\
Linear Modularity-Aware VGAE & \textbf{46.65} $\pm$ \textbf{0.94} & \textit{39.43} $\pm$ \textit{1.15} & \textit{42.86} $\pm$ \textit{1.65} & \textit{34.53} $\pm$ \textit{1.97} & \textit{85.96} $\pm$ \textit{1.24} & \textit{87.21} $\pm$ \textit{1.39} \\
Linear Modularity-Aware GAE & \textit{46.58} $\pm$ \textit{0.40} & \textbf{39.71} $\pm$ \textbf{0.41} & \textbf{43.48} $\pm$ \textbf{1.12} & \textbf{35.51} $\pm$ \textbf{1.20} & \textbf{87.18} $\pm$ \textbf{1.05} & \textit{88.53} $\pm$ \textit{1.33} \\
GCN-based Modularity-Aware VGAE & 43.25 $\pm$ 1.62 & 35.08 $\pm$ 1.88 & 41.03 $\pm$ 1.55 & \textit{33.43} $\pm$ \textit{2.17} & 84.87 $\pm$ 1.14 & 85.16 $\pm$ 1.23  \\
GCN-based Modularity-Aware GAE & 44.39 $\pm$ 0.85 & 38.70 $\pm$ 0.94 & 41.13 $\pm$ 1.35 & \textit{35.01} $\pm$ \textit{1.58} & \textit{86.90} $\pm$ \textit{1.16} & \textit{87.55} $\pm$ \textit{1.26} \\
\midrule
\underline{\textit{Standard GAE/VGAE Models}} &  &  &  &  &  &  \\
Linear VGAE & 37.12 $\pm$ 1.46 & 26.83 $\pm$ 1.68 & 32.22 $\pm$ 1.76 & 21.82 $\pm$ 1.80 & 85.69 $\pm$ 1.17 & \textbf{89.12} $\pm$ \textbf{0.82} \\
Linear GAE & 35.05 $\pm$ 2.55 & 24.32 $\pm$ 2.99 & 28.41 $\pm$ 1.68 & 19.45 $\pm$ 1.75 & 84.46 $\pm$ 1.64 & \textit{88.42} $\pm$ \textit{1.07} \\
GCN-based VGAE & 34.36 $\pm$ 3.66 & 23.98 $\pm$ 5.01 & 28.62 $\pm$ 2.76 & 19.70 $\pm$ 3.71 & 85.47 $\pm$ 1.18 & \textit{88.90} $\pm$ \textit{1.11} \\
GCN-based GAE & 35.64 $\pm$ 3.67 & 25.33 $\pm$ 4.06 & 31.30 $\pm$ 2.07 & 19.89 $\pm$ 3.07 & 85.31 $\pm$ 1.35 & \textit{88.67} $\pm$ \textit{1.24} \\
\midrule
\midrule
\underline{\textit{Other Baselines}} &  &  &  &  &  &  \\
Louvain & 42.70 $\pm$ 0.65 & 24.01 $\pm$ 1.70 & 39.09 $\pm$ 0.73 & 20.19 $\pm$ 1.73 & -- & -- \\
VGAECD& 36.11 $\pm$ 1.07 & 27.15 $\pm$ 2.05 & 33.54 $\pm$ 1.46 & 24.32 $\pm$ 2.25 & 83.12 $\pm$ 1.11 & 84.68 $\pm$ 0.98 \\
VGAECD-OPT & 38.93 $\pm$ 1.21 & 27.61 $\pm$ 1.82 & 34.41 $\pm$ 1.62 & 24.66 $\pm$ 1.98 & 82.89 $\pm$ 1.20 & 83.70 $\pm$ 1.16 \\
ARGVA & 34.97 $\pm$ 3.01 & 23.29 $\pm$ 3.21 & 28.96 $\pm$ 2.64 & 19.74 $\pm$ 3.02 & 85.85 $\pm$ 0.87 & \textit{88.94} $\pm$ \textit{0.72} \\
ARGA & 35.91 $\pm$ 3.11 & 25.88 $\pm$ 2.89 & 31.61 $\pm$ 2.05 & 20.18 $\pm$ 2.92 & 85.95 $\pm$ 0.85 & \textit{89.07} $\pm$ \textit{0.70} \\
DVGAE & 35.02 $\pm$ 2.73 & 25.03 $\pm$ 4.32 & 30.46 $\pm$ 4.12 & 21.06 $\pm$ 5.06 & 85.58 $\pm$ 1.31 & \textit{88.77} $\pm$ \textit{1.29} \\
DeepWalk & 36.58 $\pm$ 1.69 & 27.92 $\pm$ 2.93 & 30.26 $\pm$ 2.32 & 20.24 $\pm$ 3.91 & 80.67 $\pm$ 1.50 & 80.48 $\pm$ 1.28 \\
node2vec & 41.64 $\pm$ 1.25 & 34.30 $\pm$ 1.92 & 36.25 $\pm$ 1.38 & 29.43 $\pm$ 2.21 & 82.43 $\pm$ 1.23 & 81.60 $\pm$ 0.91 \\
\bottomrule
\end{tabular}}
\end{small}
\end{center}
\end{table}

\begin{figure*}[t]
\centering
\resizebox{1.03\textwidth}{!}{
  \subfigure[Linear Standard VGAE]{
  \scalebox{0.48}{\includegraphics{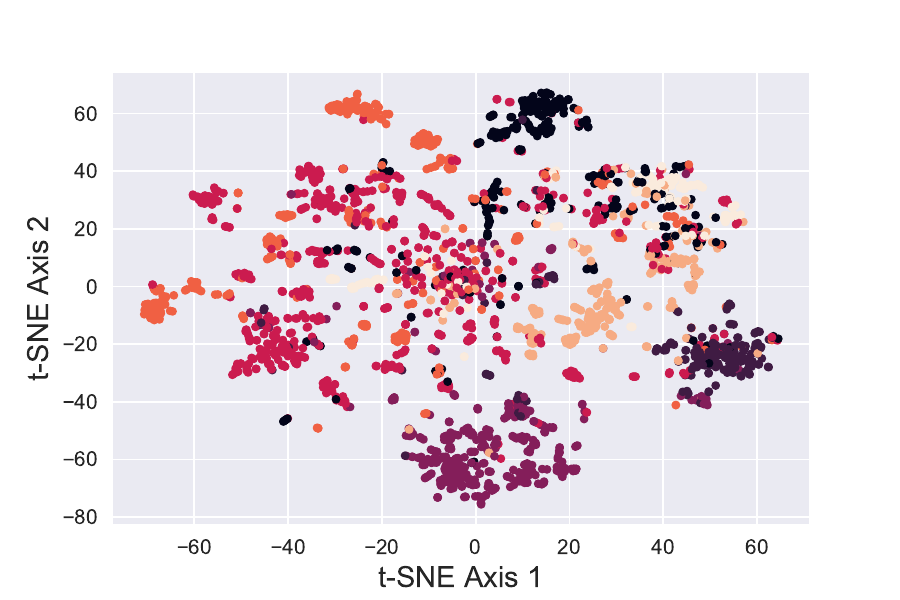}}}\subfigure[Linear Modularity-Aware VGAE]{
  \scalebox{0.48}{\includegraphics{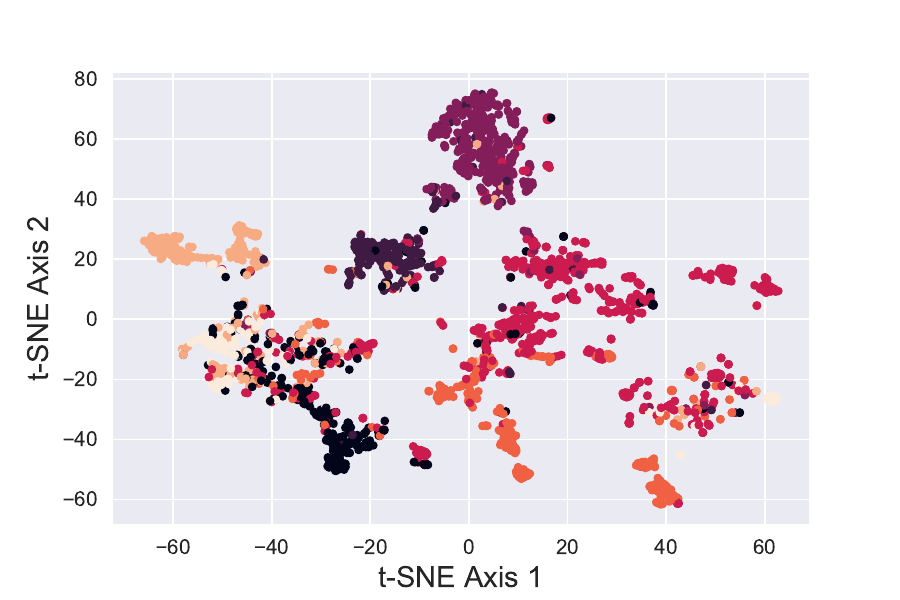}}}}
  \caption[Visualization of node embedding representations from Modularity-Aware VGAE]{Visualization of node embedding representations for the featureless Cora graph, learned by (a)~Standard VGAE, and (b)~Modularity-Aware VGAE, with linear encoders. The plots were obtained using the t-SNE method for high-dimensional data visualization. Colors denote ground truth communities, that were not available during training. Although community detection is not perfect (both methods return AMI scores $<$ 50\% in Table~\ref{tab:coraresults}), node embedding representations from (b) provide a more visible separation of these communities. Specifically, in Table~\ref{tab:coraresults}, using  Linear Modularity-Aware VGAE for community detection leads to an increase of 9 AMI points (Task 1) to 10 AMI points (Task 2) for community detection w.r.t. Linear Standard VGAE, while preserving comparable performances on link prediction (Task 2).}
  \label{visucora}
\end{figure*}

We now focus on the ``pure'' community detection task (Task 1), performed by models trained on graphs where no edges are removed for model training, as previously introduced in Section~\ref{c7s741}. The second and third columns of Table~\ref{tab:coraresults} report mean AMI and ARI scores on Cora for this task along with standard deviations over 100 runs, for Modularity-Aware GAE and VGAE models (with linear or with 2-layer GCN encoders), their standard counterparts and~other~baselines. 
We draw several conclusions from Table \ref{tab:coraresults}. Foremost, previous conclusions~\cite{choong2018learning,choong2020optimizing,salha2021fastgae,salha2019-1} on the limitations of standard GAE and VGAE models for community detection are confirmed: in Table~\ref{tab:coraresults}, these methods are notably outperformed by a direct use of the Louvain method (e.g., 42.70\% vs 34.36\% mean AMI scores for Louvain vs GCN-based VGAE). We also observe that previous GAE/VGAE extensions, reported as baselines, actually provide few empirical benefits w.r.t. standard GAE and VGAE models for this \textit{featureless} graph (e.g., only +1.81 AMI points for VGAECD-OPT\footnote{The increase is even smaller when replacing AMI scores obtained via our re-implementation of VGAECD and VGAECD-OPT (i.e., 36.11\% and 38.93\%, respectively) by AMI scores originally reported in \cite{choong2020optimizing} for these methods (i.e., 28.22\% and 37.35\%, respectively), which are lower than ours.} vs Linear VGAE). Such a result, in conjunction with the improved performances of these same baselines on graphs \textit{with features} (see thereafter), tends to confirm our initial diagnosis that various GAE/VGAE extensions for community detection mainly benefit from the presence of node features.

On the contrary, our proposed Modularity-Aware GAE and VGAE models, incorporating Louvain clusters as a prior signal in the GAE's and VGAE's encoders, significantly outperform both the use of the Louvain method alone, and the use of GAE and VGAE alone (e.g., with a top 46.65\% mean AMI for Modularity-Aware VGAE with linear encoders, and a top 39.71\% mean ARI score for Modularity-Aware GAE with linear encoders). Modularity-Aware models also compare favorably to the baselines under consideration (e.g., with +12.1 ARI points for Linear Modularity-Aware GAE w.r.t. VGAECD-OPT), while also providing less volatile results w.r.t. standard GAE/VGAE models.  Furthermore, we note that Modularity-Aware models with linear encoders tend to outperform their GCN-based counterparts and that GAE and VGAE reach comparable scores. In addition to these results, Figure~\ref{visucora} visualizes the representations learned by our models using t-SNE\footnote{We used the scikit-learn \cite{pedregosa2011scikit} implementation of this data visualization method: \href{https://scikit-learn.org/stable/modules/generated/sklearn.manifold.TSNE.html}{https://scikit-learn.org/stable/modules/generated/sklearn.manifold.TSNE.html}}~\cite{van2008visualizing}.

Overall, we obtain similar conclusions from the other graph datasets. Following the format of Table~\ref{tab:coraresults}, columns two and three of Table~\ref{tab:pubmedresults} present detailed community detection results for the featureless Pubmed graph. Table~\ref{tab:allresults} reports more summarized results for all other graph datasets under consideration, with and without node features (when available).
While Louvain outperforms standard GAE/VGAE models in 5 featureless graphs out of 7 in Table~\ref{tab:allresults} (e.g., 19.81\% vs 15.79\% mean AMI scores for Louvain vs GCN-based VGAE on Albums), our Modularity-Aware models manage to achieve either comparable or better performances w.r.t. standard models, Louvain and other baselines in the wide majority of experiments. Furthermore, throughout Table~\ref{tab:allresults}, we observe that linear encoders outperform their GCN-based counterparts in 8/10 experiments, and that VGAE models outperform GAE models in 8/10 experiments (even though performances are often relatively close, as for Cora). Lastly we emphasize that, while all tables report results for fixed embedding dimensions of $d = 16$, we reached similar conclusions for $d \in \{32, 64\}$. Although performances sometimes improved by increasing $d$, the \textit{ranking} of methods under consideration remained similar. Such a result is consistent with experiments from previous chapters. For instance, by setting $d = 64$, Deepwalk's mean AMI score increased from 36.58\% to roughly 41\% on the featureless Cora graph, while the mean AMI score from our Linear Modularity-Aware GAE model simultaneously increased from 46.58\% to 47.80\%.

\begin{table}[t]
\begin{center}
\begin{small}
\centering
\caption[Task 1 and Task 2 on Pubmed using Modularity-Aware GAE and VGAE]{Results for Task 1 and Task 2 on the featureless Pubmed graph, using Modularity-Aware GAE and VGAE with Linear and 2-layer GCN encoders, their standard GAE and VGAE counterparts, and other baselines. All node embedding models learn embedding vectors of dimension $d =16$, with other hyperparameters set as described in Section~\ref{c7s741}. Scores are averaged over 100 runs. For Task 2, link prediction results are reported from test sets (edges masked during training + same number of randomly picked unconnected node pairs).  \textbf{Bold} numbers correspond to the best performance for each score. Scores \textit{in italic} are within one standard deviation range from the best score.}
    \label{tab:pubmedresults}
   \resizebox{1.0\textwidth}{!}{
\begin{tabular}{r||cc||cc|cc}
\toprule
\textbf{Model} &  \multicolumn{2}{c}{\textbf{Task 1: Community Detection}} & \multicolumn{4}{c}{\textbf{Task 2: Joint Link Prediction and Community Detection}}\\
 & \multicolumn{2}{c}{\textbf{on complete graph}} & \multicolumn{4}{c}{\textbf{on graph with 15\% of edges being masked}}\\
\midrule
 & \textbf{AMI (in \%)} & \textbf{ARI (in \%)} &  \textbf{AMI (in \%)} &  \textbf{ARI (in \%)} &  \textbf{AUC (in \%)} & \textbf{AP (in \%)} \\ 
\midrule
\midrule
\underline{Modularity-Aware GAE/VGAE Models} &  &  &  &  &  &  \\
Linear Modularity-Aware VGAE & 28.12 $\pm$ 0.29 & 29.01 $\pm$ 0.51 & 25.93 $\pm$ 0.65 & 23.76 $\pm$ 0.49 & \textbf{85.76} $\pm$ \textbf{0.37} & 87.77 $\pm$ 0.31 \\
Linear Modularity-Aware GAE & \textit{28.54} $\pm$ \textit{0.24} & 26.36 $\pm$ 0.34 & \textbf{26.38} $\pm$ \textbf{0.43} & 21.30 $\pm$ 0.59 & 84.39 $\pm$ 0.32 & \textit{87.92} $\pm$ \textit{0.40} \\
GCN-based Modularity-Aware VGAE & 28.08 $\pm$ 0.27 & 28.14 $\pm$ 0.33 & 25.70 $\pm$ 0.86 & 22.65 $\pm$ 0.80 & 84.70 $\pm$ 0.24 & 86.64 $\pm$ 0.15 \\
GCN-based Modularity-Aware GAE & \textbf{28.74} $\pm$ \textbf{0.28} & 26.71 $\pm$ 0.47 & 25.52 $\pm$ 0.45 & 20.52 $\pm$ 0.31 & 85.07 $\pm$ 0.35 & \textit{88.27} $\pm$ \textit{0.39} \\
\midrule
\underline{Standard GAE/VGAE Models} &  &  &  &  &  &  \\
Linear VGAE & 22.16 $\pm$ 2.02 & 13.90 $\pm$ 3.47  & 21.78 $\pm$ 2.57 & 13.81 $\pm$ 3.17 & 84.57 $\pm$ 0.51 & \textbf{88.31} $\pm$ \textbf{0.44} \\
Linear GAE & 12.61 $\pm$ 4.61 & 6.37 $\pm$ 3.86  & 12.60 $\pm$ 4.67 & 6.21 $\pm$ 1.75 & 82.03 $\pm$ 0.32 & 87.71 $\pm$ 0.24 \\
GCN-based VGAE & 20.11 $\pm$ 3.05 & 13.12 $\pm$ 3.10 & 17.34 $\pm$ 2.99 & 8.71 $\pm$ 3.05 & 82.19 $\pm$ 0.88 & 87.51 $\pm$ 0.55 \\
GCN-based GAE & 20.12 $\pm$ 2.89 & 14.21 $\pm$ 2.78 & 16.75 $\pm$ 3.36 & 9.18 $\pm$ 2.71 & 82.33 $\pm$ 1.32 & 87.20 $\pm$ 0.58 \\

\midrule
\midrule
\underline{Other Baselines} &  &  &  &  &  &  \\
Louvain & 20.06 $\pm$ 0.27 & 10.34 $\pm$ 0.99 & 16.71 $\pm$ 0.46 & 8.32 $\pm$ 0.79 & -- & -- \\
VGAECD & 20.32 $\pm$ 2.95 & 13.54 $\pm$ 2.98 & 17.39 $\pm$ 3.04 & 9.21 $\pm$ 3.12 & 82.05 $\pm$ 0.90 & 87.30 $\pm$ 0.53 \\
VGAECD-OPT & 22.50 $\pm$ 1.99 & 14.58 $\pm$ 2.86 & 21.98 $\pm$ 2.46 & 15.22 $\pm$ 2.92 & 82.03 $\pm$ 0.82 &  87.41 $\pm$ 0.53 \\
ARGVA & 20.73 $\pm$ 3.10 & 13.94 $\pm$ 3.12 & 17.63 $\pm$ 3.19 & 9.19 $\pm$ 3.09 & 84.07 $\pm$ 0.55 & 87.73 $\pm$ 0.49 \\
ARGA & 20.98 $\pm$ 2.90 & 14.79 $\pm$ 2.80 & 17.21 $\pm$ 3.01 & 9.59 $\pm$ 2.76 & 83.73 $\pm$ 0.53 & \textit{87.90} $\pm$ \textit{0.45} \\
DVGAE & 23.15 $\pm$ 2.52 & 15.02 $\pm$ 3.33 & 22.10 $\pm$ 2.50 & 14.62 $\pm$ 2.96 & 83.21 $\pm$ 0.92 & \textit{88.17} $\pm$ \textit{0.49} \\
DeepWalk & \textit{28.53} $\pm$ \textit{0.43} & 29.61 $\pm$ 0.33 & 15.80 $\pm$ 1.05 & 16.16 $\pm$ 1.75 & 80.63 $\pm$ 0.42 & 81.03 $\pm$ 0.54 \\
node2vec & \textit{28.52} $\pm$ \textit{1.12} & \textbf{30.63} $\pm$ \textbf{1.14} & 23.88 $\pm$ 0.54 & \textbf{25.90} $\pm$ \textbf{0.65} & 81.03 $\pm$ 0.30 & 82.33 $\pm$ 0.41 \\
\bottomrule
\end{tabular}}
\end{small}
\end{center}
\end{table}

Interestingly, we also observe that combining the Louvain method and GAE/VGAE in our Modularity-Aware models might be empirically beneficial \textit{even when standard GAE/VGAE models initially outperform the Louvain method}. For instance in Table~\ref{tab:pubmedresults}, our Linear Modularity-Aware VGAE outperforms the Linear Standard VGAE (e.g., with 28.12\% vs 22.16\% mean AMI scores), despite the fact that this standard model initially outperformed the Louvain method (20.06\% mean AMI score). This tends to confirm that modularity-based clustering \textit{à la} Louvain complements the encoding-decoding paradigm of GAE and VGAE models, and that learning node embedding spaces from complementary criteria is empirically beneficial. On a more negative note, we nonetheless acknowledge that, on Cora, Citeseer, and Pubmed in Table~\ref{tab:allresults}, empirical gains of Modularity-Aware models are less visible on graphs \textit{equipped with node features} than on featureless graphs, which we will further discuss in our limitation section.

\begin{figure*}[t]
\centering
\resizebox{1.0\textwidth}{!}{
  \subfigure[Cora]{
  \scalebox{0.48}{\includegraphics{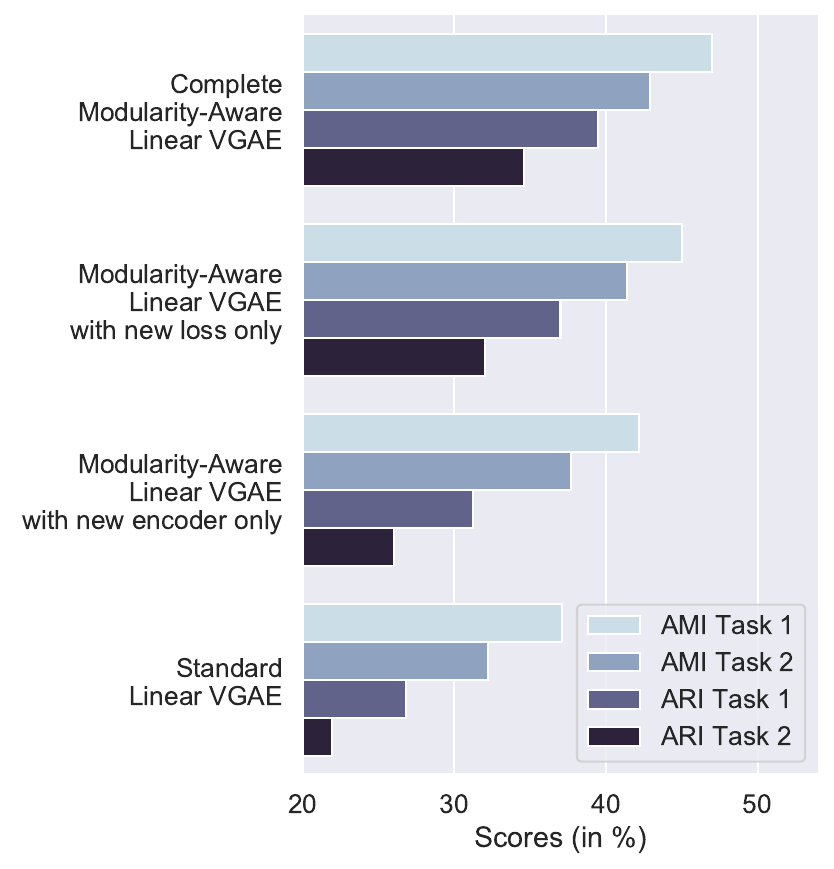}}}\subfigure[Albums]{
  \scalebox{0.48}{\includegraphics{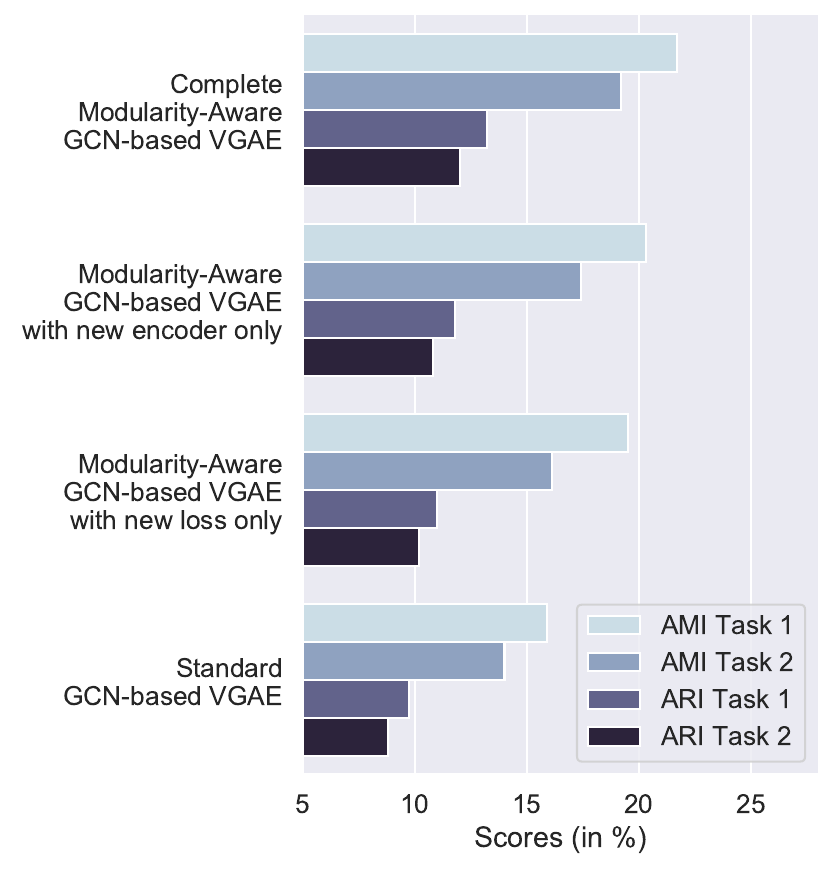}}}}
  \caption[Ablation study on the components of Modularity-Aware VGAE]{Comparison of two ``complete'' Modularity-Aware VGAE, trained on (a) featureless Cora and (b) Albums with variants of these models only leveraging the new \textit{encoder} from Section~\ref{c7s732}, or the new \textit{loss} from Section~\ref{c7s733}. We observe that incorporating any of these two components improves community detection on these two graphs w.r.t. Standard VGAE. Moreover, using both components \textit{simultaneously} leads to the best results. Note, the optimal pair ($\lambda$, $\beta$) for complete models might differ from the optimal $\lambda$ (resp. $\beta$) when incorporating the new encoder (resp. loss) only.}
    \label{fig:ablation}
\end{figure*}

As our Modularity-Aware models include two main novel components (namely our community-preserving encoders from Section~\ref{c7s732} and our revised loss from Section~\ref{c7s733}), one might wonder what the contribution of each of these components to the performance gains is. To study this question, we report in Figure~\ref{fig:ablation} the results of an \textit{ablation study}, that consisted in training variant versions of our models leveraging one of these components only\footnote{A Modularity-Aware GAE or VGAE model that leverages the novel encoder only (respectively, the novel loss only) corresponds to a particular case of a ``complete'' Modularity-Aware GAE or VGAE model, where the hyperparameter $\beta$ (respectively, the hyperparameter $\lambda$) is set~to~0.} (i.e., the encoder but not the loss, or the loss but not the encoder). Figure~\ref{fig:ablation} shows that incorporating any of these two individual components into the model improves community detection. The gain is larger for the \textit{loss} in the Cora example from Figure~\ref{fig:ablation}(a), while it is larger for the \textit{encoder} in the Albums example from Figure~\ref{fig:ablation}(b). A simultaneous use of the encoder and the loss leads to the best results in both examples, which we also confirmed on the other graphs under consideration.

\paragraph{Results for Joint Link Prediction and Community Detection (Task 2)}

We now study results for Task 2, the joint link prediction and community detection task described in Section~\ref{c7s741}, and performed on incomplete versions of the graph datasets where 15\% of edges are randomly masked. Results for this task (i.e., AMI and ARI scores from community detection on incomplete graphs, and AUC and AP scores from link prediction on test sets) are reported in the four rightmost columns of Tables~\ref{tab:coraresults},~\ref{tab:pubmedresults}~and~\ref{tab:allresults}. AMI and ARI scores from this task are also included in the ablation study in Figure~\ref{fig:ablation}.

We draw several conclusions from these additional experiments. First of all, we confirm that AMI and ARI scores decrease slightly w.r.t. Task 1, which was expected due to the absence of part of the graph structure during the training phase (e.g., from 46.65\% to 42.86\% mean AMI, for Linear Modularity-Aware VGAE on Cora in Table~\ref{tab:coraresults}). Nonetheless, the ranking of the different methods under consideration remains consistent with Task~1. In particular, our Modularity-Aware models still outperform baselines in cases where they were already outperforming in Task~1 (e.g., with a top 43.48\% mean AMI for Linear Modularity-Aware GAE on Cora  in Table~\ref{tab:coraresults}, vs 28.41\% for the standard Linear GAE and 39.09\% for the Louvain method).

Besides these confirmations, the main goal of Task 2 was to address our second research question stated in Section \ref{c7s71}: \textit{do improvements on the community detection task necessarily incur a loss in the link prediction performance or can they be jointly addressed with high accuracy?} 
Indeed, as GAE and VGAE models were originally recognized as effective link prediction methods, improving community detection while deteriorating link prediction might be undesirable, especially in problems requiring effective node embedding representations for multitask applications (see the Deezer example from Section~\ref{c7s741}). By design, our proposed encoders, losses, and selection procedure specifically aimed to avoid such a deterioration.

Empirical results confirm the ability of Modularity-Aware models to preserve comparable link prediction performances w.r.t standard GAE and VGAE models. For instance in Table~\ref{tab:coraresults}, our Linear Modularity-Aware GAE model reaches mean AUC and AP scores of 87.18\% and 88.53\%, respectively, which is comparable (or even slightly better in the case of AUC) to Linear Standard GAE (84.46\% and 88.42\%, respectively). We reach similar results for the three other Modularity-Aware models in Table~\ref{tab:coraresults}, while scores of several baselines deteriorate by a few points. Overall, all other Modularity-Aware models reported in the complete Table~\ref{tab:allresults} achieve comparable (either better, identical, or only a few points below) AUC and AP scores w.r.t. their GAE or VGAE counterparts.

\paragraph{Limitations and Possible Extensions}

As observed in the previous paragraphs, empirical gains of Modularity-Aware models are less pronounced on graphs equipped with node features (although non null in 2/3 cases). In Table~\ref{tab:allresults}, for Cora with features, we ``only'' report increases in Task 1 AMI scores of +2.63 points w.r.t. the corresponding standard VGAE model. For comparison, in the featureless case, we reported increases in Task 1 AMI scores of +11.53 points. Furthermore, our Modularity-Aware VGAE model does not surpass standard VGAE models at all on Pubmed with features. We hypothesize that the incorporation of Louvain-based prior clusters in Modularity-Aware models might be less relevant for these attributed graphs. Indeed, while Louvain only leverages the graph structure for node clustering, node features seem to play a strong role in the identification of ground truth communities for these graphs. 
\clearpage

\begin{table}[ht]
\begin{center}
\begin{small}
\centering
\caption[Task 1 and Task 2 on all graphs using Modularity-Aware GAE and VGAE]{Summarized results for Tasks 1 and 2 on all graphs. For each graph, for brevity, we only report the \textbf{best} Modularity-Inspired model (best on Task 2, among GCN \textbf{or} linear encoder, and GAE \textbf{or} VGAE), its standard counterpart, and a comparison to the Louvain baseline as well as the best other baseline (among VGAECD, VGAECD-OPT, ARGA, ARGVA, DVGAE, DeepWalk and node2vec). All node embedding models learn embedding vectors of dimension $d =16$, with other hyperparameters set as described in Section~\ref{c7s741}. Scores are averaged over 100 runs except for the larger SBM and Albums graphs (10 runs). \textbf{Bold} numbers correspond to the best performance for each score. Scores \textit{in italic} are within one standard deviation range from~the~best~score.}
    \label{tab:allresults}
     \vspace{0.2cm}
   \resizebox{1.0\textwidth}{!}{
\begin{tabular}{c|r||cc||cc|cc}
\toprule
\textbf{Dataset} & \textbf{Model} &  \multicolumn{2}{c}{\textbf{Task 1: Community Detection}} & \multicolumn{4}{c}{\textbf{Task 2: Joint Link Prediction and Community Detection}}\\
& & \multicolumn{2}{c}{\textbf{on complete graph}} & \multicolumn{4}{c}{\textbf{on graph with 15\% of edges being masked}}\\
\midrule
&  & \textbf{AMI (in \%)} & \textbf{ARI (in \%)} &  \textbf{AMI (in \%)} &  \textbf{ARI (in \%)} &  \textbf{AUC (in \%)} & \textbf{AP (in \%)} \\ 
\midrule
\midrule
& GCN-based Modularity-Aware VGAE & \textbf{74.23} $\pm$ \textbf{0.95} & \textbf{83.13} $\pm$ \textbf{0.79} & \textbf{70.42} $\pm$ \textbf{1.28} & \textbf{79.80} $\pm$ \textbf{1.12} & \textbf{91.67} $\pm$ \textbf{0.39} & \textit{92.37} $\pm$ \textit{0.41} \\
& GCN-based Standard VGAE & 73.42 $\pm$ 0.65 & 82.58 $\pm$ 0.52 & 66.90 $\pm$ 3.32 & \textit{77.23} $\pm$ \textit{3.89} & \textit{91.64} $\pm$ \textit{0.42} & \textbf{92.52} $\pm$ \textbf{0.51} \\ 
\textbf{Blogs} & Louvain & 63.43 $\pm$ 0.86 & 76.66 $\pm$ 0.70 & 57.25 $\pm$ 1.67 & 73.00 $\pm$ 1.56 & -- & -- \\
& \underline{Best other baseline:} &  &  &  &  &  & \\
& node2vec & 72.88 $\pm$ 0.87 & 82.08 $\pm$ 0.73 & 67.64 $\pm$ 1.23 & 77.03 $\pm$ 1.85 & 83.63 $\pm$ 0.34 & 79.60 $\pm$ 0.61\\
\midrule
& Linear Modularity-Aware GAE & \textbf{46.58} $\pm$ \textbf{0.40} & \textbf{39.71} $\pm$ \textbf{0.41} & \textbf{43.48} $\pm$ \textbf{1.12} & \textbf{35.51} $\pm$ \textbf{1.20} & \textbf{87.18} $\pm$ \textbf{1.05} & \textbf{88.53} $\pm$ \textbf{1.33} \\
& Linear Standard GAE & 35.05 $\pm$ 2.55 & 24.32 $\pm$ 2.99 & 28.41 $\pm$ 1.68 & 19.45 $\pm$ 1.75 & 84.46 $\pm$ 1.64 & \textit{88.42} $\pm$ \textit{1.07} \\
\textbf{Cora} & Louvain & 42.70 $\pm$ 0.65 & 24.01 $\pm$ 1.70 & 39.09 $\pm$ 0.73 & 20.19 $\pm$ 1.73 & -- & -- \\
& \underline{Best other baseline:} &  & &  &  & &  \\
& node2vec & 41.64 $\pm$ 1.25 & 34.30 $\pm$ 1.92 & 36.25 $\pm$ 1.38 & 29.43 $\pm$ 2.21 & 82.43 $\pm$ 1.23 & 81.60 $\pm$ 0.91 \\
\midrule
& Linear Modularity-Aware VGAE & \textbf{52.61} $\pm$ \textbf{1.41} & \textbf{45.74} $\pm$ \textbf{2.02} & \textbf{49.70} $\pm$ \textbf{2.04} & \textbf{43.64} $\pm$ \textbf{3.51} & \textbf{93.10} $\pm$ \textbf{0.88} & \textbf{94.06} $\pm$ \textbf{0.75} \\
\textbf{Cora} & Linear Standard VGAE & 49.98 $\pm$ 2.40 & \textit{43.15} $\pm$ \textit{4.35} & 46.90 $\pm$ 1.43 & 38.24 $\pm$ 3.56 & \textit{93.04} $\pm$ \textit{0.80} & \textit{94.04} $\pm$ \textit{0.75} \\ 
\textbf{with} & Louvain & 42.70 $\pm$ 0.65 & 24.01 $\pm$ 1.70 & 39.09 $\pm$ 0.73 & 20.19 $\pm$ 1.73 & -- & -- \\
\textbf{features} & \underline{Best other baseline:} &  &  &  &  &  & \\
& VGAECD-OPT & 50.32 $\pm$ 1.95 & \textit{43.54} $\pm$ \textit{3.23} & 47.83 $\pm$ 1.64 & 39.45 $\pm$ 3.53 & \textit{92.25} $\pm$ \textit{1.07} & 92.60 $\pm$ 0.91 \\
\midrule

& Linear Modularity-Aware VGAE & 21.28 $\pm$ 1.03 & \textbf{15.39} $\pm$ \textbf{1.06} & 19.05 $\pm$ 1.47 & \textbf{12.19} $\pm$ \textbf{1.38} & \textbf{80.84} $\pm$ \textbf{1.64} & \textbf{84.21} $\pm$ \textbf{1.21} \\
& Linear Standard VGAE & 13.83 $\pm$ 1.00 & 8.31 $\pm$ 0.89 & 11.11 $\pm$ 1.10 & 5.87 $\pm$ 0.87 & 78.26 $\pm$ 1.55 & \textit{82.93} $\pm$ \textit{1.39} \\ 
\textbf{Citeseer} & Louvain & \textbf{24.72} $\pm$ \textbf{0.27} & 9.21 $\pm$ 0.75 & \textbf{22.71} $\pm$ \textbf{0.47} & 7.70 $\pm$ 0.67 & -- & -- \\
& \underline{Best other baseline:} &  &  &  &  &  & \\
& node2vec & 18.68 $\pm$ 1.13  & \textit{14.93} $\pm$ \textit{1.15} & 14.40 $\pm$ 1.18 & \textit{12.13} $\pm$ \textit{1.53} & 76.05 $\pm$ 2.12 & 79.46 $\pm$ 1.65 \\
\midrule

& Linear Modularity-Aware VGAE  & \textbf{25.11} $\pm$ \textbf{0.94} & \textbf{15.55} $\pm$ \textbf{0.60} & \textit{22.21} $\pm$ \textit{1.24} & \textbf{12.59} $\pm$ \textbf{1.25} & 86.54 $\pm$ 1.20 & 88.07 $\pm$ 1.22 \\
\textbf{Citeseer} & Linear Standard VGAE & 17.80 $\pm$ 1.61 & 6.01 $\pm$ 1.46 & 17.38 $\pm$ 1.43 & 6.10 $\pm$ 1.51 & \textbf{89.08} $\pm$ \textbf{1.19} & \textbf{91.19} $\pm$ \textbf{0.98} \\ 
\textbf{with} & Louvain & 24.72 $\pm$ 0.27 & 9.21 $\pm$ 0.75 & \textbf{22.71} $\pm$ \textbf{0.47} & 7.70 $\pm$ 0.67 & -- & -- \\
\textbf{features} & \underline{Best other baseline:}&  &  &  &  &  & \\
& DVGAE & 20.09 $\pm$ 2.84 & 12.16 $\pm$ 2.74 & 16.02 $\pm$ 3.32 & \textit{10.03} $\pm$ \textit{4.48} & 86.85 $\pm$ 1.48 & 88.43 $\pm$ 1.23 \\
\midrule
& Linear Modularity-Aware GAE & \textbf{28.54} $\pm$ \textbf{0.24} & 26.36 $\pm$ 0.34 & \textbf{26.38} $\pm$ \textbf{0.43} & 21.30 $\pm$ 0.59 & \textbf{84.39} $\pm$ \textbf{0.32} & \textbf{87.92} $\pm$ \textbf{0.40} \\
& Linear Standard GAE & 12.61 $\pm$ 4.61 & 6.37 $\pm$ 3.86  & 12.60 $\pm$ 4.67 & 6.21 $\pm$ 1.75 & 82.03 $\pm$ 0.32 & \textit{87.71} $\pm$ \textit{0.24} \\
\textbf{Pubmed} & Louvain & 20.06 $\pm$ 0.27 & 10.34 $\pm$ 0.99 & 16.71 $\pm$ 0.46 & 8.32 $\pm$ 0.79 & -- & -- \\
& \underline{Best other baseline:} &  &  &  &  &  & \\
& node2vec & \textit{28.52} $\pm$ \textit{1.12} & \textbf{30.63} $\pm$ \textbf{1.14} & 23.88 $\pm$ 0.54 & \textbf{25.90} $\pm$ \textbf{0.65} & 81.03 $\pm$ 0.30 & 82.33 $\pm$ 0.41 \\
\midrule
& Linear Modularity-Aware VGAE & 30.09 $\pm$ 0.63 & \textbf{29.11} $\pm$ \textbf{0.65} & \textbf{29.60} $\pm$ \textbf{0.70} & \textbf{28.54} $\pm$ \textbf{0.74} & \textit{97.10} $\pm$ \textit{0.21} & \textbf{97.21} $\pm$ \textbf{0.18} \\
\textbf{Pubmed} & Linear Standard VGAE & 29.98 $\pm$ 0.41 & \textit{29.05} $\pm$ \textit{0.20} & \textit{29.51} $\pm$ \textit{0.52} & \textit{28.50} $\pm$ \textit{0.36} & \textbf{97.12} $\pm$ \textbf{0.20} & \textit{97.20} $\pm$ \textit{0.17} \\ 
\textbf{with}  & Louvain & 20.06 $\pm$ 0.27 & 10.34 $\pm$ 0.99 & 16.71 $\pm$ 0.46 & 8.32 $\pm$ 0.79 & -- & -- \\
\textbf{features} & \underline{Best other baseline:} &  &  &  &  &  & \\
& VGAECD-OPT & \textbf{32.47} $\pm$ \textbf{0.45} & \textit{29.09} $\pm$ \textit{0.42} & \textit{29.46} $\pm$ \textit{0.52} & \textit{28.43} $\pm$ \textit{0.61} & 94.27 $\pm$ 0.33 & 94.53 $\pm$ 0.36 \\
\midrule
& Linear Modularity-Aware VGAE & \textbf{48.55} $\pm$ \textbf{0.18} & \textbf{22.21} $\pm$ \textbf{0.39} & \textbf{46.10} $\pm$ \textbf{0.29} & \textbf{20.53} $\pm$ \textbf{0.38} & \textbf{95.76} $\pm$ \textbf{0.17} & \textbf{96.31} $\pm$ \textbf{0.12} \\
& Linear Standard VGAE & 46.07 $\pm$ 0.54 & 20.01 $\pm$ 0.90 & 43.38 $\pm$ 0.37 & 18.02 $\pm$ 0.66 & \textit{95.55} $\pm$ \textit{0.22} & \textit{96.30} $\pm$ \textit{0.18} \\ 
\textbf{Cora-Larger} & Louvain & 44.72 $\pm$ 0.50 & 19.46 $\pm$ 0.66 & 43.41 $\pm$ 0.52 & 19.29 $\pm$ 0.68 & -- & -- \\
& \underline{Best other baseline:}&  &  &  &  &  & \\
& DVGAE & 46.63 $\pm$ 0.56 & 20.72 $\pm$ 0.96 & 43.48 $\pm$ 0.61 & 18.45 $\pm$ 0.67 & 94.97 $\pm$ 0.23 & 95.98 $\pm$ 0.21 \\
\midrule
& Linear Modularity-Aware VGAE & \textbf{36.08} $\pm$ \textbf{0.13} & \textbf{8.11} $\pm$ \textbf{0.10} & \textbf{35.85} $\pm$ \textbf{0.20} & \textbf{8.06} $\pm$ \textbf{0.11} & \textit{82.34} $\pm$ \textit{0.38} & \textit{86.76} $\pm$ \textit{0.41} \\
& Linear Standard VGAE  & 35.01 $\pm$ 0.21 & 7.88 $\pm$ 0.15 & 30.79 $\pm$ 0.21 & 6.50 $\pm$ 0.13 & 80.11 $\pm$ 0.35 & 83.40 $\pm$ 0.36 \\ 
\textbf{SBM} & Louvain & \textit{36.06} $\pm$ \textit{0.12} & \textbf{8.11} $\pm$ \textbf{0.10} & \textit{35.84} $\pm$ \textit{0.18} & \textit{8.03} $\pm$ \textit{0.09} & -- & -- \\
& \underline{Best other baseline:} &  &  &  &  &  & \\
& DVGAE & \textit{35.90} $\pm$ \textit{0.18} & \textit{8.07} $\pm$ \textit{0.15} & 35.53 $\pm$ 0.23 & \textit{7.95} $\pm$ \textit{0.19} & \textbf{82.59} $\pm$ \textbf{0.36} & \textbf{87.08} $\pm$ \textbf{0.40} \\
\midrule
& GCN-Based Modularity-Aware VGAE & \textbf{21.64} $\pm$ \textbf{0.18} & \textbf{13.19} $\pm$ \textbf{0.09} & \textbf{19.10} $\pm$ \textbf{0.21} & \textbf{12.00} $\pm$ \textbf{0.17} & \textbf{85.40} $\pm$ \textbf{0.14} & \textit{86.38} $\pm$ \textit{0.15} \\
& GCN-Based Standard VGAE & 15.79 $\pm$ 0.32 & 9.75 $\pm$ 0.21 & 13.98 $\pm$ 0.35 & 8.81 $\pm$ 0.32 & \textit{85.37} $\pm$ \textit{0.12} & \textbf{86.41} $\pm$ \textbf{0.11} \\ 
\textbf{Albums} & Louvain & 19.81 $\pm$ 0.19 & 12.21 $\pm$ 0.09 & 17.68 $\pm$ 0.20 & 11.02 $\pm$ 0.13 & -- & -- \\
& \underline{Best other baseline:}&  &  &  &  &  & \\
& node2vec & 20.03 $\pm$ 0.24 & 12.20 $\pm$ 0.19 & 18.34 $\pm$ 0.29 & 11.27 $\pm$ 0.28 & 83.51 $\pm$ 0.17 & 84.12 $\pm$ 0.15 \\
\bottomrule
\end{tabular}}
\end{small}
\end{center}
\end{table}

\clearpage
Nevertheless, we recall that the use of the Louvain method was made without loss of generality. As explained in Section~\ref{c7s732}, our revised message passing operators would remain valid for other methods that alternatively derive a prior clustering signal. Future experiments on such alternatives (e.g., methods processing node features) could therefore improve community detection performances on these three attributed graphs. Overall, the empirical performance of our method directly depends on the quality of the underlying prior clustering method used to compute $\Ac$ and $\Ao$, which should therefore be carefully~selected. 

More broadly, our framework could also straightforwardly incorporate alternative encoders (besides linear and multi-layer GCN encoders), alternative decoders (e.g., decoders replacing inner~products by more refined graph reconstruction methods \cite{grover2019graphite,aaai20} including our gravity-inspired decoder from Chapter~\ref{chapter_5}) and alternative losses (for instance, as explained in Section~\ref{c7s733}, our modularity-inspired regularizer could be optimized in conjunction with the ELBO loss from VGAECD and VGAECD-OPT \cite{choong2018learning,choong2020optimizing} involving Gaussian mixtures). One could also replace our $k$-means step, to cluster vectors $z_i,$ with another method such as $k$-medoids \cite{park2009simple} or spectral clustering \cite{von2007tutorial} (although our preliminary experiments in this direction did not reach significantly better results). Future work considering such alternative architectures for Modularity-Aware GAE/VGAE could definitely lead to the improvement of our models.

Lastly, we would also need to extend Modularity-Aware GAE and VGAE to dynamic graphs. Indeed, while our work considered fixed graph structures, real-world graphs often evolve over time. For instance, as we will further explain and study in Chapter~\ref{chapter_9}, on the Deezer service, new albums regularly appear in the musical catalog. New nodes will therefore appear in the Albums graph. Capturing such changes, e.g., through dynamic embedding methods \cite{nguyen2014dynamic}, might permit learning more refined representations and provide effective dynamic community detection.

\section{Conclusion}
\label{c7s75}

In this chapter, we introduced an effective method for simultaneous link prediction and community detection, compatible with both the GAE and VGAE frameworks. 
This approach, referred to as Modularity-Aware GAE and VGAE, is based on a rigorous diagnosis of the shortcomings of existing approaches to this problem. 

Modularity-Aware GAE and VGAE take advantage of two elements: a theoretically grounded variant of message passing operator in the GAE and VGAE encoders, that incorporates prior cluster information, and the addition of a modularity-based loss component to the usual existing loss functions. Both elements were experimentally shown to have an individual impact on the community detection performances. We furthermore introduced a revised hyperparameter selection procedure specifically designed for joint link prediction and community detection. 
We experimentally demonstrated the effectiveness of the approach on several real-world datasets, both with node features and, crucially, featureless graphs. The results are consistently on par or better than popular baselines for both link prediction and community detection.

Last, but not least, we identified several research directions that, in future studies, should lead to the extension and the improvement of our work. In particular, we mentioned potential extensions of our approach to dynamic graphs. In Chapter~\ref{chapter_9}, we will present some applications of Modularity-Aware GAE and VGAE models to recommendation problems on Deezer, that inherently incorporate a dynamic aspect.

\section{Appendices}
\label{c7s76}

In this supplementary section, we prove the propositions of Subsection~\ref{s324}. These proofs were placed out of the main content of Chapter~\ref{chapter_7} for the sake of brevity and readability.

\subsection*{Preliminaries}\label{app:spectral_results}
We begin by introducing several theoretical results which we will use in our proofs. The specific formulations of the results in this section, i.e., Definition \ref{def:matrix_poly} and Propositions \ref{thm:poly_transf}, \ref{thm:union_of_spectra} and \ref{thm:similar}, are adapted from Lutzeyer~\cite{Lutzeyer2020}. When considering regular graphs, i.e., graphs containing only nodes of equal degree, their different graph representation matrices, such as the adjacency matrix, Laplacian matrices, and the GCN's message passing operator, are related via polynomial matrix transformations. These are now defined. 

\begin{definition} \label{def:matrix_poly} Horn and Johnson~\cite{Horn1985} define the evaluation of a polynomial $p(x) = c_l x^l + c_{l-1} x^{l-1} +\ldots + c_1 x + c_0$ at a matrix $\mati$ as
\begin{equation}
p(\mati) =c_l \mati^l + c_{l-1} \mati^{l-1} +\ldots + c_1 \mati + c_0 I.
\end{equation}
\end{definition}

Horn and Johnson~\cite{Horn1985} further include a discussion of the influence of a polynomial matrix transformation on the matrices' eigenvalues and eigenvectors, which we reproduce below. 

\begin{proposition} \label{thm:poly_transf} Let $p(\cdot)$ be a given polynomial. If $\evali$ is an eigenvalue of $\mati \in \mathbb{R}^{n \times n}$, while $\eveci$ is an associated eigenvector, then $p(\evali)$ is an eigenvalue of the matrix $p(\mati)$ and $\eveci$ is an eigenvector of $p(\mati)$ associated with $p(\evali)$.
\end{proposition}

Since we consider graphs consisting of several connected components in a multitude of our propositions, we now provide a theorem which relates the eigenvalues and eigenvectors of the whole graph to those of its connected components. 

\begin{proposition} \label{thm:union_of_spectra}
Let $\mathcal{G}$ be a graph with corresponding adjacency matrix $A$ and assume $\mathcal{G}$ to consist of $\ncluster$ connected components each with corresponding adjacency matrix $A_\clustersubscript$ for $\clustersubscript\in\{1, \ldots, \ncluster\}.$ Then, the eigenvalues of $\agcn{A}$ are equal to the union of the eigenvalues of $\agcn{A_\clustersubscript}$ over $\clustersubscript\in\{1, \ldots, \ncluster\}.$ Further, a set of eigenvectors of $\agcn{A}$ can be constructed from the eigenvector sets of $\agcn{A_\clustersubscript}$ for $\clustersubscript\in\{1, \ldots, \ncluster\}.$ 
\end{proposition}

We refer to \cite{salhagalvan2022modularity} for the proof of Proposition~\ref{thm:union_of_spectra}. To allow us to relate the spectra and eigenvectors of $A$ to the more novel GCN message passing operator $\agcn{A}$ we frequently make use of the matrix similarity relationship. The consequence of a matrix similarity relationship between matrices on their eigenvalues and eigenvectors is discussed in Proposition~\ref{thm:similar}.

\begin{proposition}\label{thm:similar}
\cite{Horn1985} If two matrices $\mati$ and $\matii$ are related via a nonsingular matrix $S$ as follows, $\mati = S^{-1}\matii S.$ Then, $\mati$ and $\matii$ have the same multiset of eigenvalues. Further, for eigenvector $v$ with corresponding eigenvalue $\evali$ of $\mati$ gives rise to an eigenvector $Sv$ of $\matii$ with equal corresponding eigenvalue $\evali.$
\end{proposition}

Equipped with such a background, we can now prove the propositions of Subsection \ref{s324}.

\subsection*{Proof of Proposition \ref{thm:Acspectrum}} \label{app:proof_Acspectrum}

Teke and Vaidyanathan~\cite{Teke2017}, among others, state that the unnormalized Laplacian matrix $L=D-A$ corresponding to a complete graph has eigenvalue $0$ with multiplicity $1$ and eigenvalue $n$ with multiplicity $n-1.$ 
Furthermore, the eigenspace corresponding to the eigenvalue $n$ is spanned by a 2-sparse set of orthogonal eigenvectors \cite{Teke2017}.
In addition, the eigenvector corresponding to the eigenvalue $0$ of the unnormalized graph Laplacian describing a connected graph is well known to be the constant eigenvector \cite{von2007tutorial}. 

Since the complete graph is regular, its degree matrix is a multiple of the identity matrix, i.e., $D=(n-1)I_n.$ Therefore, for complete graphs the following relationship holds $\agcn{A} = I_n - \frac{1}{n}L.$ Hence, from Proposition \ref{thm:poly_transf} the eigenvectors of $\agcn{A}$ and $L$ are equal and $\agcn{A}$ has the eigenvalue $1$ with multiplicity 1 and eigenvalue $0$ with multiplicity $n-1.$
Now, since $\Ac$ corresponds to a graph composed of several complete graphs, we can invoke Proposition \ref{thm:union_of_spectra} to construct the spectrum and eigenvectors of $\agcn{\Ac}$ from the the spectrum and eigenvectors of $\agcn{A}$ corresponding to a complete graph, which we just derived. Consequently, $\agcn{\Ac}$ has eigenvalues $\{\{1\}^{\ncluster}, \{0\}^{n-\ncluster}\}$ and a set of eigenvectors as described in the statement of Proposition~\ref{thm:Acspectrum}.

\subsection*{Proof of Proposition  \ref{thm:spectral_relation_A_Ac_AAc}} \label{app:proof_spectral_relation_A_Ac_AAc}

Proposition \ref{thm:union_of_spectra} can be used to extend the required result from one connected component of $\mathcal{G}$ to the full graph. Therefore, we consider only one connected component on the graph from now on. 
Let $\Acomp$ denote the adjacency matrix of this connected component, containing nodes of degree $b,$ and $\Accomp$ denote the corresponding complete component of $\Ac,$ containing nodes of degree $\ncomp-1.$ Then, the adjacency matrix of our connected component under consideration $\Acomp+\weight \Accomp$ is related to $\agcn{\Acomp + \weight \Accomp}$ as follows:
\begin{equation}
    \agcn{\Acomp+\weight \Accomp} = \frac{1}{b+\lambda (\ncomp-1) +1} \left(\Acomp+\weight \Accomp+I\right).
    \end{equation}
Therefore, from Proposition \ref{thm:poly_transf} it follows that $\Acomp+\weight \Accomp$ and $\agcn{\Acomp + \weight \Accomp}$ share eigenvectors. Similarly, the relations $\agcn{\Acomp} = \frac{1}{b+1} (\Acomp+I)$ and $\agcn{\Accomp} = \frac{1}{\ncomp} (\Accomp+I)$ together with Proposition \ref{thm:poly_transf} allow us to establish that both $\agcn{\Acomp}$ and $\Acomp$ as well as $\agcn{\Accomp}$ and $\Accomp$ each have a common set of eigenvectors.

We now make use of a result by Godsil~\cite{Godsil1993}, which states that the adjacency matrix of a graph commutes with the matrix of all ones, i.e., $\Ac+I_n,$ if and only if the graph under consideration is regular. Further, a family of matrices is a commuting family if and only if they are simultaneously diagonalizable, i.e., they share a set of eigenvectors \cite{Horn1985}.
Hence, $\Acomp$ and $\Accomp$ share a set of eigenvectors. Furthermore, this shared set of eigenvectors is also a valid set of eigenvectors for $\Acomp+\weight \Accomp,$ which, in conjunction with the above polynomial relationships, establishes the needed eigenvector relation.

In addition the eigenvalues of the sum of two simultaneously diagonalizable matrices $\mati, \matii$ with eigenvalues denoted by $\evali$ and $\evalii,$ respectively, are related \cite{Horn1985}  as follows
\begin{equation}\label{eq:diagonalisable_eval_relation}
\mathcal{S}(\mati +\matii) = \{\evali_1 + \evalii_{s(1)}, \ldots, \evali_n + \evalii_{s(n)}\},
\end{equation}
for some permutation $s(\cdot)$ defined on the set $\{1, \ldots, n\}.$ Since $\Acomp$ and $\Accomp+I_n$ are simultaneously diagonalizable their eigenvalues follow the relation in Equation \eqnref{eq:diagonalisable_eval_relation}. Now the above polynomial relationships of the GCN message passing operators to the corresponding adjacency matrices gives us the desired eigenvalue result and establish that $g_1(\mu) = \frac{b+1}{b+\lambda (\ncomp-1) +1}(\mu-1)$ and $g_2(\eta) = \frac{\lambda(\ncomp-1)+1}{b+\lambda (\ncomp-1) +1}(\eta-1)+1.$

\subsection*{Proof of Proposition \ref{thm:Aospectrum}}\label{app:proof_Aospectrum}

Hoory et al. \cite{Hoory2006} state that for $\dregc$-regular graphs the largest eigenvalue of the corresponding adjacency matrix $\Acomp$ equals $\dregc$ and the corresponding eigenvector is constant. Now the relation $\agcn{\Acomp} = \frac{1}{o+1}\left(\Acomp +I_n\right)$ in conjunction with Proposition \ref{thm:poly_transf} establish that the largest eigenvalue of $\agcn{\Acomp}$ equals 1 with a corresponding constant eigenvector. This spectrum and eigenvectors can be extended to the matrix $\agcn{\Ao}$ corresponding to a graph of several $\dregc$-regular connected components using Proposition \ref{thm:union_of_spectra}. The comparison of the derived spectrum and eigenvectors to those derived in Proposition \ref{thm:Acspectrum} completes this proof.

\part{Applications to Music Recommendation}
\label{partIII}

\chapter[Graph-Based Cold Start Similar Artists Ranking]{Graph-Based Cold Start Similar Artists Ranking}\label{chapter_8}
\chaptermark{Graph-Based Cold Start Similar Artists Ranking}

\textit{This chapter presents research conducted with Romain Hennequin, Benjamin Chapus, Viet-Anh Tran, and Michalis Vazirgiannis, and published in the proceedings of the 15\up{th} ACM Conference on Recommender Systems (RecSys 2021)~\cite{salha2021cold} where it received a ``best student paper'' honorable~mention.}

\section{Introduction}
\label{c8s81}

The previous Chapter~\ref{chapter_7} concluded the second part of this thesis, which detailed our technical contributions to improving node representation learning with GAE and VGAE models. In this Chapter~\ref{chapter_8}, we now start the Part~\ref{partIII}, which provides five additional and more ``applied'' chapters, presenting several industrial applications to music recommendation problems arising on music streaming services such as Deezer. Firstly, in this Chapter~\ref{chapter_8}, we propose a graph-based approach to tackle the \textit{cold start similar artists ranking} problem.

Music streaming services heavily rely on recommender systems to help users discover and enjoy new musical content within large catalogs of millions of songs, artists, and albums, with the general aim of improving their experience and engagement \cite{briand2021semi,mehrotra2019jointly,schedl2018current}. 
In particular, these services frequently recommend, on an artist's profile page, a ranked list of related artists that fans also listened to or liked \cite{donker2019networking,spotify2019fansalsolike,kjus2016musical}. 
Referred to as \textit{``Fans Also Like''} on Spotify and Soundcloud and as \textit{``Related''} or \textit{``Similar Artists''} on Amazon Music, Apple Music and Deezer, such a feature typically leverages \textit{learning to rank} models \cite{karatzoglou2013learning,rafailidis2017learning,schedl2018current}. It retrieves the most relevant artists according to similarity measures usually computed from usage data, e.g., from the proportion of shared listeners across artists \cite{donker2019networking,spotify2019fansalsolike}, or from more complex \textit{collaborative filtering} models \cite{jain2020survey,koren2015advances,schedl2018current} that predict similarities from the known preferences of an artist’s listeners.
It has been described as \textit{``one of the easiest ways''} to let \textit{``users discover new music''}~by~Spotify~\cite{spotify2019fansalsolike}.

However, filling up such ranked lists is especially challenging for new artists.
Indeed, while music streaming services might have access to some general descriptive information on these artists, listening data will however not be available upon their first release.
This prevents computing the aforementioned usage-based similarity measures.
As a consequence of this problem, which we refer to as \textit{cold start similar artists ranking}, music streaming services usually do not propose any \textit{``Fans Also Like''} section for these artists, until (and if ever) a sufficiently large number of usage interactions, e.g., listening sessions, has been reached. 
Besides new artists, this usage-based approach also excludes from recommendations a potentially large part of the existing catalog with too few listening data, which raises fairness concerns~\cite{corbett2018measure}. 
Furthermore, while we will focus on music streaming applications, this problem encompasses the more general \textit{cold start similar items ranking} issue, which is also crucial for media recommending other items~such~as~videos~\cite{covington2016deep}.

In this chapter, we address this problem by exploiting the fact that, as detailed in Section \ref{c8s83}, such \textit{``Fans Also Like''} features can naturally be summarized as a directed and attributed \textit{graph}, that connects each item \textit{node}, e.g., each artist, to their most similar neighbors via directed \textit{links}. Such a graph also incorporates additional descriptive information on nodes from the graph, e.g., musical information on artists. In this direction, we model cold start similar items ranking as a \textit{directed link prediction} problem \cite{salha2019-2}, for new nodes gradually added~into~this~graph.

Then, we solve this problem by leveraging our recent advances from Chapter~\ref{chapter_5}, and specifically our Gravity-Inspired GAEs and VGAEs. We propose a flexible framework which permits retrieving similar neighbors of new items from GAE/VGAE-based node embedding spaces, and where the gravity-inspired decoder acts as a ranking mechanism.  Backed by in-depth experiments on industrial data from Deezer, we show the effectiveness of our approach at addressing a real-world cold start similar artists ranking problem, outperforming several popular baselines for cold start recommendation. We publicly released our code and the industrial data~from~our~experiments. 

This chapter is organized as follows. In Section \ref{c8s82}, we introduce the cold start similar items ranking problem more precisely and mention previous works on related topics. In Section \ref{c8s83}, we present our graph-based framework to address this problem. We report and discuss our experiments on Deezer's data in Section \ref{c8s84}, and we conclude in Section \ref{c8s85}.

\section{Ranking Similar Artists/Items on Music Streaming Services}
\label{c8s82}

In this section, we introduce our ranking problem more precisely, and mention some related efforts to address cold start item recommendation problems. We voluntarily use a notation that heavily overlaps the one from the previous chapters, as the concepts introduced in this Section~\ref{c8s82} will be transposed to graphs in the remainder of the chapter (e.g., $n$, which denotes the number of items to recommend, will correspond to the number of nodes in Section~\ref{c8s83}).

\subsection{Problem Formulation}
\label{c8s821}

Throughout this chapter, we consider a catalog of $n$ recommendable items on an online service, such as music artists in our application. Each item $i$ is described by some side information summarized in an $f$-dimensional vector $x_i$. For artists, such a vector could for instance capture information related to their country of origin or their music genres. These $n$~items are assumed to be ``\textit{warm}'', meaning that the service considers that a sufficiently large number of interactions with users, e.g., likes or streams, has been reached for these items to ensure reliable usage~data~analyses. 

From these usage data, the service learns an $n \times n$ \textit{similarity matrix} $S$, where the element $S_{ij} \in [0, 1]$ captures the similarity of item $j$ w.r.t. item $i$. Examples of some possible usage-based similarity scores\footnote{Details on the computation of similarities at Deezer are provided in Section~\ref{c8s841} - without loss of generality, as our framework is valid~for~any~$S_{ij} \in [0, 1]$.} $S_{ij}$ include the percentage of users interacting with item $i$ that also interacted with item $j$ (e.g., users listening to or liking both items \cite{spotify2019fansalsolike}), mutual information scores \cite{shakibian2017mutual}, or more complex measures derived from collaborative filtering \cite{jain2020survey,koren2015advances,schedl2018current}. Throughout this chapter, we assume that similarity scores are fixed over time, which we later discuss.

\begin{figure}[t]
  \centering
  \includegraphics[width=0.95\linewidth]{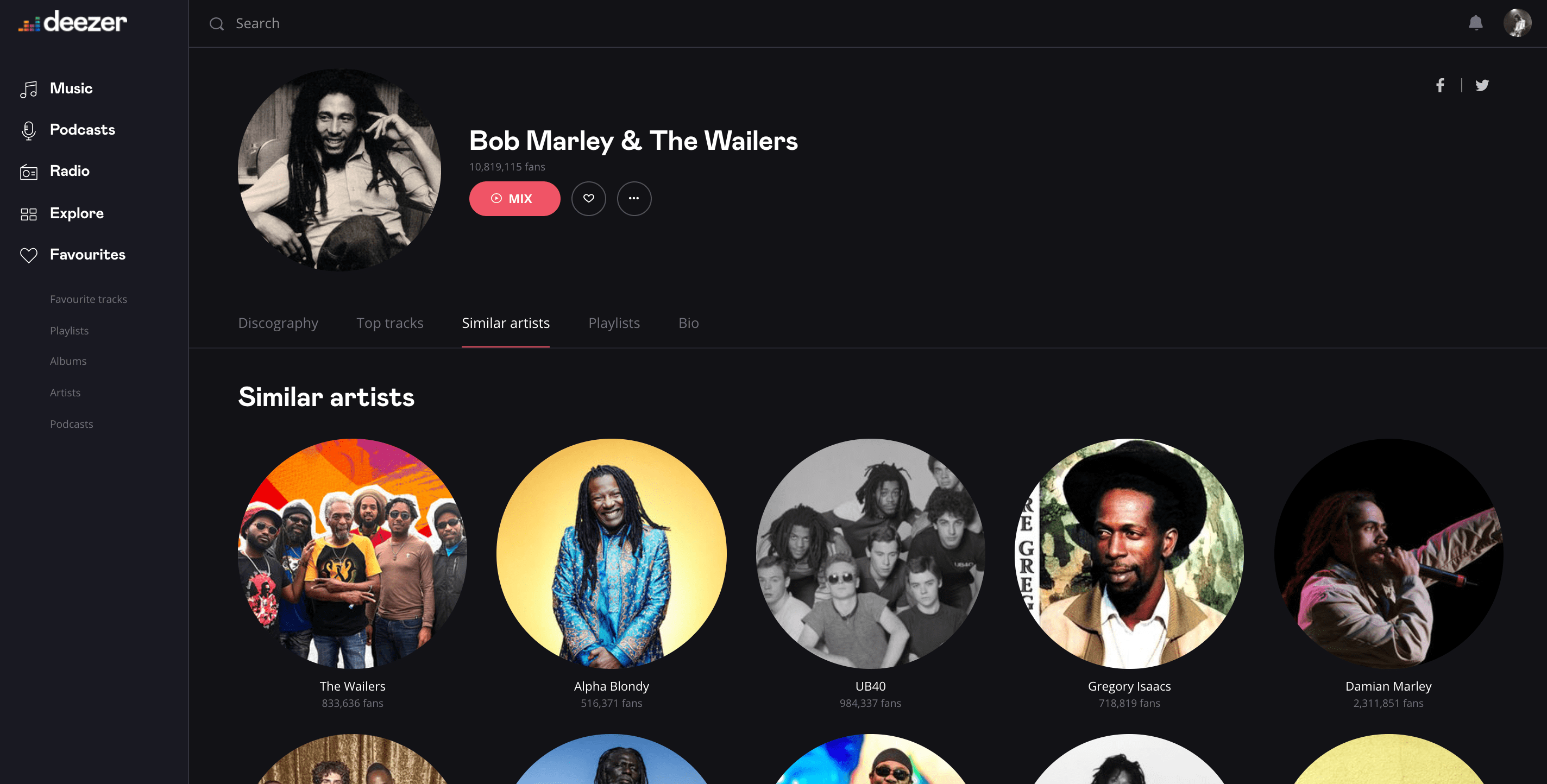}
\caption[Recommending similar artists on Deezer]{Some examples of ``similar artists'' recommended on the website version of Deezer. The mobile app proposes identical recommendations.}
  \label{c8_deezerexample}
\end{figure}

Leveraging these scores, the service proposes a \textit{similar items} feature comparable to the \textit{``Fans Also Like/Similar Items''} described in the introduction, and illustrated in Figure~\ref{c8_deezerexample}. Specifically, along with the presentation of an item $i$, it recommends a ranked list of $k$ similar items to users. They correspond to the top-$k$ items $j$ such as $j \neq i$ and with the highest~scores~$S_{ij}$.

On a regular basis, \textit{``cold''} items will appear in the catalog. While the service might have access to descriptive side information on these items, \textit{no usage data} will be available upon their first online release. This hinders the computation of usage-based similarity scores, thus excluding these items from recommendations until they become warm - if ever. 

In this chapter, we study the feasibility of effectively predicting their future similar items ranked lists, from the delivery of these items, i.e., without any usage data. This would enable offering such an important feature quicker and on a larger part of the catalog. More precisely, we answer the following research question: \textit{using only 1) the known similarity scores between warm items, and 2) the available descriptive information, how, and to which extent, can we predict the future ``Fans Also Like'' lists that would ultimately be computed~once cold items become warm?}

\subsection{Related Work}
\label{c8s822}

While collaborative filtering methods effectively learn item proximities, e.g., via the factorization of user-item
interaction matrices \cite{koren2015advances,van2013deep}, these methods usually become unsuitable for cold items without any interaction data and thus~absent from these matrices~\cite{van2013deep}. In such a setting, the simplest strategy for similar items ranking would consist in relying on \textit{popularity} metrics~\cite{schedl2018current}, e.g., to recommend the most listened artists. In the presence of descriptive information on cold items, one could also recommend items with the closest descriptions \cite{indyk1998approximate}. These heuristics are usually outperformed by \textit{hybrid} models, leveraging both item descriptions and collaborative filtering on warm items~\cite{he2016vbpr,hsieh2017collaborative,van2013deep,wang2018billion}.~They~consist~in:
\begin{itemize}
    \item learning a vector space representation (an \textit{embedding}) of warm items, where proximity aims to~reflect~user~preferences;
    \item  then, projecting cold items into this embedding, typically by learning a model to map descriptive vectors of warm items to their embedding vectors, and then applying this mapping to cold items' descriptive vectors.
\end{itemize}

Albeit under various formulations, this strategy has been transposed to Matrix Factorization~\cite{briand2021semi,van2013deep}, Collaborative Metric Learning~\cite{hsieh2017collaborative,lee2018collaborative} and Bayesian Personalized Ranking~\cite{barkan2019cb2cf,he2016vbpr}; in practice, a deep neural network often acts as the mapping model. The retrieved \textit{similar items} are then the closest ones in the embedding. Other deep learning approaches were also recently proposed for item cold start, with promising performances. DropoutNet \cite{volkovs2017dropoutnet} processes both usage and descriptive data, and is explicitly trained for cold start through a dropout \cite{srivastava2014dropout} simulation mechanism. MeLU (for Meta-Learned User preference estimator) \cite{lee2019melu} deploys a meta-learning paradigm to learn embeddings in the absence of usage data. CVAE (for Collaborative Variational Autoencoder) \cite{li2017collaborative} leverages a Bayesian generative process to sample cold embedding vectors, via a~variational~autoencoder.

While they constitute relevant baselines, these models do not rely on \textit{graphs}, contrary to our work. Graph-based recommendation has recently grown at a fast pace (see the surveys of \cite{wang2021graph,wu2020graph}), including in industrial applications~\cite{wang2018billion,ying2018graph}. Existing research widely focuses on bipartite user-item graphs \cite{wang2021graph}. Notably, STAR-GCN \cite{zhang2019star} addresses cold start by reconstructing user-item links using STAcked and Reconstructed GCNs, enhancing ideas from the GCN of Kipf and Welling~\cite{kipf2016-1} and a related extension for graph convolutional matrix completion by Berg~et~al.~\cite{berg2018matrixcomp}. Instead, recent efforts \cite{qian2} emphasized the relevance of leveraging - as we will - graphs connecting items together, along with their attributes. In this direction, the work closest to ours might be the recent DEAL (for Dual-Encoder graph embedding with ALignment) model by Hao~et~al.~\cite{ijcai2020DEAL}. Thanks to an alignment mechanism, DEAL predicts links in such graphs for new nodes having only descriptive information.
We will also compare to DEAL; we nonetheless point out that their work focused on undirected graphs, and did not consider~ranking~settings.

\section{A Graph-Based Framework for Cold-Start Similar Items Ranking}
\label{c8s83}

In this section, we present our graph-based framework to address this cold start ranking problem.

\subsection{Similar Items Ranking as a Directed Link Prediction Task}
\label{c8s831}

We argue that \textit{``Fans Also Like''} features can naturally be summarized as a graph structure with $n$ nodes and $n \times k$ edges. Nodes are warm recommendable items from the catalog, e.g., music artists with enough usage data according to the service's internal rules. Each item node points to its $k$ most similar neighbors via a link, i.e., an edge. This graph is:
\begin{itemize}
    \item \textit{directed}: edges have a direction, leading to asymmetric relations. For instance,
    while most fans of a little known reggae band might listen to Bob Marley (Marley thus appearing among their similar artists), Bob Marley's fans will rarely listen to this band, which is unlikely to appear back among Bob Marley’s own similar artists. 
    \item \textit{weighted}: among the $k$ neighbors of node $i$, some items are more similar to $i$ than others (hence the ranking). We capture this aspect by equipping each directed edge $(i,j)$ from the graph with a \textit{weight} corresponding to the similarity score $S_{ij}$. More formally, we summarize our graph structure by the $n \times n$ \textit{adjacency matrix} $A$, where the element $A_{ij} = S_{ij}$ if $j$ is one of the $k$ most similar items w.r.t. $i$, and where $A_{ij} = 0$ otherwise\footnote{Alternatively, one could consider a \textit{dense} matrix where $A_{ij} = S_{ij}$ for all pairs $(i,j)$. However, besides acting as a data cleaning process on lowest scores, sparsifying $A$ speeds up computations for the encoder introduced thereafter, whose complexity evolves linearly w.r.t. the number of edges~(see~Section~\ref{c8s833}).}.
    \item \textit{attributed}: as explained in Section \ref{c8s821}, each item $i$ is also described by a vector $x_i \in \mathbb{R}^f$. In the following, and consistently with previous chapters, we denote by $X$ the $n \times f$ matrix stacking up all descriptive vectors a.k.a. feature vectors from the graph, i.e., the $i$-th row of $X$ is $x_i$.
\end{itemize}

Then, we model the release of a cold recommendable item in the catalog \textit{as the addition of a new node} in the graph, together with its side descriptive vector. As usage data and similarity scores are unavailable for this item, it is \textit{observed as isolated}, i.e., it does not point to $k$ other nodes. In our framework, we assume that these $k$ missing directed edges~-~and their weights~- are actually \textit{masked}. They point to the $k$ nodes with the highest similarity scores, as would be identified by the service 
once it collects enough usage data to consider the item as warm, according to the service's criteria. These links and their scores, ultimately revealed, are treated as \textit{ground truth} in the remainder of this work.

From this perspective, the cold start similar items ranking problem consists in a \textit{directed link prediction} task \cite{lu2011link,schall2015link}. Specifically, we aim to predict the locations and weights - i.e., estimated similarity scores - of these $k$ missing directed edges, and at comparing predictions with the actual ground truth edges ultimately revealed, both in terms of:
\begin{itemize}
    \item \textit{prediction accuracy:} do we retrieve the correct locations of missing edges in the graph?
    \item \textit{ranking quality:} are the retrieved edges correctly ordered, in terms of similarity scores?
\end{itemize}

\subsection{From Similar Items Graphs to Directed Node Embeddings}
\label{c8s832}

As explained in Chapter~\ref{chapter_2}, locating missing links in graphs has been the objective of significant research efforts from various fields \cite{liben2007link,lu2011link,schall2015link}. While this problem has been historically addressed via the construction of hand-engineered node similarity metrics~\cite{liben2007link}, we explained in this same chapter that significant improvements were recently achieved by methods directly \textit{learning} node representations summarizing the graph structure \cite{hamilton2020graph,wu2019comprehensive}.
This includes the GAE and VGAE models at the center of this thesis, representing each node $i$ as a vector $z_i \in \mathbb{R}^d$ (with $d \ll n$) in a node embedding space where structural proximity should be preserved.

In this work, we build upon these advances and propose to learn node embeddings to tackle link prediction in our similar items graph. Specifically, we leverage our \textit{Gravity-Inspired GAEs and VGAEs} from Chapter~\ref{chapter_5}. These models are doubly advantageous for our application:
\begin{itemize}
    \item foremost, they can effectively \textit{process node attribute vectors} in addition to the graph, contrary to some popular alternatives such as DeepWalk  \cite{perozzi2014deepwalk} and standard Laplacian eigenmaps \cite{belkin2003laplacian}. As we explain thereafter in Section~\ref{c8s833}, this will help us add some cold nodes, isolated but equipped with some descriptions, into an existing warm node embedding. In other words, these models will be used in an \textit{inductive} setting that involves generalizing representations to new unseen nodes after training;
    \item simultaneously, our Gravity-Inspired GAE and VGAE models were specifically designed to address \textit{directed} link prediction from node embedding spaces, contrary to the aforementioned alternatives or to standard GAE and VGAE models from Kipf~and~Welling~\cite{kipf2016-2}. 
\end{itemize}

\subsection[Cold Start Similar Items Ranking using Gravity-Inspired GAE and VGAE]{Cold Start Similar Items Ranking using Gravity-Inspired GAE/VGAE}
\label{c8s833}

In this section, we now explain how we build upon our Gravity-Inspired GAE and VGAE models to address the cold start similar items ranking problem. As VGAEs emerged as competitive alternatives to GAEs on some link prediction experiments, including some of those from Part~\ref{partII}, we therefore saw value in considering both deterministic GAEs \textit{and} VGAEs in this chapter. We assume a good understanding of these models, and we refer in particular to our previous Chapter~\ref{chapter_5} for a broader introduction to our Gravity-Inspired GAE and VGAE models.

\paragraph{Encoding Cold Nodes with GCNs}
\label{s341}

In this chapter, and consistently with Chapter~\ref{chapter_5}, our encoders (both for Gravity-Inspired GAEs and VGAEs) will be 2-layer GCNs~\cite{kipf2016-1}, incorporating the \textit{out-degree normalized}\footnote{We recall that $\tilde{A}_{\text{out}} = (D_{\text{out}} + I_n)^{-1} (A + I_n)$ where $I_n$ is the $n \times n$ identity matrix and $D_{\text{out}}$ is the diagonal out-degree matrix~of~$A$.} version of $A$ denoted $\tilde{A}_{\text{out}}$ and a ReLU activation function~\cite{salha2019-2}. Therefore, adopting the notation from Chapter~\ref{chapter_5}, we have:
\begin{equation}
\tilde{Z} = \tilde{A}_{\text{out}}\text{ReLU} (\tilde{A}_{\text{out}}X W^{(0)}) W^{(1)},
\end{equation}
in the GAE setting. We recall that the $i$-th row of $\tilde{Z} \in \mathbb{R}^{n \times (d+1)}$ is a $(d+1)$-dimensional vector $\tilde{z}_i$. The $d$ first dimensions of $\tilde{z}_i$ correspond to the embedding vector $z_i$ of node $i$; the last dimension corresponds to the mass $\tilde{m}_{i}$ as defined in Section~\ref{c5s532}. 

Regarding the gravity-inpired VGAE model, and consistently with Chapter~\ref{chapter_5} once again, we have $\mu  =\tilde{A}_{\text{out}}\text{ReLU} (\tilde{A}_{\text{out}}X W^{(0)}_{\mu}) W^{(1)}_{\mu}$, $\log \sigma  =\tilde{A}_{\text{out}}\text{ReLU} (\tilde{A}_{\text{out}}X W^{(0)}_{\sigma}) W^{(1)}_{\sigma}$, and $\tilde{Z}$ is then sampled from $\mu$ and $\log \sigma$. 

As all outputs are $n \times (d+1)$ matrices and $X$ is an $n \times f$ matrix, then $W^{(0)}$ (or, $W^{(0)}_{\mu}$ and $W^{(0)}_{\sigma}$) is an $f \times d_{\text{hidden}}$ matrix, with $d_{\text{hidden}}$ the hidden layer dimension, and $W^{(1)}$ (or, $W^{(1)}_{\mu}$ and $W^{(1)}_{\sigma}$) is a $d_{\text{hidden}} \times (d+1)$~matrix. These GCN weights are optimized from the graph of warm recommendable items, in a similar fashion w.r.t. Chapter~\ref{chapter_5}, i.e., by iteratively optimizing a cross-entropy reconstruction loss (for Gravity-Inspired GAE) or an ELBO variational lower bound objective (for Gravity-Inspired VGAE), by gradient descent.

We rely on GCN encoders as they permit \textit{incorporating new nodes}, attributed but isolated in the graph, into an existing embedding. Indeed, let us consider a model already trained on a graph of warm items with some $A$ and $X$, leading to optimized weights $W^{(0)}$ and $W^{(1)}$ for some GCN encoder. If $m \geq 1$ cold nodes appear, along with their $f$-dimensional descriptive~vectors,~then:
\begin{itemize}
    \item $A$ becomes $A'$, an $(n+m) \times (n+m)$ adjacency matrix\footnote{Its $(n+m) \times (n+m)$ out-degree normalized version is $\tilde{A}_{\text{out}}'~ = (D_{\text{out}}^{'} + I_{(n+m)})^{-1} (A' + I_{(n+m)})$, where $D_{\text{out}}^{'}$ is the diagonal out-degree matrix of $A'$.}, with $m$ new rows and columns filled with zeros;
    \item $X$ becomes $X'$, an $(n+m) \times f$ attribute matrix, concatenating $X$ and the $f$-dimensional descriptions of the $m$ new nodes;
    \item we derive embedding vectors and masses of new nodes through a \textit{forward pass}\footnote{We note that such GCN forward pass is possible since dimensions of weight matrices $W^{(0)}$ and $W^{(1)}$ are \textit{independent} of the number of nodes. In the case of featureless nodes, i.e., with $X = I_n$ as in Equation~\eqref{eq:gcn} from Chapter~\ref{chapter_2}, the dimension of $W^{(0)}$ becomes $n \times d_{\text{hidden}}$ which therefore depends on $n$. This prevents computing forward passes in the presence of new nodes, thus making the GCN encoder \textit{transductive} by design.} into the GCN previously trained on warm nodes, i.e., by computing the $(n+m) \times (d+1)$ new embedding matrix $\tilde{Z}'~=$ $\tilde{A}^{'}_{\text{out}}~ \text{ReLU} (\tilde{A}^{'}_{\text{out}}~ X' W^{(0)}) W^{(1)}$.
\end{itemize} 
We emphasize that the choice of GCN encoders is made without loss of generality. Our framework remains valid for any inductive encoder processing new attributed nodes. In our experiments, 2-layer GCNs reached better or comparable results w.r.t. some considered alternatives, namely deeper GCNs, graph~attention~networks~\cite{velivckovic2019graph} and our linear~encoders~from~Chapter~\ref{chapter_6}.

\paragraph{Ranking Similar Items}
\label{s342}

After projecting cold nodes into the warm embedding, we use the gravity-inspired decoder to predict their masked connections. More precisely, in our experiments, we add the hyperparameter $\lambda \in \mathbb{R}^+$ from Equation~\eqref{eq:lambdac5} for flexibility. The \textit{estimated similarity weight} $\hat{A}_{ij}$ between some cold node $i$ and another node $j$ is thus:
\begin{equation}
\hat{A}_{ij} = \sigma(\underbrace{\tilde{m}_j}_{\text{influence of $j$}} - \lambda \times \underbrace{\log \|z_i - z_j\|_2^2}_{\text{proximity of $i$ and $j$}}).
\label{equation5}
\end{equation}
Then, the predicted top-$k$ most similar items of $i$ will correspond to the $k$ nodes $j$ with the highest estimated weights $\hat{A}_{ij}$.

As in Chapter~\ref{chapter_5}, we interpret Equation \eqref{equation5} in terms of influence/proximity trade-off:
\begin{itemize}
    \item the \textit{influence} part of Equation \eqref{equation5} indicates that, if two nodes $j$ and $l$ are equally close to $i$ in the embedding space (i.e., $\|z_i - z_j\|_2 = \|z_i - z_l\|_2$), then $i$ will more likely points towards the node with the largest mass (i.e., the largest \textit{``influence''}; we will compare these masses to popularity metrics in experiments from Section~\ref{c8s843});
    \item the \textit{proximity} part of Equation \eqref{equation5} indicates that, if $j$ and $l$ have the same mass (i.e., $\tilde{m}_j = \tilde{m}_l$), then $i$ will more likely points towards its closest neighbor, which could, e.g., capture a closer musical similarity for artists.
\end{itemize}
As illustrated in Section~\ref{c8s843}, tuning $\lambda$ will help us flexibly balance between these two aspects, and thus control for \textit{popularity biases}~\cite{schedl2018current} in our~recommendations.

\paragraph{On Complexity}
\label{s343}

As extensively explained throughout this thesis, training GAE/VGAE models via full-batch gradient descent requires reconstructing the entire matrix $\hat{A}$ at each training iteration, which has an $O(dn^2)$ time complexity due to the evaluation of pairwise distances~\cite{salha2019-2}. 

While we will follow such a full-batch training strategy in Section~\ref{c8s84}, our released code also implements the FastGAE method from Chapter~\ref{chapter_4} to approximate losses by decoding random subgraphs of $O(n)$ size. This permits scaling our method to large graphs with millions of nodes and edges. In our experiments, using FastGAE with degree-based sampling permits reducing training times from roughly 30 minutes to 1 minute on our machine and for the Deezer graph presented in Section~\ref{c8s841}, while preserving performances on the task presented in Section~\ref{c8s842}.

Moreover, projecting cold nodes in an embedding only requires a single forward GCN pass, with linear time complexity w.r.t. the number of edges \cite{kipf2016-1,salha2020simple}. This is another advantage of using GCNs w.r.t. more complex encoders. Lastly, retrieving the top-$k$ accelerations boils down to a nearest neighbors search in a $O(ndk)$ time, which could even be improved in our future research with approximate search methods \cite{indyk1998approximate}.

\section{Experimental Analysis}
\label{c8s84}

We now present the experimental evaluation of our graph-based framework on music artists data from the Deezer production system. We publicly released the private graph dataset used in these experiments on GitHub\footnote{\href{https://github.com/deezer/similar_artists_ranking}{https://github.com/deezer/similar\_artists\_ranking}}, as well as our source code for the experiments described thereafter. Besides making our results reproducible, such a release publicly provides a new benchmark dataset to the research community, permitting the evaluation of comparable graph-based recommender systems on~real-world~resources.

\subsection{Ranking Similar Artists on Deezer}
\label{c8s841}

\paragraph{Dataset}  We consider a directed graph of 24 270 artists with various musical characteristics (see below), extracted from the Deezer service. Each artist points towards $k=20$ other artists. They correspond, up to internal business rules, to the top-20 artists from the same graph that would be recommended by our production system on top of the \textit{``Fans Also Like/Similar Artists''} feature illustrated in Figure~\ref{c8_deezerexample}.
Each directed edge $(i,j)$ has a weight $A_{ij}$ normalized to lie in the $[0,1]$~set; for unconnected pairs, $A_{ij}= 0$.  It corresponds to the similarity score of artist $j$ w.r.t. $i$, computed on a weekly basis from usage data of millions of Deezer users. More precisely, weights are based on \textit{mutual information} scores \cite{shakibian2017mutual} from \textit{artist co-occurrences among streams}. Roughly, they compare the probability that a user listens to the two artists, to their global listening frequencies on the service, and they are normalized at the artist level through internal heuristics and business rules (some details on exact score computations are voluntarily omitted for confidentiality reasons). In the graph, edges correspond to the 20 highest scores for each node. In general, $A_{ij} \neq A_{ji}$. In particular, $j$ might be the most similar artist of $i$ while $i$ does not even appear among the top-20 of $j$.

We also have access to descriptions of these artists, either extracted through the musical content or provided by record labels. Here, each artist $i$  will be described by an attribute vector $x_i$ of dimension $f=56$, concatenating:
\begin{itemize}
    \item a 32-dimensional \textit{genre} vector. Deezer artists are described by music genres \cite{epure2020modeling}, among more than~300. 32-dimensional embedding vectors are learned from these genres, by factorizing a co-occurrence matrix based on listening usages with SVD \cite{koren2009matrix}. Then, the genre vector of an artist is the average of embedding vectors of~his/her~music~genres.
    \item a 20-dimensional \textit{country} vector. It corresponds to a one-hot encoding vector, indicating the country of origin of an artist, among the 19 most common countries on Deezer, and with a 20\up{th} category gathering all other countries.
    \item a 4-dimensional \textit{mood} vector. It indicates the average and standard deviations of the \textit{valence} and \textit{arousal} scores across an artist's discography. In a nutshell, valence captures whether each song has a positive or negative mood, while arousal captures whether each song has a calm or energetic mood \cite{delbouys2018music,russell1980circumplex}. These scores are computed internally, from audio data and using a deep neural network inspired by the work of Delbouys et al. \cite{delbouys2018music}.
\end{itemize}

While some of these features are quite general, we emphasize that the actual Deezer app also gathers more refined information on artists, e.g., from audio or textual descriptions. They are undisclosed and unused in these~experiments.

\paragraph{Problem} 
\label{s412}
We consider the following similar artists ranking problem.
Artists are split into a training set, a validation set, and a test set gathering 80\%, 10\%, and 10\% of artists, respectively. The training set corresponds to warm artists. Artists from the validation and test sets are the cold nodes: their edges are \textit{masked}, and they are therefore observed as isolated in the graph. Their 56-dimensional descriptions are available. 

We evaluate the ability of our models at retrieving these edges, with correct weight ordering. As a measure of \textit{prediction accuracy}, we will report \textit{Recall@K} scores. They indicate, for various $K$, which proportion of the 20 ground truth similar artists appear among the top-$K$ artists with the highest estimated weights. Moreover, as a measure of \textit{ranking quality}, we will also report the widely used \textit{Mean Average Precision~at}~$K$~(\textit{MAP@K}) and \textit{Normalized Discounted Cumulative Gain}~at~$K$~(\textit{NDCG@K}) scores\footnote{MAP@K and NDCG@K are computed as in  Equation~(4) of \cite{schedl2018current} (averaged over all cold artists) and in Equation~(2) of \cite{wang2013theoretical} respectively.}.

\subsection{List of Models and Baselines}
\label{c8s842}

We now describe all methods considered in our experiments. All embedding vectors have $d = 32$, which we will discuss. Also, all hyperparameters mentioned thereafter were tuned by optimizing NDCG@20 scores on the validation set.

\paragraph{Gravity-Inspired GAE/VGAE} We follow our framework from Section~\ref{c8s83} to embed cold nodes. For both GAE and VGAE, we use 2-layer GCN encoders with a 64-dimensional hidden layer, and a 33-dimensional output layer (i.e., a 32-dimensional $z_i$ vectors, plus the mass), trained for 300 epochs. We use the Adam optimizer \cite{kingma2014adam}, with a learning rate of 0.05, without dropout, performing full-batch gradient descent, and using the reparameterization trick \cite{kingma2013vae} for VGAE. We set $\lambda$ = 5  in the decoder of Equation~\eqref{equation5} and discuss the impact of $\lambda$ thereafter. Our adaptation of these models builds upon the Tensorflow~code that we developed for the experiments of Chapter~\ref{chapter_5}, which is publicly available on GitHub\footnote{\href{https://github.com/deezer/gravity_graph_autoencoders}{https://github.com/deezer/gravity\_graph\_autoencoders}}.

\paragraph{Other Methods based on the Directed Artist Graph} We compare our Gravity-Inspired GAE and VGAE models to the \textit{standard GAE and VGAE} models from Kipf and Welling~\cite{kipf2016-2}, with a similar setting as above. These models use symmetric inner product decoders, i.e., $\hat{A}_{ij} = \sigma(z^T_i z_j)$, therefore ignoring directionalities. Moreover, we implement the \textit{Source-Target GAE and VGAE} models used as a baseline in Chapter~\ref{chapter_5}. We recall that these models are similar to standard GAE and VGAE, except that they decompose the 32-dimensional $z_i$ vectors into a \textit{source} vector $z^{(s)}_i = z_{i[1: 16]}$ and a \textit{target} vector $z^{(t)}_i = z_{i[17:32]}$, and then decode edges as follows: $\hat{A}_{ij} = \sigma(z^{(s)T}_i z^{(t)}_j)$ and $\hat{A}_{ji} = \sigma(z^{(s)T}_j z^{(t)}_i)$ ($\neq \hat{A}_{ij}$ in general). They reconstruct directed links, as gravity-inspired models, and are therefore relevant baselines for our evaluation. Lastly, we also test the recent DEAL model \cite{ijcai2020DEAL} mentioned in Section \ref{c8s822}, and designed for inductive link prediction on new isolated but attributed nodes. We used the authors' PyTorch implementation, with similar attribute and structure encoders, alignment mechanism, loss and cosine predictions as their original~model~\cite{ijcai2020DEAL}.

\paragraph{Other Baselines} 

In addition, we compare our framework to four popularity-based baselines.
\textit{Popularity} recommends the $K$ most popular\footnote{Our dataset includes the \textit{popularity} rank (from 1\up{st} to $n$\up{th}) of warm artists. It is out of the $x_i$ vectors, as it is usage-based and thus unavailable for cold~artists.} artists on Deezer (with $K$ as~in~Section~\ref{s412});
\textit{Popularity by Country} recommends the $K$ most popular artists from the country of origin of the cold artist;
 \textit{In-Degree} recommends the $K$ artists with the highest in-degrees in the graph, i.e., sum of weights pointing to them; lastly,   \textit{In-Degree by Country} proceeds as \textit{In-Degree}, but on warm artists from the country of origin of the cold artist.
 
We also consider three baselines only or mainly based on descriptions $x_i$ and not on usage data. Firstly, \textit{$K$-NN} recommends the $K$ artists with closest $x_i$ vectors, from a nearest neighbors search with Euclidean distance.
Besides, \textit{$K$-NN + Popularity} and \textit{$K$-NN + In-degree} retrieve the 200 artists with closest $x_i$ vectors, then recommends the $K$ most popular ones among these 200 artists, ranked according to popularity and in-degree values respectively.

We also implement \textit{SVD+DNN}, which follows the \textit{``embedding+mapping''} strategy from Section~\ref{c8s822} by 1) computing an SVD~\cite{koren2009matrix} of the warm artists similarity matrix, learning 32-dimensional $z_i$ vectors, 2) training a 3-layer neural network (with layers of dimension 64, 32 and 32, trained with Adam \cite{kingma2014adam} and a learning rate of 0.01) to map warm $x_i$ vectors to $z_i$ vectors, and 3) projecting cold artists into the SVD embedding through this mapping. 

Lastly, among deep learning approaches from Section \ref{c8s822} (CVAE, DropoutNet, MeLU, STAR-GCN), we report results from the two best methods on our dataset, namely \textit{DropoutNet} \cite{volkovs2017dropoutnet} and \textit{STAR-GCN} \cite{zhang2019star}, using the authors' implementations with careful fine-tuning on validation artists\footnote{These last models do not process similar artists graphs, but raw \textit{user-item usage data}, either as a bipartite user-artist graph or as an interaction matrix. While Deezer can not release such fine-grained data, we nonetheless provide embedding vectors from these baselines along with our code, to reproduce our scores.}. Similar artists ranking is done via a nearest neighbors search in~embedding~spaces.

\subsection{Results and Discussion}
\label{c8s843}

\begin{table}[t]
\centering
\caption[Cold start similar artists ranking with Gravity-Inspired GAE and VGAE]{Cold start similar artists ranking with Gravity-Inspired GAE and VGAE and with all baselines. Performances are computed on test set, and averaged over 20 runs. All embedding methods verify $d = 32$. \textbf{Bold}~numbers correspond to the best scores.}
\resizebox{1.0\textwidth}{!}{
\begin{tabular}{r|ccc|ccc|ccc}
\toprule
\textbf{Method} & \multicolumn{3}{c}{\textbf{Recall@K (in \%)}} & \multicolumn{3}{c}{\textbf{MAP@K (in \%)}} & \multicolumn{3}{c}{\textbf{NDCG@K (in \%)}} \\
($d = 32$)&  \scriptsize \textbf{$K=20$} & \scriptsize \textbf{$K=100$} & \scriptsize \textbf{$K=200$} & \scriptsize \textbf{$K=20$} & \scriptsize \textbf{$K=100$} & \scriptsize \textbf{$K=200$} & \scriptsize \textbf{$K=20$} & \scriptsize \textbf{$K=100$} & \scriptsize \textbf{$K=200$} \\
\midrule
\midrule 
Popularity & 0.02 & 0.44 & 1.38 & <0.01 & 0.03 & 0.12 & 0.01 & 0.17 & 0.44 \\ 
Popularity by country & 2.76 & 12.38 & 18.98 & 0.80 & 3.58 & 6.14 & 2.14 & 6.41 & 8.76 \\ 
In-degree & 0.91 & 3.43 & 6.85 & 0.15 & 0.39 & 0.86 & 0.67 & 1.69 & 2.80 \\ 
In-degree by country & 5.46 & 16.82 & 23.52 & 2.09 & 5.43 & 7.73 & 5.00 & 10.19 & 12.64 \\ 
$K$-NN on $x_i$ & 4.41 & 13.54 & 19.80 & 1.14 & 3.38 & 5.39 & 4.29 & 8.83 & 11.22\\ 
$K$-NN + Popularity & 5.73 & 15.87 & 19.83 & 1.66 & 4.32 & 5.74 & 4.86 & 10.03 & 11.76 \\ 
$K$-NN + In-degree & 7.49 & 17.29 & 18.76 & 2.78 & 5.60 & 6.18 & 7.41 & 12.48 & 13.14  \\ 
SVD + DNN  & 6.42 $\pm$ 0.96 & 21.83 $\pm$ 1.21 & 35.01 $\pm$ 1.41 & 2.25 $\pm$ 0.67 & 6.36 $\pm$ 1.19 & 11.52 $\pm$ 1.98 & 6.05 $\pm$ 0.75 & 12.91 $\pm$ 0.92 & 17.89 $\pm$ 0.95 \\ 
STAR-GCN & 10.03 $\pm$ 0.56 & 31.45 $\pm$ 1.09 & 43.92 $\pm$ 1.10 & 3.10 $\pm$ 0.32 & 10.64 $\pm$ 0.54 & 16.62 $\pm$ 0.68 & 10.07 $\pm$ 0.40 & 21.17 $\pm$ 0.69 & 25.99 $\pm$ 0.75 \\ 
DropoutNet & 12.96 $\pm$ 0.54 & 37.59 $\pm$ 0.76 & 49.93 $\pm$ 0.82 & 4.18 $\pm$ 0.30 & 13.61 $\pm$ 0.55 & 20.12 $\pm$ 0.67 & 13.12 $\pm$ 0.68 & 25.61 $\pm$ 0.72 & 30.52 $\pm$ 0.78 \\ 
DEAL  & 12.80 $\pm$ 0.52 & 37.98 $\pm$ 0.59 & 50.75 $\pm$ 0.72 & 4.15 $\pm$ 0.25 & 14.01 $\pm$ 0.44 & 20.92 $\pm$ 0.54 & 12.78 $\pm$ 0.53 & 25.70 $\pm$ 0.62 & 30.69 $\pm$ 0.70 \\ 
GAE & 7.30 $\pm$ 0.51 & 25.92 $\pm$ 0.95 & 40.37 $\pm$ 1.11 & 2.81 $\pm$ 0.29 & 7.97 $\pm$ 0.47 & 14.24 $\pm$ 0.67 & 6.32 $\pm$ 0.39 & 15.54 $\pm$ 0.66 & 20.94 $\pm$ 0.72 \\ 
VGAE & 10.01 $\pm$ 0.52 & 34.00 $\pm$ 1.06 & 49.72 $\pm$ 1.14 & 3.53 $\pm$ 0.27 & 11.68 $\pm$ 0.52 & 19.46 $\pm$ 0.70 & 10.09 $\pm$ 0.58 & 21.37 $\pm$ 0.73 & 27.31 $\pm$ 0.75 \\ 
Sour.-Targ. GAE & 12.21 $\pm$ 1.30 & 39.52 $\pm$ 3.53 & 56.25 $\pm$ 3.57 & 4.62 $\pm$ 0.81 & 14.67 $\pm$ 2.33 & 23.60 $\pm$ 2.85 & 12.42 $\pm$ 1.39 & 25.45 $\pm$ 3.37 & 31.80 $\pm$ 3.38 \\ 
Sour.-Targ. VGAE & 13.52 $\pm$ 0.64 & 42.68 $\pm$ 0.69 & 59.51 $\pm$ 0.76 & 5.19 $\pm$ 0.31 & 16.07 $\pm$ 0.40 & 25.48 $\pm$ 0.55 & 13.60 $\pm$ 0.73 & 27.81 $\pm$ 0.56 & 34.19 $\pm$ 0.59 \\ 
\midrule%
\textbf{Gravity GAE} & \textbf{18.33} $\pm$ \textbf{0.45} & \textbf{52.26} $\pm$ \textbf{0.90} & \textbf{67.85} $\pm$ \textbf{0.98} & \textbf{6.64} $\pm$ \textbf{0.25} & \textbf{21.19} $\pm$ \textbf{0.55} & \textbf{30.67} $\pm$ \textbf{0.68} & \textbf{18.64} $\pm$ \textbf{0.47} & \textbf{35.77} $\pm$ \textbf{0.66} & \textbf{41.42} $\pm$ \textbf{0.68} \\ 
\textbf{Gravity VGAE} & 16.59 $\pm$ 0.50 & 49.51 $\pm$ 0.78 & 65.70 $\pm$ 0.75 & 5.66 $\pm$ 0.35 & 19.07 $\pm$ 0.57 & 28.66 $\pm$ 0.59 & 16.74 $\pm$ 0.55 & 33.34 $\pm$ 0.66 & 39.29 $\pm$ 0.64 \\ 
\bottomrule
\end{tabular}}
\label{tableresults}
\end{table}

\paragraph{Performances} Table~\ref{tableresults} reports mean performance scores for all models, along with standard deviations
over 20 runs for models with randomness due to weights initialization in GCNs or neural networks. Popularity and In-degree appear as the worst baselines. Their scores significantly improve by focusing on the country of origin of cold artists (e.g., with a Recall@100 of 12.38\% for Popularity by country, v.s. 0.44\% for Popularity).

Besides, we observe that methods based on a direct $K$-NN search from $x_i$ attributes are outperformed by the more elaborated cold start methods leveraging both attributes and warm usage data. In particular, DropoutNet, as well as the graph-based DEAL, reach stronger results than SVD+DNN and STAR-GCN. They also surpass standard GAE and VGAE (e.g., with a +6.46 gain in average NDCG@20 score for DEAL w.r.t. GAE), but not the GAE/VGAE extensions that explicitly model edges directionalities, i.e., source-target GAE/VGAE and, even more, Gravity-Inspired GAE/VGAE, that provide the best recommendations. It emphasizes the effectiveness of our framework, both in terms of prediction accuracy (e.g., with a top 67.85\% average Recall@200 for Gravity-Inspired GAE) and of ranking quality (e.g., with a top 41.42\% average NDCG@200 for this same method).

Moreover, while all embedding methods from Table~\ref{tableresults} verify $d =32$, we point out that gravity-inspired models remained superior on our tests with $d=64$ and $d=128$. Also, VGAE methods tend to outperform their deterministic counterparts for standard and source-target models, while the contrary conclusion emerges on gravity-inspired models. This confirms the value of considering both settings when testing these models on novel applications. 

\begin{figure}[t]
  \centering
  \includegraphics[width=1.0\linewidth]{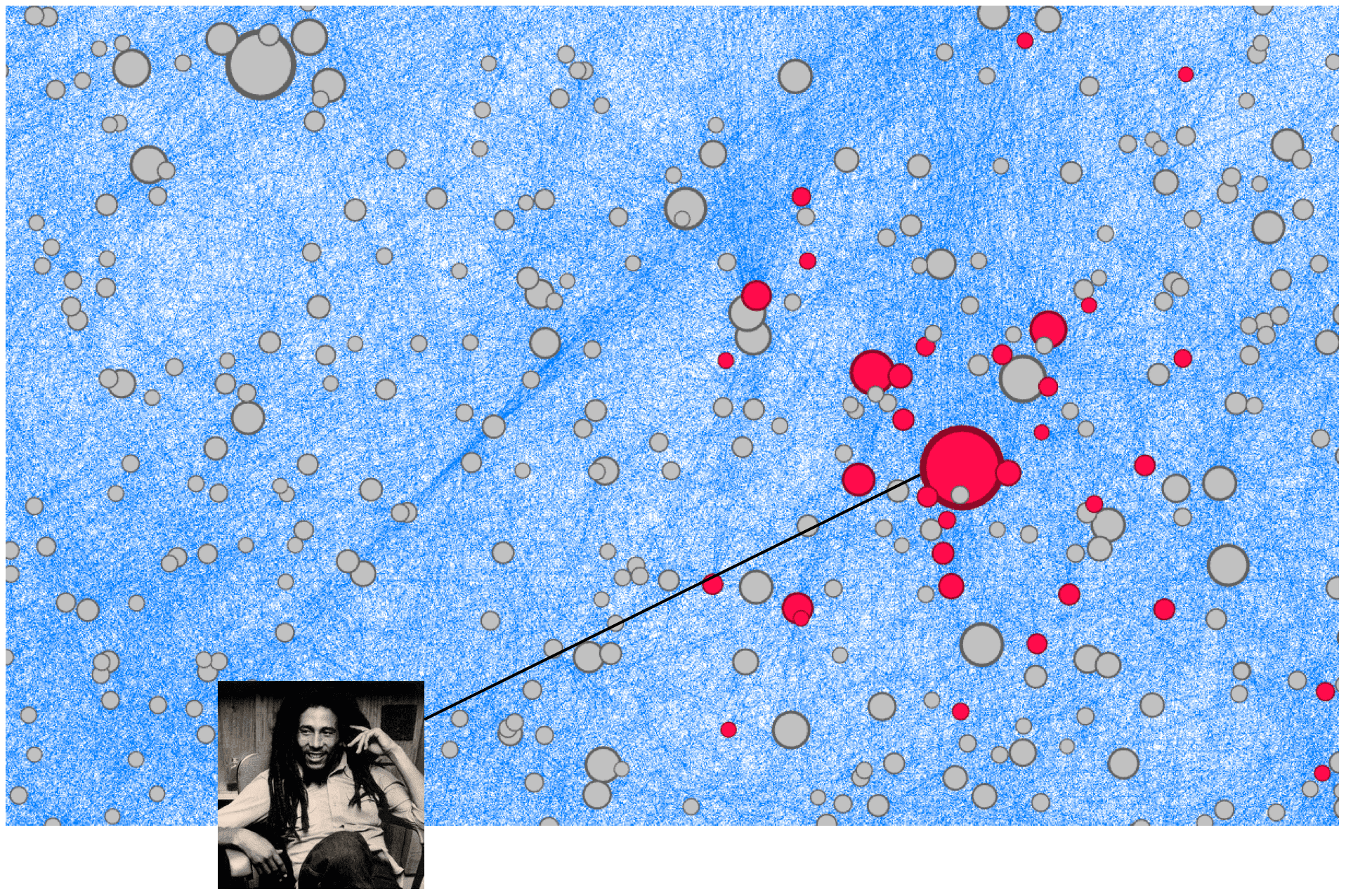}
\caption[Visualization of GAE-based embedding representations of music artists]{Visualization of embedding representations of some music artists from the Gravity-Inspired GAE model. Nodes are scaled using masses $\tilde{m}_i$, and node separation is based on distances in the embedding, using multidimensional scaling and with \cite{fruchterman1991graph}. Red~nodes correspond to \textit{Jamaican reggae} artists, appearing in the same neighborhood.}
  \label{visualization}
\end{figure}

\begin{figure*}[t]

\minipage{0.48\textwidth}
\centering
\begin{footnotesize}
\captionof{table}[Pearson and Spearman correlation coefficients of $\tilde{m}_i$ w.r.t. popularity]{Pearson and Spearman correlation coefficients of masses~$\tilde{m}_i$ from the Gravity-Inspired GAE model, w.r.t. artist-level reversed popularity ranks (i.e., the higher the more popular) on Deezer and to three node importance measures: the in-degree (i.e., the sum of edges coming into the node), the betweenness centrality \cite{salha2019-2} and the PageRank \cite{page1999pagerank}. Coefficients were computed on the~training~set.}
\label{correlations}
  \vspace{0.1cm}
\resizebox{1.0\textwidth}{!}{
\begin{tabular}{c|c|c}
    \toprule
\textbf{Node-level} & \textbf{Pearson} & \textbf{Spearman} \\ \textbf{measures} & \textbf{correlation} & \textbf{correlation}\\
\midrule
\midrule
\textbf{Popularity Rank} & $0.208$ & $0.206$ \\
\textbf{Popularity Rank by Country} & $0.290$ & $0.336$ \\
\midrule
\textbf{In-degree Centrality} & $0.201$ & $0.118$ \\
\textbf{Betweenness Centrality} & $0.109$ & $0.079$\\
\textbf{PageRank Score} & $0.272$ & $0.153$\\
    \bottomrule
\end{tabular}
}
\end{footnotesize}
\endminipage\hfill
\minipage{0.51\textwidth}
\centering
  \includegraphics[width=0.94\linewidth]{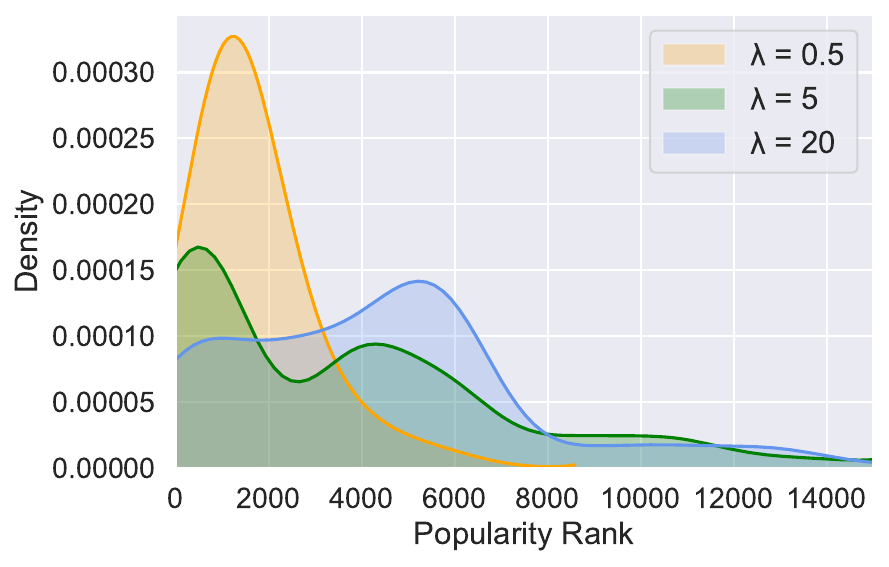}
  \vspace{-0.05cm}
  \caption[Popularity ranks of recommended artists]{Popularity rank of the \textit{most popular} artist recommended to each test artist, among their top-20. Distributions obtained from Gravity-Inspired GAE models with varying hyperparameters $\lambda$.}
  \label{impactlambda}
\endminipage
\end{figure*}

\paragraph{On the mass parameter} In Figure~\ref{visualization}, we visualize some artists and their estimated masses. At first glance, one might wonder whether nodes $i$ with the largest masses~$\tilde{m}_i$, as Bob~Marley in Figure~\ref{visualization}, simply correspond to the most popular artists on Deezer. Table~\ref{correlations} shows that, while masses are indeed positively correlated to popularity and to various graph-based node importance measures, these correlations are not perfect, which highlights that our models do not exactly learn any of these metrics. 

Furthermore, as in Chapter~\ref{chapter_5}, replacing masses $\tilde{m}_i$  by any of these measures during training (i.e., by optimizing $z_i$ vectors with the mass of $i$ being fixed, e.g., as its PageRank \cite{page1999pagerank} score) diminishes performances (e.g., more than -6 points in NDCG@200, in the case of PageRank), which confirms that jointly learning embedding vectors and masses is optimal. 

Lastly, qualitative investigations also tend to reveal that masses, and thus relative node attractions, vary across the location in the graph. Local influence correlates with popularity but is also impacted by various culture or music-related factors such as countries and genres. As an illustration, the successful samba/pagode Brazilian artist Thiaguinho, out of the top-100 most popular artists from our training set, has a larger mass than American pop star Ariana Grande, appearing among the top-5 most popular ones. While numerous Brazilian pagode artists point towards Thiaguinho, American pop music is much broader and all pop artists do not point towards Ariana Grande despite her popularity.

\paragraph{Popularity/diversity trade-off} 
The gravity-inspired decoder from Equation~\eqref{equation5} enables us to flexibly address popularity biases when ranking similar artists. More precisely:
\begin{itemize}
    \item setting $\lambda \rightarrow 0$ increases the relative importance of the \textit{influence} part of equation (\ref{equation5}). Thus, the model will highly rank the most massive nodes. As illustrated in Figure~\ref{impactlambda}, this results in recommending \textit{more popular} music artists;
    \item on the contrary, increasing $\lambda$ diminishes the relative importance of masses in predictions, in favor of the actual node \textit{proximity}. As illustrated in Figure~\ref{impactlambda}, this tends to increase the recommendation of~\textit{less~popular}~content.
\end{itemize}
Setting $\lambda = 5$ leads to optimal scores in our application (e.g., with a 41.42\% NDCG@200 for our AE, v.s. 35.91\% and 40.31\% for the same model with $\lambda=1$ and $\lambda=20$ respectively). Balancing between popularity and diversity is often desirable for industrial-level recommender systems \cite{schedl2018current}. Gravity-inspired decoders flexibly permit such a balancing.

\paragraph{Impact of attributes} So far, all models processed the complete 56-dimensional attribute vectors $x_i$, concatenating information on music genres, countries, and moods. 

\begin{figure}[H]
  \centering
  \includegraphics[width=1.0\linewidth]{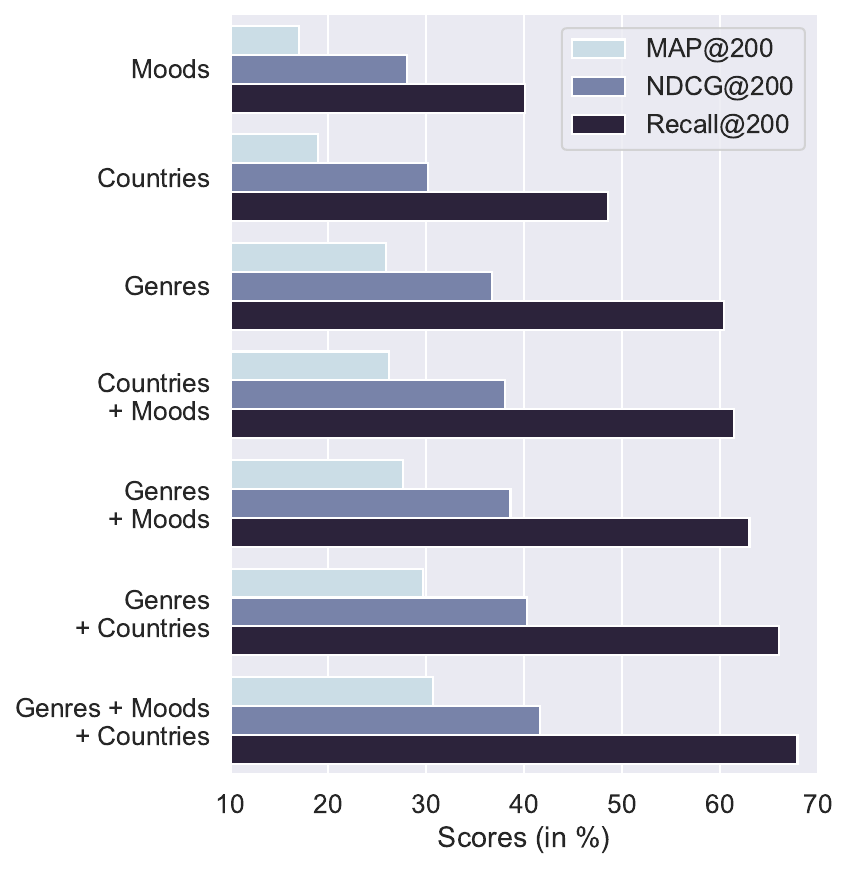}
  \caption[MAP, NDCG and Recall scores depending on input features]{Performances of Gravity-Inspired GAE, when only trained from subsets of artist-level attributes, among countries, music genres and moods.}
  \label{impactfeatures}
\end{figure}

\begin{multicols}{2}
In Figure~\ref{impactfeatures}, we assess the actual impact of each of these descriptions on performances, for our Gravity-Inspired GAE. Assuming only one attribute (genres, countries \textit{or} moods) is available during training, genres-aware models return the best performances, in terms of MAP, NDCG, and Recall scores.

Moreover, we observe that adding moods to country vectors leads to larger gains than adding moods to genre vectors. This could reflect how some of our music genres, such as \textit{speed metal}, already capture some valence or arousal characteristics. Lastly, Figure~\ref{impactfeatures} confirms that gathering all three descriptions provides the best performances, corresponding to those reported in Table~\ref{tableresults}.

\columnbreak

\end{multicols}

\paragraph{Possible Improvements (on Models)} Despite promising results in this chapter, assuming fixed similarity scores over time might sometimes be unrealistic, as some user preferences could actually evolve. Capturing such changes, e.g., through \textit{dynamic graph embeddings}, might permit providing even more refined recommendations. Also, during training we kept $k=20$ edges for each artist, while one could consider varying $k$ at the artist level, i.e., adding more (or fewer) edges, depending on the actual musical relevance of each link. One could also compare different metrics, besides mutual information, when constructing the ground truth graph. Lastly, we currently rely on a single GCN forward pass to embed cold artists which, while being fast and simple, might also be limiting. Future studies on more elaborated approaches, e.g., to incrementally update GCN weights when new nodes appear, could also improve~our~framework.

\paragraph{Possible Improvements (on Evaluation)} Our evaluation focused on the prediction of \textit{ranked lists for cold artists}. This permits filling up their \textit{"Fans Also Like/Similar Artists"} sections, which was our main goal in this chapter and of the paper associated with this work~\cite{salha2021cold}. On the other hand, future internal investigations could also aim to measure to which extent the inclusion of new nodes in the embedding space impacts the existing \textit{ranked lists for warm artists}. Such an additional evaluation, e.g., via an online test on Deezer, could assess which cold artists actually enter these lists, and whether the new recommendations 1)~are more diverse, according to some music or culture-based criteria, and 2)~improve user engagement on the service.

\section{Conclusion}
\label{c8s85}

In this chapter, we modeled the challenging cold start similar items ranking problem as a link prediction task, in a directed and attributed graph summarizing information from \textit{``Fans Also Like''} features. We presented an effective framework to address this task, transposing our recent advances on Gravity-Inspired GAE and VGAE models to recommender systems. 

Backed by in-depth experiments on artists available on the Deezer service, we emphasized the practical benefits of our approach, both in terms of recommendation accuracy, ranking quality, and flexibility. As a consequence, we plan to A/B test such models in production on the actual Deezer app in 2022, in combination with the FastGAE method from Chapter~\ref{chapter_4} to address scalability concerns.

Last, but not least, along with the paper associated with this work~\cite{salha2021cold}, we publicly released our source code, as well as the industrial data from our experiments. We hope that this release of industrial resources will benefit future research on graph-based cold start recommendation. In particular, we already identified several directions that, in future studies, should lead to the improvement of our approach.

\chapter[Music Graph Embedding and Clustering for Recommendation]{Music Graph Embedding and Clustering~for~Recommendation}\label{chapter_9}

\textit{This chapter provides a broader overview of how Deezer internally leverages similarity graphs for music recommendation. It presents and discusses practical (ongoing and future) applications of GAE and VGAE models.}

\section{Introduction}
\label{c9s91}

In Chapter~\ref{chapter_8}, we focused on quite specific top-$k$ directed graphs. In this Chapter~\ref{chapter_9}, we now give a more general overview of how Deezer leverages other pre-processed \textit{similarity graphs} to recommend musical content to users on the service. Unlike the previous chapters of this thesis, the work presented below is not associated with a research publication. It rather aims to discuss more practical industrial applications and, as a consequence, some technical details on production-facing algorithms will be voluntarily omitted for confidentiality reasons.

Firstly, in Section~\ref{c9s92}, we explain how these similarity graphs have historically played a central role in Deezer's recommender systems, to detect communities of similar artists. Then, in Section~\ref{c9s93}, we show the benefits of adopting GAEs and VGAEs, integrating our advances from previous chapters. We focus on community detection in graphs of artists and albums with recommendation purposes, through offline experiments on data extracted from Deezer. 
At the end of Section~\ref{c9s93}, we also mention several ongoing and future plans to A/B test these methods, but also to extend them to other graphs, e.g., to graphs of music tracks and of users, as well as to culture-specific graphs and to dynamic graphs. We conclude in Section~\ref{c9s94}.


\section{Music Graph Representations at Deezer}
\label{c9s92}

In this section, we describe how, at the beginning of this PhD project, similar \textit{artists} graphs were already used at Deezer to recommend music to millions of users.

\subsection{Connecting Artists based on Usage Data}
\label{c9s921}

\begin{figure}[t]
    \centering
    \includegraphics[width=\textwidth]{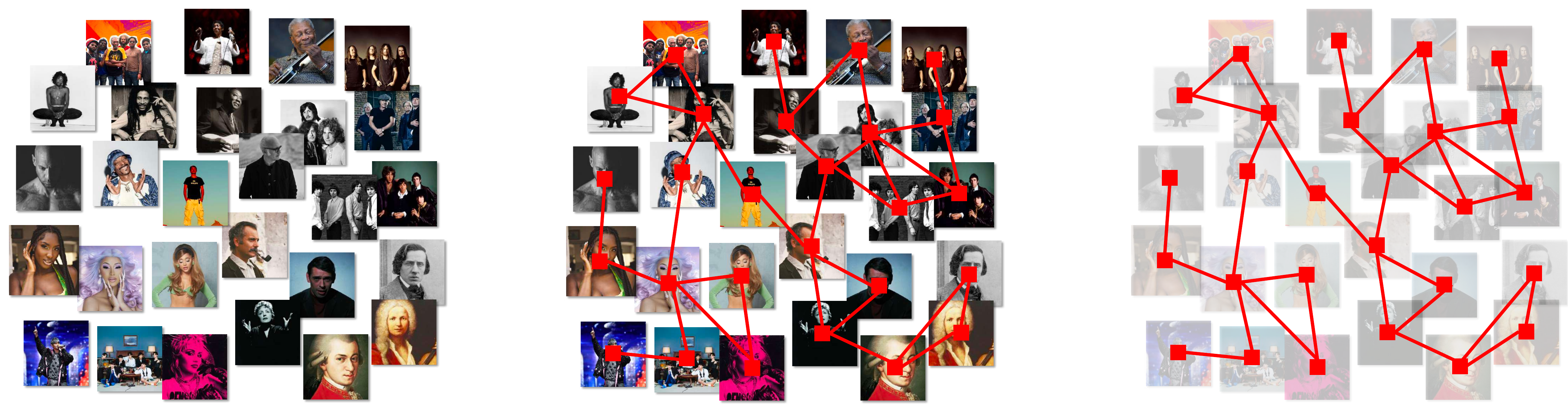}
    \caption[An example of a similar artists graph]{An example of a similar artists graph. While connections between artists are not naturally given (\textit{left}), a graph of similar artists can be constructed, e.g., by connecting artists that are simultaneously listened to or liked by numerous users (\textit{middle}). This graph can be used as an abstraction (\textit{right}), e.g., for music recommendation.}
    \label{fig:graphdeezerintroc9}
\end{figure}

Deezer historically leveraged two types of similar artists graphs computed from usage data:
\begin{itemize}
    \item the first ones correspond to the top-$k$ \textit{directed} graphs previously described in Section~\ref{c8s84} from Chapter~\ref{chapter_8}. We recall that, in such graphs, each node is an artist, pointing towards its $k$ most similar neighbors (for a pre-selected $k$, e.g., $k = 20$ in  Chapter~\ref{chapter_8}, and up to $k = 150$ in Deezer's internal databases) in terms of co-listening data. As described in this previous chapter, each directed edge $(i,j)$ has a weight corresponding to a ``similarity score'' for the artist $j$ w.r.t. the artist $i$;
    \item the second ones are refined versions of the above  top-$k$ directed graphs. The Deezer team performs several pre-processing operations, consisting in adding or removing edges between artists from these graphs. The general objective of these operations is to facilitate the use of these graphs for recommendation purposes and, in particular, to ease the extraction of homogeneous artist communities (see Section~\ref{c9s922}). Among others, several internal heuristics permit consolidating the existing links through external information, e.g., by using ``likes'' or the country of origin of each artist. Moreover, some pre-processing steps aim to limit popularity biases~\cite{schedl2018current} and to ensure that node degrees are relatively homogeneous\footnote{This objective contrasts with the ranking problem from Chapter~\ref{chapter_8} where, on the contrary, having nodes of various degrees was desirable and was (explicitly or implicitly) leveraged by several methods considered~in~this~chapter.}. The exact procedure will not be detailed in this thesis. We nonetheless emphasize that the graphs resulting from these operations are \textit{undirected}, contrary to those~from~Chapter~\ref{chapter_8} but consistently with the illustrative example reported in the above Figure~\ref{fig:graphdeezerintroc9}. 
\end{itemize}

\begin{figure}[t]
    \centering
    \includegraphics[width=1.0\textwidth]{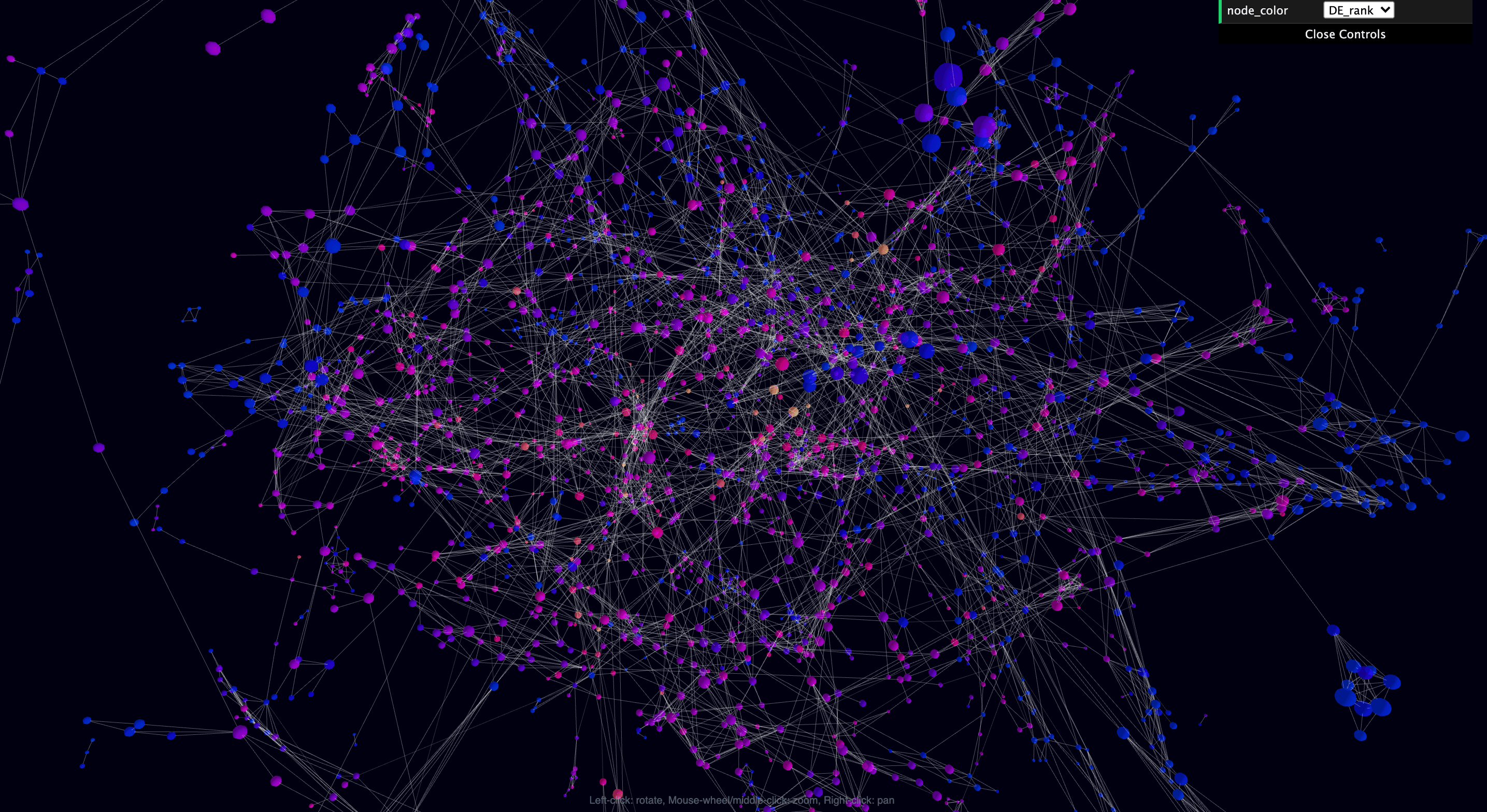}
    \caption[3D visualization of a similar artists graph (1/2)]{Visualization of a similar artists graph, computed internally at Deezer and based on the  ``3D Force-Directed Graph'' visualization tool, using the ThreeJS and WebGL JavaScript libraries and available on GitHub: \href{https://github.com/vasturiano/3d-force-graph}{https://github.com/vasturiano/3d-force-graph}. By clicking on a node, one can display information about the corresponding artist. In this figure, colors denote the popularity rank of each artist on Deezer, in Germany.}
    \label{fig:c9_deezer3}
\end{figure}

\begin{multicols}{2}

\begin{figure}[H]
    \centering
    \includegraphics[width=0.45\textwidth]{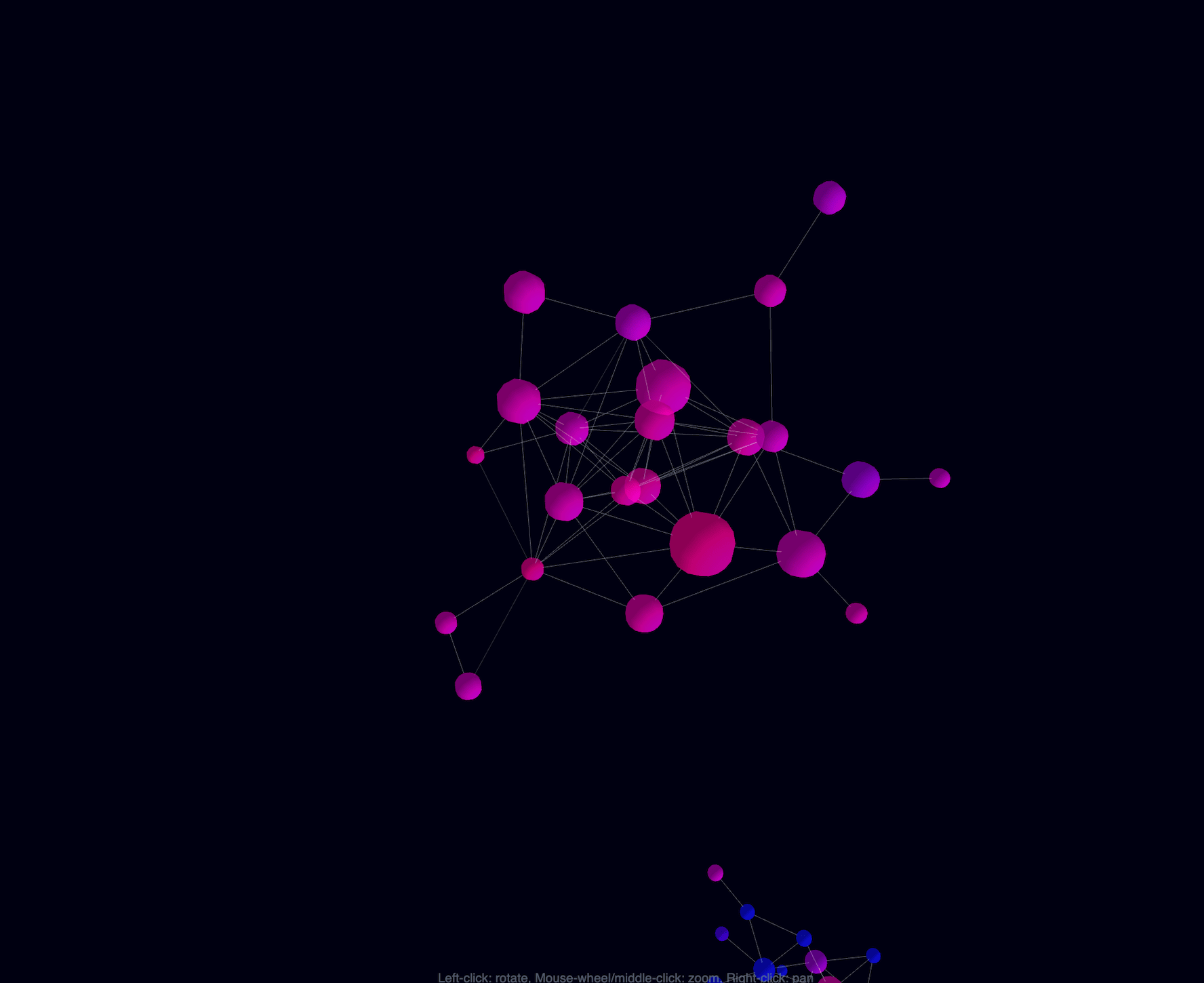}
    \caption[3D visualization of a similar artists graph (2/2)]{Zoom on an isolated group of interconnected artists, extracted from the similar artists graphs visualized in Figure~\ref{fig:c9_deezer3}. All nodes from this group correspond to Turkish pop-folk artists, such as Tarkan and Yavuz Bing{\"o}l. They are connected in the graph under consideration as they are simultaneously listened to and liked by the same users on Deezer.}
    \label{fig:c9_deezer10}
\end{figure}

These graphs are computed and updated on a weekly basis, from the usage data of millions of active users on Deezer. While the catalog includes millions of artists\footnote{While Deezer publicly advertises a catalog of 73 million tracks, the exact number of artists on the service remains a private and undisclosed information at the time of writing (i.e., in late 2021).}, at the beginning of this PhD project the Deezer team had only considered smaller versions of these graphs, with roughly 100~000 nodes corresponding to the most popular artists. Such a restriction was due to scalability constraints, but also to the desire to focus on the most reliable data for the recommendation applications described in the next sections. Figures~\ref{fig:c9_deezer3}~and~\ref{fig:c9_deezer10} provide some illustrative 3D visualizations of such similar artists graphs (specifically, of the refined undirected graphs described in the second bullet point of the previous page). 
\columnbreak

\end{multicols}

\subsection{Artist Community Detection for Music Recommendation}
\label{c9s922}

The undirected graphs described in Section~\ref{c9s921} have historically played a central role in Deezer’s recommender systems. 
At the time of writing, the Deezer team still leverages these graphs to extract clusters of similar artists using graph-based community detection methods, with the general aim of providing recommendations based on usage data. The intuition behind this strategy is the following: if some users already listened to several artists from a cluster (i.e., a community), then other unlistened artists from this same cluster could be~recommended~to~them. 

\begin{figure}[H]
    \centering
    \includegraphics[width=0.33\textwidth]{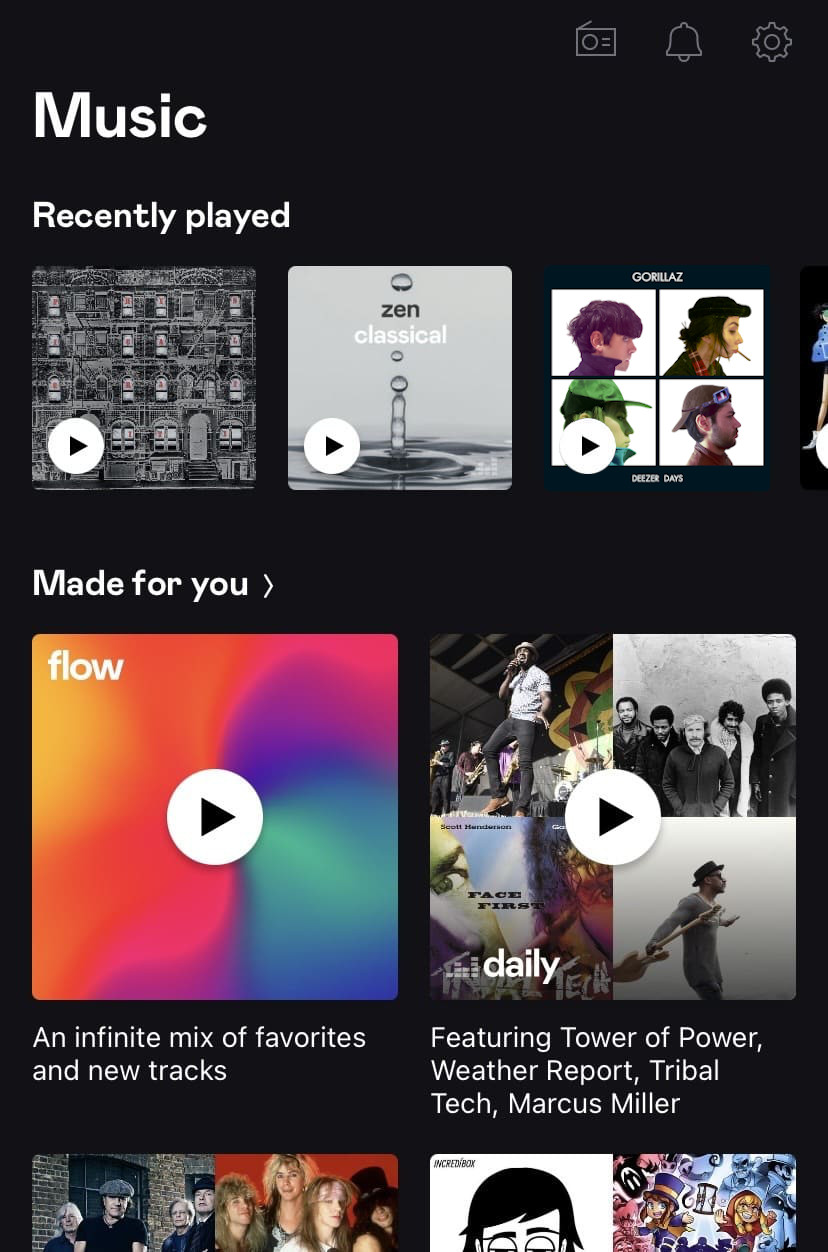}
    \caption[The Flow feature on Deezer]{The ``Flow'' feature on Deezer, along with other daily mixes of recommended music tracks.}
    \label{fig:flow}
\end{figure}

\begin{multicols}{2}
For instance, graph-based artist communities are a key component of Deezer's \textit{``Flow''} feature. As illustrated in Figure~\ref{fig:flow}, the Flow takes the appearance of a button on the homepage. A click on this button launches a virtually infinite playlist mixing the user's favorite music tracks with some recommended new tracks. Tracklists are generated by a recommendation algorithm processing the aforementioned artist communities, through a pipeline involving several operations and other data sources. While technical details on the Flow are omitted in this thesis, we emphasize that previous internal investigations at Deezer revealed the benefits of relying on graph-based methods to learn these communities,~w.r.t. \linebreak
\columnbreak

\end{multicols}
\vspace{-1.25cm}
more standard methods, e.g., some strategies consisting in clustering artists from some usage-based input features via a $k$-means algorithm\footnote{We refer to the examples of Figure~\ref{fig:twomoons_c2} from Chapter~\ref{chapter_2} for an illustration of how graph-based methods (specifically, spectral clustering, in this figure) can leverage the ``connectivity'' between data points and identify community structures in settings where a standard $k$-means would fail.}. Specifically, at Deezer, artist communities are currently computed using the Louvain algorithm~\cite{blondel2008louvain}, that we presented~in~Chapter~\ref{chapter_7}. 

\begin{figure}[t]
    \centering
    \includegraphics[width=0.85\textwidth]{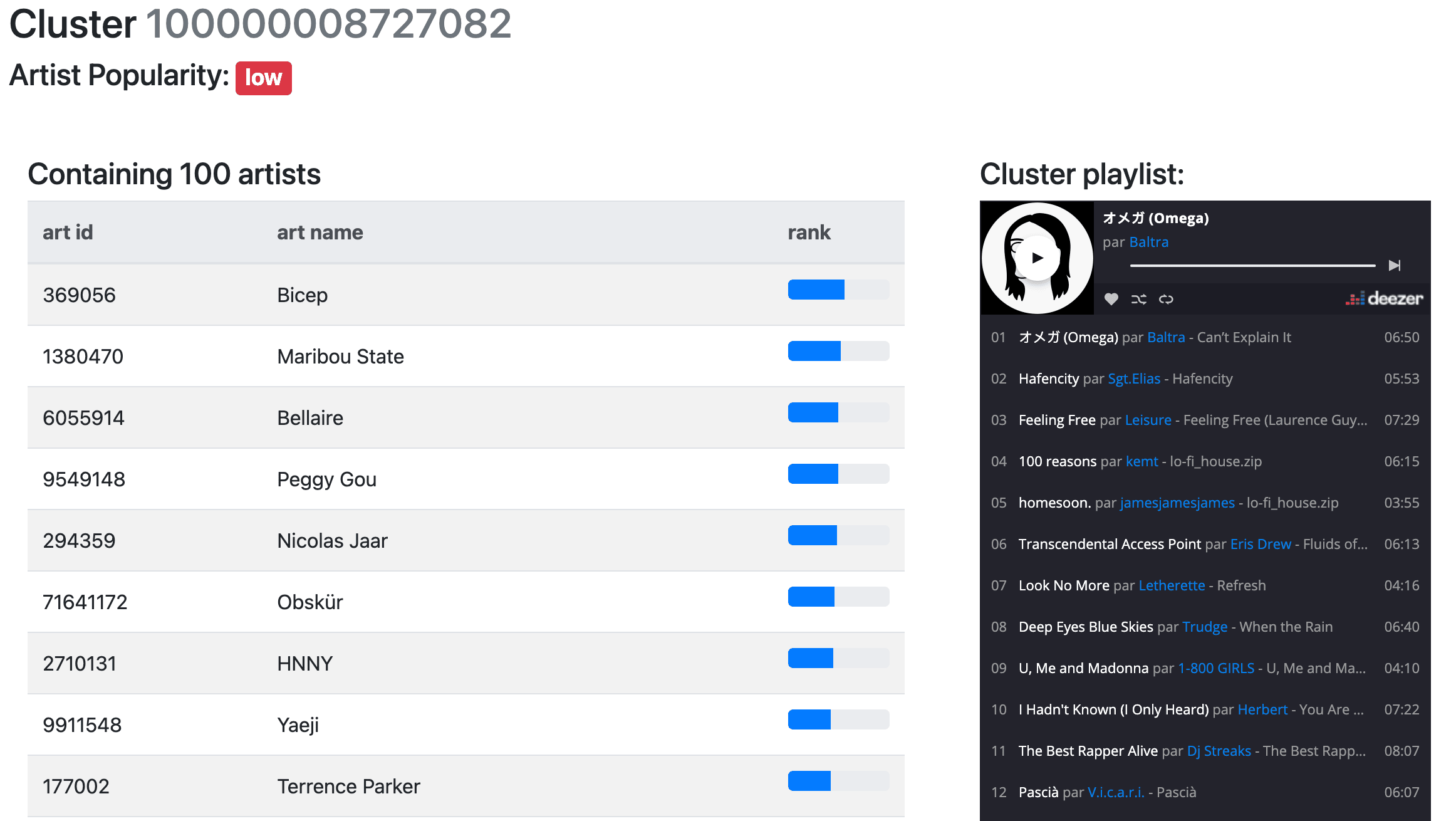}
    \caption[Generating playlists from artist communities]{An example of an artist community, computed at Deezer using the Louvain algorithm~\cite{blondel2008louvain} on a similar artists graph, and visualized through an internal tool developed by our Recommendation team. This community mainly gathers DJs and ``house music'' artists. This internal tool also generates a playlist mixing music tracks from artists of the community under consideration, using algorithms that we do not describe in this thesis.}
    \label{fig:c9_deezer6}
\end{figure}

Such a choice has several advantages. As explained in Chapter~\ref{chapter_7}, the Louvain algorithm is relatively fast and scalable, as it runs in $O(n\log n)$ time~\cite{blondel2008louvain}. Besides, its iterative modularity maximization procedure permits retrieving a hierarchical clustering of artists. This is convenient as Deezer often wants to merge or, on the contrary, to split some artist communities in post-processing steps, for recommendation purposes. Using the Louvain algorithm for artist community detection also suffers from several limitations, which will be presented and discussed in Section~\ref{c9s93}. In Figure~\ref{fig:c9_deezer6}, we provide an example of an artist community, visualized through an internal tool developed by data scientists from Deezer's Recommendation team.

\subsection{Graphs of Albums, Users, Songs}
\label{c9s923}

At the beginning of this PhD project, Deezer had mainly studied graphs of similar \textit{artists}. Nonetheless, we emphasize that the general approach from Section~\ref{c9s921}, consisting in connecting artists based on usage data (e.g., artists simultaneously listened to by numerous users), can be extended to create other graphs. For instance, one could consider different granularity levels from the musical catalog, and aim to create usage-based graphs of similar \textit{albums} or similar \textit{music tracks}. One could also study graphs of similar \textit{users}, e.g., by connecting users with the most similar listening histories on the service. As explained in Chapter~\ref{chapter_1}, Deezer users can also ``follow'' each other on the service, hence creating a large \textit{social graph}. These two graphs could be used in conjunction, e.g., to derive graph-based communities of similar users. While we did not examine such an approach so far, it constitutes a relevant direction for future work\footnote{We refer in particular to the two projects presented in Chapters~\ref{chapter_11}~and~\ref{chapter_12}. Both of them will directly rely on clusters of users, and could therefore benefit from such graph-based user communities.}.

\section{Improving Music Recommendation with GAE and VGAE}
\label{c9s93}

In this section, we study the potential benefits of leveraging the GAE and VGAE models developed in Part~\ref{partII} of this thesis for music recommendation.

\subsection{From Music Graphs to Node Embedding Spaces}
\label{c9s931}

We consider the replacement of the Louvain algorithm by a GAE/VGAE-based detection of communities in the similarity graphs described in Section~\ref{c9s92}. 
In addition to obtaining communities, running GAE and VGAE models instead of the Louvain algorithm would permit learning low-dimensional node embedding representations of the musical entities under consideration (e.g., of artists or albums). These representations could act as feature vectors for these entities in other machine learning problems, and be helpful for data visualization.

More importantly, the standard Louvain algorithm only relies on the graph structure to extract communities, which is limiting as our similarity graphs are often enriched by node-level descriptions. For instance, in the graph of Chapter~\ref{chapter_8}, artists were also described by their own vectors providing information on their music genres, moods, and countries of origin. 

On the other hand, as explained in previous chapters, GAE and VGAE models can process such node \textit{attributes} a.k.a. node \textit{features}.  Chapter~\ref{chapter_8} already emphasized how GAE and VGAE models can effectively learn artist embeddings that simultaneously capture usage (through edges in the graph) and musical (through node features) information on artists. In this section, we now aim to analyze to which extent this aspect could also improve the detection of communities of artists, or of other musical entities, and enhance Deezer's recommender systems.

\subsection{Experimental Setting}
\label{c9s932}

In this direction, we now present the experimental setting of some offline community detection experiments, conducted on data extracted from Deezer's production system.

\paragraph{Task and Datasets} 
As is the case for the previous community detection experiments from this thesis, our task will consist in extracting a partition of nodes from a graph into $K$ communities, for a given value of $K$. We consider~two~\textit{undirected}~graphs:
\begin{itemize}
    \item a similar \textit{artists} graph, including $n = $ 77~656 artists from the Deezer catalog, connected via $m = $ 377~591 edges. This graph was computed in September 2021 from usage data of millions of active users, and using the pre-processing steps from Section~\ref{c9s921};
    \item a larger similar \textit{albums} graph, including $n =$ 2~503~985 albums from the Deezer catalog, connected via $m =$ 25~039~155 edges. This graph was also computed in September 2021. It corresponds to the ``Albums'' graphs used in the experiments of Chapter~\ref{chapter_7}.
\end{itemize}

Artists and albums are also described by 128-dimensional vectors $x_i$, which will act as node features in GAE/VGAE models. We obtained these vectors from the factorization of a \textit{pointwise mutual information} matrix among music tracks, using singular value decomposition (SVD). The matrix is constructed from the co-occurrences of music tracks in diverse musical collections on Deezer, e.g., music playlists, listening sessions, and lists~of~favorite~tracks. The SVD returns a 128-dimensional vector for each track; we subsequently derive the 128-dimensional vector of each artist/album by averaging the vectors from their corresponding music tracks.

\paragraph{Evaluation Metrics} We do not have access to a list of artist/album communities that should actually be recommended together to users. As a consequence, in this chapter, we assume the unavailability of ground truth node labels for evaluation. This prevents computing AMI and ARI scores, as we did in Part~\ref{partII} of this thesis. In these offline experiments, we instead report three alternative metrics, to evaluate the models described in~the~next~paragraph:
\begin{itemize}
    \item first and foremost, we report the \textit{modularity} score $Q$ associated with each node partition into $K$ communities. The modularity, defined in Definition~\ref{def:modularity} from Chapter~\ref{chapter_7}, acts as an unsupervised graph-based measure of the quality of a node partition, in terms of intra- and inter-community edge density;
    \item then, we evaluate the \textit{musical homogeneity} of each community, in terms of music genres. As in Section~\ref{c8s841} from Chapter~\ref{chapter_8}, nodes are associated with 32-dimensional music genre embedding vectors, computed internally at Deezer. For each community detection method, we compute $H$, the average intra-community standard deviation of these vectors, as a measure of intra-community musical homogeneity;
    \item lastly, we consider the artists and albums liked by a set of 2~000 Deezer users in October 2021, i.e., the month after computing the two similarity graphs of this study. We denote by $\mathcal{F}_u$ the set of artists (or, of albums) liked by the user~$u$ during the period. Furthermore, we denote by $N_{uk} = \sum_{i \in \mathcal{F}_u} \mathds{1}_{(i \in C_k)}$ the number of artists (or albums) from $\mathcal{F}_u$ belonging to the community $C_k$, with $k \in \{1,\dots,K\}$. We compute:
    \begin{equation}
        P_u = \frac{1}{K} \sum_{k = 1}^K  \mathds{1}_{N_{uk} > 0},
    \end{equation}
    i.e., the percentage of communities including at least one favorite artist (or album) of $u$. A low value of $P_u$ indicates that a few communities summarize the musical preferences of $u$. This is desirable in the context of our recommender systems (see Section~\ref{c9s92}), as unlistened artists/albums from the same few communities would more likely constitute relevant recommendations for this user. Specifically, in our experiments, we consider $P = \frac{1}{\text{2000}} \sum_{u} P_u$, the average value of $P_u$ over the set of users.
    \end{itemize}

\paragraph{Models} Using these evaluation metrics, we compare the communities obtained from the Louvain method~\cite{blondel2008louvain} to those obtained via several~GAE~and~VGAE~models:
\begin{itemize}
    \item the \textit{standard GAE and VGAE} models from Kipf and Welling~\cite{kipf2016-2} with two-layer GCN encoders and inner product decoders;
    \item their simplified counterparts, \textit{Linear GAE and VGAE}, introduced in Chapter~\ref{chapter_6}; 
    \item our improved models for community detection, \textit{Modularity-Aware GAE and VGAE}, with consistent decoders/losses w.r.t. Chapter~\ref{chapter_7}. We consider two variants of these models, with either 1) the linear encoders from Chapter~\ref{chapter_6}, or 2) 2-layer GCN encoders, incorporating our revised message passing operator $A + \lambda A_s$ on the first layer only, as in Chapter~\ref{chapter_7}.
\end{itemize}

All hyperparameters were optimized according to the model selection procedure proposed in Section~\ref{c7s733} from Chapter~\ref{chapter_7}. We set $d = 32$ for all models. Multi-layer GCN encoders include a 64-dimensional hidden layer. We trained models for 500 iterations (resp., 600 iterations), with a learning rate of 0.01 (resp., of 0.005) for the similar artists graph (resp., for the similar albums graph), without dropout and using the Adam optimizer~\cite{kingma2014adam}. For all Modularity-Aware GAE and VGAE models and for both graphs, we used the same values of $\lambda$, $\beta$, $\gamma$ and $s$ as those reported in Table~\ref{tab:hyperparameterstable} from Chapter~\ref{chapter_7}. To overcome scalability issues, we used the FastGAE method from Chapter~\ref{chapter_4}, and trained all models by decoding stochastic subgraphs of size $n_{(S)} =$ 10~000, with degree-based sampling and with $\alpha = 1$ for both graphs. Our experiments therefore incorporate and combine several of the technical contributions presented throughout this thesis.

\subsection{Offline Evaluation of GAE and VGAE Models}
\label{c9s933}

Table~\ref{c9offlineevaluation} reports our results on the two graphs. For the sake of readability, we report \textit{relative} scores w.r.t. the Louvain method, currently used in our production systems. We recall that, in these experiments, we aim to \textit{maximize} $Q$ but to \textit{minimize} $H$ and $P$.

\paragraph{Results} The Louvain method produces the communities associated with the highest modularity ($Q$) values. We underline that $Q$ is the criterion explicitly optimized by the Louvain greedy algorithm~\cite{blondel2008louvain}. While the GAE and VGAE models underperform according to this criterion, most of them reach better $H$ and $P$ scores. Better $H$ scores indicate that the GAE/VGAE-based communities are more homogeneous in terms of music genres. Better $P$ scores indicate that user preferences tend to be summarized by fewer GAE/VGAE-based communities. 

We now compare the different GAE and VGAE models. Firstly, we observe that our Modularity-Aware GAEs and VGAEs, specifically designed for improved community detection, provide the best $H$ and $P$ scores. They also obtain better $Q$ scores than standard GAEs and VGAEs. Such a result is not surprising, as these models incorporate the Louvain method as a pre-processing step (see Section~\ref{c7s732}) as well as a modularity-inspired regularization term (see Section~\ref{c7s733}).  Moreover, as in Chapter~\ref{chapter_6}, the results obtained using linear encoders are either on par or relatively close to those obtained using 2-layer GCN encoders. Lastly, on both graphs, VGAE models tend to slightly outperform their GAE counterparts, although most results are still close as in our previous experiments from this thesis. 

In our experiments, all GAE and VGAE models processed the aforementioned SVD-based node features. In future tests, these features could be completed with some additional information on artists and albums. For instance, one could incorporate information on music genres, as we did in Chapter~\ref{chapter_8} (in this Chapter~\ref{chapter_9}, music genres were alternatively used for \textit{evaluation}). We would expect the addition of music genres into node features to improve the homogeneity scores~$H$. Lastly, while our experiments set $K = 100$ or $1000$ in Table~\ref{c9offlineevaluation}, we note that we obtained consistent conclusions with other values, in our preliminary tests. At the time of writing, Deezer selects the optimal number and sizes of communities through heuristics undisclosed~in~this~thesis.

\begin{table}[t]

\centering
\caption[Learning communities of similar albums and artists]{Community detection on similar Deezer artists/albums graphs, using the Louvain method and several GAE and VGAE models. All GAE and VGAE models learn embedding vectors of dimension $d = 32$, with other hyperparameters described in Section~\ref{c9s932}. All models learn $K = 100$ (resp., $K = 1000$) communities of artists (resp., of albums). Scores are averaged over 10 runs. We report relative scores (in \%) w.r.t. the Louvain baseline. We recall that we aim to \textit{maximize} $Q$ but to \textit{minimize} $H$ and $P$. \textbf{Bold} numbers correspond to the best scores.}
\resizebox{1.0\textwidth}{!}{
\begin{tabular}{c|ccc|ccc}
\toprule
\textbf{Model} & \multicolumn{3}{c}{\textbf{Artists}} & \multicolumn{3}{c}{\textbf{Albums}} \\
 &  \footnotesize \textbf{Modularity $Q$ } & \footnotesize \textbf{Homogeneity $H$} & \footnotesize \textbf{Coverage $P$} & \footnotesize \textbf{Modularity $Q$ } & \footnotesize \textbf{Homogeneity $H$} & \footnotesize \textbf{Coverage $P$} \\
\midrule
\midrule 
Louvain & \textbf{100} & 100 & 100 & \textbf{100} & 100 & 100 \\ 
\midrule
Linear Standard GAE & 88.2 & 96.9 & 97.6 & 78.3 & 101.4 & 95.1 \\
Linear Standard VGAE & 89.4 & 96.8 & 97.4 & 80.1 & 98.5 & 94.8 \\
GCN-based Standard GAE & 89.4 & 96.6 & 97.6 & 79.7 & 100.9 & 95.4 \\
GCN-based Standard VGAE &  90.9 & 96.0 & 97.5 & 80.6 & 98.2 & 94.7 \\
\midrule
Linear Mod.-Aware GAE & 93.6 & 96.0 & 97.1 & 92.4 & 98.9 & 94.3 \\
Linear Mod.-Aware VGAE & 94.2 & 96.0 & \textbf{96.9} & 92.7 & 97.8 & 93.7 \\
GCN-based Mod.-Aware GAE & 93.9 & 95.3 & 97.2 & 92.6 & 98.7 & 94.1 \\
GCN-based Mod.-Aware VGAE & 94.6 & \textbf{94.4} & 97.0 & 93.1 & \textbf{97.5} & \textbf{93.6} \\
\bottomrule
\end{tabular}
\label{c9offlineevaluation}
}
\end{table}

\paragraph{Discussion} Overall, these results are promising. The best communities for music recommendation (see Section~\ref{c9s92}) are not always the densest ones, e.g., the ones associated with the highest modularity. At the price of a lower modularity, GAEs and VGAEs from Table~\ref{c9offlineevaluation} manage to learn artist/album communities with a slightly higher musical homogeneity. More importantly, their lower $P$ scores suggest that GAE/VGAE-based communities are more in phase with our recommendation strategy, i.e.,  that if users already listened to several artists from a community, then other unlistened artists from the same community could~be~recommended~to~them.

Undoubtedly, these results remain very preliminary. To verify whether GAEs and VGAEs indeed learn ``better'' communities for recommendation, the Deezer team might try to incorporate them into production-facing recommender systems such as the Flow. For instance, an online A/B test could measure whether the GAE/VGAE-based communities improve some key performance indicators (e.g., whether Deezer users listen to the recommended tracks longer, or, skip the recommended tracks less often) w.r.t. Louvain-based communities. At the time of writing, such investigations were still undone. They open the way for some future work and online tests, that might lead to a model deployment on the Deezer service in 2022.

\paragraph{Extensions} In addition to these online A/B tests on Deezer, we also plan to extend these analyses to graphs of \textit{users}, previously mentioned in Section~\ref{c9s931}. As the FastGAE method permits scaling GAE and VGAE models to large graphs with millions of nodes and edges, future work could also aim to train models on graphs with a larger number of artists or of albums, as well as on large similar \textit{music tracks} graphs with up to 73 million nodes, and on more general \textit{knowledge graphs} of musical entities. In such graphs, as detailed in Chapter~\ref{chapter_1}, music tracks would be connected to artists, albums, music genres, or record labels, that would themselves be connected together through various semantic links.

Another research direction could consist in learning \textit{country-specific} or \textit{culture-specific} music graph representations. As we will develop in Chapter~\ref{chapter_10}, people from different countries and/or different cultures perceive music differently~\cite{Sordo2008}. This can result in different listening behaviors on music streaming services. For instance, two similar artists graphs, constructed only from the usage data of French users and Japanese users, respectively, might be structurally different. Taking into account these differences by considering country/culture-specific graphs might permit providing more refined recommendations. In a similar fashion, future work could also consider \textit{context-specific} similarity graphs. As music
consumption highly depends on the listening context~\cite{tran2021hierarchical}, some node pairs (e.g., some artist pairs) might only be connected (i.e., co-listened) in some contexts (e.g., based on the activity, mood, or time of the day).

\subsection{Towards Dynamic Music Graph Embedding and Clustering}
\label{c9s934}

In this section, we continue to discuss extensions of our methods. We provide more details on an ongoing project, related to GAE/VGAE-based dynamic node embedding and clustering.

\paragraph{From Static to Dynamic Graphs} In the experiments of Section~\ref{c9s933}, the graph structures under consideration were \textit{fixed}. In reality, as previously mentioned in Chapter~\ref{chapter_8}, new artists (as well as new albums, music tracks, etc) regularly appear on the Deezer service. As a consequence, new nodes should be frequently incorporated into the corresponding similarity graphs. On the contrary, some artists (or albums, music tracks, etc) are sometimes removed from the catalog. In addition, users' listening habits on Deezer might evolve over time, resulting in new or removed edges between these nodes. Therefore, these similarity graphs are \textit{dynamic} by nature.

Our (static) definition of a graph from Definition~\ref{defgraph} can be straightforwardly extended to represent dynamic structures. In the scientific literature, \textit{dynamic graphs} are often summarized by $T$ discrete snapshots of standard graphs, depicting the structural evolution over time~\cite{kazemi2020representation}:
\begin{equation}
    \mathcal{G}_1 = (\mathcal{V}_1, \mathcal{E}_1),~\mathcal{G}_2 = (\mathcal{V}_2, \mathcal{E}_2),~\dots,~\mathcal{G}_T = (\mathcal{V}_T, \mathcal{E}_T),
\end{equation}
They are associated with $T$ adjacency matrices $A_1$, $A_2$, $\dots$, $A_T$. As an illustration, Deezer computes similarity graphs on a weekly basis (see Section~\ref{c9s92}). Hence, each similar artist graph $\mathcal{G}_t$ corresponds to a snapshot of the Deezer catalog and users' listening data during~the~$t$-th~week.

\paragraph{Dynamic Music Graph Clustering} Currently, our production system computes artist communities every week \textit{from scratch}, by running the Louvain algorithm on every new similarity graphs. In a similar fashion, one could consider training a completely new GAE/VGAE model every week. While being simple, this approach suffers from~two~limitations:
\begin{itemize}
    \item firstly, this approach is computationally \textit{inefficient}. In our applications, $\mathcal{G}_{t}$ will often be relatively similar to $\mathcal{G}_{t-1}$. For instance, the new nodes (e.g., the new artists) appearing during a week usually represent much less than 1\% of the catalog. Intuitively, the information learned on $\mathcal{G}_{t-1}$ should still be relevant to extract communities from $\mathcal{G}_{t}$. Training an entire model from scratch every week might not be necessary;
    \item secondly, this approach is not \textit{stable} over time. By training a new model on $\mathcal{G}_{t}$, there is no theoretical guarantee that the resulting communities will be ``close'' to those obtained at week $t-1$. Yet, for consistency reasons, ensuring relatively stable recommendations over time is usually desirable for music streaming services such as Deezer.
\end{itemize}


\paragraph{Extending GAE and VGAE Models to Dynamic Graphs}
In 2020, the research internship of Raphaël Ginoulhac~\cite{ginoulhac2020stage}, that we supervised at Deezer with Benjamin Chapus, aimed to address these limitations. During his internship at Deezer, Raphaël reviewed the recent advances on {dynamic node embedding and clustering} (we refer the interested reader to the recent survey of Kazemi et al.~\cite{kazemi2020representation}). He subsequently investigated several strategies to extend GAEs and VGAEs to dynamic graphs. Through an experimental analysis on similar artists graphs, he revealed the empirical effectiveness of two simple methods:
\begin{itemize}
    \item \textbf{HotStart GAE/VGAE}: this method is inspired by the work of Goyal et~al.~\cite{goyal2018dyngem}. It consists in 1) training a GAE/VGAE model on an initial graph $\mathcal{G}_0$, then 2) using the optimized weights of such a model as the \textit{initial weights} of the GAE/VGAE model associated with a new graph $\mathcal{G}_1$, and 3) fine-tuning this new model for a few training iterations only\footnote{In Chapter~\ref{chapter_8}, we trained our models on a graph of ``warm'' artists, and used the same model to learn embeddings for new ``cold'' nodes from this graph. This approach actually constitutes a particular case of HotStart GAE/VGAE, where the model is \textit{not} fine-tuned after a modification of the graph structure.}. The procedure is repeated over time, i.e., the final weights of the $(t-1)$-th model correspond to the initial weights of the $t$-th model, for any $t > 1$;
    \item \textbf{KL-Reg VGAE}: this alternative method for VGAE models consists in modifying the ELBO objective from Equation~\eqref{elbo}, so that the Kullback-Leibler divergence~\cite{kullback1951information} term forces the posterior distribution at time $t$, say $q^{(t)}$, to be close to the one computed at time $t-1$, say $q^{(t-1)}$, instead of the unit Gaussian prior $\mathcal{N}(0,I_d)$. Specifically, we firstly train a VGAE model, with a standard ELBO objective, on an initial graph $\mathcal{G}_0$. Then, when training the $t$-th VGAE on $\mathcal{G}_t$ with $t > 0$, we maximize, by gradient ascent:
    \begin{equation}
    \mathcal{L}^{(t)}_{\text{VGAE}} = \mathbb{E}_{q^{(t)}(Z | A,X)} \Big[\log
p^{(t)}(A | Z,X)\Big] - \mathcal{D}_{KL}\Big(q^{(t)}(Z | A,X)||q^{(t-1)}(Z | A,X)\Big).
    \end{equation}
\end{itemize}
Raphaël's experiments tend to confirm that both methods are computationally efficient, i.e., that only a few training iterations are required for the $t$-th model to converge when $\mathcal{G}_t$ is relatively similar to $\mathcal{G}_{t-1}$. Simultaneously, and in a comparable way, both methods are more stable than the standard approach consisting in training new models from scratch every week. Specifically, in a majority of experiments, the AMI scores of communities extracted from  $\mathcal{G}_t$ w.r.t. those extracted on $\mathcal{G}_{t-1}$ increase when using the HotStart GAE/VGAE or KL-Reg VGAE methods. 

Nonetheless, such results still require further theoretical analyses, as well as an empirical validation on external datasets and a proper comparison to existing baselines~\cite{goyal2018dyngem,kazemi2020representation}. While we chose not to fully report these results in their preliminary state in this thesis, they will motivate more investigations in the upcoming months, potentially leading to a scientific publication and/or to a model deployment on the Deezer service. 

\section{Conclusion}
\label{c9s94}

In this chapter, we provided a broader overview of how Deezer has historically leveraged similarity graphs for community-based music recommendation. Then, through offline experiments on data extracted from Deezer's production system, we showed the benefits of adopting GAE and VGAE models, integrating our technical contributions from Part~\ref{partII}. 

While our conclusions are promising, we acknowledge that the results presented in this chapter are the most preliminary of this thesis. They still require further empirical validation, including through online A/B tests on the service. With the Deezer team, we will consider launching such tests in the upcoming months, i.e., after the end of this PhD project.

Lastly, in this chapter, we mentioned several ongoing and future plans to extend our methods and our analyses to other graphs. This includes our graphs of users, which have been quite neglected so far, as well as more general knowledge graphs, culture-specific graphs, and context-specific graphs. As the structures under consideration in our work are dynamic by nature, we also aim to pursue our efforts towards more efficient dynamic node embedding methods.

\chapter[Modeling the Music Genre Perception across Language-Bound Cultures]{Modeling the Music Genre Perception across  Language-Bound Cultures}\label{chapter_10}
\chaptermark{Modeling the Music Genre Perception across Cultures}

\textit{This chapter presents research conducted with Elena V. Epure, Manuel Moussallam, and Romain Hennequin, and published in the proceedings of the 2020 Conference on Empirical Methods in Natural Language Processing (EMNLP 2020)~\cite{epure2020modeling}. This chapter also mentions the Muzeeglot prototype~\cite{epure2020muzeeglot}, internally developed with Felix Voituret and based on this research.}

\section{Introduction}
\label{c10s1}

The first two chapters of this Part~\ref{partIII}, i.e., Chapters~\ref{chapter_8}~and~\ref{chapter_9}, presented several applications of GAE and VGAE models to similarity graphs. In this Chapter~\ref{chapter_10}, we now focus on quite different graph structures, that correspond to music genre \textit{ontologies}. As explained in Chapter~\ref{chapter_1} and further detailed in the following sections, these ontologies are graphs of conceptually related music genres, connected through various relation-specific edges \cite{schreiber2016genre}. In this chapter, we propose to leverage these representations to model the music genre perception across~cultures. 

Specifically, we consider the following problem: the music genre perception expressed through human annotations of artists or albums varies significantly across (language-bound) cultures~\cite{Ferwerda2016InvestigatingTR}. As we will show, these variations cannot be modeled as mere translations since we also need to account for cultural differences in the music genre perception~\cite{epure2020modeling}. This is an important issue for music streaming services such as Deezer, as these variations impact a wide range of music information retrieval and recommendation tasks, ranging from language-aware music genre auto-tagging to localized playlist captioning and music genre-driven recommendations~\cite{epure2020multilingual}. To address this problem, we would need to 
tag music items, such as artists, albums, or tracks, with music genre representations that capture such differences in perception,~which~is~challenging~\cite{epure2020modeling,epure2021taln}.

Previous research works at Deezer already studied the relations between music genres and the way music items are associated with music genres. Hennequin~et~al.~\cite{hennequinaudio} managed to learn music genre embedding representations from audio data, with application to audio-based disambiguation of music genre annotations (i.e., identifying situations where different tags actually refer to the same genre, or, on the contrary, where the same genre is identified by different tags). Epure et al.~\cite{Epure2019} proposed to use knowledge bases and parallel annotations for music genre ``translation'' across several inconsistent genre tag systems, but only~considered~English~genres. 

In this work, published in a more NLP-focused venue (EMNLP 2020), we aim to complement these studies, by  \textit{analyzing the feasibility of obtaining relevant cross-lingual, culture-specific music genre annotations based only on language-specific semantic representations}, namely on 1)~\textit{graph ontologies}, and 2) NLP-based \textit{distributed word embeddings} of music genres, acting as attribute vectors of genres/nodes in these graphs. Our study, focused on six languages and on genre annotations obtained from the Wikipedia online encyclopedia, shows that unsupervised cross-lingual music genre annotation is feasible with high accuracy when combining both types of representations, using either GAE/VGAE-inspired models or~\textit{retrofitting}~\cite{faruqui2015}. Besides, along with the paper associated with this work~\cite{epure2020modeling}, we publicly released our source code, as well as the dataset (scraped and processed from the internet) that we used for our experiments.

This chapter is organized as follows. In Section~\ref{c10s2}, we introduce the problem of subjectivity in the music genre perception in detail. We also formally describe the cross-lingual music genre annotation problem we aim to solve in this work. To address this problem, we subsequently present, in Section~\ref{c10s3}, our methods to learn music genre representations by combining graph ontologies and word embeddings. We report and discuss our experimental setting and our results in Section~\ref{c10s4}. At the end of this same section, we also mention the Muzeeglot prototype~\cite{epure2020muzeeglot}, internally developed at Deezer to visualize our multilingual music genre embedding representations. We conclude in Section~\ref{c10s5}. In Section~\ref{c10s6}, we provide an additional proof, placed out of the ``main'' chapter for the sake of brevity and readability.

\section{The Cross-Lingual Music Genre Annotation Problem}
\label{c10s2}

In this section, we detail the research problem we aim to address in this chapter.

\subsection{Modeling the Music Genre Perception}
\label{c10s21}

A prevalent approach to culturally study music genres starts with a common set of music items, e.g., artists, albums, or music tracks, and assumes that the same music genres would be associated with the items in all cultures \cite{Ferwerda2016InvestigatingTR,skowron2017predicting}.
However, music genres are subjective.
Cultures themselves and individual musicological backgrounds influence the music genre perception, which can differ among individuals \cite{Lee2013KPopGA,Sordo2008}.  
For instance, a Westerner may relate \emph{funk} to \emph{soul} and \emph{jazz}, while a Brazilian to \emph{baile funk}, which is a type of \textit{rap} \cite{hennequinaudio}.
Thus, accounting for cultural differences in the music genre perception could give a more grounded basis for such cultural studies. It could also help music streaming services associate their musical catalog with more relevant music genres tags, and, among other applications, provide better genre-driven recommendations to users (see, e.g., our experiments from Chapters~\ref{chapter_8}~and~\ref{chapter_9}, that both involved music genre-related data). However, ensuring both a common set of music items and culture-sensitive annotations with a broad coverage of music genres is strenuous~\cite{bogdanov2019acousticbrainz}.

To address this challenge, we study the feasibility of cross-culturally annotating music items with music genres, without relying on a parallel corpus.
In this work, culture is related to \textit{a community speaking the same language}\footnote{This assumption conveniently frames the problem. Nonetheless, we acknowledge that it is quite strong for some languages, e.g., French and Spanish, if the origin of music genres annotators is not specified.} \cite{kramsch1998language}.
The specific research question we build upon is: 
\textit{assuming consistent patterns of music genres association with music items within cultures, can a mapping between these patterns be learned by relying on language-specific semantic representations?}
It is worth noting that, since music genres fall within the class of Culture-Specific Items \cite{Aixela1996,Newmark1988}, cross-lingual annotation, in this case, cannot be framed as standard translation, as one also needs to model the dissimilar perception of music~genres~across~cultures.

Our work focuses on six languages from four language families: Germanic (English and Dutch), Romance (Spanish and French), Japonic (Japanese), and Slavic (Czech); and on two types of language-specific semantic representations: music genre ontologies and word embeddings. In the following Section~\ref{c10s22}, we formalize the cross-lingual annotation task. Then, in Section~\ref{c10s23}, we describe the Wikipedia-based test corpus we used in this work, to evaluate the music genre representations that we will subsequently present in Section~\ref{c10s3}.

\subsection{Problem Formulation}
\label{c10s22}

Our cross-lingual music genre annotation task consists of inferring, for music items, tags in a target language $L_t$, knowing tags in a source language $L_s$ according to some source (e.g., Wikipedia, in this study).
For instance, knowing the English music genres of \emph{Fatboy Slim} (\emph{big beat}, \emph{electronica}, \emph{alternative rock}), the goal is to predict \emph{rave} and \emph{rock alternativo} in Spanish.
As shown in this example and in the one from Section~\ref{c10s21}, the problem goes beyond translation and instead targets a model able to map concepts, potentially dissimilar, across~languages~and~cultures.

Formally, given $S$ a set of tags in a language $L_s$, $\mathcal{P}$ the partitions of $S$ and $T$ a set of tags in a language $L_t$, a mapping scoring function $f: \mathcal{P}(S) \rightarrow \mathbb{R}^{|T|}$ can attribute a prediction score to each target tag, relying on subsets of source tags drawn from $S$ \cite{Epure2019,epure2020modeling,hennequinaudio}.
The produced score incorporates the degree of relatedness of each particular input source tag to the target tag.
A common approach to compute relatedness in distributional semantics relies on cosine similarity.
Thus, for some source tags $\{ s_{1}, ..., s_K \}$  and a target tag $t$, $f$ can be defined as:
\begin{equation}
f_t(\{s_{1}, s_{2}, \dots, s_K\}) = \sum_{k=1}^{K}\frac{{z_{s_{k}}}^T z_t}{||z_{s_{k}}||_2 || z_t||_2},
\label{eq:transfinit}
\end{equation}
where $z_{s_{k}}$ and $ z_t$ are some $d$-dimensional embedding vectors representing each $s_{k}$ and $t$, respectively (see Section~\ref{c10s3}), and where $|| \cdot ||_2$ denotes the Euclidean norm. 


\subsection{Evaluation Corpus}
\label{c10s23}

\begin{CJK}{UTF8}{min}

\begin{figure}[t]
    \centering
    \includegraphics[width=0.8\textwidth]{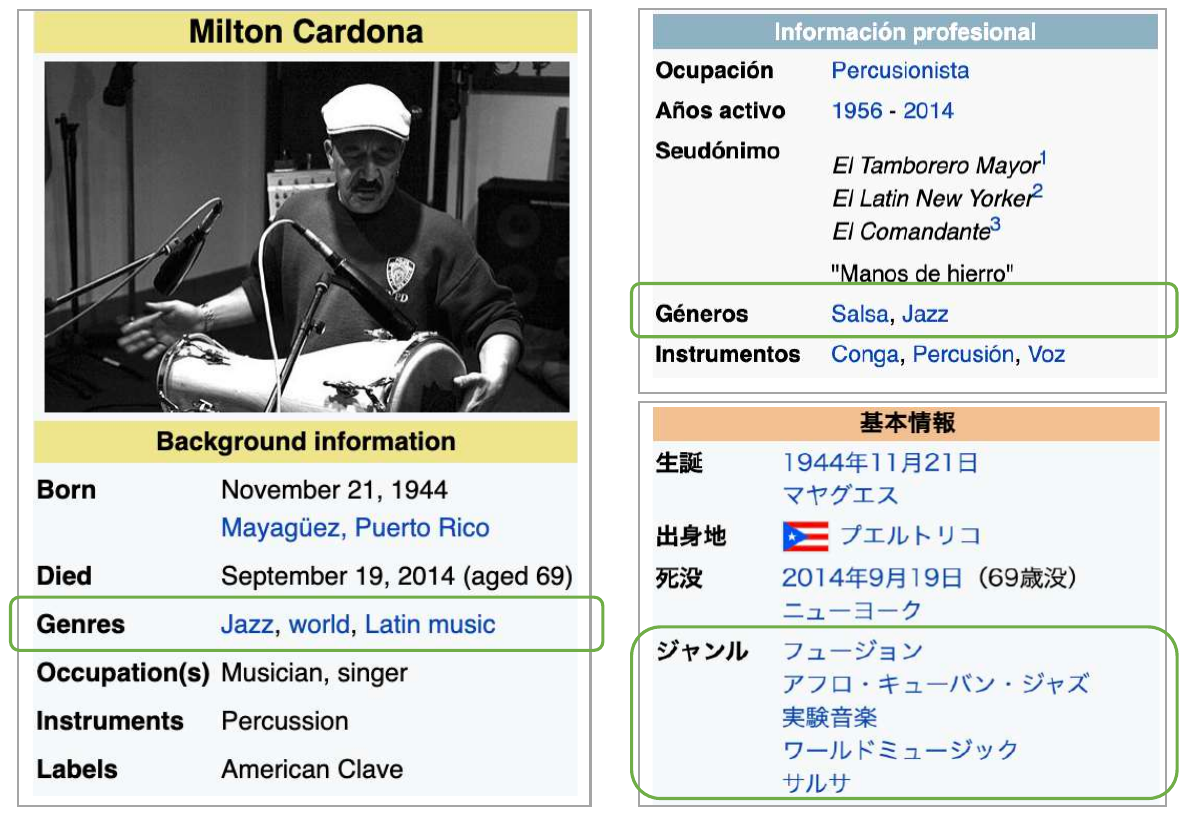}
    \caption[Infoboxes of Milton Cardona, in English, Spanish, and Japanese]{Wikipedia infoboxes of Puerto Rican artist Milton Cardona, in English, Spanish, and Japanese. Some music genre annotations are culture-specific, such as \textit{World}/ワールドミュージック which is present in English and Japanese but not in Spanish, or \textit{実験音楽} (\textit{Experimental Music}) in Japanese only.}
    \label{fig:emnlpexample}
\end{figure}

\end{CJK}

We identify Wikipedia\footnote{\href{https://en.wikipedia.org}{https://en.wikipedia.org}}, the online multilingual encyclopedia, to be particularly relevant to our study. Indeed, Wikipedia records worldwide music artists and their discographies, with a frequent mention of their music genres.
By manually checking the Wikipedia pages of music items, we observed that their music genres vary significantly across languages. 
For instance, \textit{Knights of Cydonia}, a single by \textit{Muse}, was annotated in Spanish as \textit{progressive rock}, while in Dutch as \textit{progressive metal} and \textit{alternative rock}. In Figure~\ref{fig:emnlpexample}, we provide another example of different annotations in English, Spanish, and Japanese from Wikipedia infoboxes.

As Wikipedia's writing is localized, the various cultures of its contributors can lead to differences in the multilingual content on the same topic, particularly for subjective matters. We refer the interested reader to the study of Pfeil et al.~\cite{Pfeil2006}, arguing that Wikipedia's contributions expose cultural differences aligned with the real-world ones. Therefore, Wikipedia appears as a suitable source for assembling a test corpus.

Using DBpedia \cite{Auer_2007} as a proxy to Wikipedia, we collected music items such as artists and albums, annotated with music genres in at least two of the six languages (English, Dutch, French, Spanish, Czech, and Japanese).
We targeted the \emph{MusicalWork}, \emph{MusicalArtist} and \emph{Band} DBpedia resource types, and we only kept music items that were annotated with music genres which appeared at least $15$ times in the corpus.
Our final corpus includes 63~246 music items, and was publicly released along with the paper associated with this work~\cite{epure2020modeling}. The number of annotations for each language pair is presented in Table \ref{tab:corpus_items}.
We also show in Table \ref{tab:no_genres} the number of unique music genres per language in the corpus and the average number of tags for each music item.

The English and Spanish languages use the most diverse tags.
This can be because more annotations exist in these languages, in comparison to Czech, which has the least annotations and least diverse tags.
However, the mean number of tags per item appears relatively high for Czech, while Japanese has the smallest average number of tags per item.

\begin{table}[t]
\centering
\caption[Number of music items for each language pair]{Number of music items for each language pair.}
\label{tab:corpus_items} 
\resizebox{0.55\textwidth}{!}{
\begin{tabular}{c|ccccc}
\toprule
\textbf{Language}  & Dutch  & French  & Spanish  & Czech & Japanese \\
\midrule
\midrule
English  &  12~604  & 28~252 & 32~891 & 4~772 & 14~752 \\
Dutch  & & 7~139 & 7~689 & 1~885 & 3~426 \\
French  & & &  15~616 & 3~046 & 8~622 \\
Spanish  & & & & 3~245 & 7~644 \\
Japanese  & & & & & 2~065 \\
\bottomrule
\end{tabular}
}
\end{table}

\section{Learning Music Genre Representations}
\label{c10s3}

This work assesses the possibility of obtaining relevant cross-lingual music genre annotations, able to capture cultural differences too, by relying on language-specific semantic representations. Again, such representations would be valuable for music streaming services such as Deezer, e.g., to improve localized content tagging and music genre-driven recommendation.

\subsection{General Overview}
\label{c10s30}

The first step of our method consists in obtaining initial embedding representations for music genres by leveraging \textit{pre-trained multilingual word embeddings} (Section \ref{c10s32}). However, directly using these embeddings for cross-lingual genre annotation is prone to underperform, as:
\begin{itemize}
    \item the embeddings often correspond to the most common word senses (e.g., \textit{country} can refer to \textit{nation}, and \textit{rock} could be closer in meaning to \textit{stone}), and not to their musical senses;
    \item some music genres could contain rare words which are absent from the pre-trained model vocabulary, resulting in potentially unknown tag embeddings.
\end{itemize}

To address these issues, we complement our word embeddings with semantics from knowledge bases that expose concept relations. Specifically, in a second step, we assemble \textit{graph ontologies} to represent music genre relations, as described in Section \ref{c10s31}.
Then, we aim to adjust the initial genre embeddings to encode relations from these graphs, ensuring the domain adaptation. For this, but also to learn embeddings for concepts with unknown vocabulary words, we consider two different approaches based on retrofitting and on GAE/VGAE-inspired models, and presented in Sections~\ref{c10s33}~and~\ref{c10s34}, respectively.

We note that, in contrast to our unsupervised approach, mapping patterns of associating music genres with music items could have also been enabled with a parallel corpus~\cite{Epure2019}. However, gathering a corpus that includes all music genres for each pair of languages is challenging~\cite{epure2020modeling}.

\subsection{Initializing Music Genre Embedding Representations}
\label{c10s32}

\paragraph{Multilingual Word Embeddings}
Under the formalism introduced in Section \ref{c10s22}, our main goal relates to quantifying the degree of relatedness of two textual tags. 
This task is widely popular in the NLP community and contemporary approaches resort to expressing their relatedness via the distance between their corresponding word embeddings \cite{grave-etal-2018-learning,mikolov2018advances,pennington2014glove}. The mapping of words with embedding vectors is guided by the \textit{distributional hypothesis} \cite{harris1954}, which states that words occurring in similar contexts are likely to have similar meanings.

To measure the relatedness of multilingual words using embeddings obtained from monolingual corpora, an \textit{alignment} \cite{joulin-etal-2018-loss} between the language-specific embedding spaces is required. Through this alignment, we ensure that multilingual word embeddings are projected into a common space where they are comparable. Practically, a mapping function between two monolingual word embedding spaces is learned, for instance by using a bilingual lexicon~\cite{mikolov2013exploiting}.

In the experiments reported in this chapter, we leverage word embeddings obtained from the multilingual \textit{fastText} (FT) model proposed by Grave et~al.~\cite{grave-etal-2018-learning}, which we aligned by using the method described by Joulin~et~al.~\cite{joulin2017}. Embedding vectors are of dimension $d = 300$. We emphasize that this choice is not restrictive. In our EMNLP 2020 paper, we provide additional results for music genre word embeddings obtained from the recent BERT~\cite{devlin-etal-2019-bert}, LASER~\cite{Artetxe-Schwenk-2019} and XLM~\cite{lample2019cross} models, together with a comparison of these different methods. We omit these additional results in this thesis, for the sake of brevity.

\paragraph{Music Genre Embeddings}
Starting from these multilingual fastText word vectors, we now discuss our strategy to initialize our music genre embedding representations.
Music genres can contain multiple words.
We claim that the \textit{compositionality principle}, stating that the meaning of a multi-word expression is dictated by the meaning of each word, often holds for our case.
For instance, \textit{Dance pop} is related to \textit{dance} and \textit{pop} or \textit{Balada romántica} is a type of \textit{ballad} which is \textit{romantic}\footnote{Exceptions from the principle also exist, e.g., \textit{hard rock}.}.
The contemporary approach for compositional embeddings consists in learning a function which derives the embeddings of a multi-word expression from the embeddings of its words~\cite{shwartz-2019-systematic}. In this work, we consider two music genre initialization strategies:
\begin{itemize}
    \item the first one, denoted \textit{avg}, simply consists in \textit{averaging word embeddings}.
Formally, let $\mathcal{V} = \{c_1, c_2, \dots, c_n\}$ be the multilingual vocabulary, $c_i$ being a concept (i.e., a genre) composed of at least one word. 
We aim to compute $\hat{Q}\in \mathbb{R}^{n \times d}$, the embedding matrix for the vocabulary $\mathcal{V}$, where the $i$-th row $\hat{q}_i \in \mathbb{R}^d$ denotes the embedding of $c_i$.
If $c_i$ is composed of the following words, $\{w_1, w_2, \dots, w_M\}$, $\hat{q}_i$ can be computed as $\frac{1}{M}\sum_{m=1}^M{z_{w_m}}$, where $z_{w_m}$ is the embedding vector of $w_m$.
We note that, if $c_i$ contains words absent from the pretrained fastText vocabulary, the $d$-dimensional null vector $0_d$ is used by default;
\item the second one, denoted \textit{sif}, exploits the fact that some words in a compounded expression may be more illustrative than others.
The more frequently a word is observed in a corpus, the more likely it is that the word is common for a language and semantically less informative (e.g., \textit{music} in \textit{post industrial music}). 
Thus, the compositional embedding computation of a multi-word expression can be modified such that the contribution of each word embedding is inversely proportional to its frequency.
Pre-trained word embeddings are generally released sorted by decreasing word corpus frequency. 
Let $r_{w_m}$ be the rank of $w_m$ in this vocabulary.
Then, based on Mandelbrot's generalization~\cite{mandelbrot1953informational} of Zipf's~law~\cite{Zipf1949}, its frequency $f_{w_m}$ can be estimated as follows: $f_{w_
{m}} = 1 / (r_{w_{m}} + 2.7)$.

In the following, we rely on the \textit{smooth inverse frequency} (\textit{sif}) based averaging proposed by Arora et al.~\cite{arora2016simple} to compute the multi-word expression embeddings.
This method is aligned with our previous observations and proven highly effective compared to more complex neural network-based models on a large diversity of NLP tasks \cite{arora2016simple}.
Given $f_{w_{m}}$ the estimated frequency of the word $w_m$ and $a$ a fixed hyperparameter\footnote{Arora et al.~\cite{arora2016simple} argue that $a=10^{-3}$ is a suitable choice for various pre-trained models.}, $\hat{q}_i$ is~computed~as:
\begin{equation}
  \overline{q}_i = \frac{1}{M}\sum_{m=1}^M{\frac{a}{a + f_{w_{m}}}z_{w_m}},  
\label{eq:sif1}
\end{equation}
\begin{equation}
    \hat{q}_i = \overline{q}_i - uu^T\overline{q}_i,
\end{equation}
where $u$ is the first singular vector from the SVD of $\overline{Q}$ obtained with the Equation~\eqref{eq:sif1}.

\end{itemize}

Using one of these two methods, we can derive initial music genre embedding representations. In the remainder of Section~\ref{c10s3}, we will improve these (\textit{avg}- or \textit{sif}-based) embeddings using retrofitting or GAE/VGAE-inspired models, to encode relations from the ontologies presented in the next Section~\ref{c10s31}.

\subsection{Music Genre Graph Ontologies}
\label{c10s31}

Conceptually, music genres are interconnected entities. Researchers often use \textit{ontologies} to represent music genre relations and enrich the music genre definitions \cite{Lisena2018,raimond2007music,schreiber2016genre}. These ontologies are graphs of conceptually related music genres, acting as nodes and connected through various relation-specific edges. For instance, \textit{rap west coast} is a subgenre of \textit{hip hop}, while \textit{punk} and \textit{electronic music} are the origin~of~\textit{synthpunk}.

\begin{multicols}{2}

In this chapter, we consider Wikipedia-based ontologies, extracted from the internet using the procedure described below. Wikipedia extensively documents worldwide music genres, relating them through a coherent set of relation types across languages, including the aforementioned subgenre and origin relations.
Though the relations types are the same per language, the actual music genres and the way they are related can differ and expose cultural differences in the music genre perception, as emphasized by Pfeil et al.~\cite{Pfeil2006}. We further describe how we crawled these ontologies, by relying on DBpedia as in Section~\ref{c10s23}.
\columnbreak
\vspace{-0.2cm}
\captionof{table}[Number of unique music genres in the evaluation corpus and in the ontologies]{Number of unique music genres in the evaluation corpus (presented in Section~\ref{c10s23}) and in the graph ontologies (presented in Section~\ref{c10s31}).}\label{tab:no_genres}
\vspace{0.4cm}
\resizebox{0.5\textwidth}{!}{
\begin{tabular}{c|c|c}
\toprule
 & \multicolumn{2}{c}{\textbf{Number of unique music genres}}  \\ 
\textbf{Language} & in the evaluation  & in the \\
  & corpus \textit{(avg. per item)} & ontologies \\
\midrule \midrule
English & 558 \textit{(2.12 $\pm$ 1.34)} & 10~748 \\
Dutch & 204 \textit{(1.71 $\pm$ 1.06)} & 1~529 \\
French  & 364 \textit{(1.75 $\pm$ 1.06)} & 2~905 \\
Spanish  & 525 \textit{(2.11 $\pm$ 1.34)} & 3~988\\
Czech  & 133 \textit{(2.23 $\pm$ 1.34)} & 1~418\\
Japanese  & 192 \textit{(1.51 $\pm$ 1.11)} & 1~609\\
\bottomrule
\end{tabular}
}
\end{multicols}
\vspace{-0.3cm}
Firstly, for each language, we constituted a seed list using two sources: the DBpedia resources of type \textit{MusicGenre} and their aliases linked through the \textit{wikiPageRedirects} relation; the music genres discovered when collecting the evaluation corpus (introduced in Section~\ref{c10s23}) and their aliases.
Then, music genres were fetched by visiting the DBpedia resources linked to the seeds through the relations: \textit{wikiPageRedirects}, \textit{musicSubgenre}, \textit{stylisticOrigin}, \textit{musicFusionGenre} and \textit{derivative}\footnote{We present these relations by their English names, which may be translated in DBpedia in other languages.}.
The seed list was updated each time, allowing the crawling to continue until no new resource was found. In DBpedia, resources are sometimes linked to their equivalents in other languages through the relation \textit{sameAs}.
In most experiments, we will rely on monolingual music genres ontologies (i.e., on six isolated graphs). However, we also collected the cross-lingual links between music genres to include a translation baseline for cross-lingual annotation, i.e., for each music genre in a source language, we predicted its equivalent in a target language using DBpedia's \textit{sameAs} (see Section~\ref{c10s4}). 

The number of unique music genres discovered in each language is presented in Table~\ref{tab:no_genres}. As our evaluation corpus from Section~\ref{c10s23}, these graphs were publicly released along with the paper associated with this work~\cite{epure2020modeling}. We note that the number of nodes is significantly larger than the number of genres in the evaluation corpus, emphasizing the challenge of constituting a parallel corpus that would cover all language-specific music genres.

\subsection{Improving Music Genre Representations with Retrofitting}
\label{c10s33}

In this section, and in the following one, we present our two strategies to improve the initial music genre embedding vectors from Section~\ref{c10s32}, by encoding information from the graph ontologies of Section~\ref{c10s31} into these vectors. Firstly, we present a method referred to as \textit{retrofitting}~\cite{faruqui2015} in this section, while the following one will focus on GAE/VGAE-inspired approaches.

Retrofitting \cite{faruqui2015} is a method to refine word embedding vectorial representations by considering the relations between words as defined by a graph, e.g., an ontology. It consists in modifying the word embeddings to become closer to the representations of the concepts to which they are related. Ever since the original work of Faruqui et~al.~\cite{faruqui2015}, many uses of retrofitting have been explored to semantically specialize word embeddings in relations such as synonyms or antonyms \cite{kiela-etal-2015-specializing,kim-etal-2016-adjusting}, in other languages than a source one \cite{ponti-etal-2019-cross} or in specific domains \cite{hangya-etal-2018-two}.

Formally, let $\mathcal{G}=(\mathcal{V}, \mathcal{E})$ be a graph including the concepts $\mathcal{V}$ and the semantic relations between these concepts $\mathcal{E} \subseteq \mathcal{V} \times \mathcal{V}$.
The goal of retrofitting is to learn new concept embeddings $q_i \in \mathbb{R}^d$, stacked up in the matrix $Q \in \mathbb{R}^{n \times d}$, with $n=|\mathcal{V}|$ and $d$ the embedding dimension.
The learning starts by initializing each $q_i$, i.e., the new embedding vector for a concept $i\in \mathcal{V}$, to $\hat{q}_i$, the initial embedding vector. Then, it consists in iteratively updating $q_i$ until convergence, as follows:
\begin{equation}
    q_i \leftarrow{} \frac{\sum_{j:(i,j) \in \mathcal{E}}{(\beta_{ij}+ \beta_{ji})q_j} + \alpha_i \hat{q}_i}{\sum_{j:(i,j) \in \mathcal{E}}{(\beta_{ij}+\beta_{ji})} + \alpha_i}.
    \label{eq:update}
\end{equation}
The $\alpha$ and $\beta$ terms are positive scalars weighting the importance of the initial embedding vector and of the ones from related concepts in computations, respectively. Equation~\eqref{eq:update} was reached through the minimization of the convex retrofitting loss: 
\begin{equation}
\Phi(Q) = \sum_{i \in \mathcal{V}}\alpha_i ||q_i - \hat{q}_i||^2_2 
+ \sum_{i =1}^{n} \sum_{(i, j)\in \mathcal{E}}{\beta_{ij}||q_i - q_j||^2_2},
\label{retrofitting}
\end{equation}
using the Jacobi method \cite{saad2003iterative}. We note that Equation~\eqref{eq:update} is actually a corrected version of the iterative procedure initially proposed by Faruqui et~al.~\cite{faruqui2015}. For a concept $i$, not only $\beta_{ij}$ appears in it, but also $\beta_{ji}$ (which was omitted in the original formula of Faruqui et~al~\cite{faruqui2015}).

The further modifications that we make regard the parameters $\alpha$ and $\beta$.
For each $i\in \mathcal{V}$, Faruqui et~al.~\cite{faruqui2015} originally fixed $\alpha_i$ to $1$, and $\beta_{ij}$ to $\frac{1}{D_{ii}}$ (with $D_{ii}$ the degree of $i$, as in Definition~\ref{def:degree_mat}) for $(i,j) \in \mathcal{E}$ or $0$ otherwise. However, while many word embedding models can handle unknown words, some concepts may still have unknown initial vectors, depending on the model's choice.
For this case, expanded retrofitting \cite{speer2016ensemble} has been proposed, considering $\alpha_i=0$, for each concept $i$ with unknown initial distributed vector, and $\alpha_i=1$ for the other ones: $q_i$ is initialized to $0_d$ and updated by averaging the embeddings of its related concepts~at~each~iteration.

We adopt the same approach to initialize the $\alpha_i$ parameters. We also adjust the $\beta_{ij}$ to weight the importance of each related concept embedding depending on the relation semantics in our genre ontologies. Specifically, we distinguish between ``equivalence'' and~``relatedness''~as~follows:
\begin{equation}
\overline{\beta}_{ij} = \left\{
        \begin{array}{ll}
        1 & : (i,j) \in \mathcal{E}_{\text{equi}} \subset \mathcal{E},\\
        \beta_{ij} & : (i,j) \in \mathcal{E}_{\text{rel}} = \mathcal{E} \setminus \mathcal{E}_{\text{equi}}, \\ 
        0, & : (i,j) \not\in \mathcal{E},
        \end{array}
    \right.
\end{equation}
where $\mathcal{E}_{\text{equi}}$ contains the ``equivalence'' relation type \textit{wikiPageRedirects}, and
$\mathcal{E}_{\text{rel}}$ contains the ``relatedness'' relation types \textit{stylisticOrigin}, \textit{musicSubgenre}, \textit{derivative}~and~\textit{musicFusionGenre}.

Finally, we want to highlight an important aspect of retrofitting.
Previous studies \cite{Fang2019,hayes-2019-just,speer2016ensemble} claim that, while the retrofitting updating procedure converges, the results depend on the order in which the updates are made.  We prove in the supplementary Section~\ref{c10s6} that the retrofitting loss is actually \textit{strictly convex} when at least one initial concept vector is known in each connected component.
Hence, with this condition satisfied, retrofitting always converges to the same solution, independently of the ordering of updates.

\subsection{Improving Music Genre Representations with GAE and VGAE}
\label{c10s34}

In essence, the retrofitting procedure consists in learning reinforced embedding representations of some nodes (corresponding to some music genres), by jointly leveraging 1)~their connections in a graph, and 2)~some initial vectorial representations of these nodes, in an unsupervised fashion. In the previous chapters, and on numerous occasions, we also achieved such a combination by training GAE and VGAE models. Therefore, while our EMNLP 2020 paper~\cite{epure2020modeling} mainly focused on improving genre representations with retrofitting, leveraging GAE and VGAE models appears as a natural alternative strategy to consider, in the context of this PhD thesis. Formally:
\begin{itemize}
 \item the graph $\mathcal{G} = (\mathcal{V},\mathcal{E})$, on which the GAE or VGAE model should be trained, is the one gathering all music genres from the six languages under consideration, as nodes, and their connections in the six ontologies, as edges. For consistency with the previous retrofitting method, we consider the \textit{wikiPageRedirects}, \textit{stylisticOrigin}, \textit{musicSubgenre}, \textit{derivative} and \textit{musicFusionGenre} links, but not the DBpedia's \textit{sameAs} links. Therefore, $\mathcal{G}$ actually includes six isolated subgraphs, corresponding to our six monolingual~genre~ontologies;
    \item the feature vector $x_i \in \mathbb{R}^f$ of each node $i \in \mathcal{V}$, which acts as the input layer representation of  $i$ in the context of a GAE or VGAE model, corresponds, in this application, to the initial word embedding vector of the music genre $i$, i.e., $x_i = \hat{q}_i \in \mathbb{R}^d$.  We observe that, in this application, we have $f = d$, i.e., the input and output representations will have the same dimension;
    \item the final embedding representation $z_i \in \mathbb{R}^d$ of each node $i \in \mathcal{V}$ corresponds to the vector denoted as $q_i  \in \mathbb{R}^d$ in the previous section.
\end{itemize}

Nonetheless, as we argue in the next two paragraphs, a direct use of standard GAEs or VGAEs is prone to underperform in this setting. This motivates several modifications of the encoder and decoder components of these models, which we describe in the same paragraphs.

\paragraph{Encoder} Unlike previous chapters, in our training graph ontologies, nodes can be connected through edges of five \textit{different natures}. Our definition of a graph from Definition~\ref{defgraph} can be extended to represent such structures, by replacing the notation $\mathcal{G} = (\mathcal{V},\mathcal{E})$~by: $\mathcal{G} = (\mathcal{V}, \mathcal{E}_1, \mathcal{E}_2, \dots, \mathcal{E}_5)$,
where each $\mathcal{E}_i \subset \mathcal{E}$ (with $i \in \{1, \dots, 5\}$ and $\bigcup_i \mathcal{E}_i = \mathcal{E}$) only includes the edges of a particular type, among \textit{wikiPageRedirects}, \textit{stylisticOrigin}, \textit{musicSubgenre}, \textit{derivative} and \textit{musicFusionGenre}. For the sake of simplicity, and as in Section~\ref{c10s33}, we can also alternatively consider:
\begin{equation}
\mathcal{G} = (\mathcal{V}, \mathcal{E}_{\text{equi}}, \mathcal{E}_{\text{rel}}),
\label{equirel}
\end{equation}
where, as previously defined, $\mathcal{E}_{\text{equi}}$ contains edges of ``equivalence'' relation type \textit{wikiPageRedirects}, and
$\mathcal{E}_{\text{rel}}$ contains edges of the ``relatedness'' relation types \textit{stylisticOrigin}, \textit{musicSubgenre}, \textit{derivative}~and~\textit{musicFusionGenre}. Then, extending our Definition~\ref{def:adj}, we summarize $\mathcal{G}$ by two adjacency matrices $A_{\text{equi}}$ and $A_{\text{rel}}$, indicating edges of each subgroup. 

In this work, we aim to capture the dichotomy of Equation~\eqref{equirel} in the message passing operations of our GAE/VGAE encoders. We expect such an aspect to improve our representations, w.r.t. a standard encoding scheme ignoring the different natures of edges. Specifically, in our experiments, we employ the \textit{linear encoders} from Chapter~\ref{chapter_6} (both for the sake of simplicity, and as they returned comparable results w.r.t. multi-layer GCNs in our preliminary tests). However, we replace the operation from Definition~\ref{def:linearencoder} of Chapter~\ref{chapter_6} (i.e., $Z = \tilde{A} X W$) by\footnote{Here, for clarity of exposition, we discuss the deterministic GAE framework. However, the changes are equally applicable to the VGAE framework, for which $Z$ has to be replaced by $\mu$ and $\log \sigma$ as in Equation~\eqref{linvgaec6equation}.}:
\begin{equation}
Z = \tilde{A}_{\text{rel}} X W_{\text{rel}} + \tilde{A}_{\text{equi}} X W_{\text{equi}},
\label{newencoder}
\end{equation}
for some weight matrices $W_{\text{rel}} \in \mathbb{R}^{d \times d}$ and $W_{\text{equi}} \in \mathbb{R}^{d \times d}$, tuned by gradient descent/ascent to optimize the loss/objective presented in the next paragraph. Consistently with the notation adopted throughout this thesis, $X \in \mathbb{R}^{n \times d}$ is the node feature matrix, stacking up all $x_i$ (i.e., $\hat{q}_i$) vectors, and $Z \in \mathbb{R}^{n \times d}$ is the node embedding matrix, stacking up all $z_i$ (i.e., $q_i$) vectors. We adopt an out-degree normalization (see Definition~\ref{def:norm_c6}) for the two adjacency matrices.

\paragraph{Decoder} In this chapter, our main goal is \textit{not} to precisely reconstruct our graph ontologies from a music genre embedding space. In particular, consistently with Equation~\eqref{eq:transfinit}, we would want the music genres of a given music item in different languages to have large inner products in the embedding space. As these genres are not connected in $\mathcal{G}$, a standard GAE/VGAE with an inner product decoder and a reconstruction loss/objective based on the observed edges in $\mathcal{G}$ is unlikely to achieve such a result. As a consequence, we also adapt our decoding schemes, by instead incorporating the retrofitting loss function presented in Section~\ref{c10s33}.

Specifically, we propose a modified GAE-inspired model\footnote{One could wonder whether the models proposed in this section should be referred to as graph \textit{auto}encoders. On the one hand, they do not aim to reconstruct an original graph, which was the key objective of standard autoencoders. On the other hand, they remain conceptually very close to GAE and VGAE models~\cite{kipf2016-2}. For these reasons, we chose to refer to them as ``GAE-inspired'' and ``VGAE-inspired'' models, throughout this chapter.}. After encoding nodes as in Equation~\eqref{newencoder}, the ``decoder'' of such a model does not reconstruct the entire graph. It only evaluates $\Phi(Z)$, as defined in Equation~\eqref{retrofitting}, from the node embedding matrix $Z$. Then, during training, we tune $W_{\text{rel}}$ and $W_{\text{equi}}$ by iteratively minimizing $\Phi(Z)$, by gradient descent.

In a similar fashion, we also propose a modified VGAE-inspired model. After encoding nodes as above (and sampling the node embedding matrix $Z$, in this probabilistic setting), we tune weight matrices by iteratively maximizing, by gradient ascent, the following modified objective, that replaces the standard ELBO from Equation~\eqref{elbo} of Chapter~\ref{chapter_2}:
    \begin{equation}
    \mathcal{L}_{\text{VGAE-insp}} = -\Phi(Z) - \lambda \mathcal{D}_{KL}\Big(q(Z | A,X)||\tilde{p}(Z)\Big).
    \label{newelbo}
    \end{equation}
In Equation~\eqref{newelbo}, the $-\Phi(Z)$ term replaces the ``reconstruction-based'' expectation of the standard ELBO from Equation~\eqref{elbo}. We also modify the Kullback-Leibler divergence term. In Equation~\eqref{newelbo}, $\tilde{p}$ corresponds to $\mathcal{N}(x_i, I_d)$ priors on the distribution of each $z_i$ vector. They replace the unit Gaussian priors, and aim to ensure that the $z_i$ embedding vectors remain relatively close to the initial vectors $x_i$ (i.e., $\tilde{q}_i$), consistently with the retrofitting loss. The scaling hyperparameter $\lambda \geq 0$ permits ensuring that both terms are comparable\footnote{Is anyone reading this? Send me a message! I would be glad to hear that someone has read my thesis so~meticulously.}.

\section{Experimental Analysis}
\label{c10s4}

In this section, we present and discuss our experimental analysis of the proposed methods, using the evaluation corpus scraped from DBpedia and presented in the previous Section~\ref{c10s23}.

\subsection{Experimental Setting}
\label{c10s41}

Cross-lingual music genre annotation, as formalized in Section~\ref{c10s22}, is a multi-label prediction task.
For evaluation, we report the mean and standard deviations of AUC scores, macro-averaged \cite{BRADLEY19971145}, using $3$-fold cross-validation.
For each language, we apply an iterative split \cite{Sechidis:2011} of the test corpus that balances the number of samples and the tag distributions across the folds.
We pre-process the music genres by either replacing special characters with space (\textit{\_-/,}) or removing them (\textit{()':.!\$}).
For Japanese, we introduce spaces between tokens obtained with Mecab\footnote{\href{https://taku910.github.io/mecab/}{https://taku910.github.io/mecab/}}.
Embedding vectors are then computed from pre-processed tags.

We test two translation baselines, one based on Google Translate\footnote{\href{https://translate.google.com}{https://translate.google.com}} and one directly leveraging the aforementioned DBpedia's \textit{SameAs} relations for genre translation.
In this case, the source music genres are mapped to a single or no target music genre. Furthermore, we consider both \textit{avg}-based and \textit{sif}-based variants of our fastText (FT) word embeddings, together with their corresponding retrofitted versions, adopting the hyperparameters described in the above sections. We also train GAE-inspired and VGAE-inspired models (with the modified linear encoders and loss/objective of Section~\ref{c10s34}) on both variants, using the Adam optimizer, without dropout, with a learning rate of 0.01 and for 200 training iterations. We released our source code on GitHub, along with datasets including our ontologies, embedding vectors, and evaluation corpus\footnote{\href{https://github.com/deezer/CrossCulturalMusicGenrePerception}{https://github.com/deezer/CrossCulturalMusicGenrePerception}}.

\textit{Disclaimer: the data pre-processing operations and retrofitting experiments were conducted by Elena V. Epure, lead author of the paper associated with this work. The GAE/VGAE-inspired experiments, absent from this paper, were subsequently designed and added for this thesis.}

\subsection{Results and Discussion}
\label{c10s42}

\paragraph{Results}

\begin{table}[t]
\centering
\caption[Cross-lingual music genre annotation]{Macro-AUC scores (in \%) for cross-lingual music genre annotation on our DBpedia-based evaluation corpus, presented in Section~\ref{c10s23}. The first part corresponds to baselines. The second part corresponds to (initial) fastText embedding vectors, with \textit{avg} or \textit{sif} averaging. The third part corresponds to methods combining graph ontologies and \textit{sif}-based fastText embedding vectors. \textbf{Bold} numbers correspond to the best scores.}
\label{tab:results}
\resizebox{0.9\textwidth}{!}{
\begin{tabular}{c|cc|cc|cc}
\toprule
\textbf{Language Pair} & \textbf{Google} & \textbf{DBpedia's} & \textbf{FT}$_{\text{avg}}$ & \textbf{FT}$_{\text{sif}}$  & \textbf{VGAE} & \textbf{Retrofitting} \\
Source $\rightarrow$ Target & \textbf{Translate} & \textbf{``SameAs''} &  &  & + \textbf{FT}$_{sif}$ & + \textbf{FT}$_{sif}$\\
\midrule
\midrule
English $\rightarrow$ Dutch & 59.9 $\pm$ 0.3 & {72.2 $\pm$ 0.2} & 75.2 $\pm$ 0.2   & {86.5  $\pm$ 0.1} & 89.4 $\pm$ 0.2 & \textbf{90.0  $\pm$ 0.1} \\
English $\rightarrow$ French & 58.4 $\pm$ 0.1 & {70.0  $\pm$ 0.2} &  78.6 $\pm$ 0.2   & {87.4  $\pm$ 0.3} & \textbf{90.8  $\pm$ 0.2} & \textbf{90.8  $\pm$ 0.2} \\
English $\rightarrow$ Spanish & 56.9 $\pm$ 0.0 & {65.4  $\pm$ 0.2} & 77.5 $\pm$ 0.1   & {86.9  $\pm$ 0.2} & \textbf{90.0 $\pm$ 0.2} & 89.9  $\pm$ 0.1 \\
English $\rightarrow$ Czech & 60.6 $\pm$ 0.6 & {78.4  $\pm$ 0.6} & 74.5 $\pm$ 0.6 & {88.6  $\pm$ 0.4} & 90.0 $\pm$ 0.4 & \textbf{90.4  $\pm$ 0.3} \\
English $\rightarrow$ Japanese & 60.9 $\pm$ 0.1 & {70.4  $\pm$ 0.2} & 69.6 $\pm$ 0.5 & {80.8  $\pm$ 0.3} & 86.1 $\pm$ 0.3 & \textbf{86.7  $\pm$ 0.3} \\ 
\hline
Dutch $\rightarrow$ English & 53.5 $\pm$ 0.1 & {56.7  $\pm$ 0.3} & 73.8 $\pm$ 0.5 & {79.8  $\pm$ 0.4} & \textbf{84.3  $\pm$ 0.3} & \textbf{84.3  $\pm$ 0.1} \\
Dutch $\rightarrow$ French & 54.4 $\pm$ 0.2 & {60.0  $\pm$ 0.4} & 63.9 $\pm$ 0.5  & {79.3  $\pm$ 0.8} & 81.0 $\pm$ 0.8  & \textbf{81.5  $\pm$ 0.7} \\
Dutch $\rightarrow$ Spanish & 53.1 $\pm$ 0.2 & {56.8  $\pm$ 0.2} & 63.7 $\pm$ 0.4 & {77.7  $\pm$ 0.5} & 79.9 $\pm$ 0.5 & \textbf{80.5  $\pm$ 0.4} \\
Dutch $\rightarrow$ Czech & {57.8 $\pm$ 0.1} & 50.0  $\pm$ 0.0 & 65.1 $\pm$ 0.3 & {80.6  $\pm$ 0.2} & 82.6 $\pm$ 0.6 & \textbf{83.4  $\pm$ 0.5} \\
Dutch $\rightarrow$ Japanese & 57.5 $\pm$ 0.4 & {62.7  $\pm$ 0.4} & 64.8 $\pm$ 0.2 & {74.9  $\pm$ 1.0} & 79.1 $\pm$ 0.9 & \textbf{80.0  $\pm$ 0.7} \\ \hline
French $\rightarrow$ Dutch & 58.6 $\pm$ 0.1 & {65.3  $\pm$ 0.4} & 67.7 $\pm$ 1.0 & {81.9  $\pm$ 0.4} & 83.7 $\pm$ 0.4 & \textbf{84.7  $\pm$ 0.3} \\
French $\rightarrow$ English &  55.3 $\pm$ 0.0 & {59.7  $\pm$ 0.2} & 76.2 $\pm$ 0.2 & {83.0  $\pm$ 0.2} & 87.4 $\pm$ 0.2 & \textbf{87.7  $\pm$ 0.1} \\
French $\rightarrow$ Spanish & 54.1 $\pm$ 0.1 & {59.0  $\pm$ 0.1} & 71.0 $\pm$ 0.2 & {81.8  $\pm$ 0.3} & 84.9 $\pm$ 0.3 & \textbf{85.3  $\pm$ 0.2} \\
French $\rightarrow$ Czech & 59.1 $\pm$ 0.3 & {70.0  $\pm$ 0.6} & 70.4 $\pm$ 0.8 & {83.9  $\pm$ 0.4} & 86.5 $\pm$ 0.5 & \textbf{87.2  $\pm$ 0.3} \\
French $\rightarrow$ Japanese & 59.1 $\pm$ 0.2 & {64.7  $\pm$ 0.5} & 71.1 $\pm$ 0.3 & {77.9  $\pm$ 0.1} & 80.5 $\pm$ 0.5 & \textbf{81.4  $\pm$ 0.3} \\ \hline
Spanish $\rightarrow$ Dutch & 59.8 $\pm$ 0.3 & {67.2  $\pm$ 0.2} & 68.5 $\pm$ 0.5 & {82.8  $\pm$ 0.9} & 85.2 $\pm$ 0.6 & \textbf{85.9  $\pm$ 0.6} \\
Spanish $\rightarrow$ French & 57.4 $\pm$ 0.2 & {64.8  $\pm$ 0.3} & 70.8 $\pm$ 0.4 & {85.0  $\pm$ 0.3} & \textbf{87.6  $\pm$ 0.5} & 87.5  $\pm$ 0.3 \\
Spanish $\rightarrow$ English & 57.0 $\pm$ 0.1 & {61.7  $\pm$ 0.0} & 75.3 $\pm$ 0.1  & {84.7  $\pm$ 0.2} & \textbf{88.8  $\pm$ 0.3} & \textbf{88.8  $\pm$ 0.3} \\
Spanish $\rightarrow$ Czech & 60.3 $\pm$ 0.2 & {72.2  $\pm$ 0.4} & 68.9 $\pm$ 0.8 & {85.6  $\pm$ 0.6} & 87.1 $\pm$ 0.6 & \textbf{88.0  $\pm$ 0.4} \\
Spanish $\rightarrow$ Japanese & 60.9 $\pm$ 0.1 & {67.0  $\pm$ 0.5} & 65.2 $\pm$ 0.4 & {78.3  $\pm$ 0.6} & 82.5 $\pm$ 0.6 & \textbf{83.1  $\pm$ 0.6} \\ \hline
Czech $\rightarrow$ Dutch & {57.6 $\pm$ 0.6} & 50.0  $\pm$ 0.0 & 68.5 $\pm$ 1.0 & {78.3 $\pm$ 0.9} & 79.9 $\pm$ 1.0 & \textbf{81.1  $\pm$ 1.2} \\
Czech $\rightarrow$ French & 54.2 $\pm$ 0.2 & {60.0  $\pm$ 0.3} & 64.5 $\pm$ 0.9 & 78.5  $\pm$ 0.2 & 80.4 $\pm$ 0.7 & \textbf{81.4  $\pm$ 0.3} \\
Czech $\rightarrow$ Spanish & 53.7 $\pm$ 0.4 & {56.9  $\pm$ 0.3} & 65.3 $\pm$ 0.9 & {77.7  $\pm$ 0.8} & 80.7 $\pm$ 0.8 & \textbf{81.6  $\pm$ 0.9} \\
Czech $\rightarrow$ English & 54.2 $\pm$ 0.2 & {57.1  $\pm$ 0.1} & 70.3 $\pm$ 0.5 & {78.9  $\pm$ 0.1} & 83.4 $\pm$ 0.6 & \textbf{84.5  $\pm$ 0.4} \\
Czech $\rightarrow$ Japanese & 58.6 $\pm$ 0.2 & {64.0  $\pm$ 0.3} & 67.1 $\pm$ 1.1 & {76.9  $\pm$ 0.1} & 79.3 $\pm$ 0.6 & \textbf{80.5  $\pm$ 0.5} \\ \hline
Japanese $\rightarrow$ Dutch & 54.8 $\pm$ 0.4 & {61.6  $\pm$ 1.1} & 62.0 $\pm$ 0.5 & {72.8  $\pm$ 1.0} & 74.0 $\pm$  0.7 & \textbf{76.9  $\pm$ 0.3} \\
Japanese $\rightarrow$ French & 53.3 $\pm$ 0.2 & {58.4  $\pm$ 0.2} & 66.0 $\pm$ 1.1 & {73.7  $\pm$ 0.6} & 76.4 $\pm$ 0.4 & \textbf{77.8  $\pm$ 0.1} \\
Japanese $\rightarrow$ Spanish & 52.7 $\pm$ 0.1 & {55.9  $\pm$ 0.4} & 63.2 $\pm$ 0.3 & {73.9  $\pm$ 0.4} & 77.5 $\pm$ 0.6 & \textbf{78.8  $\pm$ 0.5} \\
Japanese $\rightarrow$ Czech & 56.1 $\pm$ 0.5 & {65.7  $\pm$ 0.7} & 61.7 $\pm$ 1.3 & {77.5  $\pm$ 0.2} & 78.7 $\pm$ 0.7 & \textbf{80.7  $\pm$ 0.4} \\
Japanese $\rightarrow$ English & 52.5 $\pm$ 0.1 & {55.8  $\pm$ 0.1} & 72.1 $\pm$ 1.0 & {75.6  $\pm$ 0.3} & 80.2 $\pm$ 0.8 & \textbf{81.6  $\pm$ 0.8} \\
\bottomrule
\end{tabular}
}
\end{table}
Table \ref{tab:results} reports our results for cross-lingual music genre annotation, for all possible ``source $\rightarrow$ target'' language pairs. We observe that a standard translation via Google Translate leads to the lowest results, being outperformed by a knowledge-based translation using DBpedia's \textit{SameAs} links, more adapted to this music domain.
The results confirm that a naive translation fails to capture the dissimilar music genre perception across cultures.

On the contrary, the \textit{sif}-based fastText initial word embedding representations (FT$_{\text{sif}}$) can model quite well the varying music genre annotation across languages, w.r.t. the aforementioned baselines. FT$_{\text{sif}}$ embeddings also consistently outperform their FT$_{\text{avg}}$ counterparts for all language pairs, which confirms the relevance of smooth inverse frequency averaging for this application. In the experiments reported in \cite{epure2020modeling}, we reached consistent conclusions when considering vectors obtained from the BERT, LASER, and XLM models, instead of fastText.

The last two columns of Table \ref{tab:results} report the results of cross-lingual annotation when using the FT$_{\text{sif}}$ vectors retrofitted to monolingual music genre ontologies, and when combining FT$_{\text{sif}}$ vectors and ontologies using the VGAE-inspired approach. The latter reaches comparable results w.r.t. its GAE-inspired counterpart, which is therefore not reported in Table~\ref{tab:results} for brevity. In a nutshell, for all three methods (GAE-inspired, VGAE-inspired, and retrofitting), the domain adaptation of initial word embeddings improves our scores across all pairs of languages.

Nonetheless, in this application, retrofitting tends to outperform our GAE/VGAE-inspired models. Such a result emphasizes the effectiveness of the (arguably simpler) retrofitting approach. We also postulate that some of our current design choices might hurt the performances of GAE/VGAE-inspired models. In particular, in future experiments, one might try to improve the results of Table \ref{tab:results} by including DBpedia's \textit{sameAs} cross-lingual links, instead of training GAE/VGAE-inspired models on a graph consisting of six monolingual isolated subgraphs. Also, while we only separated ``equivalence'' edges from ``relatedness'' edges in our encoders, one might instead optimize different weight matrices for each of our five (or six, with \textit{sameAs}) relation types.

\paragraph{Modeling the Music Genre Perception: Discussion}
\begin{CJK}{UTF8}{ipxm}
Our results confirm that using translation to produce cross-lingual music genre annotations is limited, as it does not consider the culturally divergent perception of music genres.
Instead, by leveraging language-specific semantic representations, combined with GAE/VGAE or retrofitting, one can model this phenomenon rather well.
For instance, from \emph{Milton Cardona}'s music genres in Spanish, \emph{salsa} and \emph{jazz}, our Retrofitting+FT$_{\text{sif}}$ model correctly predicts the equivalent of \emph{fusion} (フュージョン) in Japanese. Yet, while a thorough qualitative analysis requires more work, preliminary exploration suggests that larger gaps in perception might still be inadequately modeled.
For instance, for \emph{Santana}'s album \emph{Welcome} tagged with \emph{jazz} in Spanish, it does not predict \emph{pop} in French.
\end{CJK}

Regarding the scores per language, we obtained the lowest ones for Japanese as the source.
We could explain this by either a more challenging test corpus or still incompatible embeddings in Japanese, possibly because of the quality of the individual embedding models for this language and the completeness of the Japanese music genre ontology. Also, we did not notice improvements for pairs of languages from the same family, e.g., French and Spanish. However, we would need a sufficiently large parallel corpus exhaustively annotated in all languages to reliably compare the performance for pairs of languages from the same family or different ones.

Finally, we noticed that given two languages $L_1$ and $L_2$, with more music genre embeddings in $L_1$ than in $L_2$ (from both the ontologies and the corpus), the results of mapping annotations from $L_1$ to $L_2$ seem always better than the results from $L_2$ to $L_1$.
This observation explains two trends in Table \ref{tab:results}.
Firstly, the scores achieved for English or Spanish as the source, the languages with the largest number of music genres, are the best.
Secondly, the results for the same pair of languages could vary a lot, depending on the role each language plays, i.e., source or target. One possible explanation is that the prediction from languages with fewer music genre tags such as $L_2$ towards languages with more music genre tags such as $L_1$ is more challenging because the target language contains more specific or rare annotations.
For instance, when checking the results per tag from Dutch to English, we observed that among the tags with the lowest scores, we found \textit{moombahton}, \textit{zeuhl}, or \textit{candombe}.
However, other common music genres, such as \textit{latin music} or \textit{hard rock}, were also poorly predicted, showing that other causes exist too. Is the unbalanced number of music genres used in annotations a cultural consequence?  Related work \cite{Ferwerda2016InvestigatingTR} seems to support this hypothesis.
Then could we design a better mapping function that leverages the unbalanced numbers of music genres in cross-cultural annotations? 
We will dedicate a thorough investigation of these questions as future work.

\subsection{The Muzeeglot Web Interface}
\label{c10s43}

\begin{figure}[t]
    \centering
    \includegraphics[width=0.9\textwidth]{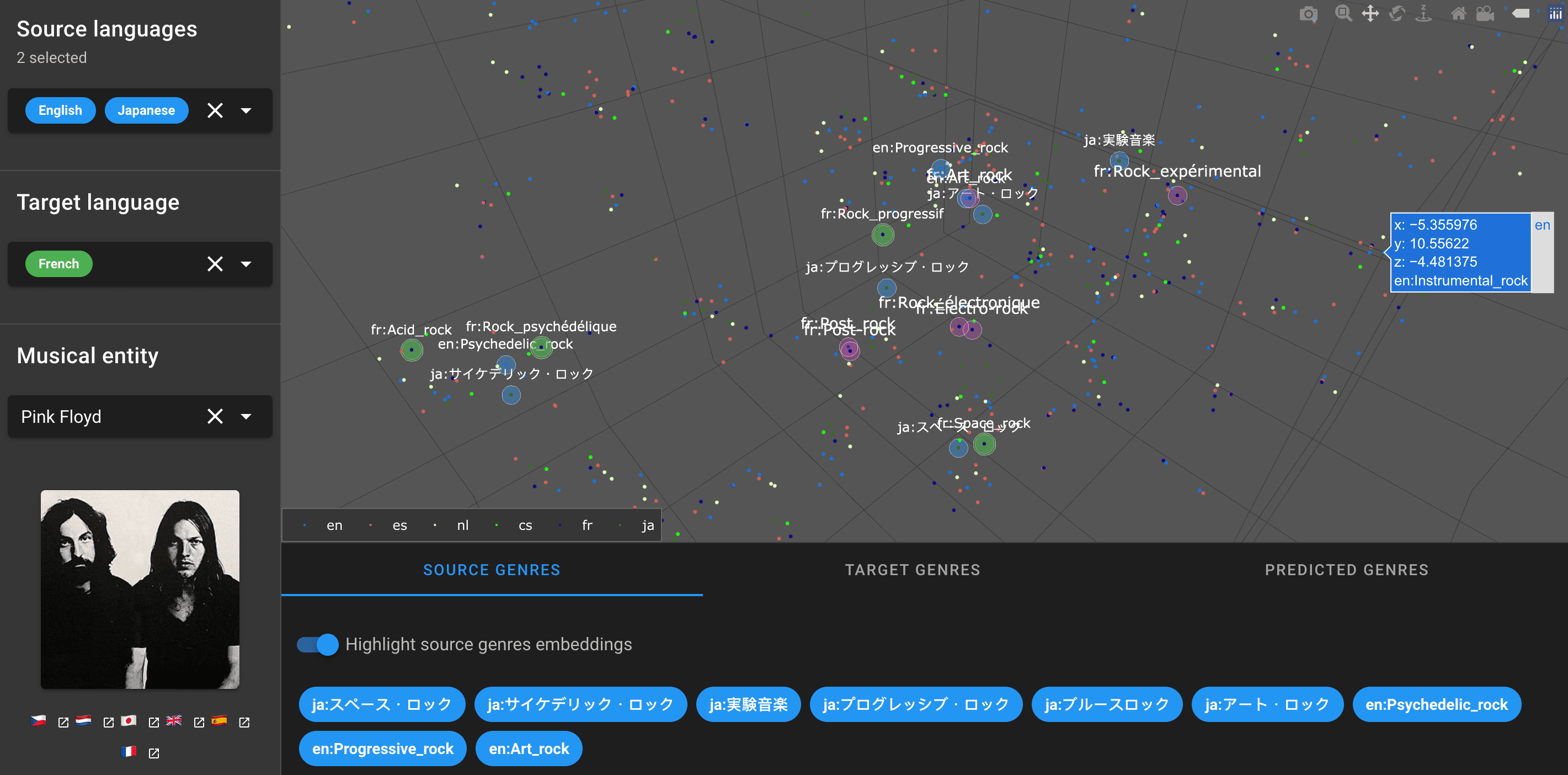}
    \caption[The Muzeeglot web interface]{The Muzeeglot web interface. Visualization of the music genres associated with the English band Pink Floyd on DBpedia, in English and in Japanese (two pre-selected  source languages), together with their Retrofitting+FT$_{\text{sif}}$ embedding representations. By clicking on the ``Predicting genres'' tab, one would display the predicted music genres in French (the pre-selected target language).}
    \label{fig:muzeeglot1}
\end{figure}

To support the research presented in this chapter, and in addition to our source code and datasets, our team publicly released the Muzeeglot\footnote{\href{https://github.com/deezer/muzeeglot}{https://github.com/deezer/muzeeglot}} prototype~\cite{epure2020muzeeglot}. As illustrated in Figure~\ref{fig:muzeeglot1}, Muzeeglot is a web interface. Based on a REST API developed in Python with FastAPI and on a frontend developed with VueJS~\cite{epure2020muzeeglot} by our research engineer Felix Voituret, and incorporating our models from this research, Muzeeglot permits visualizing multilingual music genre embeddings representations. For some pre-selected music entities from our corpus, and for some pre-selected source and target languages, Muzeeglot displays the corresponding ``ground truth'' music genres according to DBpedia, along with our predicted genres in the target language(s). 

\section{Conclusion}
\label{c10s5}

In this chapter, we aimed to model the music genre perception across language-bound cultures, to address a cross-lingual music genre annotation problem. Our study focused on six languages from four language families, and two common approaches to semantically represent concepts: music genre graph ontologies, and distributed word embeddings. We empirically showed that unsupervised cross-lingual music genre annotation is feasible with high accuracy when combining both types of semantic representations, using either a GAE/VGAE-inspired approach, or an alternative simple yet effective method based on retrofitting.

This is an important result for music streaming services such as Deezer. Indeed, in future work, our embedding representations could permit tagging music items with genre representations that capture cultural differences, which could improve several genre-related music information retrieval and music recommendation tasks, including the ones mentioned in Section~\ref{c10s1}. Besides, our work also provides a methodological framework to study the annotation behavior across language-bound cultures in other domains. Hence, the effectiveness of language-specific concept representations to model the culturally diverse perception could be further probed.

\section{Appendices}
\label{c10s6}

In this supplementary section, we prove the strict convexity of the retrofitting loss function, as claimed at the end of Section~\ref{c10s33}, This result was placed out of the main content of Chapter~\ref{chapter_10} for the sake of brevity and readability.

\begin{theorem}
 Let $\mathcal{V}$ be a finite vocabulary with $|\mathcal{V}| = n$.
Let $\mathcal{G} = (\mathcal{V},\mathcal{E})$ be an ontology represented as a directed graph which encodes semantic relationships between vocabulary words.
Further, let $\hat{\mathcal{V}} \subseteq \mathcal{V}$ be the subset of words which have non-zero initial distributed representations, $\hat{q}_i$. 
The goal of retrofitting is to learn the matrix $Q \in \mathbb{R}^d$, stacking up the new embeddings $q_i \in \mathbb{R}^d$ for each $i \in \mathcal{V}$.
The loss function to be minimized is:
\begin{equation}
\Phi(Q) = \sum_{i \in \hat{\mathcal{V}}}\alpha_i ||q_i - \hat{q}_i||^2_2  + \sum_{i =1}^{n} \sum_{(i, j)\in \mathcal{E}}{\beta_{ij}||q_i - q_j||^2_2 },
\end{equation}
where the $\alpha_i$ and $\beta_{ij}$ are positive scalars. 
Assuming that each connected component of $\mathcal{G}$ includes at least one word from $\hat{\mathcal{V}}$, the loss function $\Phi$ is strictly convex w.r.t. $Q$.
\end{theorem}

\begin{proof}
First of all, let $\hat{Q}$ denote the $n \times d$ matrix whose $i$-th row corresponds to $\hat{q}_i$ if $i \in \hat{\mathcal{V}}$, and to the $d$-dimensional null vector $0_d$ otherwise. 
Let $A$ denote the $n \times n$ diagonal matrix verifying $A_{ii} = \alpha_i$ if $i \in \hat{\mathcal{V}}$ and $A_{ii} = 0$ otherwise. 
Let $B$ denote the $n\times n$ symmetric matrix such as, for all $(i, j) \in \{1,...,n\}^2$ with $i \neq j$, $B_{ij} = B_{ji} = -\frac{1}{2} (\beta_{ij} + \beta_{ji})$ and $B_{ii} = \sum_{j=1,j\neq i}^n |B_{ij}|$. 
With such a notation, and with $\text{Tr}(\cdot)$ denoting the trace operator for square matrices, we have:
\begin{align*}
\sum_{i \in \hat{\mathcal{V}}} \alpha_i ||q_i - \hat{q}_i||^2_2 
&= \text{Tr}\Big( (Q - \hat{Q})^T A (Q - \hat{Q}) \Big) \\ 
&=  \text{Tr}\Big(Q^T A Q - \hat{Q}^T A Q - Q^T A \hat{Q} + \hat{Q}^T A \hat{Q}\Big).
\end{align*}
Also:
\begin{equation*}
\sum_{i =1}^{n} \sum_{(i, j)\in \mathcal{E}}{\beta_{ij}||q_i - q_j||^2_2 } = \text{Tr}\Big( Q^T B Q \Big).
\end{equation*}

Therefore, as the trace is a linear mapping, we have:
\begin{equation*}
\Phi(Q) = \text{Tr}\Big( Q^T (A + B)Q\Big) +  \text{Tr}\Big(\hat{Q}^T A \hat{Q} - \hat{Q}^T A Q - Q^T A \hat{Q}\Big).
\end{equation*}

Then, we note that $A + B$ is a weakly diagonally dominant matrix (WDD) as, by construction, $\forall i \in \{1,...,n\}, |(A+B)_{ii}| \geq \sum_{j \neq i} |(A+B)_{ij}|$. 
Also, for all $i \in \hat{\mathcal{V}}$, the inequality is strict, as $|(A+B)_{ii}| = \alpha_i + \sum_{j \neq i} |B_{ij}| > \sum_{j \neq i} |(A+B)_{ij}| = \sum_{j \neq i} |B_{ij}|$, which means that, for all $i \in \hat{\mathcal{V}}$, row $i$ of $A + B$ is strictly diagonally dominant (SSD). 
Assuming that each connected component of the graph $\mathcal{G}$ includes at least one node from $\hat{\mathcal{V}}$, we conclude that $A +B$ is a weakly chained diagonally dominant matrix \cite{azimzadeh2016weakly}, i.e., that:
\begin{itemize}
    \item $A+B$ is WDD;
    \item for each $i \in \mathcal{V}$ such that row $i$ is not SSD, there exists a \textit{walk} in the graph whose adjacency matrix is $A+B$ (two nodes $i$ and $j$ are connected if $(A+B)_{ij} = (A+B)_{ji} \neq 0$), starting from $i$ and ending at a node associated with a SSD row.
\end{itemize}
Such matrices are nonsingular \cite{azimzadeh2016weakly}, which implies that $Q \rightarrow Q^T (A + B) Q$ is a positive-definite quadratic form. As $A + B$ is a symmetric positive-definite matrix, there exists a matrix $M$ such that $A + B = M^T M$. Therefore, denoting $||\cdot||_F^2$ the squared Frobenius matrix norm:
\begin{equation*}
\text{Tr}\Big( Q^T (A + B) Q \Big) = \text{Tr}\Big( Q^T  M^T M Q \Big) = || Q M ||_F^2
\end{equation*}
which is strictly convex w.r.t. $Q$ due to the strict convexity of the squared Frobenius norm~\cite{dattorro2005convex}. 
Since the sum of strictly convex functions of $Q$ (first trace in $\Phi(Q)$) and linear functions of $Q$ (second trace in $\Phi(Q)$) is still strictly convex w.r.t. $Q$, we conclude that the loss function $\Phi$ is strictly convex w.r.t. $Q$.
\end{proof}

\paragraph{Corollary 10.2.} The retrofitting update procedure is insensitive to the order in which nodes are updated.

\begin{proof}
The aforementioned update procedure for $Q$ \cite{faruqui2015} is derived from the Jacobi iteration procedure \cite{bengio2006label,saad2003iterative} and converges for any initialization.
Such a convergence result is discussed in Bengio et~al.~\cite{bengio2006label}. 
It can also be directly verified in our specific setting by checking that each irreducible element of $A+B$, i.e., each connected component of the underlying graph constructed from this matrix, is irreducibly diagonally
dominant (see Section~4.2.3 in Saad~\cite{saad2003iterative}) and then by applying Theorem 4.9 from Saad~\cite{saad2003iterative} on each of these components.
Besides, due to its strict convexity w.r.t. $Q$, the loss function $\Phi$ admits a unique global minimum. 
Consequently, the retrofitting update procedure will converge to the same embedding matrix regardless of the order in which nodes are updated.
\end{proof}

\chapter[Carousel Personalization with Contextual Bandits]{Carousel Personalization with Contextual Bandits}\label{chapter_11}
\chaptermark{Carousel Personalization with Contextual Bandits}

\textit{This chapter presents research conducted with Walid Bendada and Théo Bontempelli, and published as a short paper in the proceedings of the 14\up{th} ACM Conference on Recommender Systems (RecSys 2020)~\cite{bendada2020carousel} where it received a ``best short paper'' honorable~mention.}

\section{Introduction}

As explained in the introduction, several other projects less related to GAE and VGAE models were also carried out during the last three years. We chose to present two of them in these last two chapters, as they directly relate to music recommendation, and as they led to online A/B tests and subsequently to model deployment on the Deezer service. They complement the previous chapters by providing a larger overview of some of Deezer's strategies and production-facing algorithms to recommend music. Specifically, in this Chapter~\ref{chapter_11}, we present some research on \textit{carousel personalization} at Deezer.

\begin{figure}[H]
    \centering
    \includegraphics[width=0.5\textwidth]{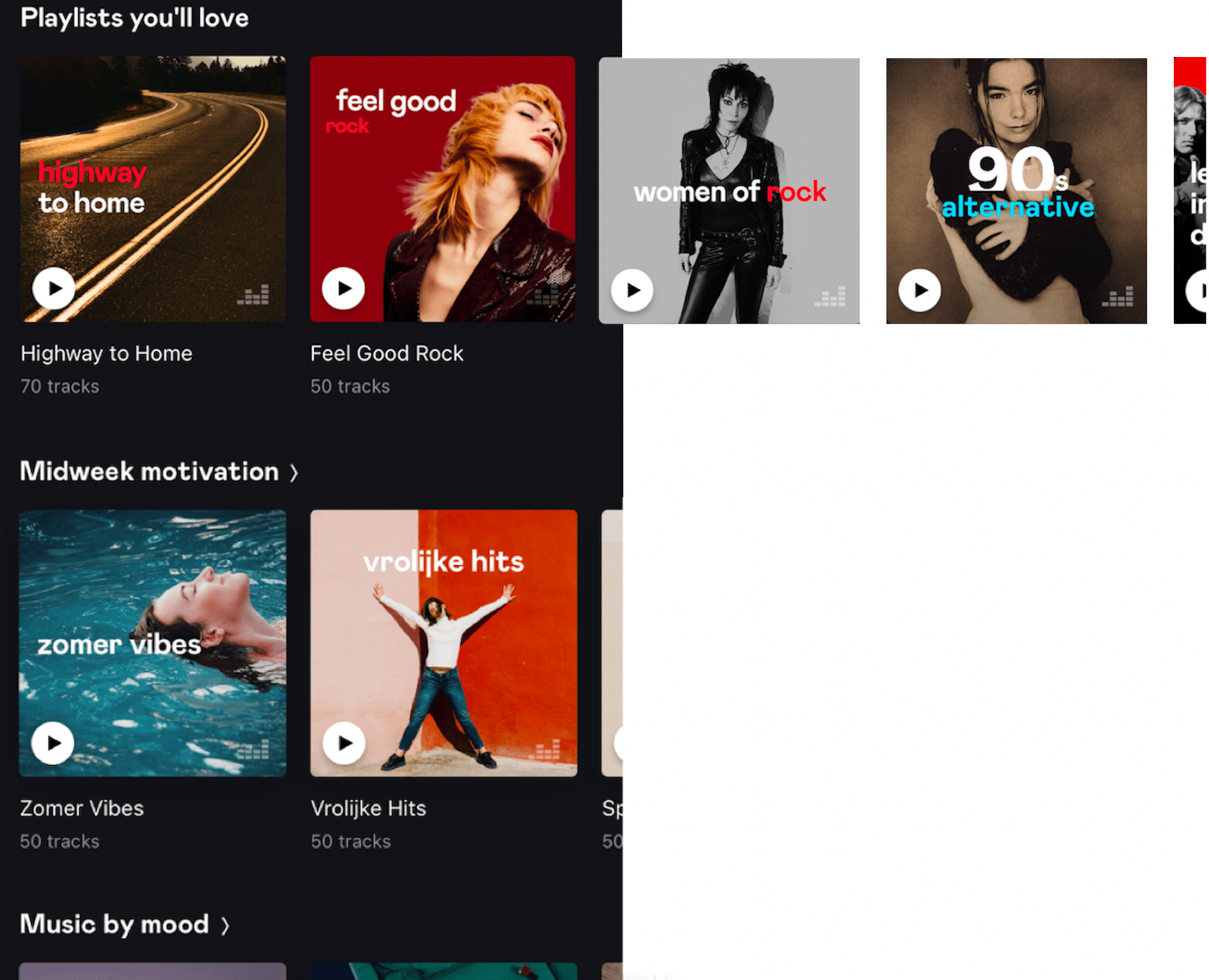}
    \caption[Personalized carousels on Deezer]{An example of a personalized swipeable carousel on the Deezer mobile app, to recommend playlists. These playlists were created by professional curators from Deezer with the purpose of complying with a specific music genre, cultural area, or mood. A few playlists are initially displayed to each user, who can click on them or swipe on the screen to see some of the additional recommended playlists from the carousel.}
    \label{fig:carousel_c11}
\end{figure}

Many mobile apps and websites, notably from the music streaming industry, leverage \textit{swipeable carousels} to display recommended content on their homepages. These carousels, also referred to as \textit{sliders} or \textit{shelves} \cite{mcinerney2018explore}, consist in ranked lists of items or \textit{cards} (albums, artists, playlists...). A few cards are initially displayed to the users, who can click on them or swipe on the screen to see some of the additional cards from the carousel. Selecting and ranking the most relevant cards to display is a challenging task \cite{gruson2019offline,huang2019contextual,ma2015introduction,mcinerney2018explore}, as the catalog size is usually significantly larger than the number of available slots in a carousel, and as users have different preferences. While being close to \textit{slate recommendation} \cite{ie2019slateq,kale2010non,swaminathan2017off} and to \textit{learning to rank} settings \cite{liu2009learning,pereira2019online,radlinski2008learning}, carousel personalization also requires dealing with user feedback to adaptively improve the recommended content via \textit{online learning} strategies \cite{anantharam1987asymptotically,chen2016combinatorial,hoi2018online}, and integrating that some cards from the carousel might not be seen by users due to the swipeable structure. Figure~\ref{fig:carousel_c11} provides an illustration of a swipeable carousel on Deezer.
\begin{multicols}{2}
In this chapter, we model carousel personalization as a \textit{multi-armed bandit with multiple plays}~\cite{anantharam1987asymptotically} learning problem. Within our framework, we account for important characteristics of real-world carousels, notably by considering that services have access to contextual information on user preferences, that they might not know which cards are actually seen by users, and that feedback data from carousels might not be available in real time. We show the effectiveness of our approach by addressing a large-scale carousel-based playlist recommendation task on Deezer. With the paper associated with this work~\cite{bendada2020carousel}, we released industrial data from our experiments and an open-source environment to simulate comparable carousel personalization learning problems.
\columnbreak
\end{multicols}
\vspace{-0.5cm}
This chapter is organized as follows. In Section~\ref{c11s2}, we introduce and formalize our multi-armed bandit framework for carousel personalization. We detail our data, our playlist recommendation task and our experimental setting in Section~\ref{c11s3}. We present and discuss our results in Section~\ref{c11s4}, and we conclude in Section~\ref{c11s5}.

\section{A Contextual Multi-Armed Bandit Framework for Carousel Personalization}
\label{c11s2}

In this section, we review key notions on multi-armed bandits and introduce our framework. We note that this Chapter~\ref{chapter_11}, as the upcoming Chapter~\ref{chapter_12}, will introduce quite different concepts w.r.t. those studied in previous chapters on GAE and VGAE models. We will therefore need to introduce a new \textit{notation} for several of these concepts. Nonetheless, when possible, we will try to remain consistent w.r.t. previous chapters (e.g., $K$ will still denote the number of clusters; $d$ will denote the dimension of embedding vectors, etc).

\subsection{Background on Multi-Armed Bandits with Multiple Plays}
\label{c11s21}

Multi-armed bandits are among the most famous instances of sequential decision making problems \cite{slivkins2019introduction,sutton2011reinforcement}. Multi-armed bandits \textit{with multiple plays} \cite{anantharam1987asymptotically,komiyama2015optimal} involve $N_{\text{arm}}$ entities called \textit{arms}. At each round $t = 1, 2, . . . , T$, a \textit{forecaster} has to select \textit{a set} $S_t \subset \{1,...,N_{\text{arm}}\}$ of $L < N_{\text{arm}}$ arms (while $L = 1$ in the \textit{single play} version of the problem \cite{slivkins2019introduction}). The forecaster then receives some \textit{rewards} from the selected arms, that we assume to be binary. The reward associated with an arm $i \in S_t$ is a sample drawn from a Bernoulli($p_i$) distribution, with $p_i \in [0,1]$ being an unknown parameter. Bernoulli distributions of arms $1,...,N_{\text{arm}}$ are assumed independent, which we later discuss. 

The objective of the forecaster is to maximize the sum of rewards received from the selected arms over time. It requires identifying the optimal set:
\begin{equation}
\Omega^*(L) \subset \{1,...,N_{\text{arm}}\}
\end{equation}
of the $L$ arms associated with the top-$L$ highest Bernoulli parameters, i.e., the $L$ highest expected rewards, as fast as possible.

In such problems, the forecaster faces an \textit{exploration-exploitation dilemma}. As the environment does \textit{not} reveal the rewards of the unselected arms, the forecaster needs to try all arms over time to identify the best ones (\textit{exploration}). However, selecting underperforming arms also leads to lower expected rewards, which encourages the forecaster to repeatedly select the assumed best ones (\textit{exploitation}). Over the past years, several strategies have been proposed and studied, providing efficient trade-offs between these two opposite objectives when sequentially selecting sets $S_t$. Notable examples include the Upper Confidence Bound (UCB) \cite{auer2002finite,chen2016combinatorial,lai1985asymptotically,wang2018regional} and Thompson Sampling (TS) \cite{chapelle2011empirical,komiyama2015optimal,thompson1933likelihood} algorithms (see Section \ref{c11s3}). The \textit{expected cumulative regret}:
\begin{equation}
    \text{Reg}(T)= \sum_{t=1}^T \Big( \sum_{i \in \Omega^*(L)} p_i - \sum_{i \in S_t} p_i \Big),
\end{equation} 
which represents the expected total loss endured by the forecaster by selecting non-optimal sets of arms at rounds 1 to $T$, is a common measure to compare the performances of strategies addressing this top-$L$ best arms identification problem \cite{anantharam1987asymptotically,chen2016combinatorial,komiyama2015optimal,slivkins2019introduction,sutton2011reinforcement,wang2018regional}.

\subsection{Carousel Personalization with Multi-Armed Bandits}
\label{c11s22}

Throughout this chapter, the $N_{\text{arm}}$ arms will correspond to a list of $N_{\text{arm}}$ cards/items, such as a catalog of albums or playlists in a music streaming app. They can be recommended to $n$ users through a swipeable carousel containing $L \ll N_{\text{arm}}$ slots. As users have various preferences, different cards can be displayed to different users. The $L$ recommended cards from the carousel of each user, i.e., the $L$ \textit{selected arms} for each user, are updated at regular intervals or \textit{rounds}, whose frequency depends on the technical constraints of the platform.

We aim to optimize \textit{display-to-stream rates}, i.e., to identify the $L$ cards for which each user is the most likely to click and then to \textit{stream} the underlying content, \textit{at least once} during the round. When a card $i$ is displayed to a user $u$, such streaming activity, i.e., a reward of 1, occurs during the round with an unknown probability $p_{ui} \in [0,1]$. Here, we assume that the number of cards, the number of users, and the display-to-stream probabilities $p_{ui}$ are fixed. We later discuss these assumptions.

A naive way to tackle this problem would consist in simultaneously running $n$ standard bandit algorithms, aiming to individually identify the top-$L$ cards with the highest $p_{ui}$ probabilities for each user $u$. This approach is actually unsuitable and would require a too long training time to reach convergence. Indeed, the number of display-to-stream parameters to estimate would be $N_{\text{arm}} \times n$, which is very large in practice as platforms often have millions of active users. In Section~\ref{c11s23}, we describe two strategies to address this problem by leveraging \textit{contextual information} on user preferences.

\subsection{Leveraging Contextual Information on User Preferences}
\label{c11s23}

\paragraph{Semi-Personalization via User Clustering}
Firstly, let us assume that we have access to a \textit{clustering of users}, constructed from users' past behaviors on the platform. Each user belongs to one of the $K$ groups $C_1, C_2,...,C_K$ with $K \ll n$. For instance, on a music streaming app, users from the same group would have homogeneous musical tastes. We propose to assume that users from the same group have identical expected display-to-stream probabilities for each card:
\begin{equation}
\forall c \in \{C_1,...,C_K\}, \forall u \in c, \forall i \in \{1,...,N_{\text{arm}}\}, p_{ui} = p_{ci}.
\end{equation}
Then, we run $K$ bandit algorithms, one for each cluster, to identify the top-$L$ best cards to recommend to each group. This strategy reduces the number of parameters to estimate to $N_{\text{arm}} \times K$, which is significantly fewer than $N_{\text{arm}} \times n$ in practice. Moreover, thanks to such users gathering, platforms receive more feedback on each displayed card w.r.t. the previous naive setting. This ensures a faster and more robust identification of optimal sets. However, the empirical performance of this strategy also strongly depends on the quality of the underlying user clustering.

\paragraph{Contextual Multi-Armed Bandits}

Instead of relying on clusters, let us now assume that we directly have access to a $d$-dimensional attribute vector $x_u \in \mathbb{R}^d$ for each user $u$. These vectors aim to summarize user preferences on the platform, e.g., their musical tastes (in terms of genres, moods, countries...) for a music streaming app. We assume that the expected display-to-stream probabilities of a user $u$ are functions of his/her attribute vector:
\begin{equation}
\forall i \in \{1,...,N_{\text{arm}}\}, p_{ui} = \sigma(x_u^T \theta_i),
\end{equation}
where the $\theta_i \in \mathbb{R}^d$ are $d$-dimensional weight vectors to learn for each of the $N_{\text{arm}}$ arms, and where $\sigma(\cdot)$ is the sigmoid function: $\sigma(x) = 1/(1 + e^{-x})$. This corresponds to the \textit{contextual bandit} setting \cite{agarwal2014taming,chu2011contextual,li2010contextual}, a popular learning paradigm for online recommender systems \cite{gruson2019offline,li2010contextual,li2011unbiased,mcinerney2018explore,qin2014contextual,tang2014ensemble,wang2017efficient,zhou2016latent,zoghi2017online}. Strategies to learn weight vectors are detailed, e.g., in \cite{chapelle2011empirical,mcinerney2018explore}. As $d \ll n$ in practice, such strategy also significantly reduces the number of parameters, to $N_{\text{arm}} \times d$. By design, users with similar preferences will have close expected display-to-stream probabilities. Moreover, all $n$ users can end up with different optimal carousels, contrary to the semi-personalized clustering approach.

\subsection{Capturing Characteristics of Real-World Carousels}
\label{c11s24}

In our framework, we also aim to capture other important characteristics of real-world swipeable carousels. In particular, while standard bandit algorithms usually consider that the forecaster receives rewards (0 or 1) from \textit{each} of the $L$ selected arms at each round, in our setting some selected cards might actually \textit{not} be seen by users. As illustrated in Figure~\ref{fig:carousel_c11}, only a few cards, say $L_{\text{init}} < L$, are initially displayed on a user's screen. The user needs to swipe right to see additional cards. As we later verify, ignoring this important aspect, and thus returning a reward of 0 for all unclicked cards at each round whatever their rank in the carousel, would lead to underestimating display-to-stream probabilities.

In this chapter, we assume that we do \textit{not} exactly know how many cards were seen by each user. Such an assumption is consistent with Deezer's actual usage data and is realistic. Indeed, on many real-world mobile apps carousels, users do not click on any button to discover additional cards, but, instead, need to continuously swipe left and right on the screen. As a consequence, the card display information is ambiguous, and is technically hard to track with accuracy.

To address this problem, we consider and later evaluate a \textit{cascade-based arm update} model. We draw inspiration from the \textit{cascade model} \cite{craswell2008experimental}, a popular approach to represent user behaviors when facing ranked lists of recommended items in an interface, with numerous applications and extensions \cite{katariya2016dcm,kveton2015cascading,lagree2016multiple,zong2016cascading}. At each round, we consider that:
\begin{itemize}
    \item an active user who did not stream any card during the round only saw the $L_{\text{init}}$ first ones;
    \item an active user who streamed the i\up{th} card, with $i \in \{1,...,L\}$, saw all cards from ranks 1 to $\text{max}(L_{\text{init}},i)$.
\end{itemize}
For instance, let $L_{\text{init}}=3$ and $L=12$. The reward vectors obtained from users who a) did not stream during the round, b) only streamed the 2\up{nd} card, and c) streamed the 2\up{nd} and 6\up{th} cards, are as follows, with $X$ denoting no reward:
\begin{align}
    &a:~[0, 0, 0, X, X , X, X, X, X, X, X, X] \nonumber \\
    &b:~[0, 1, 0, X, X, X, X, X, X, X, X, X] \nonumber \\
    &c:~[0, 1, 0, 0, 0, 1, X, X, X, X, X, X]
\end{align}

Lastly, to be consistent with real-world constraints, we assume that rewards are not processed on the fly but by \textit{batch}, at the end of each round, e.g., every day. We study the impact of such \textit{delayed batch feedback} in our upcoming experiments.

\subsection{Related Work}
\label{c11s25}

Bandits are very popular models for online recommendation \cite{li2010contextual,li2011unbiased,nguyen2014dynamic,pereira2019online,qin2014contextual,radlinski2008learning,tang2014ensemble,wang2017efficient,zhou2016latent}. In particular, McInterney~et~al.~\cite{mcinerney2018explore} and Gruson~et~al.~\cite{gruson2019offline} also recently studied carousel personalization in mobile apps. McInterney~et~al.~\cite{mcinerney2018explore} introduced a contextual bandit close to ours. However, their approach focuses more on explainability, they do not model cascade-based displays as we did in Section~\ref{c11s24}, and do not integrate semi-personalized strategies. Gruson~et~al.~\cite{gruson2019offline} also considered contextual bandits inspired by \cite{mcinerney2018explore} for playlist recommendation in carousels, but did not provide details on their models. They instead aimed to predict the online ranking of these models from various offline evaluations. Last, other different sets of ordered items have been studied \cite{ie2019slateq,jiang2018beyond,kale2010non,swaminathan2017off,zong2016cascading}.

\section{Application to Carousel Personalization on Deezer}
\label{c11s3}

In the following, we empirically evaluate and discuss the effectiveness of our framework.

\subsection{Experimental Setting: Carousel-Based Playlist Recommendation}
\label{c11s31}

We study a large-scale carousel-based \textit{playlist recommendation} task on the Deezer mobile app. We consider $N_{\text{arm}} = 862$ playlists, that were created by professional curators from Deezer with the purpose of complying with a specific music genre, cultural area, or mood, and that are among the most popular ones on the service. Playlists' cover images constitute the cards that can be recommended to users on the app homepage in a carousel, updated on a daily basis, with $L = 12$ available slots and $L_{\text{init}}=3$ cards initially displayed.

To determine which method would best succeed in making users click and stream the displayed playlists, extensive experiments were conducted in two steps. Firstly, \textit{offline} experiments simulating users' responses to carousel-based recommendations were run, on a simulation environment and on data that we both publicly release\footnote{\label{code} Data and code are available at: \href{https://github.com/deezer/carousel_bandits}{https://github.com/deezer/carousel\_bandits}} with our paper (see Section \ref{c11s32}). We believe that such industrial data and code release  will benefit the research community. Then, an \textit{online} large-scale A/B test was run on the Deezer app to validate the findings of offline experiments.

\subsection{A Simulation Environment and Dataset for Offline Evaluation}
\label{c11s32}

For offline experiments, we designed a simulated environment in Python based on 974 960 fully anonymized Deezer users. We release a dataset in which each user $u$ is described by a feature vector $x_u$ of dimension $d = 97$, computed internally by factorizing the interaction matrix between users and songs as described in \cite{hu2008collaborative} and then adding a bias term. A $k$-means clustering with $K = 100$ clusters was also performed to assign each user to a single cluster. In addition, for each user-playlist pair, we release a ``ground~truth'' display-to-stream probability $p_{ui} = \sigma(x_u^T \theta_i)$ where, as in the work of Chapelle~and~Li~\cite{chapelle2011empirical}, the $d$-dimensional vectors $\theta_i$ were estimated by fitting a logistic regression on a click data history from January 2020.

Simulations proceed as follows. At each round, a random subset of users (20 000, in the following) is presented to several sequential algorithms a.k.a. policies to be evaluated. These policies must then recommend an ordered set of $L = 12$ playlists to each user. Streams, i.e., positive binary rewards, are generated according to the aforementioned display-to-stream probabilities and to a configurable cascading browsing model capturing that users explore the carousel from left to right and might not see all recommended playlists. At the end of each round, all policies update their model based on the set of users and on binary rewards received from displayed playlists. Expected cumulative regrets of policies \cite{sutton2011reinforcement} w.r.t. the optimal top-$L$ playlists sets according to $p_{ui}$ probabilities are computed.

\subsection{List of Multi-Armed Bandit Algorithms}
\label{c11s33}

\def \randomAlgo{\textit{random}}
\def \exploreThenCommitAlgo[#1]{\textit{etc-seg#1}}
\def \exploreThenCommitExplore[#1]{\textit{etc-seg-explore#1}}
\def \exploreThenCommitExploit[#1]{\textit{etc-seg-exploit#1}}
\def \epsilonGreedyAlgo[#1]{\textit{$\epsilon$-greedy-seg#1}}
\def \epsilonGreedyExplore[#1]{\textit{$\epsilon$-greedy-seg-explore#1}}
\def \epsilonGreedyExploit[#1]{\textit{$\epsilon$-greedy-seg-exploit#1}}
\def \klucbAlgo{\textit{kl-ucb-seg}}
\def \TSAlgo[#1]{\textit{ts-seg#1}}
\def \LinearTSAlgo[#1]{\textit{ts-lin#1}}

In our experiments, we evaluate semi-personalized versions of several popular sequential decision making algorithms/policies, using the provided $K=100$ clusters, and compare their performances against fully-personalized methods. As detailed in Section \ref{c11s23}, users within a given cluster share parameters for all semi-personalized policies; they are the ones whose names end with \textit{-seg} in the following list. We consider the following methods:

\begin{itemize}
    \item \randomAlgo{}: a simple baseline that randomly recommends $L$ playlists to each user.
    \item \epsilonGreedyAlgo[]: recommends playlists randomly with probability $\epsilon$, otherwise recommends the top-$L$ with the highest mean observed rewards. Two versions, \epsilonGreedyExplore[] ($\epsilon$=0.1) and \epsilonGreedyExploit[] ($\epsilon$=0.01) are evaluated.
    \item \exploreThenCommitAlgo[]: an \textit{explore then commit} strategy, similar to \randomAlgo{} until all arms have been played $n$ times, then recommends the top-$L$ playlists. Two versions, \exploreThenCommitExplore[] ($n=100$) and \exploreThenCommitExploit[] ($n=20$) are evaluated.
    \item \klucbAlgo{}: the Upper Confidence Bound (UCB) strategy \cite{auer2002finite,chen2016combinatorial,lai1985asymptotically}, that tackles the exploration-exploitation trade-off by computing confidence intervals for the estimation of each arm probability, then selecting the $L$ arms with the highest upper confidence bounds. Here, we use KL-UCB bounds \cite{garivier2011kl}, tailored for Bernoulli rewards. 
    \item \TSAlgo[]: the Thompson Sampling (TS) strategy \cite{chapelle2011empirical,thompson1933likelihood}, in which estimated display-to-stream probabilities are samples drawn from Beta distributions \cite{thompson1933likelihood}, whose parameters are updated at each round in a Bayesian fashion, such that variance tends towards zero and expectation converges to empirical mean as more rewards are observed. Two versions, \textit{ts-seg-naive} (prior distributions are Beta$(1,1)$, i.e., Uniform$(0,1)$) and \textit{ts-seg-pessimistic} (priors are Beta$(1,99)$) are evaluated. As the UCB algorithm \cite{chen2016combinatorial}, TS is backed by strong theoretical guarantees \cite{komiyama2015optimal} on  speeds of expected cumulative regrets in the multi-armed bandit with multiple plays setting.
    \item \LinearTSAlgo[]: an extension of TS \cite{chapelle2011empirical} to the linear contextual framework from Section~\ref{c11s23}. We follow the method of Chapelle~and~Li~\cite{chapelle2011empirical} to learn $\theta_{i}$ vectors for each arm $i$ from Gaussian prior distributions. Two versions, \textit{ts-lin-naive} (0 means for all dimensions of the prior) and \textit{ts-lin-pessimistic} (-5 mean for the bias dimension prior) are evaluated.
\end{itemize}

By default, policies always abide by the cascade model introduced in Section \ref{c11s24}, meaning they do not update the parameters relative to recommended playlists that the cascade model labels as unseen. For comparison, we also implemented versions of these policies that do not abide by this behavior. In the following, they are labeled \textit{no-cascade}.
\section{Experimental Results}
\label{c11s4}

\subsection{Offline Evaluation}
\label{c11s41}

\begin{figure}[t]
\centering
  \includegraphics[width=1.0\linewidth]{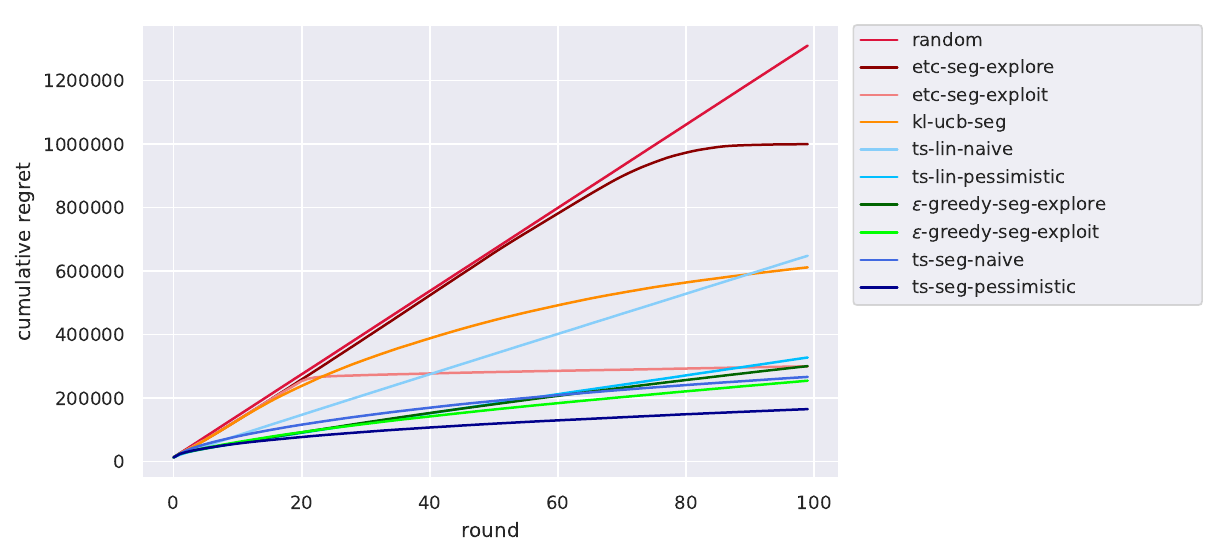}
  \caption[Offline evaluation: top-12 playlist recommendation]{Offline evaluation of top-12 playlist recommendation using our simulation environment: expected cumulative regrets of policies over 100 simulated rounds. The empirical gain of \textit{ts-seg-pessimistic} w.r.t. others is statistically significant at the 1\% level (p-value $<$ 0.01).}

  \label{fig:overall_results}
\end{figure}

\paragraph{Semi-Personalization vs Personalization} 
Figure \ref{fig:overall_results} provides cumulative regrets over 100 rounds for the different policies, recommending playlists via our offline environment.
Both \exploreThenCommitAlgo[-explore]{} and \exploreThenCommitAlgo[-exploit] behave as badly as \randomAlgo{} in the exploration phase, then, shortly after starting to exploit, they both reach competitive performances as illustrated by the brutal flattening of their cumulative regret curves, with \exploreThenCommitAlgo[-exploit] transitioning 50 rounds earlier. 
The latter strategy also outperforms \klucbAlgo{}, which shape suggests slow learning throughout the whole experiment. 
Moreover, both \LinearTSAlgo[-pessimistic] and \LinearTSAlgo[-naive] appear to stabilize to non-flat linear cumulative regret curves after only a few rounds. Pessimistic policies are overall more effective than their naive counterparts, which is due to their lower prior display-to-stream probabilities, which are more realistic.
Overall, several semi-personalized policies eventually outclassed fully-personalized alternatives, with \TSAlgo[-pessimistic] already outperforming them all at the end of the first 25 rounds\footnote{We point out that, in a recent work based on our data and simulation environment, Jeunen~and~Goethals~\cite{jeunen2021top} nonetheless managed to improve the performance of \textit{ts-lin-pessimistic} by adopting different hyperparameters.}. This method manages to effectively exploit information and to quickly \textit{rank} playlists, which is an interesting result, as fully-personalized contextual models were actually the only ones able to learn the \textit{exact} display-to-stream probabilities (see generative process in Section~\ref{c11s32}), and as both frameworks have comparable numbers of parameters ($N_{\text{arm}}\times K$ vs $N_{\text{arm}} \times d$). While fully-personalized methods have been the focus of previous works on carousel recommendation \cite{gruson2019offline,mcinerney2018explore}, our experiments emphasize the empirical benefit of semi-personalization via user clustering that, assuming good underlying clusters, might appear as a suitable alternative for such large-scale real-world applications.

\paragraph{Impact of Delayed Batch Feedback}
In our experiments, to be consistent with real-world constraints,  rewards are not processed on the fly but by batch, at the end of each round. We observe that, for semi-personalization, such setting tends to favor stochastic policies, such as the \textit{ts-seg} or \epsilonGreedyAlgo[] ones, w.r.t. deterministic ones such as \klucbAlgo{}. Indeed, as \klucbAlgo{} selects arms in a deterministic fashion, it always proposes the same playlists to all users of the same cluster until the round is over. On the contrary, stochastic policies propose different playlists sets within the same cluster, ensuring a wider exploration during the round, which might explain why \klucbAlgo{} underperforms in our experiments.

\begin{figure}[H]
\centering
  \includegraphics[width=1\linewidth]{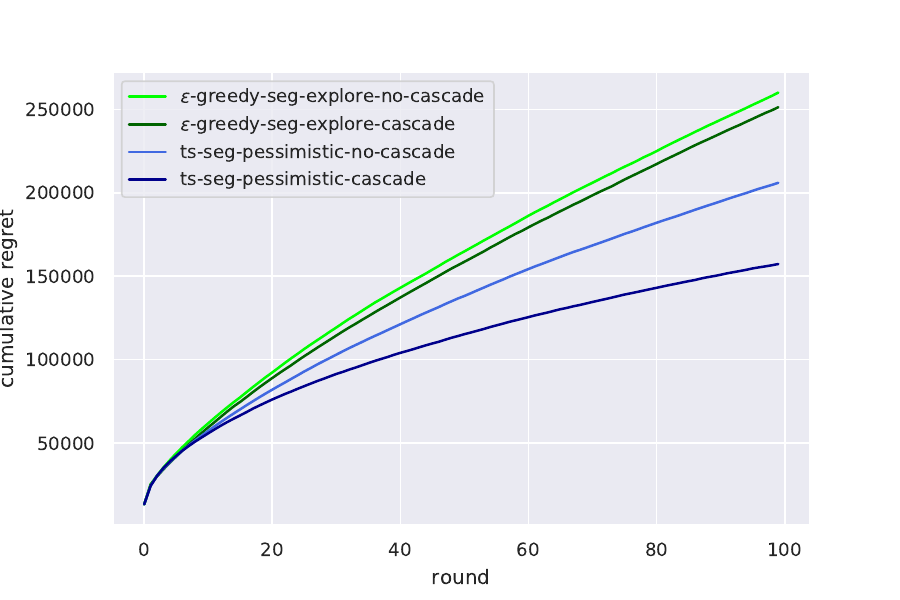}
  \caption[Offline evaluation: cascade vs no-cascade]{Offline evaluation using our simulation environment: comparison of cascade vs no-cascade, over 100 simulated rounds. Differences at final round are statistically significant at the 1\% level (p-value $<$ 0.01).}
  \label{fig:cascade_results}
\end{figure}

\begin{multicols}{2}
\paragraph{Cascade vs No-Cascade}
All policies from Figure~\ref{fig:overall_results} abide by the cascade model introduced in Section~\ref{c11s24}. In Figure~\ref{fig:cascade_results}, we report results from follow-up experiments. They measure the empirical benefit of taking into account this cascading behavior of users when browsing a sequence of playlists. We compared policies to alternatives that ignored the cascade model, and thus returned a 0 reward for all unstreamed playlists at each round, whatever their rank in the carousel. Only two policy pairs are displayed in Figure~\ref{fig:cascade_results}~for~brevity.
\columnbreak
\end{multicols}
\vspace{-1cm}
 For both of them, the \textit{no-cascade} variant is outperformed by policies integrating our proposed cascade-based update model from Section~\ref{c11s24}. This result validates the relevance of capturing such a phenomenon for our carousel-based personalization problem.

\subsection{Online Evaluation}
\label{c11s42}

\begin{figure}[H]
\centering
  \includegraphics[width=1\linewidth]{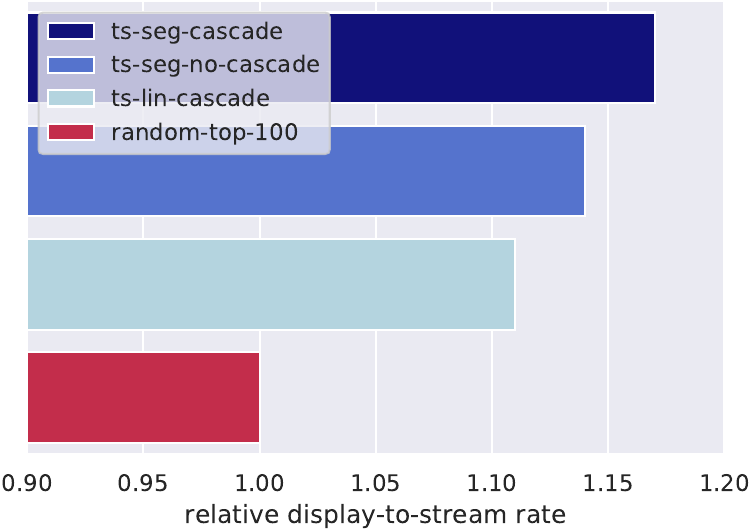}
\caption[Online A/B test on Deezer's carousels]{Online A/B test evaluation on Deezer: relative display-to-stream gains w.r.t. \textit{random-top-100} baseline (see Section \ref{c11s42}). Differences are statistically significant at the 1\% level (p-value $<$ 0.01).}\label{fig:online_results}
\end{figure}

\begin{multicols}{2}
An industrial-scale A/B test has been run in February 2020 on millions of users, to verify whether results from the simulations would hold on the actual Deezer mobile app. The 12 recommended playlists from each user's carousel were updated on a daily basis on the Deezer app. Due to industrial constraints, only a subset of policies, from (naive) TS, were tested in production. Also, for confidentiality reasons, we do not report the exact number of users involved in each cohort, nor the precise display-to-stream rates. Instead, results are expressed in Figure~\ref{fig:online_results} in relative terms.
\columnbreak

\end{multicols}
\vspace{-0.5cm}

Specifically, results are expressed in terms of relative display-to-stream rates gains w.r.t. \textit{random-top-100}, an internal baseline that randomly recommends 12 playlists from a subset of 100, pre-selected for each cluster from internal heuristics. 
Results confirm the superiority of the proposed multi-armed bandit framework for personalization, notably the semi-personalized strategy, and the empirical benefit of integrating a cascade model for arms updates, although users might actually have more complex behaviors on the platform.

\section{Conclusion}
\label{c11s5}

In this chapter, we modeled carousel personalization as a contextual multi-armed bandit problem with multiple plays. By addressing a challenging large-scale playlist recommendation task on Deezer, we highlighted the benefits of our framework, notably the integration of the cascade model and semi-personalization via user clustering. 

Along with the paper associated with this work~\cite{bendada2020carousel}, we publicly released a private dataset of user preferences for curated playlists on Deezer, and an open-source environment to recreate comparable learning problems. At the time of writing, this release already benefited research on carousel personalization. Several articles explicitly mentioned our study \cite{felicioni2021methodology,felicioni2021measuring,ferrari2021optimizing,lo2021page,yan2021two} and/or used our code and data \cite{claffeyevaluating,jeunen2021top} to address various other carousel-based problems.

Despite the promising results presented in this chapter, there is still room for improvement. In particular, we assumed that the number of users and cards was fixed throughout the rounds, which is a limit. In Chapter~\ref{chapter_12}, we will consider the \textit{user cold start} problem and explain how Deezer handles the arrival of new users on the service.

Moreover, future work could also benefit from the advances in GAE and VGAE models presented in previous chapters. For instance, our clustering of users, currently done through a standard $k$-means, could be improved by applying community detection techniques from Chapters~\ref{chapter_7}~and~\ref{chapter_9} to a graph of users (e.g., an artificial graph of ``similar'' users with a comparable listening history, or a social graph of users connected together through ``follow'' connections on the service). The $d$-dimensional vector characterizing each user in Section~\ref{c11s23} could also correspond to GAE-based or VGAE-based node embedding representations.

Lastly, in this chapter, we assumed that arms/cards distributions were fixed and independent, which might be unrealistic. A playlist's relative interest might depend on its neighbors in the carousel, and \textit{individually} selecting the top-$L$ playlists does not always lead to the best \textit{set} of $L$ playlists, e.g., in terms of musical diversity. Future research in this direction would definitely lead to the improvement of carousel personalization.


\chapter[A Semi-Personalized System for User Cold Start Recommendation]{A Semi-Personalized System for User~Cold~Start~Recommendation}\label{chapter_12}
\chaptermark{A Semi-Personalized System for User Cold Start Recommendation}

\textit{This chapter presents research conducted with Léa Briand, Walid Bendada, Mathieu Morlon, and Viet-Anh Tran, and published in the proceedings of the 27\up{th} ACM SIGKDD Conference on Knowledge Discovery and Data Mining (KDD 2021)~\cite{briand2021semi}.}

\section{Introduction}
\label{c12s1}

A prevalent approach to recommend personalized content on online services such as Deezer is \textit{collaborative filtering} which, broadly, consists in predicting the preferences of a user within a set of items by leveraging the known preferences of some similar users \cite{covington2016deep,koren2015advances,schedl2018current,su2009survey}. In particular, several recent works emphasized the empirical effectiveness of \textit{latent models} for collaborative filtering at addressing industrial-level challenges \cite{covington2016deep,gomez2015netflix,jacobson2016music,smith2017two}. Analogously to the node embedding methods developed throughout this thesis, these models aim to directly learn vector space representations, i.e., \textit{embeddings} of users and items where proximity should reflect user preferences, typically via the factorization of a user-item interaction matrix \cite{he2016fast,koren2015advances,koren2009matrix} or with neural networks architectures processing usage data~\cite{covington2016deep,mongia2020deep,wang2015collaborative}.

However, the performances of these models tend to significantly degrade for new users who only had few interactions with the catalog \cite{cao2020improving,lika2014facing,schedl2018current}. They might even become unsuitable for users with no interaction at all, who are absent from user-item matrices in standard algorithms \cite{bobadilla2012collaborative,gope2017survey,lee2019melu}. This is commonly referred to as the \textit{user cold start} problem \cite{bobadilla2012collaborative,lee2019melu,lika2014facing,schedl2018current,volkovs2017dropoutnet}. Yet, recommending relevant content to these new users is crucial for online services. Indeed, a new user facing low-quality recommendations might have a bad first impression and decide to stop~using~the~service. 

In this last chapter, we present the system recently deployed on Deezer to address this problem. The solution starts from an existing large-scale latent model for collaborative filtering, periodically trained on Deezer's \textit{warm} users (see Section \ref{c12s3}). It automatically integrates \textit{cold} users into the existing embedding space, by collecting heterogeneous sources of demographic and interaction information on these users at registration day, processed by a deep neural network, and by leveraging a segmentation of warm users to strengthen the final representations and provide semi-personalized recommendations to cold users \textit{by the end of their registration day}.

The proposed system is suitable for an online production use on a large-scale app such as Deezer. Throughout this chapter, we show its practical impact and its empirical effectiveness at predicting the future musical preferences of cold users, through both offline experiments on data extracted from Deezer and an online A/B test on \textit{carousels} from the previous Chapter~\ref{chapter_11}. We also emphasize how this system enables us to provide more interpretable music recommendations. Along with the paper associated with this work~\cite{briand2021semi}, we publicly released our source code as well as anonymized usage data of Deezer users from our offline experiments.

This chapter is organized as follows. In Section \ref{c12s2}, we introduce the user cold start problem more precisely and mention previous research efforts on this topic. In Section \ref{c12s3}, we present our semi-personalized recommender system. We report and discuss our experimental setting, our data, and our results in Sections~\ref{c12s4}~and~\ref{c12s5}, and we conclude in Section~\ref{c12s6}.

\section{User~Cold~Start~Recommendation~on~Music~Streaming~Apps}
\label{c12s2}

In this section, we provide a precise formulation of the problem we aim to address. We also give an overview of the existing related work. Some of the mentioned approaches will constitute relevant baselines to evaluate the effectiveness of our system.

\subsection{Problem Formulation}
\label{c12s21}

Throughout this paper, we consider a catalog of $m$ music tracks available on Deezer. We assume that the catalog remains fixed over time, which we later discuss. At time~$t$, Deezer gathers $n_t$ \textit{warm} users who, according to some criteria internally fixed by our data scientists, had a sufficiently large number of interactions with the catalog, e.g., enough listening sessions, to be used in the training of our recommender systems. We consider a \textit{latent model for collaborative filtering} \cite{covington2016deep,koren2015advances,koren2009matrix,mongia2020deep}. From the observed  warm user-track interactions and following the processes described in Section~\ref{c12s3}, this model learns a vector space representation of both users and tracks. In this \textit{embedding} space, each user $i$ and track $j$ are represented by $d$-dimensional vectors (with $d \ll m$ and $d \ll n_t$), say $u_i \in \mathbb{R}^d$ and $v_j \in \mathbb{R}^d$, capturing musical preferences. To recommend relevant new tracks to the user~$i$, we leverage user-item similarity measures $f(u_i,v_j)$ in this space, encoding user-item affinities. These measures are typically based on an inner-product or a cosine similarity \cite{koren2009matrix}.  The model is updated at regular time intervals to take into account the evolution of preferences.

Everyday, new users, referred to as \textit{cold} (analogously to \textit{cold artists} from Chapter~\ref{chapter_8}), will register to the service. They will only have few to no interactions with the catalog during their registration day. As explained in the introduction, a straightforward inclusion of these cold users in the aforementioned latent model is unsuitable \cite{bobadilla2012collaborative,gope2017survey,lee2019melu,lika2014facing}. Waiting for them to become warm users, according to internal criteria, is also undesirable: indeed, recommending relevant content as soon as possible is crucial, as new users facing low-quality recommendations might make up their mind on this first impression and quickly stop using the service. 

As a consequence, we aim to address the following problem: \textit{given an existing latent model for collaborative filtering learning an embedding space from a set of warm users, how can we effectively include new cold users into this same space, by the end of their registration day on Deezer?}
In this chapter, we will evaluate the estimated embedding vectors of cold users by assessing their ability at predicting the future musical preferences of these users on Deezer after their registration day, through the evaluation tasks and metrics presented in~Section~\ref{c12s4}.

\subsection{Related Work}
\label{c12s22}
The user cold start problem has initiated significant research efforts over the past decade. In the following, we provide an overview of the most relevant work w.r.t. our approach. We refer the interested reader to some recent surveys \cite{gope2017survey,mu2018survey,zhang2019deep} for a more exhaustive review of the existing literature, and to \cite{mu2018survey,smith2017two,van2013deep,wang2018billion}  as well as Chapter~\ref{chapter_8} for a presentation of the related \textit{item cold start} problem, which will be out of our scope in most of this Chapter~\ref{chapter_12}.

A prevalent strategy to address the user cold start problem in the total absence of usage data consists in relying on metadata related to new users, and notably on demographic information (such as the age or country of the user) collected during registration \cite{fernandez2016alleviating,lam2008addressing,lika2014facing,mu2018survey,yanxiang2013user}. In particular, various approaches aim to cluster warm users, and subsequently assign cold users to existing clusters by leveraging these metadata \cite{cao2020improving,felicio2017multi,lika2014facing,mu2018survey,su2009survey,yanxiang2013user}. Building upon these works, the model we present in Section~\ref{c12s3} will also leverage demographic information and incorporate a clustering component.

Besides, one can enrich such systems by explicitly asking new users to rate items from the catalog through interview processes, leading to hybrid models based on preferences and side information \cite{gope2017survey,mu2018survey,shi2017local}. On the industry side, Netflix \cite{gomez2015netflix} and Spotify~\cite{jacobson2016music} are famous examples of services implementing such \textit{onboarding} session for new users. As explained in Section~\ref{c12s3}, Deezer also adopted this strategy. As the inclusion of an onboarding is not always possible in production, Fel{\'\i}cio~et~al.~\cite{felicio2017multi} propose to use bandit algorithms to assign cold users to warm user segments, while other studies \cite{lin2013addressing,shapira2013facebook} resort to social media data to connect~similar~users.

Sometimes, cold users do have a few interactions with the catalog on their registration day. In this case, exploiting such usage signal, in addition to side information, can significantly improve recommendations \cite{bobadilla2012collaborative,he2017neural,schedl2018current,su2009survey}. In particular, several recent works emphasized the effectiveness of \textit{deep learning} models at dealing with such heterogeneous settings \cite{bharadhwaj2019meta,bobadilla2012collaborative,covington2016deep,lee2019melu,mu2018survey,volkovs2017dropoutnet,zhang2019deep}. 
Notably, Covington~et~al.~\cite{covington2016deep} explain how a deep neural network, processing various user-item interactions (including watched videos, search queries...) and demographic information to learn embedding vectors, improved the YouTube recommender system. To represent interactions, they average various latent representations of watched/searched items from the same user session, allowing features to have the same dimension for each user. In Section~\ref{c12s3}, we will draw inspiration from their approach to pre-process the user features serving as input to our own model. However, a direct comparison to \cite{covington2016deep} will be impossible, as no complete description nor implementation of the YouTube recommender system was made publicly available. Several other deep learning approaches were already mentioned in Chapter~\ref{chapter_8}. Here, we provide more details on two of these methods that will be used in our experiments:

\begin{itemize}
    \item DropoutNet \cite{volkovs2017dropoutnet} emerged as one of the most powerful latent collaborative filtering models, addressing cold start while preserving performances for warm users. This neural network takes into account usage and content data, and explicitly simulates the cold start situation during training by applying \textit{dropout} \cite{srivastava2014dropout}, alternatively to user and item embedding layers. DropoutNet relies on the assumption that data is \textit{missing at random}, with the risk of introducing biased predictions \cite{little2019statistical}. Also, it equally considers different types of positive feedback; in Section~\ref{c12s3}, we will furthermore consider negative feedback.
Besides, both warm and cold users' embeddings are learned during training, whereas our system will directly incorporate cold users in an existing and fixed embedding space of warm users;
\item recently, \textit{meta-learning} methods for cold start have also been proposed \cite{bharadhwaj2019meta,lee2019melu,zhang2019deep}. Notably, optimization-based algorithms consider each user as a learning task. From a set of global parameters ensuring an initialization of the recommender system, other local parameters are progressively updated while the user interacts with items to capture his/her preferences. In particular, Lee~et~al.~\cite{lee2019melu} introduce MeLU (for \textit{Meta-Learned User preference estimator}, already mentioned in Chapter~\ref{chapter_8}), a neural network architecture following such a meta-learning paradigm and learning preferences from the concatenation of user and item information. When new users interact with some items, then \textit{local} parameters of the neural network are updated to refine predictions for these users. In the absence of usage data, users will still be associated with an embedding vector and receive a recommended list of items, thanks to \textit{global} updates of all~layers~of~MeLU.

\end{itemize}

\section{A Semi-Personalized System to Address User Cold Start}
\label{c12s3}

\begin{figure}[t]
  \centering
  \includegraphics[width=1.0\linewidth]{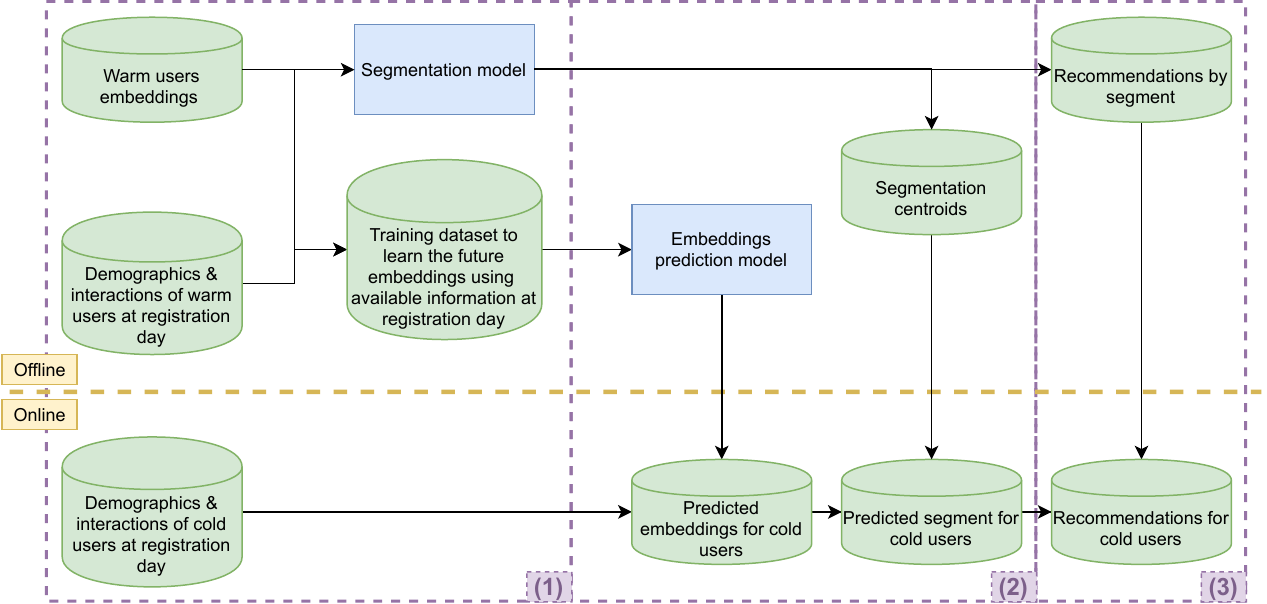}
  \caption[A semi-personalized system for user cold start recommendation]{The semi-personalized user cold start recommendation framework available for online requests in our production environment, and described throughout Section~\ref{c12s3}. (\textit{1})~Demographics and user-item interactions of a new user are concatenated, as described in Section \ref{c12s32}, to (\textit{2}) predict a user embedding vector from warm user embeddings described in Section~\ref{c12s31}. From the estimated user preferences, the new user is assigned to a segment of warm users. (\textit{3}) Combining the online predicted segment with the pre-computed top items by segment, cold users benefit from semi-personalized recommendations.}
\label{figOurProductionEnv}
\end{figure}

In this section, we present the system deployed in 2020 on Deezer to address the user cold start problem, as formulated in Section~\ref{c12s21}. The architecture of our framework is summarized in Figure \ref{figOurProductionEnv}, and discussed thereafter.

\subsection{Representing Musical
Preferences of Warm Users}
\label{c12s31}

We recall that our objective is to effectively incorporate, \textit{by the end of their registration day}, a set of cold users in an existing embedding space trained on warm users. In the following, we introduce two different strategies to learn such a space on Deezer's data. Although some technical details on computations are omitted for confidentiality reasons, some embedding vectors from both models will be released with this paper (see Section~\ref{c12s4}). Moreover, experimental results on both spaces will be reported in the next section.

\paragraph{UT-ALS Embeddings}
Latent models for collaborative filtering can approximate a preference matrix between users and items from the product of two low-rank matrices, respectively stacking up latent vector representations, a.k.a. embedding vectors, of users and items \cite{koren2015advances,su2009survey}. At Deezer, we consider a \textit{user-track (UT) interaction matrix} summarizing interactions between millions of active users and music tracks from the catalog. The \textit{affinity score} between user $i$ and music track $j$, i.e., the entry $(i,j)$ of the matrix, is computed from various signals, including the number of streams and the potential addition of the music track (or the corresponding album or artist) to a playlist of favorites.
The final entry is refined in accordance with internal heuristic rules. Then, we rely on a \textit{weighted matrix factorization}, specifically by using the \textit{alternating least squares} (ALS) method \cite{koren2009matrix}, to map both users and music tracks to a joint latent space of dimension $d = 256$. These vector representations will be referred to as \textit{UT-ALS embeddings} in the remainder of this paper.

\paragraph{TT-SVD Embeddings}

Models inspired by word2vec \cite{mikolov2018advances} rely on the \textit{distributional hypothesis} \cite{pennington2014glove} to map items co-occurring in similar contexts to geometrically close embedding vectors. Levy and Goldberg~\cite{levy2014neural} show that word2vec with negative sampling implicitly factorizes a shifted \textit{pointwise mutual information} (PMI) matrix using \textit{singular value decomposition} (SVD)~\cite{koren2009matrix}. In this paper, we also consider a PMI matrix, based on the co-occurrences of music tracks in diverse music collections on Deezer, such as music playlists. Then, we factorize this \textit{track-track (TT) matrix} using a distributed implementation of SVD\footnote{\href{https://github.com/criteo/Spark-RSVD}{https://github.com/criteo/Spark-RSVD}}, leading to embedding vectors of dimension $d = 128$ for each music track\footnote{Embedding dimensions of UT-ALS and TT-SVD have been optimized independently for recommendation, and are therefore different (256 vs 128). The choice of SVD vs ALS factorization is also driven by internal optimizations on Deezer's data. In experiments, we will simply exploit these vectors, independently, for cold start.}. Finally, we derive embedding vectors for warm users, by averaging music track vectors over their listening history on Deezer. These vector representations, different from \textit{UT-ALS embeddings}, will be referred to as \textit{TT-SVD embeddings} in the remainder of this paper.

\paragraph{Warm User Segmentation}
On top of UT-ALS or TT-SVD user embeddings, our system also computes a segmentation of warm users, by running a $k$-means algorithm, with $k =$ 1 000 clusters/segments, in the embedding space. Each user segment is represented by its \textit{centroid}, i.e., by the average of its user embedding vectors. In production, a list of the most popular music items to recommend among each warm user segment is also pre-computed.

\subsection{Predicting the Preferences of Cold Users}
\label{c12s32}

\begin{figure}[t]
  \centering
  \includegraphics[width=1.\linewidth]{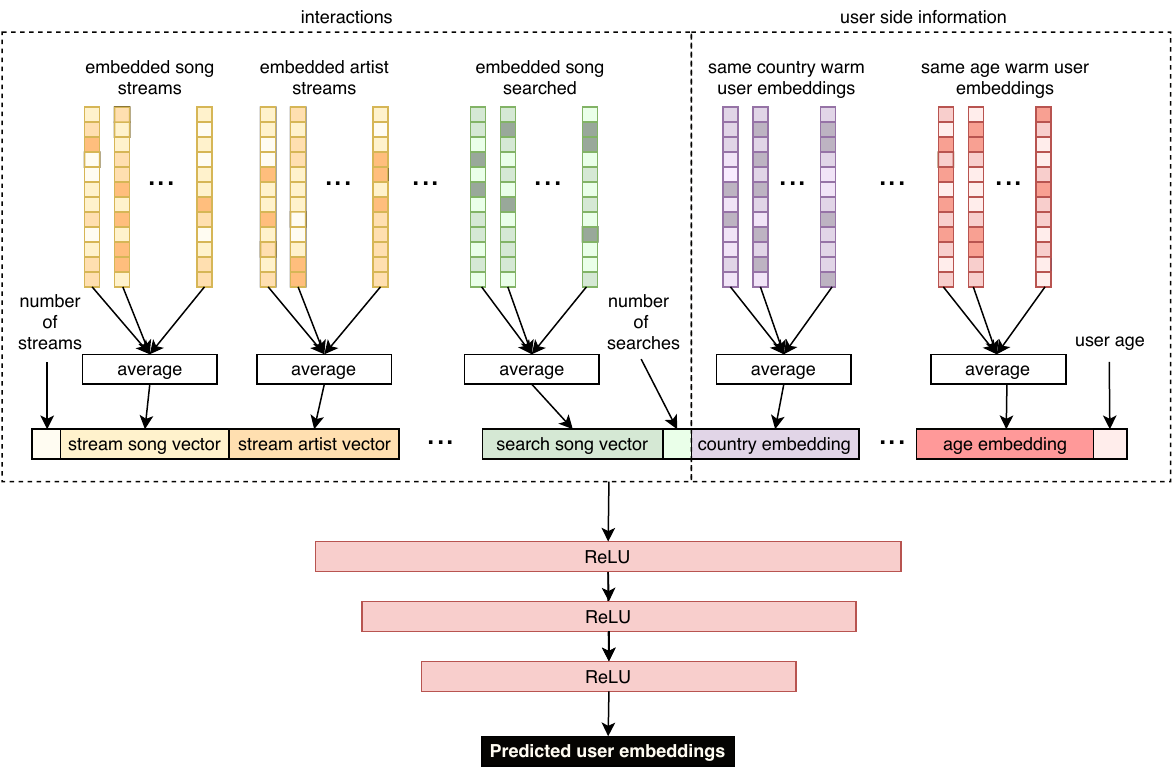}
  \caption[Prediction of user embeddings from heterogeneous data]{Prediction of user embeddings from heterogeneous data. \textit{Top left}: embedding vectors of activated items in usage data, including onboarding (Figure~\ref{onboarding}), are aggregated as described in Section~\ref{c12s32}. \textit{Top right}: we enrich representations with demographic information. \textit{Bottom}: after pre-processing the dense input features vector, a deep neural network model, trained as in Section~\ref{c12s32}, predicts user embedding vectors in either the UT-ALS embedding space or the TT-SVD embedding space from Section~\ref{c12s31}.}
  \label{embeddingsPredictionFigure}
\end{figure}

In the following we present our model, illustrated in Figure~\ref{embeddingsPredictionFigure}, to integrate cold users into these embedding spaces, and subsequently predict their future musical preferences on Deezer.

\paragraph{Overall Strategy}
Firstly, we gather data from various sources, presented in the next paragraph and referred to as \textit{input features}. They can be collected for warm users and (at least partially for) cold users. Then, we train a neural network (see the ``Model Training'' paragraph thereafter) to map input features of warm users to their (either UT-ALS or TT-SVD) embedding vectors. Last, through a forward pass on this trained neural network, we predict embedding vectors for cold users from their input features. Cold users are therefore integrated into the existing latent space alongside warm users and each track of the catalog. This will permit computing cold user-track similarities, and even similarities between cold and warm users, which we leverage for clustering (see the ``Semi-Personalization'' paragraph thereafter).

\paragraph{Input features}
During registration on Deezer, all users specify their age and country of origin, which we include in input features. This information is enriched with \textit{country embedding} and \textit{age embedding} vectors which, as illustrated in Figure \ref{embeddingsPredictionFigure}, are the average of embedding vectors of warm users from respectively the same country and age class. 
As hybrid models mentioned in Section~\ref{c12s22}, we complement this side information with data retrieved from user-item interactions, limiting to interactions at registration day (if any). These interactions include  positive or negative) \textit{explicit} and \textit{implicit} signals, including streaming activity, searches, skips and likes. As the granularity of items is crucial in music~\cite{schedl2018current}, we also compute such signals at the album, artist, and playlist levels, deriving embedding vectors for such music entities by averaging the relevant track embeddings (e.g., by averaging the tracks of an album, or part of the discography of an artist).

\begin{figure}[H]
  \centering
  \includegraphics[width=0.7\linewidth]{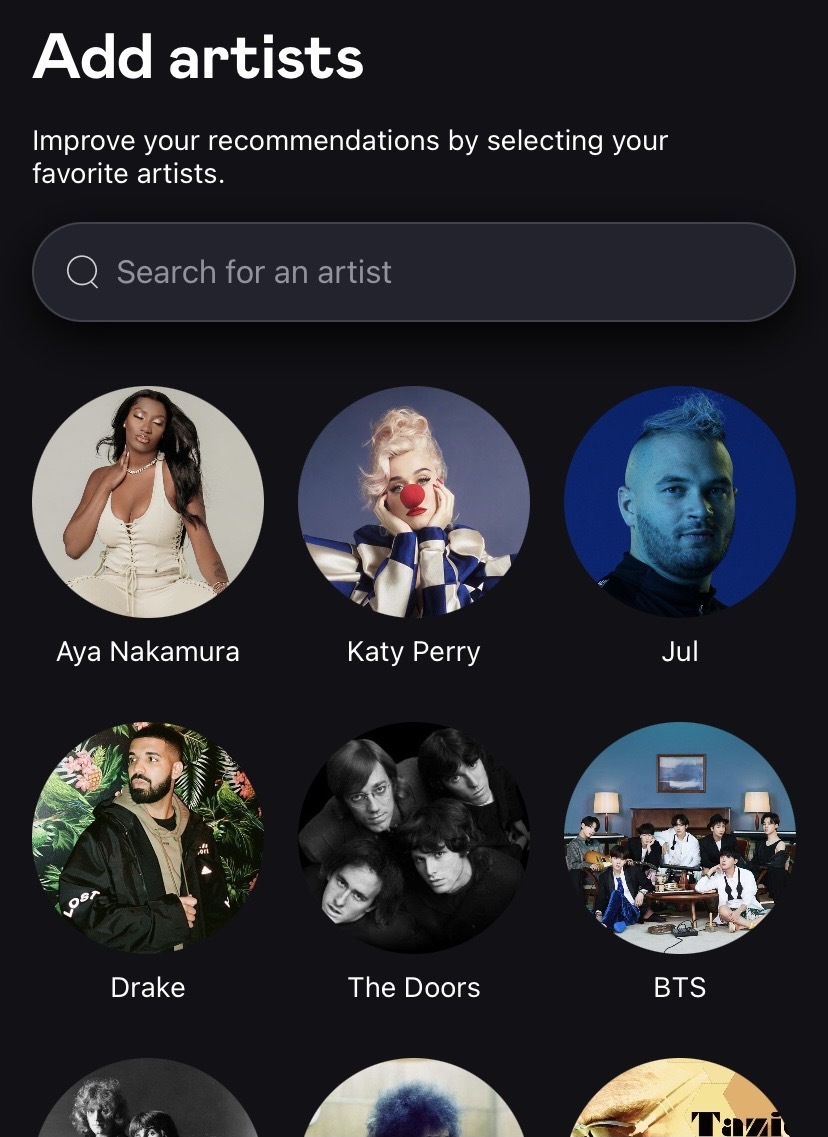}
  \caption[Overview of the onboarding process]{Overview of the \textit{onboarding} process, displayed to new users on the Deezer service.}
\label{onboarding}
\end{figure}

\begin{multicols}{2}
Then, for each type of interaction and music entity, embedding vectors are averaged, as illustrated in Figure~\ref{embeddingsPredictionFigure}. We obtain \textit{fixed-size representations} for both demographics and interactions, i.e., independent of the number of modalities or interactions, which is crucial for scalability in a production environment. 
For some new users, some (or possibly all) types of interactions at registration day may be missing; the corresponding representations are replaced arbitrarily by null vectors. To avoid such a situation, Deezer proposes an \textit{onboarding} process for newly registered users, proposing them to add artists from various music genres to their list of favorites, as illustrated in Figure~\ref{onboarding}.
\columnbreak
\end{multicols}
\vspace{-0.5cm}

\paragraph{Model Training} 
Fixed-size representations of demographics and interactions are concatenated to form a unique dense input vector of dimension 5139 (when considering UT-ALS embeddings) or 2579 (TT-SVD embeddings). It constitutes the input layer of a feedforward neural network, with three hidden layers of dimensions 400, 300, and 200, respectively, and an output layer of dimension $d = 256$ (when considering UT-ALS embeddings) or $128$ (TT-SVD embedding). We use ReLU activations at each layer except the output, followed by \textit{batch normalization} \cite{goodfellow2016deep}. We train the model on warm users, by iteratively minimizing the \textit{mean squared error} between the predicted user embeddings and their actual value in UT-ALS or TT-SVD spaces, by \textit{stochastic gradient descent} \cite{goodfellow2016deep} with a learning rate of 0.001, batch sizes of 512, and 100 (respectively 130) epochs for TT-SVD (resp. UT-ALS). 

\paragraph{Semi-Personalization}
Our model integrates cold users into the existing embedding space alongside warm users and music tracks. Therefore, one could provide \textit{fully personalized} music recommendation to each of those cold users, by retrieving the most \textit{similar} tracks for each user via an exact or an approximate \textit{nearest neighbors} search and some similarity measure. However, as we will empirically show in our experiments, such a strategy can still lead to noisy results for users with very few to no usage data. As a consequence, our system instead adopts a \textit{semi-personalized} recommendation strategy. On top of our neural network predictions, we include cold users into the pre-computed \textit{warm user segmentation} previously described.
Specifically, each cold user is assigned to the warm cluster whose centroid is the closest w.r.t. the predicted embedding vector of this cold user. We subsequently recommend the pre-computed most popular tracks among warm users from the cluster. Our framework is summarized in Figure \ref{figOurProductionEnv}.

\paragraph{Model Deployment}
\label{c12s33}
This system is suitable for online production use. At Deezer, the real-time inference service to predict user embeddings is a Golang web server, deployed in a Kubernetes cluster. The web service wraps the onnxruntime library\footnote{\href{https://www.onnxruntime.ai/}{https://www.onnxruntime.ai/}}, a fast engine for running ONNX machine learning models. It permits fast predictions of cold users' embeddings via forward passes on already trained neural networks. Models are trained offline using PyTorch, on an NVIDIA GTX 1080 GPU and an Intel Xeon Gold 6134 CPU, and then exported to ONNX format and stored on Hadoop. Embeddings of music tracks (from which we also derive embeddings of artists, albums, or playlists), as well as warm segment centroids, are exported weekly in tables in a Cassandra cluster, exposed via a JSON REST service. Embeddings and serialized models are weekly updated to take into account changes in the catalog and preferences, and weekly~exported~as~well. 

\section{Application to Offline Prediction of Listening Data on Deezer}
\label{c12s4}

In the remainder of this chapter, we evaluate the performance and impact of our system on our data. Firstly, in this section, we focus on offline experiments.

\subsection{Experimental Setting: Predicting Future Preferences of Cold Users}
\label{c12s41}

Our system permits recommending musical content to cold users. Through experiments on an offline dataset of Deezer active users, described thereafter, we evaluate to which extent the proposed recommendations at registration day would have matched the actual musical preferences of a set of users on their first month on the service. Specifically, we compute the 50 most relevant music tracks for each user of the dataset, from our model and registration day's input features (described in the previous section). 
We compare them to the tracks listened to by each user during their next 30 days on Deezer, using three standard recommendation metrics: the \textit{Precision}, the \textit{Recall}, as well as the \textit{Normalized Discounted Cumulative Gain} (NDCG) as a measure of ranking quality~\cite{schedl2018current}. 

\subsection{Dataset and List of Models}
\label{c12s42}

\paragraph{Dataset} 
For offline experiments, we extracted a dataset of 100 000 fully anonymized Deezer users. Among them, 70 000 are \textit{warm} users. They are associated with demographic information (country and self-reported age), as well as their respective UT-ALS and TT-SVD embedding vectors. These vectors correspond to those actually computed by our latent collaborative filtering models on the Deezer production system on November 1\up{st}, 2020, from millions of active warm users. Our dataset also includes the UT-ALS and TT-SVD embedding representations of the 50 000 most popular anonymized music tracks on Deezer. 

The remaining 30 000 users are \textit{cold} users, who registered on Deezer on the first week of November 2020, and subsequently listened to at least 50 music tracks \textit{on their first month on the service (excluding registration day)}. They are split into a validation set and a test set of respectively 20~000 and 10~000 users.
For each cold user, we collected demographic information as well as the list of artists selected during the onboarding, and the lists of available streams, skips, bans, searches, additions to favorites relative to music tracks, artists, albums, and playlists \textit{at registration day only}. 77\% of cold users from this dataset streamed at least once at registration day, whereas 95\% of cold users fulfilled one of the aforementioned interactions. Thus, for the remaining 5\%, only demographic information is available at registration day. Lastly, the dataset includes the tracks listened to by cold users during their next 30 days on Deezer, among the 50~000~tracks.

Along with our paper~\cite{briand2021semi}, we publicly released\footnote{Data and code are available on: \href{https://github.com/deezer/semi_perso_user_cold_start }{https://github.com/deezer/semi\_perso\_user\_cold\_start}} this dataset, as well as the code corresponding to our offline experiments. Besides making our results reproducible, it publicly provides a relevant real-world benchmark dataset to evaluate and compare future cold start models on actual (albeit anonymized) usage data. We, therefore, hope that this open-source release of industrial resources will favor and benefit future research and applications on user cold start problems.

\paragraph{Models} 

We report results from two versions of our system (one trained on the UT-ALS embeddings of the 70 000 warm users, and one trained on their TT-SVD embeddings) on the task presented in Section~\ref{c12s31}. For each embedding space, we simultaneously evaluate:
\begin{itemize}
    \item the \textit{semi-personalized} recommendations, which are actually used in production at Deezer. In this case, the 50 recommended tracks of each user will correspond to the 50 most popular tracks of his/her \textit{user segment}, as detailed in \ref{c12s32};
    \item the \textit{fully-personalized} ones, which directly leverage the predicted embedding vectors of each cold user from the neural network. In this case, we recommend, for each cold user, the 50 \textit{nearest neighbors} music tracks w.r.t. his/her vector in the embedding space, according to a cosine similarity.
\end{itemize}

Moreover, although the main objective of the applied paper associated with this work~\cite{briand2021semi} was to show the very practical impact of a deployed system and not to chase the state of the art, we also report the performances of several baseline methods as a point of comparison. Foremost, we consider a \textit{Popularity} baseline. This chart-based method recommends the most popular songs to cold users. Besides, to motivate the need for a careful modeling of user cold start, we consider the \textit{Registration Day Streams} method, which includes cold users in the embedding in a more straightforward way. It estimates a cold user's embedding vector by averaging embedding vectors of music tracks listened to at registration day (and relying on popularity in the absence of any stream), then recommends 50 tracks to each cold user via a nearest neighbors search. Moreover, a third ablation baseline, denoted \textit{Input Features Clustering} in the following, will consist in
getting rid of our neural network model and directly rely on the \textit{input features} of Section~\ref{c12s32}. This method will also cluster all users into segments via a $k$-means, but from their stacked input features, i.e., the large vector illustrated in Figure \ref{embeddingsPredictionFigure}. Recommended tracks will correspond to the 50 most popular among warm users of each cluster; as users are no longer in the same space as tracks, we do not evaluate any fully-personalized version of this baseline.

Lastly, we evaluate DropoutNet \cite{volkovs2017dropoutnet} and MeLU \cite{lee2019melu}, two deep learning models described in Section~\ref{c12s22}. We selected these two methods as they are simultaneously \textit{1)} among the most promising approaches, to the best of our knowledge, \textit{2)} scalable to large datasets, and \textit{3)}~publicly available online\footnote{Public implementations are available on \href{https://github.com/layer6ai-labs/DropoutNet}{https://github.com/layer6ai-labs/DropoutNet} and \href{https://github.com/hoyeoplee/MeLU}{https://github.com/hoyeoplee/MeLU} respectively.}. They process the same input features as our model, with the notable exception that DropoutNet only processes positive user-track interactions (i.e., not skips nor bans, representing approximately 12\% of all interactions in our dataset). We carefully tuned each model using the validation set. For each model, we simultaneously evaluate \textit{fully-personalized} recommendations where, as for our system, we recommend to each cold user his/her 50 most similar music tracks, as well as \textit{semi-personalized} recommendations leveraging a warm user segmentation similar to ours. We consider two variants of each model from the last two paragraphs, respectively trained on UT-ALS and TT-SVD embeddings.

\subsection{Offline Evaluation}
\label{c12s43}

\begin{table}[t]

\centering
\caption[Offline prediction of the future musical preferences of cold users]{Offline prediction of the future musical preferences of cold users on Deezer.}
\resizebox{1.0\textwidth}{!}{
\begin{tabular}{c|ccc|ccc}
\toprule
\textbf{Model} & \multicolumn{3}{c}{\textbf{TT-SVD Embeddings}} & \multicolumn{3}{c}{\textbf{UT-ALS Embeddings}} \\
&  \footnotesize \textbf{Precision@50 (in \%)} & \footnotesize \textbf{Recall@50 (in \%)} & \footnotesize \textbf{NDCG@50 (in \%)} & \footnotesize \textbf{Precision@50 (in \%)} & \footnotesize \textbf{Recall@50 (in \%)} & \footnotesize \textbf{NDCG@50 (in \%)}\\
\midrule
\midrule 
Popularity & 8.92 $\pm$ 0.21 & 3.01 $\pm$ 0.08 & 9.72 $\pm$ 0.20 & 8.92 $\pm$ 0.21 & 3.01 $\pm$ 0.08 & 9.72 $\pm$ 0.20 \\ 
Registration Day Streams & 9.30 $\pm$ 0.22 & 3.48 $\pm$ 0.06 & 9.73 $\pm$ 0.23 & 
16.88 $\pm$ 0.46 & 5.99 $\pm$ 0.11 & 17.72 $\pm$ 0.43 \\ 
Input Features Clustering & 8.84 $\pm$ 0.22 & 2.97 $\pm$ 0.08 & 9.75 $\pm$ 0.23 & 8.85 $\pm$ 0.22 & 2.98 $\pm$ 0.08 & 9.75 $\pm$ 0.22 \\ 
DropoutNet Full-Pers. & 10.04 $\pm$ 0.27 & 3.75 $\pm$ 0.11 & 10.46 $\pm$ 0.29 & 16.30 $\pm$ 0.50 & 5.77 $\pm$ 0.50 & 17.62 $\pm$ 0.54 \\ 
DropoutNet Semi-Pers. & 20.85 $\pm$ 0.35 & 7.55 $\pm$ 0.12 & 22.61 $\pm$ 0.36 & \textbf{19.83 $\pm$ 0.32} & 6.93 $\pm$ 0.16 & \textbf{21.55 $\pm$ 0.44}  \\
MeLU Full-Pers. & 15.00 $\pm$ 0.40 & 5.12 $\pm$ 0.17 & 16.79 $\pm$ 0.45 & 13.92 $\pm$ 0.36 & 4.71 $\pm$ 0.12 & 15.49 $\pm$ 0.39 \\ 
MeLU Semi-Pers. & 19.66 $\pm$ 0.36 & 6.87 $\pm$ 0.15 & 21.63 $\pm$ 0.40 & 19.35 $\pm$ 0.43 & 6.71 $\pm$ 0.14 & 21.33 $\pm$ 0.45 \\ 
\midrule%
\textbf{Deezer Full-Pers. (ours)} & 9.58 $\pm$ 0.18 & 3.53 $\pm$ 0.03 & 9.77 $\pm$ 0.17 & 18.50 $\pm$ 0.43 & 6.63 $\pm$ 0.10 & 20.22 $\pm$ 0.41 \\ \textbf{Deezer Semi-Pers. (ours)} & \textbf{22.75 $\pm$ 0.32} & \textbf{8.26 $\pm$ 0.15} & \textbf{24.59 $\pm$ 0.30} 

& 19.00 $\pm$ 0.42 & \textbf{6.93 $\pm$ 0.10} & 20.38 $\pm$ 0.45  \\ 
\bottomrule
\end{tabular}
\label{offlineevaluation}
}
\end{table}

Table \ref{offlineevaluation} reports performance scores of all models on the UT-ALS and TT-SVD embeddings, along with standard deviations over ten iterations. As the \textit{Popularity} baseline is independent of embedding vectors, it obtains the same scores for the two embeddings. In both settings, \textit{Popularity} is the worst method, although performances are still fairly good for such a~simple~strategy.

We first focus on TT-SVD embeddings. The \textit{Registration Day Streams} baseline hardly beats \textit{Popularity} for these embeddings, which emphasizes the limits of a direct use of sparse usage data in cold start settings. On the contrary, our proposed semi-personalized system provides significant improvements (e.g., a +14.86 NDCG points increase w.r.t. \textit{Registration Day Streams}), and even reaches competitive results w.r.t. DropoutNet and MeLU. Overall, semi-personalized methods outperform their fully-personalized variants (e.g., a +13.17 precision points increase for \textit{Deezer Semi-Pers.} vs \textit{Deezer Full-Pers.}). This confirms the empirical relevance of our user segmentation strategy, and that, while the studied methods \textit{could} provide fully personalized recommendations to cold users, this strategy can lead to noisier results on real-world applications such as ours. Lastly, our system provides better results than a direct use of the \textit{input features} vector (\textit{Input Features Clustering} baseline), which confirms the relevance of our modeling step on top of this vector. We considered replacing our 3-layer neural network with alternative architectures between \textit{Input Features Clustering} and \textit{Deezer Semi-Pers.}, notably with a 1-layer neural network, i.e., a simpler linear regression model, but did not reach comparable results.

Regarding the UT-ALS embeddings, most conclusions are consistent w.r.t. TT-SVD. \textit{Deezer Semi-Pers.} reaches quite comparable or better results w.r.t. alternatives. Among the key differences, we highlight that, while \textit{Deezer, DropoutNet and MeLU Semi-Pers.} are overperforming, the fully-personalized \textit{Registration Day Streams}, \textit{DropoutNet Full-Pers.} and \textit{Deezer Full-Pers.} obtain significantly stronger results than on TT-SVD embeddings (e.g., a 18.50\% precision score for \textit{Deezer Full-Pers.} on UT-ALS, vs 9.58\% on TT-SVD). UT-ALS embeddings are constructed from a user-item interaction matrix, an approach appearing as better suited for nearest neighbors search.

To conclude on Table \ref{offlineevaluation}, we emphasize that, while all scores might seem relatively low, they are actually encouraging considering the intrinsic complexity of the evaluation task (predicting a few listened tracks, among 50 000). Overall, the TT-SVD version of \textit{Deezer Semi-Pers.} reaches the best results. We note that the choice of @50 metrics is not restrictive. We reached consistent model rankings for @25 and @100 scores; the number of items to recommend is a selectable parameter in our public implementation.

\begin{figure}[H]
  \centering
  \includegraphics[width=1.0\linewidth]{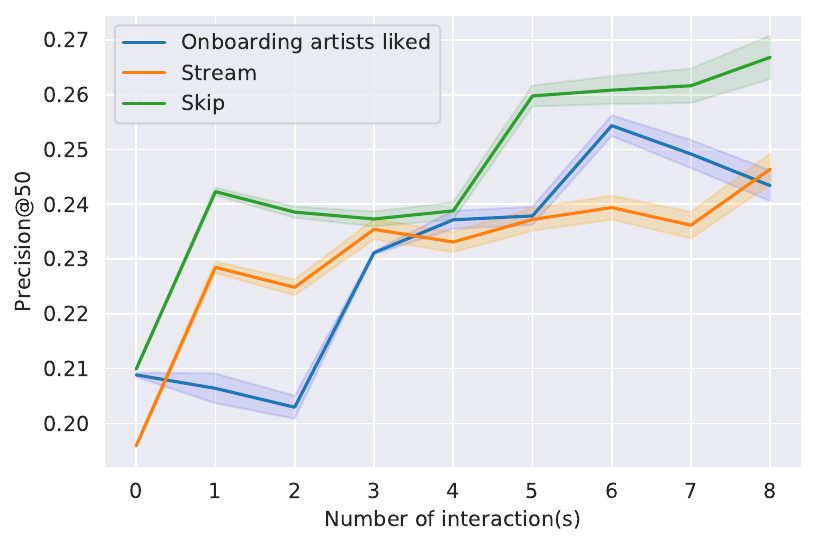}
  \caption[Precision@50 scores depending on the activity at registration day]{Precision@50 scores of Deezer Semi-Pers. depending on the number of artists liked during the onboarding, of streams, and of skips \textit{at registration day}.}
\label{interactions_analysis}
\end{figure}

To go further, Figure \ref{interactions_analysis} reports the mean precision@50 scores obtained by our semi-personalized system trained on TT-SVD embeddings, depending on available interactions at~registration~day. 
\begin{multicols}{2}
We observe that our model returns better results for users who streamed (positive signal) or skipped (negative signal) more tracks at registration day. Liking at least three artists during the onboarding session also improves recommendations, which tends to confirm the relevance of this strategy. On the contrary, our experiments show that users without \textit{any} interaction (5\% of them, for which only demographics are available) get a lower average precision@50 score of 17.74\%, 5.01 points below the global average.

\begin{figure}[H]
  \centering
  \includegraphics[width=1.0\linewidth]{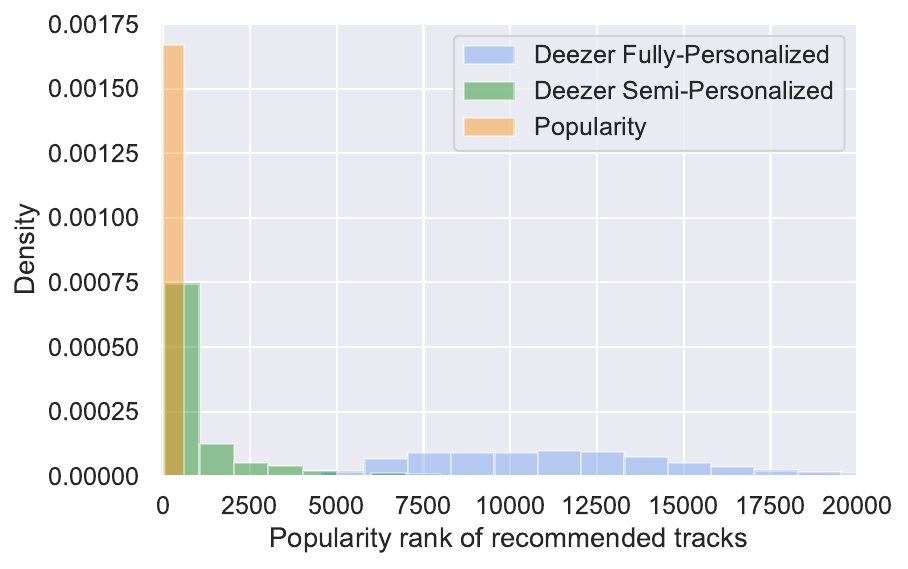}
  \caption[Popularity of the music tracks recommended to cold users]{Distribution of music tracks recommended to cold users, by popularity rank on Deezer.}
  \label{popularity_bias_svdembed}
\end{figure}

Lastly, Figure~\ref{popularity_bias_svdembed} reports the popularity distribution of recommended tracks from our system, trained on TT-SVD embeddings and for users from the top 5 countries on Deezer. We aim to assess whether better performance inevitably means recommending popular tracks, which is referred to as the \textit{popularity bias} \cite{park2008long}. Our semi-personalized system stands out from the popularity baseline, and mainly recommends tracks among the 5 000 most popular in the dataset, out of 50 000. 
\columnbreak

\end{multicols}
\vspace{-0.5cm}
The fully-personalized variant of our system permits recommending even less \textit{mainstream} tracks, but, as we showed, this might come at the price of noisier recommendations.

\section[Application to Online Recommendation on Deezer]{Application to Online Recommendation on Deezer via Personalized Carousels}
\label{c12s5}

We now report our online evaluation on the Deezer app, leveraging \textit{carousels} from Chapter~\ref{chapter_11}. 

\subsection{Experimental Setting: Cold Start Carousel Personalization}
\label{c12s51}

In addition to these experiments on data extracted from Deezer, online tests were run to check whether our conclusions would hold on the actual Deezer app. On our homepage, we do not directly recommend music tracks, as in our offline experimental setting, but instead \textit{musical collections} such as albums and music playlists. As a consequence, our online tests will rather consist in \textit{recommending playlists} to cold users, using the \textit{carousels}~\cite{bendada2020carousel} fully described in Chapter~\ref{chapter_11}. 

 We recall that carousels are ranked and swipeable lists of playlists cards. Deezer displays 12 recommended playlists to each user through these carousels. They are updated on a daily basis on the app. The embedding vectors of these playlists are computed from averages of the TT-SVD embeddings of music tracks from each playlist and from internal heuristics. They are leveraged in carousels through a TS extension now in production (technical details are voluntarily omitted). All playlists were created by professional curators from Deezer, with the purpose of complying with a specific music genre, cultural area, or mood. 

\subsection{Online Evaluation and Interpretation of Recommendations}
\label{c12s52}

\begin{figure}[H]
\includegraphics[width=1.0\linewidth]{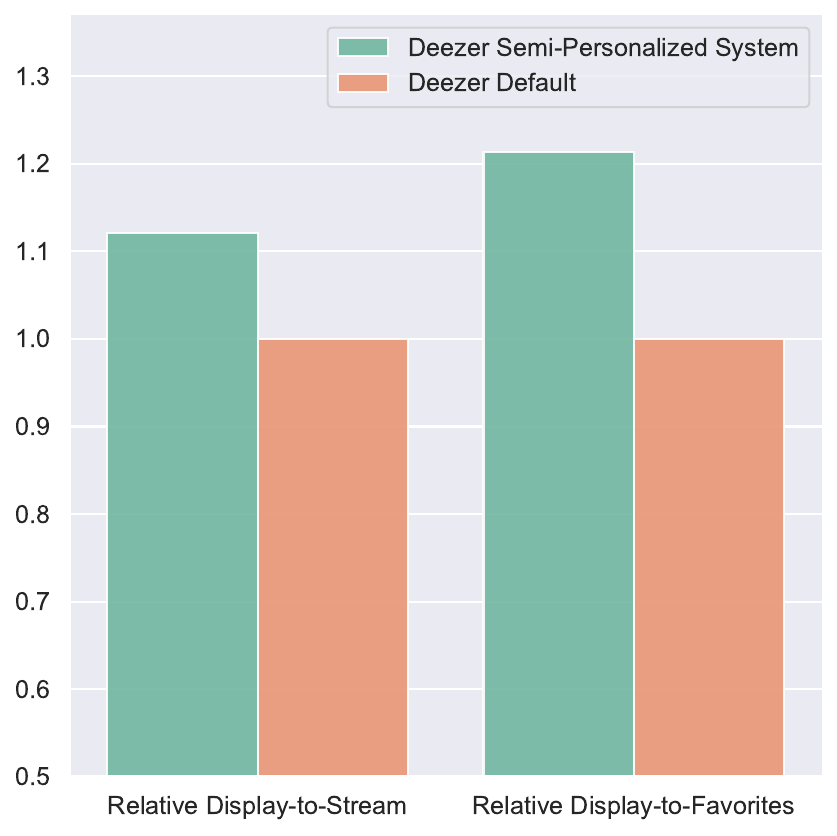}
\caption[Online A/B test on Deezer's carousels]{Online A/B test: relative \textit{display-to-stream} and  \textit{display-to-favorites} rates w.r.t. internal baseline. Differences are statistically significant at the 1\% level (p-value $<$ 0.01).}
\label{abtest}
\end{figure}

A large-scale A/B test has been run for a month on Deezer in 2020, on new cold users registering during this period. Due to industrial constraints, testing all model variants in production was impossible. Also, for confidentiality reasons, we do not report the exact number of users involved~in~each~cohort. 
\begin{multicols}{2}
Results are expressed in \textit{relative} terms w.r.t. the performance of \textit{Deezer Default}, a previous production system estimating cold user embeddings by countries and from internal heuristics. We observe in Figure \ref{abtest} that our new (TT-SVD-based) semi-personalized system leads to significant improvements of the relative \textit{display-to-stream} and \textit{display-to-favorite} rates, i.e., it permits selecting playlists on which cold users are more likely to click on and then to \textit{stream} the underlying content or \textit{add it to their list of favorite content}. Such results validate the relevance of our proposed system, and emphasize its practical impact on industrial-scale applications in a global music streaming app such as Deezer. In 2021, the system is still used in production at Deezer.
\columnbreak

\end{multicols}

Our semi-personalized system also enables us to provide \textit{interpretable} recommendations on Deezer. 
Indeed, carousel personalization relies on user embedding vectors that, for cold users, are linked to centroids of warm user segments. Therefore, recommended playlists can be described via the characteristics of these segments, such as the most common country or age class of warm users from each segment, or the most common music genres among their favorite artists. 

Table \ref{recoexamples} reports a few illustrative examples of descriptions of user segments, along with music playlists that were actually recommended to cold users from these segments on the Deezer app. Providing interpretable recommendations is often desirable for industrial applications, both for data scientists dealing with opaque model predictions, and for users as a way to improve their satisfaction and trust in the system \cite{afchar2020making}.

\begin{table}[t]
\centering
\caption[Examples of user segments from Deezer's production system]{Examples of user segments from Deezer's production system, described by most common country, age class and music genres among favorite tracks, together with some playlists recommended on the service.}
\label{recoexamples}
\resizebox{1.0\textwidth}{!}{
\begin{tabular}{c|c}
\toprule
\textbf{Description of the segment} & \textbf{Playlists recommended on Deezer} \\
\midrule
\midrule
\makecell{France \\ 18-24 y.o.\\ Rap, Hip-Hop}& \raisebox{-.5\height}{\includegraphics[width=0.75\textwidth]{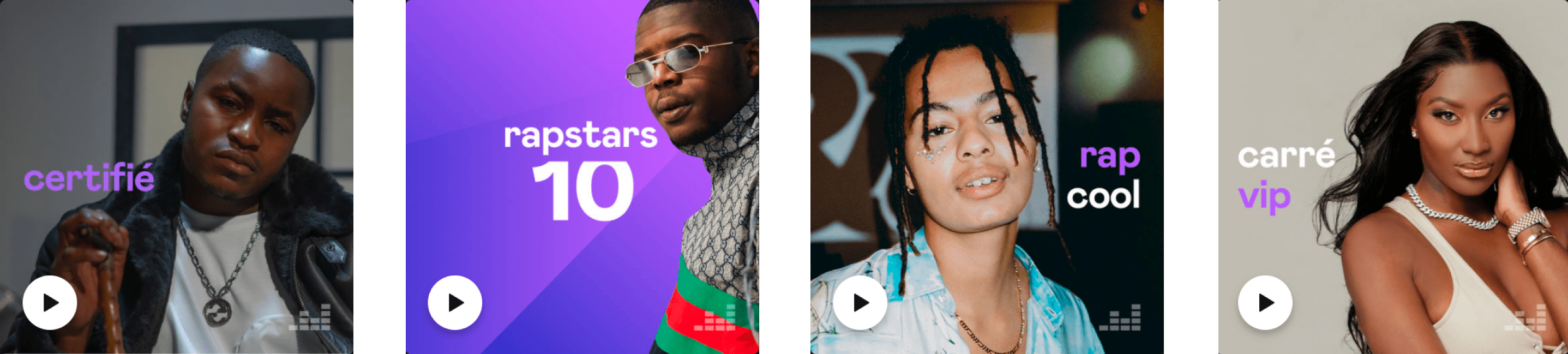}} \\
\midrule
\makecell{Brazil \\ 25-34 y.o. \\ Sertanejo}& \raisebox{-.5\height}{\includegraphics[width=0.75\textwidth]{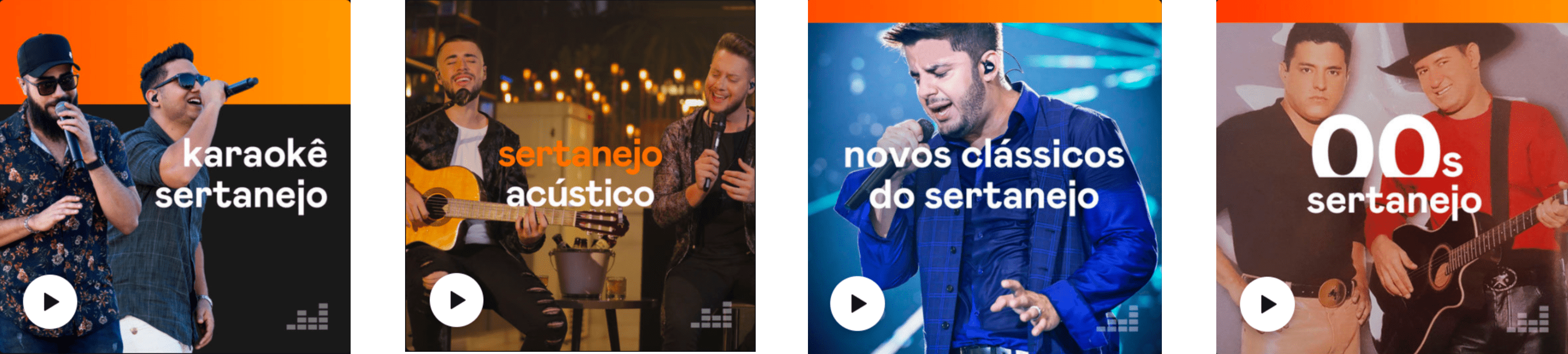}} \\
\midrule
\makecell{Germany \\ 35-49 y.o. \\ Schlager, Pop}& \raisebox{-.5\height}{\includegraphics[width=0.75\textwidth]{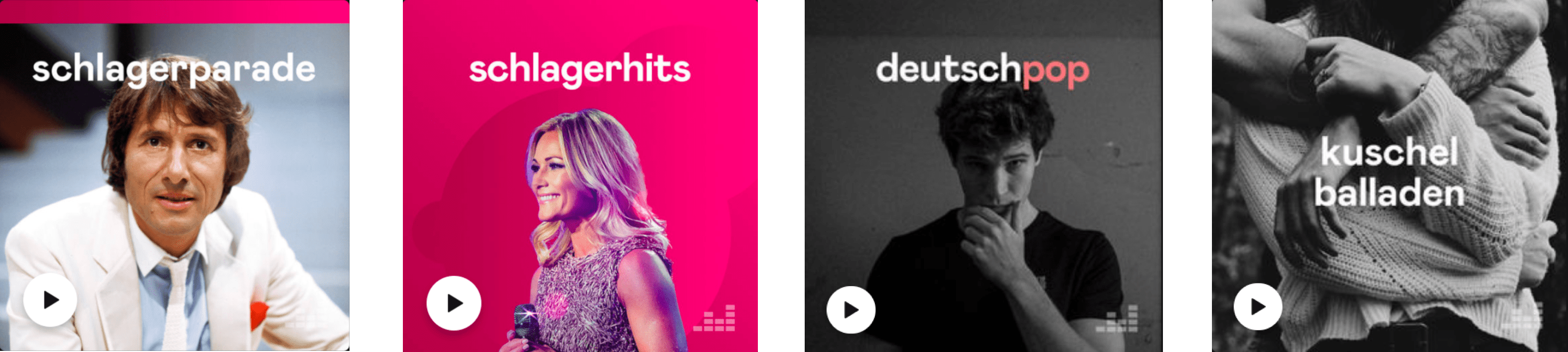}} \\
\bottomrule
\end{tabular}
}
\end{table}

\section{Conclusion}
\label{c12s6}

In this chapter, we presented the semi-personalized system recently deployed in production at Deezer to address the challenging user cold start problem. We demonstrated the tangible impact of this system, through both offline and online experiments on music recommendation tasks. Moreover, although the main focus of our work~\cite{briand2021semi} was on practical impact and not on chasing the state of the art, we also showed that our approach is competitive w.r.t. powerful user cold start models from the recent scientific literature.

Along with the paper associated with this work~\cite{briand2021semi}, we publicly released our code, as well as the dataset used in our offline experiments, providing information on 100~000 anonymized users and their interactions with the Deezer catalog. We hope that this open-source release of industrial resources will enable future research on user cold start recommendation.

For instance, in this paper, we assumed that, while the number of users increases over time, the musical catalog remains fixed, which is a limit, currently addressed at Deezer via internal heuristics extending our system. Future studies on a more effective incorporation of new releases in our semi-personalized system would decidedly improve its empirical effectiveness. In this direction, leveraging the GAE and VGAE models from Chapter~\ref{chapter_8}, combined with the FastGAE method from Chapter~\ref{chapter_4} for scalability, seems to be a promising approach, especially considering the recent successes of graph-based recommender systems on large-scale services such as Alibaba~\cite{wang2018billion} or Pinterest~\cite{ying2018graph}, including in cold start settings. 

Besides, future studies could also consider different embedding sizes for users and items \cite{he2017neural}, or the inclusion of temporal and contextual information in our neural network. Lastly, as we rely on user segments, we also believe that future work on a more effective user clustering, e.g., by leveraging graph-based methods, could strengthen our system and permit providing even more refined recommendations to new users. 

\fancyhead[LE]{\textbf{\nouppercase{\leftmark}} } 
\fancyhead[LO]{\textbf{\nouppercase{\leftmark}  }} 

\bookmarksetup{startatroot}
\chapter*{Conclusion}\label{sec_conclusion} \addcontentsline{toc}{chapter}{Conclusion}
\markboth{Conclusion}{}

Graph autoencoders (GAEs) and variational graph autoencoders (VGAEs) emerged as two powerful families of node embedding methods. Nonetheless, at the beginning of this PhD project, i.e., in 2018, existing variants of GAEs and VGAEs were still suffering from several fundamental limitations. As a consequence, leveraging these models for industrial-level applications, e.g., for music recommendation at Deezer, was still a challenging task.  The primary objective of this thesis was to address these limitations, with the general aim of improving GAEs and VGAEs, and of facilitating their use for real-world problems involving node embedding representations.

Specifically, we firstly aimed to improve their scalability. While standard GAE and VGAE models were limited to medium-size graphs with a few thousand nodes and edges, due to their quadratic computational complexity, we proposed two effective methods to overcome this issue. They are based on graph degeneracy and stochastic subgraph decoding, respectively. We provided the first application of these models to graphs with up to millions of nodes and edges. Besides, while standard variants of GAEs and VGAEs were limited to undirected graphs, we introduced Gravity-Inspired GAEs and VGAEs, an effective extension to process directed graphs that are ubiquitous in real-world problems. Our models achieved competitive empirical results w.r.t. popular alternatives on several directed link prediction tasks.

Throughout this thesis, we also considered extensions of GAEs and VGAEs for dynamic graphs and graphs with edges of different natures. Furthermore, by introducing Linear GAEs and VGAEs, we emphasized that multi-layer GCN-based GAEs and VGAEs are often unnecessarily complex, and we subsequently proposed to simplify them. This is an important aspect in the context of an industrial thesis, as simpler models are often preferred in production environments. Simultaneously, we provided some recommendations to improve the evaluation of complex GAE and VGAE models. Lastly, we introduced Modularity-Aware GAEs and VGAEs to improve community detection using these models, while jointly preserving their initially good performances on~link~prediction.

The second objective of this thesis was to evaluate our proposed approaches on industrial graphs extracted from the Deezer service. We put the emphasis on graph-based music recommendation problems, often formulated as link prediction or community detection tasks. In particular, we explained how, to this day, Deezer leverages similarity graphs constructed from usage data for music recommendation. We highlighted how this music streaming service can benefit from the GAE and VGAE models introduced in this thesis, e.g., to improve the detection of artists/albums communities to recommend to users. We also proposed a graph-based approach to rank similar artists on Deezer, leveraging our Gravity-Inspired GAEs and VGAEs. Using this approach, we successfully addressed the challenging cold start similar artists ranking problem, with an experimental evaluation on data extracted from Deezer’s production system. In another research study, we leveraged music genres graph ontologies to effectively model the music genre perception across language-bound cultures. To finish, in the last two chapters of this thesis, we presented two additional projects, less related to GAEs and VGAEs but providing a larger
overview of music recommendation at Deezer. We introduced, described, and analyzed some production-facing algorithms developed during the period covered by this PhD project. To this day, they are still running on the Deezer service and recommend music to millions of active users.

Overall, the majority of our experiments led to promising results. We believe and hope that the research conducted over the last three years will benefit some future projects and, in particular, will ease the application of graph autoencoders to large-scale real-world problems involving the use of node embedding representations. To encourage the usage of our methods, we publicly released our code along with each article published in the context of this thesis. Along with five of these articles, we also publicly released new datasets, that were either directly extracted from Deezer’s private resources, or scraped and processed from the internet.

Moreover, at the end of each chapter of this thesis, we aimed to describe the potential limitations of our methods. They open the way for future research on the corresponding topics. For instance, we acknowledged that the FastGAE method could underperform on very sparse graphs, and on directed graphs if the method relies on (undirected) core-based or degree-based sampling probabilities. We explained how our recommender systems from Chapters~\ref{chapter_11}~and~\ref{chapter_12} could be improved by considering the inclusion of graph-based methods in the learning process, and how the ones from Chapters~\ref{chapter_8}~and~\ref{chapter_9} still require online A/B tests and deployments, undone at the time of writing. Another aspect that we consistently noticed in our experiments regards the absence of clear empirical differences between GAE and VGAE models. While VGAE models seem to slightly outperform their deterministic counterparts in the majority of our community detection experiments, scores of GAE models are nonetheless fairly close. As we conclude this PhD thesis, the question of when one should favor VGAEs over GAEs remains quite open and deserves future research. Lastly, we recall that the GAE and VGAE frameworks are very general. Therefore, in the upcoming years, one could most likely improve our empirical results, e.g., by replacing our encoders (mostly, linear or multi-layer GCN models) with novel GNN architectures from the scientific literature, or by investigating better-suited prior latent distributions and generative models in the case of VGAEs.

The above considerations admittedly constitute relevant research directions for future work. Nonetheless, we believe that the most interesting improvements should come, not necessarily from more complex encoders, but rather from better data and better evaluations. Experiments from this thesis tend to suggest that a better initial representation of nodes could improve the resulting GAE/VGAE-based node embeddings at least as much as a more powerful GNN encoder. As an illustration, the performances of our GAE and VGAE models from Chapters~\ref{chapter_8}~and~\ref{chapter_9} strongly depend on how artists or albums are connected in the training graphs under consideration. Therefore, we believe that music streaming services, and other media aiming to leverage such graph-based methods, should redouble their efforts to define better ``similarity'' scores. As argued throughout this thesis, enriching these training graphs, e.g., with culture-specific information or with contextual information, could also decidedly improve our recommender systems. 
An even more ambitious direction for future research would consist in developing models that directly \textit{learn} such initial graph connections from the massive amounts of usage data and item descriptions collected on online services, i.e., without any arbitrary pre-processing step or any business-driven assumption on what ``similarity'' should mean. Concurrently, several of our studies could also benefit from more extensive evaluations. For instance, in the aforementioned Chapter~\ref{chapter_8}, we did not assess the impact of the inclusion of cold nodes in embedding spaces on the \textit{``Fans Also Like''} lists of \textit{warm} artists. In a future study, one could aim to assess whether these cold artists emerge on the recommended lists of warm artists, and therefore to which extent GAE and VGAE models could permit recommending more diverse musical content on music streaming services.  

At the beginning of this PhD project three years ago, the graph representation learning field was already growing at a fast pace.
As we are now concluding this thesis, we observe a gain of interest for graph representation learning methods in the music streaming industry and, overall, in the research communities working on the specific domains of applications covered at Deezer, i.e., recommender systems and music information retrieval. For instance, at the recent 15\up{th} ACM Conference on Recommender Systems (RecSys 2021) where our work received a ``best student paper'' honorable mention~\cite{salha2021cold}, the opening session was marked by a keynote presentation from Max Welling, one of the co-authors of the GCN, GAE and VGAE models~\cite{kipf2016-2,kipf2016-1}, on graph neural networks for recommendation. A month later, at the 22\up{nd} International Society for Music Information Retrieval Conference (ISMIR 2021) which occurred during the writing of this thesis, researchers from the music streaming service Pandora presented a graph-based approach to compute artist similarities, using a GNN model~\cite{korzeniowski2021artist}. Their work also received a ``best paper'' honorable mention. At the same conference, researchers from the music streaming service Spotify presented another multi-task graph-based model, with application to music recommendation~\cite{saravanou2021multi}. 
These recent studies complement this thesis, which mainly focused on GAE/VGAE-based node embedding representations and their applications to link prediction and community detection tasks. In agreement with our research, they confirm the impact, the relevance, and the potential of graph representation learning methods to tackle numerous real-world problems emerging on music streaming services, and, more broadly, on large-scale media providing and recommending content to millions of users.

\setstretch{1}

\clearpage
\phantomsection
\addcontentsline{toc}{chapter}{Bibliography}

\begin{scriptsize}

\bibliography{chapters/references}
\bibliographystyle{acm}
\end{scriptsize}

\fancyhead[LE]{\textbf{Appendix \thechapter.\ \nouppercase{\leftmark}} } 
\fancyhead[LO]{\textbf{\thesection.\ \nouppercase{\rightmark}} } 



\end{onehalfspace}
\end{document}